# Artificial Intelligence Index Report 2022

Stanford University
Human-Centered
Artificial Intelligence

# INTRODUCTION TO THE AI INDEX REPORT 2022

Welcome to the fifth edition of the AI Index Report! The latest edition includes data from a broad set of academic, private, and nonprofit organizations as well as more self-collected data and original analysis than any previous editions, including an expanded technical performance chapter, a new survey of robotics researchers around the world, data on global AI legislation records in 25 countries, and a new chapter with an in-depth analysis of technical AI ethics metrics.

The AI Index Report tracks, collates, distills, and visualizes data related to artificial intelligence. Its mission is to provide unbiased, rigorously vetted, and globally sourced data for policymakers, researchers, executives, journalists, and the general public to develop a more thorough and nuanced understanding of the complex field of AI. The report aims to be the world's most credible and authoritative source for data and insights about AI.

## FROM THE CO-DIRECTORS

This year's report shows that AI systems are starting to be deployed widely into the economy, but at the same time they are being deployed, the ethical issues associated with AI are becoming magnified. Some of this is natural—after all, we tend to care more about the ethical aspects of a given technology when it is being rolled out into the world. But some of it is bound up in the peculiar traits of contemporary AI—larger and more complex and capable AI systems can generally do better on a broad range of tasks while also displaying a greater potential for ethical concerns.

This is bound up with the broad globalization and industrialization of AI—a larger range of countries are developing, deploying, and regulating AI systems than ever before, and the combined outcome of these activities is the creation of a broader set of AI systems available for people to use, and reductions in their prices. Some parts of AI are not very globalized, though, and our ethics analysis reveals that many AI ethics publications tend to concentrate on English-language systems and datasets, despite AI being deployed globally.

If anything, we expect the above trends to continue: 103% more money was invested in the private investment of AI and AI-related startups in 2021 than in 2020 ($96.5 billion versus $46 billion).

**Jack Clark and Ray Perrault**

# TOP TAKEAWAYS

## Private investment in AI soared while investment concentration intensified:

- **The private investment in AI in 2021 totaled around $93.5 billion—more than double the total private investment in 2020,** while the number of newly funded AI companies continues to drop, from 1051 companies in 2019 and 762 companies in 2020 to 746 companies in 2021. **In 2020, there were 4 funding rounds worth $500 million or more; in 2021, there were 15.**

## U.S. and China dominated cross-country collaborations on AI:

- Despite rising geopolitical tensions, the United States and China had the greatest number of cross-country collaborations in AI publications from 2010 to 2021, **increasing five times since 2010**. The collaboration between the two countries **produced 2.7 times more publications** than between the United Kingdom and China—the second highest on the list.

## Language models are more capable than ever, but also more biased:

- Large language models are setting new records on technical benchmarks, but new data shows that larger models are also more capable of reflecting biases from their training data. **A 280 billion parameter model developed in 2021 shows a 29% increase in elicited toxicity over a 117 million parameter model considered the state of the art as of 2018.** The systems are growing significantly more capable over time, though as they increase in capabilities, so does the potential severity of their biases.

## The rise of AI ethics everywhere:

- Research on fairness and transparency in AI has exploded since 2014, **with a fivefold increase in related publications** at ethics-related conferences. Algorithmic fairness and bias has shifted from being primarily an academic pursuit to becoming firmly embedded as a mainstream research topic with wide-ranging implications. **Researchers with industry affiliations contributed 71% more publications year over year** at ethics-focused conferences in recent years.

## AI becomes more affordable *and* higher performing:

- **Since 2018, the cost to train an image classification system has decreased by 63.6%, while training times have improved by 94.4%.** The trend of lower training cost but faster training time appears across other MLPerf task categories such as recommendation, object detection and language processing, and favors the more widespread commercial adoption of AI technologies.

## Data, data, data:

- Top results across technical benchmarks have increasingly relied on the use of extra training data to set new state-of-the-art results. **As of 2021, 9 state-of-the-art AI systems out of the 10 benchmarks in this report are trained with extra data.** This trend implicitly favors private sector actors with access to vast datasets.

## More global legislation on AI than ever:

- An AI Index analysis of legislative records on AI in 25 countries shows that the number of bills containing "artificial intelligence" that were **passed into law grew from just 1 in 2016 to 18 in 2021.** Spain, the United Kingdom, and the United States passed the highest number of AI-related bills in 2021 with each adopting three.

## Robotic arms are becoming cheaper:

- An AI Index survey shows that **the median price of robotic arms has decreased by 46.2% in the past five years—from $42,000 per arm in 2017 to $22,600 in 2021.** Robotics research has become more accessible and affordable.

# Steering Committee

### Co-Directors

Jack Clark
Anthropic, OECD

Raymond Perrault
SRI International

### Members

Erik Brynjolfsson
Stanford University

John Etchemendy
Stanford University

Terah Lyons

James Manyika
Google, University of Oxford

Juan Carlos Niebles
Stanford University, Salesforce

Michael Sellitto
Stanford University

Yoav Shoham
(Founding Director)
Stanford University, AI21 Labs

# Staff and Researchers

### Research Manager and Editor in Chief

Daniel Zhang
Stanford University

### Research Associate

Nestor Maslej
Stanford University

### Affiliated Researchers

Andre Barbe
The World Bank

Latisha Harry
Independent Consultant

Helen Ngo
Cohere

Ellie Sakhaee
Microsoft

### Graduate Researcher

Benjamin Bronkema-Bekker
Stanford University

## How to Cite This Report

Daniel Zhang, Nestor Maslej, Erik Brynjolfsson, John Etchemendy, Terah Lyons, James Manyika, Helen Ngo, Juan Carlos Niebles, Michael Sellitto, Ellie Sakhaee, Yoav Shoham, Jack Clark, and Raymond Perrault, “The AI Index 2022 Annual Report,” AI Index Steering Committee, Stanford Institute for Human-Centered AI, Stanford University, March 2022.

## Public Data and Tools

The AI Index 2022 Report is supplemented by raw data and an interactive tool. We invite each reader to use the data and the tool in a way most relevant to their work and interests.

- Raw data and charts: The public data and high-resolution images of all the charts in the report are available on Google Drive.
- Global AI Vibrancy Tool: We redesigned the Global AI Vibrancy Tool this year with a better visualization to compare up to 29 countries across 23 indicators.

## AI Index and Stanford HAI

The AI Index is an independent initiative at the Stanford Institute for Human-Centered Artificial Intelligence (HAI).

Artificial Intelligence Index

Stanford University
Human-Centered
Artificial Intelligence

The AI Index was conceived within the One Hundred Year Study on AI (AI100).

We welcome feedback and new ideas for next year.
Contact us at AI-Index-Report@stanford.edu.

## Supporting Partners

Google

OpenAI

Open Philanthropy

## Analytics and Research Partners

Bloomberg Government

CSET CENTER for SECURITY and EMERGING TECHNOLOGY

CRA Computing Research Association

Emsi | burning glass

LinkedIn

McKinsey & Company

NETBASE QUID We make the world make sense

WiML Women in Machine Learning

# Contributors

We want to acknowledge the following individuals by chapter and section for their contributions of data, analysis, advice, and expert commentary included in the AI Index 2022 Report:

### Research and Development

Sara Abdulla, Catherine Aiken, Jack Clark, James Dunham, Nezihe Merve Gürel, Nestor Maslej, Ray Perrault, Sarah Tan, Daniel Zhang

### Technical Performance

Jack Clark, David Kanter, Nestor Maslej, Deepak Narayanan, Juan Carlos Niebles, Konstantin Savenkov, Yoav Shoham, Daniel Zhang

### Technical AI Ethics

Jack Clark, Nestor Maslej, Helen Ngo, Ray Perrault, Ellie Sakhaee, Daniel Zhang

### The Economy and Education

Betsy Bizot, Erik Brynjolfsson, Jack Clark, John Etchemendy, Murat Erer, Akash Kaura, Julie Kim, Nestor Maslej, James Manyika, Brenden McKinney, Julia Nitschke, Ray Perrault, Brittany Presten, Tejas Sirohi, Bledi Taska, Rucha Vankudre, Daniel Zhang

### AI Policy and Governance

Amanda Allen, Benjamin Bronkema-Bekker, Jack Clark, Latisha Harry, Taehwa Hong, Cameron Leuthy, Terah Lyons, Nestor Maslej, Ray Perrault, Michael Sellitto, Teruki Takiguchi, Daniel Zhang

### Conference Attendance

Terri Auricchio (ICML), Christian Bessiere (IJCAI), Meghyn Bienvenu (KR), Andrea Brown (ICLR), Alexandra Chouldechova (FAccT), Nicole Finn (ICCV, CVPR), Enrico Gerding (AAMAS), Carol Hamilton (AAAI), Seth Lazar (FAccT), Max Qing Hu Meng (ICRA), Jonas Martin Peters (UAI), Libor Preucil (IROS), Marc'Aurelio Ranzato (NeurIPS), Priscilla Rasmussen (EMNLP, ACL), Hankz Hankui Zhuo (ICAPS)

### Global AI Vibrancy Tool

Andre Barbe, Latisha Harry, Daniel Zhang

### Robotics Survey

Pieter Abbeel, David Abbink, Farshid Alambeigi, Farshad Arvin, Nikolay Atanasov, Ruzena Bajcsy, Philip Beesley, Tapomayukh Bhattacharjee, Jeannette Bohg, David J. Cappelleri, Qifeng Chen, I-Ming Chen, Jack Cheng, Cynthia Chestek, Kyujin Cho, Dimitris Chrysostomou, Steve Collins, David Correa, Brandon DeHart, Katie Driggs-Campbell, Nima Fazeli, Animesh Garg, Maged Ghoneima, Tobias Haschke, Kris Hauser, David Held, Yue Hu, Josie Hughes, Soo Jeon, Dimitrios Kanoulas, Jonathan Kelly, Oliver Kroemer, Changliu Liu, Ole Madsen, Anirudha Majumdar, Genaro J. Martinez, Saburo Matunaga, Satoshi Miura, Norrima Mokhtar, Elena De Momi, Chrystopher Nehaniv, Christopher Nielsen, Ryuma Niiyama, Allison Okamura, Necmiye Ozay, Jamie Paik, Frank Park, Karthik Ramani, Carolyn Ren, Jan Rosell, Jee-Hwan Ryu, Tim Salcudean, Oliver Schneider, Angela Schoellig, Reid Simmons, Alvaro Soto, Peter Stone, Michael Tolley, Tsu-Chin Tsao, Michiel van de Panne, Andy Weightman, Alexander Wong, Helge Wurdemann, Rong Xiong, Chao Xu, Geng Yang, Junzhi Yu, Wenzhen Yuan, Fu Zhang, Yuke Zhu

We thank the following organizations and individuals who provided data for inclusion in the AI Index 2022 Report:

## Organizations

**Bloomberg Government**
Amanda Allen, Cameron Leuthy

**Center for Security and Emerging Technology, Georgetown University**
Sara Abdulla, Catherine Aiken, James Dunham

**Computing Research Association**
Betsy Bizot

**Emsi Burning Glass**
Julia Nitschke, Bledi Taska, Rucha Vankudre

**Intento**
Grigory Sapunov, Konstantin Savenkov

**LinkedIn**
Murat Erer, Akash Kaura

**McKinsey Global Institute**
Brenden McKinney, Brittany Presten

**MLCommons**
David Kanter

**NetBase Quid**
Julie Kim, Tejas Sirohi

**Women in Machine Learning**
Nezihe Merve Gürel, Sarah Tan

We also would like to thank Jeanina Casusi, Nancy King, Shana Lynch, Jonathan Mindes, Stacy Peña, Michi Turner, and Justin Sherman for their help in preparing this report, and Joe Hinman, Travis Taylor, and the team at Digital Avenues for their efforts in designing and developing the AI Index and HAI websites.

# Table of Contents

**ACCESS THE PUBLIC DATA**

# REPORT HIGHLIGHTS

## CHAPTER 1: RESEARCH AND DEVELOPMENT

- Despite rising geopolitical tensions, the United States and China had the greatest number of cross-country collaborations in AI publications from 2010 to 2021, **increasing five times since 2010.** The collaboration between the two countries **produced 2.7 times more publications** than between the United Kingdom and China—the second highest on the list.
- In 2021, China continued to lead the world in the number of AI journal, conference, and repository publications—**63.2% higher than the United States** with all three publication types combined. In the meantime, the United States **held a dominant lead among major AI powers** in the number of AI conference and repository citations.
- From 2010 to 2021, **the collaboration between educational and nonprofit organizations produced the highest number of AI publications,** followed by the collaboration between private companies and educational institutions and between educational and government institutions.
- **The number of AI patents filed in 2021 is more than 30 times higher than in 2015,** showing a compound annual growth rate of 76.9%.

## CHAPTER 2: TECHNICAL PERFORMANCE

- **Data, data, data:** Top results across technical benchmarks have increasingly relied on the use of extra training data to set new state-of-the-art results. **As of 2021, 9 state-of-the-art AI systems out of the 10 benchmarks in this report are trained with extra data.** This trend implicitly favors private sector actors with access to vast datasets.
- **Rising interest in particular computer vision subtasks:** In 2021, the research community saw a greater level of interest in more specific computer vision subtasks, such as medical image segmentation and masked-face identification. **For example, only 3 research papers tested systems against the Kvasir-SEG medical imaging benchmark prior to 2020. In 2021, 25 research papers did.** Such an increase suggests that AI research is moving toward research that can have more direct, real-world applications.
- **AI has not mastered complex language tasks, *yet:*** AI already exceeds human performance levels on basic reading comprehension benchmarks like SuperGLUE and SQuAD by 1%–5%. **Although AI systems are still unable to achieve human performance on more complex linguistic tasks such as abductive natural language inference (aNLI), the difference is narrowing. Humans performed 9 percentage points better on aNLI in 2019. As of 2021, that gap has shrunk to 1.**

- **Turn toward more general reinforcement learning:** For the last decade, AI systems have been able to master narrow reinforcement learning tasks in which they are asked to maximize performance in a specific skill, such as chess. **The top chess software engine now exceeds Magnus Carlsen's top ELO score by 24%. However, in the last two years AI systems have also improved by 129% on more general reinforcement learning tasks (Procgen) in which they must operate in novel environments.** This trend speaks to the future development of AI systems that can learn to think more broadly.
- **AI becomes more affordable *and* higher performing: Since 2018, the cost to train an image classification system has decreased by 63.6%, while training times have improved by 94.4%.** The trend of lower training cost but faster training time appears across other MLPerf task categories such as recommendation, object detection and language processing, and favors the more widespread commercial adoption of AI technologies.
- **Robotic arms are becoming cheaper:** An AI Index survey shows that **the median price of robotic arms has decreased by 46.2% in the past five years—from $42,000 per arm in 2017 to $22,600 in 2021.** Robotics research has become more accessible and affordable.

## CHAPTER 3: TECHNICAL AI ETHICS

- **Language models are more capable than ever, but also more biased:** Large language models are setting new records on technical benchmarks, but new data shows that larger models are also more capable of reflecting biases from their training data. **A 280 billion parameter model developed in 2021 shows a 29% increase in elicited toxicity over a 117 million parameter model considered the state of the art as of 2018.** The systems are growing significantly more capable over time, though as they increase in capabilities, so does the potential severity of their biases.
- **The rise of AI ethics everywhere:** Research on fairness and transparency in AI has exploded since 2014, **with a fivefold increase in related publications** at ethics-related conferences. Algorithmic fairness and bias has shifted from being primarily an academic pursuit to becoming firmly embedded as a mainstream research topic with wide-ranging implications. **Researchers with industry affiliations contributed 71% more publications year over year** at ethics-focused conferences in recent years.
- **Multimodal models learn multimodal biases:** Rapid progress has been made on training multimodal language-vision models which exhibit new levels of capability on joint language-vision tasks. These models have set new records on tasks like image classification and the creation of images from text descriptions, but they also reflect societal stereotypes and biases in their outputs—**experiments on CLIP showed that images of Black people were misclassified as nonhuman at over twice the rate of any other race.** While there has been significant work to develop metrics for measuring bias within both computer vision and natural language processing, this highlights the need for metrics that provide insight into biases in models with multiple modalities.

## CHAPTER 4: THE ECONOMY AND EDUCATION

- New Zealand, Hong Kong, Ireland, Luxembourg, and Sweden are the countries or regions with the highest growth in AI hiring from 2016 to 2021.
- In 2021, California, Texas, New York, and Virginia were states with the highest number of AI job postings in the United States, with **California having over 2.35 times the number of postings as Texas**, the second greatest. Washington, D.C., had the greatest rate of AI job postings compared to its overall number of job postings.
- **The private investment in AI in 2021 totaled around $93.5 billion—more than double the total private investment in 2020**, while the number of newly funded AI companies continues to drop, from 1051 companies in 2019 and 762 companies in 2020 to 746 companies in 2021. **In 2020, there were 4 funding rounds worth $500 million or more; in 2021, there were 15.**
- "Data management, processing, and cloud" received the greatest amount of private AI investment in 2021—**2.6 times the investment in 2020**, followed by "medical and healthcare" and "fintech."
- In 2021, the United States led the world in both total private investment in AI and the number of newly funded AI companies, **three and two times higher**, respectively, than China, the next country on the ranking.
- Efforts to address ethical concerns associated with using AI in industry remain limited, according to a McKinsey survey. **While 29% and 41% of respondents recognize "equity and fairness" and "explainability" as risks while adopting AI, only 19% and 27% are taking steps to mitigate those risks.**
- In 2020, **1 in every 5 CS students who graduated with PhD degrees specialized in artificial intelligence/machine learning,** the most popular specialty in the past decade. From 2010 to 2020, the majority of AI PhDs in the United States headed to industry while a small fraction took government jobs.

## CHAPTER 5: AI POLICY AND GOVERNANCE

- An AI Index analysis of legislative records on AI in 25 countries shows that the number of bills containing "artificial intelligence" that were **passed into law grew from just 1 in 2016 to 18 in 2021.** Spain, the United Kingdom, and the United States passed the highest number of AI-related bills in 2021, with each adopting three.
- The federal legislative record in the United States shows a sharp increase in the total number of proposed bills that relate to AI from 2015 to 2021, **while the number of bills passed remains low, with only 2% ultimately becoming law.**
- State legislators in the United States **passed 1 out of every 50 proposed bills** that contain AI provisions in 2021, while the number of such bills proposed **grew from 2 in 2012 to 131 in 2021.**
- In the United States, the current congressional session (the 117th) is on track to record the greatest number of AI-related mentions since 2001, **with 295 mentions by the end of 2021, half way through the session, compared to 506 in the previous (116th) session.**

Artificial Intelligence Index Report 2022

## CHAPTER 1:

# Research & Development

## CHAPTER 1:
# Chapter Preview

ACCESS THE PUBLIC DATA

# Overview

Research and development is an integral force that drives the rapid progress of artificial intelligence (AI). Every year, a wide range of academic, industry, government, and civil society experts and organizations contribute to AI R&D via a slew of papers, journal articles, and other AI-related publications, conferences on AI or on particular subtopics like image recognition or natural language processing, international collaboration across borders, and the development of open-source software libraries. These R&D efforts are diverse in focus and geographically dispersed.

Another key feature of AI R&D, making it somewhat distinct from other areas of STEM research, is its openness. Each year, thousands and thousands of AI publications are released in the open source, whether at conferences or on file-sharing websites. Researchers will openly share their findings at conferences; government agencies will fund AI research that ends up in the open source; and developers use open software libraries, freely available to the public, to produce state-of-the-art AI applications. This openness also contributes to the globally interdependent and interconnected nature of modern AI R&D.

This first chapter draws on multiple datasets to analyze key trends in the AI research and development space in 2021. It first looks at AI publications, including conference papers, journal articles, patents, and repositories. It then analyzes AI conference attendance. And finally, it examines AI open-source software libraries used in the R&D process.

## CHAPTER HIGHLIGHTS

- Despite rising geopolitical tensions, the United States and China had the greatest number of cross-country collaborations in AI publications from 2010 to 2021, **increasing five times since 2010.** The collaboration between the two countries **produced 2.7 times more publications** than between the United Kingdom and China—the second highest on the list.

- In 2021, China continued to lead the world in the number of AI journal, conference, and repository publications—**63.2% higher than the United States** with all three publication types combined. In the meantime, the United States **held a dominant lead among major AI powers** in the number of AI conference and repository citations.

- From 2010 to 2021, **the collaboration between educational and nonprofit organizations produced the highest number of AI publications**, followed by the collaboration between private companies and educational institutions and between educational and government institutions.

- **The number of AI patents filed in 2021 is more than 30 times higher than in 2015**, showing a compound annual growth rate of 76.9%.

This section draws on data from the Center for Security and Emerging Technology (CSET) at Georgetown University. CSET maintains a merged corpus of scholarly literature that includes Digital Science's Dimensions, Clarivate's Web of Science, Microsoft Academic Graph, China National Knowledge Infrastructure, arXiv, and Papers with Code. In that corpus, CSET applied a classifier to identify English-language publications related to the development or application of AI and ML since 2010.[1]

# 1.1 PUBLICATIONS[2]

## OVERVIEW

The figures below capture the total number of English-language AI publications globally from 2010 to 2021—by type, affiliation, cross-country collaboration, and cross-industry collaboration. The section also breaks down publication and citation data by region for AI journal articles, conference papers, repositories, and patents.

## Total Number of AI Publications

Figure 1.1.1 shows the number of AI publications in the world. From 2010 to 2021, the total number of AI publications doubled, growing from 162,444 in 2010 to 334,497 in 2021.

NUMBER of AI PUBLICATIONS in the WORLD, 2010–21
Source: Center for Security and Emerging Technology, 2021 | Chart: 2022 AI Index Report

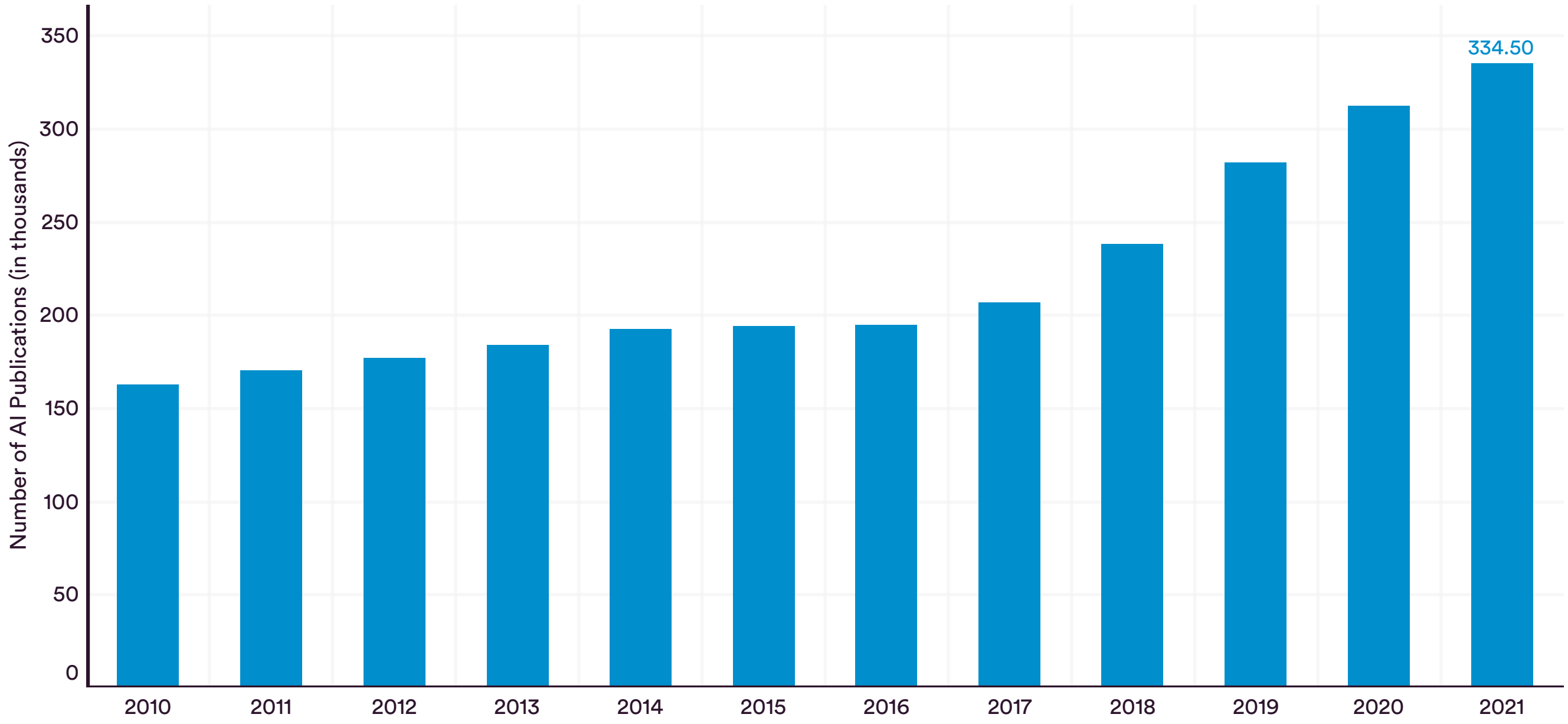

Figure 1.1.1

1 See the Appendix for more information on CSET's methodology. Due to the change in data provider and classification method, the publication trend/data may be different from past reports. For more on the challenge of defining AI and correctly capturing relevant bibliometric data, see the AI Index team's discussion in the paper "Measurement in AI Policy: Opportunities and Challenges."
2 The number of AI publications in 2021 in this section may be lower than the actual count due to the lag in the collection of publication metadata by aforementioned databases.

## By Type of Publication

Figure 1.1.2 shows the variation overtime in the types of AI publications released globally. In 2021, 51.5% of all AI documents published were journal articles, 21.5% were conference papers, and 17.0% were from repositories. Books, book chapters, theses, and unknown document types comprised the remaining 10.1% of publications. While journal and repository publications have grown 2.5 and 30 times, respectively, in the past 12 years, the number of conference papers has declined since 2018.

NUMBER of AI PUBLICATIONS by TYPE, 2010–21
Source: Center for Security and Emerging Technology, 2021 | Chart: 2022 AI Index Report

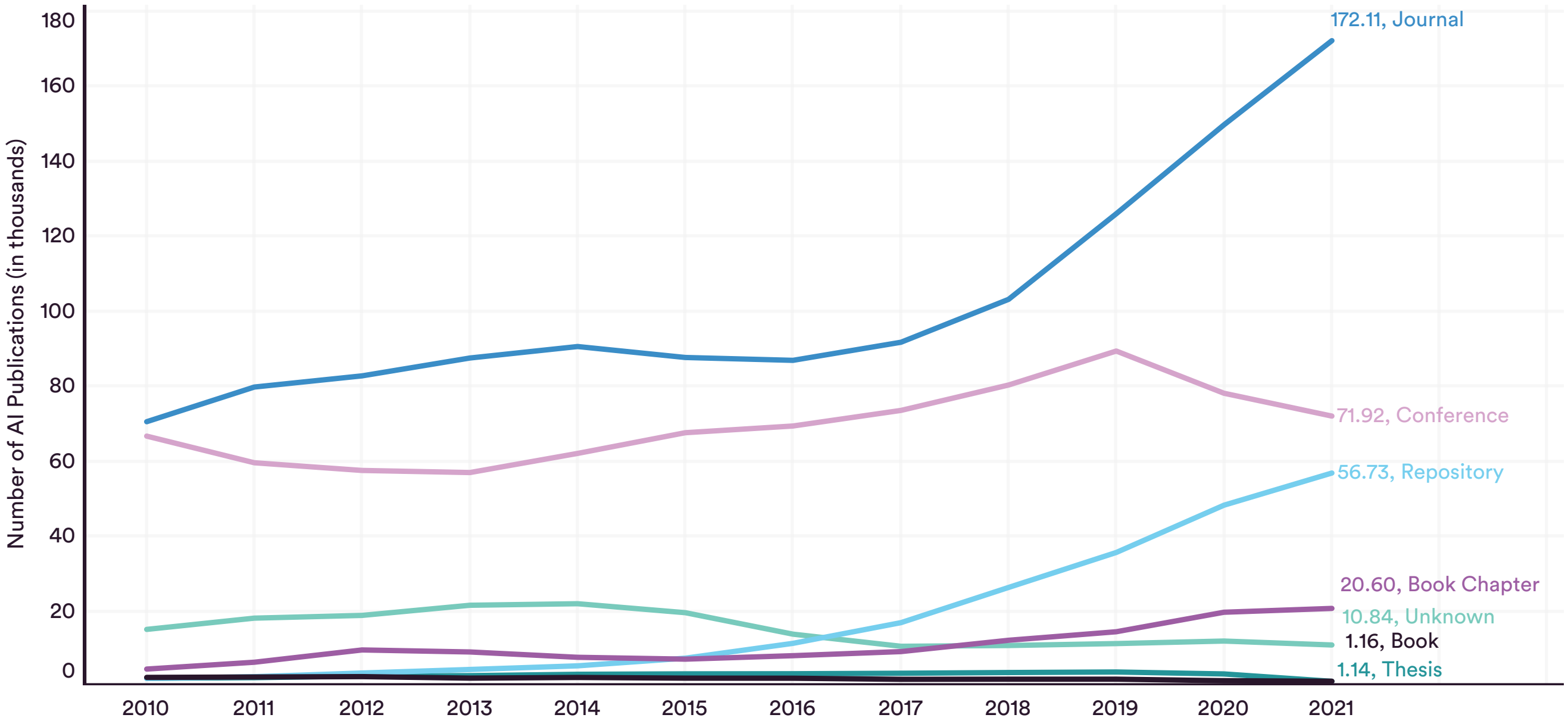

Figure 1.1.2

## By Field of Study

Figure 1.1.3 shows that publications in pattern recognition and machine learning have more than doubled since 2015. Other areas strongly influenced by deep learning, such as computer vision, data mining and natural language processing, have shown smaller increases.

NUMBER of AI PUBLICATIONS by FIELD of STUDY (EXCLUDING OTHER AI), 2010–21
Source: Center for Security and Emerging Technology, 2021 | Chart: 2022 AI Index Report

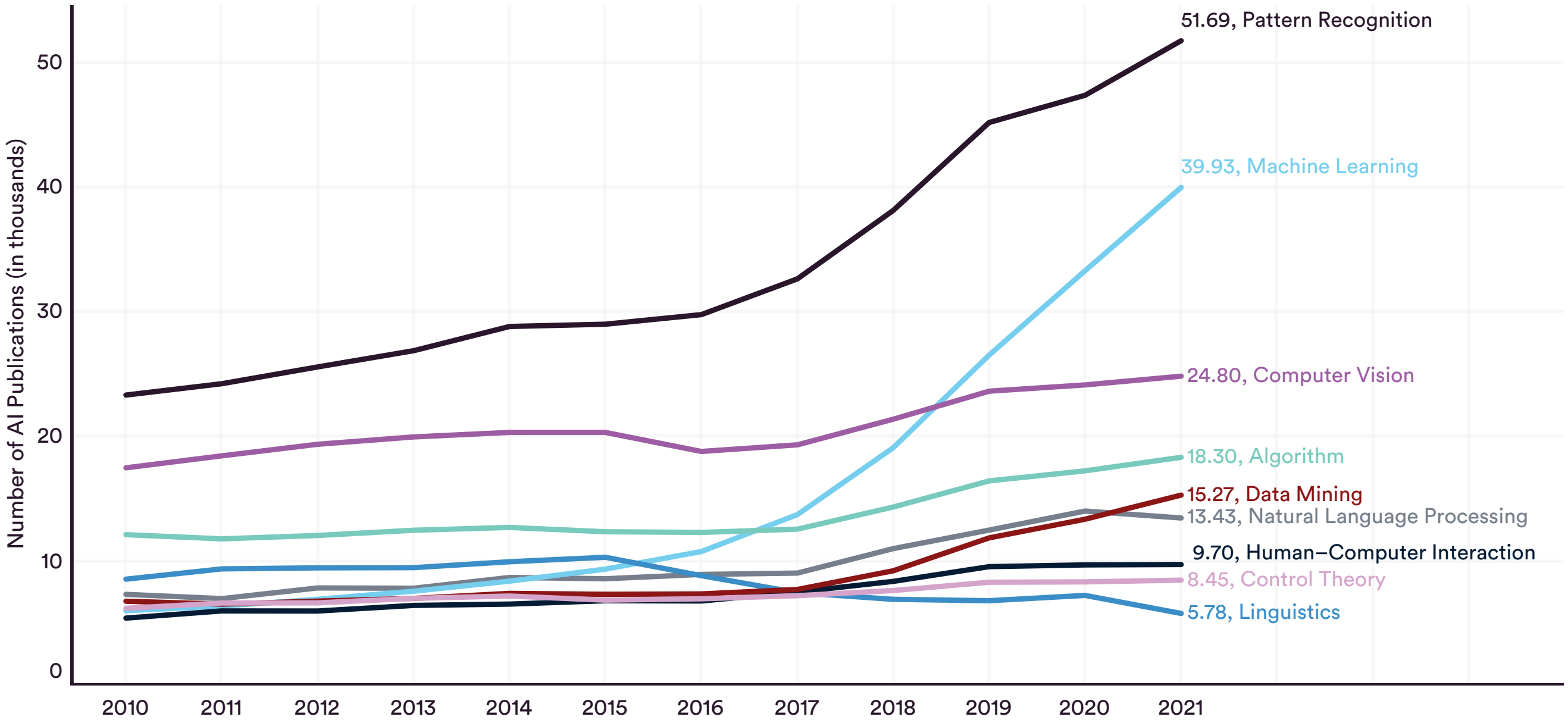

Figure 1.1.3

## By Sector

This section shows the number of AI publications affiliated with industry, education, government, and nonprofit in the world (Figure 1.1.4a), China (Figure 1.1.4b), the United States (Figure 1.1.4c), and the European Union plus the United Kingdom (Figure 1.1.4d).[3] The education sector dominates in each of the regions. The level of company participation is highest in the United States, then in the European Union. China is the only area in which the share of education has been rising.

### AI PUBLICATIONS (% of TOTAL) by SECTOR, 2010–21

Source: Center for Security and Emerging Technology, 2021 | Chart: 2022 AI Index Report

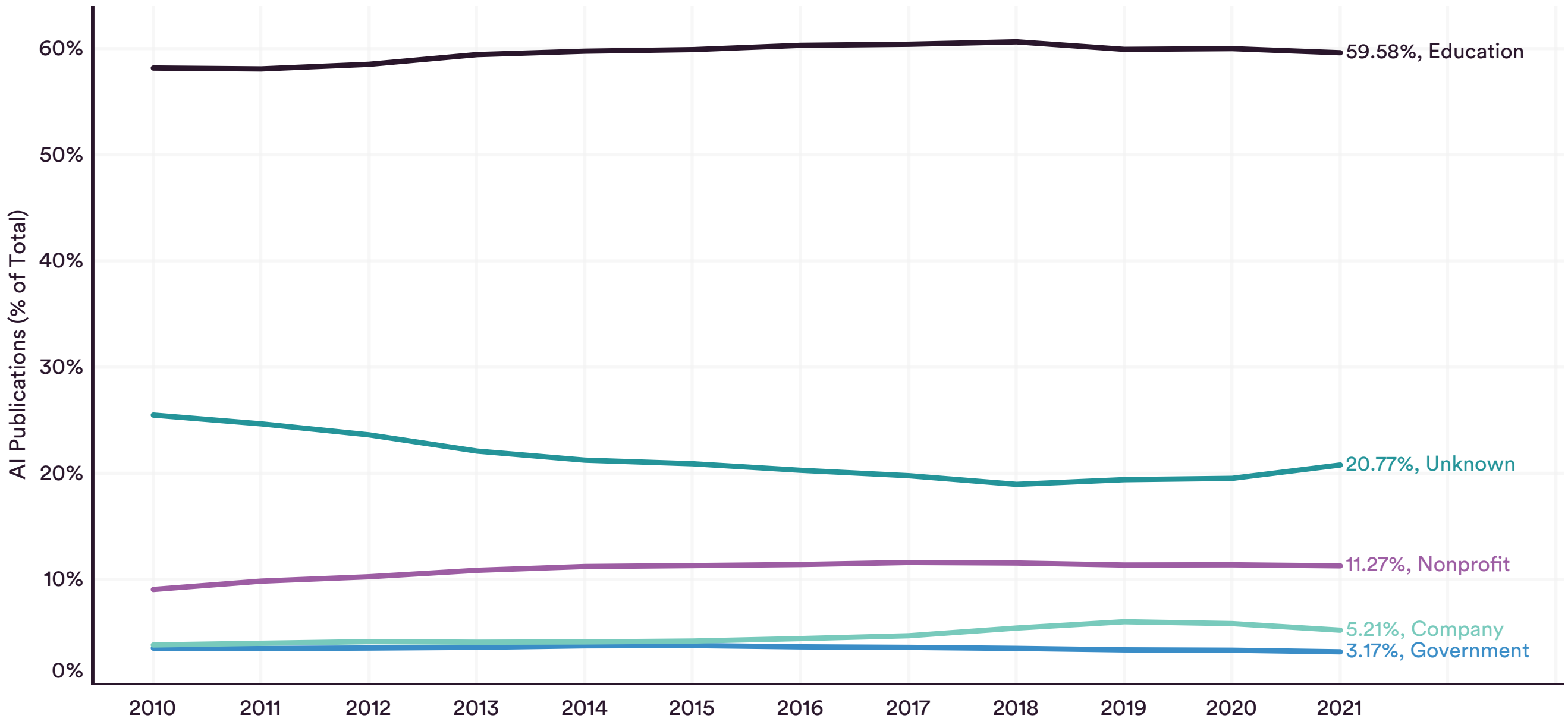

Figure 1.1.4a

### AI PUBLICATIONS in UNITED STATES (% of TOTAL) by SECTOR, 2010–21

Source: Center for Security and Emerging Technology, 2021 | Chart: 2022 AI Index Report

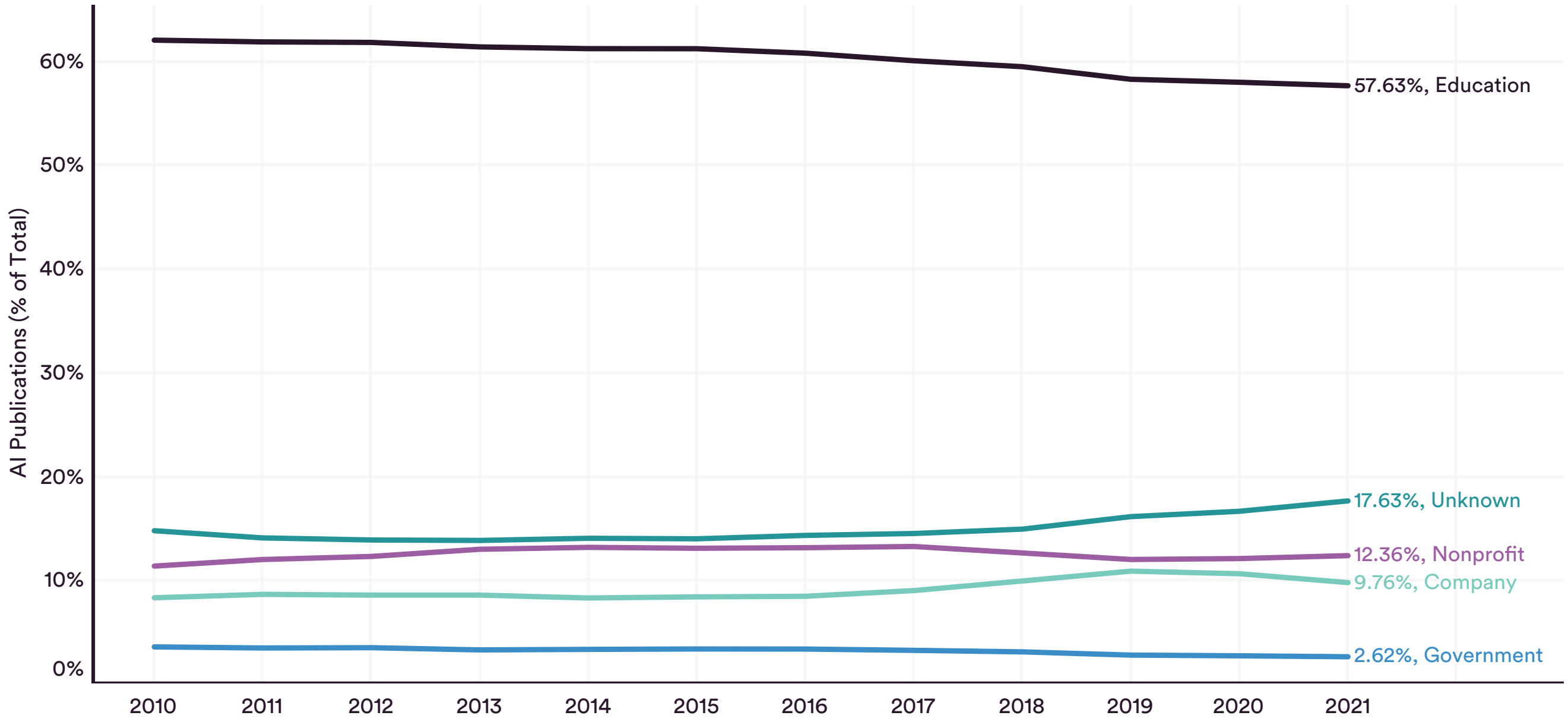

Figure 1.1.4b

3 The categorization is adapted based on the Global Research Identifier Database (GRID). See definitions of each category here. Healthcare, including hospitals and facilities, are included under nonprofit. Publications affiliated with state-sponsored universities are included in the education sector.

## AI PUBLICATIONS in CHINA (% of TOTAL) by SECTOR, 2010–21

Source: Center for Security and Emerging Technology, 2021 | Chart: 2022 AI Index Report

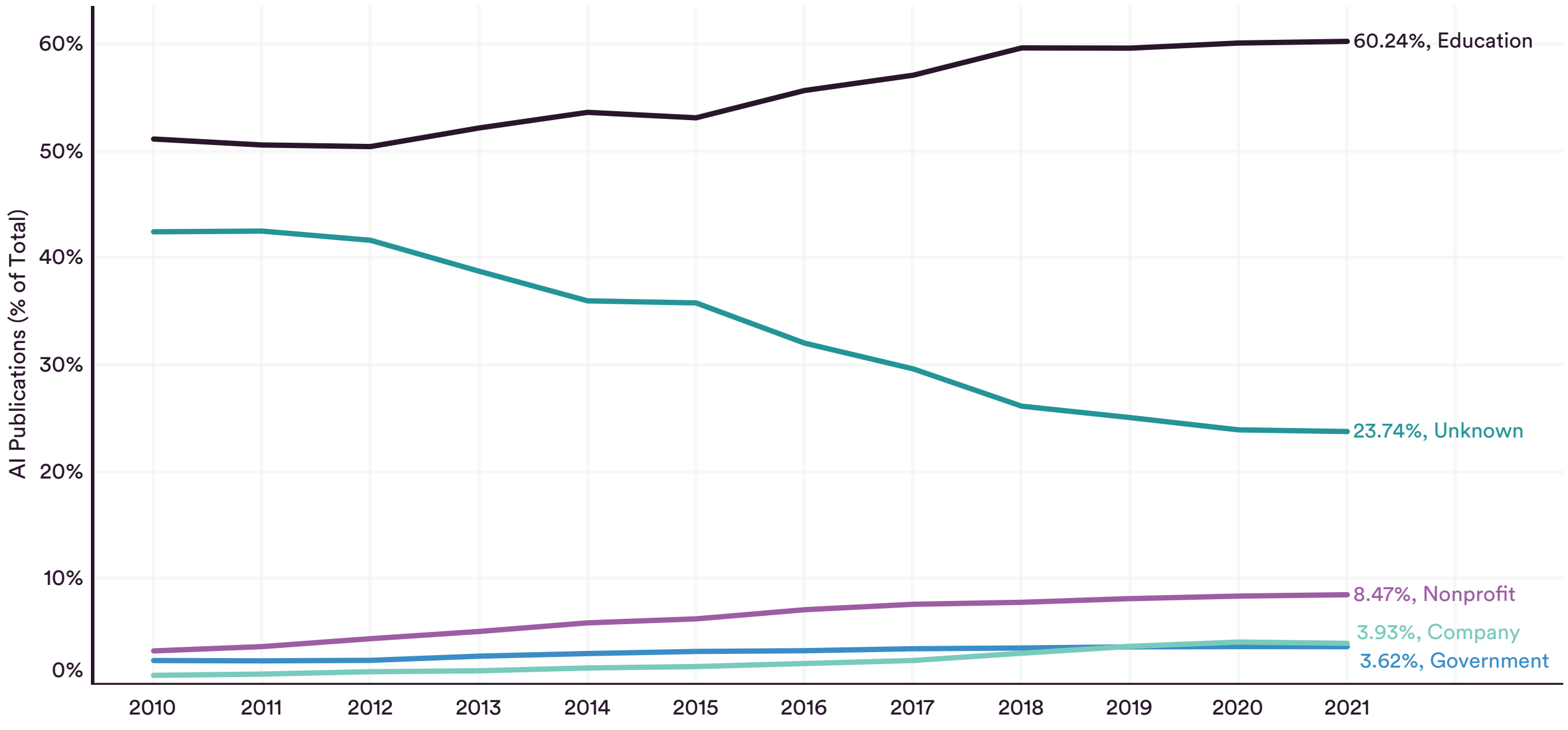

Figure 1.1.4c

## AI PUBLICATIONS in EUROPEAN UNION and UNITED KINGDOM (% of TOTAL) by SECTOR, 2010–21

Source: Center for Security and Emerging Technology, 2021 | Chart: 2022 AI Index Report

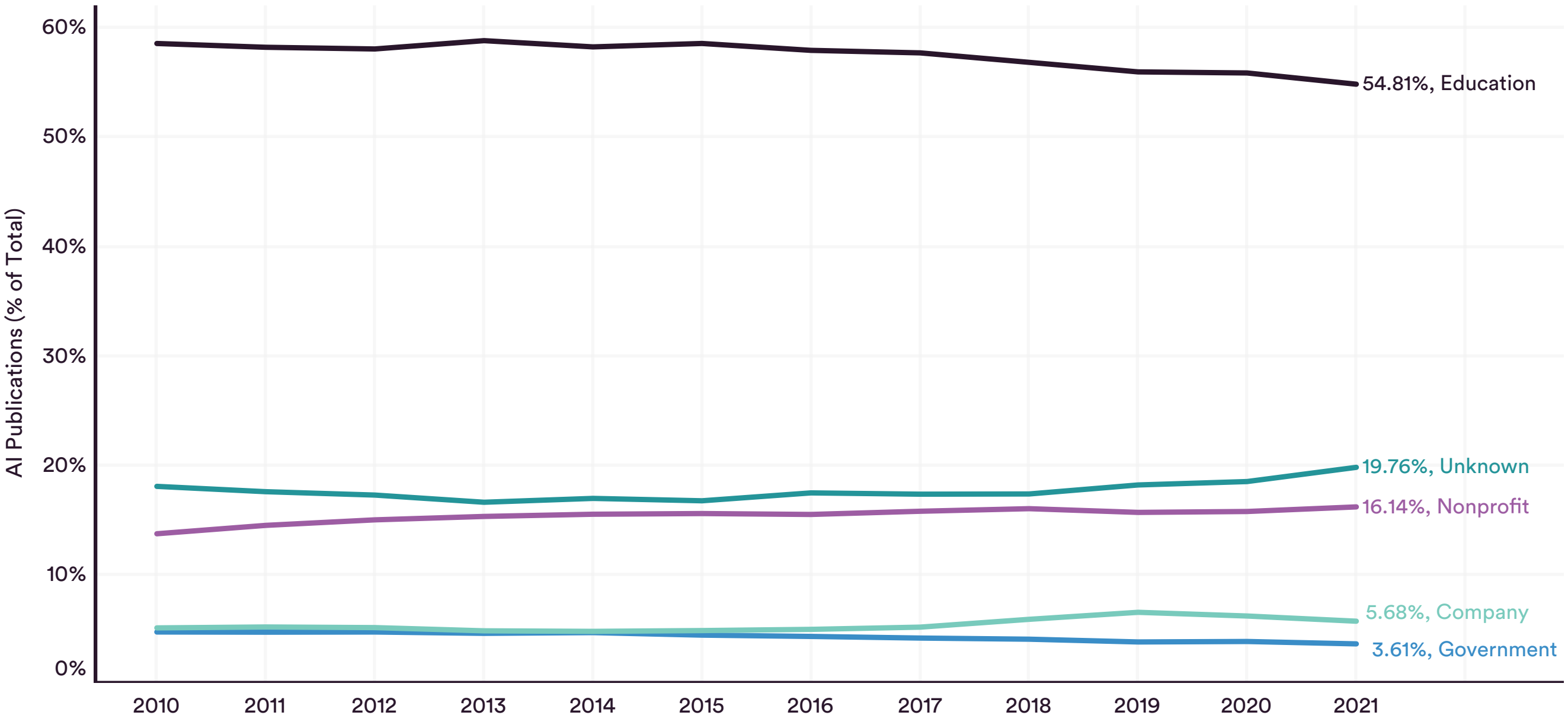

Figure 1.1.4d

## Cross-Country Collaboration

Cross-border collaborations between academics, researchers, industry experts, and others are a key component of modern STEM development that accelerate the dissemination of new ideas and the growth of research teams. Figures 1.1.5a and 1.1.5b depict the top cross-country AI collaborations from 2010 to 2021. CSET counted cross-country collaborations as distinct pairs of countries across authors for each publication (e.g., four U.S. and four Chinese-affiliated authors on a single publication are counted as one U.S.-China collaboration; two publications between the same authors counts as two collaborations).

By far, the greatest number of collaborations in the past 12 years took place between the United States and China, increasing five times since 2010. The next largest set of collaborations is between the United Kingdom and both the United States and China, which have increased more than three times since 2010. In 2021, the number of collaborations between the United States and China was 2.7 times greater than between the United Kingdom and China.

**UNITED STATES and CHINA COLLABORATIONS in AI PUBLICATIONS, 2010–21**
Source: Center for Security and Emerging Technology, 2021 | Chart: 2022 AI Index Report

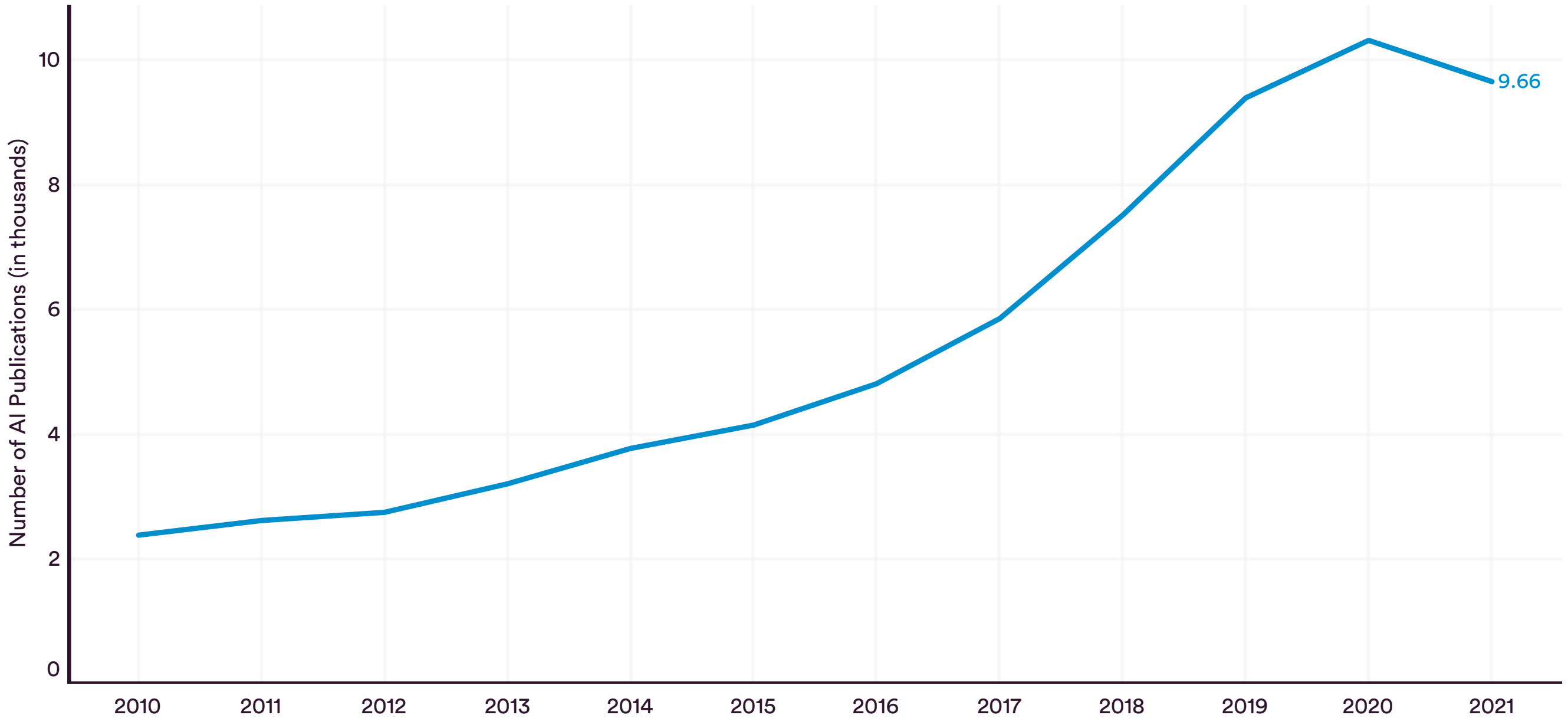

Figure 1.1.5a

**CROSS-COUNTRY COLLABORATIONS in AI PUBLICATIONS (EXCLUDING U.S. and CHINA), 2010–21**
Source: Center for Security and Emerging Technology, 2021 | Chart: 2022 AI Index Report

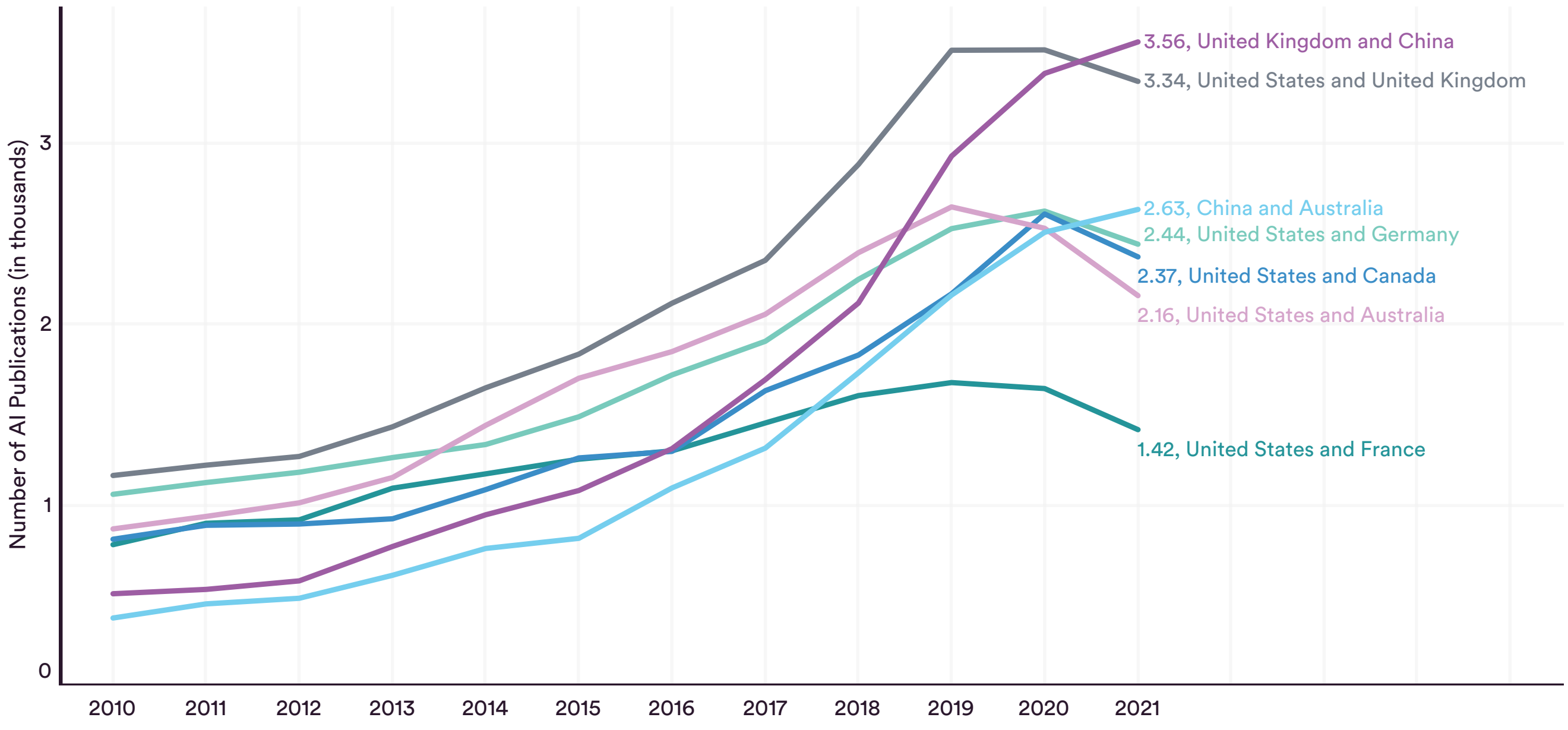

Figure 1.1.5b

## Cross-Sector Collaboration

The increase in AI research outside of universities has brought collaborations between universities and other industries. Figure 1.1.6 shows that in 2021, educational institutions and nonprofits had the greatest number of collaborations (29,839), followed by companies and educational institutions (11,576), and governments and educational institutions (8,087). There were 2.5 times as many collaborations between educational institutions and nonprofits in 2021 as between educational institutions and companies.

**CROSS-SECTOR COLLABORATIONS in AI PUBLICATIONS, 2010–21**
Source: Center for Security and Emerging Technology, 2021 | Chart: 2022 AI Index Report

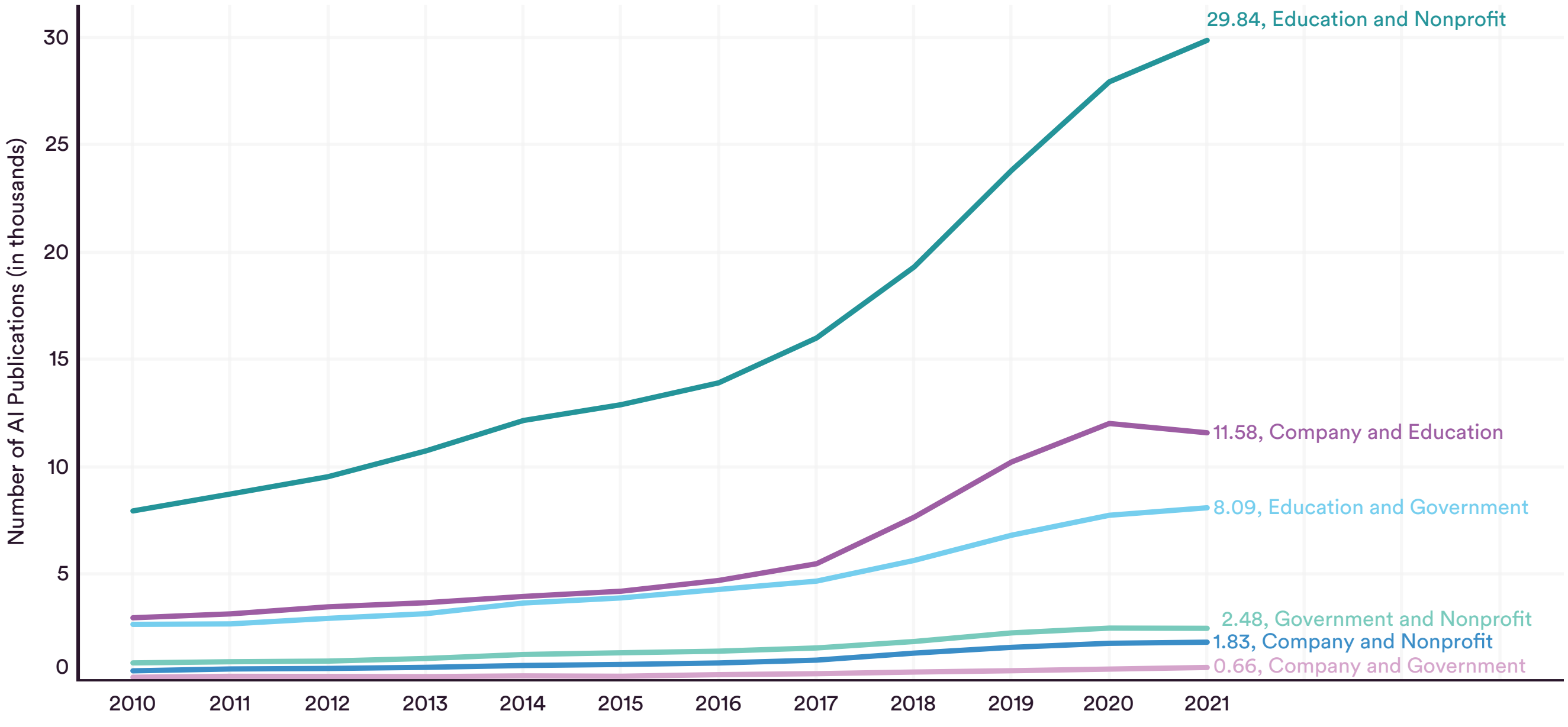

Figure 1.1.6

## AI JOURNAL PUBLICATIONS

### Overview

After growing only slightly from 2010 to 2015, the number of AI journal publications grew almost 2.5 times since 2015 (Figure 1.1.7). As a percentage of all journal publications, as captured in Figure 1.1.8, AI journal publications in 2021 were about 2.5% of all publications, compared to 1.5% in 2010.

**NUMBER of AI JOURNAL PUBLICATIONS, 2010–21**
Source: Center for Security and Emerging Technology, 2021 | Chart: 2022 AI Index Report

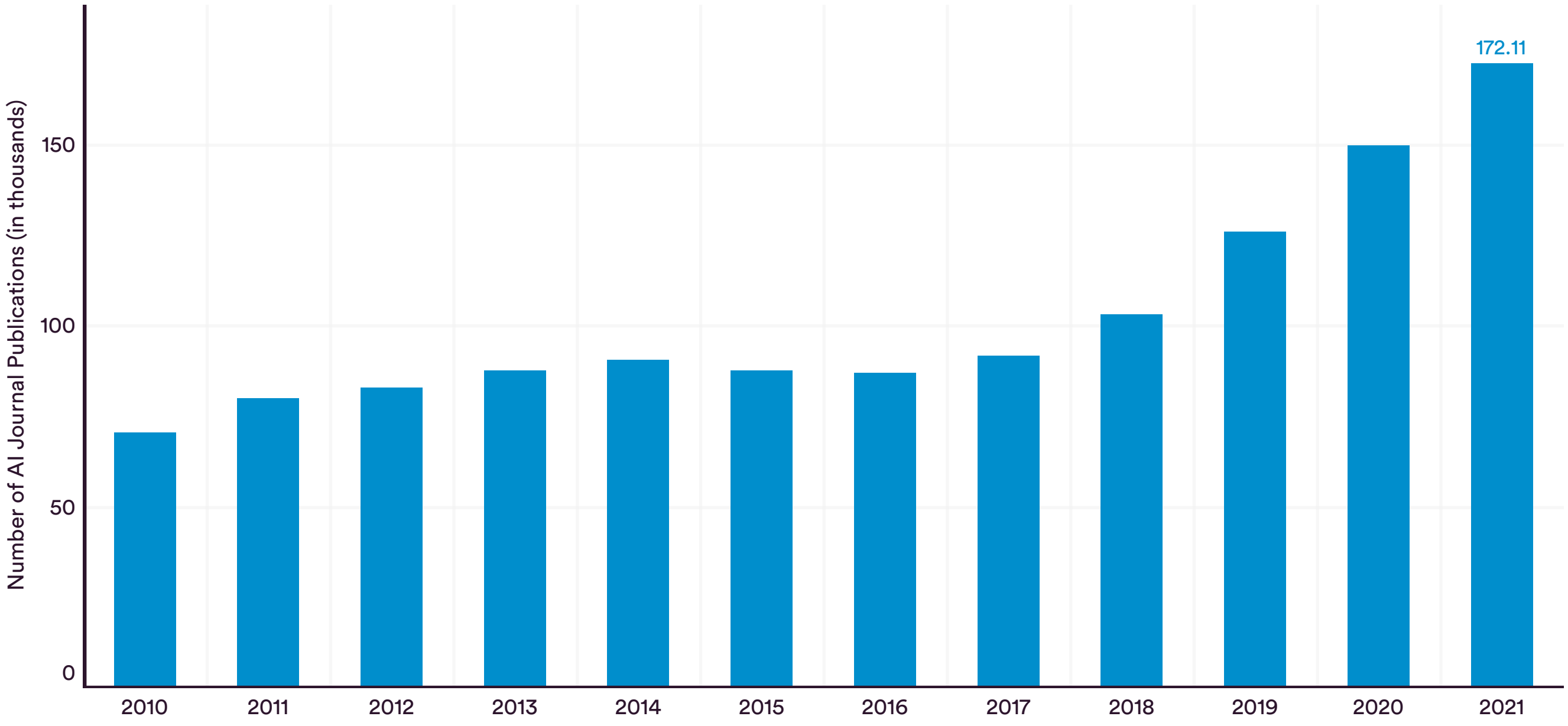

Figure 1.1.7

**AI JOURNAL PUBLICATIONS (% of TOTAL JOURNAL PUBLICATIONS), 2010–21**
Source: Center for Security and Emerging Technology, 2021 | Chart: 2022 AI Index Report

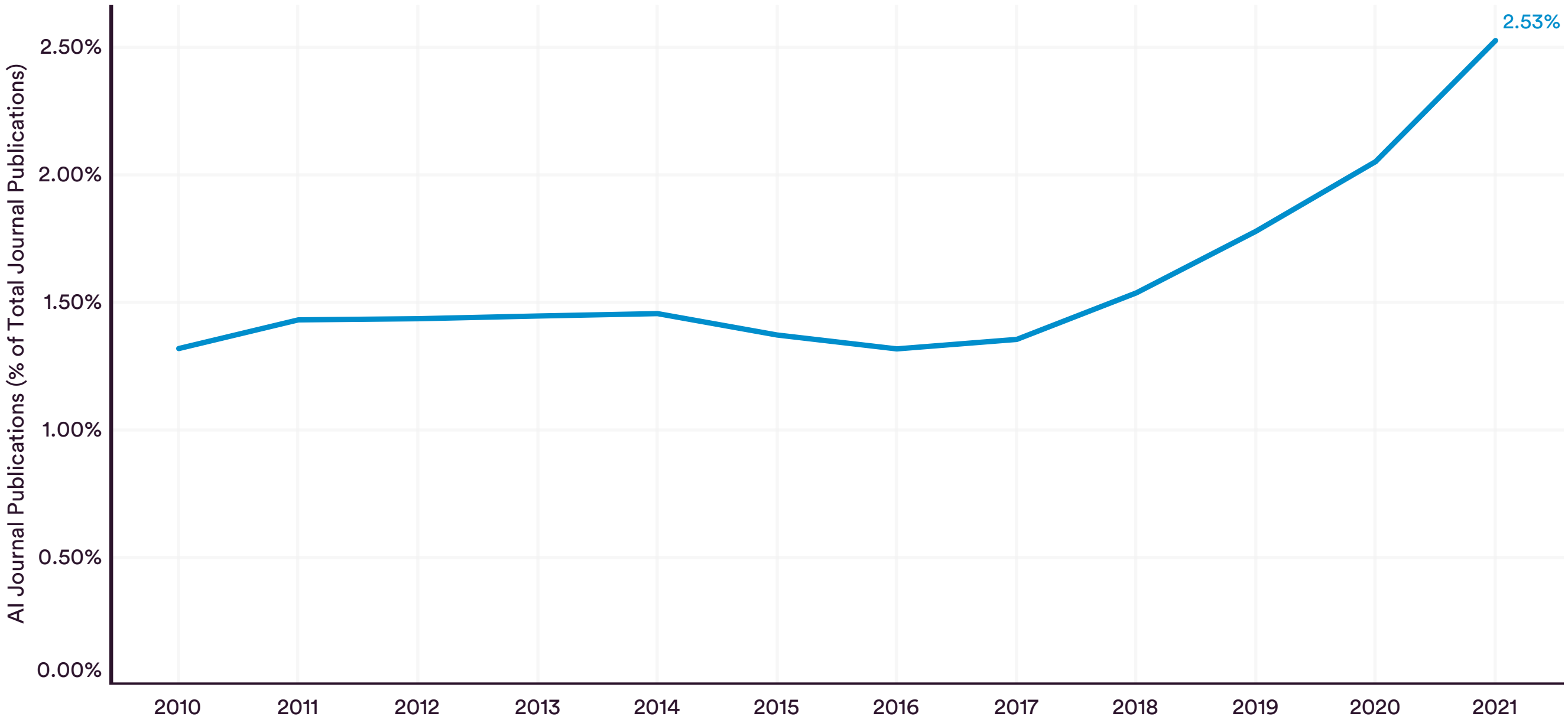

Figure 1.1.8

## By Region[4]

Figure 1.1.9 shows the share of AI journal publications by region between 2010 and 2021. In 2021, East Asia and Pacific leads with 42.9%, followed by Europe and Central Asia (22.7%) and North America (15.6%). In addition, South Asia and the Middle East and North Africa saw the most significant growth as their number of AI journal publications grew around 12 and 7 times, respectively, in the last 12 years.

### AI JOURNAL PUBLICATIONS (% of WORLD TOTAL) by REGION, 2010–21

Source: Center for Security and Emerging Technology, 2021 | Chart: 2022 AI Index Report

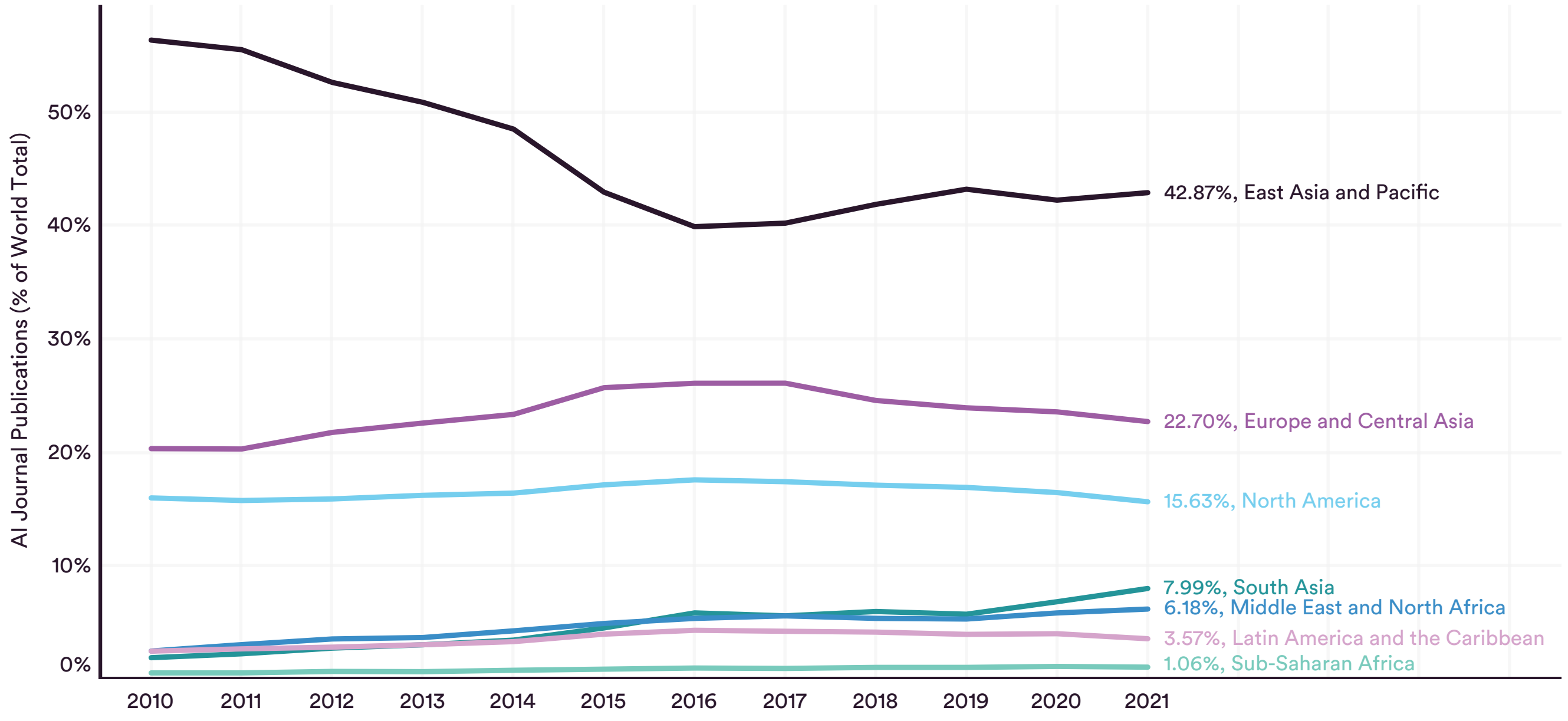

Figure 1.1.9

4 Regions in this chapter are classified according to the World Bank analytical grouping.

## By Geographic Area[5]

Figure 1.1.10 breaks down the share of AI journal publications over the past 12 years by three major AI powers. China has remained the leader throughout, with 31.0% in 2021, followed by the European Union plus the United Kingdom at 19.1% and the United States at 13.7%

### AI JOURNAL PUBLICATIONS (% of WORLD TOTAL) by GEOGRAPHIC AREA, 2010–21

Source: Center for Security and Emerging Technology, 2021 | Chart: 2022 AI Index Report

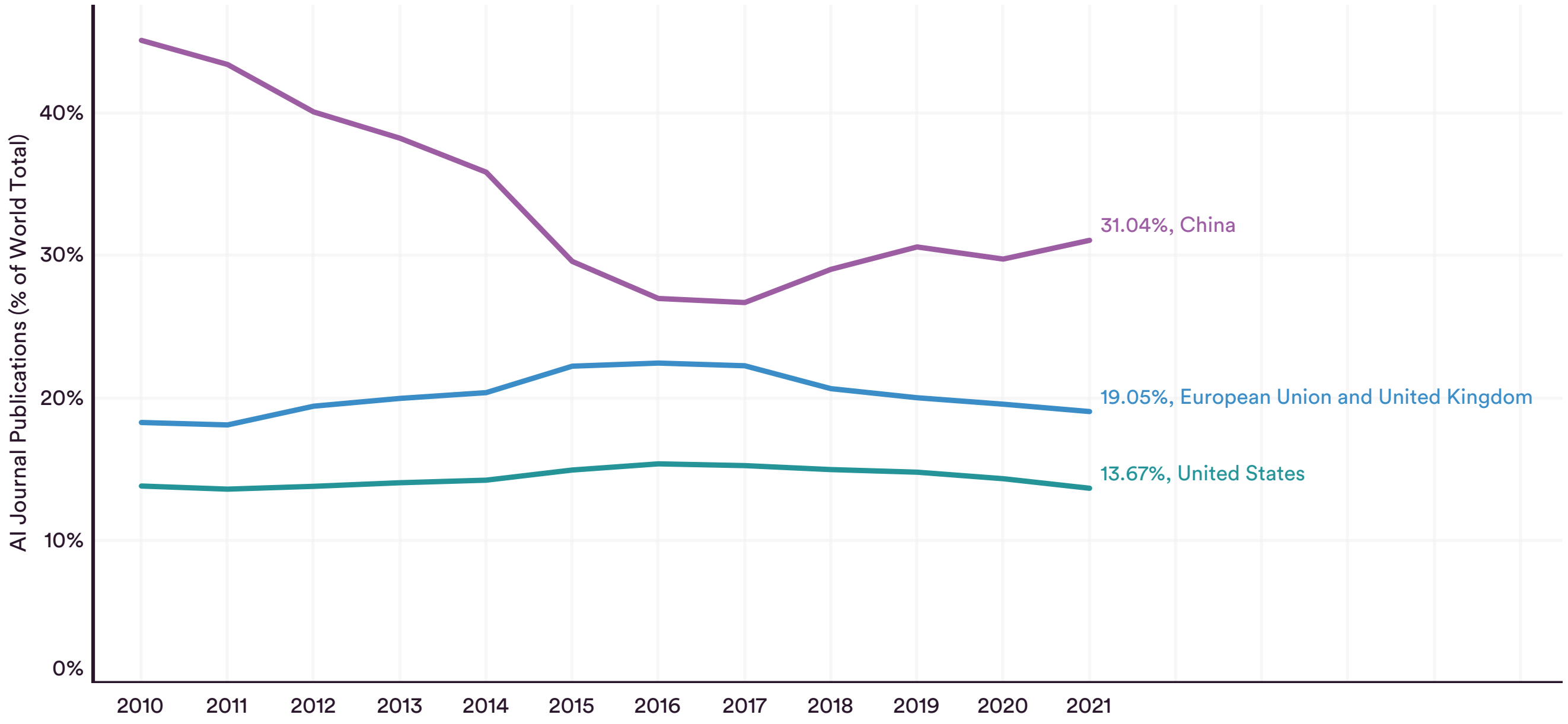

Figure 1.1.10

5 Geographic areas in this chapter combine the number of publications between the European Union and the United Kingdom to reflect the historically strong association between them with regards to research collaboration.

## Citations

On the number of citations of AI journal publications, China's share has gradually increased while those of the European Union plus the United Kingdom and the United States have decreased. The three geographic areas combined accounted for more than 66% of the total citations in the world.

**AI JOURNAL CITATIONS (% of WORLD TOTAL) by GEOGRAPHIC AREA, 2010–21**

Source: Center for Security and Emerging Technology, 2021 | Chart: 2022 AI Index Report

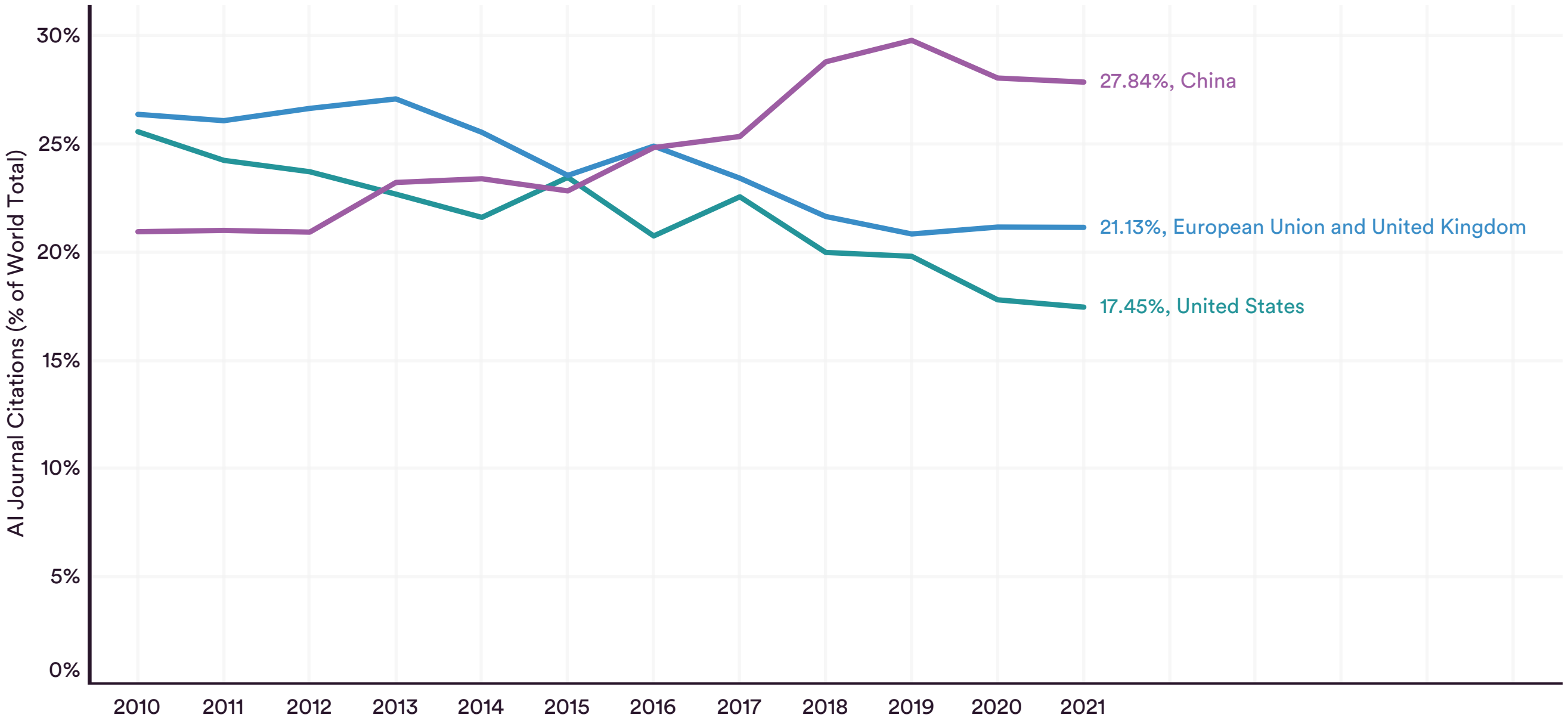

Figure 1.1.11

## AI CONFERENCE PUBLICATIONS

### Overview

The number of AI conference publications peaked in 2019, and fell about 19.4% below the peak in 2021 (Figure 1.1.12). Despite the decline in the total numbers, however, the share of AI conference publications among total conference publications in the world has increased by more than five percentage points since 2010 (Figure 1.1.13).

**NUMBER of AI CONFERENCE PUBLICATIONS, 2010–21**
Source: Center for Security and Emerging Technology, 2021 | Chart: 2022 AI Index Report

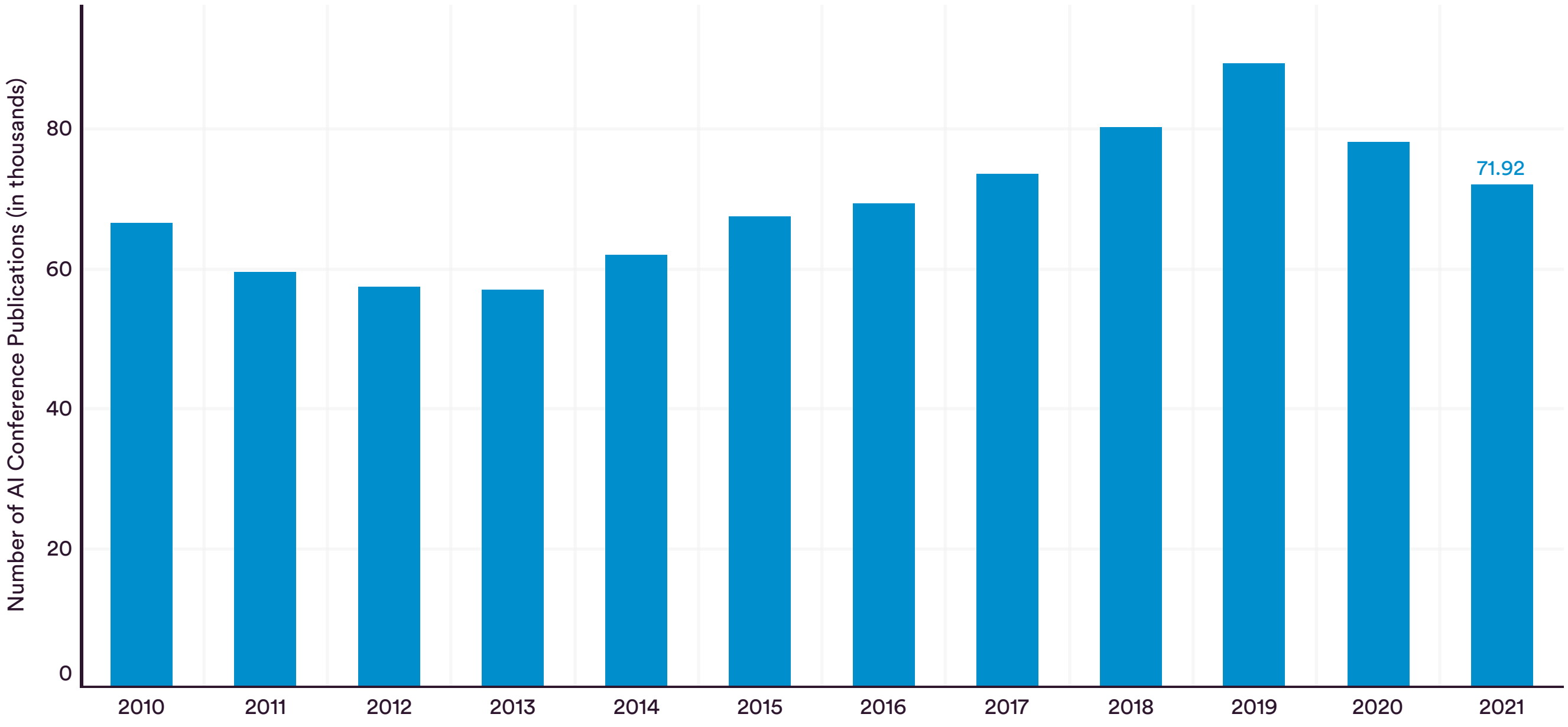

Figure 1.1.12

**AI CONFERENCE PUBLICATIONS (% of TOTAL CONFERENCE PUBLICATIONS), 2010–21**
Source: Center for Security and Emerging Technology, 2021 | Chart: 2022 AI Index Report

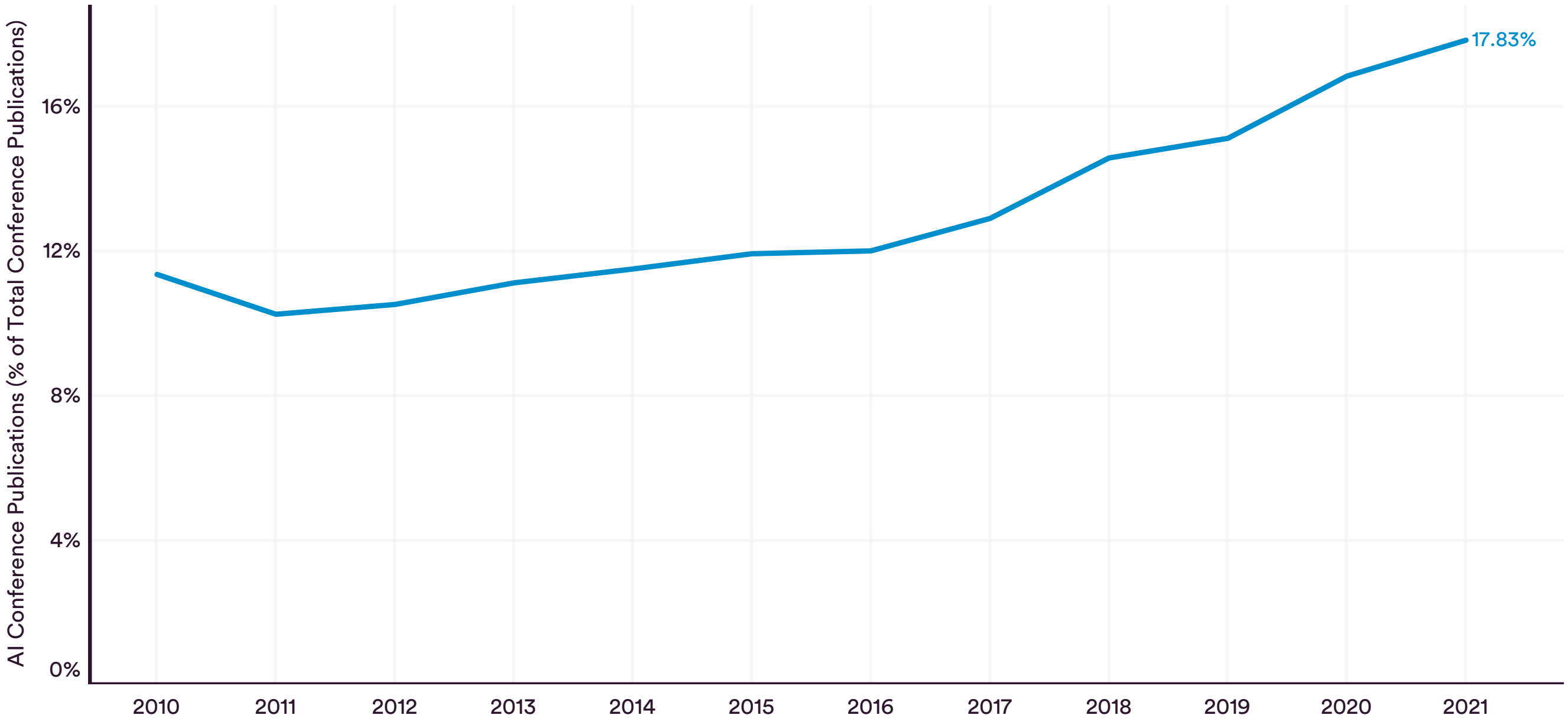

Figure 1.1.13

## By Region

Figure 1.1.14 shows the number of AI conference publications by region. Similar to the trend in AI journal publication, East Asia and Pacific, Europe and Central Asia, and North America account for the world's highest numbers of AI conference publications. Specifically, the share represented by East Asia and Pacific continues to rise since taking the lead in 2014, accounting for 40.4% in 2021, followed by Europe and Central Asia (23.0%) and North America (19.0%). The percentage of AI conference publications in South Asia saw a noticeable rise in the past 12 years, from 4.0% in 2010 to 10.4% in 2021.

**AI CONFERENCE PUBLICATIONS (% of WORLD TOTAL) by REGION, 2010–21**

Source: Center for Security and Emerging Technology, 2021 | Chart: 2022 AI Index Report

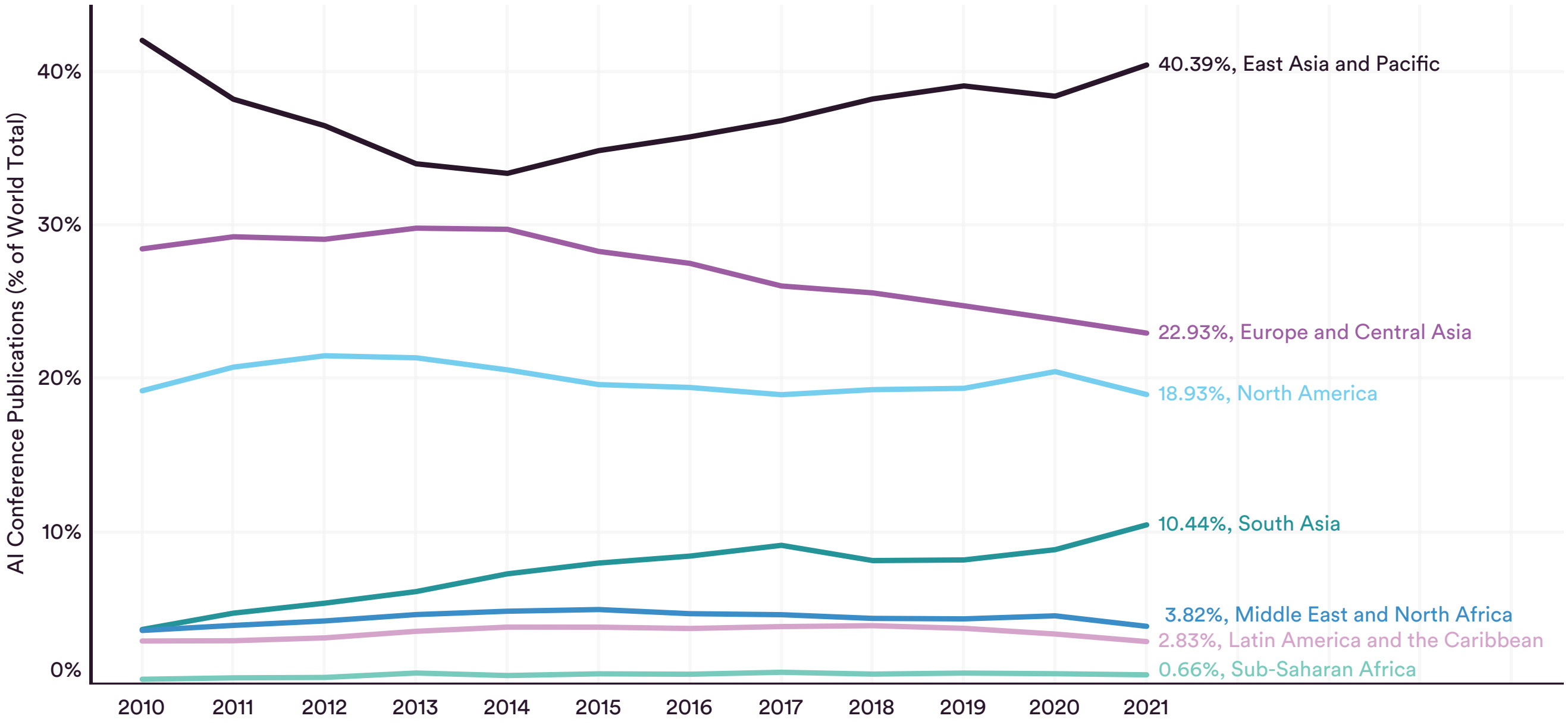

Figure 1.1.14

## By Geographic Area

In 2021, China produced the greatest share of the world's AI conference publications at 27.6%, opening an even greater lead than in 2020, while the European Union plus the United Kingdom followed at 19.0% and the United States came in third at 16.9% (Figure 1.1.15).

**AI CONFERENCE PUBLICATIONS (% of WORLD TOTAL) by GEOGRAPHIC AREA, 2010–21**

Source: Center for Security and Emerging Technology, 2021 | Chart: 2022 AI Index Report

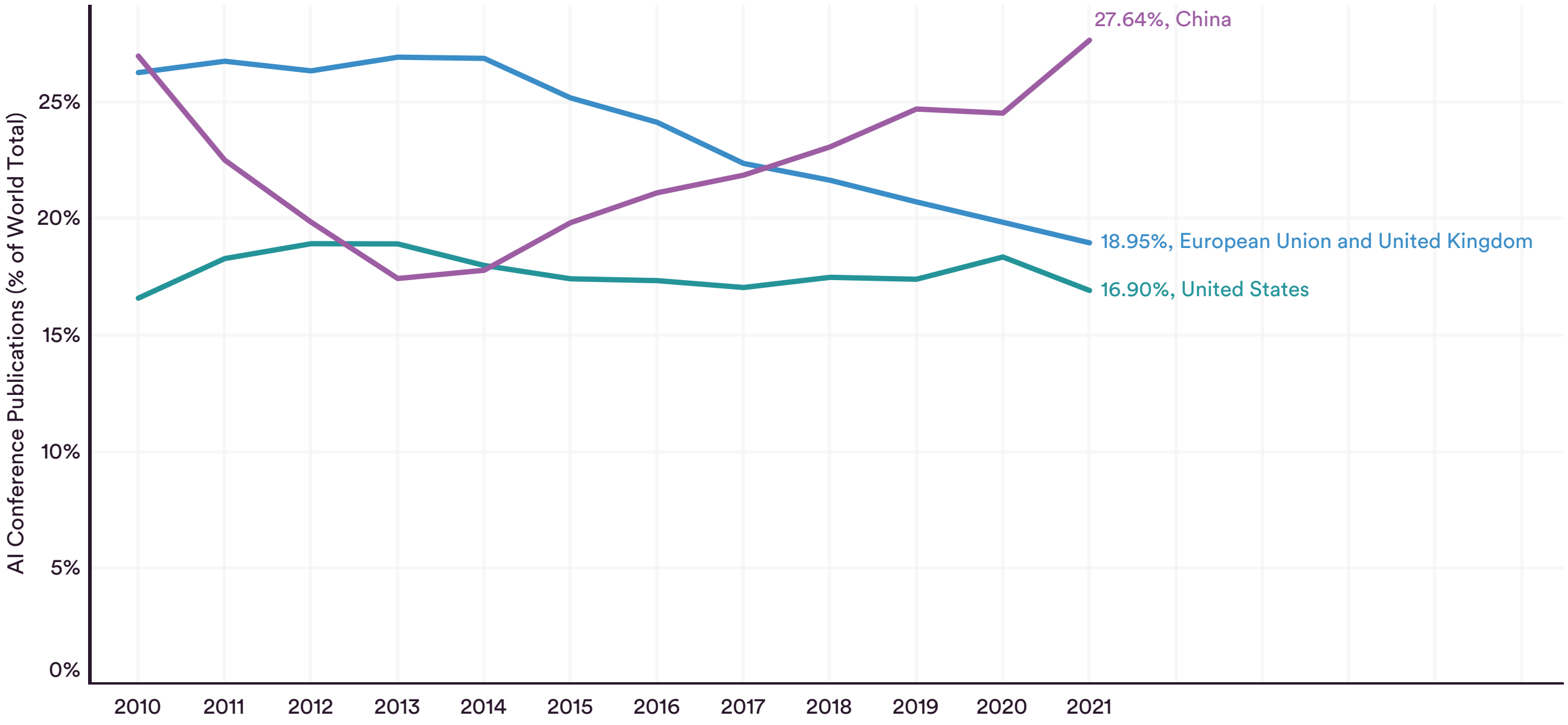

Figure 1.1.15

## Citations

Despite China producing the most AI conference publications in 2021, Figure 1.1.16 shows that the United States led among the major powers with respect to the number of AI conference citations, with 29.5% in 2021, followed by the European Union plus the United Kingdom (23.3%) and China (15.3%).

AI CONFERENCE CITATIONS (% of WORLD TOTAL) by GEOGRAPHIC AREA, 2010–21

Source: Center for Security and Emerging Technology, 2021 | Chart: 2022 AI Index Report

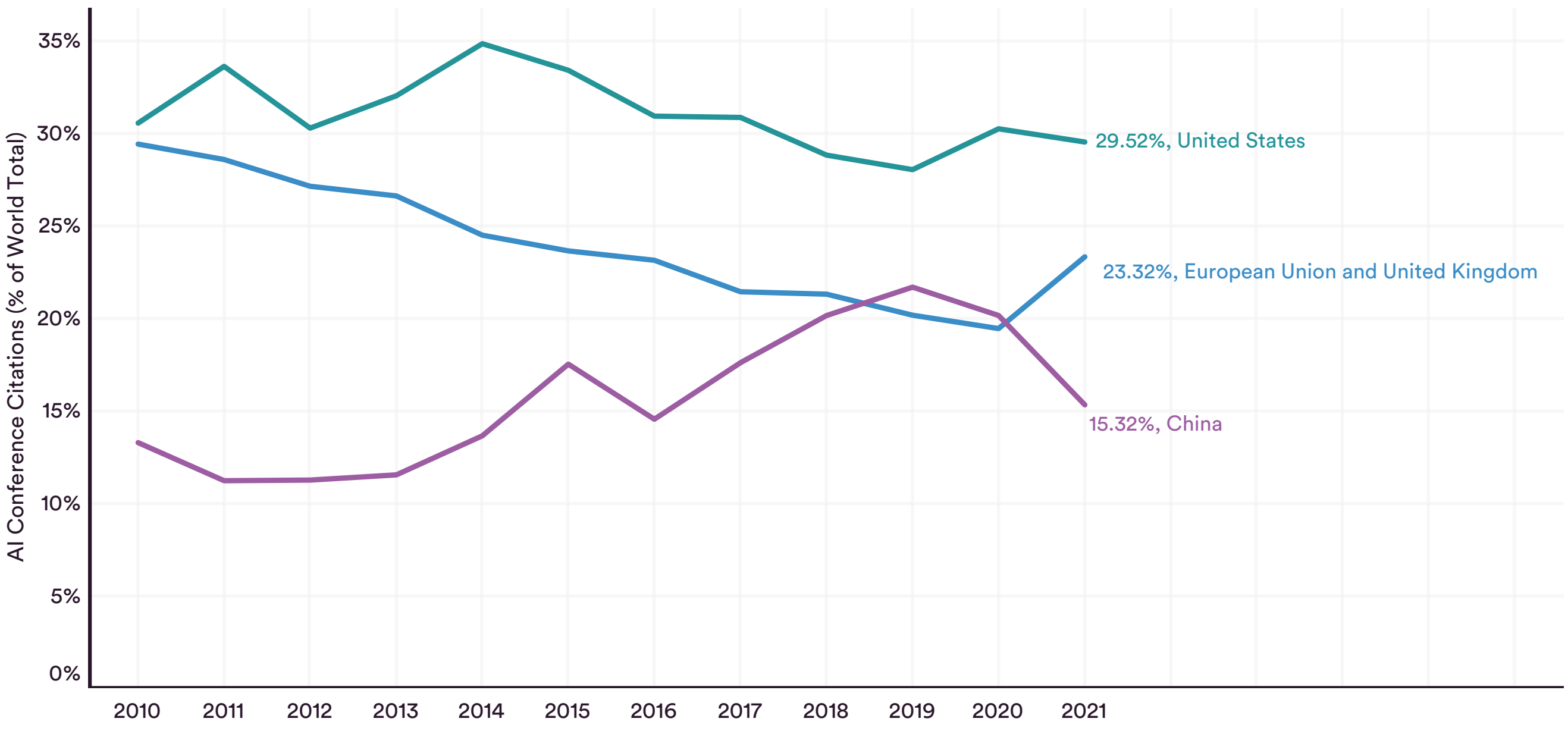

Figure 1.1.16

## AI REPOSITORIES

### Overview

Publishing pre-peer-reviewed papers on repositories of electronic preprints (such as arXiv and SSRN) has become a popular way among AI researchers to disseminate their work outside traditional avenues for publications. Those repositories allow researchers to share their findings before submitting them to journals and conferences, which greatly accelerates the cycle of information discovery. The number of AI repository publications grew almost 30 times in the past 12 years (Figure 1.1.17), now account for 15.3% of all repository publications (Figure 1.1.18).

**NUMBER of AI REPOSITORY PUBLICATIONS, 2010–21**

Source: Center for Security and Emerging Technology, 2021 | Chart: 2022 AI Index Report

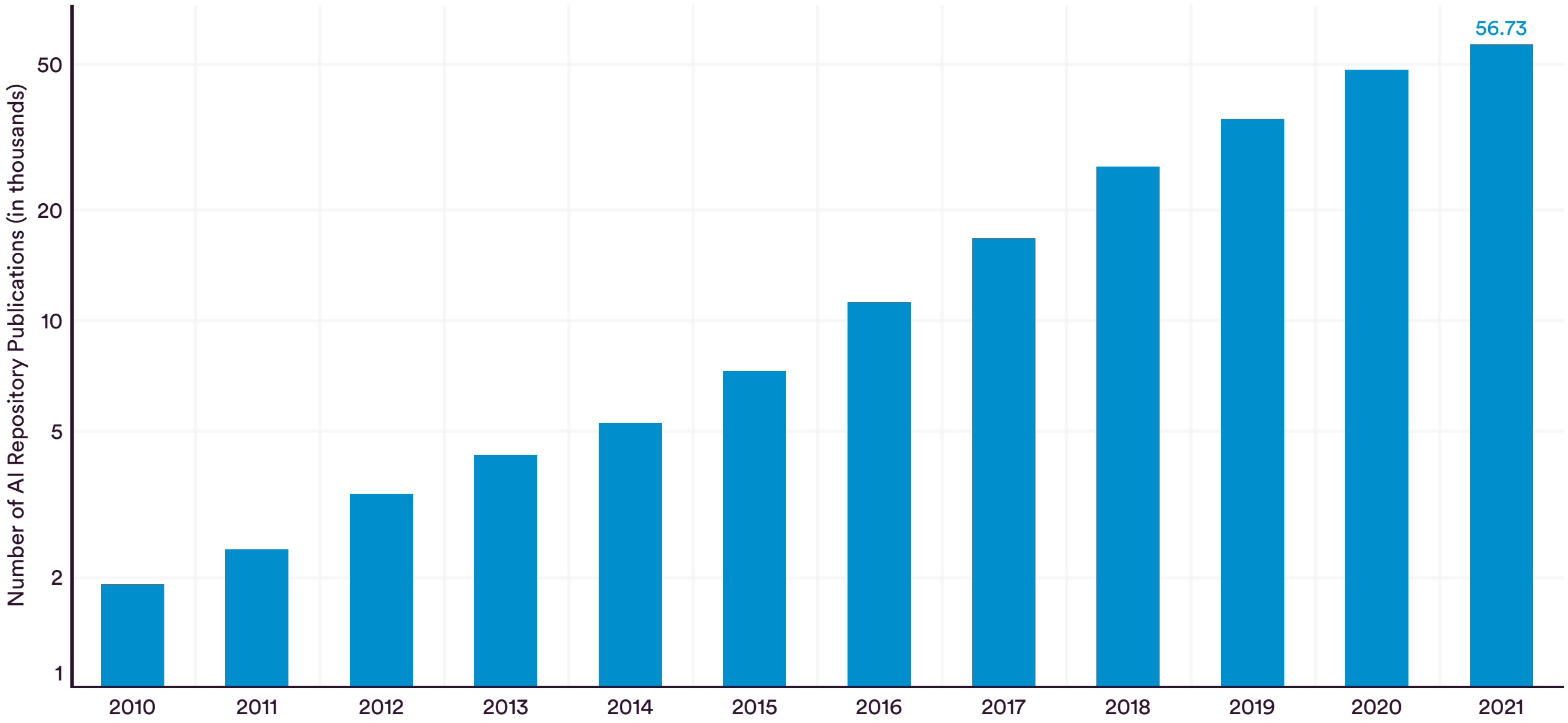

Figure 1.1.17

**AI REPOSITORY PUBLICATIONS (% of TOTAL REPOSITORY PUBLICATIONS), 2010–21**

Source: Center for Security and Emerging Technology, 2021 | Chart: 2022 AI Index Report

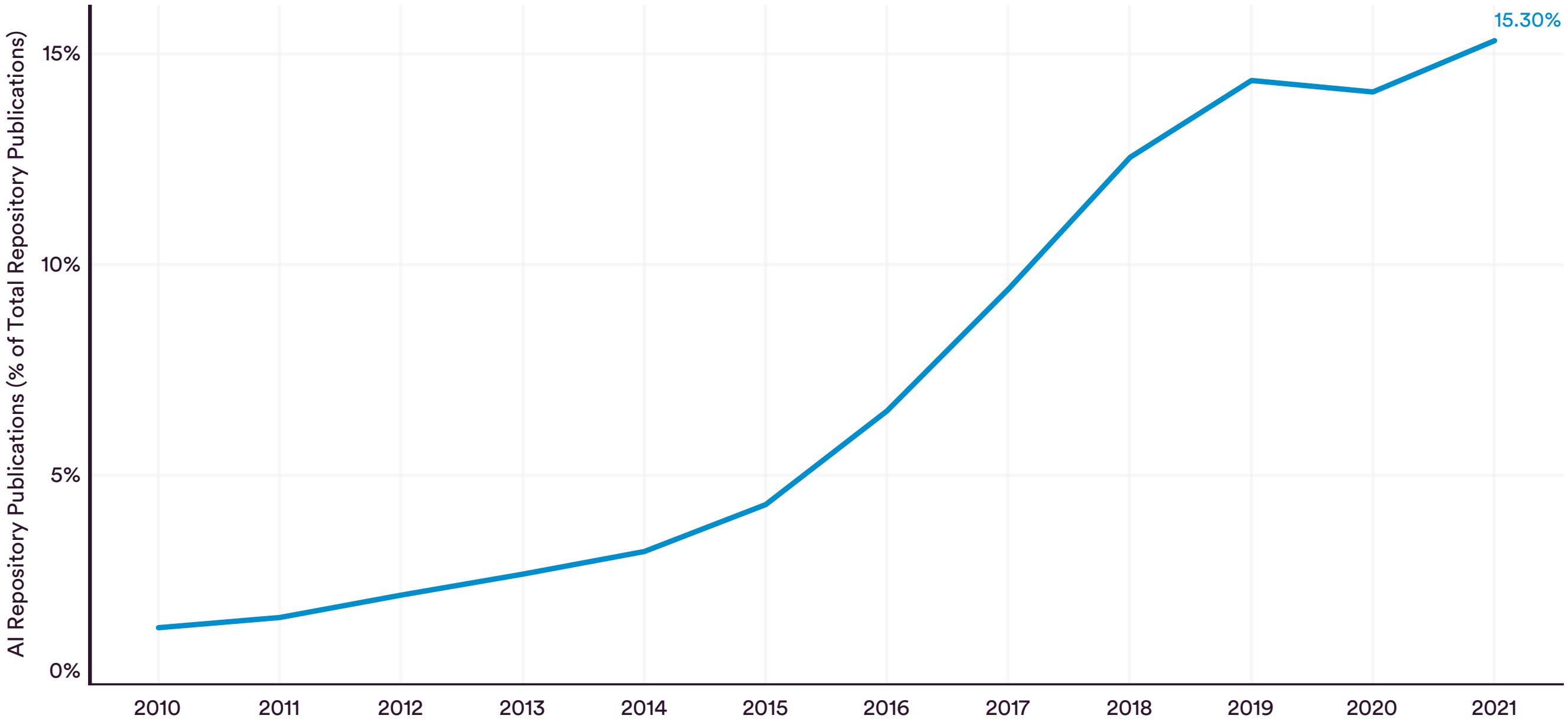

Figure 1.1.18

## By Region

The analysis by region in Figure 1.1.19 shows that North America has maintained a steady lead in the share of AI repository publications in the world since 2014 while that of Europe and Central Asia has declined. Since 2013, the share of East Asia and Pacific has grown significantly.

**AI REPOSITORY PUBLICATIONS (% of WORLD TOTAL) by REGION, 2010–21**

Source: Center for Security and Emerging Technology, 2021 | Chart: 2022 AI Index Report

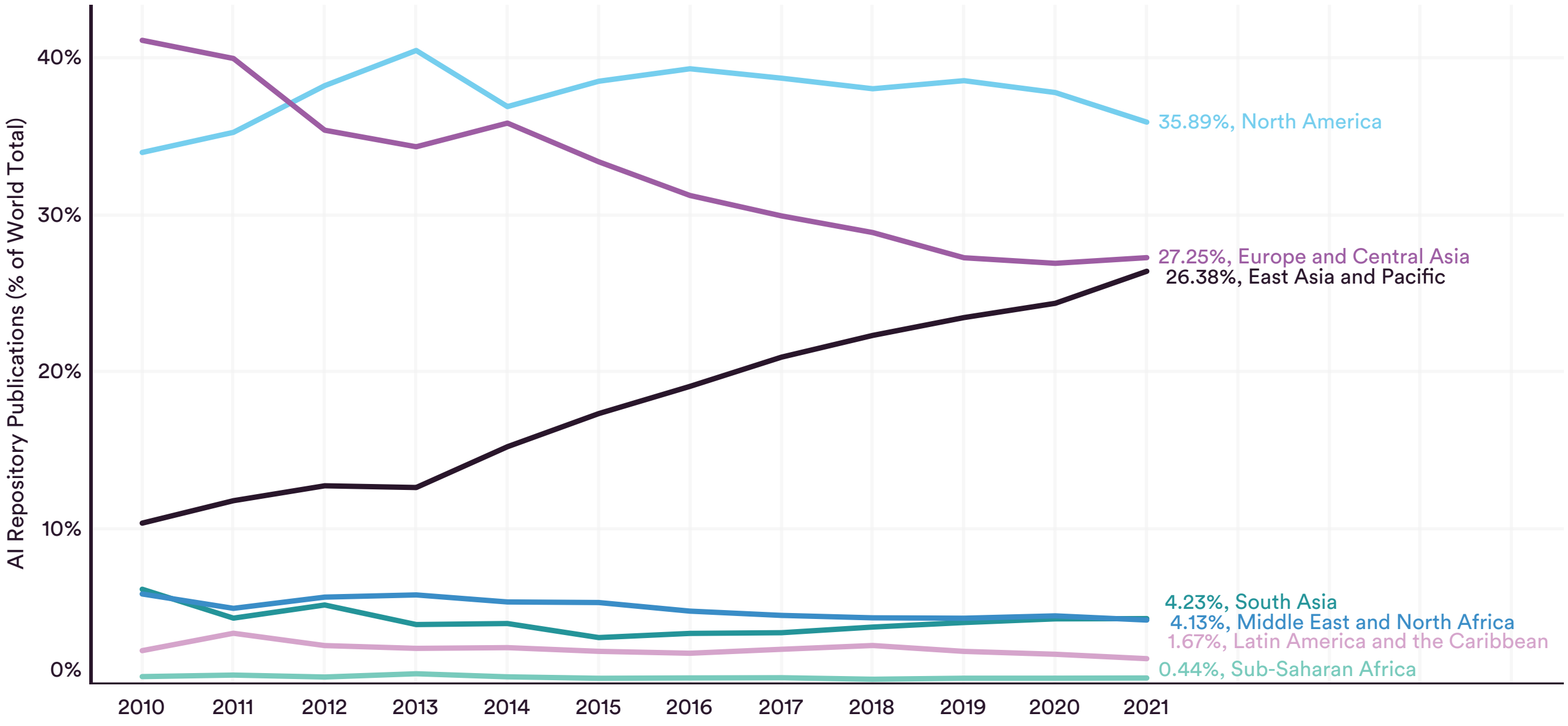

Figure 1.1.19

## By Geographic Area

While the United States has held the lead in the percentage of AI repository publications in the world since 2011, China is catching up while the European Union plus the United Kingdom's share continues to drop (Figure 1.1.20). In 2021, the United States accounted for 32.5% of the world's AI repository publications—a higher percentage compared to journal and conference publications, followed by the European Union plus the United Kingdom (23.9%) and China (16.6%).

**AI REPOSITORY PUBLICATIONS (% of WORLD TOTAL) by GEOGRAPHIC AREA, 2010–21**

Source: Center for Security and Emerging Technology, 2021 | Chart: 2022 AI Index Report

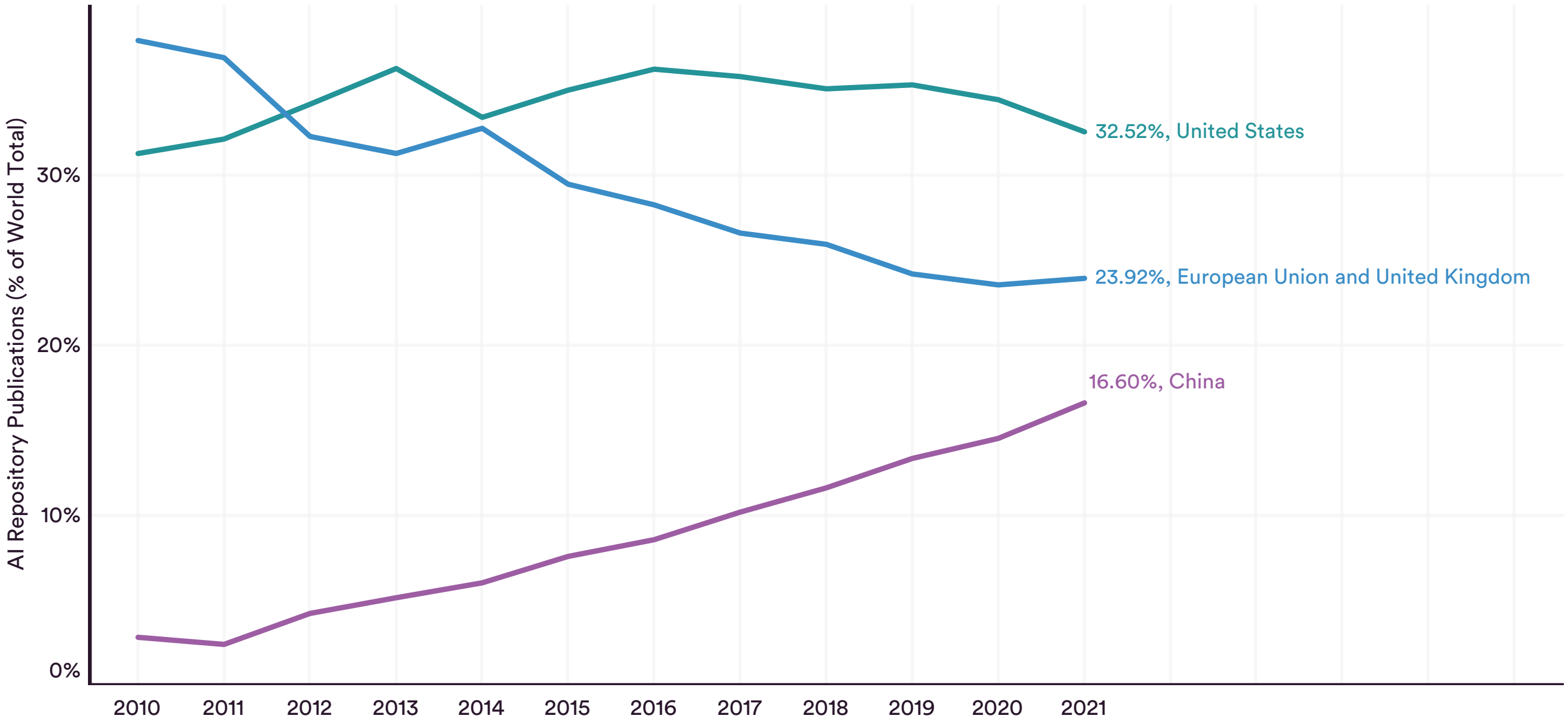

Figure 1.1.20

## Citations

On the citations of AI repository publications, Figure 1.1.21 shows that the United States tops the list with 38.6% of overall citations in 2021, establishing a dominant lead over the European Union plus the United Kingdom (20.1%) and China (16.4%).

**AI REPOSITORY CITATIONS (% of WORLD TOTAL) by GEOGRAPHIC AREA, 2010–21**

Source: Center for Security and Emerging Technology, 2021 | Chart: 2022 AI Index Report

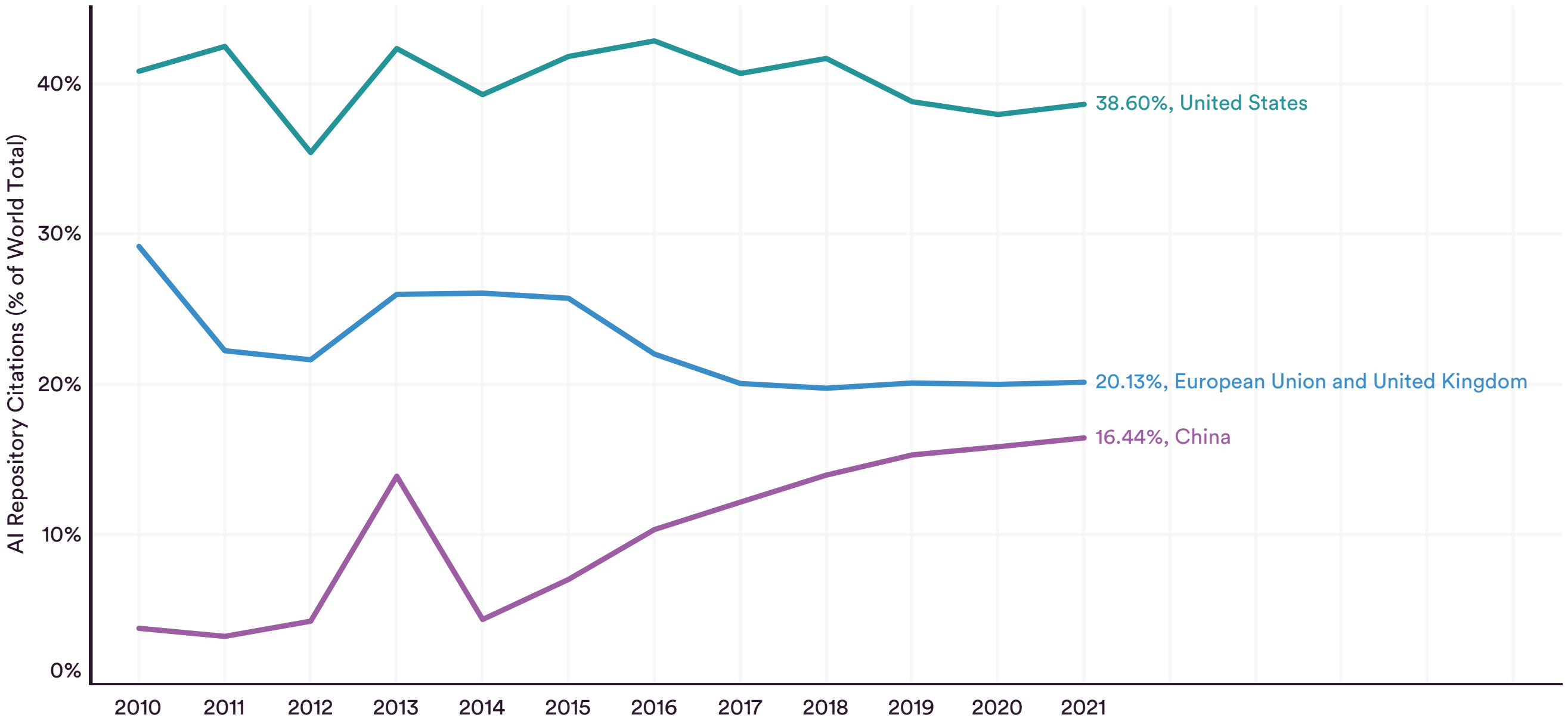

Figure 1.1.21

## AI PATENTS

This section draws on data from CSET and 1790 Analytics on patents relevant to AI development and applications—indicated by Cooperative Patent Classification (CPC)/International Patent Classification (IPC) codes and keywords. Patents were grouped by country and year and then counted at the "patent family" level, before CSET extracted year values from the most recent publication date within a family.

### Overview

Figure 1.1.22 captures the number of AI patents filed from 2010 to 2021. The number of patents filed in 2021 is more than 30 times higher than in 2015, showing a compound annual growth rate of 76.9%.

**NUMBER of AI PATENT FILINGS, 2010–21**

Source: Center for Security and Emerging Technology, 2021 | Chart: 2022 AI Index Report

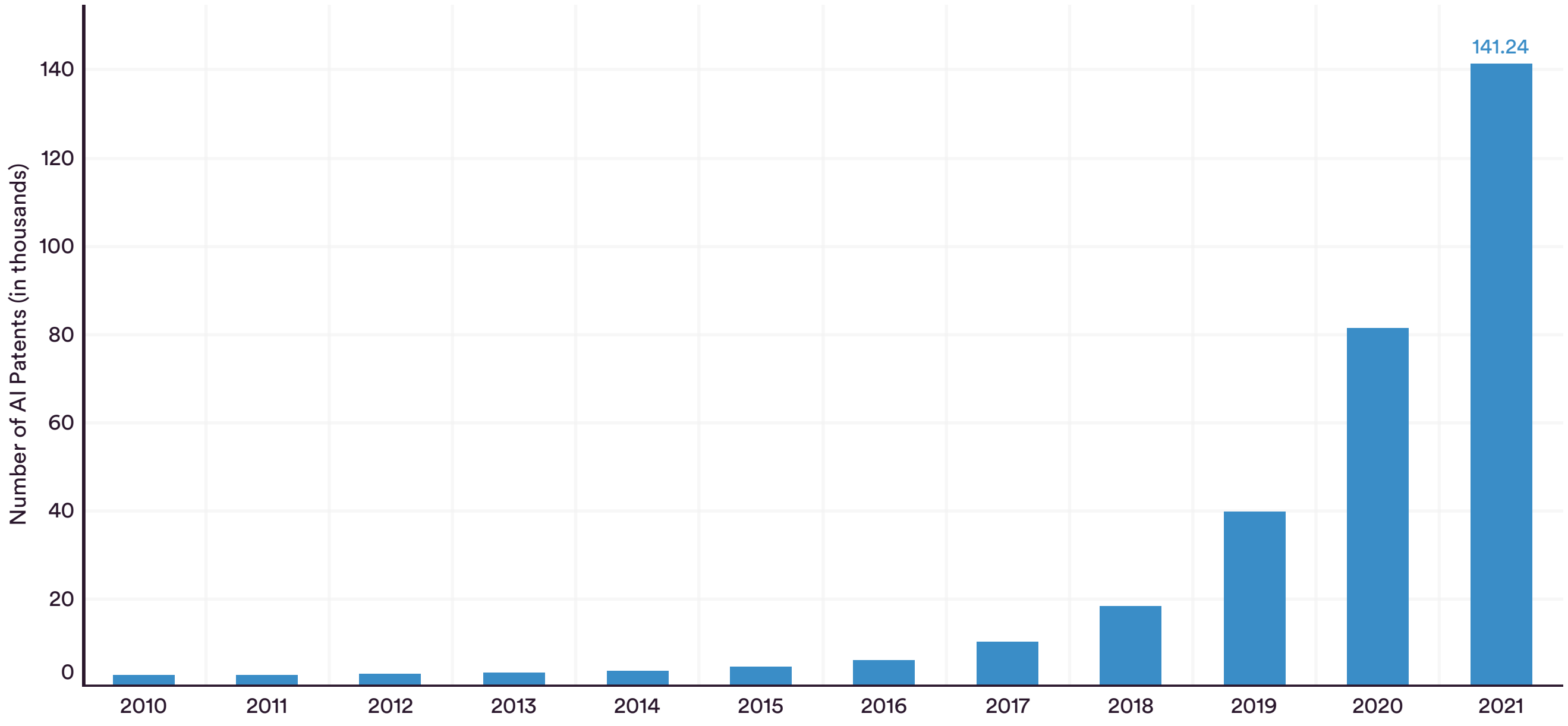

Figure 1.1.22

## By Region and Application Status

Figure 1.1.23a breaks down AI patent filings by region. The share of East Asia and Pacific took off in 2014 and led the rest of the world in 2021 with 62.1% of all patent applications, followed by North America and Europe and Central Asia. In terms of granted patents in those regions, North America leads with 57.0%, followed by East Asia and Pacific (31.0%), and Europe and Central Asia (11.3%) (Figure 1.1.23b). The other regions combine to make up roughly 1% of world patents (Figure 1.1.23c).

**AI PATENT FILINGS (% of WORLD TOTAL) by REGION, 2010–21**

Source: Center for Security and Emerging Technology, 2021 | Chart: 2022 AI Index Report

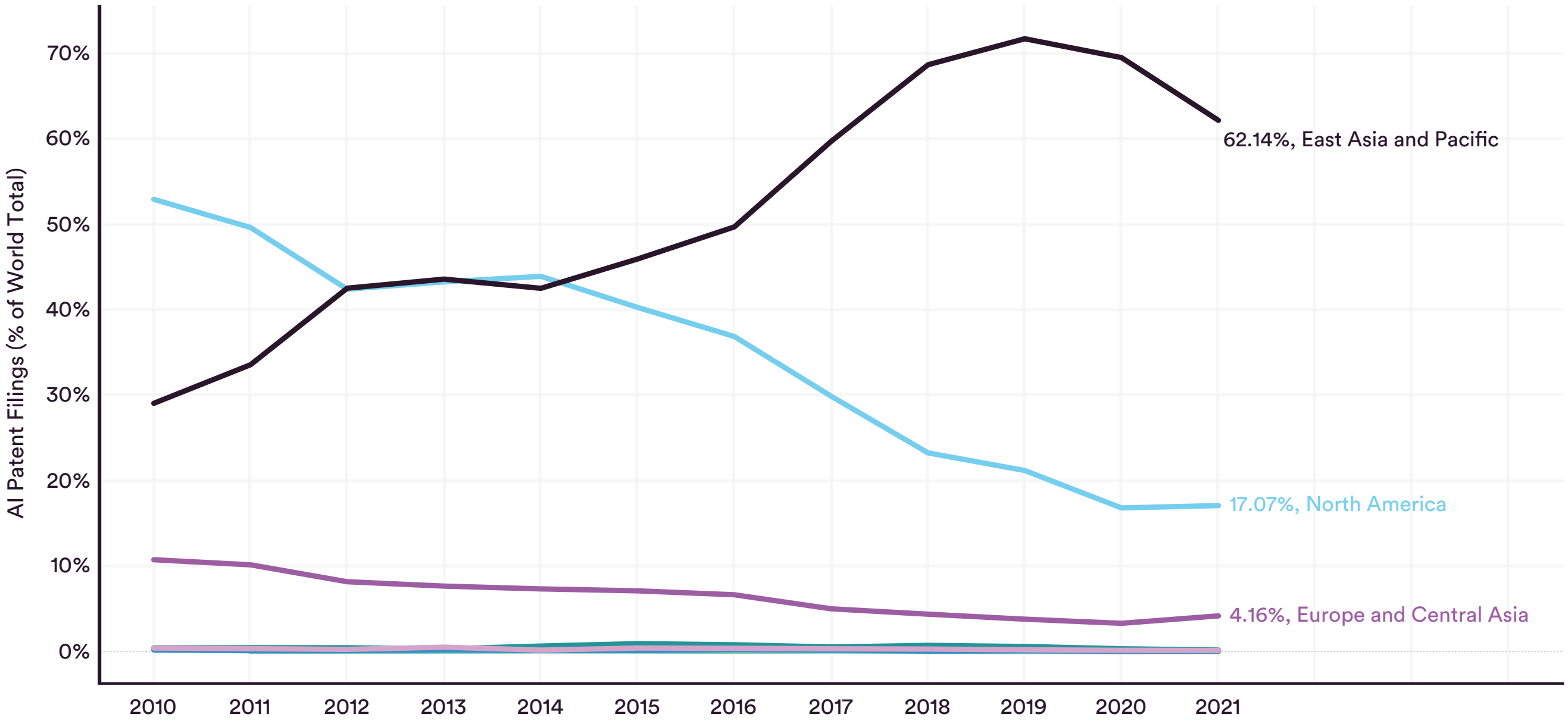

Figure 1.1.23a

## GRANTED AI PATENTS (% of WORLD TOTAL) by REGION, 2010–21 (Part 1)

Source: Center for Security and Emerging Technology, 2021 | Chart: 2022 AI Index Report

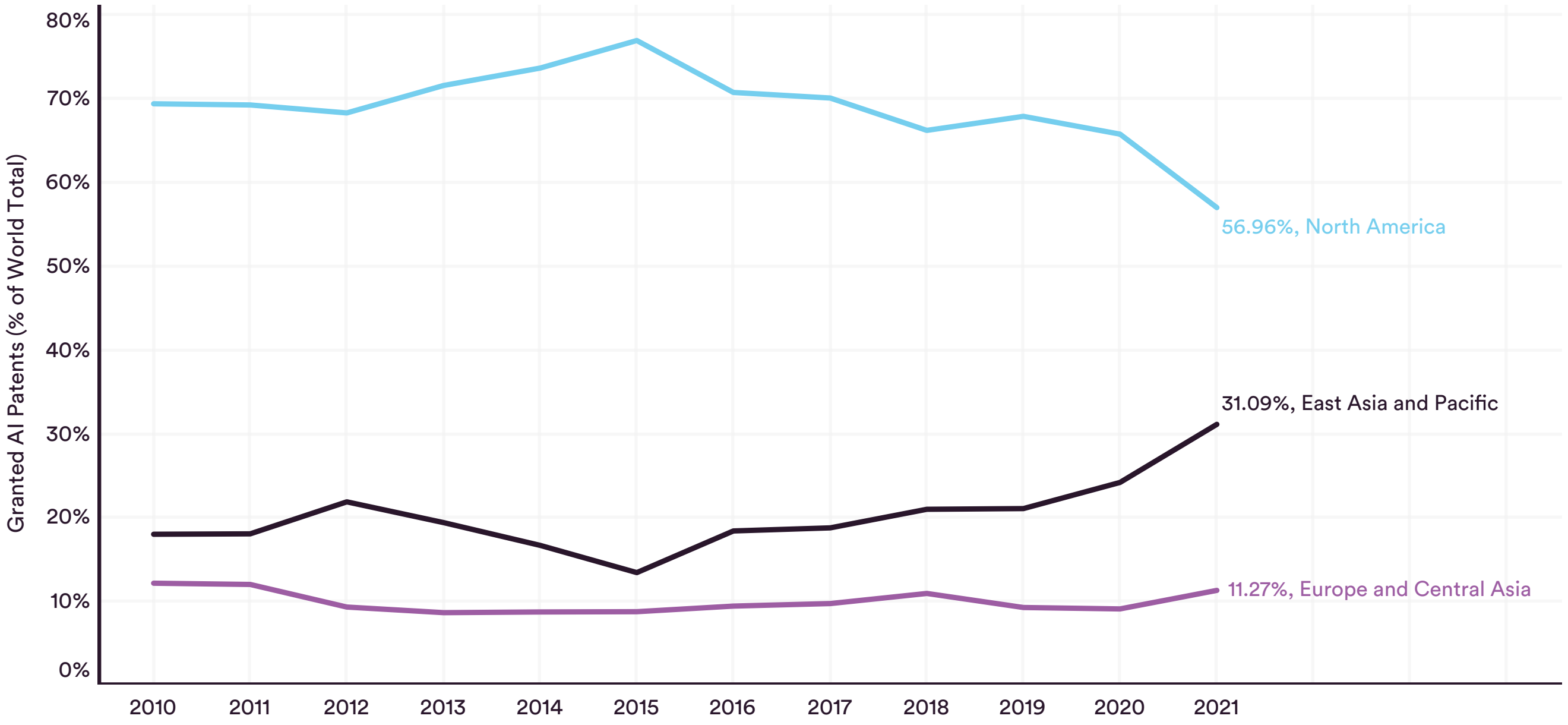

Figure 1.1.23b

## GRANTED AI PATENTS (% of WORLD TOTAL) by REGION, 2010–21 (Part 2)

Source: Center for Security and Emerging Technology, 2021 | Chart: 2022 AI Index Report

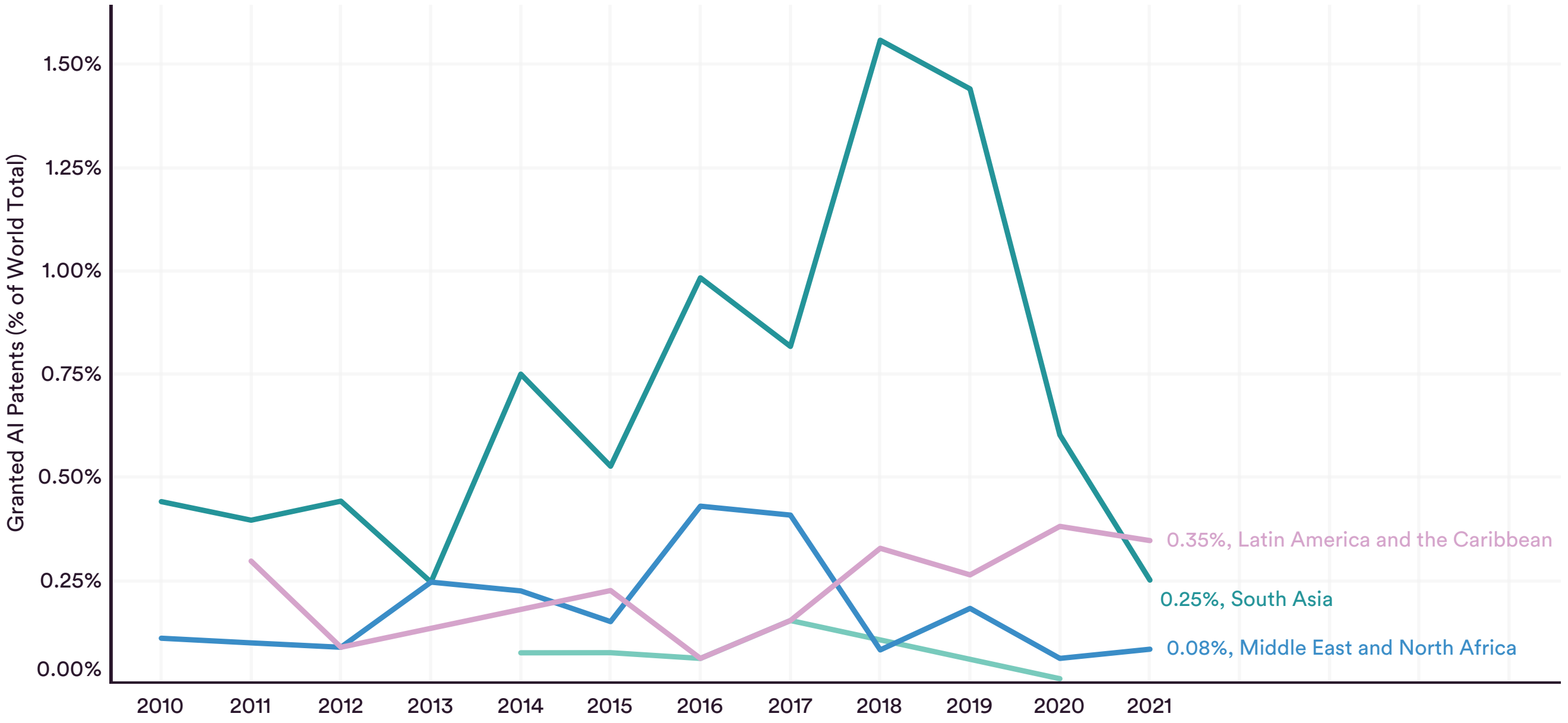

Figure 1.1.23c

## By Geographic Area and Application Status

Trends revealed by the regional analysis can also be observed in AI patent data broken down by geographic area (Figure 1.1.24a and Figure 1.1.24b). China is now filing over half of the world's AI patents and being granted about 6%, about the same as the European Union plus the United Kingdom. The United States, which files almost all the patents in North America, does so at one-third the rate of China. Figure 1.1.24c shows that compared to the increasing numbers of AI patents applied and granted, China has far greater numbers of patent applications (87,343 in 2021) than those granted (1,407 in 2021).

### AI PATENT FILINGS (% of WORLD TOTAL) by GEOGRAPHIC AREA, 2010–21

Source: Center for Security and Emerging Technology, 2021 | Chart: 2022 AI Index Report

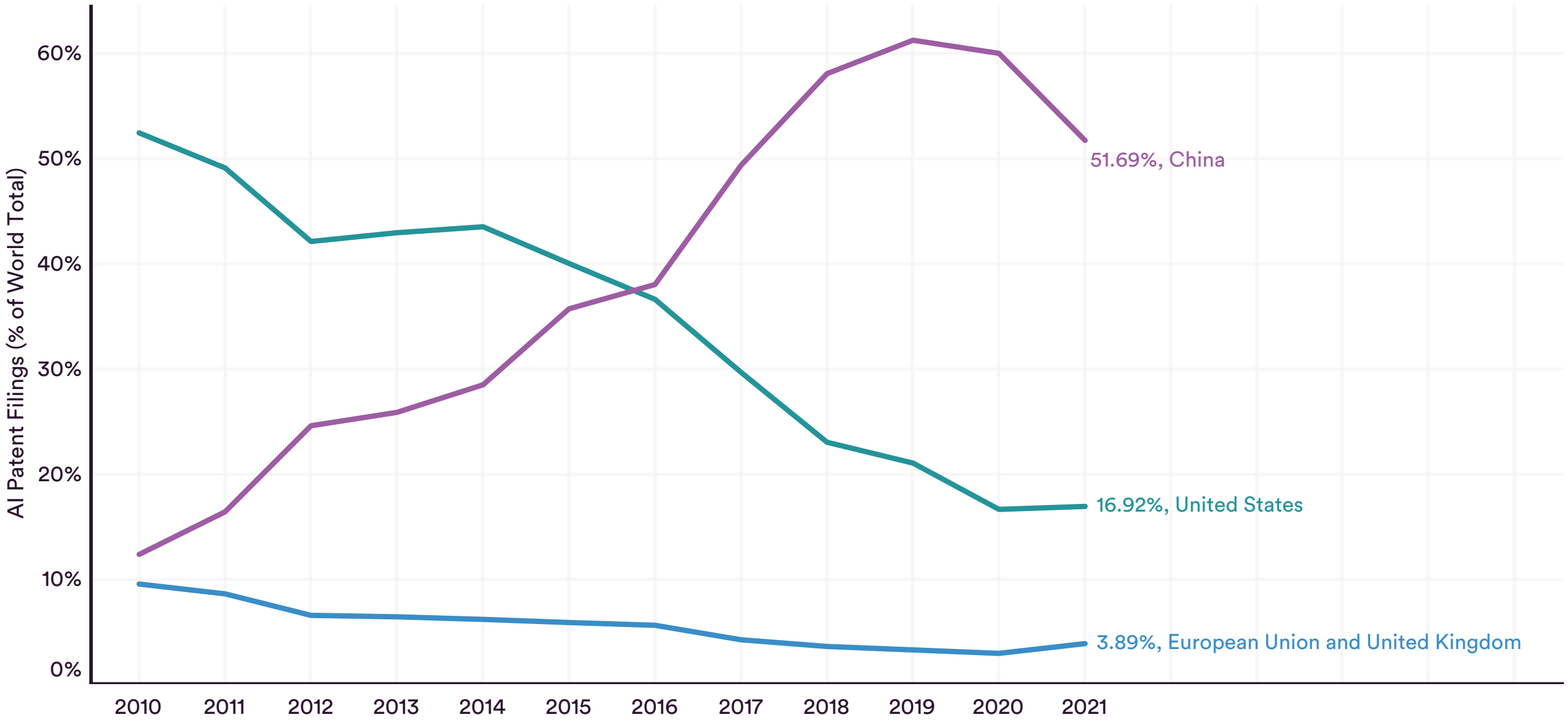

Figure 1.1.24a

### GRANTED AI PATENTS (% of WORLD TOTAL) by GEOGRAPHIC AREA, 2010–21

Source: Center for Security and Emerging Technology, 2021 | Chart: 2022 AI Index Report

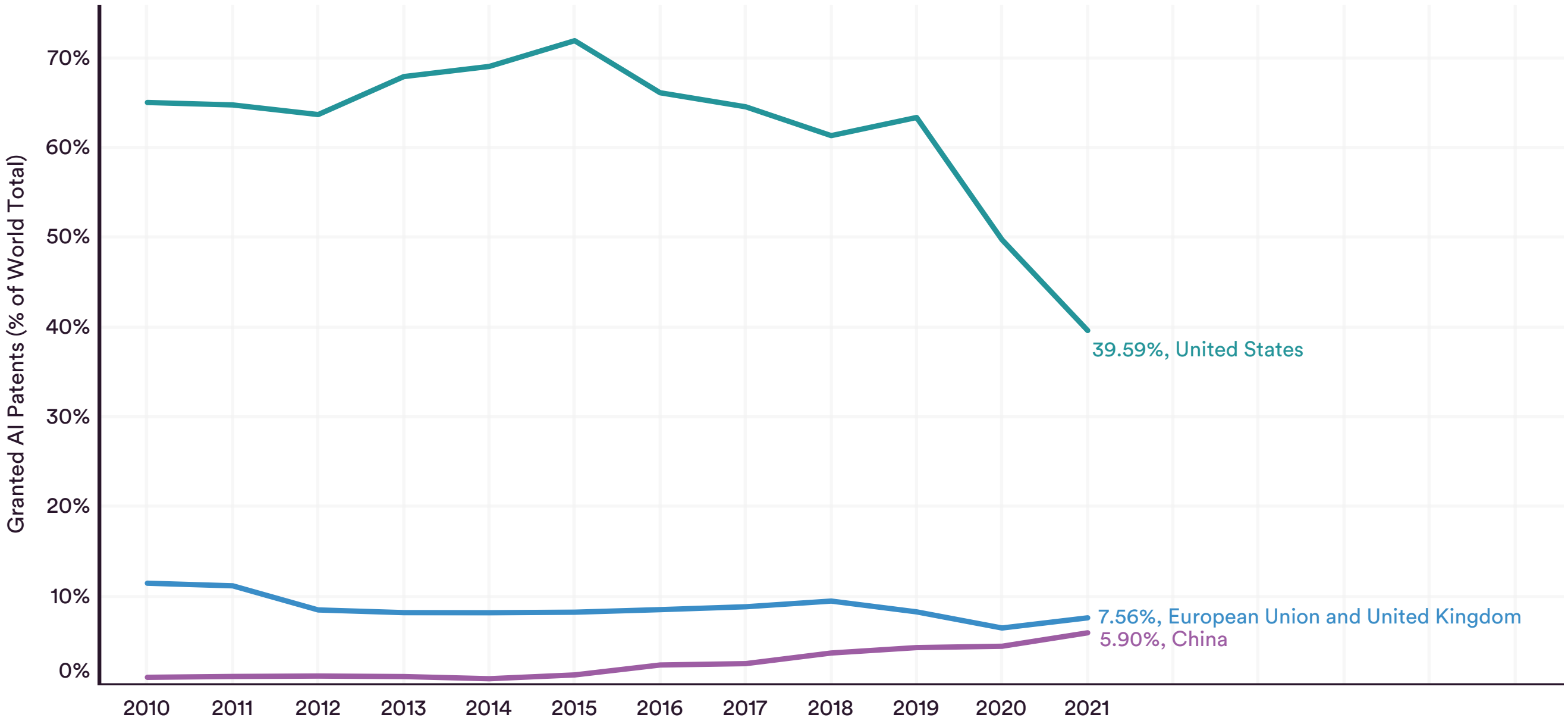

Figure 1.1.24b

## AI PATENTS by APPLICATION STATUS by GEOGRAPHIC AREA, 2010–21

Source: Center for Security and Emerging Technology, 2021 | Chart: 2022 AI Index Report

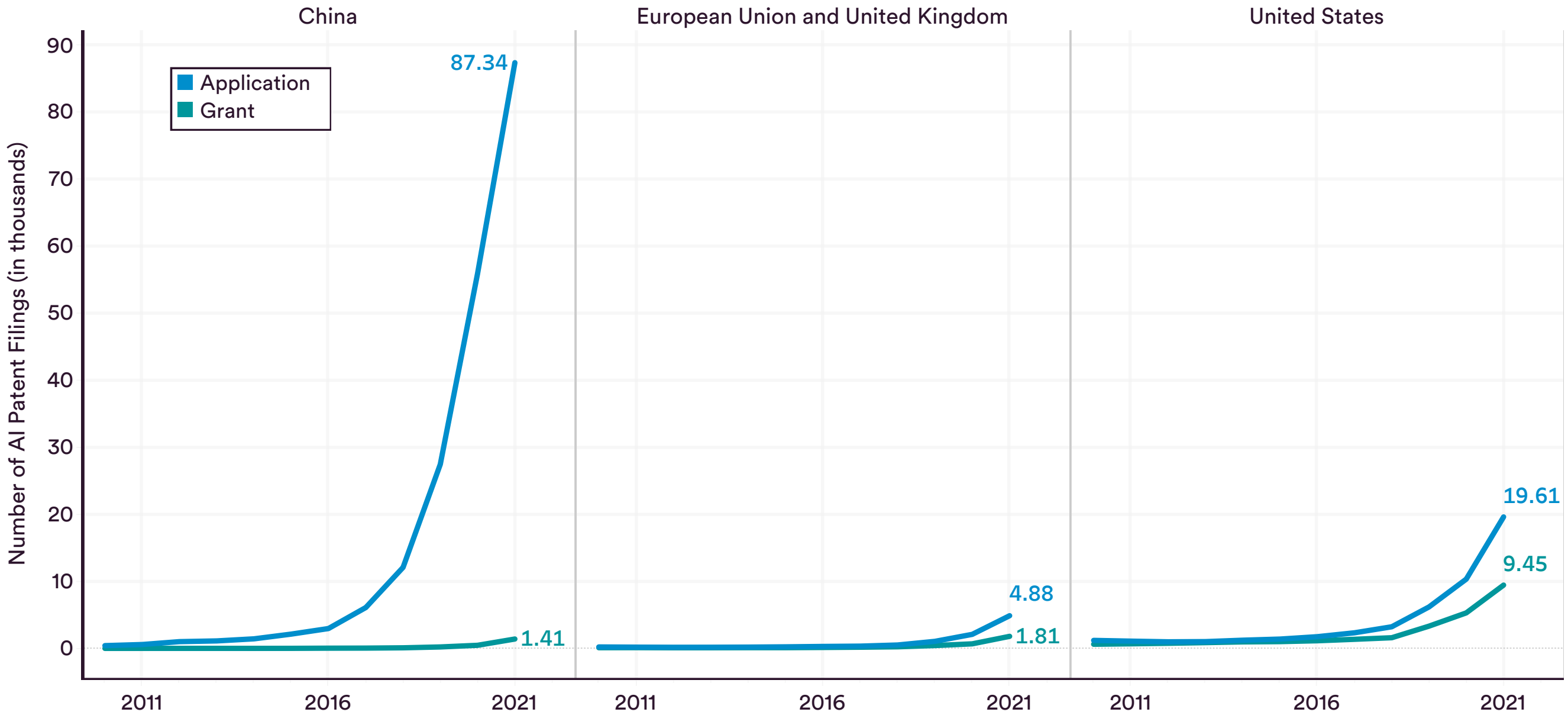

Figure 1.1.24c

**AI conferences are key venues for researchers to publish and communicate their work, as well as to connect with peers and collaborators. Conference attendance is an indication of broader industrial and academic interest in a scientific field. In the past 20 years, AI conferences have grown not only in size but also in number and prestige. This section presents data on the trends in attendance at major AI conferences, covering more conferences (16) than previous Index reports.**

# 1.2 CONFERENCES

## CONFERENCE ATTENDANCE

Similar to 2020, most AI conferences were offered virtually in 2021. Only the International Conference on Robotics and Automation (ICRA) and the Conference on Empirical Methods in Natural Language Processing (EMNLP) were held using a hybrid format. Conference organizers reported measuring the exact attendance numbers at a virtual conference is difficult, as virtual conferences allow for higher attendance of researchers from all around the world.

Figure 1.2.1 shows that attendance at top AI conferences in 2021 was relatively consistent with 2020, with more than 88,000 participants worldwide. Figure 1.2.2 and Figure 1.2.3 show the attendance data for individual conferences, with 16 major AI conferences split into two categories: large AI conferences with over 2,500 attendees and small AI conferences with fewer than 2,500 attendees.[6]

**NUMBER of ATTENDEES at SELECT AI CONFERENCES, 2010–21**

Source: Conference Data, 2021 | Chart: 2022 AI Index Report

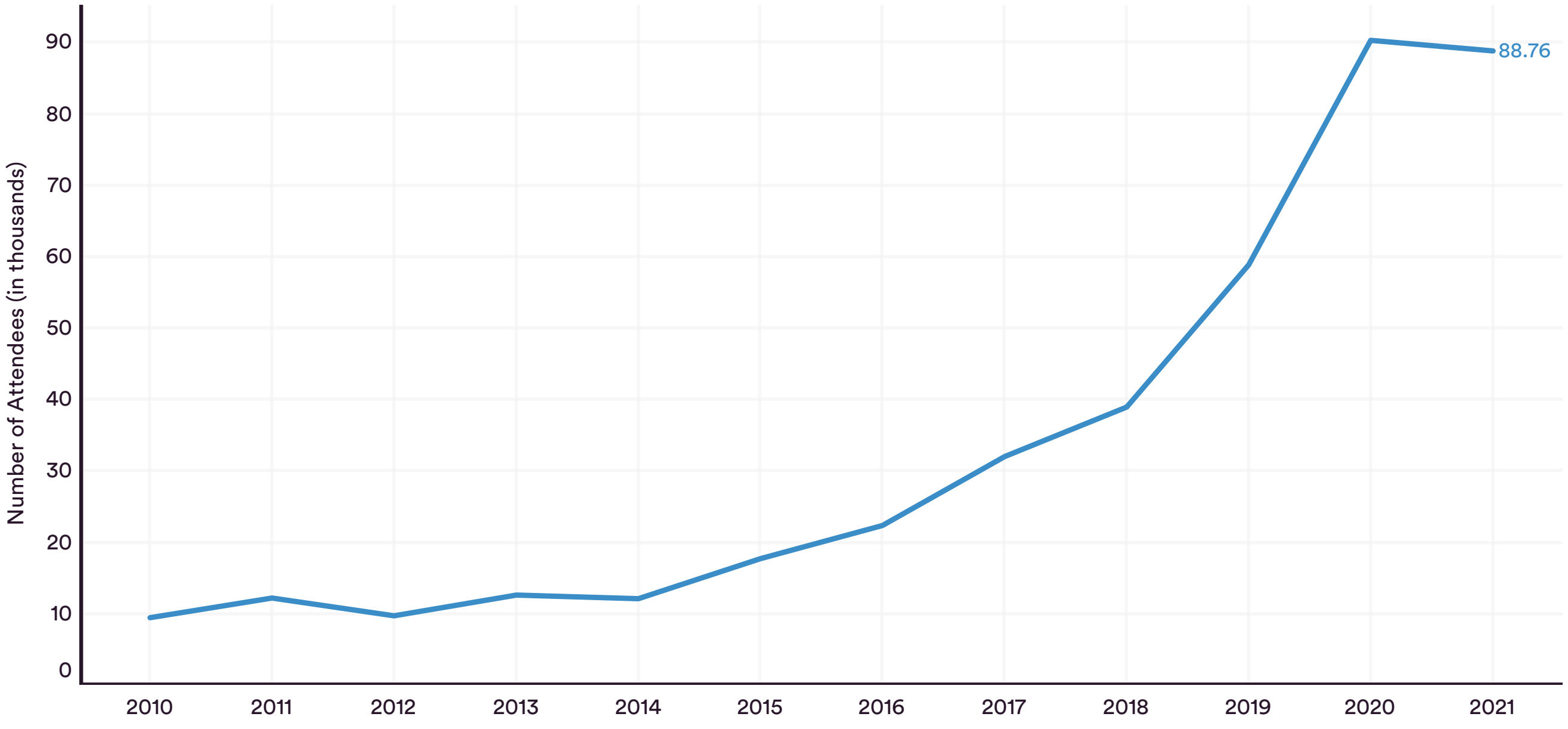

Figure 1.2.1

6 The International Conference on Machine Learning (ICML) used the number of session visitors as a proxy for the number of conference attendees, which explains the high attendance count in 2021. The International Conference on Intelligent Robots and Systems (IROS) extended the virtual conference to allow users to watch events for up to three months, which explains the high attendance count in 2020. For the AAMAS conference, the attendance in 2020 is based on the number of users on site reported by the platform that recorded the talks and managed the online conference, while the 2021 number is for total registrants.

## ATTENDANCE at LARGE AI CONFERENCES, 2010–21

Source: Conference Data, 2021 | Chart: 2022 AI Index Report

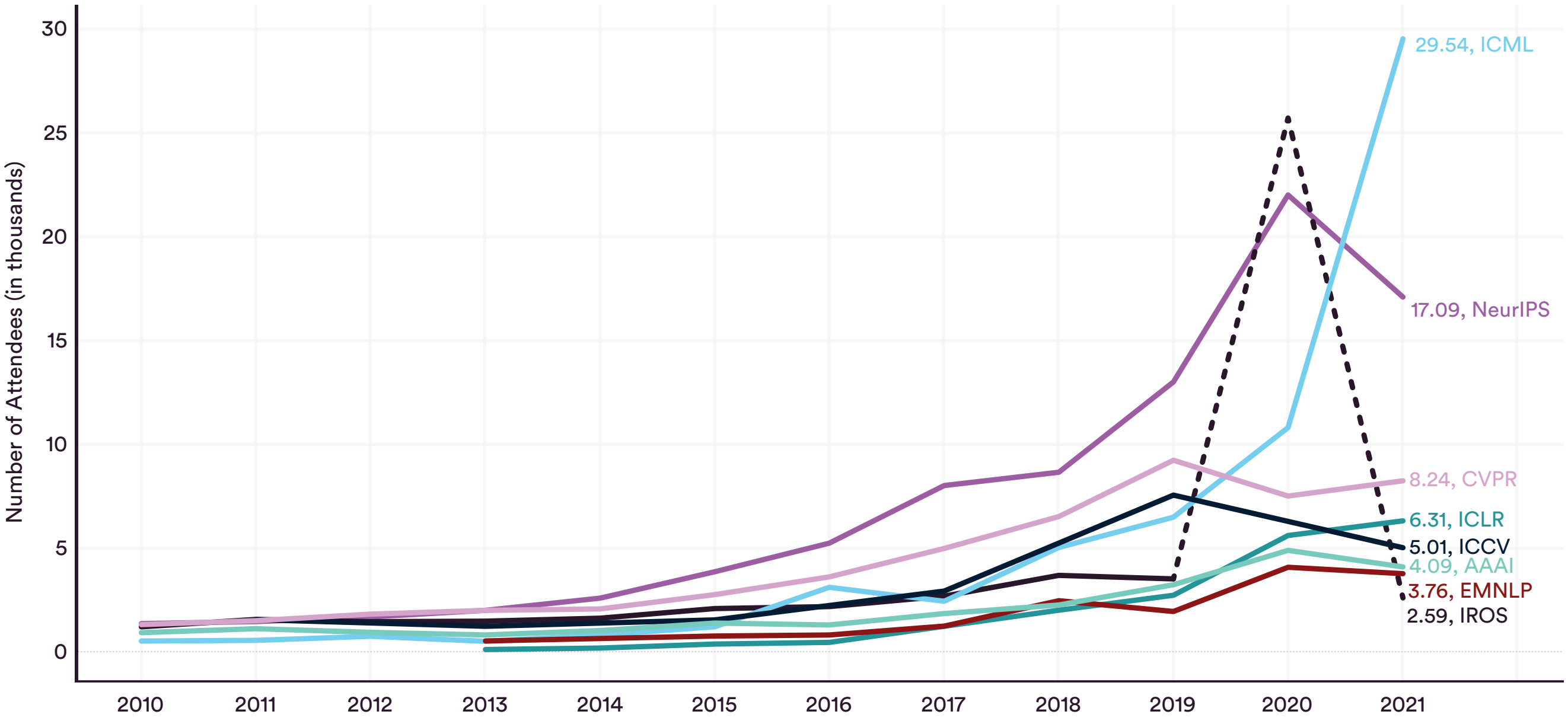

Figure 1.2.2

## ATTENDANCE at SMALL AI CONFERENCES, 2010–21

Source: Conference Data, 2021 | Chart: 2022 AI Index Report

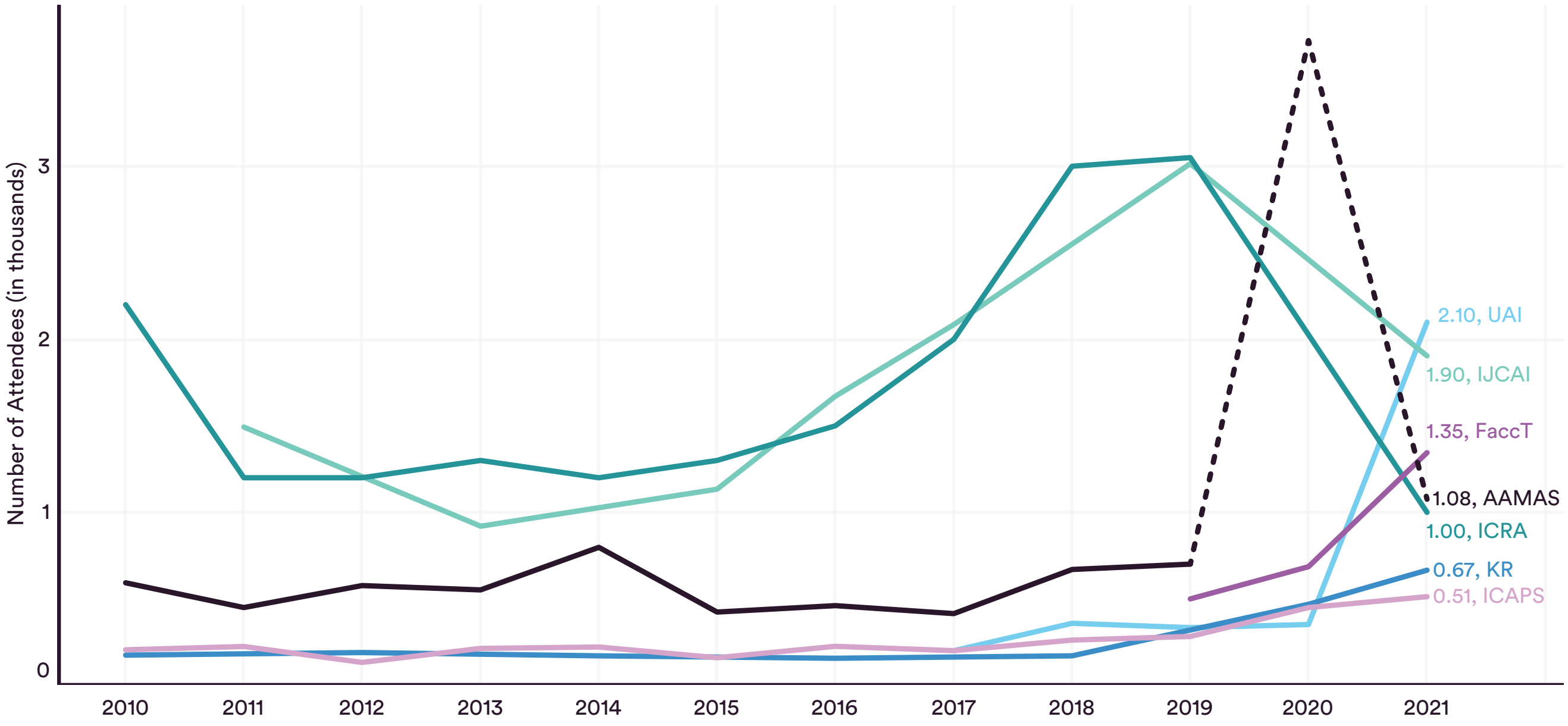

Figure 1.2.3

## WOMEN IN MACHINE LEARNING (WIML) NEURIPS WORKSHOP

Founded in 2006, Women in Machine Learning is an organization dedicated to supporting and increasing the impact of women in machine learning. This section presents data from its annual technical workshop colocated with NeurIPS. Starting in 2020, WiML has also been hosting the Un-Workshop, which aims to advance research via collaboration and interaction among participants from diverse backgrounds at ICML.

### Workshop Participants

The number of participants attending the WiML workshop has steadily increased since it was first introduced in 2006. For the 2021 edition, Figure 1.2.4 shows an estimate of 1,486 attendees over all workshop sessions, counted as the number of unique individuals who accessed the virtual workshop platform at neurips.cc. The 2021 WiML Workshop at NeurIPS happened as multiple sessions over three days, which was a change in format from 2020. As in 2020, the workshop was held virtually due to the pandemic.

ATTENDANCE at NEURIPS WOMEN in MACHINE LEARNING WORKSHOP, 2010–21

Source: Women in Machine Learning, 2021 | Chart: 2022 AI Index Report

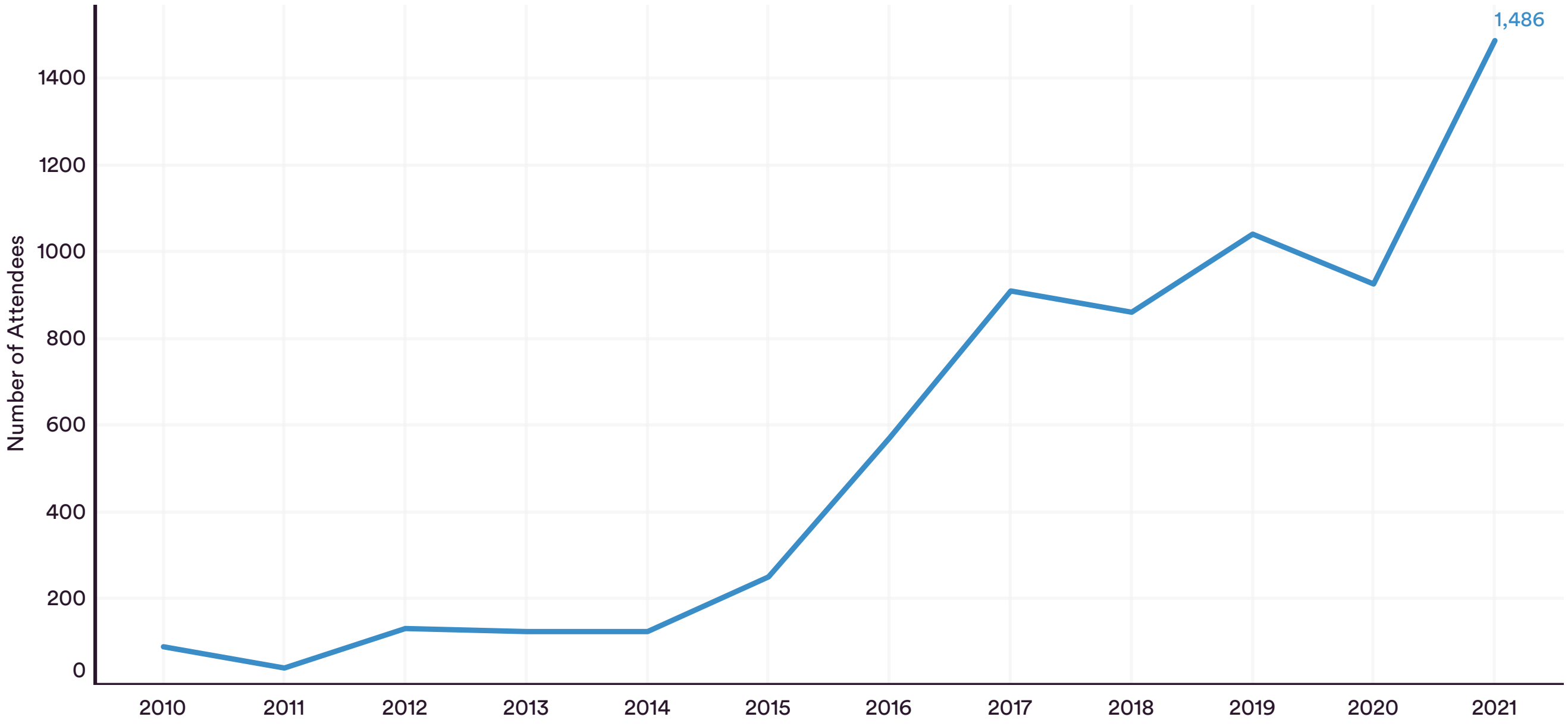

Figure 1.2.4

## Demographics Breakdown

This section shows the continent of residence and professional position breakdowns of the 2021 workshop participants based on a survey filled by participants who consented to have such information aggregated. Among the survey respondents, more than half of the survey respondents were from North America, followed by Europe (19.9%), Asia (16.2%), and Africa (7.3%) (Figure 1.2.5). Figure 1.2.6 shows that Ph.D. students made up almost half of the survey participants, while the share of university faculty is around 1.2%. Researcher scientists/engineers, data scientists/engineers, and software engineers were among the most commonly held professional positions.

**CONTINENT of RESIDENCE of PARTICIPANTS at NEURIPS WOMEN in MACHINE LEARNING WORKSHOP, 2021**

Source: Women in Machine Learning, 2021 | Chart: 2022 AI Index Report

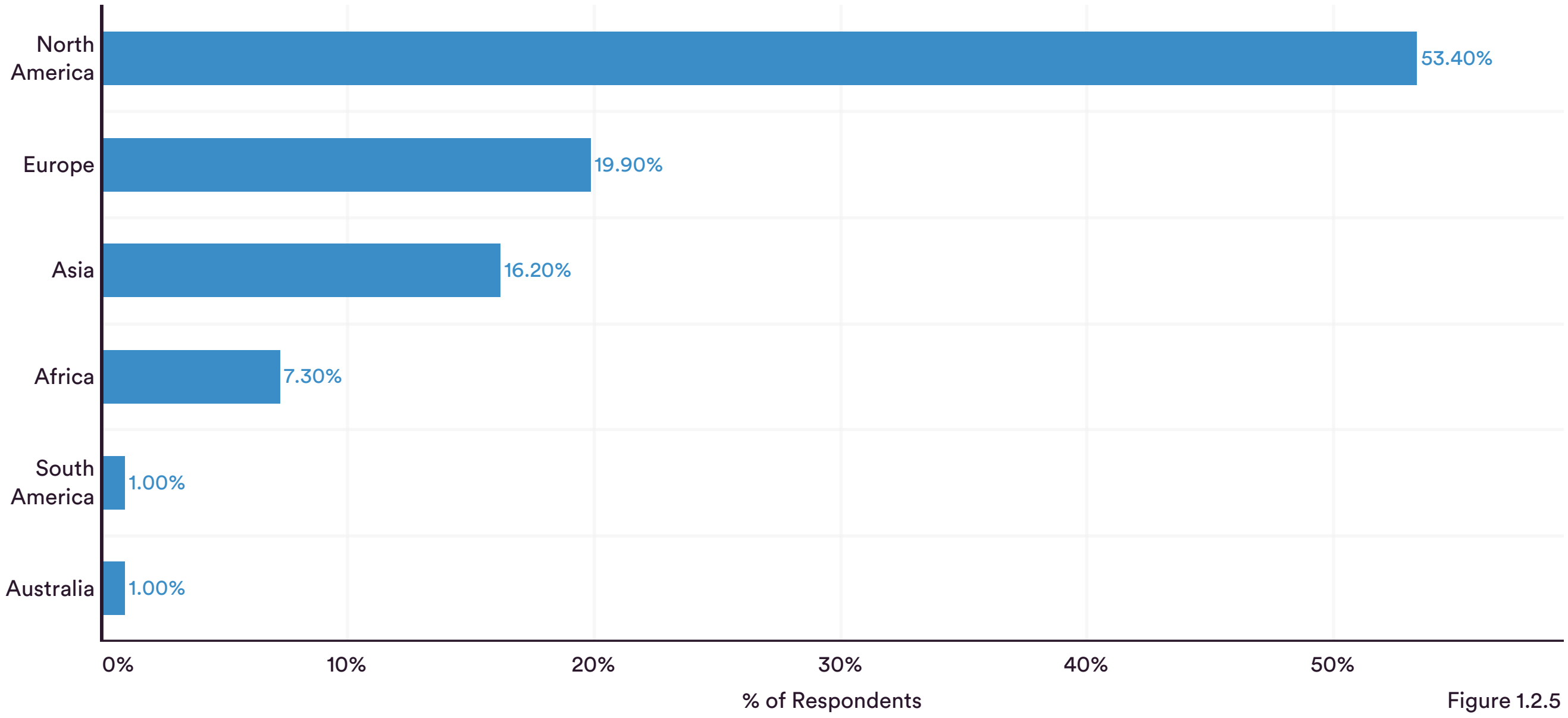

Figure 1.2.5

**PROFESSIONAL POSITIONS of PARTICIPANTS at NEURIPS WOMEN in MACHINE LEARNING WORKSHOP, 2021**

Source: Women in Machine Learning, 2021 | Chart: 2022 AI Index Report

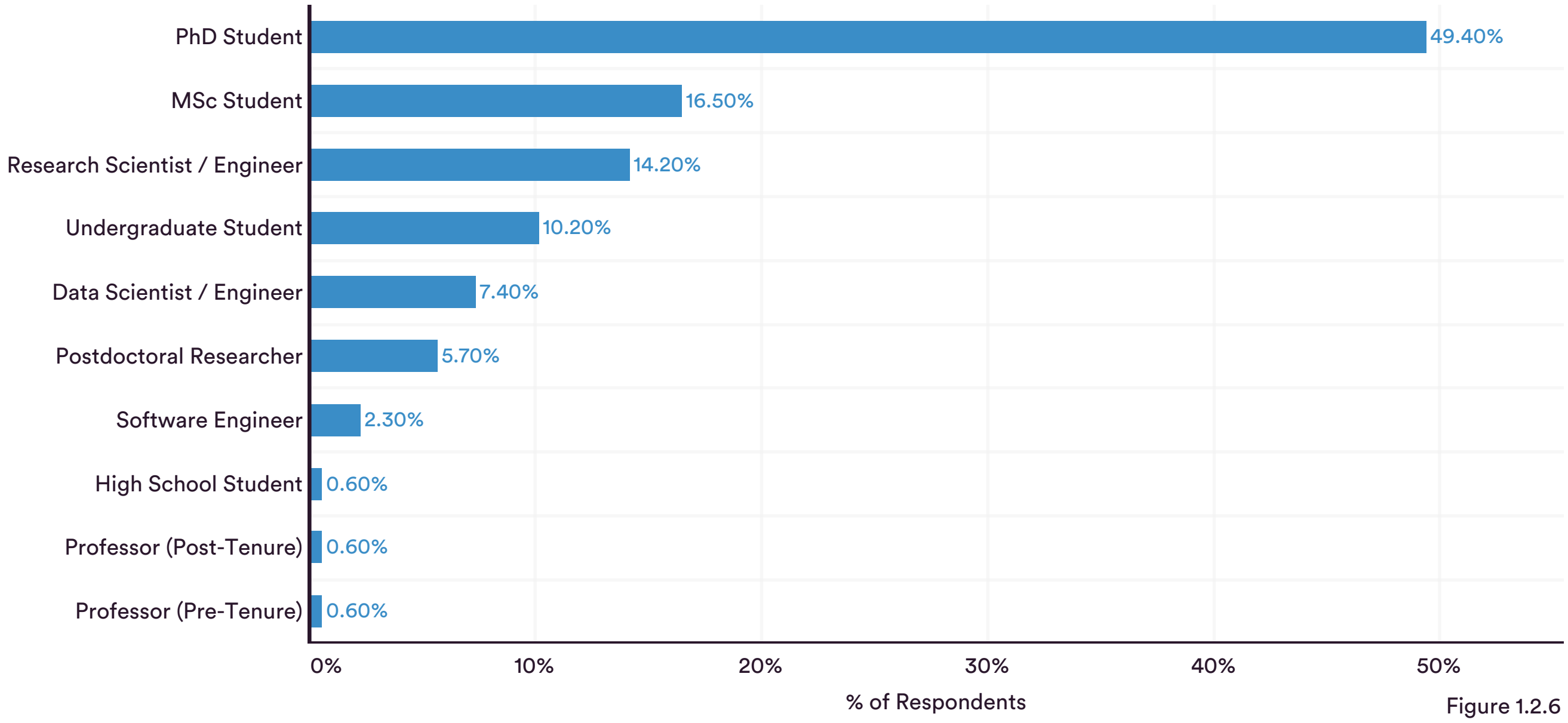

Figure 1.2.6

A software library is a collection of computer code that is used to create applications and products. Popular AI-specific software libraries—such as TensorFlow and PyTorch—help developers create their AI solutions quickly and efficiently. This section analyzes the popularity of software libraries through GitHub data.

# 1.3 AI OPEN-SOURCE SOFTWARE LIBRARIES

## GITHUB STARS

Figures 1.3.1 and 1.3.2 reflect the number of users of GitHub open-source AI software libraries from 2015 to 2021. TensorFlow remained by far the most popular in 2021, with around 161,000 cumulative GitHub stars—a slight increase over 2020. TensorFlow was about three times as popular in 2021 as the next-most-starred GitHub open-source AI software library, OpenCV, which was followed by Keras, PyTorch, and Scikit-learn. Figure 1.3.2 shows library popularity for libraries with fewer than 40,000 GitHub stars—led by FaceSwap with around 40,000 stars, followed by 100-Days-Of-ML-Code, AiLearning, and BVLC/caffe.

**NUMBER of GITHUB STARS by AI LIBRARY (OVER 40K STARS), 2014–21**
Source: GitHub, 2021 | Chart: 2022 AI Index Report

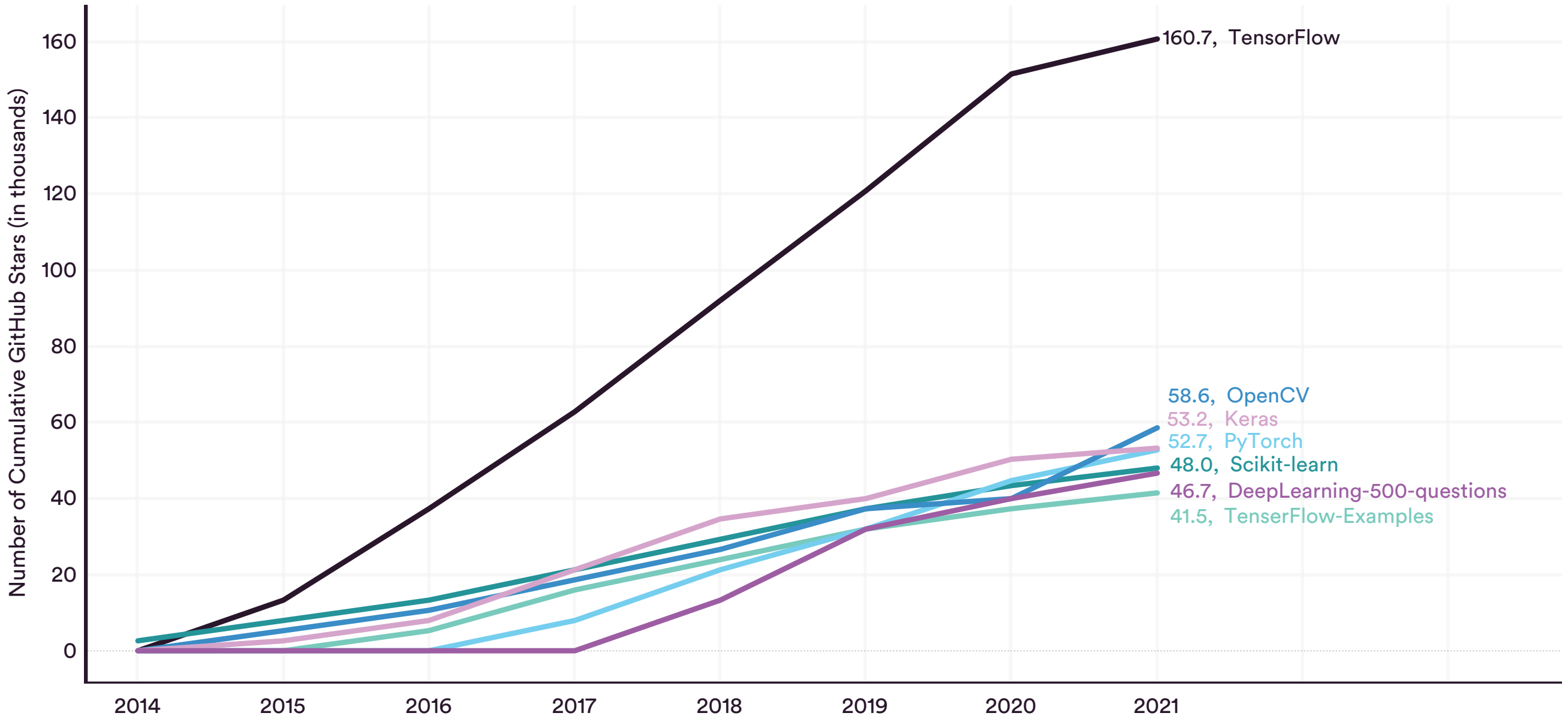

Figure 1.3.1

## NUMBER of GITHUB STARS by AI LIBRARY (UNDER 40K STARS), 2014–21

Source: GitHub, 2021 | Chart: 2022 AI Index Report

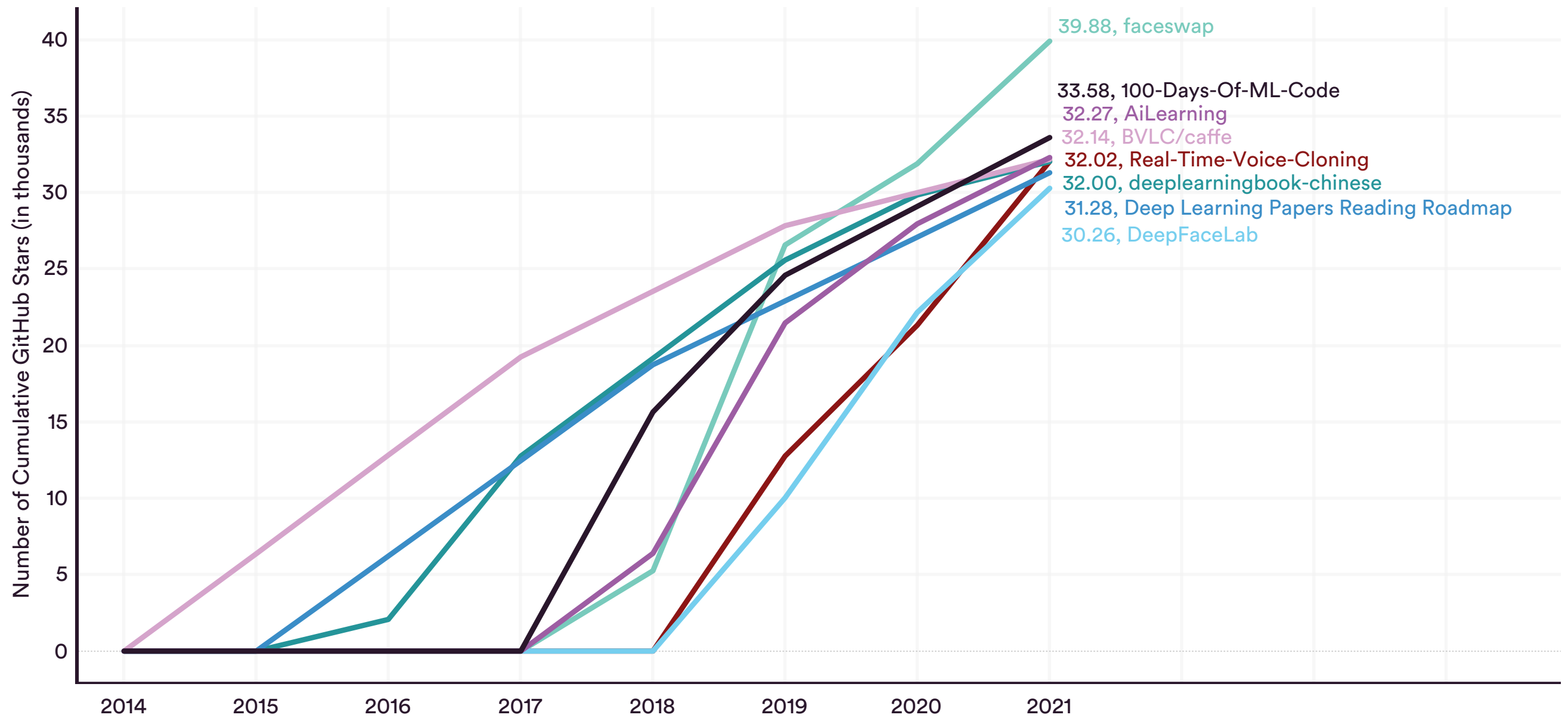

Figure 1.3.2

Artificial Intelligence
Index Report 2022

# CHAPTER 2:
# Technical Performance

## CHAPTER 2:

# Chapter Preview

**ACCESS THE PUBLIC DATA**

# CHAPTER 2: CHAPTER PREVIEW (CONT'D)

ACCESS THE PUBLIC DATA

# Overview

This year, the technical performance chapter includes more analysis than ever before of the technical progress in various subfields of artificial intelligence, including trends in computer vision, language, speech, recommendation, reinforcement learning, hardware, and robotics. It uses a number of quantitative measurements, from common AI benchmarks and prize challenges to a field-wide survey, to highlight the development of top-performing AI systems.

# CHAPTER HIGHLIGHTS

- **Data, data, data:** Top results across technical benchmarks have increasingly relied on the use of extra training data to set new state-of-the-art results. **As of 2021, 9 state-of-the-art AI systems out of the 10 benchmarks in this report are trained with extra data.** This trend implicitly favors private sector actors with access to vast datasets.

- **Rising interest in particular computer vision subtasks:** In 2021, the research community saw a greater level of interest in more specific computer vision subtasks, such as medical image segmentation and masked-face identification. **For example, only 3 research papers tested systems against the Kvasir-SEG medical imaging benchmark prior to 2020. In 2021, 25 research papers did.** Such an increase suggests that AI research is moving toward research that can have more direct, real-world applications.

- **AI has not mastered complex language tasks, *yet:*** AI already exceeds human performance levels on basic reading comprehension benchmarks like SuperGLUE and SQuAD by 1%–5%. **Although AI systems are still unable to achieve human performance on more complex linguistic tasks such as abductive natural language inference (aNLI), the difference is narrowing. Humans performed 9 percentage points better on aNLI in 2019. As of 2021, that gap has shrunk to 1.**

- **Turn toward more general reinforcement learning:** For the last decade, AI systems have been able to master narrow reinforcement learning tasks in which they are asked to maximize performance in a specific skill, such as chess. **The top chess software engine now exceeds Magnus Carlsen's top ELO score by 24%. However, in the last two years AI systems have also improved by 129% on more general reinforcement learning tasks (Procgen) in which they must operate in novel environments.** This trend speaks to the future development of AI systems that can learn to think more broadly.

- **AI becomes more affordable *and* higher performing: Since 2018, the cost to train an image classification system has decreased by 63.6%, while training times have improved by 94.4%.** The trend of lower training cost but faster training time appears across other MLPerf task categories such as recommendation, object detection and language processing, and favors the more widespread commercial adoption of AI technologies.

- **Robotic arms are becoming cheaper:** An AI Index survey shows that **the median price of robotic arms has decreased by 46.2% in the past five years—from $42,000 per arm in 2017 to $22,600 in 2021.** Robotics research has become more accessible and affordable.

**Computer vision is the subfield of AI that teaches machines to understand images and videos. There is a wide range of computer vision tasks, such as image classification, object recognition, semantic segmentation, and face detection. As of 2021, computers can outperform humans on a plethora of computer vision tasks. Computer vision technologies have a variety of important real-world applications, such as autonomous driving, crowd surveillance, sports analytics, and video-game creation.**

# 2.1 COMPUTER VISION—IMAGE

## IMAGE CLASSIFICATION

Image classification refers to the ability of machines to categorize what they see in images (Figure 2.1.1). In a practical sense, image recognition systems can help cars identify objects in their surroundings, doctors detect tumors, and factory managers spot production defects. The past decade has seen tremendous advances in the technical capacity of image recognition systems, especially as researchers have embraced more machine learning techniques. Moreover, progress in algorithmic, hardware, and data technologies has meant that image recognition has become more affordable, widely applicable, and accessible than ever before.

**A DEMONSTRATION OF IMAGE CLASSIFICATION**
Source: Krizhevsky, 2020

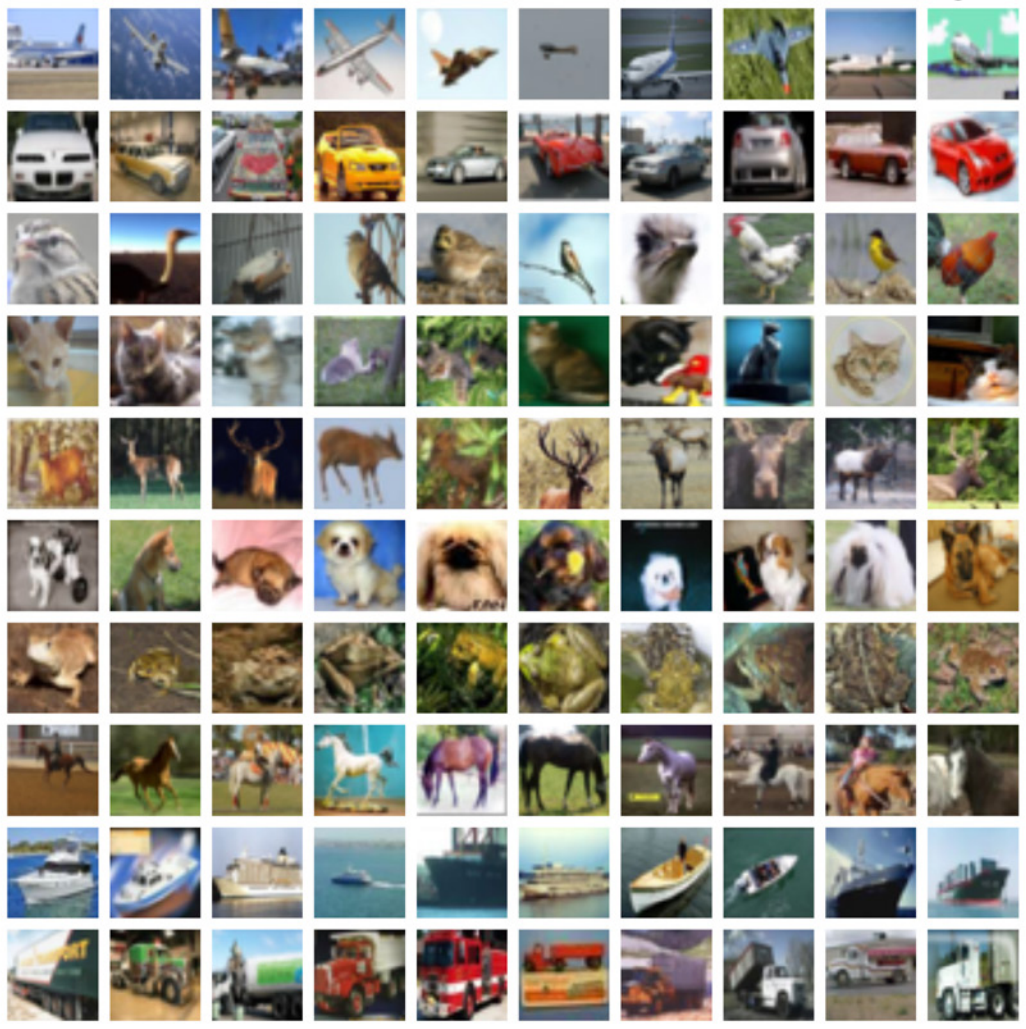

Figure 2.1.1

### ImageNet

ImageNet is a database that includes over 14 million images across 20,000 categories publicly available to researchers working on image classification problems. Created in 2009, ImageNet is now one of the most common ways scientists benchmark improvement on image classification.

### ImageNet: Top-1 Accuracy

Benchmarking on ImageNet is measured through accuracy metrics, which quantify how frequently AI systems assign the right labels to the given images. Top-1 accuracy measures the rate at which the top prediction made by a classification model for a given image matches the image's actual target label. In recent years, it has become increasingly common to improve the performance of systems on ImageNet by pretraining them with additional data from other image datasets.

As of late 2021, the top image classification system makes on average 1 error for every 10 classification attempts on Top-1 accuracy compared to an average of 4 errors for every 10 attempts in late 2012 (Figure 2.1.2). In 2021, the top pretrained system was CoAtNets, produced by researchers on the Google Brain Team.

### ImageNet: Top-5 Accuracy

Top-5 accuracy considers whether any of the model's 5 highest probability answers align with the image label. As highlighted in Figure 2.1.3, AI systems presently achieve near perfect Top-5 estimation. Currently, the state-of-the-art performance on Top-5 accuracy with pretraining is 99.0%, achieved in November 2021 by Microsoft Cloud and Microsoft AI's Florence-CoSwim-H model.

Improvements in Top-5 accuracy on ImageNet seem to be plateauing, which is perhaps unsurprising. If your system is classifying correctly 98 or 99 out of 100 times, there is only so much higher you can go.

## IMAGENET CHALLENGE: TOP-1 ACCURACY

Source: Papers with Code, 2021; arXiv, 2021 | Chart: 2022 AI Index Report

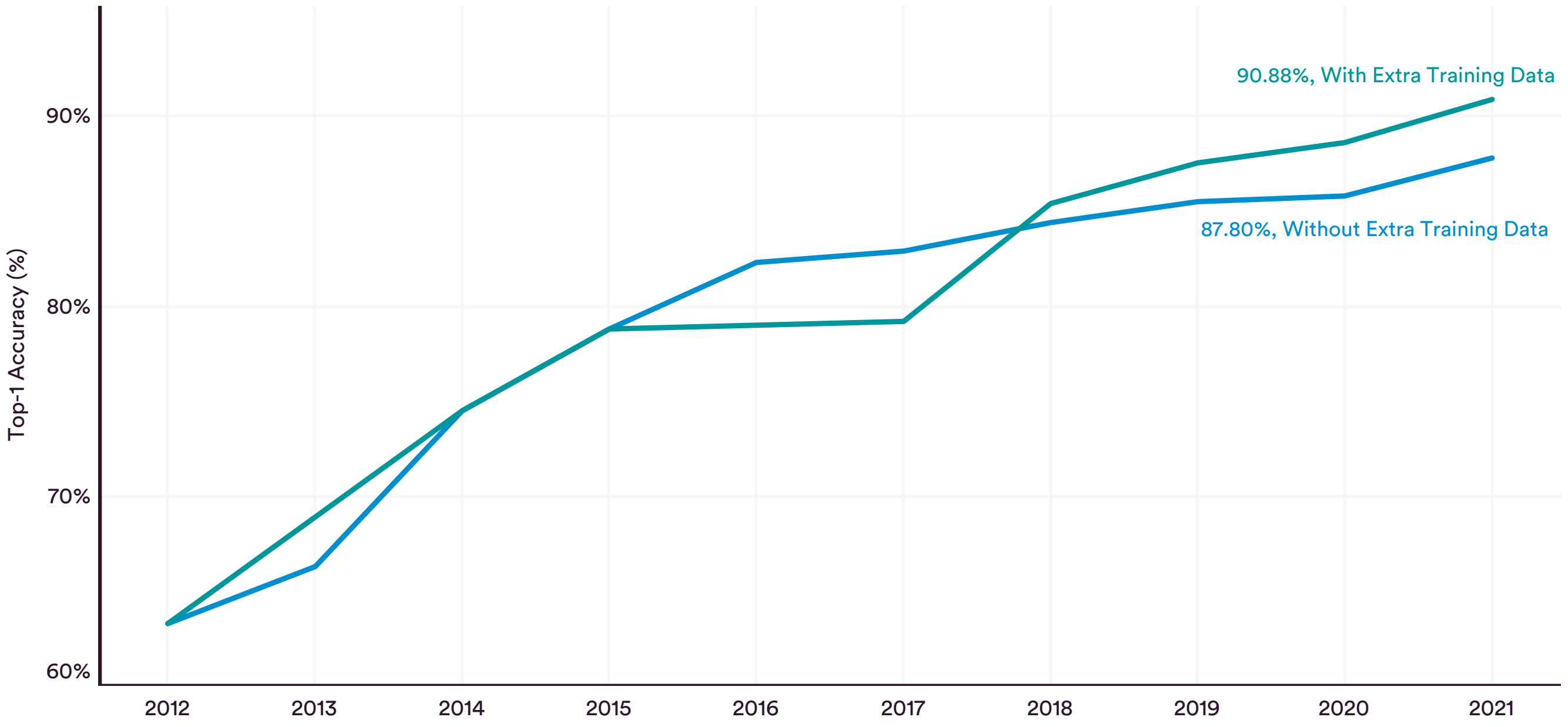

Figure 2.1.2

## IMAGENET CHALLENGE: TOP-5 ACCURACY

Source: Papers with Code, 2021; arXiv, 2021 | Chart: 2022 AI Index Report

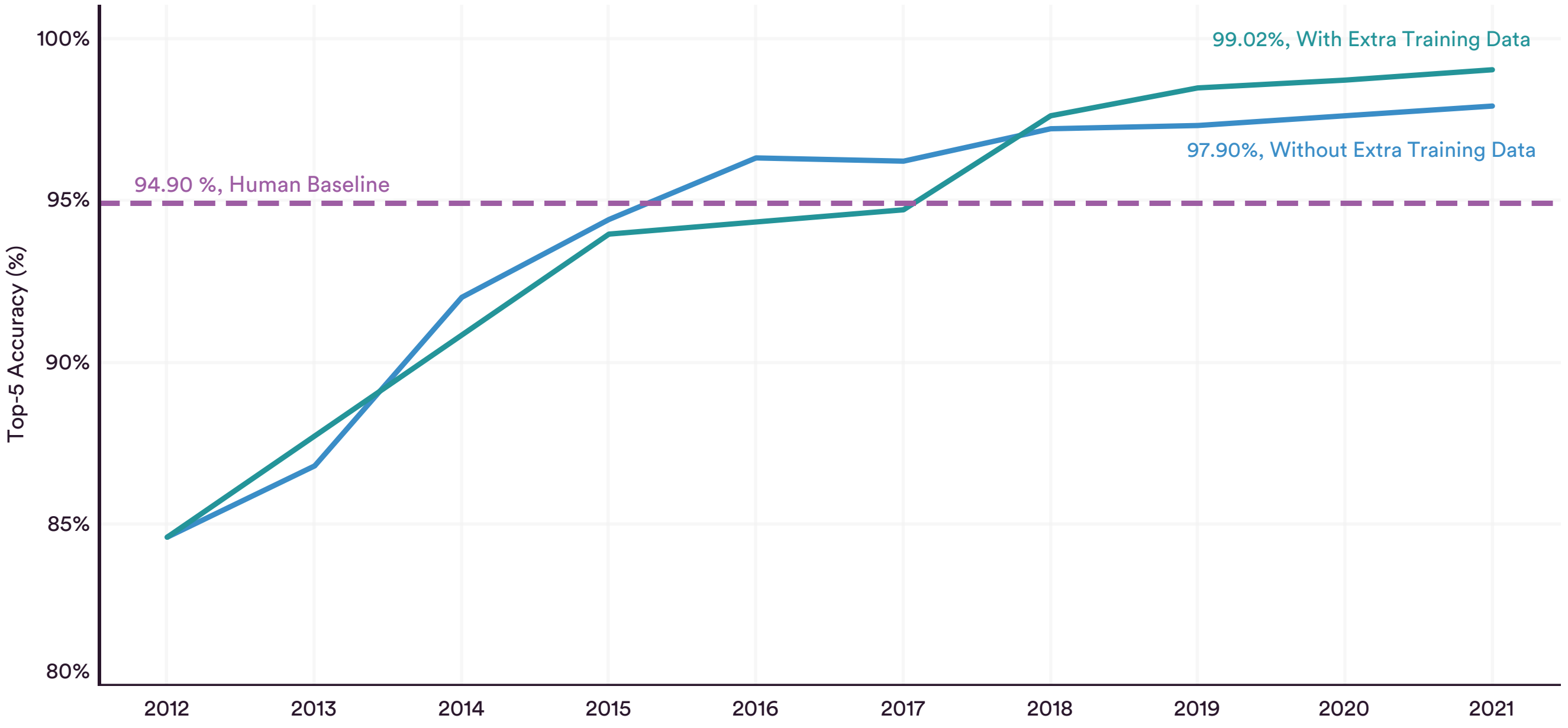

Figure 2.1.3

## IMAGE GENERATION

Image generation is the task of generating images that are indistinguishable from real ones. Image generation can be widely useful in generative domains where visual content has to be created, for example entertainment (companies like NVIDIA have already used image generators to create virtual worlds for gaming), fashion (designers can let AI systems generate different design patterns), and healthcare (image generators can synthetically create novel drug compounds). Figure 2.1.4 illustrates progress made in image generation by presenting several human faces that were synthetically generated by AI systems in the last year.

### GAN PROGRESS ON FACE GENERATION

Source: Goodfellow et al., 2014; Radford et al., 2016; Liu & Tuzel, 2016; Karras et al., 2018; Karras et al., 2019; Goodfellow, 2019; Karras et al., 2020; AI Index, 2021; Vahdat et al., 2021

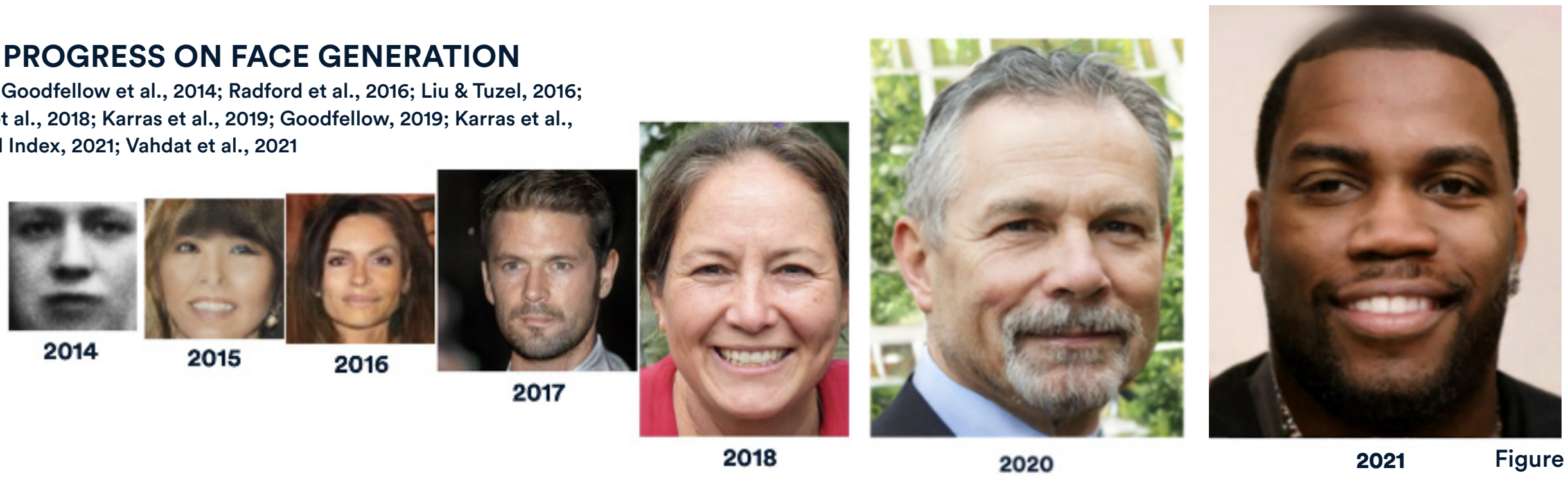

Figure 2.1.4

### STL-10: Fréchet Inception Distance (FID) Score

The Fréchet Inception Distance score tracks the similarity between an artificially generated set of images and the real images from which it was generated. A low score means that the generated images are more similar to the real ones, and a score of zero indicates that the fake images are identical to the real ones.

Figure 2.1.5 documents the gains generative models have made in FID on the STL-10 dataset, one of the most widely cited datasets in computer vision. The state-of-the-art model on STL-10 developed by researchers at the Korea Advanced Institute of Science and Technology as well as the University of Seoul posted a FID score of 7.7, significantly better than the state-of-the-art result from 2020.

### STL-10: FRÉCHET INCEPTION DISTANCE (FID) SCORE

Source: Papers with Code, 2021; arXiv, 2021 | Chart: 2022 AI Index Report

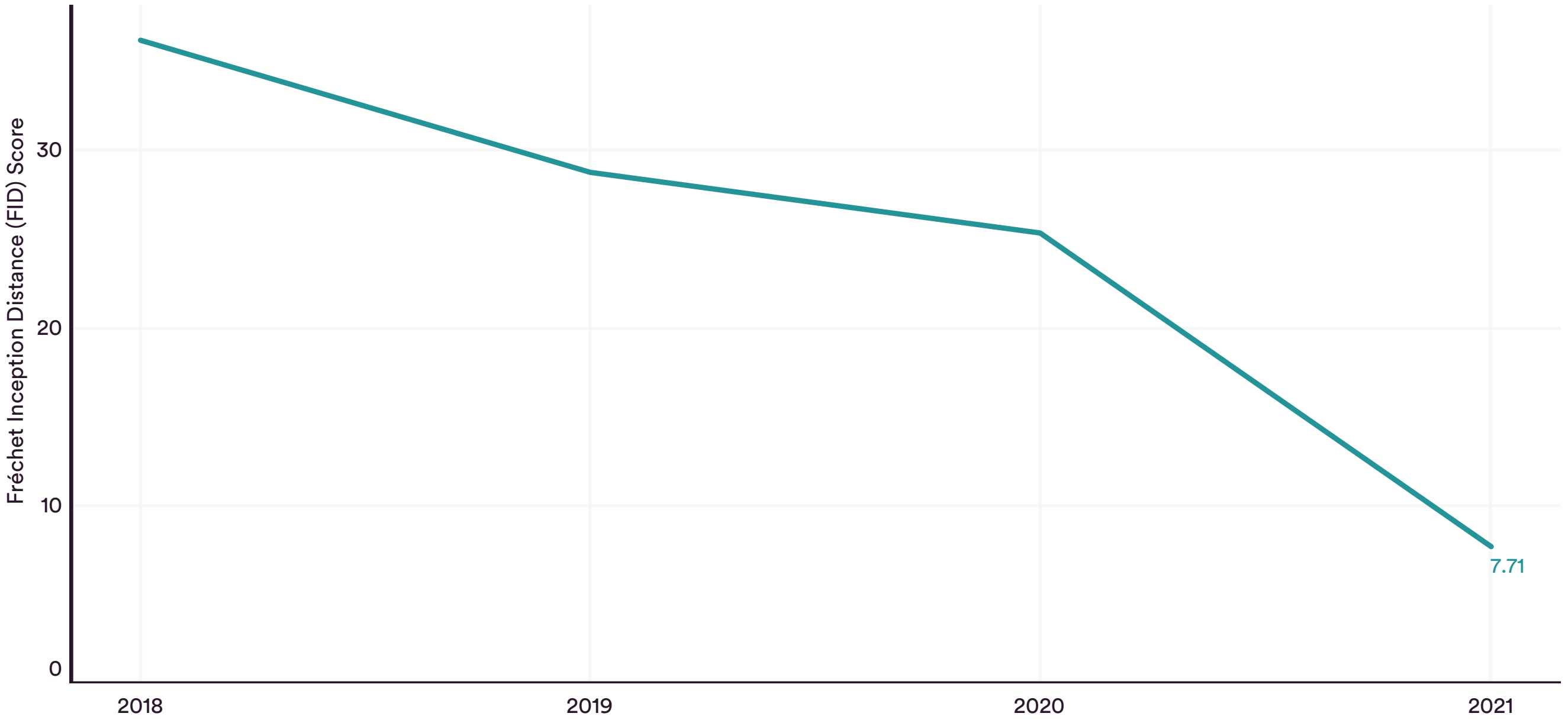

Figure 2.1.5

### CIFAR-10: Fréchet Inception Distance (FID) Score

Progress on image generation can also be benchmarked on CIFAR-10, a dataset of 60,000 color images across 10 different object classes. The state-of-the-art results on CIFAR-10 posted in 2021 were achieved by researchers from NVIDIA.

The FID scores achieved by the top image generation models are much lower on CIFAR-10 than STL-10. This difference is likely attributable to the fact that CIFAR-10 contains images of much lower resolution (32 x 32 pixels) than those on STL-10 (96 x 96 pixels).

**CIFAR-10: FRÉCHET INCEPTION DISTANCE (FID) SCORE**
Source: Papers with Code, 2021; arXiv, 2021 | Chart: 2022 AI Index Report

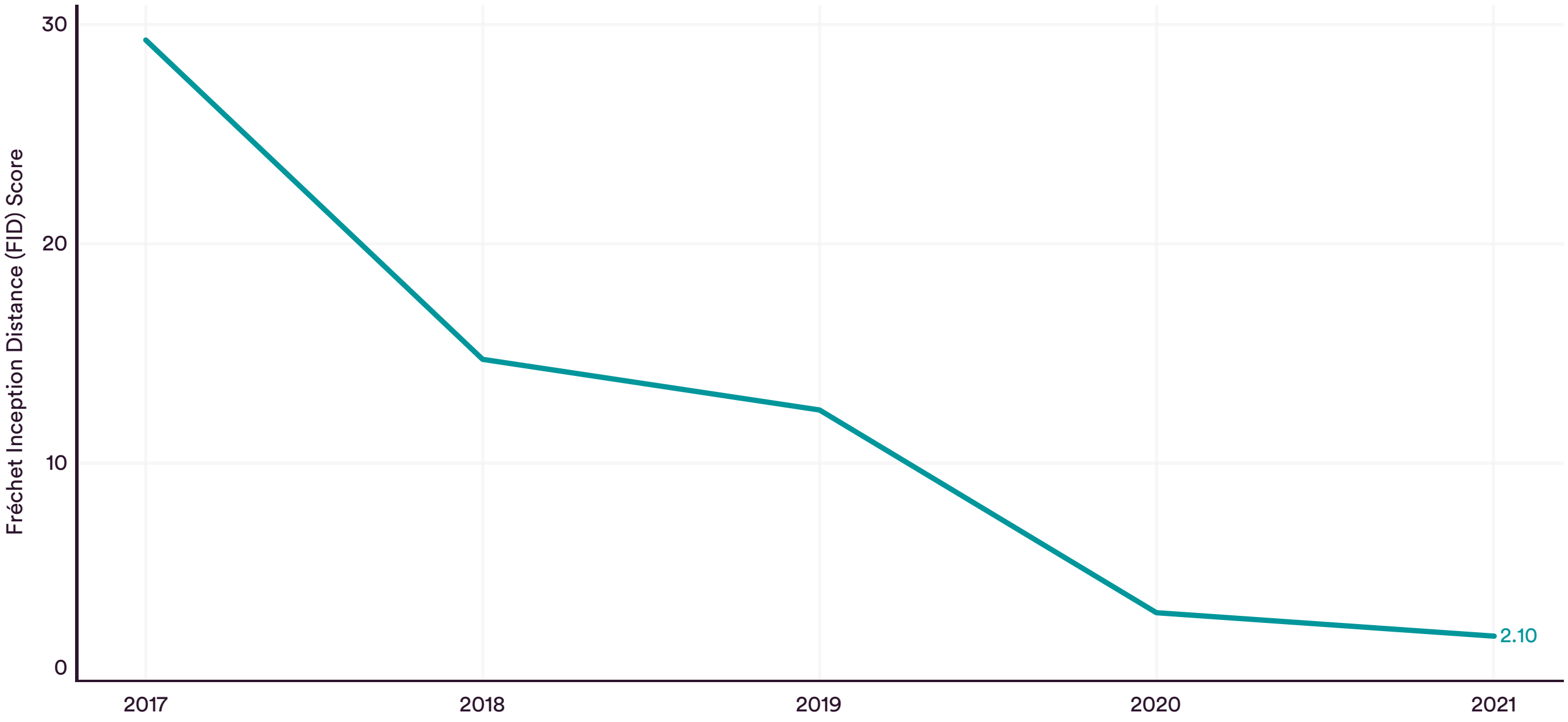

Figure 2.1.6

## DEEPFAKE DETECTION

Many AI systems can now generate fake images that are indistinguishable from real ones. A related technology involves superimposing one person's face onto another, creating a so-called "deepfake." Deepfakes are used for purposes ranging from advertising to generating misogynistic pornography and disinformation (in 2018, for example, a deepfake video of Barack Obama uttering profanities about Donald Trump was circulated online over 2 million times). In the last few years, AI researchers have sought to keep up with improving deepfake technologies by crafting stronger deepfake detection algorithms.

### FaceForensics++

FaceForensics++ is a deepfake detection benchmarking dataset that contains approximately 1,000 original video sequences sourced from YouTube videos. Progress on FaceForensics++ is measured in terms of accuracy: the percentage of altered images an algorithm can correctly identify.

Although FaceForensics++ was introduced in 2019, researchers have tested previously existing deepfake detection methods on the dataset in order to track progress over time in deepfake detection (Figure 2.1.7). In the last decade, AI systems have become better and better at detecting deepfakes. In 2012, the top-performing systems could correctly identify 69.9% of deepfakes across all four FaceForencis++ datasets. In 2021, that number increased to 97.7%.[1]

FACEFORENSICS++: ACCURACY
Source: arXiv, 2021 | Chart: 2022 AI Index Report

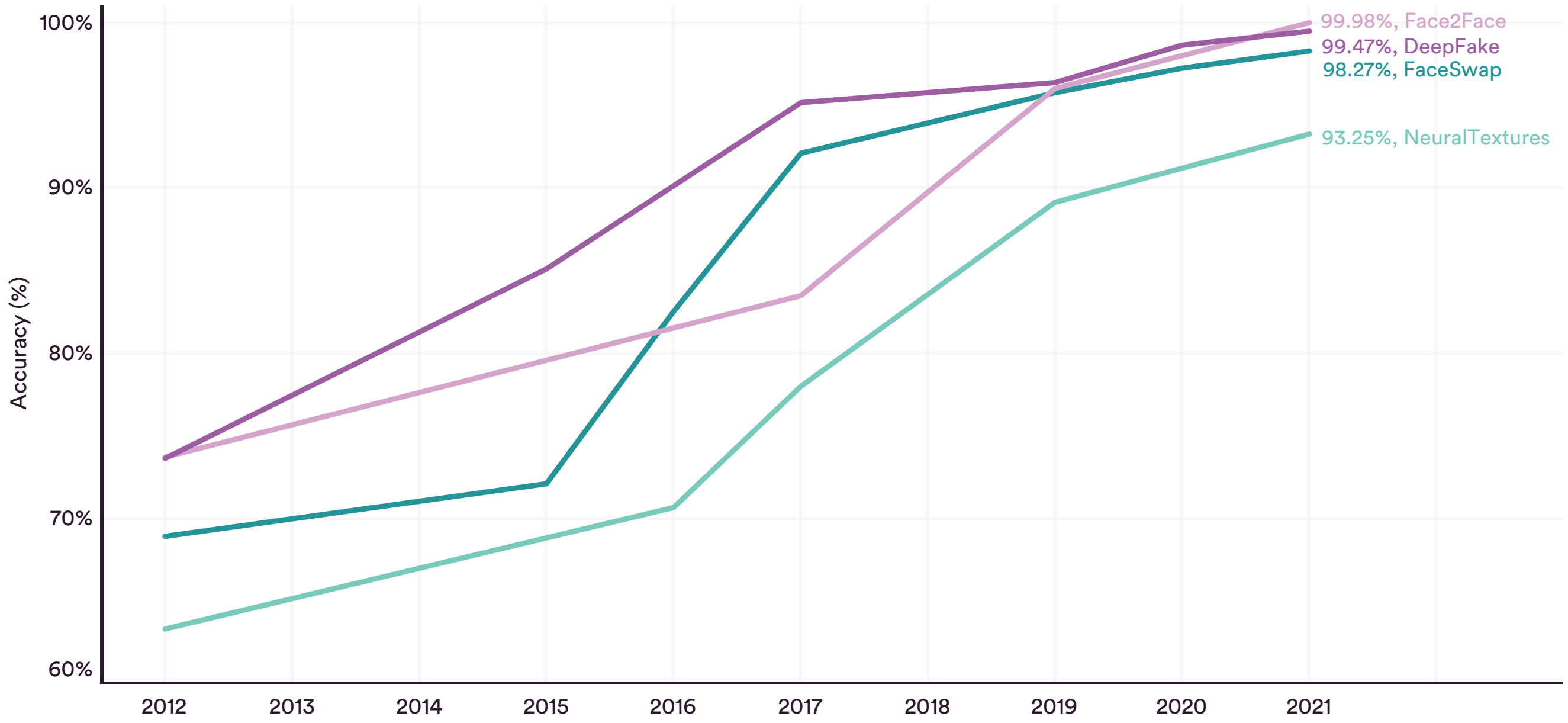

Figure 2.1.7

1 These numbers were taken by averaging performance across all four FaceForensics++ datasets.

## Celeb-DF

The Celeb-DF deepfake detection dataset is composed of 590 original celebrity YouTube videos manipulated into 5,639 deepfakes. Celeb-DF was introduced in 2019. In 2021, the top score on Celeb-DF was 76.9 and came from researchers at the University of Science and Technology of China and Alibaba group (Figure 2.1.8).

Top detection models perform significantly worse (by 20 percentage points) on Celeb-DF than FaceForensics++, suggesting that Celeb-DF is a more challenging dataset to test out techniques on. As deepfake technologies continue to improve in the upcoming years, it will be important to continue monitoring progress on Celeb-DF and other similarly challenging deepfake detection datasets.

**CELEB-DF: AREA UNDER CURVE SCORE (AUC)**
Source: arXiv, 2021 | Chart: 2022 AI Index Report

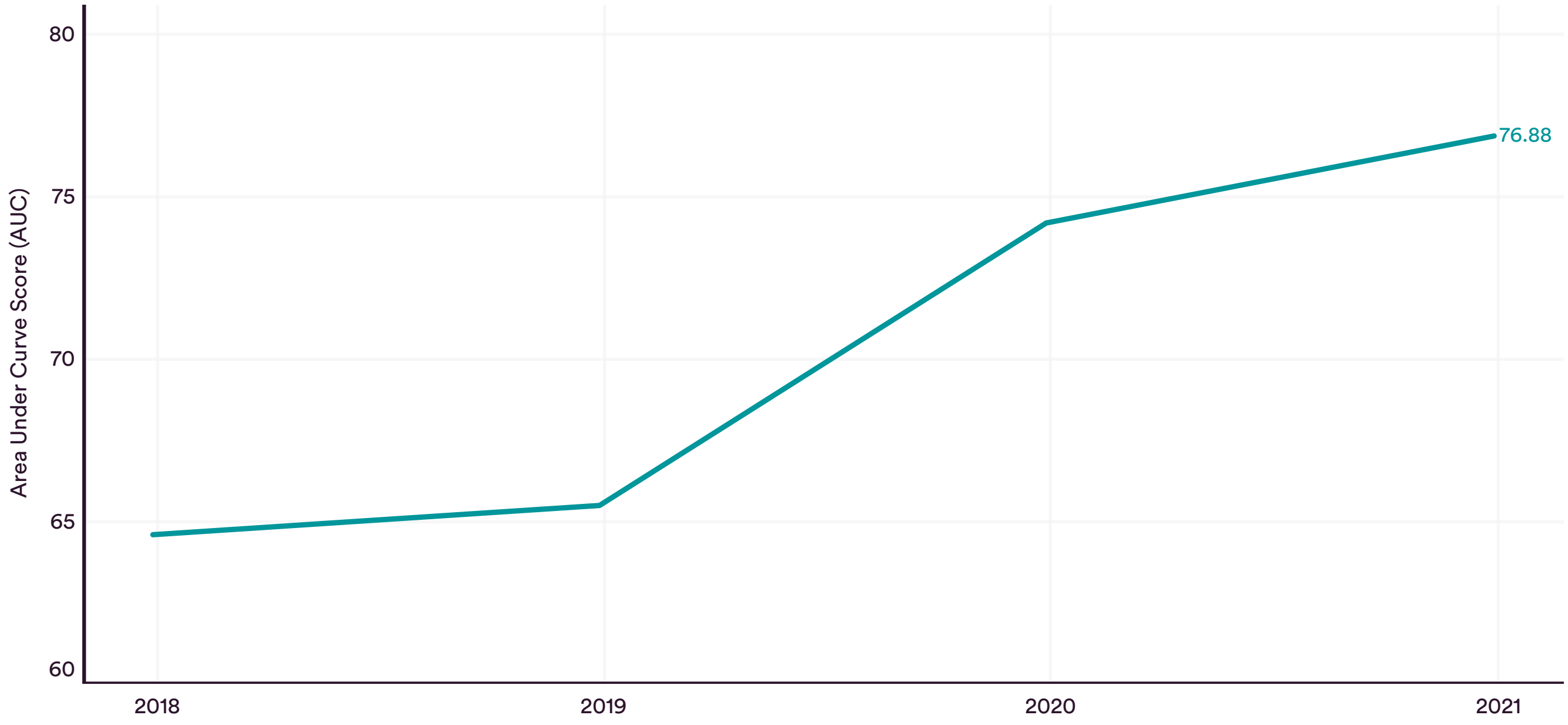

Figure 2.1.8

## HUMAN POSE ESTIMATION

Human pose estimation is the task of estimating different positions of human body joints (arms, head, torso, etc.) from a single image (Figure 2.1.9), and then combining these estimates to correctly label the pose the human is taking.

Human pose estimation can be used to facilitate activity recognition for purposes such as sports analytics, crowd surveillance, CGI development, virtual environment design, and transportation (for example, identifying the body language signs of an airport runway controller).

**A DEMONSTRATION OF HUMAN POSE ESTIMATION**
Source: Cao et al., 2019

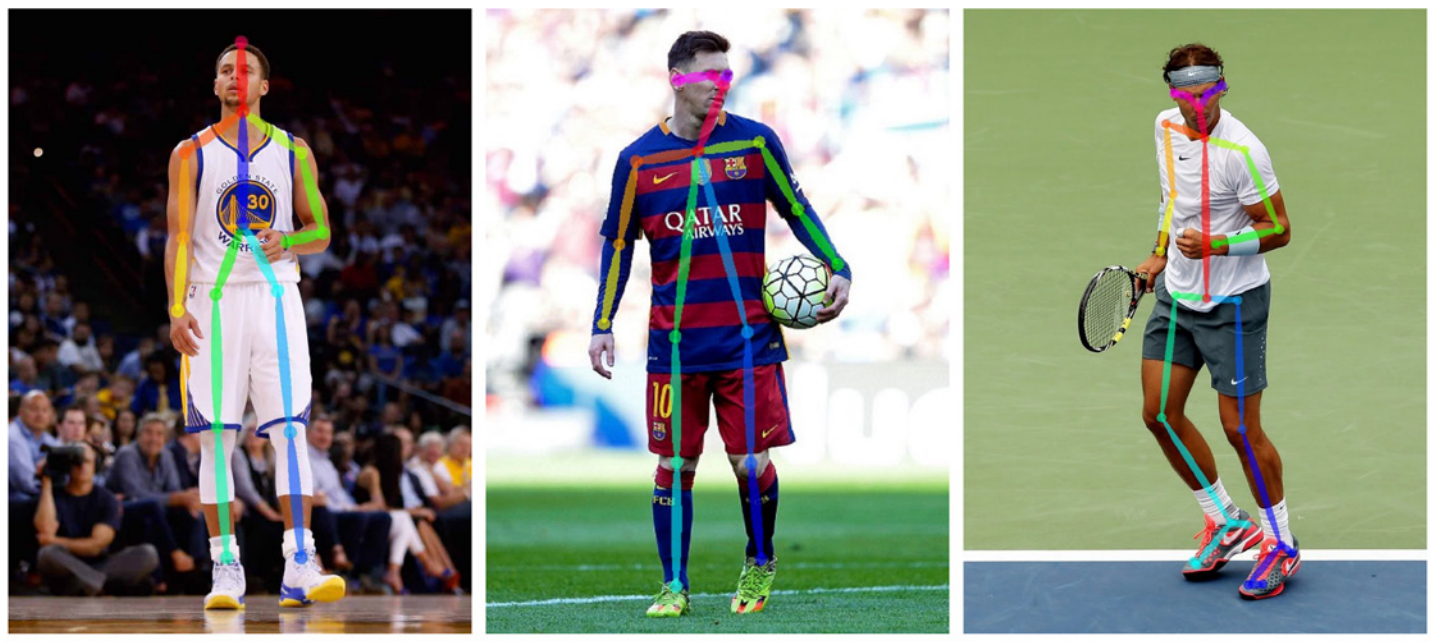

Figure 2.1.9

## Leeds Sports Poses: Percentage of Correct Keypoints (PCK)

The Leeds Sports Poses dataset contains 2,000 images collected from Flickr of athletes playing a sport. Each image includes information on 14 different body joint locations. Performance on the Leeds Sports Poses benchmark is assessed by the percentage of correctly estimated keypoints.

In 2021, the top-performing human pose estimation model correctly identified 99.5% of keypoints on Leeds Sports Poses (Figure 2.1.10). Given that maximum performance on Leeds Sports is 100.0%, more challenging benchmarks for human pose estimation will have to be developed, as we are close to saturating the benchmark.

**LEEDS SPORTS POSES: PERCENTAGE of CORRECT KEYPOINTS (PCK)**
Source: Papers with Code, 2021; arXiv, 2021 | Chart: 2022 AI Index Report

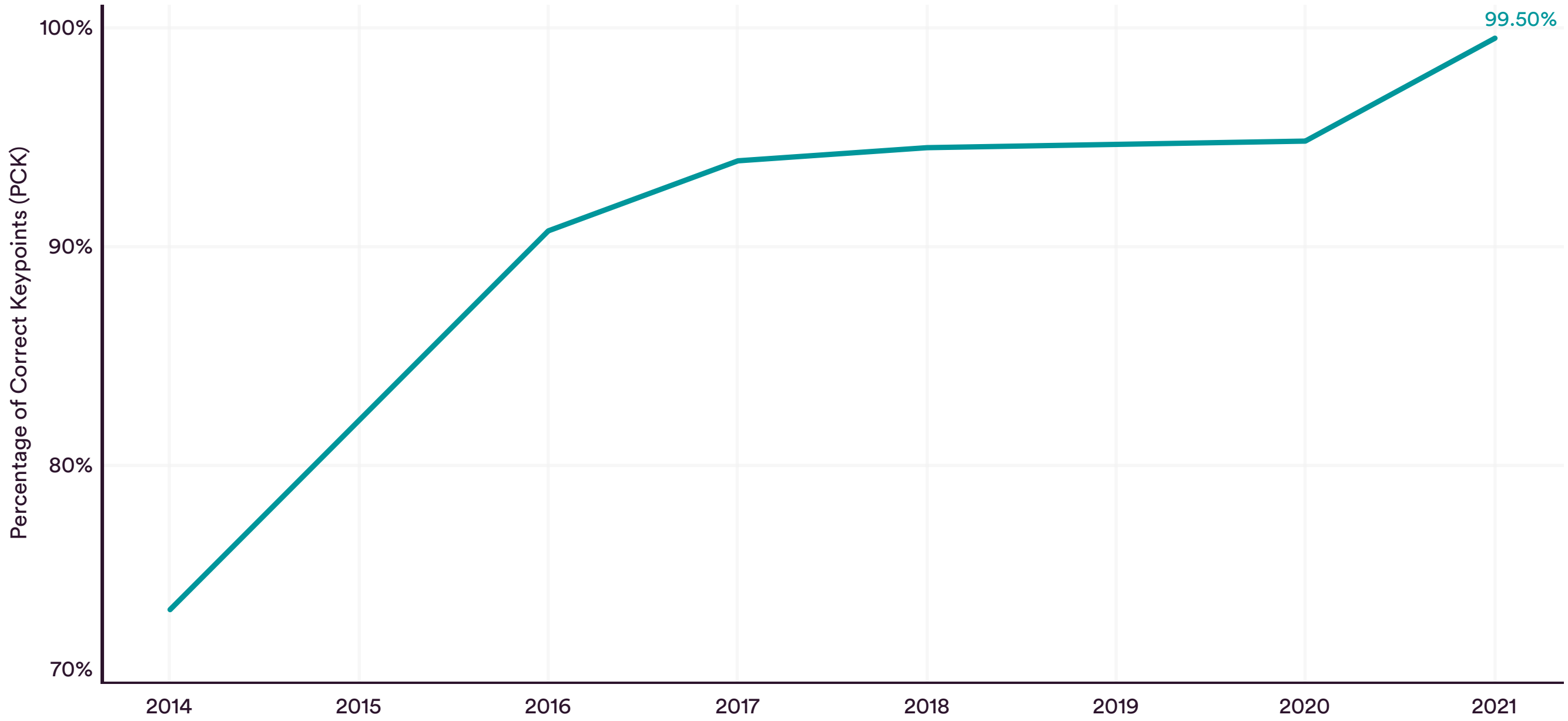

Figure 2.1.10

## Human3.6M: Average Mean Per Joint Position Error (MPJPE)

3D human pose estimation is a more challenging type of pose estimation, where AI systems are asked to estimate poses in a three- rather than two-dimensional space. The Human3.6M dataset tracks progress in 3D human pose estimation. Human3.6M is a collection of over 3.6 million images of 17 different types of human poses (talking on the phone, discussing, and smoking, etc.). Performance on Human3.6M is measured in average mean per joint position error in millimeters, which is the average difference between an AI model's position estimations and the actual position annotation.

In 2014, the top-performing model was making an average per joint error of 16 centimeters, half the size of a standard school ruler. In 2021, this number fell to 1.9 centimeters, less than the size of an average paper clip.

**HUMAN3.6M: AVERAGE MEAN PER JOINT POSITION ERROR (MPJPE)**
Source: Papers with Code, 2021; arXiv, 2021 | Chart: 2022 AI Index Report

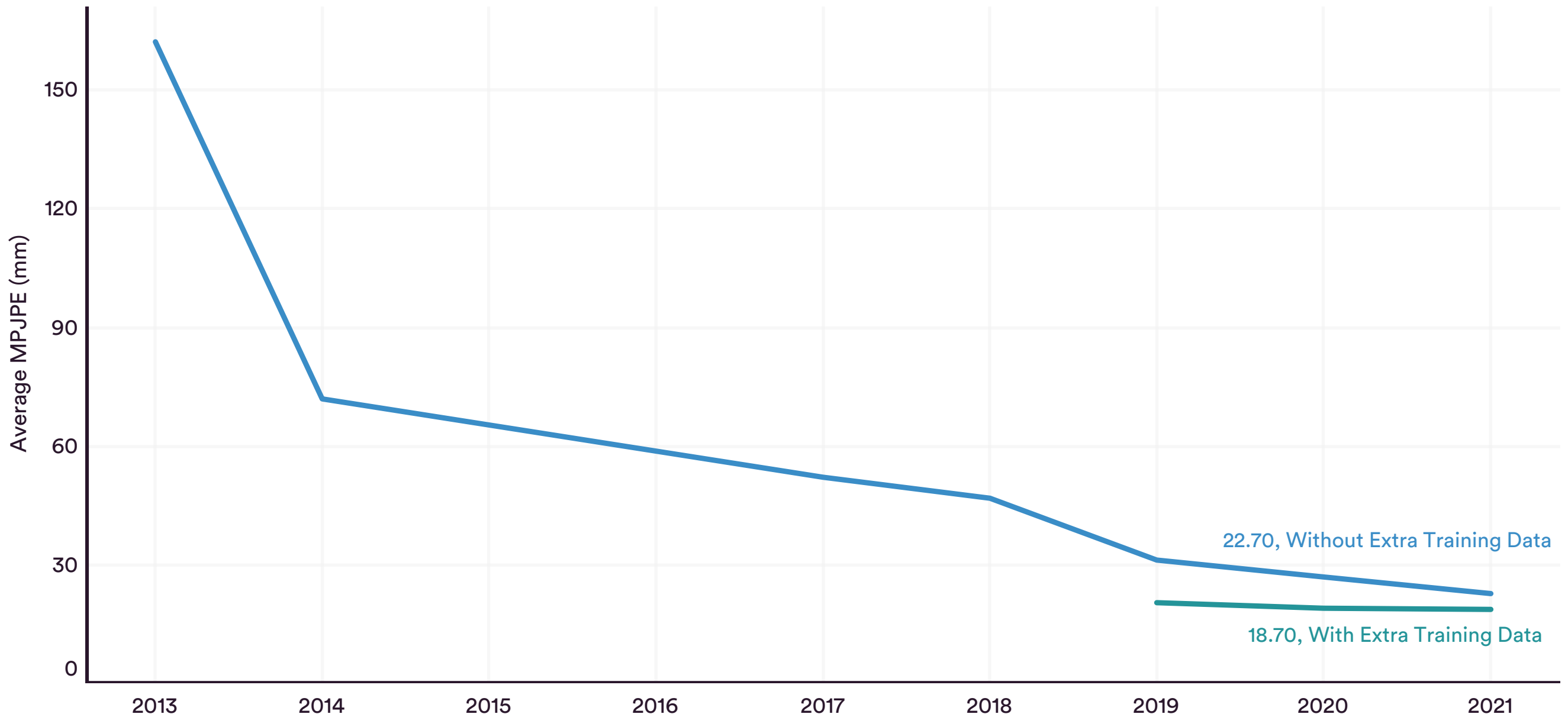

Figure 2.1.11

## SEMANTIC SEGMENTATION

Semantic segmentation is the task of assigning individual image pixels a category (such as person, bicycle, or background) (Figure 2.1.12). A plethora of real world domains require pixel-level image segmentation such as autonomous driving (identifying which parts of the image a car sees are pedestrians and which parts are roads), image analysis (distinguishing the foreground and background in photos), and medical diagnosis (segmenting tumors in lungs).

### A DEMONSTRATION OF SEMANTIC SEGMENTATION

Source: Visual Object Classes Challenge, 2012

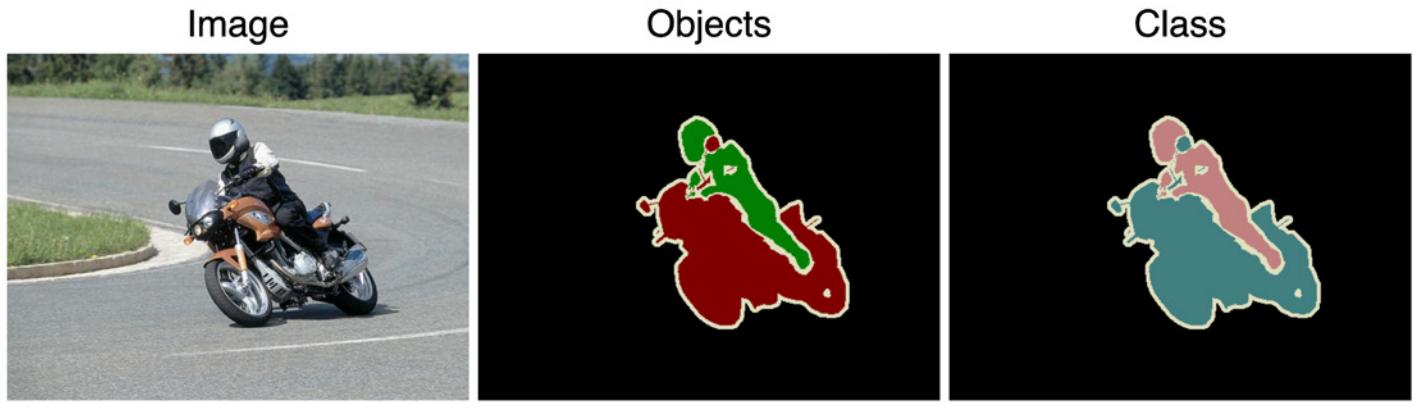

Figure 2.1.12

### Cityscapes

The Cityscapes dataset contains images of urban street environments from 50 cities, taken during the daytime in different seasons, and it allows for evaluation on a wide range of semantic segmentation tasks (instance-level, panoptic, and 3D vehicle).

The task to which most researchers submit is pixel-level semantic labeling, a challenge in which AI systems must semantically label an image on a per-pixel level. Challengers are evaluated on the intersection-over-union (IoU) metric, with a higher IoU score corresponding to better segmentation accuracy. In practical terms, a higher score means that a greater proportion of the image segments predicted by the model overlap with the image's actual segments.

The top-performing AI systems on Cityscapes in 2021 reported scores that are 14.6 percentage points higher than those in 2015. As with other computer vision tasks, the top performing models on Cityscapes have been pretrained on additional training data in the last few years.

### CITYSCAPES CHALLENGE, PIXEL-LEVEL SEMANTIC LABELING TASK: MEAN INTERSECTION-OVER-UNION (IOU)

Source: Cityscapes Challenge, 2021 | Chart: 2022 AI Index Report

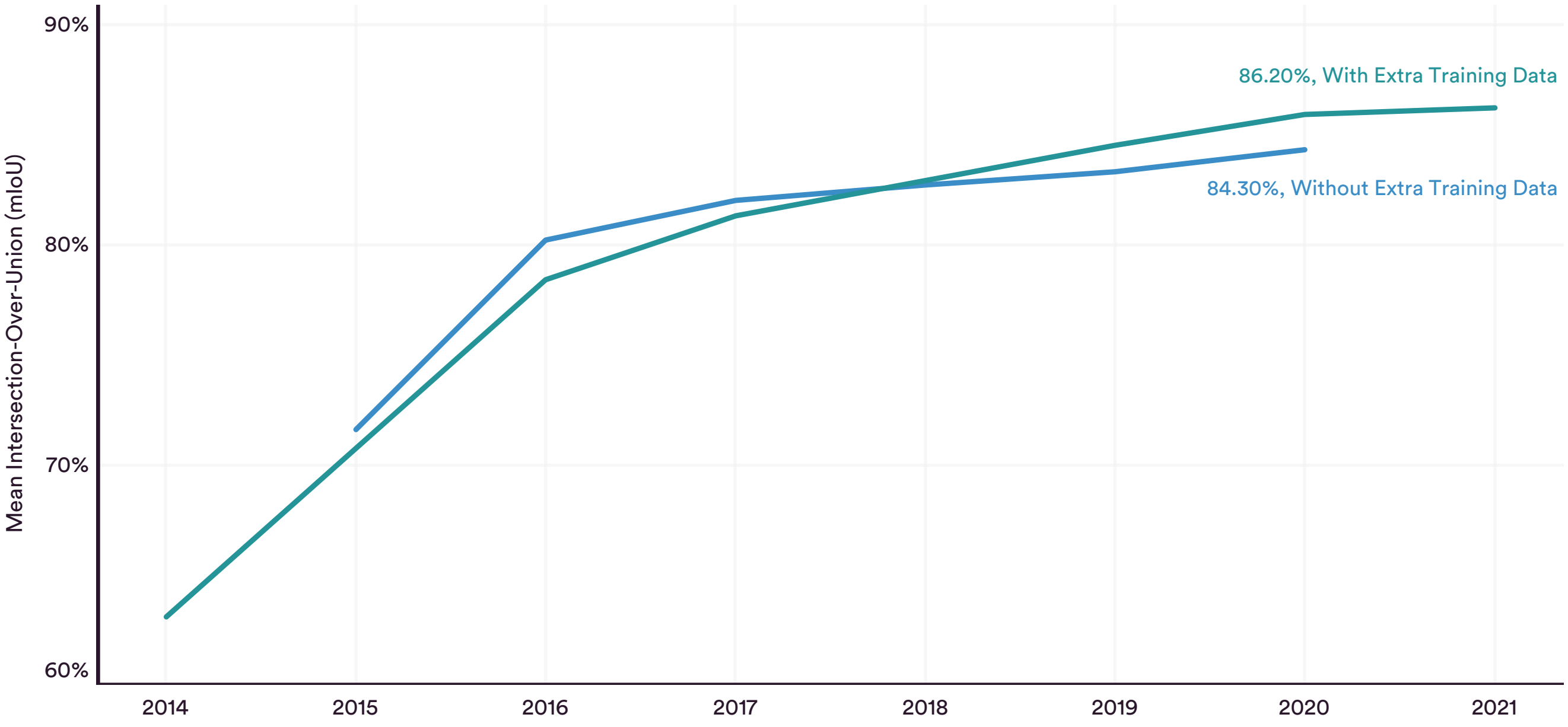

Figure 2.1.13

## MEDICAL IMAGE SEGMENTATION

Medical image segmentation refers to the ability of AI systems to segment objects of interest, such as organs, lesions, or tumors, in medical images (Figure 2.1.14). Technical progress in this task is vital to streamlining medical diagnoses. Advances in medical image segmentation mean doctors can spend less time on diagnosis and more time treating patients.

### CVC-ClinicDB and Kvasir-SEG

CVC-ClinicDB is a dataset that includes over 600 high-resolution images from 31 colonoscopies. Kvasir-SEG is a public dataset of 1,000 high-resolution gastrointestinal polyp images that were manually segmented by doctors and cross-verified by professional gastroenterologists. Both datasets are used to track progress in medical image segmentation. Performance is measured in mean DICE, which represents the average overlap between the polyp segments identified by an AI system and the actual polyp segments.

AI systems are now capable of correctly segmenting colonoscopy polyps at a rate of 94.2% on CVC-ClinicDB, representing an 11.9 percentage point improvement since 2015, and a 1.8 percentage improvement since 2020 (Figure 2.1.15). Similar progress has been made on Kvasir-SEG, where presently the top-performing AI model can accurately segment gastrointestinal polyps at a rate of 92.2%. The top results on both the CVC-ClinicDB and Kvasir-SEG benchmarks were achieved by the MSRF-Net model, one of the first convolutional neural networks designed specifically for medical image segmentation.

**A DEMONSTRATION OF KIDNEY SEGMENTATION**
Source: Kidney and Kidney Tumor Segmentation, 2021

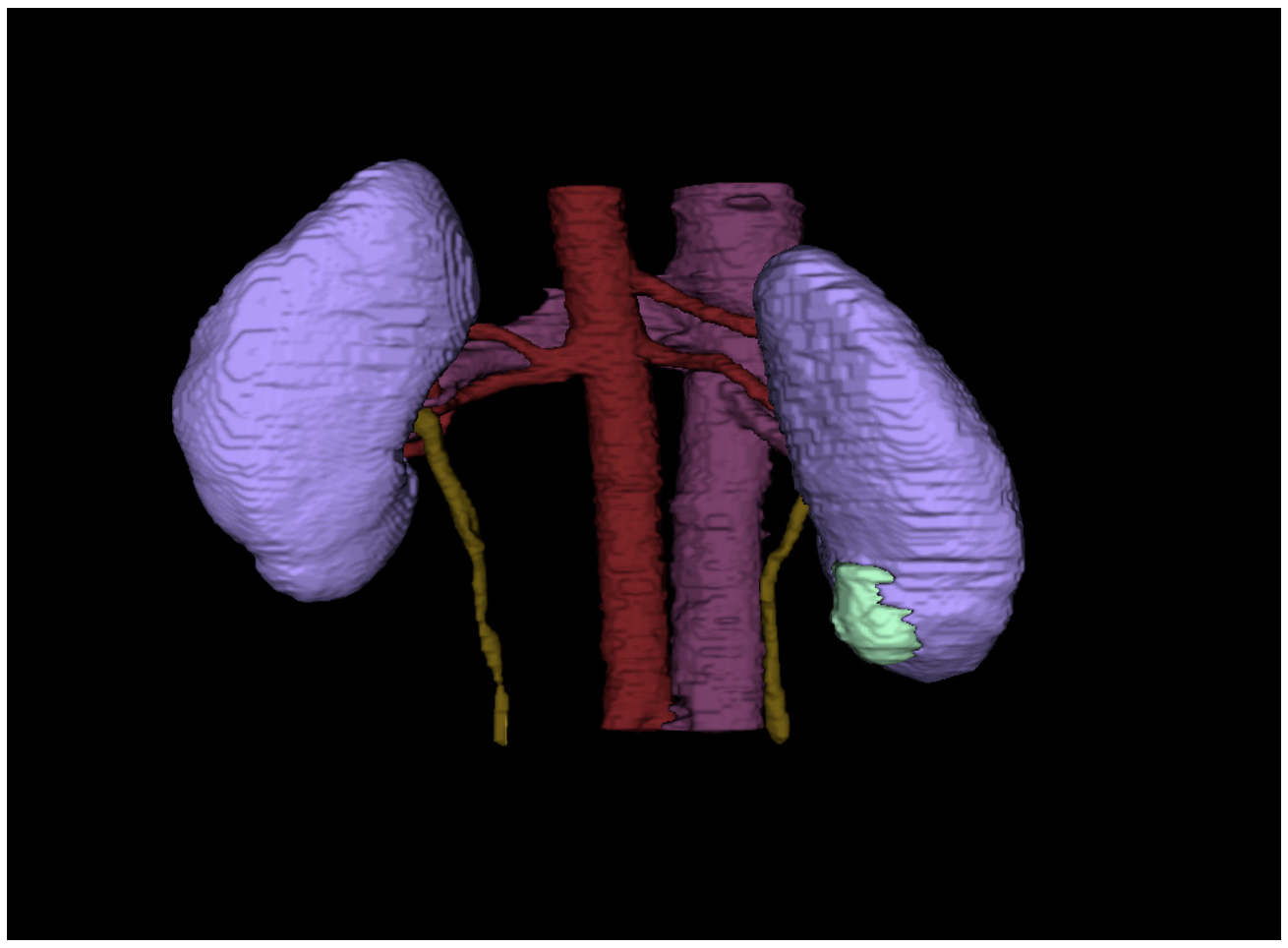

Figure 2.1.14

**CVC-CLINICDB: MEAN DICE**
Source: Papers with Code, 2021; arXiv, 2021 |Chart: 2022 AI Index Report

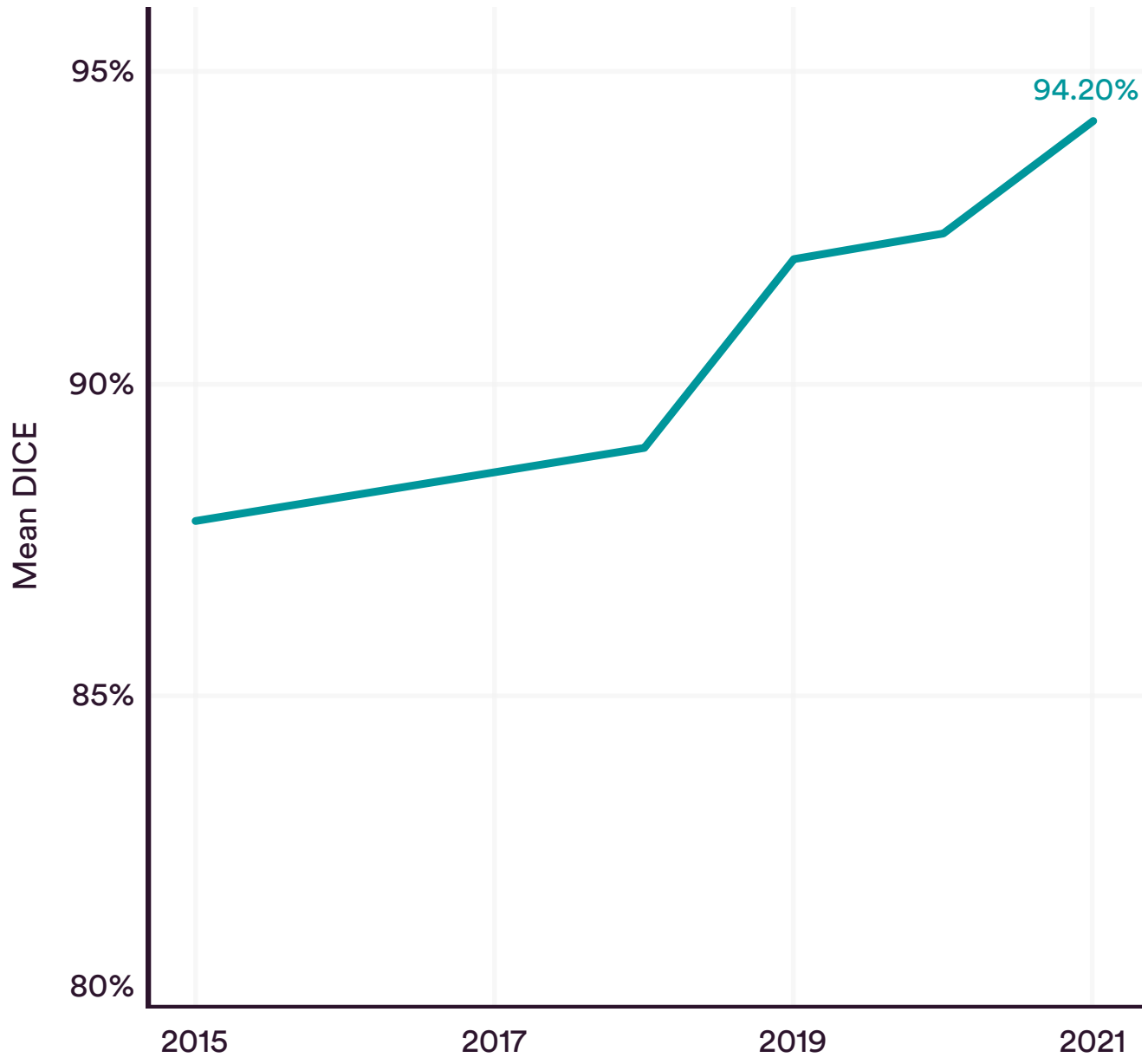

Figure 2.1.15a

**KVASIR-SEG: MEAN DICE**
Source: Papers with Code, 2021; arXiv, 2021 | Chart: 2022 AI Index Report

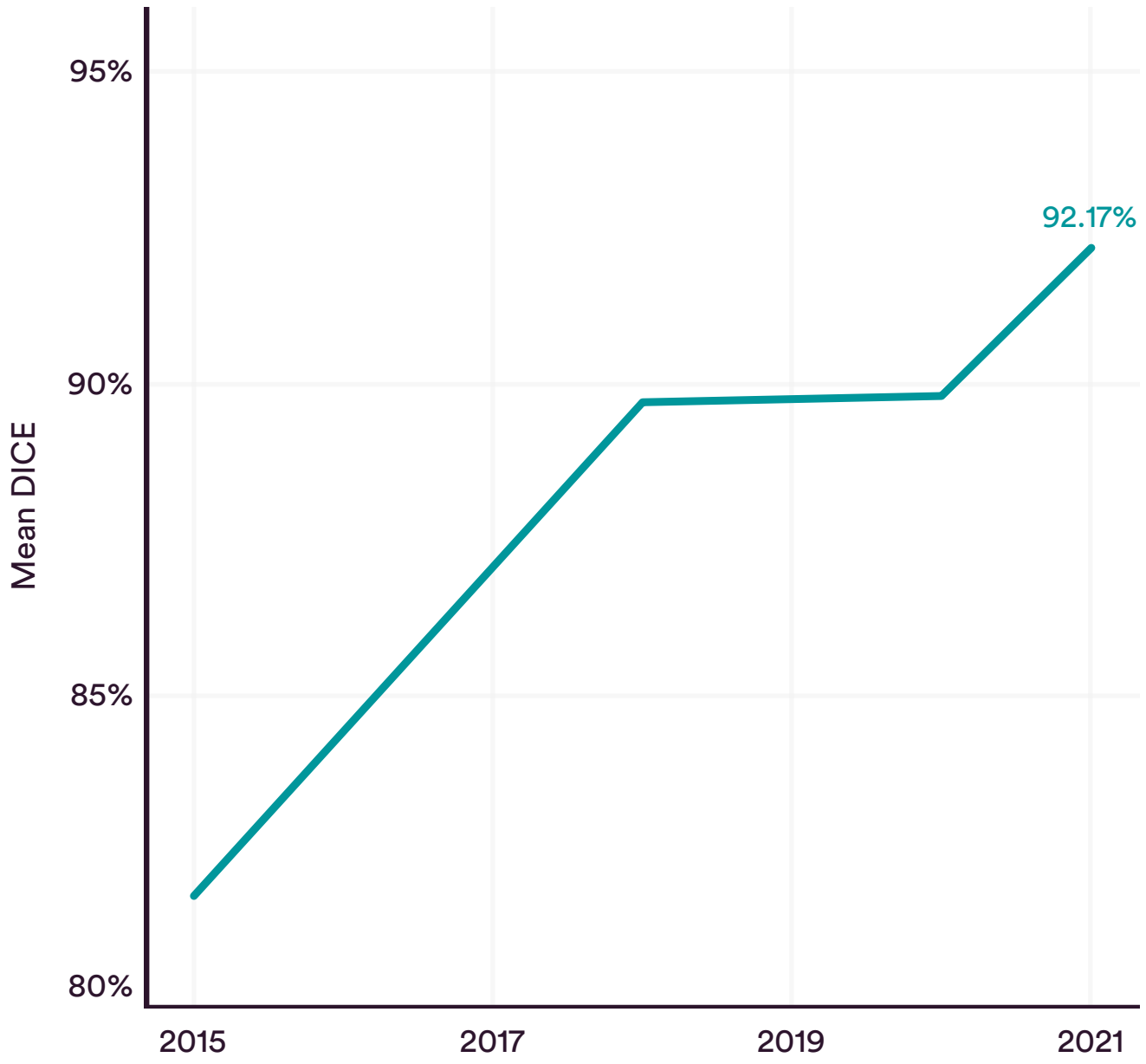

Figure 2.1.15b

The Kvasir-SEG benchmark also points to the explosion of interest in medical image segmentation. Prior to 2020, the dataset was referenced in only three academic papers. In 2020 that number rose to six, and in 2021 it shot up to 25. Last year also saw the hosting of KiTS21 (the Kidney and Kidney Tumor Segmentation Challenge), which challenged medical researchers from academia and industry to create the best systems for automatically segmenting renal tumors and the surrounding anatomy of kidneys.

## FACE DETECTION AND RECOGNITION

In facial detection, AI systems are tasked with identifying individuals in images or videos. Although facial recognition technology has existed for several decades, the technical progress in the last few years has been significant. Some of today's top-performing facial recognition algorithms have a near 100% success rate on challenging datasets.

Facial recognition can be used in transportation to facilitate cross-border travel, in fraud prevention to protect sensitive documents, and in online proctoring to identify illicit examination behavior. The greatest practical promise of facial recognition, however, is in its potential to aid security, which makes the technology extremely appealing to militaries and governments all over the world (e.g., 18 out of 24 U.S. government agencies are already using some kind of facial recognition technology).

### National Institute of Standards and Technology (NIST) Face Recognition Vendor Test (FRVT)

The National Institute of Standards and Technology's Face Recognition Vendor Test measures how well facial recognition algorithms perform on a variety of homeland security and law enforcement tasks, such as face recognition across photojournalism images, identification of child trafficking victims, deduplication of passports, and cross-verification of visa images. Progress on facial recognition algorithms is measured according to the false non-match rate (FNMR) or the error rate (the frequency with which a model fails to match an image to a person).

In 2017, some of the top-performing facial recognition algorithms had error rates of over 50.0% on certain FRVT tests. As of 2021, none has posted an error rate greater than 3.0%. The top-performing model across all datasets in 2021 (visa photos) registered an error rate of 0.1%, meaning that for every 1,000 faces, the model correctly identified 999.

**NATIONAL INSTITUTE OF STANDARDS AND TECHNOLOGY (NIST) FACE RECOGNITION VENDOR TEST (FRVT): VERIFICATION ACCURACY by DATASET**

Source: National Institute of Standards and Technology, 2021 | Chart: 2022 AI Index Report

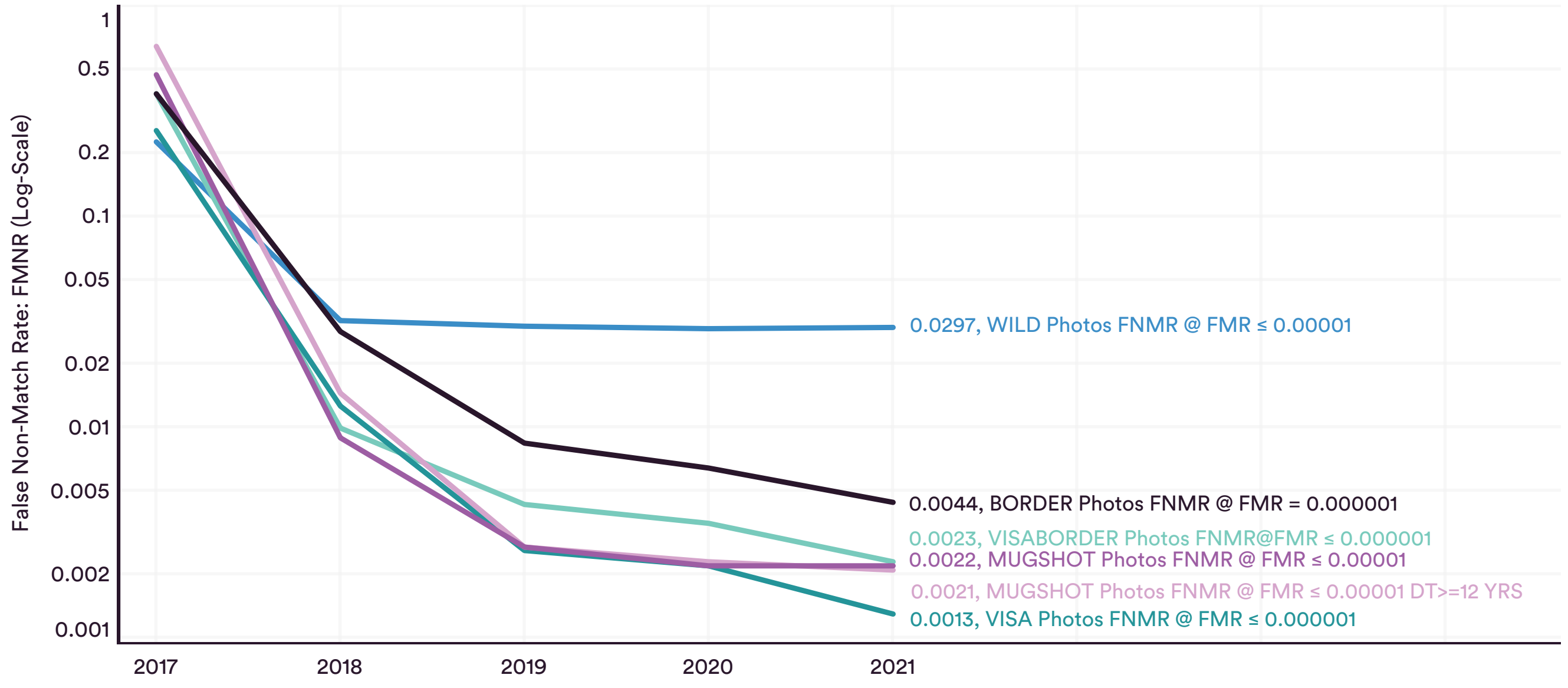

Figure 2.1.16

## FACE DETECTION: EFFECTS OF MASK-WEARING

### Face Recognition Vendor Test (FRVT): Face-Mask Effects

Facial recognition has become a more challenging task with the onset of the COVID-19 pandemic and accompanying mask mandates. The face-mask effects test asks AI models to identify faces on two datasets of visa border photos, one of which includes masked faces, the other which does not.

Three important trends can be gleaned from the FRVT face-mask test: (1) Facial recognition systems still perform relatively well on masked faces; (2) the performance on masked faces is worse than on non-masked faces; and (3) the gap in performance has narrowed since 2019.

**Although facial recognition technology has existed for several decades, the technical progress in the last few years has been significant. Some of today's top-performing facial recognition algorithms have a near 100% success rate on challenging datasets.**

NIST FRVT FACE MASK EFFECTS: FALSE-NON MATCH RATE
Source: National Institute of Standards and Technology, 2021 | Chart: 2022 AI Index Report

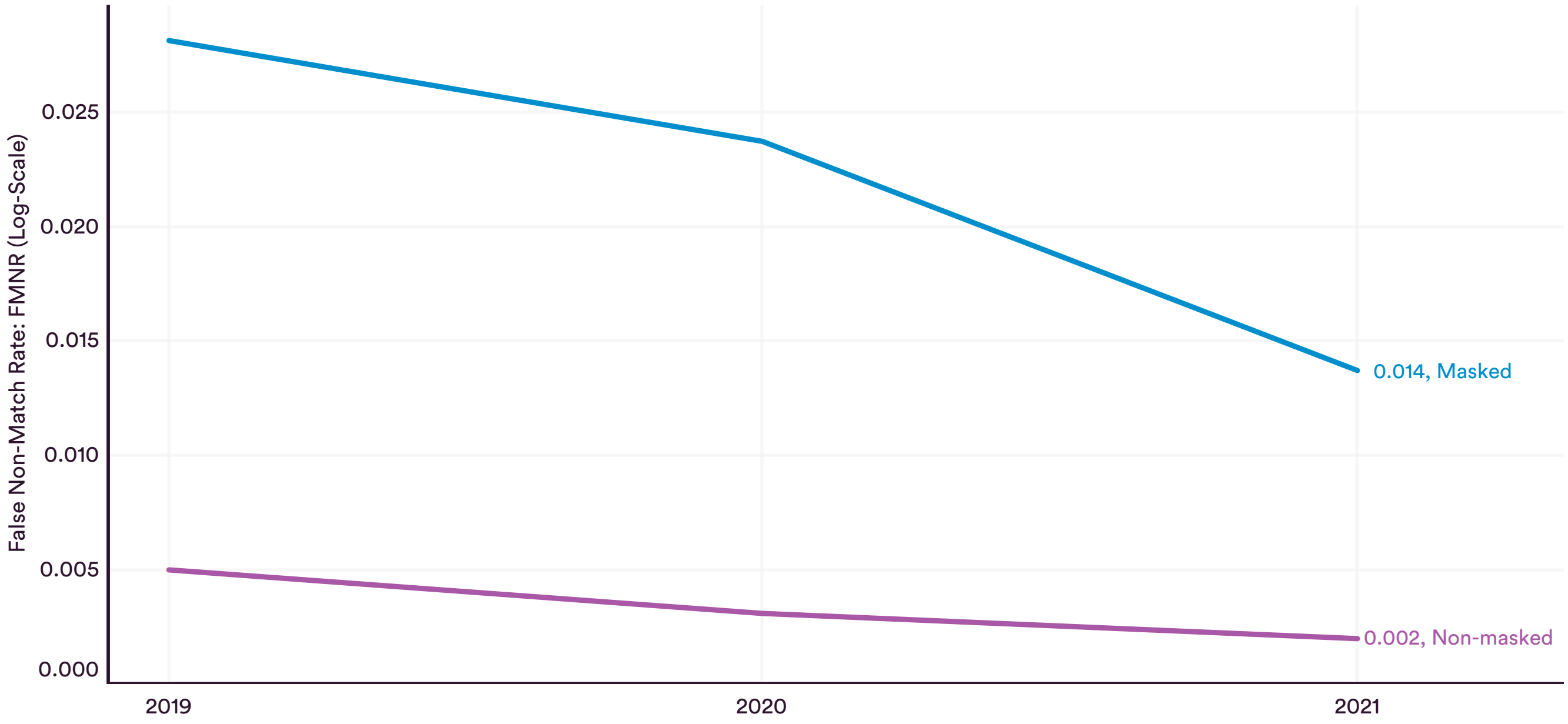

Figure 2.1.17

# Masked Labeled Faces in the Wild (MLFW)

In 2021, researchers from the Beijing University of Posts and Telecommunications released a facial recognition dataset of 6,000 masked faces in response to the new recognition challenges posed by large-scale mask-wearing.

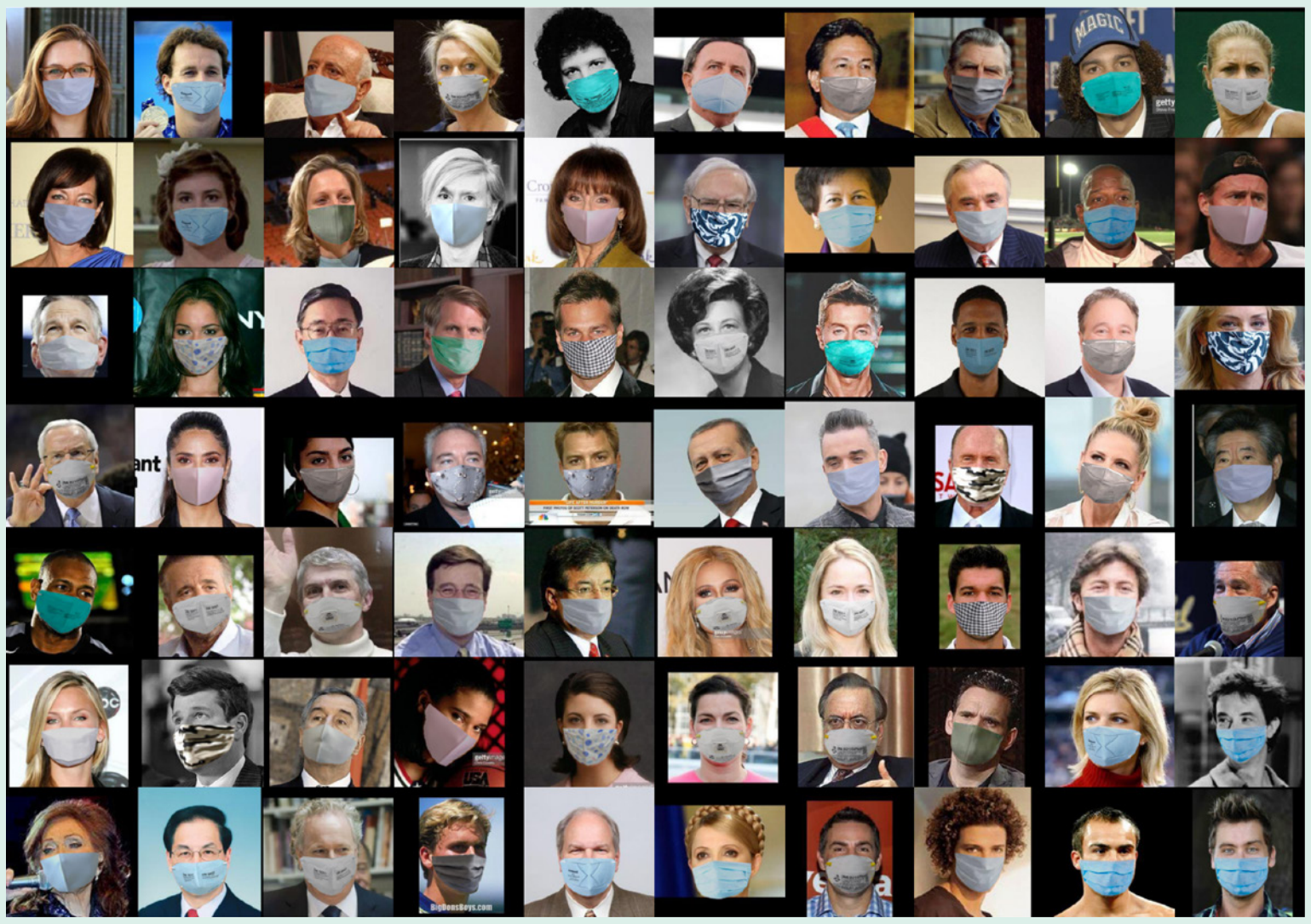

**EXAMPLES OF MASKED FACES IN THE MASKED LABELED FACES IN THE WILD (MLFW) DATABASE**
Source: Wang et al., 2021
Figure 2.1.18

As part of the dataset release, the researchers ran a series of existing state-of-the-art detection algorithms on a variety of facial recognition datasets, including theirs, to determine how much detection performance decreased when faces were masked. Their estimates suggest that top methods perform 5 to 16 percentage points worse on masked faces compared to unmasked ones. These findings somewhat confirm the insights from the FRVT face-mask tests: Performance deteriorates when masks are included, but not by an overly significant degree.

**STATE-OF-THE-ART FACE DETECTION METHODS on MASKED LABELED FACES IN THE WILD (MLFW): ACCURACY**
Source: Wang et. al, 2021 | Chart: 2022 AI Index Report

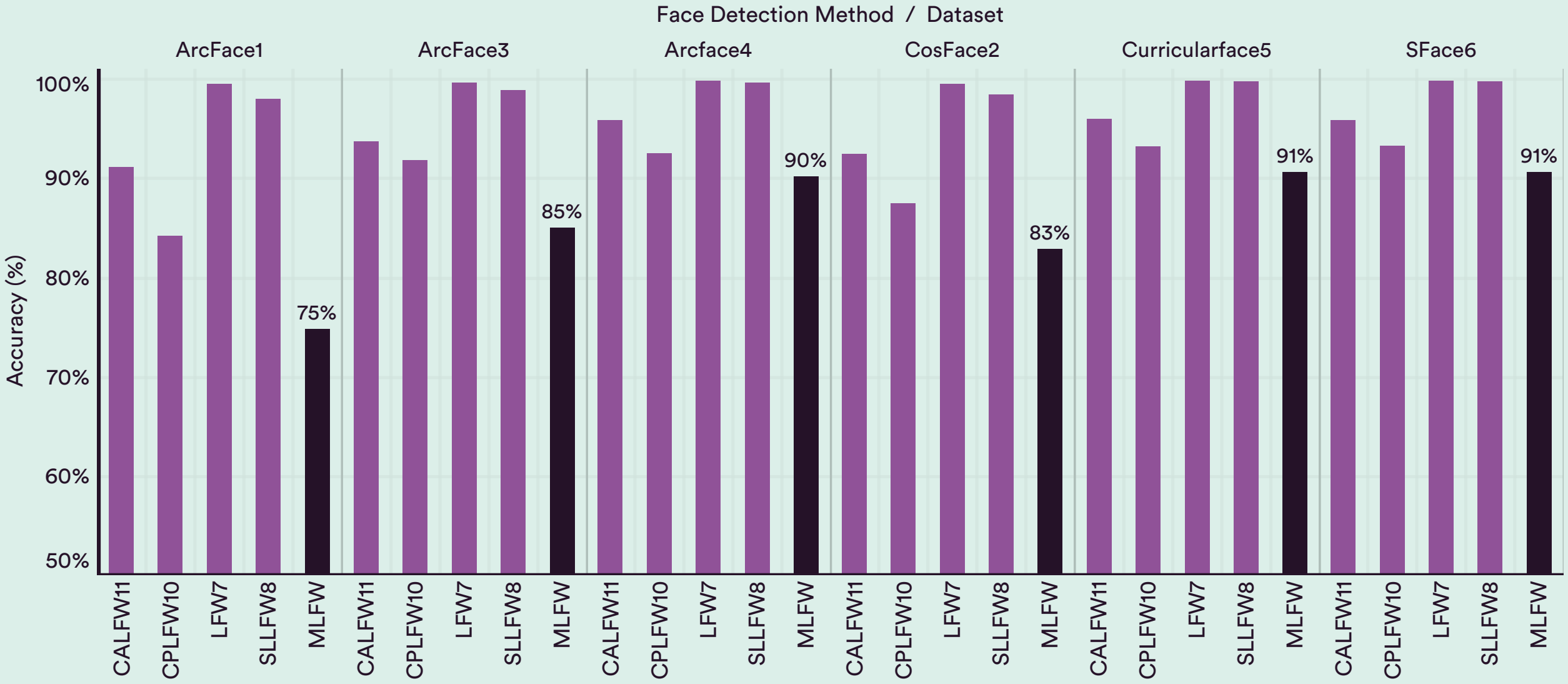

Figure 2.1.19

## VISUAL REASONING

Visual reasoning assesses how well AI systems can reason across a combination of visual and textual data. Visual reasoning skills are essential in developing AI that can reason more broadly. Existing AI can already execute certain narrow visual tasks better than humans, such as classifying images, detecting faces, and segmenting objects. But many AI systems struggle when challenged to reason more abstractly—for example, generating valid inferences about the actions or motivations of agents in an image (Figure 2.1.20).

### AN EXAMPLE OF A VISUAL REASONING TASK

Source: Goyal et al., 2021

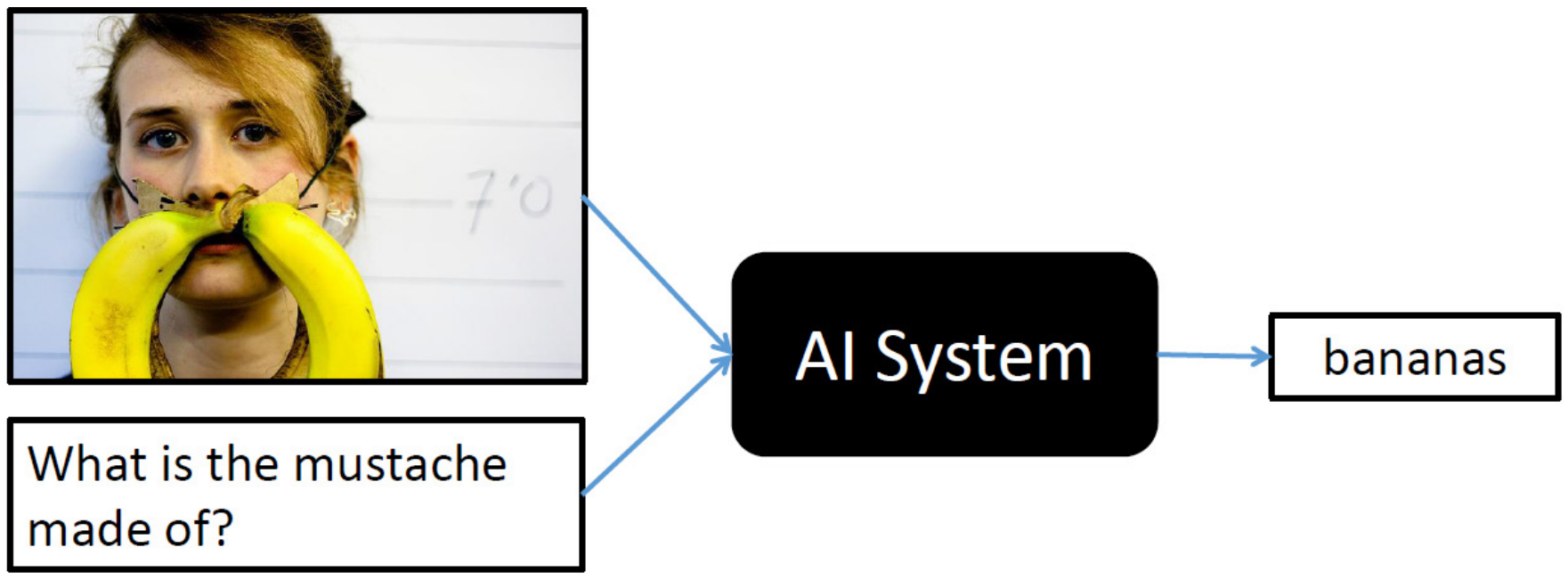

Figure 2.1.20

### Visual Question Answering (VQA) Challenge

In the Visual Question Answering challenge, AI systems are tasked with answering open-ended questions about images (Figure 2.1.21). To answer the questions at a high level, AI systems must have a combined understanding of language, vision, and commonsense reasoning.

### SAMPLE QUESTIONS IN THE VISUAL QUESTION ANSWERING (VQA) CHALLENGE

Source: Goyal et al., 2017

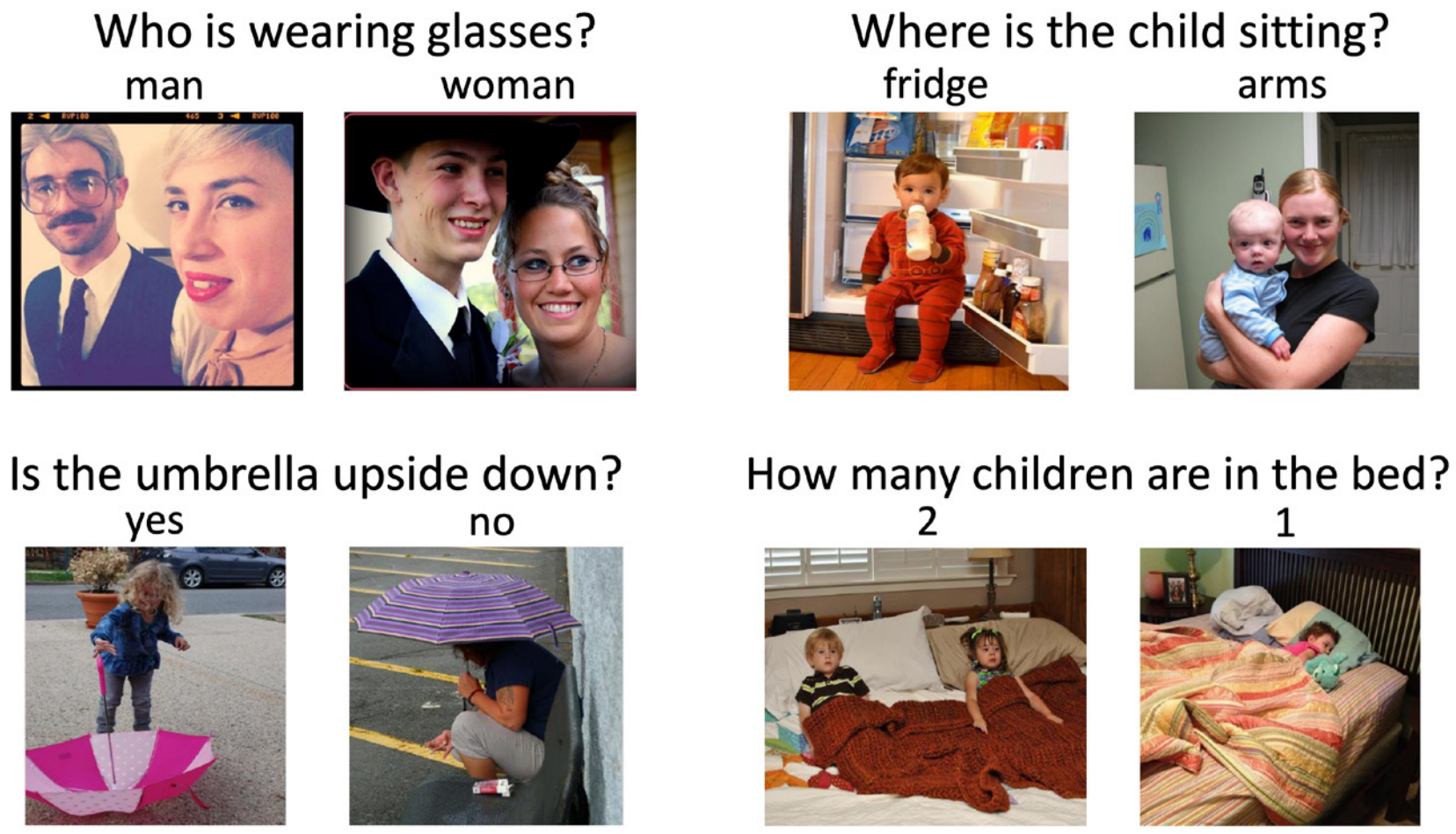

Figure 2.1.21

In the six years since the VQA challenge began, there has been a 24.4 absolute percentage point improvement in state-of-the-art performance. In 2015, the top-performing systems could correctly answer only 55.4% of questions (Figure 2.1.22). As of 2021, top performance stood at 79.8%—close to the human baseline of 80.8%.

## VISUAL QUESTION ANSWERING (VQA) CHALLENGE: ACCURACY

Source: VQA Challenge, 2021 | Chart: 2022 AI Index Report

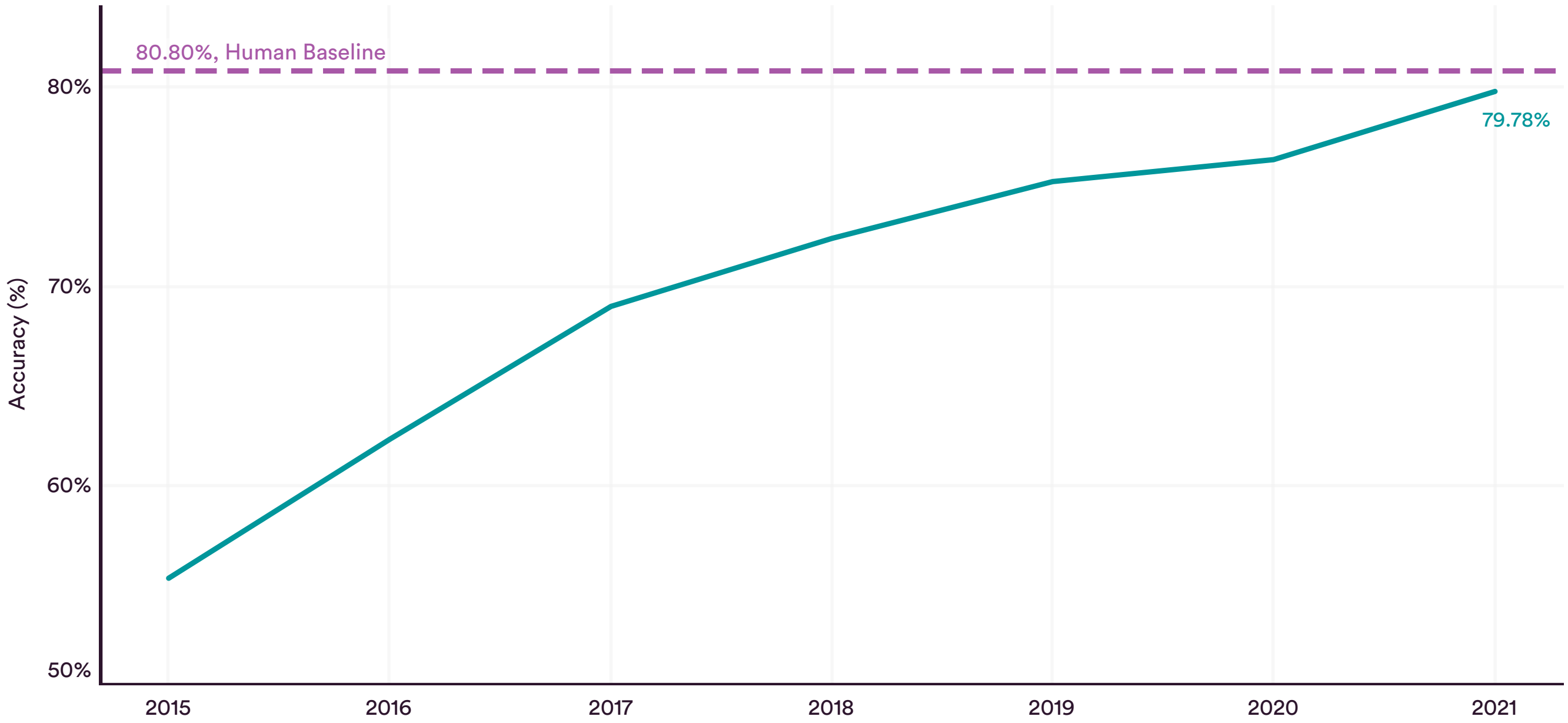

Figure 2.1.22

**Video analysis concerns reasoning or task operation across sequential frames (videos), rather than single frames (images). Video computer vision has a wide range of use cases, which include assisting criminal surveillance efforts, sports analytics, autonomous driving, navigation of robots, and crowd monitoring.**

# 2.2 COMPUTER VISION—VIDEO

## ACTIVITY RECOGNITION

A fundamental subtask in video computer vision is activity recognition: identifying the activities that occur in videos. AI systems have been challenged to classify activities that range from simple actions, like walking, waving, or standing, to ones that are more complex and contain multiple steps, like preparing a salad (which requires an AI system to recognize and chain together discrete actions like cutting tomatoes, washing the greens, applying dressing, etc.)

### Kinetics-400, Kinetics-600, Kinetics-700

Kinetics-400, Kinetics-600, and Kinetics-700 are a series of datasets for benchmarking video activity recognition. Each dataset includes 650,000 large-scale, high-quality video clips from YouTube that display a wide range of human activities, and asks AI systems to classify an action from a possible set of 400, 600, and 700 categories, respectively. Some of the new and more challenging activity classes added to the Kinetics-700 series include pouring wine, playing the oboe, and making latte art.

**EXAMPLE CLASSES FROM THE KINETICS DATASET**
Source: Kay et al., 2017

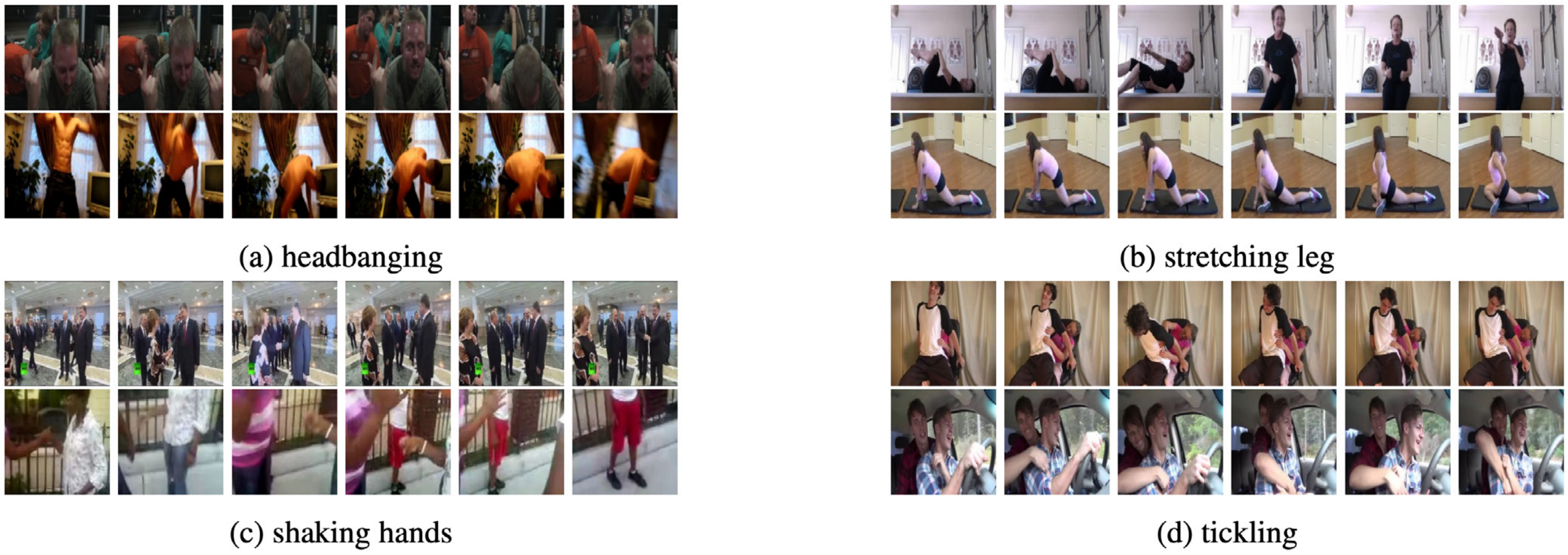

Figure 2.2.1

As of 2022, one model tops all three Kinetics datasets. MTV, a collaboration between Google Research, Michigan State University, and Brown University, released in January 2022, achieved a 89.6% Top-1 accuracy on the 600 series, 89.1% accuracy on the 400 series, and 82.20% accuracy on the 700 series (Figure 2.2.2). The most striking aspect about technical progress on Kinetics is how rapidly the gap has narrowed between performance on the datasets. In 2020, the gap between performance on Kinetics-400 and Kinetics-700 was 27.14 percentage points. In one short year, that gap has narrowed to 7.4 points, which means that performance on the newer, harder dataset is occurring more rapidly than performance on the easier dataset and suggests that the easier ones are starting to asymptote.

**KINETICS-400, KINETICS-600, KINETICS-700: TOP-1 ACCURACY**

Source: Papers with Code, 2021; arXiv, 2021 | Chart: 2022 AI Index Report

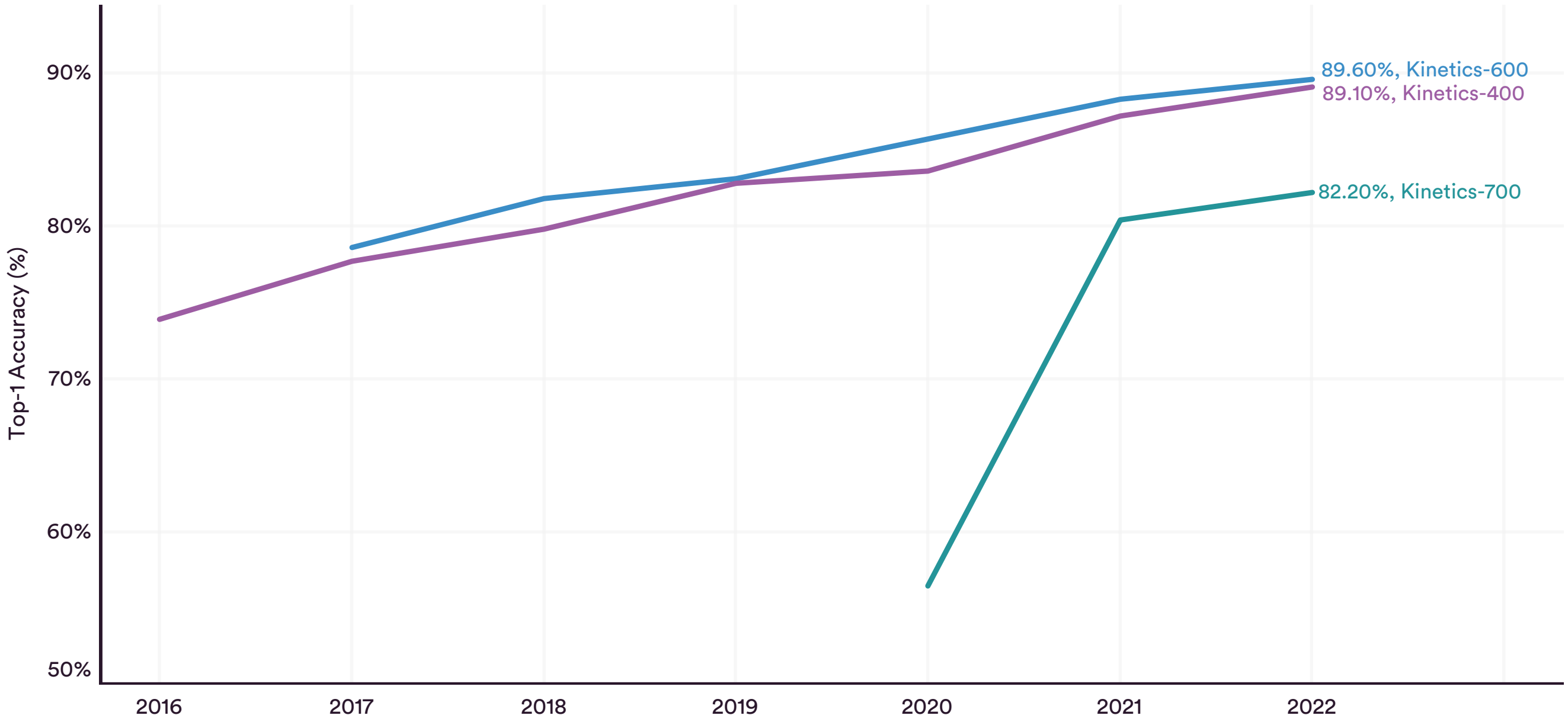

Figure 2.2.2

## ActivityNet: Temporal Action Localization Task

ActivityNet is a video dataset for human activity understanding that contains 700 hours of videos of humans doing 200 different activities (long jump, dog walking, vacuuming, etc.). For an AI system to successfully complete the ActivityNet Temporal Action Localization Task (TALT) task, it has to execute two separate steps: (1) localization (identify the precise interval during which the activity occurs); and (2) recognition (assign the correct category label). Temporal action localization is one of the most complex and difficult tasks in computer vision. Performance on TALT is measured in terms of mean average precision, with a higher score indicating greater accuracy.

As of 2021 the top-performing model on TALT, developed by HUST-Alibaba, scores 44.7%, a 26.9 percentage point improvement over the top scores posted in 2016 when the challenge began (Figure 2.2.3). Although state-of-the-art results on the task have been posted for each subsequent year, the gains have become increasingly small.

**ACTIVITYNET, TEMPORAL ACTION LOCALIZATION TASK: MEAN AVERAGE PRECISION (mAP)**
Source: ActivityNet, 2021 | Chart: 2022 AI Index Report

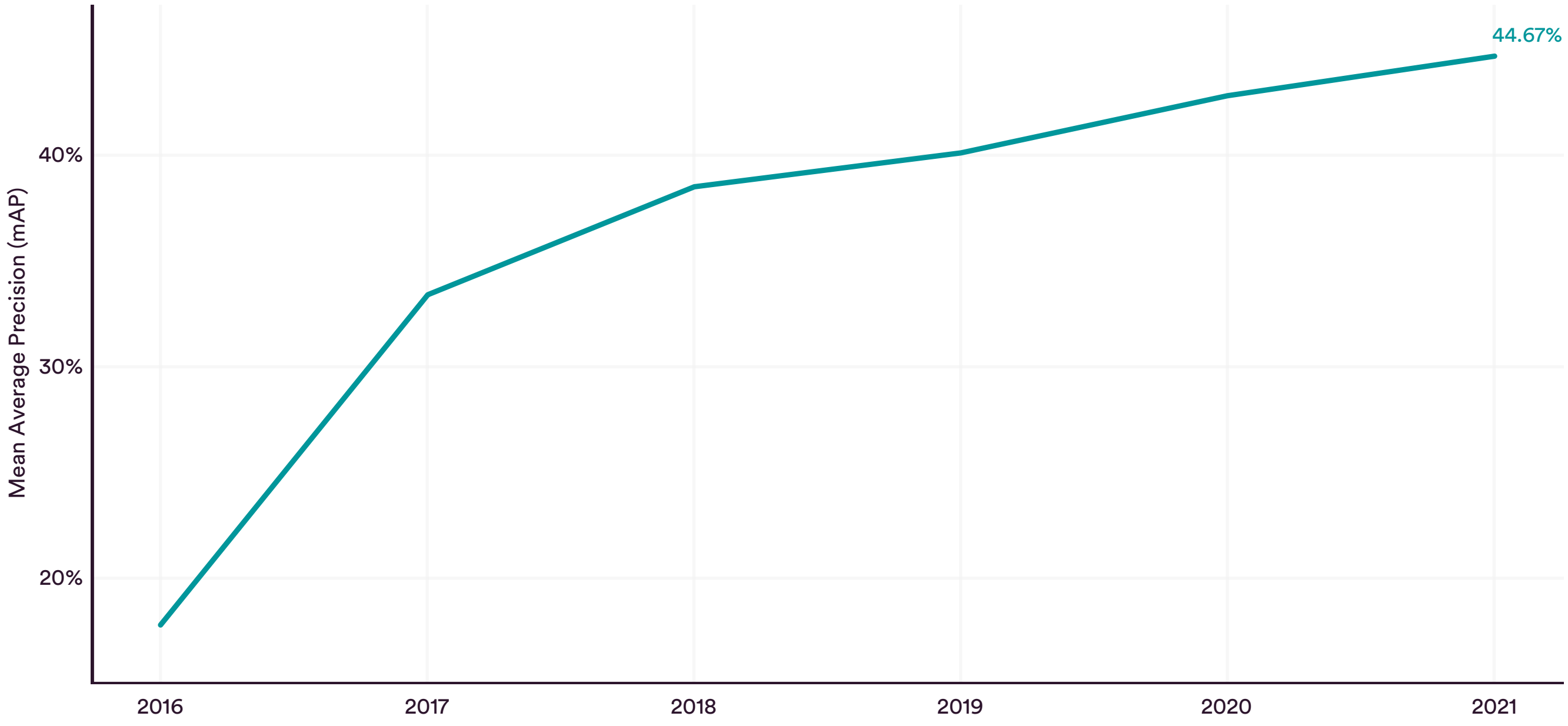

Figure 2.2.3

## OBJECT DETECTION

Object detection is the task of identifying objects within an image (Figure 2.2.4). There are different philosophies bearing on priority, speed, and accuracy that guide the design of object detection systems. Systems that train quickly might be more efficient but are less accurate. Those that are more accurate might perform better but take longer to process a video. This tradeoff between speed and accuracy is also reflected in the types of object detection methods pioneered in the last decade. There are one-stage methods which prioritize speed, such as SSD, RetinaNet, and YOLO, and two-stage methods which prioritize accuracy, such as Mask R-CNN, Faster R-CNN, and Cascade R-CNN.

### A DEMONSTRATION OF HOW OBJECT DETECTION APPEARS TO AI SYSTEMS

Source: COCO, 2020

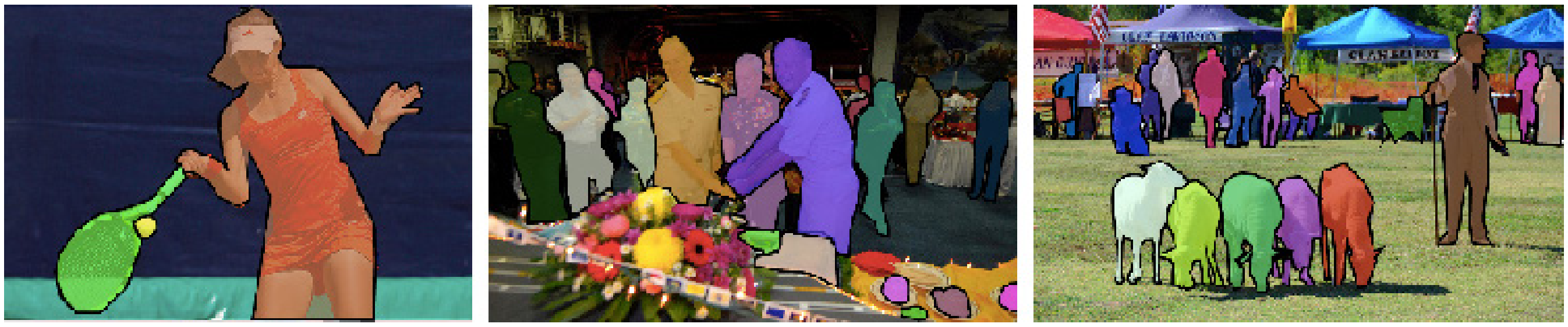

Figure 2.2.4

## Common Object in Context (COCO)

Microsoft's Common Object in Context (COCO) object detection dataset contains over 328,000 images across more than 80 object categories. There are many accuracy metrics used to track performance on object detection, but for the sake of consistency, this section and the majority of this report considers mean average precision (mAP50).

Since 2016, there has been a 23.8 percentage point improvement on COCO object detection, with this year's top model, GLIP, registering a mean average precision of 79.5%.[2] Figure 2.2.5 illustrates how the use of extra training data has taken over object detection, much as it has with other domains of computer vision.

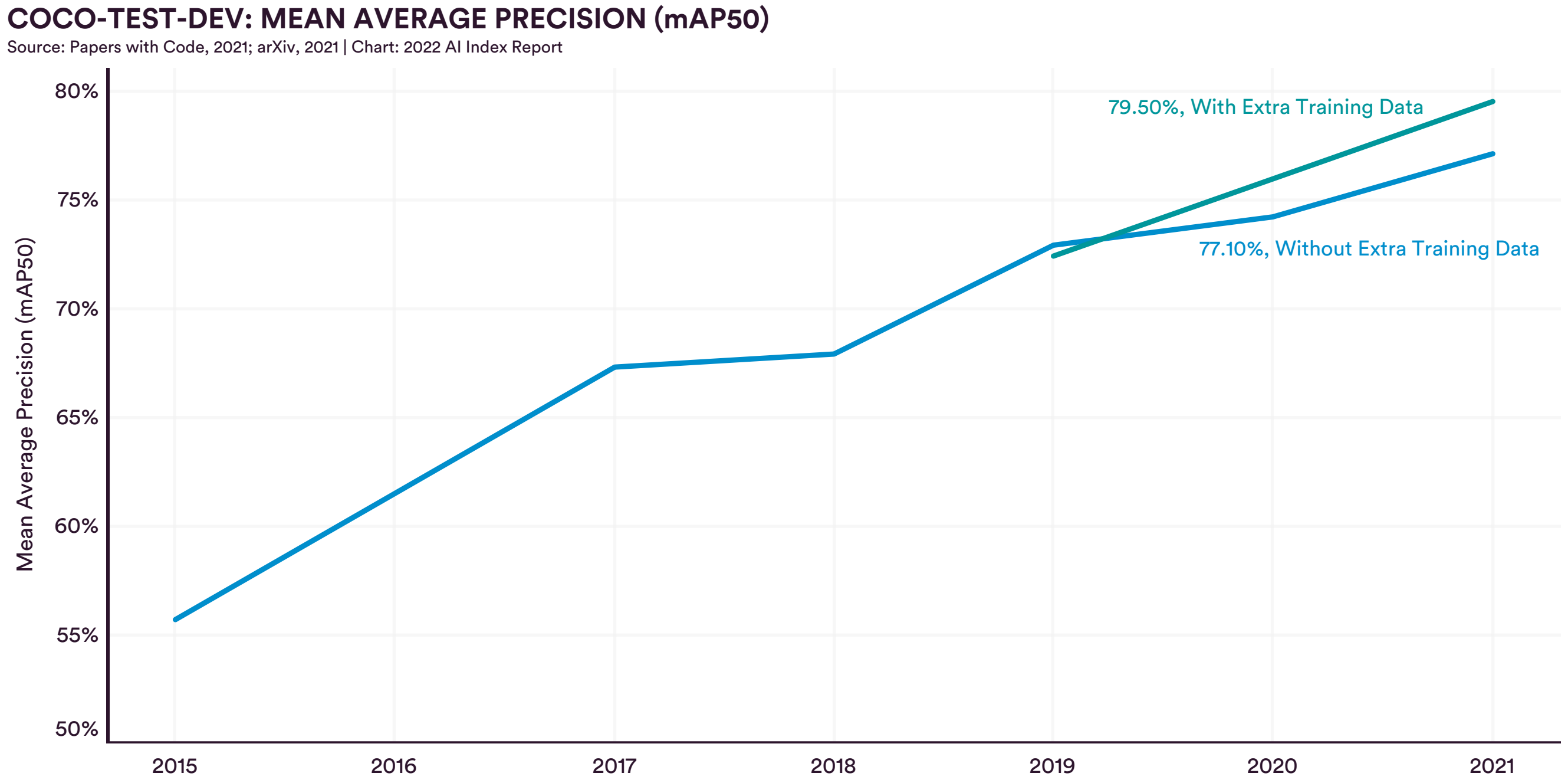

Figure 2.2.5

2 GLIP (Grounded Language-Image Pretraining), a model designed to master the learning of language contextual visual representations, was a collaboration of researchers from UCLA, Microsoft Research, University of Washington, University of Wisconsin–Madison, Microsoft Cloud, Microsoft AI, and International Digital Economy Academy.

## You Only Look Once (YOLO)

You Only Look Once is an open-source object detection model that emphasizes speed (inference latency) over absolute accuracy.

Over the years, there have been different iterations of YOLO, and Figure 2.2.6 plots the performance of YOLO object detectors versus the absolute top performing detectors on the COCO dataset. YOLO detectors have become much better in terms of performance since 2017 (by 28.4 percentage points). Second, the gap in performance between YOLO and the best-performing object detectors has narrowed. In 2017 the gap stood at 11.7%, and it decreased to 7.1% in 2021. In the last five years, object detectors have been built that are both faster and better.

**STATE OF THE ART (SOTA) vs. YOU ONLY LOOK ONCE (YOLO): MEAN AVERAGE PRECISION (mAP50)**

Source: arXiv, 2021; GitHub, 2021 | Chart: 2022 AI Index Report

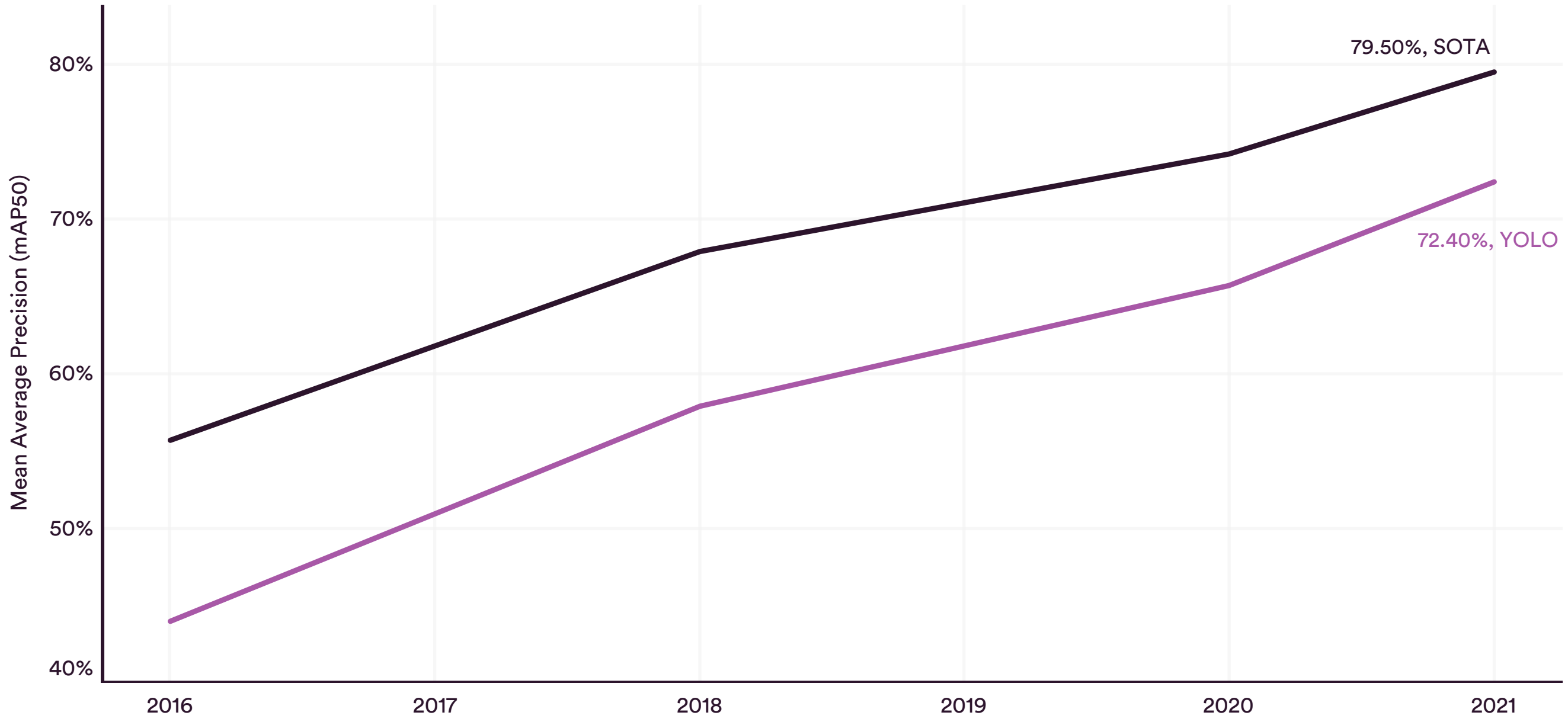

Figure 2.2.6

## Visual Commonsense Reasoning (VCR)

The Visual Commonsense Reasoning challenge is a relatively new benchmark for visual understanding. VCR asks AI systems to answer challenging questions about scenarios presented from images, and also to provide the reasoning behind their answers (unlike the VQA challenge, which only requires an answer). The dataset contains 290,000 pairs of multiple-choice questions, answers, and rationales from 110,000 image scenarios taken from movies. Figure 2.2.7 illustrates the kinds of questions posed in the VCR.

### A SAMPLE QUESTION OF THE VISUAL COMMONSENSE REASONING (VCR) CHALLENGE

Source: Zellers et al., 2018

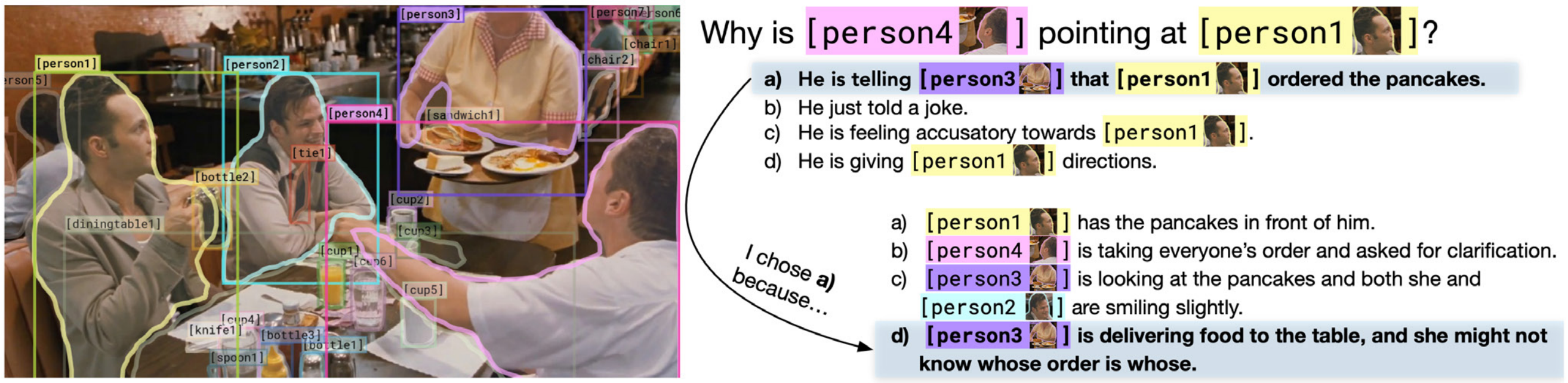

Figure 2.2.7

Performance on VCR is measured in the Q->AR score, which aggregates how well machines can choose the right answer for a given multiple-choice question (Q->A) and then select the correct rationale for the answer (Q->R).

Since the challenge debuted, AI systems have become better at visual commonsense reasoning, although they still lag far behind human levels of performance (Figure 2.2.8). At the end of 2021, the best mark on VCR stood at 72.0, a score that represents a 63.6% increase in performance since 2018. Although progress has been made since the challenge was launched, improvements have become increasingly marginal, suggesting that new techniques may need to be invented to significantly improve performance.

### VISUAL COMMONSENSE REASONING (VCR) TASK: Q->AR SCORE

Source: VCR Leaderboard, 2021 | Chart: 2022 AI Index Report

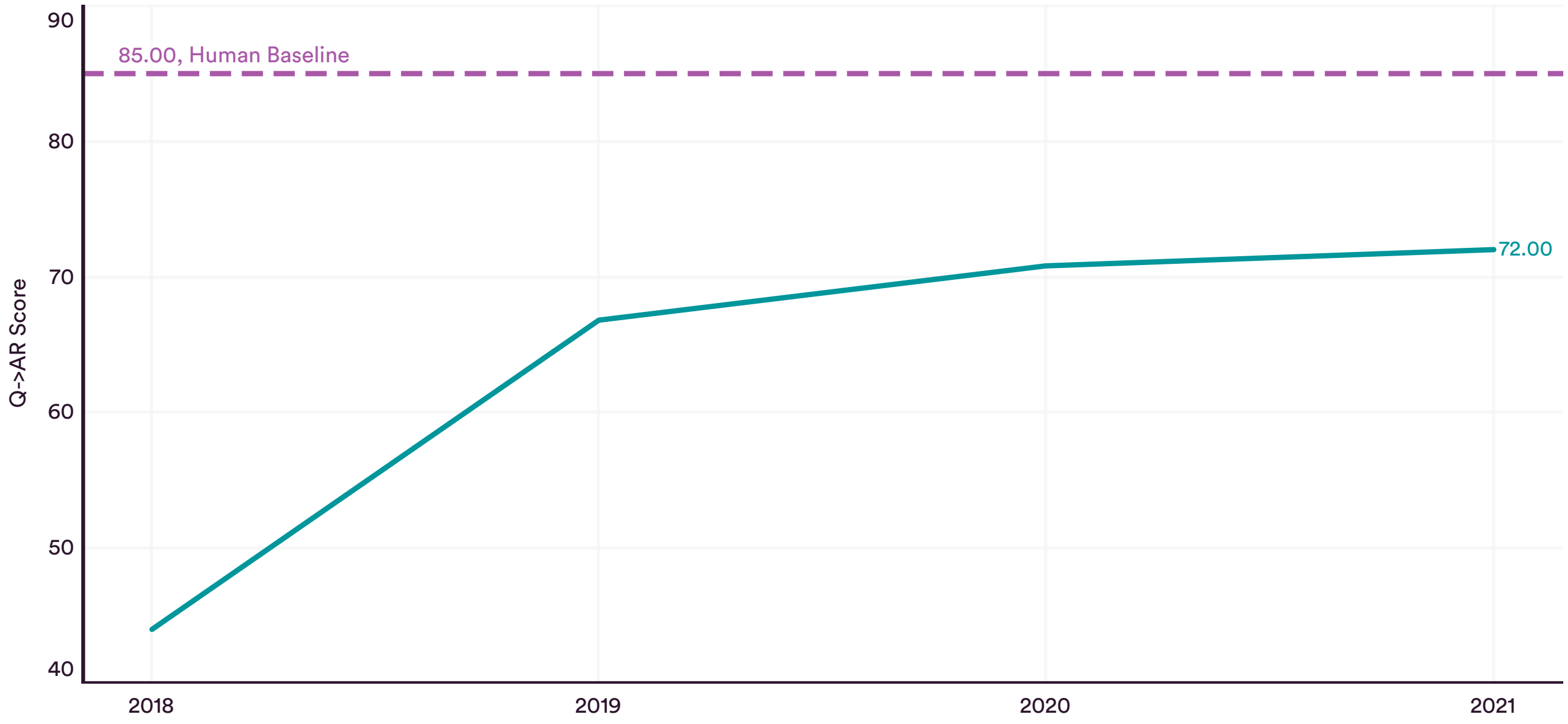

Figure 2.2.8

**Natural language processing (NLP) is a subfield of AI, with roots that stretch back as far as the 1950s. NLP involves research into systems that can read, generate, and reason about natural language. NLP evolved from a set of systems that in its early years used handwritten rules and statistical methodologies to one that now combines computational linguistics, rule-based modeling, statistical learning, and deep learning.**

**This section looks at progress in NLP across several language task domains, including: (1) English language understanding; (2) text summarization; (3) natural language inference; (4) sentiment analysis; and (5) machine translation. In the last decade, technical progress in NLP has been significant: The adoption of deep neural network–style machine learning methods has meant that many AI systems can now execute complex language tasks better than many human baselines.**

# 2.3 LANGUAGE

## ENGLISH LANGUAGE UNDERSTANDING

English language understanding challenges AI systems to understand the English language in various contexts, such as sentence understanding, yes/no reading comprehension, reading comprehension with logical reasoning, etc.

### SuperGLUE

SuperGLUE is a single-number metric that tracks technical progress on a diverse set of linguistic tasks (Figure 2.3.1).

As part of the benchmark, AI systems are tested on eight different tasks (such as answering yes/no questions, identifying causality in events, and doing commonsense reading comprehension), and their performance on these tasks is then averaged into a single score. SuperGLUE is the successor to GLUE, an earlier benchmark that also tests on multiple tasks. SuperGLUE was released in May 2019 after AI systems began to saturate the GLUE metric, creating demand for a harder benchmark.

**A SET OF SUPERGLUE TASKS**[3]
Source: Wang et al., 2019

| | |
|---|---|
| **BoolQ** | **Passage:** *Barq's – Barq's is an American soft drink. Its brand of root beer is notable for having caffeine. Barq's, created by Edward Barq and bottled since the turn of the 20th century, is owned by the Barq family but bottled by the Coca-Cola Company. It was known as Barq's Famous Olde Tyme Root Beer until 2012.*<br>**Question:** *is barq's root beer a pepsi product* **Answer:** `No` |
| **CB** | **Text:** *B: And yet, uh, I we-, I hope to see employer based, you know, helping out. You know, child, uh, care centers at the place of employment and things like that, that will help out. A: Uh-huh. B: What do you think, do you think we are, setting a trend?*<br>**Hypothesis:** *they are setting a trend* **Entailment:** `Unknown` |
| **COPA** | **Premise:** *My body cast a shadow over the grass.* **Question:** *What's the CAUSE for this?*<br>**Alternative 1:** *The sun was rising.* **Alternative 2:** *The grass was cut.*<br>**Correct Alternative:** `1` |

Figure 2.3.1

3 For the sake of brevity, this figure only displays 3 of the 8 tasks.

At the top of the SuperGLUE leaderboard sits the SS-MoE model with a state-of-the-art score of 91.0 (Figure 2.3.2), which exceeds the human performance score of 89.8 given by the SuperGLUE benchmark developers. The fact that progress on SuperGLUE was achieved so rapidly suggests that researchers will need to develop more complex suites of natural language tasks to challenge the next generation of AI systems.

**SUPERGLUE: SCORE**
Source: SuperGLUE Leaderboard, 2021 | Chart: 2022 AI Index Report

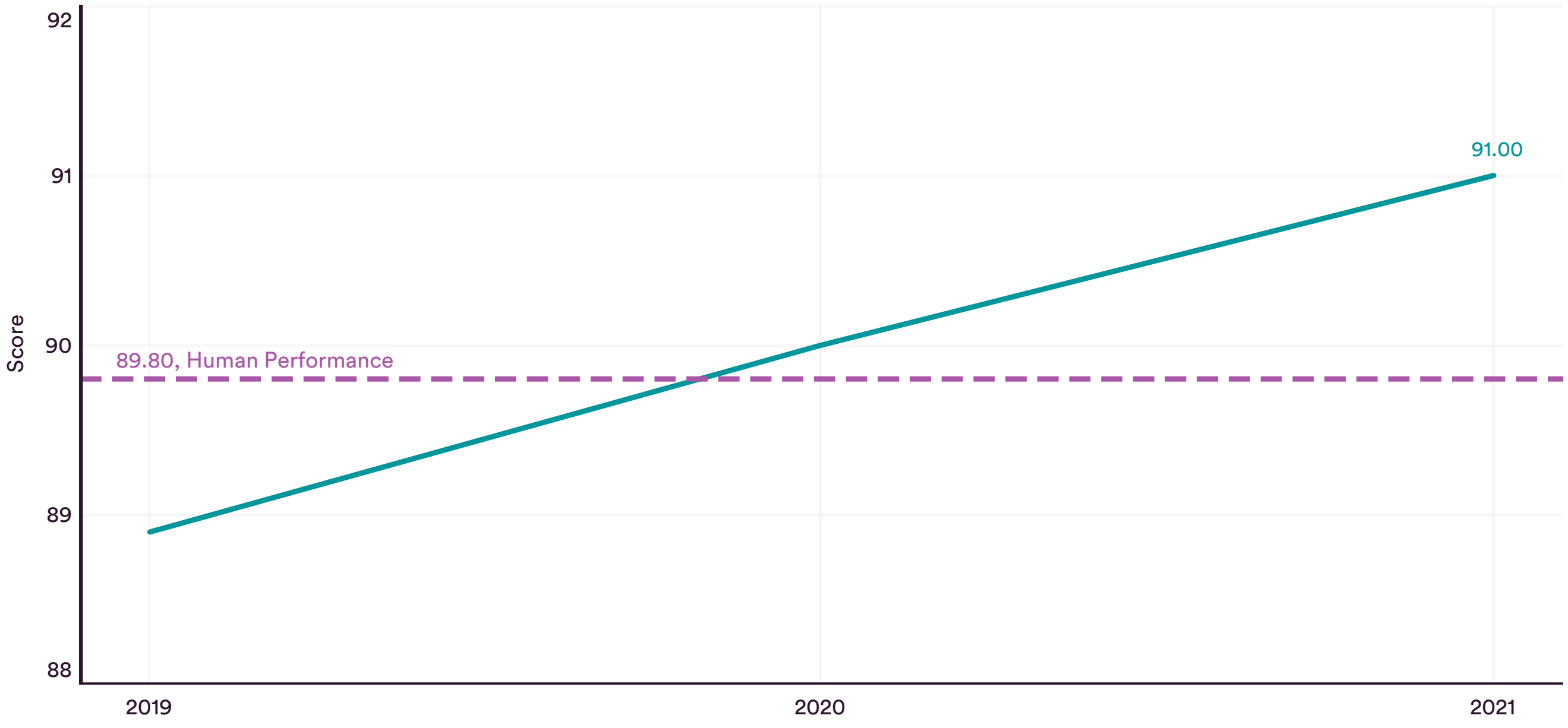

Figure 2.3.2

## Stanford Question Answering Dataset (SQuAD)

The Stanford Question Answering Dataset (SQuAD) benchmarks performance on reading comprehension. The dataset includes 107,785 question-and-answer pairs taken from 536 Wikipedia articles. Performance on SQuAD is measured by the F1 score, which is the average overlap between the AI system's answers and the actual correct answers: The higher the score, the better the performance.

As was the case with GLUE, AI systems improved so rapidly on SQuAD that only two years after launching SQuAD in 2016, researchers released SQuAD 2.0. This second version included more challenging reading comprehension tasks, namely a set of 50,000 unanswerable questions that were written in a way to appear answerable (Figure 2.3.3).

**HARDER QUESTIONS ADDED TO STANFORD QUESTION ANSWERING DATASET (SQUAD) 2.0**
Source: Rajpurkar et al., 2018

**Article:** Endangered Species Act
**Paragraph:** " *. . . Other legislation followed, including the Migratory Bird Conservation Act of 1929, a 1937 treaty prohibiting the hunting of right and gray whales, and the Bald Eagle Protection Act of 1940. These later laws had a low cost to society—the species were relatively rare—and little opposition was raised.*"

**Question 1:** "*Which laws faced significant opposition?*"
**Plausible Answer:** *later laws*

**Question 2:** "*What was the name of the 1937 treaty?*"
**Plausible Answer:** *Bald Eagle Protection Act*

Figure 2.3.3

At the end of 2021, the leading scores on SQuAD 1.1 and SQuAD 2.0 stood at 95.7 and 93.2, respectively (Figure 2.3.4). Although these scores are state of the art, they are marginal improvements over the previous year's top scores (0.4% and 0.2%). Both SQuAD datasets have seen a trend whereby immediately after the initial launches, human-performance-exceeding scores were realized and then followed by small, plateau-like increases.

**SQUAD 1.1 and SQUAD 2.0: F1 SCORE**
Source: SQuAD 1.1 and SQuAD 2.0, 2021 | Chart: 2022 AI Index Report

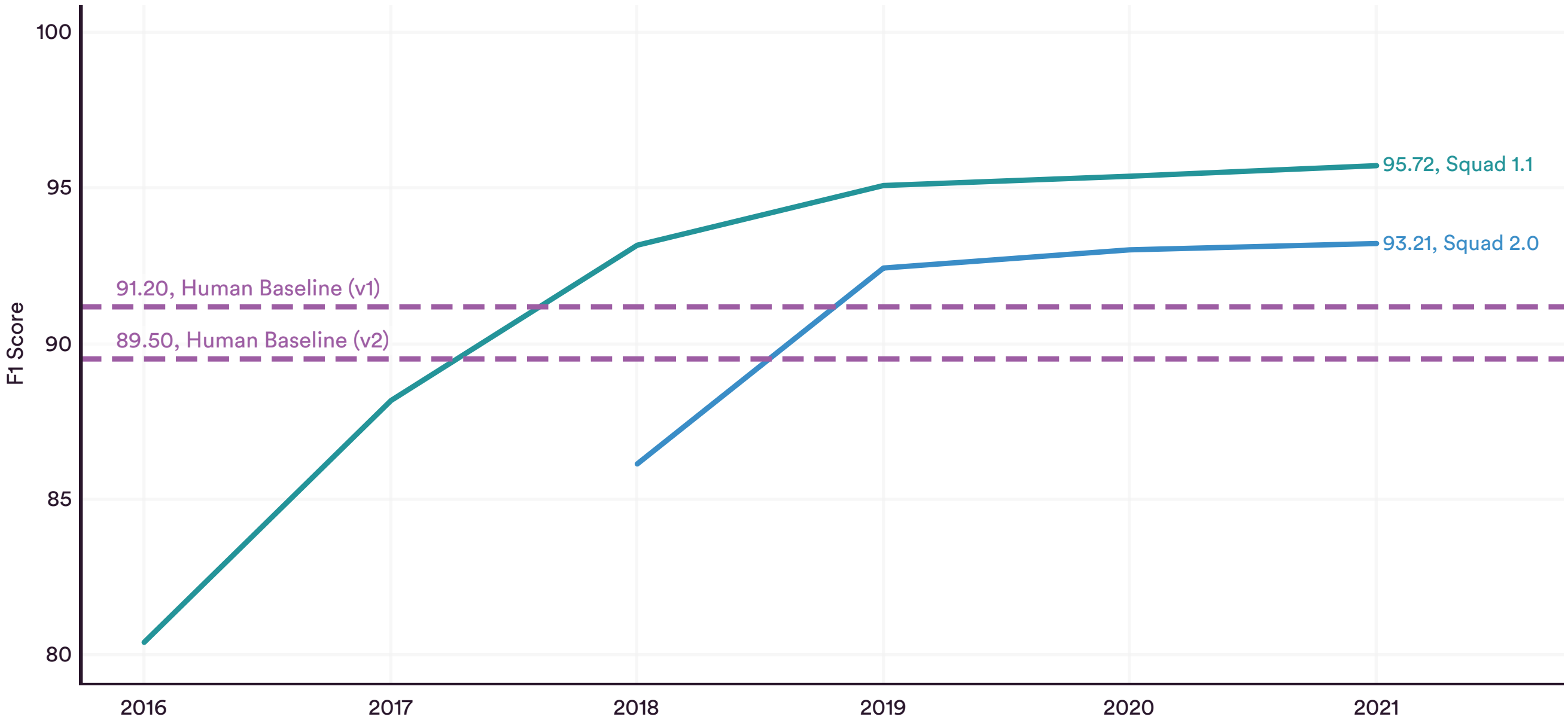

Figure 2.3.4

## Reading Comprehension Dataset Requiring Logical Reasoning (ReClor)

The plateauing progress on benchmarks like SQuAD suggests that NLP models need to be tested on more complex linguistic challenges like those offered by ReClor. Created in 2020 by computer scientists from the National University of Singapore, ReClor requires that AI systems engage in reading comprehension that also necessitates logical reasoning. The ReClor dataset is made up of logical reasoning questions from the LSAT, the entrance exam for law schools in the United States and Canada (Figure 2.3.5).

**A SAMPLE QUESTION IN READING COMPREHENSION DATASET REQUIRING LOGICAL REASONING (RECLOR)**
Source: Yu et al., 2020

**Context:**
In jurisdictions where use of headlights is optional when visibility is good, drivers who use headlights at all times are less likely to be involved in a collision than are drivers who use headlights only when visibility is poor. Yet Highway Safety Department records show that making use of headlights mandatory at all times does nothing to reduce the overall number of collisions.
**Question:** Which one of the following, if true, most helps to resolve the apparent discrepancy in the information above?
**Options:**
A. In jurisdictions where use of headlights is optional when visibility is good, one driver in four uses headlights for daytime driving in good weather.
B. Only very careful drivers use headlights when their use is not legally required.
C. The jurisdictions where use of headlights is mandatory at all times are those where daytime visibility is frequently poor.
D. A law making use of headlights mandatory at all times is not especially difficult to enforce.
**Answer:** B

Figure 2.3.5

There are two sets of questions on ReClor, easy and hard, with AI systems being judged on accuracy based on the percentage of questions they answer correctly (Figure 2.3.6). Although AI systems are presently capable of achieving a relatively high level of performance on the easy set of questions, they struggle on the hard set. In 2021, the top-performing model on ReClor (hard set) scored 69.3%, roughly 22.5 percentage points lower than the top-performing model on the easy set. Datasets like ReClor suggest that while NLP models can execute straightforward reading comprehension tasks, they face more difficulty when those tasks are coupled with logical reasoning requirements.

**READING COMPREHENSION DATASET REQUIRING LOGICAL REASONING (RECLOR): ACCURACY**

Source: ReClor Leaderboard, 2021 | Chart: 2022 AI Index Report

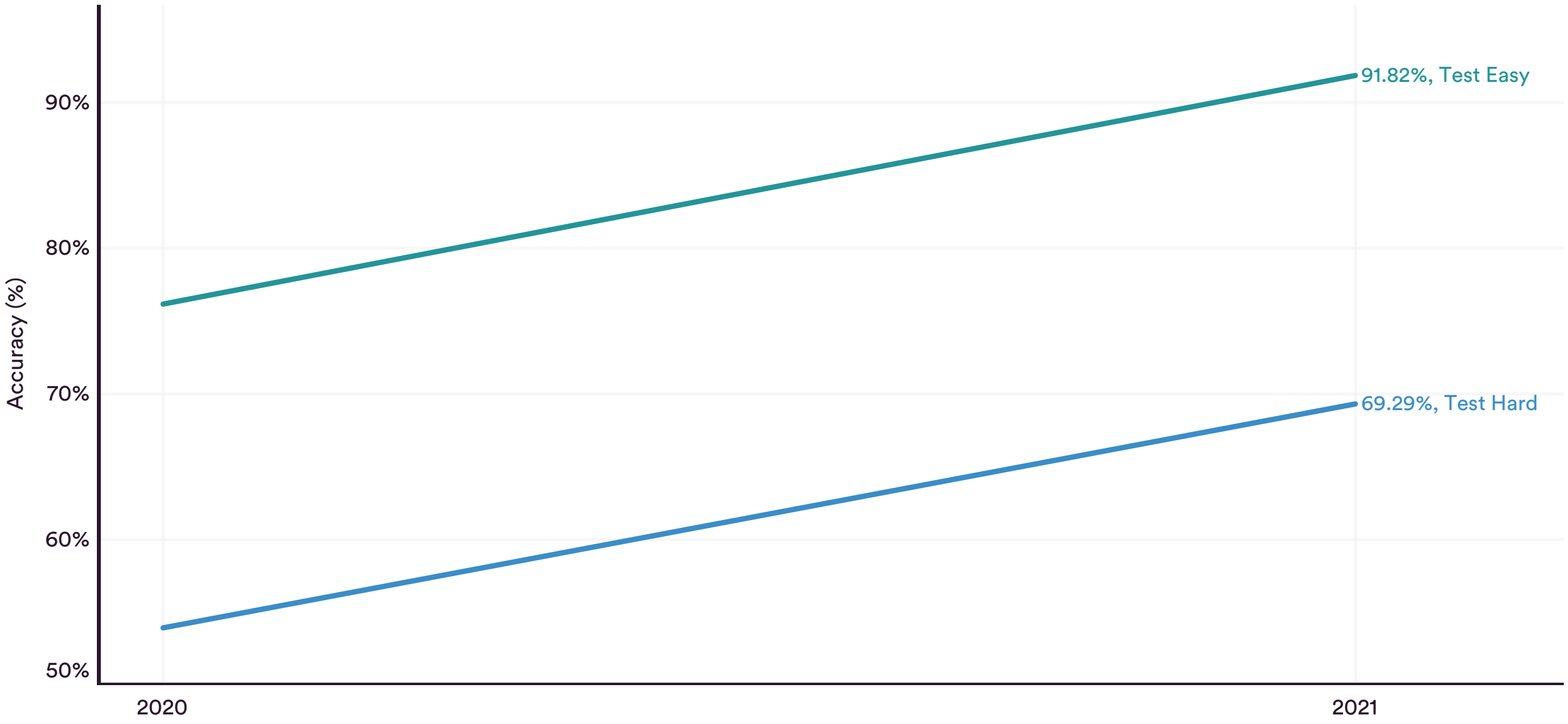

Figure 2.3.6

## TEXT SUMMARIZATION

Text summarization is the challenge of synthesizing a piece of text while capturing its core content. Summarizing texts is an important component of text classification, reading comprehension, and information dissemination; however, when done manually by humans, it is time- and labor-intensive. Developing AI systems that can functionally summarize texts has a number of practical use cases, from aiding universities in classifying academic papers to helping lawyers generate case summaries.

Progress in text summarization is often scored on ROUGE (Recall-Oriented Understudy for Gisting Evaluation). ROUGE calculates the overlap between a summary produced by an AI system and the reference summary produced by a human. The higher the ROUGE score, the greater the overlap and the more accurate the summary.

### arXiv

ArXiv is a text summarization benchmark dataset that contains over 27,770 different papers from arXiv, the open-access repository of scientific papers. In the five years since benchmarking on arXiv began, AI text summarization models have improved their performance by 47.1% (Figure 2.3.7). However, as is the case with other natural language benchmarks, progress seems to be plateauing.

**ARXIV: ROUGE-1**

Source: Papers with Code, 2021; arXiv, 2021 | Chart: 2022 AI Index Report

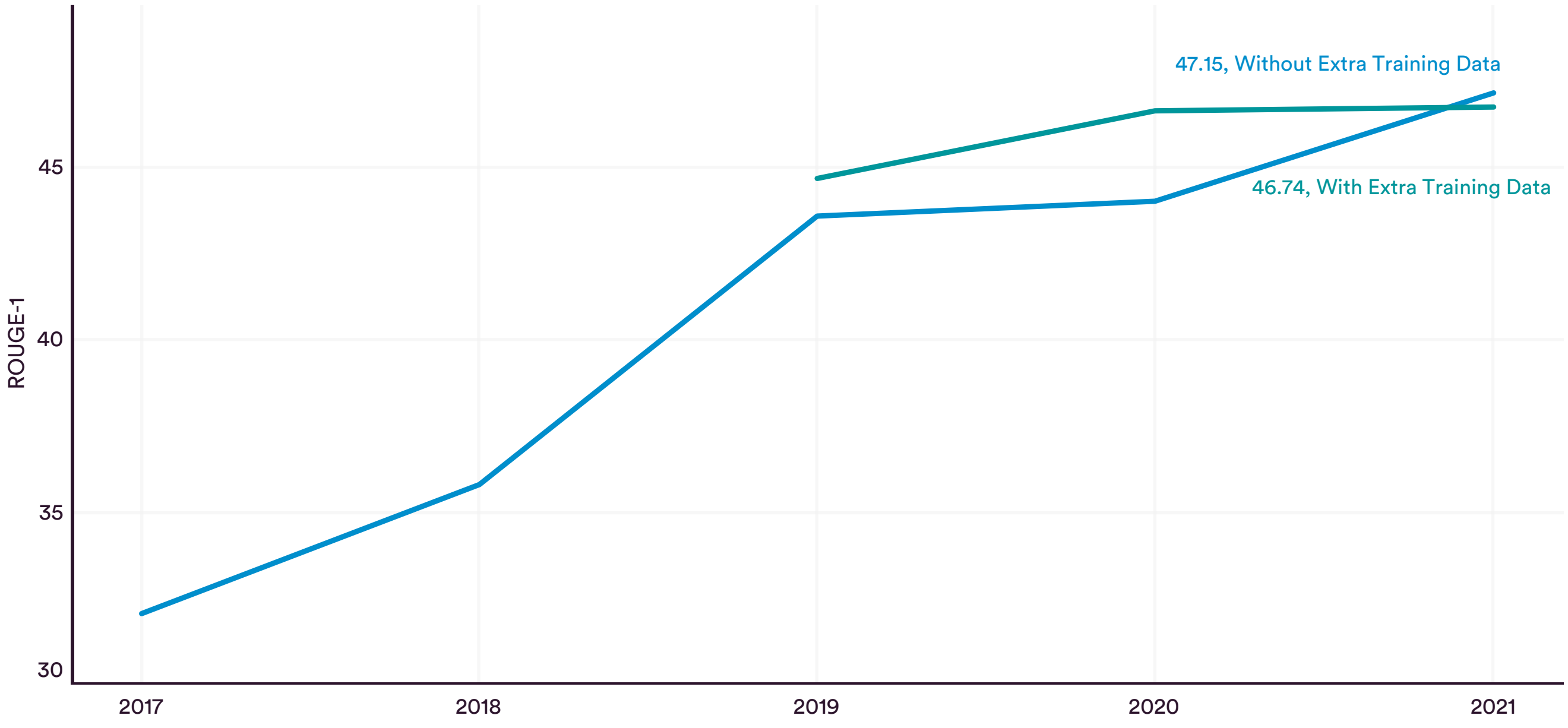

Figure 2.3.7

## PubMed

PubMed is a text summarization dataset consisting of 19,717 scientific publications from the PubMed database of scientific papers. Progress on PubMed validates the trend seen on arXiv: There has been significant improvement on text classification since 2017 (34.6%), but recently the pace of that progress has slowed (Figure 2.3.8). In 2021, the top-performing model on PubMed was HAT (hierarchical attention transformer model), created by researchers at Birch AI and the University of Washington.

**PUBMED: ROUGE-1**

Source: Papers with Code, 2021; arXiv, 2021 | Chart: 2022 AI Index Report

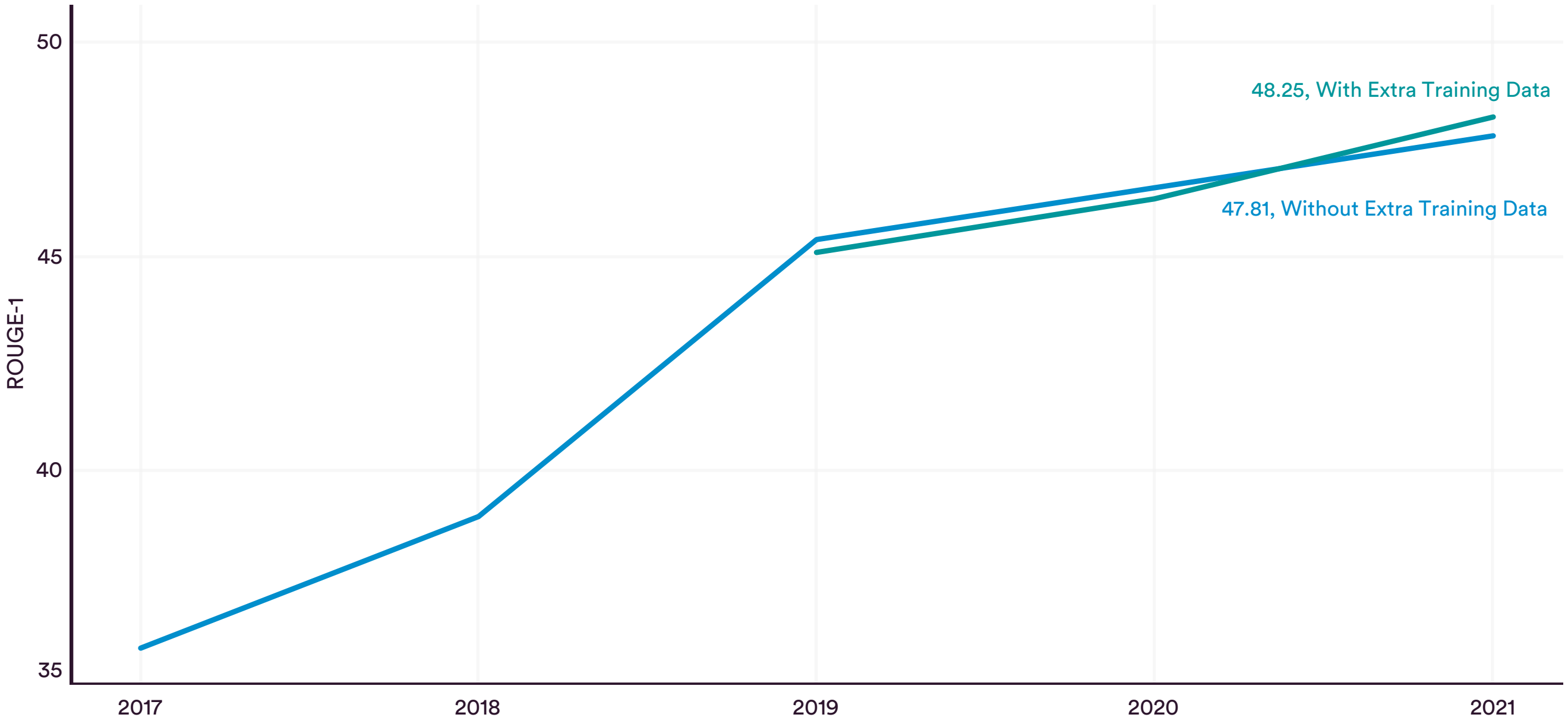

Figure 2.3.8

## NATURAL LANGUAGE INFERENCE

Natural language inference is the task of determining whether, given a premise, a hypothesis is true (entailment), false (contradiction), or undetermined (neutral). This skill is also known as textual entailment as it requires determining whether a particular premise logically entails a hypothesis. Natural language inference necessitates language-processing skills, such as named entity recognition (understanding the words you see), as well as being able to use commonsense knowledge to distinguish between reasonable and unreasonable inferences.

### Stanford Natural Language Inference (SNLI)

The Stanford Natural Language Inference (SNLI) dataset contains around 600,000 sentence pairs (premise and associated hypothesis) that are labeled as either entailment, contradiction, or neutral. As part of this challenge, AI systems are asked whether premises logically entail certain hypotheses (Figure 2.3.9). Performance on SNLI is measured in accuracy based on the percentage of questions answered correctly.

**QUESTIONS AND LABELS IN STANFORD NATURAL LANGUAGE INFERENCE (SNLI)**
Source: Bowman et al., 2015

| | | |
|---|---|---|
| A man inspects the uniform of a figure in some East Asian country. | **contradiction**<br>C C C C C | The man is sleeping |
| An older and younger man smiling. | **neutral**<br>N N E N N | Two men are smiling and laughing at the cats playing on the floor. |
| A black race car starts up in front of a crowd of people. | **contradiction**<br>C C C C C | A man is driving down a lonely road. |
| A soccer game with multiple males playing. | **entailment**<br>E E E E E | Some men are playing a sport. |
| A smiling costumed woman is holding an umbrella. | **neutral**<br>N N E C N | A happy woman in a fairy costume holds an umbrella. |

Figure 2.3.9

The top-performing model on SNLI is Facebook AI USA's EFL, which in April 2021 posted a score of 93.1% (Figure 2.3.10).

**STANFORD NATURAL LANGUAGE INFERENCE (SNLI): ACCURACY**
Source: Papers with Code, 2021; arXiv, 2021 | Chart: 2022 AI Index Report

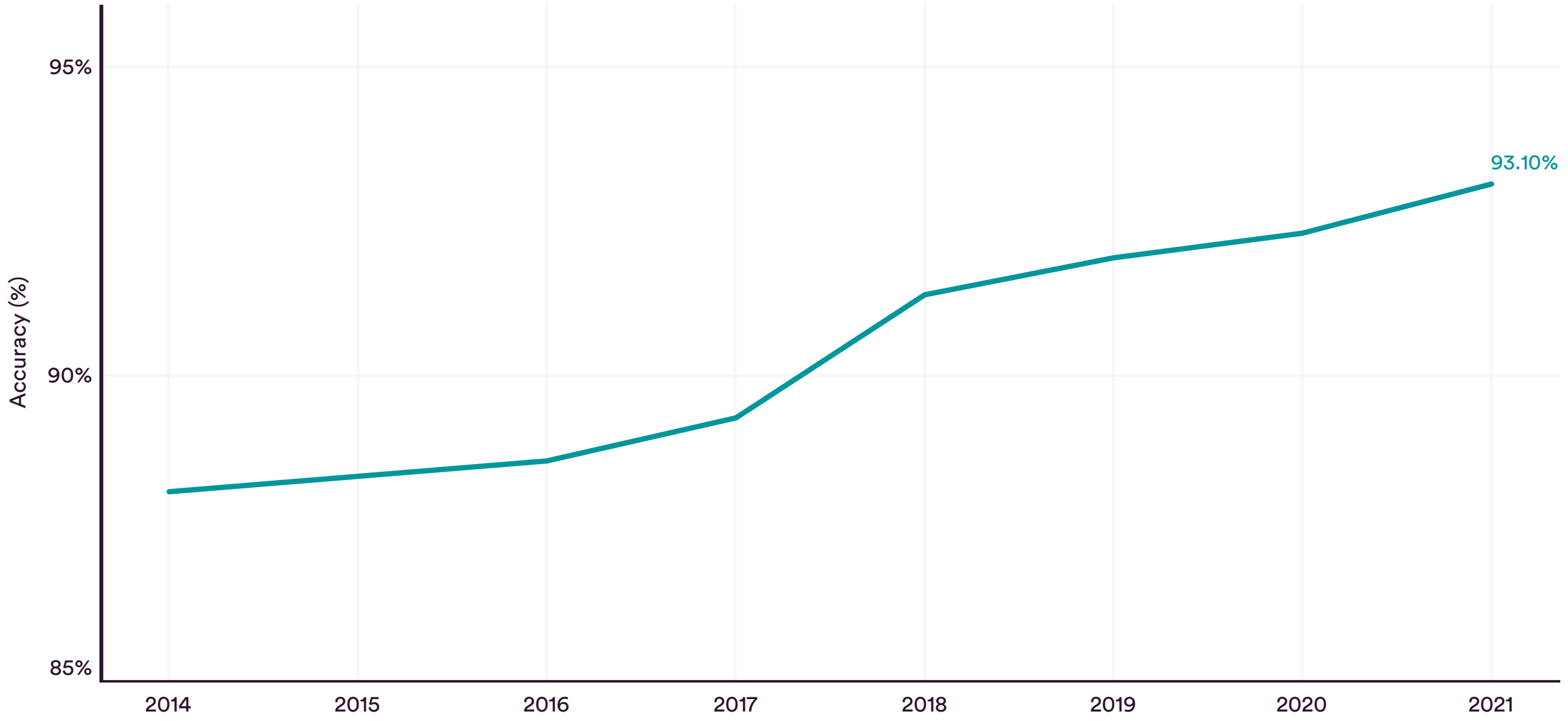

Figure 2.3.10

## Abductive Natural Language Inference (aNLI)

Abductive natural language inference is a more difficult type of textual entailment. Abductive inference requires drawing the most plausible conclusion from a context of limited information and uncertain premises. For instance, if Jenny were to return from work and find her home in a disheveled state and then recall that she left a window open, she can plausibly infer that a burglar broke in and caused the mess.[4] Although abduction is regarded as an essential element in how humans communicate with one another, few studies have attempted to study the abductive ability of AI systems.

ANLI, a new benchmark for abductive natural language inference created in 2019 by the Allen Institute for AI, comes with 170,000 premise and hypothesis pairs. Figure 2.3.11 illustrates the types of statements included in the dataset.

**EXAMPLE QUESTIONS IN ABDUCTIVE NATURAL LANGUAGE INFERENCE (ANLI)**
Source: Allen Institute for AI, 2021

*Obs1:* It was a gorgeous day outside.

*Obs2:* She asked her neighbor for a jump-start.

*Hyp1:* **Mary decided to drive to the beach, but her car would not start due to a dead battery.**

*Hyp2:* It made a weird sound upon starting.

Figure 2.3.11

4 This particular example of abductive commonsense reasoning is taken from Bhagavatula et al. ( 2019), the first paper that investigates the ability of AI systems to perform language-based abductive reasoning.

AI performance on abductive commonsense reasoning has increased by 7.7 percentage points since 2019; however, the top AI systems, while close, are unable to achieve human performance levels (Figure 2.3.12). Abductive reasoning is therefore still a challenging linguistic task for AI systems.

**ABDUCTIVE NATURAL LANGUAGE INFERENCE (aNLI): ACCURACY**
Source: Allen Institute for AI, 2021 | Chart: 2022 AI Index Report

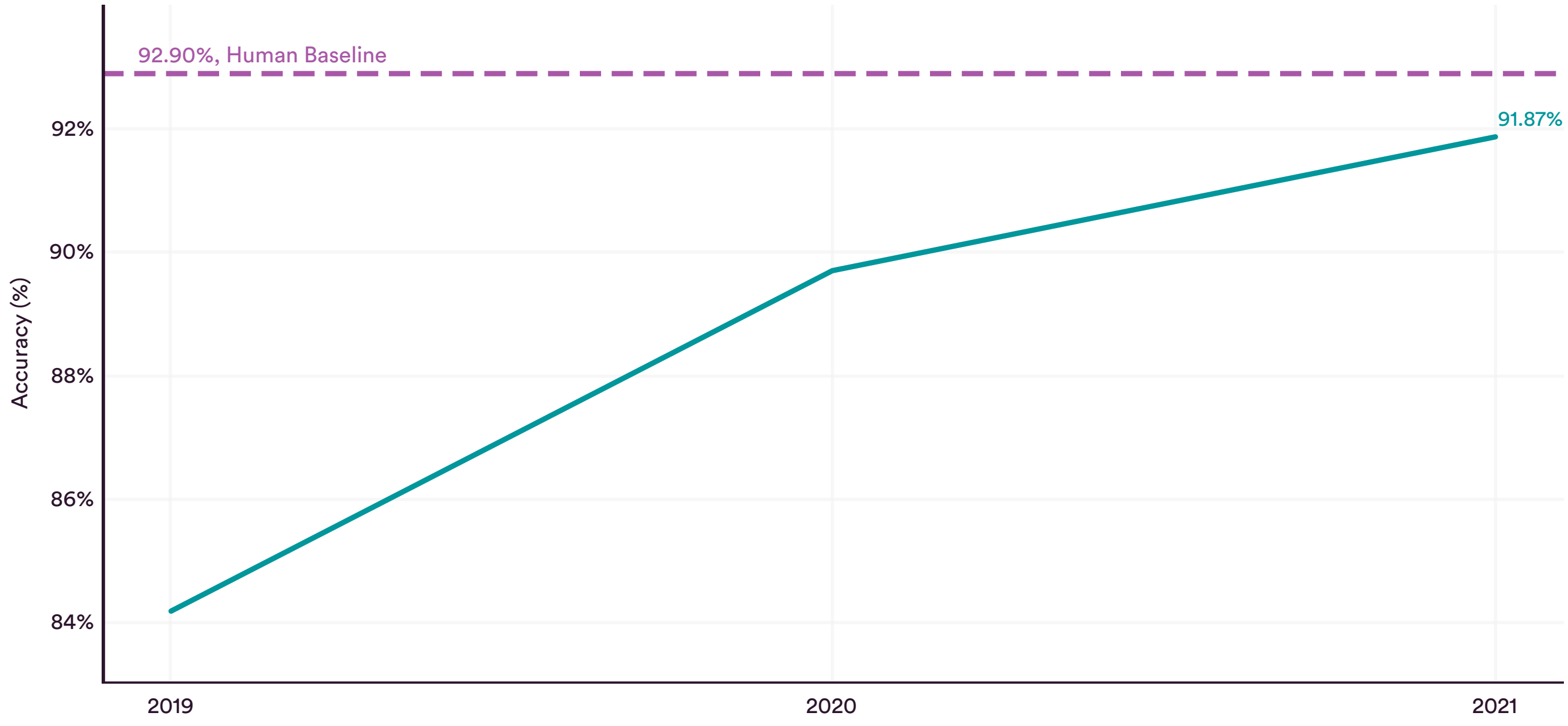

Figure 2.3.12

## SENTIMENT ANALYSIS

Sentiment analysis is the task of using NLP techniques to identify the sentiment (very negative, negative, neutral, positive, very positive) of a given text. Sentiment analysis can be straightforward if sentences are worded clearly and unambiguously, such as "I dislike winter weather." However, sentiment analysis can become more challenging when AI systems encounter sentences with flipped structures or negations, such as "to say that disliking winter weather is not really my thing is completely inaccurate."

Sentiment analysis has many commercial use cases, from parsing customer reviews and field survey responses to identifying the emotional states of customers.

### SemEval 2014 Task 4 Sub Task 2

The SemEval 2014 Task 4 Sub Task 2 is a benchmark for sentiment analysis that asks machines to engage in sentiment analysis. This specific task tests whether AI systems can identify the sentiment associated with particular aspects of a text, rather than the sentiment of entire sentences or paragraphs (Figure 2.3.13).

**A SAMPLE SEMEVAL TASK**
Source: Pontiki et al., 2014

For example:

"I loved their **fajitas**" → {fajitas: *positive*}
"I hated their **fajitas**, but their **salads** were great" → {fajitas: *negative*, salads: *positive*}
"The **fajitas** are their first plate" → {fajitas: *neutral*}
"The **fajitas** were great to taste, but not to see" → {fajitas: *conflict*}

Figure 2.3.13

The SemEval dataset is composed of 7,686 restaurant and laptop reviews, whose emotional polarities have been rated by humans. On SemEval, AI systems are tasked with assigning the right sentiment labels to particular components of the text, with their performance measured in terms of the percentage of the labels they correctly assign.

In the past seven years, AI systems have become much better at sentiment analysis. As of last year, top-performing systems estimate sentiment correctly 9 out of 10 times, whereas in 2016, they made correct estimates only 7 out of 10 times. As of 2021, the state-of-the-art scores on SemEval stood at 88.6%, realized by a team of Chinese researchers from South China Normal University and Linklogis Co. Ltd. (Figure 2.3.14).

**SEMEVAL 2014 TASK 4 SUB TASK 2: ACCURACY**

Source: Papers with Code, 2021; arXiv, 2021 | Chart: 2022 AI Index Report

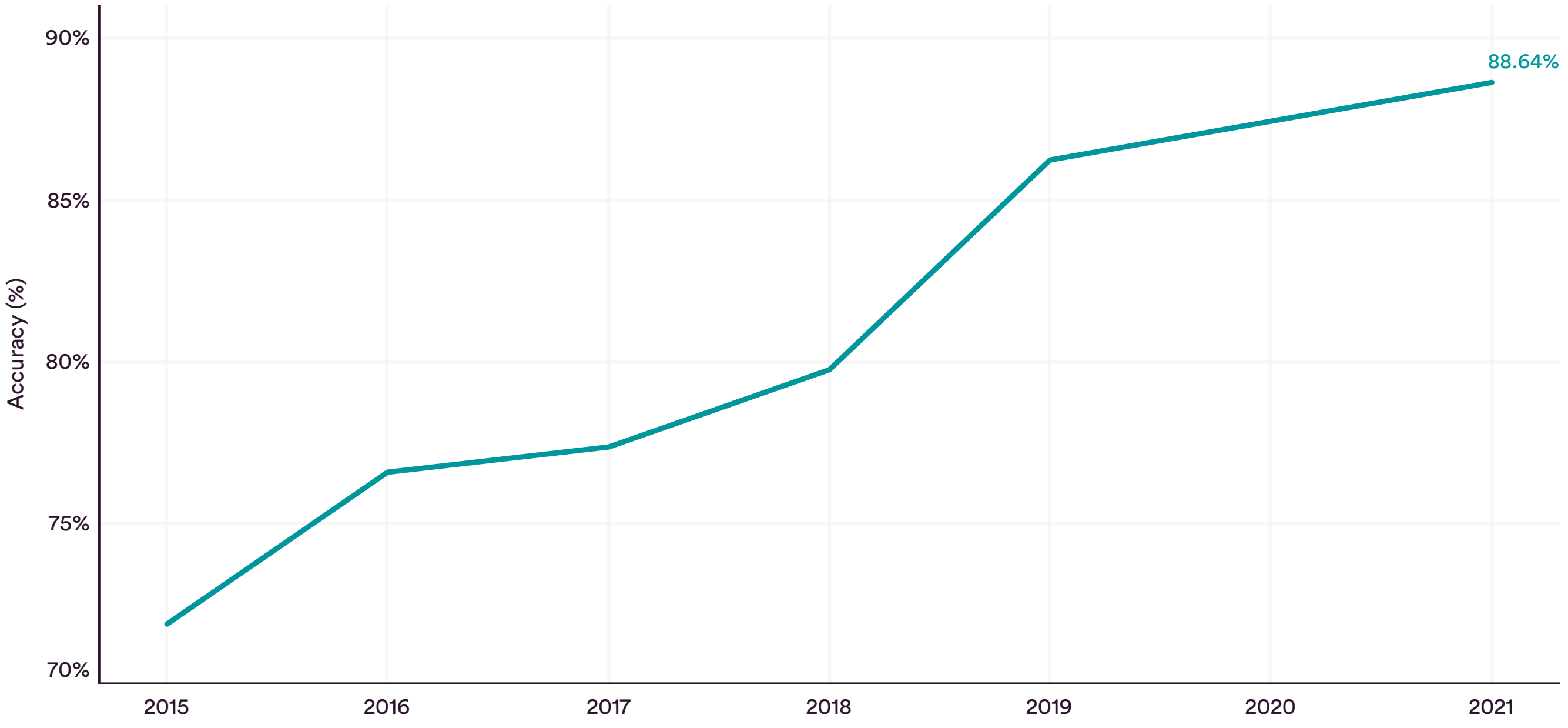

Figure 2.3.14

## MACHINE TRANSLATION (MT)

Machine translation examines how AI software can translate languages. In domains where fluency in multiple languages is required, machine translation can be extremely impactful. The European Union, for example, is required to translate all its cross-national policy documents into the 24 languages of its member states. Using machine translators can save time, improve efficiency, and lead to more consistent outcomes.

Since 2017, neural networks have taken over machine translation. Unlike their predecessors, neural translators learn from a series of prior translation tasks and predict the likelihood of a sequence of words. Neural translation models have revolutionized the field of machine translation not only because they do not require human supervision, but also because they produce the most accurate translations. As a result, they have been widely deployed by search engines and social networks.

## WMT 2014, English-German and English-French

The WMT 2014 family of datasets, first introduced at the Meeting of the Association for Computational Linguistics (ACL) 2014, consist of different kinds of translation tasks, including translation between English-French and English-German language pairs. A machine's translation capabilities are measured by the Bilingual Evaluation Understudy, or BLEU, score, which compares the extent to which a machine-translated text matches a reference human-generated translation. The higher the score, the better the translation.

Both the English-French and English-German WMT 2014 benchmarks showcase the significant progress made in AI machine translation over the last decade (Figure 2.3.15). Since submissions began, there has been a 23.7% improvement in English-French and a 68.1% improvement in English-German translation ability. Relatively speaking, although performance improvements have been more significant on the English-German language pair, absolute translation ability remains meaningfully higher on English-French translation.

**WMT2014, ENGLISH-FRENCH: BLEU SCORE**

Source: Papers with Code, 2021; arXiv, 2021 | Chart: 2022 AI Index Report

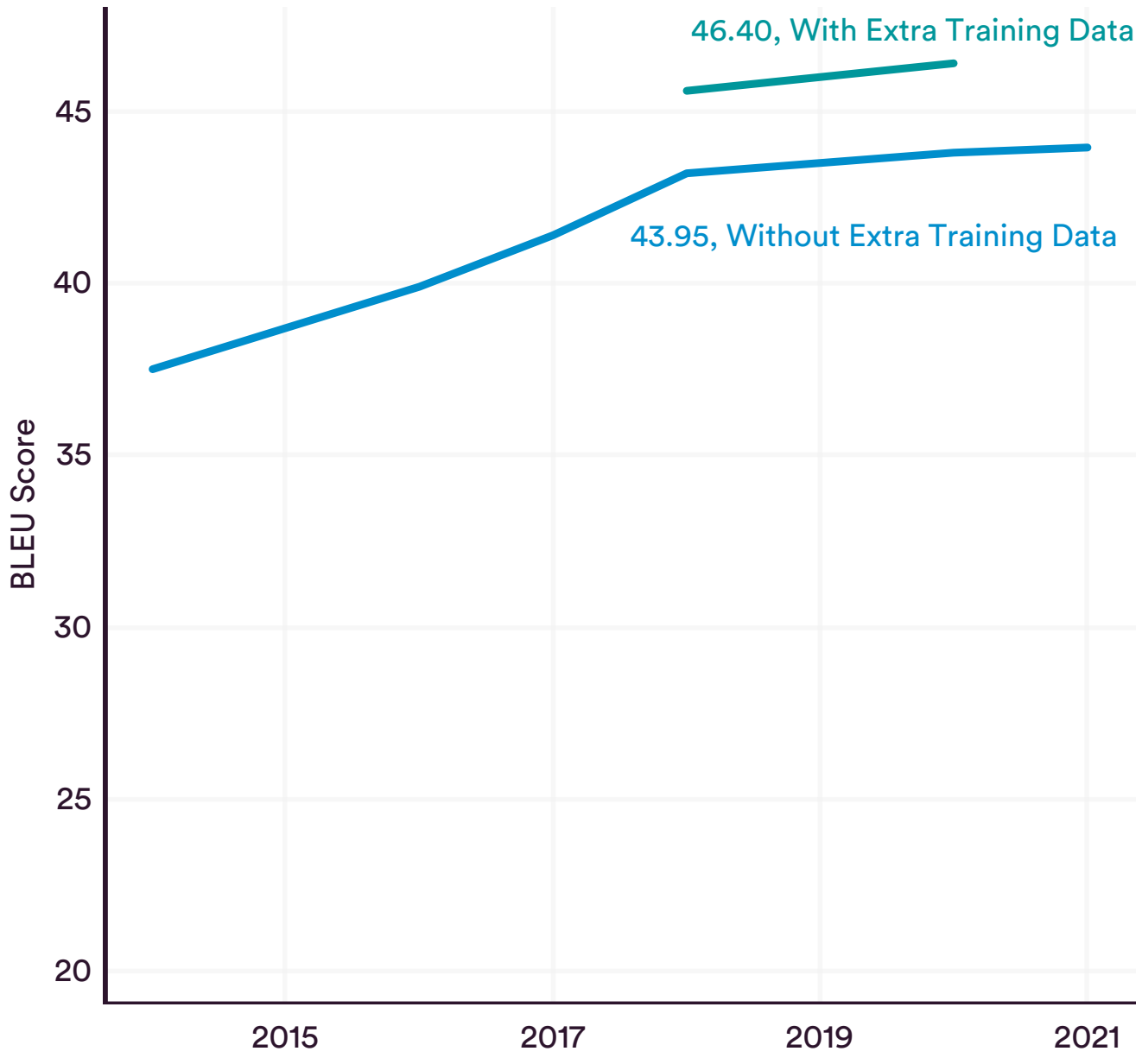

**WMT2014, ENGLISH-GERMAN: BLEU SCORE**

Source: Papers with Code, 2021; arXiv, 2021 | Chart: 2022 AI Index Report

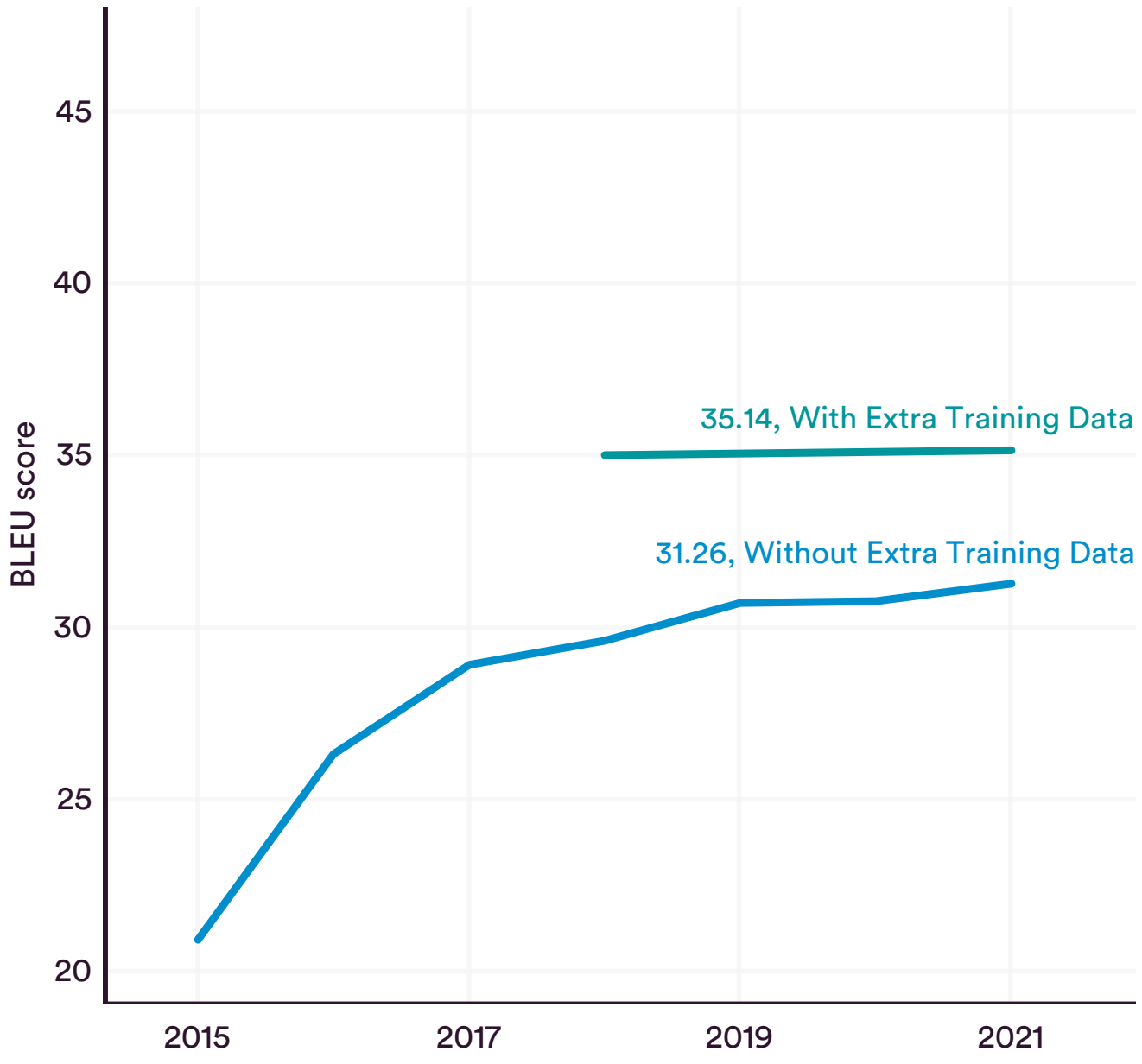

Figure 2.3.15

### Number of Commercially Available MT Systems

The growing interest in machine translation is also reflected in the rise of commercial machine translation services such as Google Translate. Since 2017, there has been a nearly fivefold increase in the number of commercial machine translators on the market, according to Intento (Figure 2.3.16). 2021 also saw the introduction of three open-source machine translation services (M2M-100, mBART, and OPUS). The emergence of publicly available, high-functioning machine translation services speaks to the increasing accessibility of such services and bodes well for anybody who routinely relies on translation.

NUMBER of INDEPENDENT MACHINE TRANSLATION SERVICES
Source: Intento, 2021 | Chart: 2022 AI Index Report

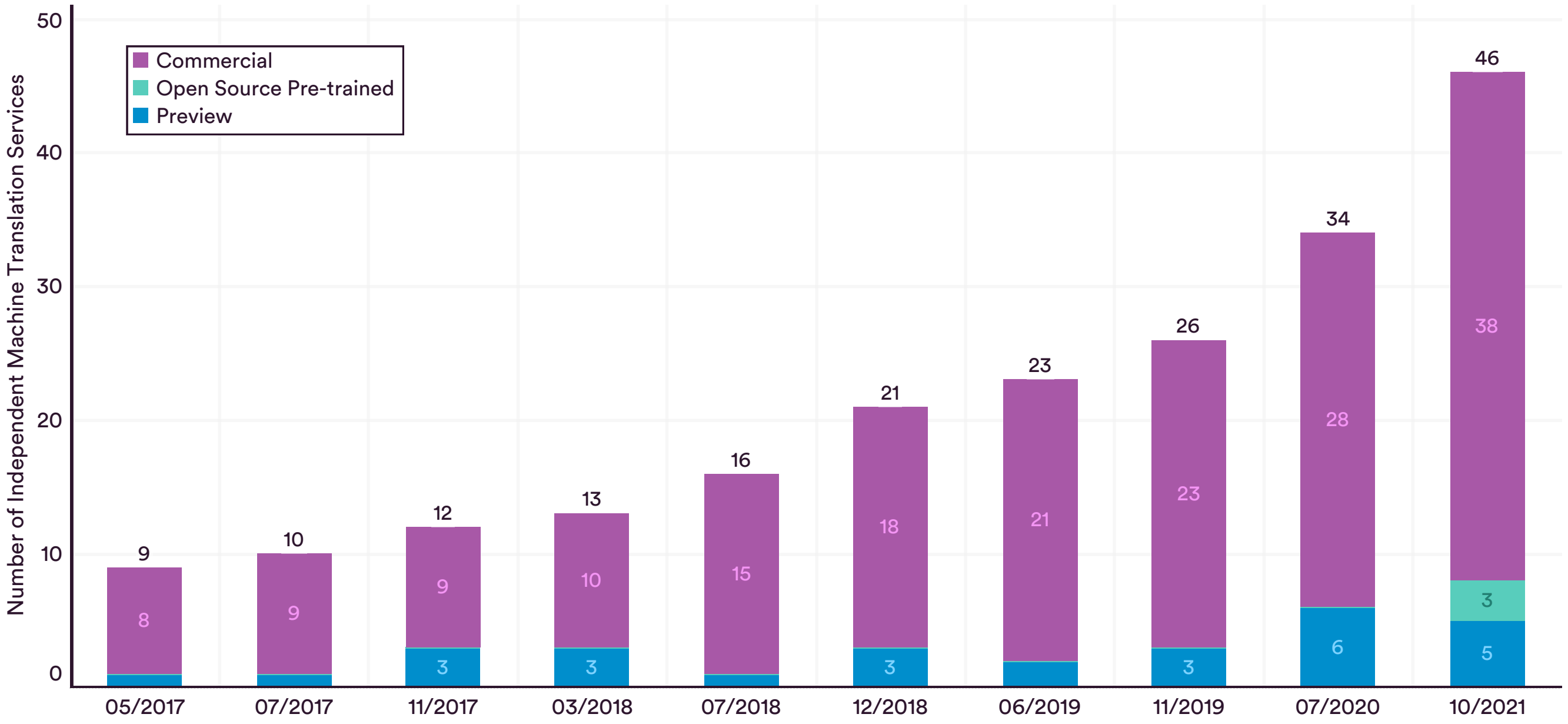

Figure 2.3.16

**Another important domain of AI research is the analysis, recognition, and synthesis of human speech. In this AI subfield, AI systems are typically rated on their ability to recognize speech and identify words and convert them into text; and also to recognize speakers and identify the individuals speaking. Modern home assistance tools, such as Siri, are one of the many examples of commercially applied AI speech technology.**

# 2.4 SPEECH

## SPEECH RECOGNITION

Speech recognition is the process of training machines to recognize spoken words and convert them into text. Research in this domain began at Bell Labs in the 1950s, when the world was introduced to the automatic digit recognition machine (named "Audrey"), which could recognize a human saying any number from zero to nine. Speech recognition has come a long way since then and in the last decade has benefited tremendously from deep-learning techniques and the availability of rich speech recognition datasets.

### Transcribe Speech: LibriSpeech (Test-Clean and Other Datasets)

Introduced in 2015, LibriSpeech is a speech transcription database that contains around 1,000 hours of 16 khz English speech taken from a collection of audiobooks. On LibriSpeech, AI systems are asked to transcribe speech to text and then measured on word error rate, or the percentage of words they fail to correctly transcribe.

LibriSpeech is subdivided into two datasets. First, there is LibriSpeech Test Clean, which includes higher-quality recordings. Performance on Test Clean suggests how well AI systems can transcribe speech in ideal conditions. Second, there is LibriSpeech Test Other, which includes lower-quality recordings. Performance on Test Other is indicative of transcription performance in environments where sound quality is less than ideal.

AI systems perform incredibly well on LibriSpeech, so much so that progress appears to be plateauing (Figure 2.4.1). A state-of-the-art result on the Test Clean dataset was not realized in 2021, which speaks to the fact that the error rate of the top system was already low at 1.4%. For every 100 words that top-performing transcription models heard, they correctly transcribed 99.

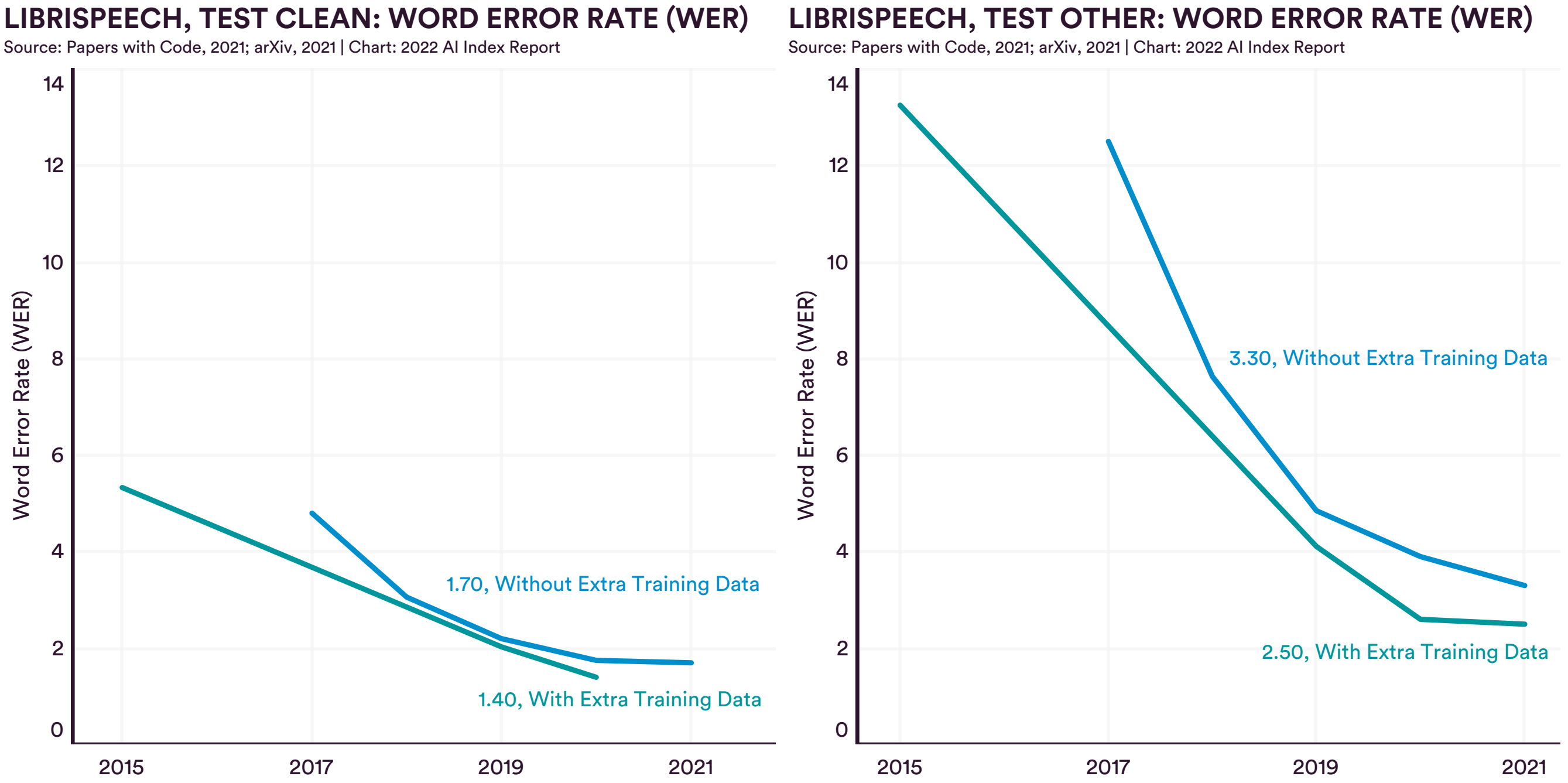

Figure 2.4.1

Performance on the Test Other dataset, while worse than Test Clean, was still relatively poor. The state-of-the-art results on Test Other were realized by the W2V-BERT model, an MIT and Google Brain collaboration, which posted an error rate of 2.0%.

## VoxCeleb

VoxCeleb is a large-scale audiovisual dataset of human speech for speaker recognition, which is the task of matching certain speech with a particular individual. Each year, the makers of VoxCeleb host a speaker verification challenge. A low score or equal error rate on the VoxCeleb challenge is indicative of an AI system that makes few errors in its attribution of speech to particular individuals.[5] Figure 2.4.2 plots performance over time on VoxCeleb-1, the original VoxCeleb dataset. Since 2017, performance on VoxCeleb has improved: Systems that once reported equal error rates of 7.8% now report errors that are less than 1.0%.

**VOXCELEB: EQUAL ERROR RATE (EER)**
Source: VoxCeleb, 2021 | Chart: 2022 AI Index Report

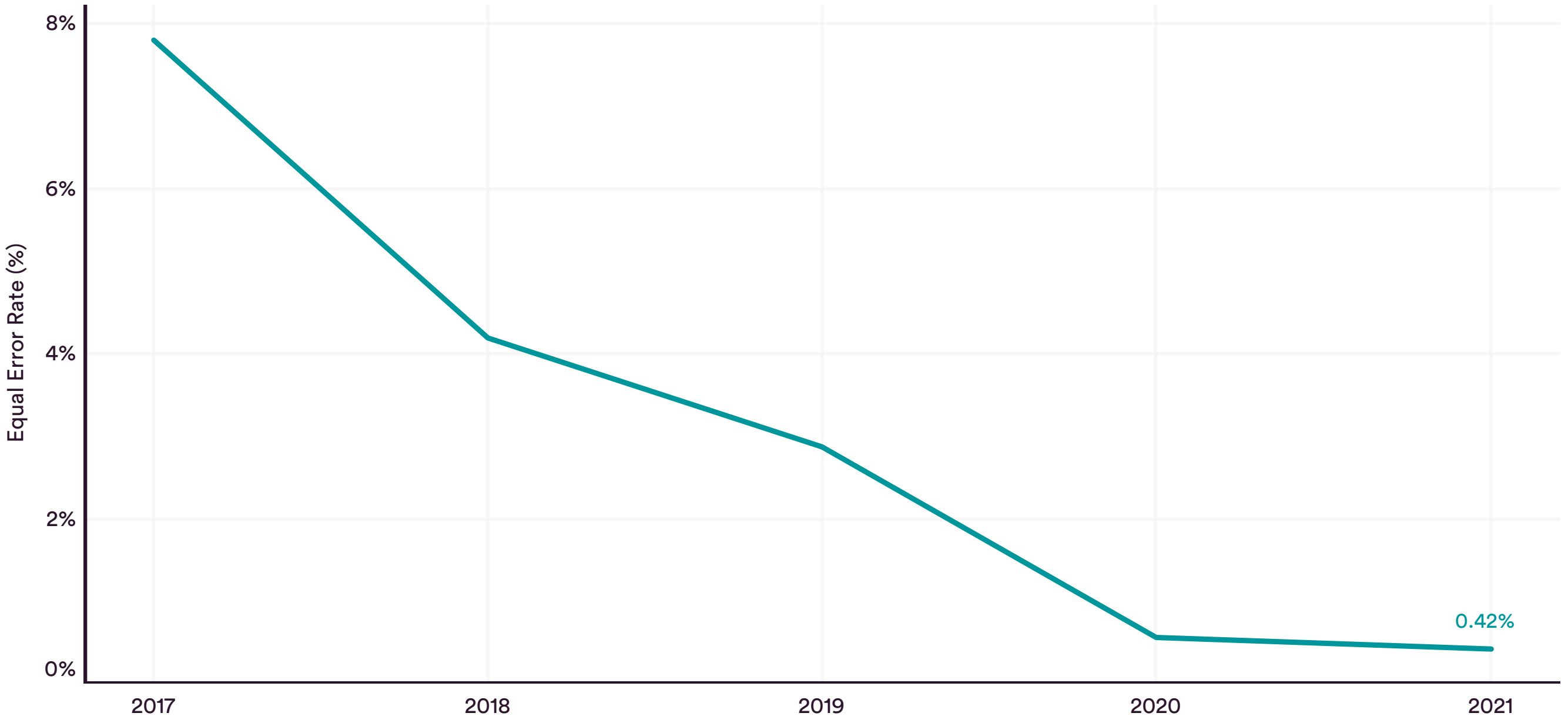

Figure 2.4.2

5 The Equal Error Rate (EER) is not only a measure of the false positive rate, assigning a bad label, but also the false negative rate (failure to assign the correct label).

**Recommendation is the task of suggesting items that might be of interest to a user, such as movies to watch, articles to read, or products to purchase. Recommendation systems are crucial to businesses, such as Amazon, Netflix, Spotify, and YouTube. For example, one of the earliest open recommendation competitions in AI was the Netflix Prize; hosted in 2009, it challenged computer scientists to develop algorithms that could accurately predict user ratings for films based on previously submitted ratings.**

# 2.5 RECOMMENDATION

## Commercial Recommendation: MovieLens 20M

The MovieLens 20M dataset contains around 20 million movie ratings for 27,000 movies from 138,000 users. The ratings are taken from MovieLens (a movie recommendation platform), and AI systems are challenged to see if they can predict a user's movie preferences based on their previously submitted ratings. The metric used to track performance on MovieLens is Normalized Discounted Cumulative Gain (nDCG), which is a measure of ranking quality. A higher nDCG score means that an AI system delivers more accurate recommendations.

Since 2018, top models now perform roughly 5.2% better on MovieLens 20M than they did in 2018 (Figure 2.5.1). In 2021, the state-of-the-art system on MovieLens 20M posted an nDCG of 0.448 and came from researchers at the Czech Technical University in Prague.

**MOVIELENS 20M: NORMALIZED DISCOUNTED CUMULATIVE GAIN@100 (nDCG@100)**
Source: Papers with Code, 2021; arXiv, 2021 | Chart: 2022 AI Index Report

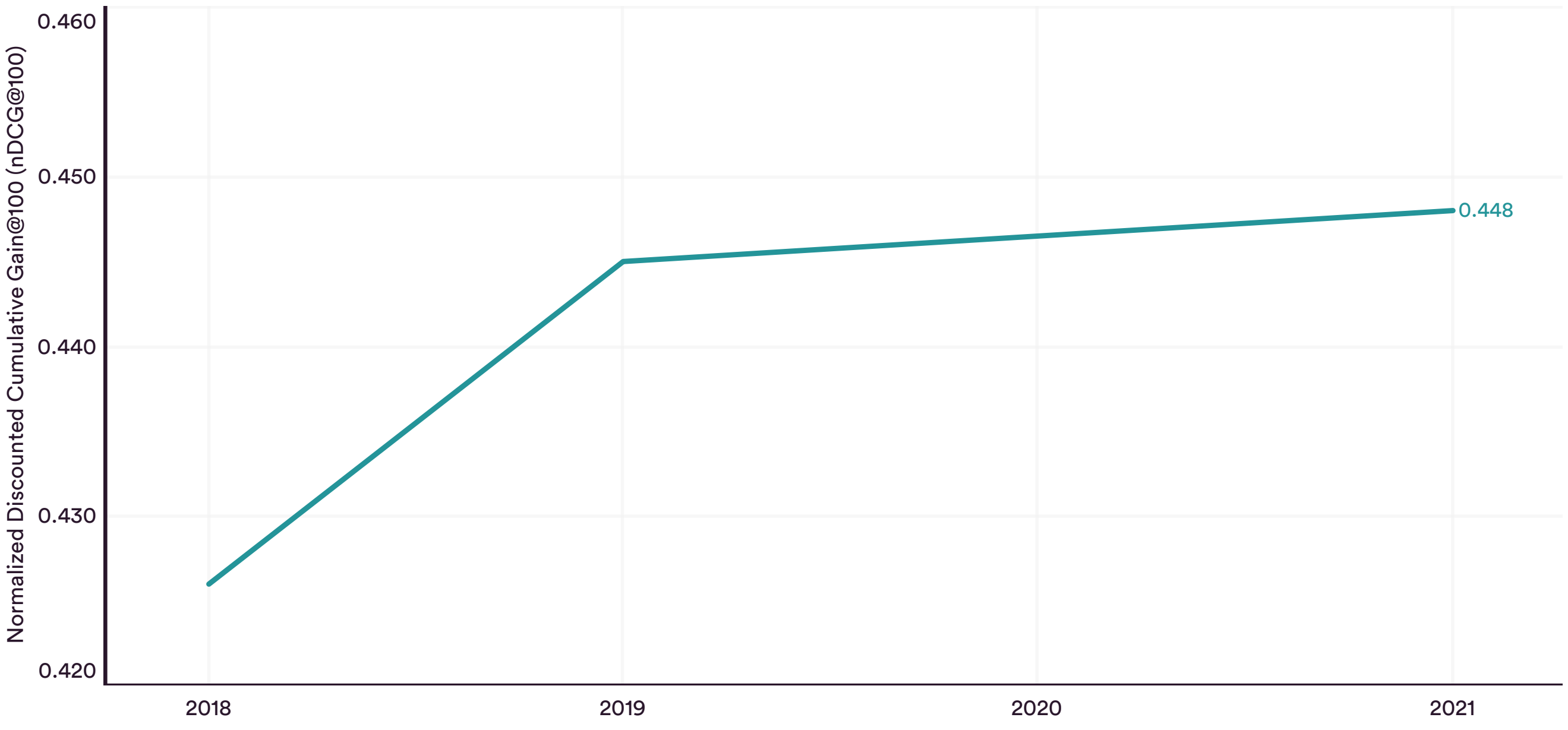

Figure 2.5.1

## Click-Through Rate Prediction: Criteo

Click-through rate prediction is the task of predicting the likelihood that something on a website, say an advertisement, will be clicked. In 2014, the online advertising platform Criteo launched an open click-through prediction challenge. Included as part of this challenge dataset was information on a million ads that were displayed during a 24-day period, whether they were clicked, and additional information on their characteristics. Since the competition launched, the Criteo dataset has been widely used to test recommender systems. On Criteo, systems are measured on area under the curve (AUC). A higher AUC means a better click-through prediction rate and a stronger recommender system.

Performance on Criteo also indicates that recommender systems have been slowly and steadily improving in the past decade. Last year's top model (Sina Weibo Corp's MaskNet) performed 1.8% higher on Criteo than the top model from 2016. An improvement of 1.8% may seem small in absolute terms, but it can be a valuable margin in the commercial world.

A limit of the Criteo and MovieLens benchmarks is that they are primarily academic measures of technical progress in recommendation (Figure 2.5.2). Most of the research work on recommendation occurs in commercial settings. Given that companies have an incentive to keep their recommendation improvements proprietary, the academic metrics included in this section might not be complete measures of the technical progress made in recommendation.

**CRITEO: AREA UNDER CURVE SCORE (AUC)**
Source: Papers with Code, 2021; arXiv, 2021 | Chart: 2022 AI Index Report

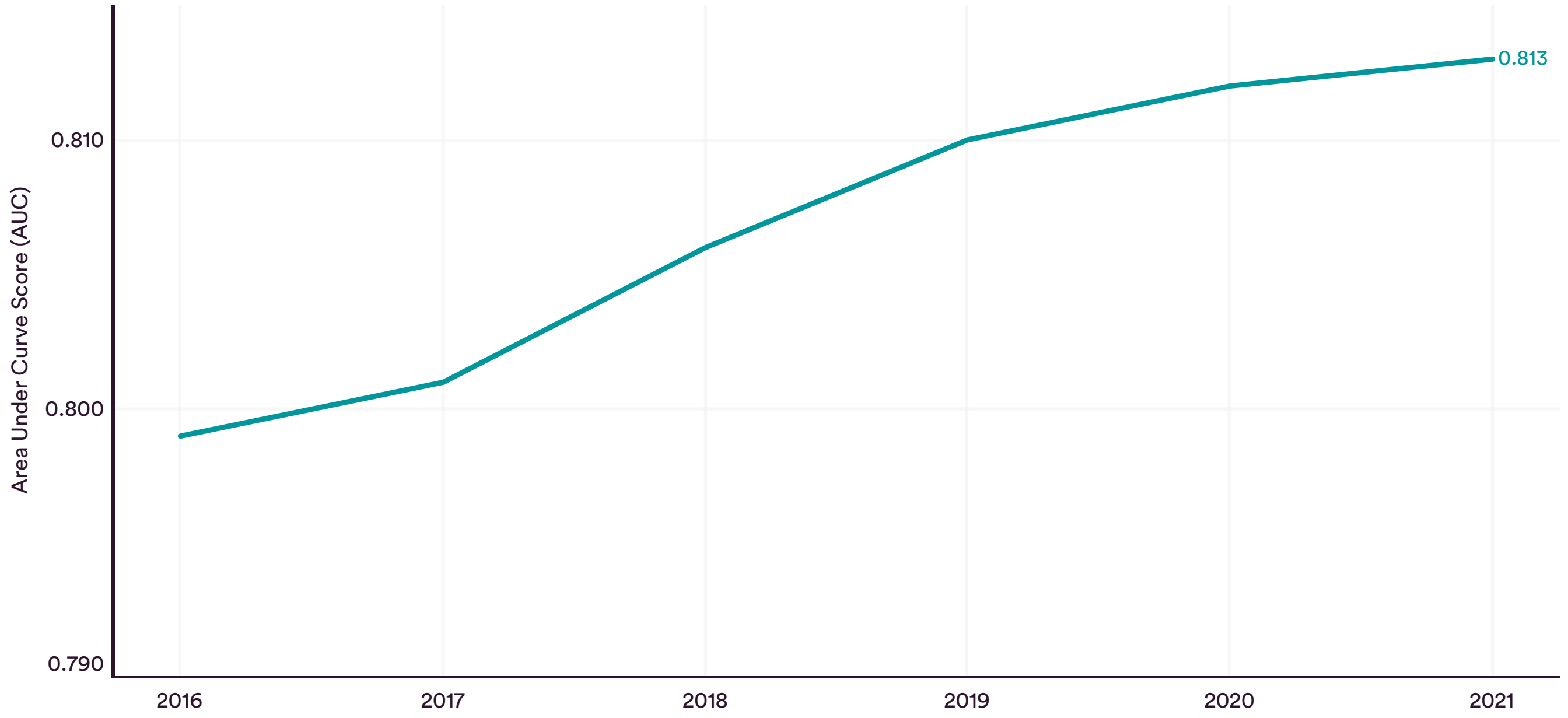

Figure 2.5.2

In reinforcement learning, AI systems are trained to maximize performance on a given task by interactively learning from their prior actions. Researchers train systems to optimize by rewarding them if they achieve a desired goal and then punishing them if they fail. Systems experiment with different strategy sequences to solve their stated problem (e.g., playing chess or navigating through a maze) and select the strategies which maximize their rewards.

Reinforcement learning makes the news whenever programs like DeepMind's AlphaZero demonstrate superhuman performance on games like Go and Chess. However, reinforcement learning is useful in any commercial domain where computer agents need to maximize a target goal or stand to benefit from learning from previous experiences. Reinforcement learning can help autonomous vehicles change lanes, robots optimize manufacturing tasks, or time-series models predict future events.

# 2.6 REINFORCEMENT LEARNING

## REINFORCEMENT LEARNING ENVIRONMENTS

A reinforcement learning environment is a computer platform where AI agents are challenged to maximize their performance on a defined task. Unlike other AI tasks which require systems to train on a dataset, reinforcement learning necessitates that AI systems have an environment in which they can test various strategies and, in the process, identify the set of strategies that will maximize rewards.

### Arcade Learning Environment: Atari-57

Introduced in 2013, the Arcade Learning Environment is an interface that includes various Atari 2600 game environments (such as "Pac-Man," "Space Invaders," and "Frogger") in which AI agents are challenged to optimize performance. To enable standard comparisons, researchers typically report average performance on the ALE across a suite of 57 games. There are various metrics in which performance is measured, but one of the most common is mean human-normalized score. A human-normalized score of 0% represents random performance, and a 100% score represents average human performance. The mean human-normalized score is then the average human-normalized score achieved by an AI system.

In late 2019, DeepMind's MuZero algorithm achieved state-of-the-art performance on Atari-57. MuZero not only performed 48.3% better on Atari-57 than the previous best-performing model, but it also set a new world record on Go and achieved superhuman performance on chess and shogi.

**Creating reinforcement learning models that are both high performing and highly efficient is an important step in the commercial deployment of reinforcement learning.**

In 2021, however, researchers from Tsinghua University and ByteDance launched the GDI-H3 model, which surpassed (and nearly doubled) MuZero's performance on Atari-57 (Figure 2.6.1). Moreover, GDI-H3 achieved this performance with less training. It only used 200 million training frames, whereas MuZero used 20 billion: GDI-H3 was twice as effective and one hundred times more efficient. Creating reinforcement learning models that are both high performing and highly efficient is an important step in the commercial deployment of reinforcement learning.

ATARI-57: MEAN HUMAN-NORMALIZED SCORE
Source: Papers with Code, 2021; arXlv, 2021 | Chart: 2022 AI Index Report

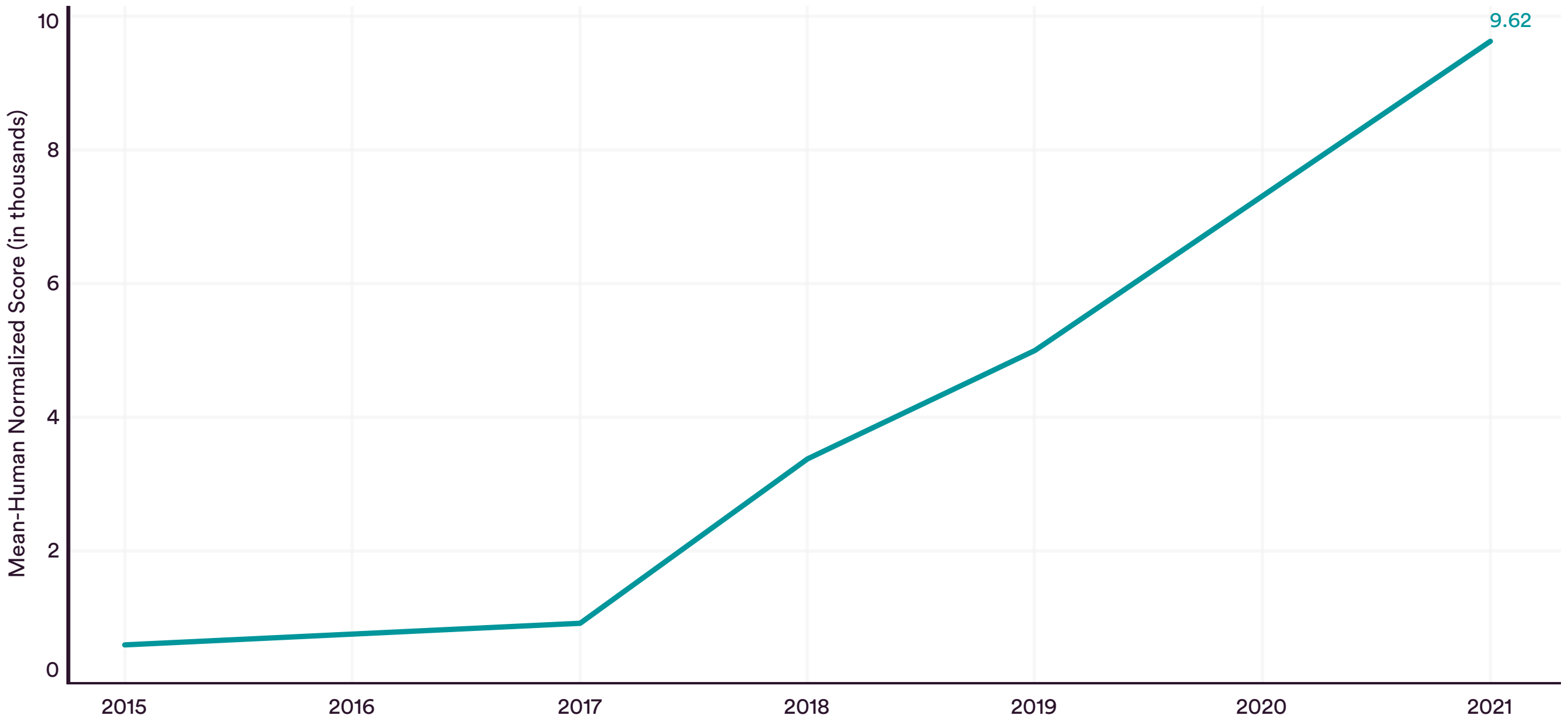

Figure 2.6.1

## Procgen

Procgen is a new reinforcement learning environment introduced by OpenAI in 2019. It includes 16 procedurally generated video-game-like environments specifically designed to test the ability of reinforcement learning agents to learn generalizable skills (Figure 2.6.2). Procgen was developed to overcome some of the criticisms leveled at benchmarks like Atari that incentivized AI systems to become narrow learners that maximized capacity in one particular skill. Procgen encourages broad learning by introducing a reinforcement learning environment that emphasizes high diversity and forces AI systems to train in generalizable ways. Performance on Procgen is measured in terms of mean-normalized score. Researchers typically train their systems on 200 million training runs and report an average score across the 16 Procgen games. The higher the system scores, the better the system.

A SCREENSHOT OF THE 16 GAME ENVIRONMENTS IN PROCGEN
Source: Cobbe et al. 2019

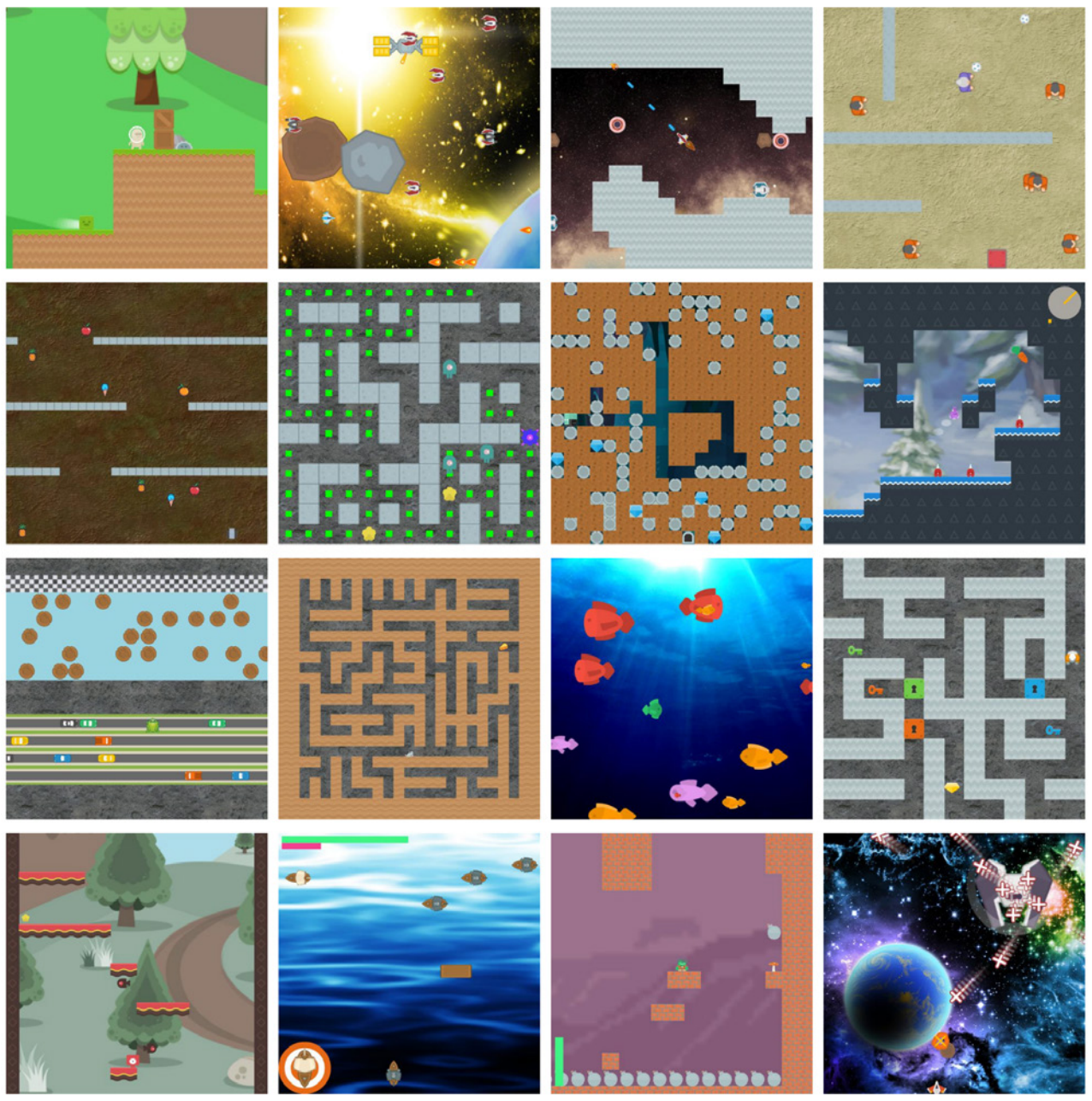

Figure 2.6.2

In November 2021, the MuZero model from DeepMind posted a state-of-the-art score of 0.6 on Procgen. DeepMind's results were a 128.6% improvement over the baseline performance established in 2019 when the environment was first released. Rapid progress on such a diverse benchmark signals that AI systems are slowly improving their capacity to reason in broader environments.

**PROCGEN: MEAN-NORMALIZED SCORE**

Source: arXiv, 2021 | Chart: 2022 AI Index Report

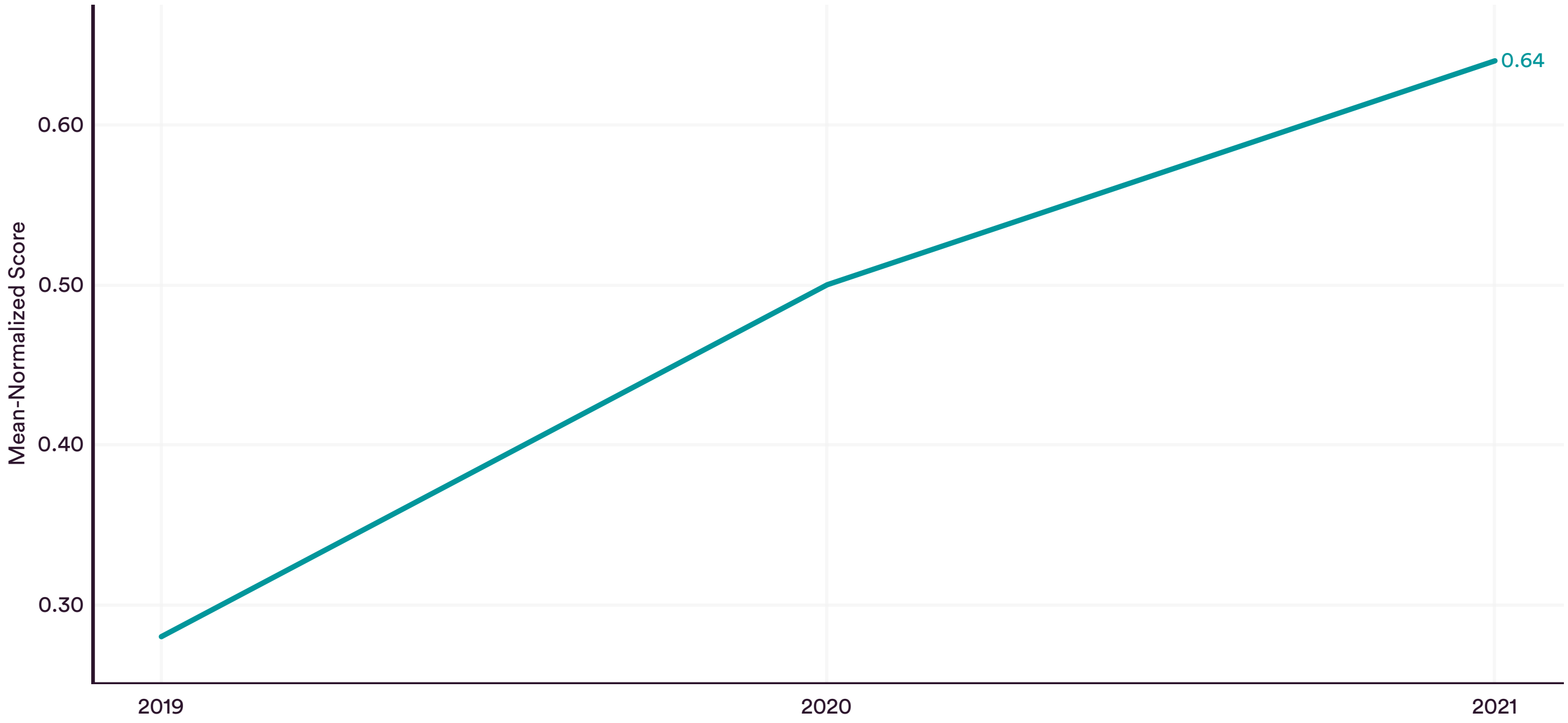

Figure 2.6.3

## Human Games: Chess

Progress in reinforcement learning can also be captured by the performance of the world's top chess software engines. A chess engine is a computer program that is trained to play chess at a high level by analyzing chess positions. The performance of chess engines is ranked on Elo, a method for identifying the relative skill levels of players in zero-sum games like chess: A higher score means a stronger player.

One caveat is that tracking the performance of chess engines is not a complete reflection of general reinforcement learning progress; chess engines are specifically trained to maximize performance on chess. Other popular reinforcement learning systems, like DeepMind's AlphaZero, are capable of playing a broader range of games, such as shogi and Go, and have in fact beaten some of the top-ranked chess engines. Nevertheless, looking at chess engine performance is an effective way to relativize the progress made in AI and compare it to a widely understandable human baseline.

Computers surpassed human performance in chess a long time ago, and since then have not stopped improving (Figure 2.6.4). By the mid-1990s, the top chess engines exceeded expert-level human performance, and by the mid-2000s they surpassed the peak performance of Magnus Carlsen, one of the best chess players of all time. Magnus Carlsen's 2882 Elo, recorded in 2014, is the highest level of human chess performance ever documented. As of 2021, the top chess engine has exceeded that level by 24.3%.

**CHESS SOFTWARE ENGINES: ELO SCORE**
Source: Swedish Computer Chess Association, 2021 | Chart: 2022 AI Index Report

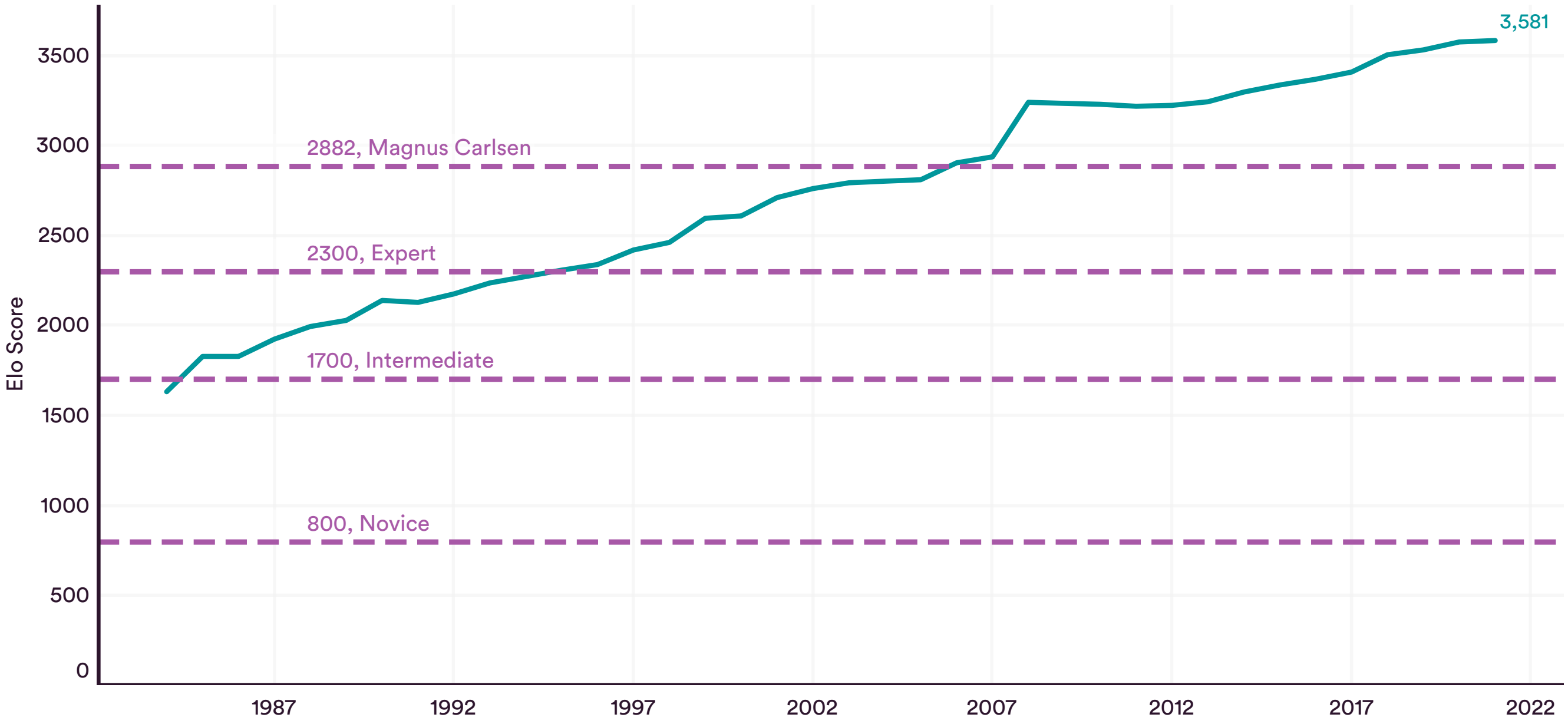

Figure 2.6.4

**In evaluating technical progress in AI, it is relevant not only to consider improvements in technical performance but also the speed of operation. As this section shows, AI systems continue to improve in virtually every skill category. This performance is often realized by increasing parameters and training systems on greater amounts of data. However, all else being equal, models that use more parameters and source more data will take longer to train. Longer train times mean slower real-world deployment. Given that the potential of increased training times can be offset by stronger and more robust computational infrastructures, it is important to keep track of progress in the hardware that powers AI systems.**

# 2.7 HARDWARE

## MLPerf: Training Time

MLPerf is an AI training competition run by the ML Commons organization. In this challenge, participants train systems to execute various AI tasks (image classification, image segmentation, natural language processing, etc.) using a common architecture. Entrants are then ranked on their absolute wall clock time, which is how long it takes for the system to train.

Since the MLPerf competitions began in December 2018, two key trends have emerged: (1) Training times for virtually every AI skill category have massively decreased; while (2) AI hardware robustness has substantially increased. Top-performing hardware systems can reach baseline levels of performance in task categories like recommendation, light-weight objection detection, image classification, and language processing in under a minute.

Figure 2.7.2 more precisely profiles the magnitude of improvement across each skill category since MLPerf first introduced the category.[6] For example, training times on image classification increased roughly twenty-seven-fold, as top times dropped from 6.2 minutes in 2018 to 0.2 minutes (or 13.8 seconds) in 2021. It might be hard to fathom the magnitude of a 27-time decrease in training time, but in real terms it is the difference between waiting one hour for a bus versus a little more than two minutes.

**MLPERF TRAINING TIME of TOP SYSTEMS by TASK: MINUTES**
Source: MLPerf, 2021 | Chart: 2022 AI Index Report

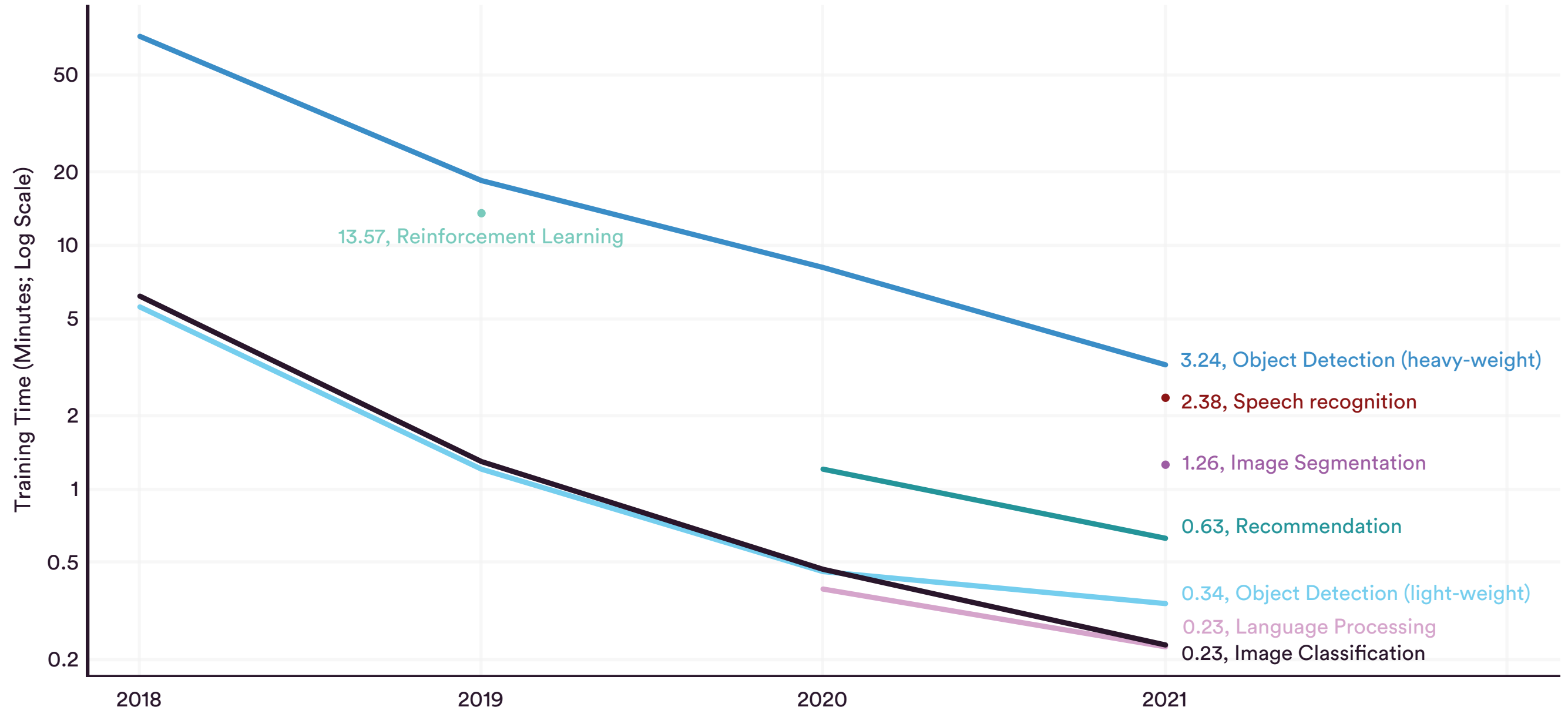

Figure 2.7.1

6 The solitary point for reinforcement learning on Figure 2.7.1 indicates that a faster time was not registered in the MLPerf competitions in 2020 or 2021. The solitary points for speech recognition and image segmentation are indicative of the fact that those AI subtask categories were added to the MLPerf competition in 2021.

## Top-performing hardware systems can reach baseline levels of performance in task categories like recommendation, light-weight objection detection, image classification, and language processing in under a minute.

**MLPERF: SCALE of IMPROVEMENT across TASK**
Source: MLPerf, 2021 | Chart: 2022 AI Index Report

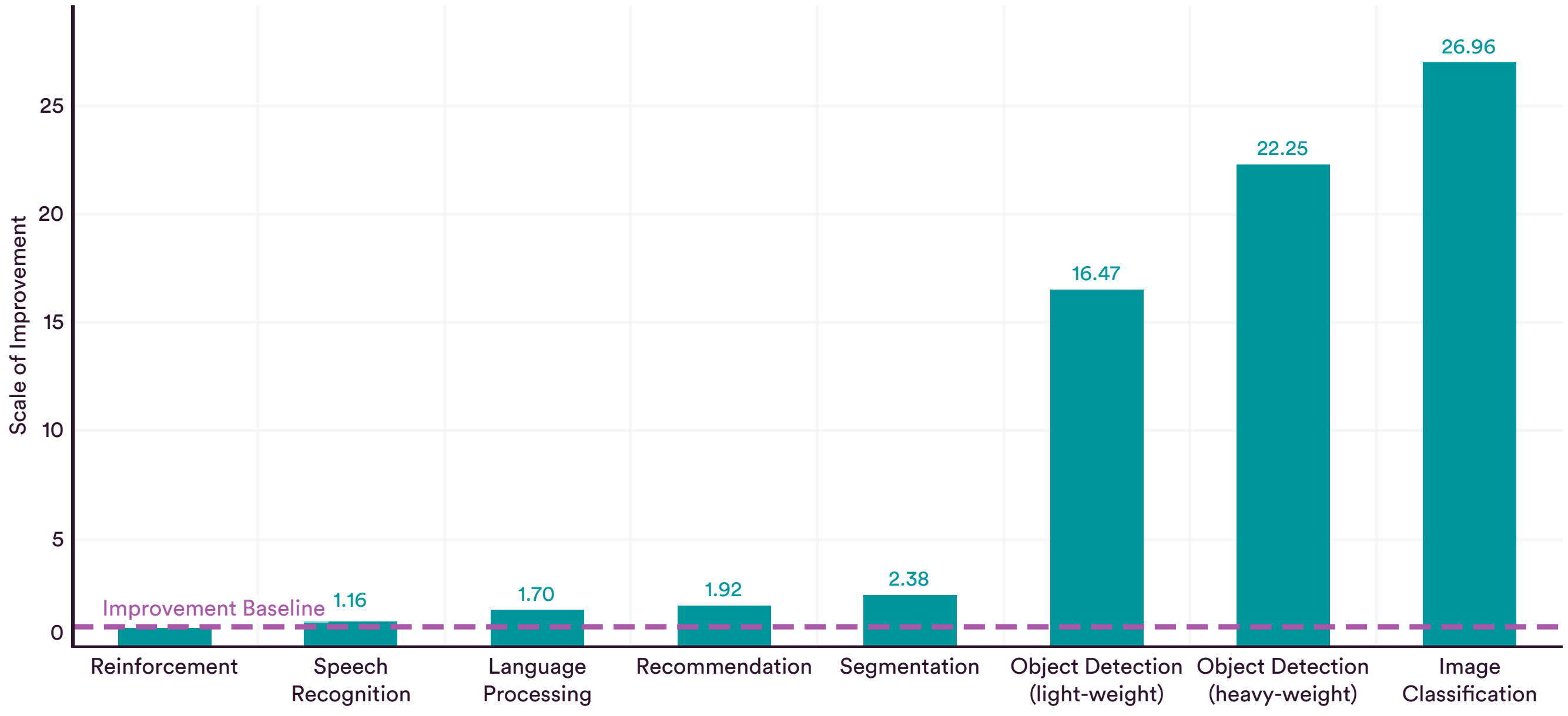

Figure 2.7.2

## MLPerf: Number of Accelerators

The cross-task improvements in training time are being driven by stronger underlying hardware systems, as shown in Figure 2.7.3. Since the competition began, the highest number of accelerators used—where an accelerator is a chip used predominantly for the machine learning component of a training run, such as a GPU or a TPU—and the mean number of accelerators used by the top system increased roughly 7 times while the mean number of accelerators used by all entrants increased 3.5 times. Most notable, however, is the growing gap between the average number of accelerators used by the top-performing systems and the average accelerators used by all systems. This gap was 9 times larger at the end of 2021 than it had been in 2018. This growth clearly means that, on average, building the fastest systems requires the most powerful hardware.

### MLPERF HARDWARE: ACCELERATORS

Source: MLPerf, 2021 | Chart: 2022 AI Index Report

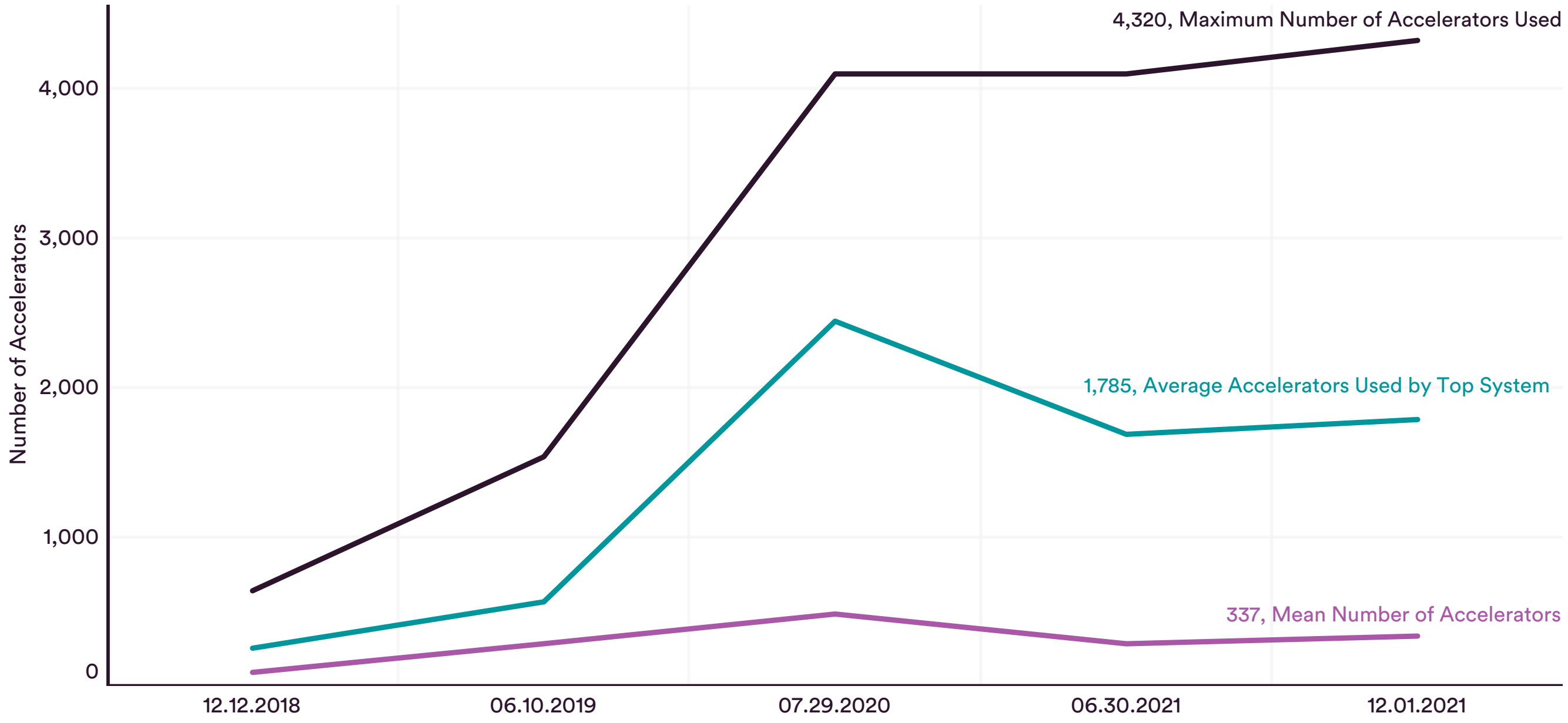

Figure 2.7.3

## IMAGENET: Training Cost

Stronger hardware has not necessarily meant costlier training. Figure 2.7.4 plots the lowest training cost per year on MLPerf's image classification subtask (ImageNet). In 2021, it cost only $4.6 to train a high-performing image classification system. This cost is marginal, especially when compared to the $1,112.6 it cost to train a similarly performing system in 2017. In simpler terms, in four short years, image classification training costs have decreased by a factor of 223.

**IMAGENET: TRAINING COST (to 93% ACCURACY)**
Source: AI Index and Narayanan, 2021 | Chart: 2022 AI Index Report

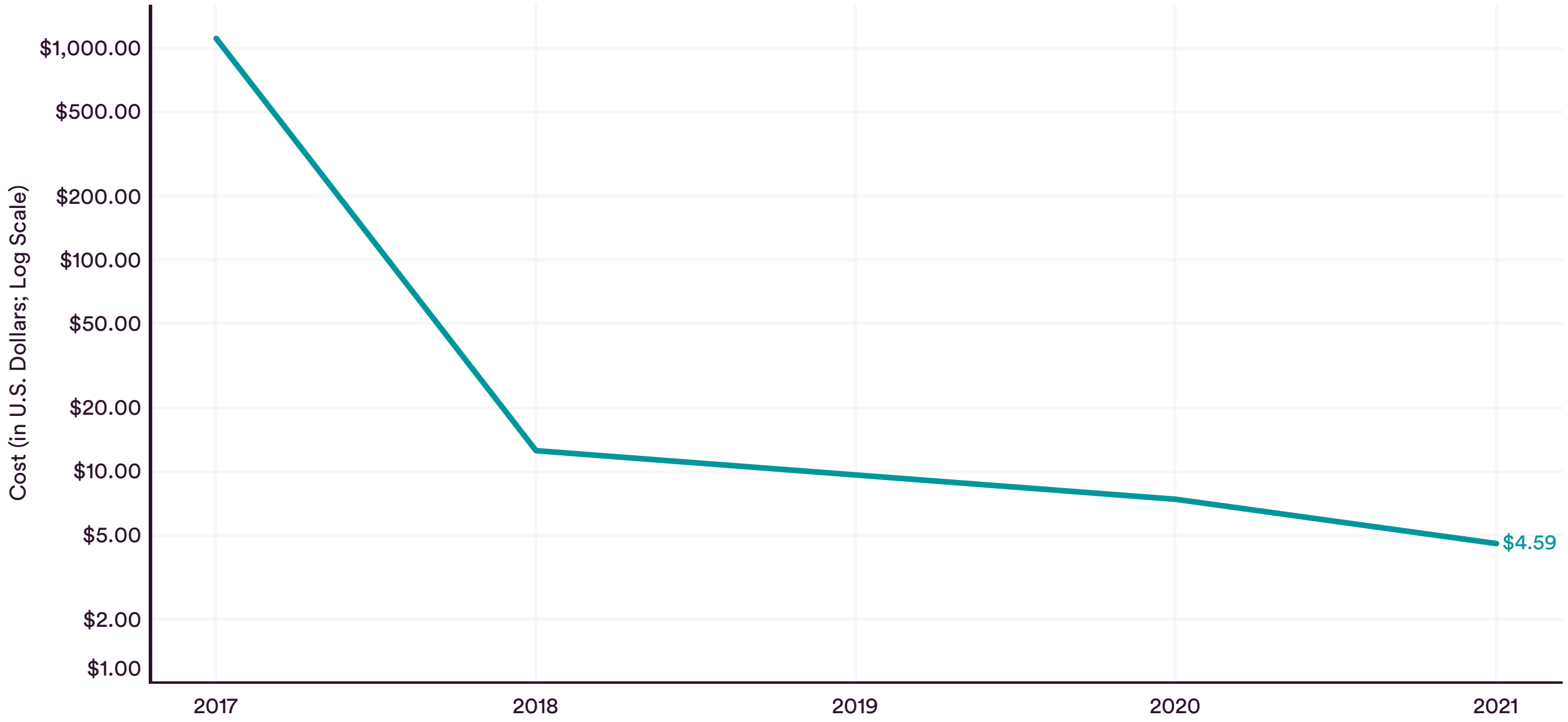

Figure 2.7.4

In 2021, the AI Index developed a survey that asked professors who specialize in robotics at top-ranked universities around the world and in emerging economies about changes in the pricing of robotic arms as well as the uses of robotic arms in research labs. The survey was completed by 101 professors and researchers from over 40 universities and collected data on 117 robotic arm purchase events from 2016 to 2022. The survey results suggest that there has been a notable decline in the price of robotic arms since 2016.

# 2.8 ROBOTICS

## Price Trends in Robotic Arms[7]

The survey results show a clear downward trend in the pricing of robotic arms in the last seven years. In 2017, the median price of a robotic arm was $42,000. Since then, the price has declined by 46.2% to reach roughly $22,600 in 2021 (Figure 2.8.1). Figure 2.8.2, which plots the distribution of robotic arm prices, paints a similar picture: Despite some high-priced outliers, there has been a clear downward trend since 2017 in the price of robotic arms.

**MEDIAN PRICE of ROBOTIC ARMS, 2017–21**
Source: AI Index, 2022 | Chart: 2022 AI Index Report

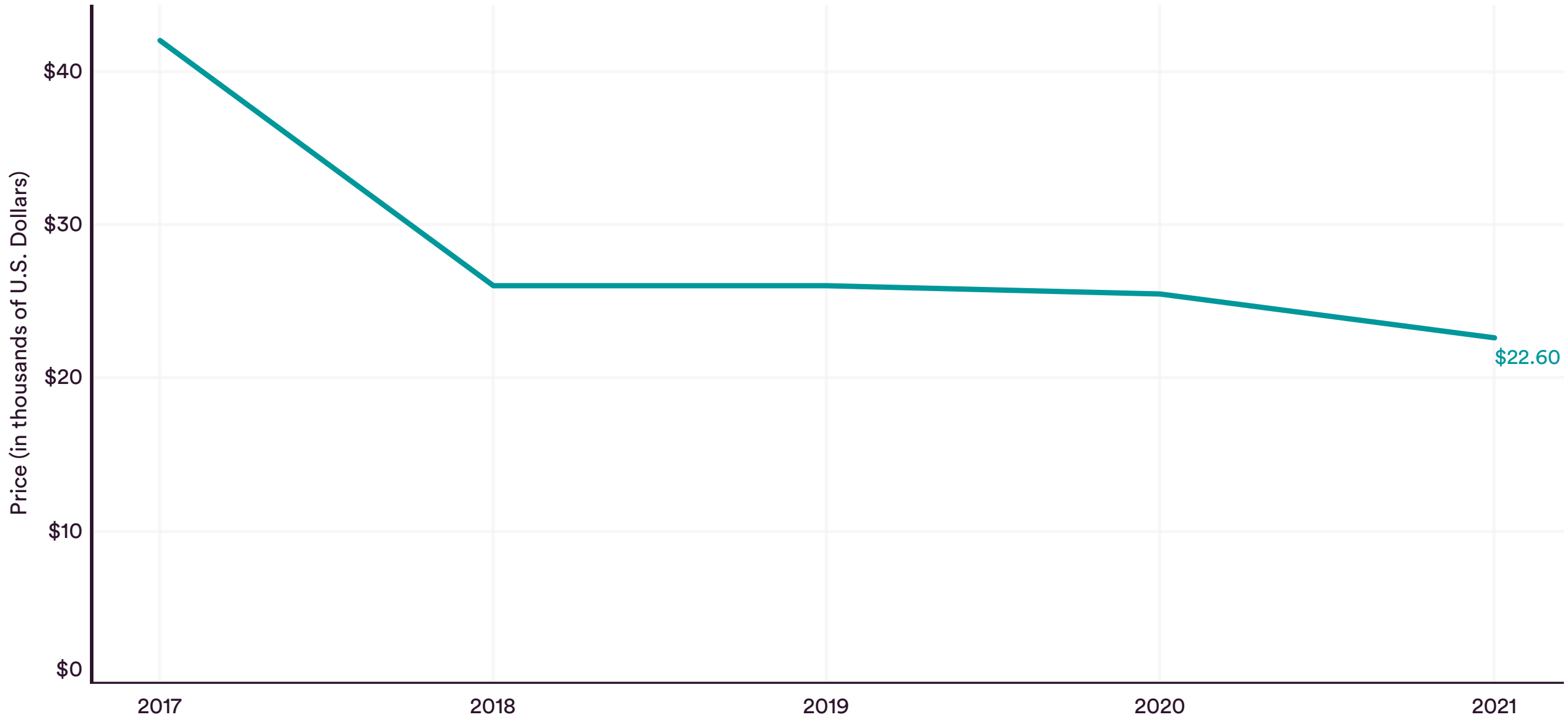

Figure 2.8.1

7 We have corrected Figure 2.8.1 and Figure 2.8.2 after noticing a data filtering issue with the survey result. The correct chart has since been updated. View the appendix here for links to the data. In addition, note that academic researchers may get a discount when purchasing robotic arms so prices are lower than retail.

## DISTRIBUTION of ROBOTIC ARM PRICES, 2017–21

Source: AI Index, 2022 | Chart: 2022 AI Index Report

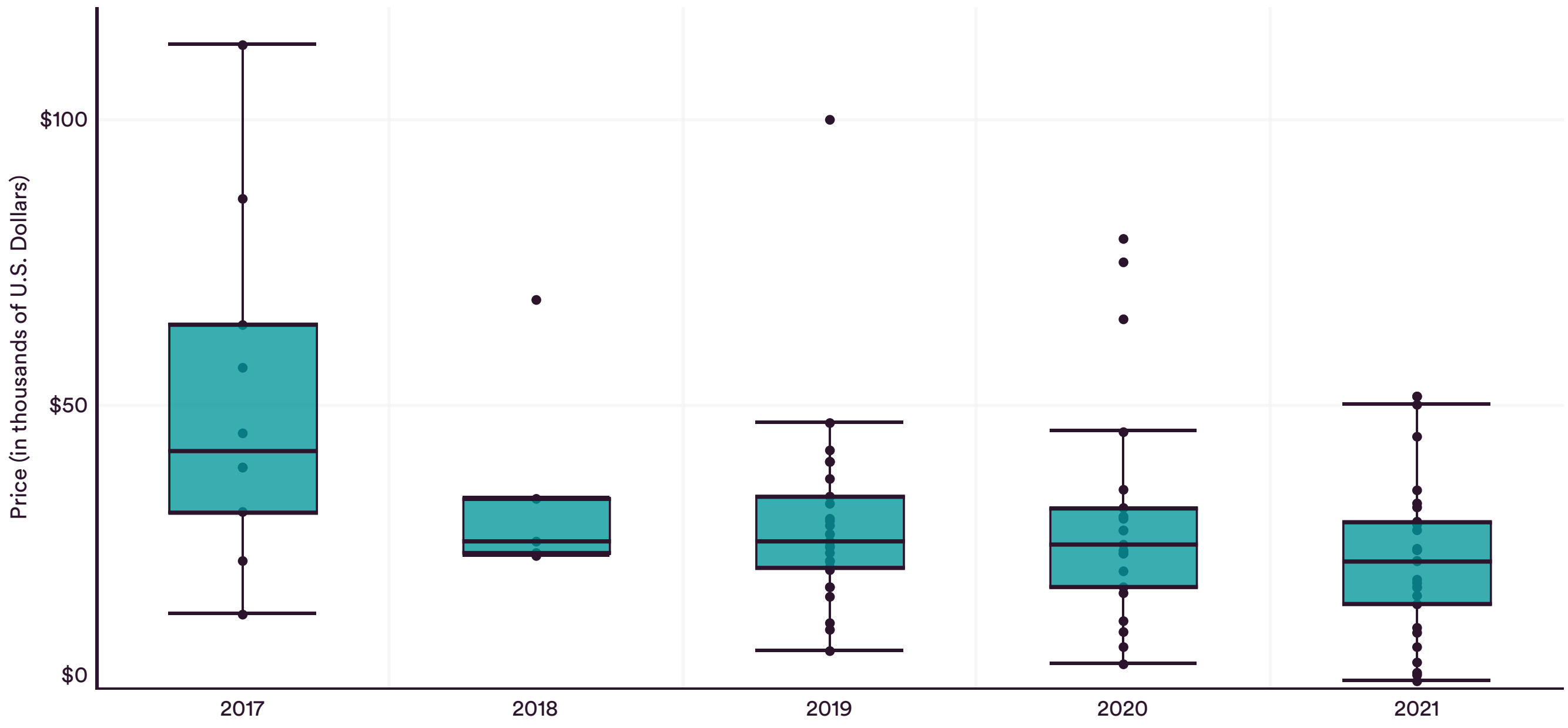

Figure 2.8.2

## AI Skills Employed by Robotics Professors

In the survey, the AI Index also asked robotics professors to what extent they employ AI skills in their research. Responses revealed that both deep learning and reinforcement learning are popular AI skills employed by roboticists. More specifically, 67.0% of professors reported using deep learning and 46.0% reported using reinforcement learning.

## AI SKILLS EMPLOYED by ROBOTICS PROFESSORS

Source: AI Index, 2022 | Chart: 2022 AI Index Report

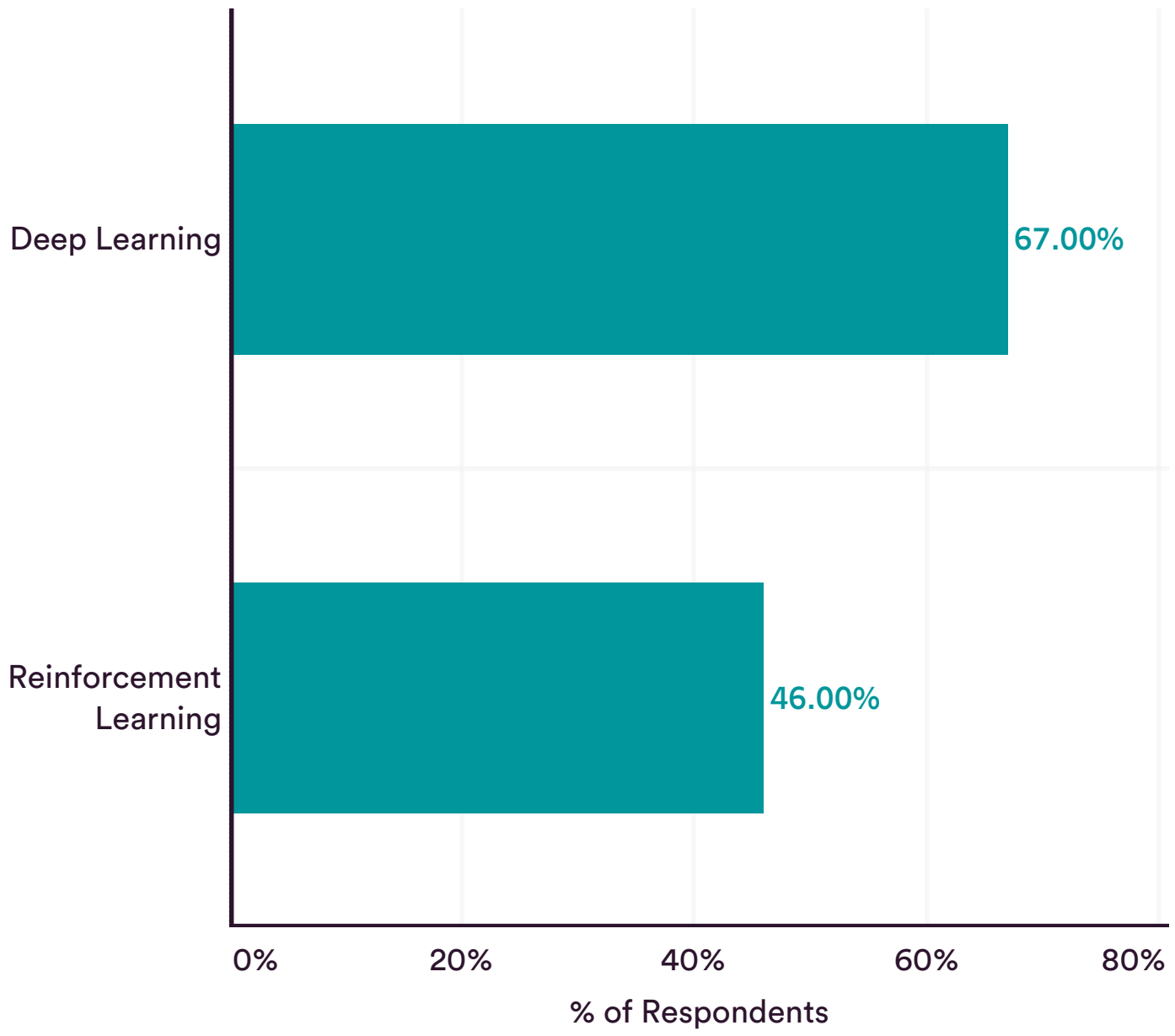

Figure 2.8.3

Artificial Intelligence Index Report 2022

# CHAPTER 3:
# Technical AI Ethics

Text and Analysis by
Helen Ngo and Ellie Sakhaee

## CHAPTER 3:

# Chapter Preview

**ACCESS THE PUBLIC DATA**

# Overview

In recent years, AI systems have started to be deployed into the world, and researchers and practitioners are reckoning with their real-world harms. Some of these harms include commercial facial recognition systems that discriminate based on race, résumé screening systems that discriminate on gender, and AI-powered clinical health tools that are biased along socioeconomic and racial lines. These models have been found to reflect and amplify human social biases, discriminate based on protected attributes, and generate false information about the world. These findings have increased interest within the academic community in studying AI ethics, fairness, and bias and prompted industry practitioners to direct resources toward remediating these issues, and attracted attention from the media, governments, and the people who use and are affected by these systems.

This year, the AI Index highlights metrics which have been adopted by the community for reporting progress in eliminating bias and promoting fairness. Tracking performance on these metrics alongside technical capabilities provides a more comprehensive perspective on how fairness and bias change as systems improve, which will be important to understand as systems are increasingly deployed.

## ACKNOWLEDGMENT

The AI Index would like to thank all those involved in research and advocacy around the development and governance of responsible AI. This chapter builds upon the work of scholars from across the AI ethics community, including those working on measuring technical capabilities as well those focused on shaping thoughtful societal norms. There is much more work to be done, but we are inspired by the progress made by this community and its collaborators.

Publications cited in this Chapter include:

Sandhini Agarwal, Gretchen Krueger, Jack Clark, Alec Radford, Jong Wook Kim, and Miles Brundage. 2021. Evaluating CLIP: Towards Characterization of Broader Capabilities and Downstream Implications. arXiv preprint arXiv:2108.02818.

Jack Bandy and Nicholas Vincent. 2021. Addressing “Documentation Debt” in Machine Learning Research: A Retrospective Datasheet for Book Corpus. arXiv preprint arXiv:2105.05241.

Abeba Birhane, Vinay Uday Prabhu, and Emmanuel Kahembwe. 2021. Multimodal Datasets: Misogyny, Pornography, and Malignant Stereotypes. arXiv preprint arXiv:2110.01963.

Sebastian Borgeaud, Arthur Mensch, Jordan Hoffmann, Trevor Cai, Eliza Rutherford, Katie Millican, George van den Driessche, Jean-Baptiste Lespiau, Bogdan Damoc, Aidan Clark, Diego de Las Casas, Aurelia Guy, Jacob Menick, Roman Ring, Tom Hennigan, Saffron Huang, Loren Maggiore, Chris Jones, Albin Cassirer, Andy Brock, Michela Paganini, Geoffrey Irving, Oriol Vinyals, Simon Osindero, Karen Simonyan, Jack W. Rae, Erich Elsen, and Laurent Sifre. 2021. Improving Language Models by Retrieving from Trillions of Tokens. arXiv preprint arXiv:2112.04426.

Nikhil Garg, Londa Schiebinger, Dan Jurafsky, and James Zou. 2017. Word Embeddings Quantify 100 Years of Gender and Ethnic Stereotypes. arXiv preprint arXiv:1711.08412.

Samuel Gehman, Suchin Gururangan, Maarten Sap, Yejin Choi, and Noah A.Smith. 2020. RealToxicityPrompts: Evaluating Neural Toxic Degeneration in Language Models. arXiv preprint arXiv:2009.11462.

Wei Guo and Aylin Caliskan. 2020. Detecting Emergent Intersectional Biases: Contextualized Word Embeddings Contain a Distribution of Human-like Biases. arXiv preprint arXiv:2006.03955.

Aylin Caliskan Islam, Joanna J. Bryson, and Arvind Narayanan. 2016. Semantics Derived Automatically from Language Corpora Necessarily Contain Human Biases. arXiv preprint arXiv:1608.07187.

Opher Lieber, Or Sharir, Barak Lenz, and Yoav Shoham. 2021. Jurassic-1: Technical Details and Evaluation. (2021). https://uploads-ssl.webflow.com/60fd4503684b466578c0d307/61138924626a6981ee09caf6_jurassic_tech_paper.pdf

Chandler May, Alex Wang, Shikha Bordia, Samuel R. Bowman, and Rachel Rudinger. 2019. On Measuring Social Biases in Sentence Encoders. arXiv preprint arXiv:1903.10561.

Moin Nadeem, Anna Bethke, and Siva Reddy. 2020. StereoSet: Measuring Stereotypical Bias in Pretrained Language Models. arXiv preprint arXiv:2004.09456.

Reiichiro Nakano, Jacob Hilton, Suchir Balaji, Jeff Wu, Long Ouyang, Christina Kim, Christopher Hesse, Shantanu Jain, Vineet Kosaraju, William Saunders, Xu Jiang, Karl Cobbe, Tyna Eloundou, Gretchen Krueger, Kevin Button, Matthew Knight, Benjamin Chess, John Schulman. WebGPT: Browser-Assisted Question-Answering with Human Feedback. 2021. arXiv preprint arXiv:2112.09332.

Nikita Nangia, Clara Vania, Rasika Bhalerao, and Samuel R. Bowman. 2020. CrowS-Pairs: A Challenge Dataset for Measuring Social Biases in Masked Language Models. arXiv preprint arXiv:2010.00133.

Long Ouyang, Jeff Wu, Xu Jiang, Diogo Almeida, Carroll L. Wainwright, Pamela Mishkin, Chong Zhang, Sandhini Agarwal, Katarina Slama, Alex Ray, John Schulman, Jacob Hilton, Fraser Kelton, Luke Miller, Maddie Simens, Amanda Askell, Peter Welinder, Paul Christiano, Jan Leike, Ryan Lowe. Training Language Models to Follow Instructions with Human Feedback. 2022. arXiv preprint arXiv:2203.02155.

Jack W. Rae, Sebastian Borgeaud, Trevor Cai, Katie Millican, Jordan Hoffmann, H. Francis Song, John Aslanides, Sarah Henderson, Roman Ring, Susannah Young, Eliza Rutherford, Tom Hennigan, Jacob Menick, Albin Cassirer, Richard Powell, George van den Driessche, Lisa Anne Hendricks, Maribeth Rauh, Po-Sen Huang, Amelia Glaese, Johannes Welbl, Sumanth Dathathri, Saffron Huang, Jonathan Uesato, John Mellor, Irina Higgins, Antonia Creswell, Nat McAleese, Amy Wu, Erich Elsen, Siddhant M. Jayakumar, Elena Buchatskaya, David Budden, Esme Sutherland, Karen Simonyan, Michela Paganini, Laurent Sifre, Lena Martens, Xiang Lorraine Li, Adhiguna Kuncoro, Aida Nematzadeh, Elena Gribovskaya, Domenic Donato, Angeliki Lazaridou, Arthur Mensch, Jean-Baptiste Lespiau, Maria Tsimpoukelli, Nikolai Grigorev, Doug Fritz, Thibault Sottiaux, Mantas Pajarskas, Toby Pohlen, Zhitao Gong, Daniel Toyama, Cyprien de Masson d'Autume, Yujia Li, Tayfun Terzi, Vladimir Mikulik, Igor Babuschkin, Aidan Clark, Diego de Las Casas, Aurelia Guy, Chris Jones, James Bradbury, Matthew Johnson, Blake A. Hechtman, Laura Weidinger, Iason Gabriel, William S. Isaac, Edward Lockhart, Simon Osindero, Laura Rimell, Chris Dyer, Oriol Vinyals, Kareem Ayoub, Jeff Stanway, Lorrayne Bennett, Demis Hassabis, Koray Kavukcuoglu, and Geoffrey Irving. 2021. Scaling Language Models: Methods, Analysis & Insights from Training Gopher. arXiv preprint arXiv:2112.11446.

Gabriel Stanovsky, Noah A. Smith, and Luke Zettlemoyer. 2019. Evaluating Gender Bias in Machine Translation. arXiv preprint arXiv:1906.00591.

Ryan Steed and Aylin Caliskan. 2020. Image Representations Learned With Unsupervised Pre-Training Contain Human-like Biases. arXiv preprint arXiv:2010.15052.

Laura Weidinger, John Mellor, Maribeth Rauh, Conor Griffin, Jonathan Uesato, Po-Sen Huang, Myra Cheng, Mia Glaese, Borja Balle, Atoosa Kasirzadeh, Zac Kenton, Sasha Brown, Will Hawkins, Tom Stepleton, Courtney Biles, Abeba Birhane, Julia Haas, Laura Rimell, Lisa Anne Hendricks, William S. Isaac, Sean Legassick, Geoffrey Irving, and Iason Gabriel. 2021. Ethical and social risks of harm from Language Models. arXiv preprint arXiv:2112.04359.

Johannes Welbl, Amelia Glaese, Jonathan Uesato, Sumanth Dathathri, John Mellor, Lisa Anne Hendricks, Kirsty Anderson, Pushmeet Kohli, Ben Coppin, and Po-Sen Huang. 2021. Challenges in Detoxifying Language Models. arXiv preprint arXiv:2109.07445.

Albert Xu, Eshaan Pathak, Eric Wallace, Suchin Gururangan, Maarten Sap, and Dan Klein. 2021. Detoxifying Language Models Risks Marginalizing Minority Voices. arXiv preprint arXiv:2104.06390.

Pei Zhou, Weijia Shi, Jieyu Zhao, Kuan-Hao Huang, Muhao Chen, Ryan Cotterell, and Kai-Wei Chang. 2019. Examining Gender Bias in Languages with Grammatical Gender. arXiv preprint arXiv:1909.02224.

# CHAPTER HIGHLIGHTS

- **Language models are more capable than ever, but also more biased:** Large language models are setting new records on technical benchmarks, but new data shows that larger models are also more capable of reflecting biases from their training data. **A 280 billion parameter model developed in 2021 shows a 29% increase in elicited toxicity over a 117 million parameter model considered the state of the art as of 2018.** The systems are growing significantly more capable over time, though as they increase in capabilities, so does the potential severity of their biases.

- **The rise of AI ethics everywhere:** Research on fairness and transparency in AI has exploded since 2014, **with a fivefold increase in related publications** at ethics-related conferences. Algorithmic fairness and bias has shifted from being primarily an academic pursuit to becoming firmly embedded as a mainstream research topic with wide-ranging implications. **Researchers with industry affiliations contributed 71% more publications year over year** at ethics-focused conferences in recent years.

- **Multimodal models learn multimodal biases:** Rapid progress has been made on training multimodal language-vision models which exhibit new levels of capability on joint language-vision tasks. These models have set new records on tasks like image classification and the creation of images from text descriptions, but they also reflect societal stereotypes and biases in their outputs—**experiments on CLIP showed that images of Black people were misclassified as nonhuman at over twice the rate of any other race.** While there has been significant work to develop metrics for measuring bias within both computer vision and natural language processing, this highlights the need for metrics that provide insight into biases in models with multiple modalities.

**Significant research effort has been invested over the past five years into creating datasets, benchmarks, and metrics designed to measure bias and fairness in machine learning models. Bias is often learned from the underlying training data for an AI model; this data can reflect systemic biases in society or the biases of the humans who collected and curated the data.**

# 3.1 META-ANALYSIS OF FAIRNESS AND BIAS METRICS

Algorithmic bias is commonly framed in terms of allocative and representation harms. Allocative harm occurs when a system unfairly allocates an opportunity or resource to a specific group, and representation harm happens when a system perpetuates stereotypes and power dynamics in a way that reinforces subordination of a group. Algorithms are broadly considered fair when they make predictions that neither favor nor discriminate against individuals or groups based on protected attributes which cannot be used for decision-making due to legal or ethical reasons (e.g., race, gender, religion).

To better understand the landscape of algorithmic bias and fairness, the AI Index conducted original research to analyze the state of the field. As shown in Figure 3.1.1, the number of metrics for measuring bias and fairness along ethical dimensions of interest has grown steadily since 2018. For this graph, the number of fairness and bias metrics published has been cited in at least one other work.[1]

**NUMBER of AI FAIRNESS and BIAS METRICS, 2016–21**
Source: AI Index, 2021 | Chart: 2022 AI Index Report

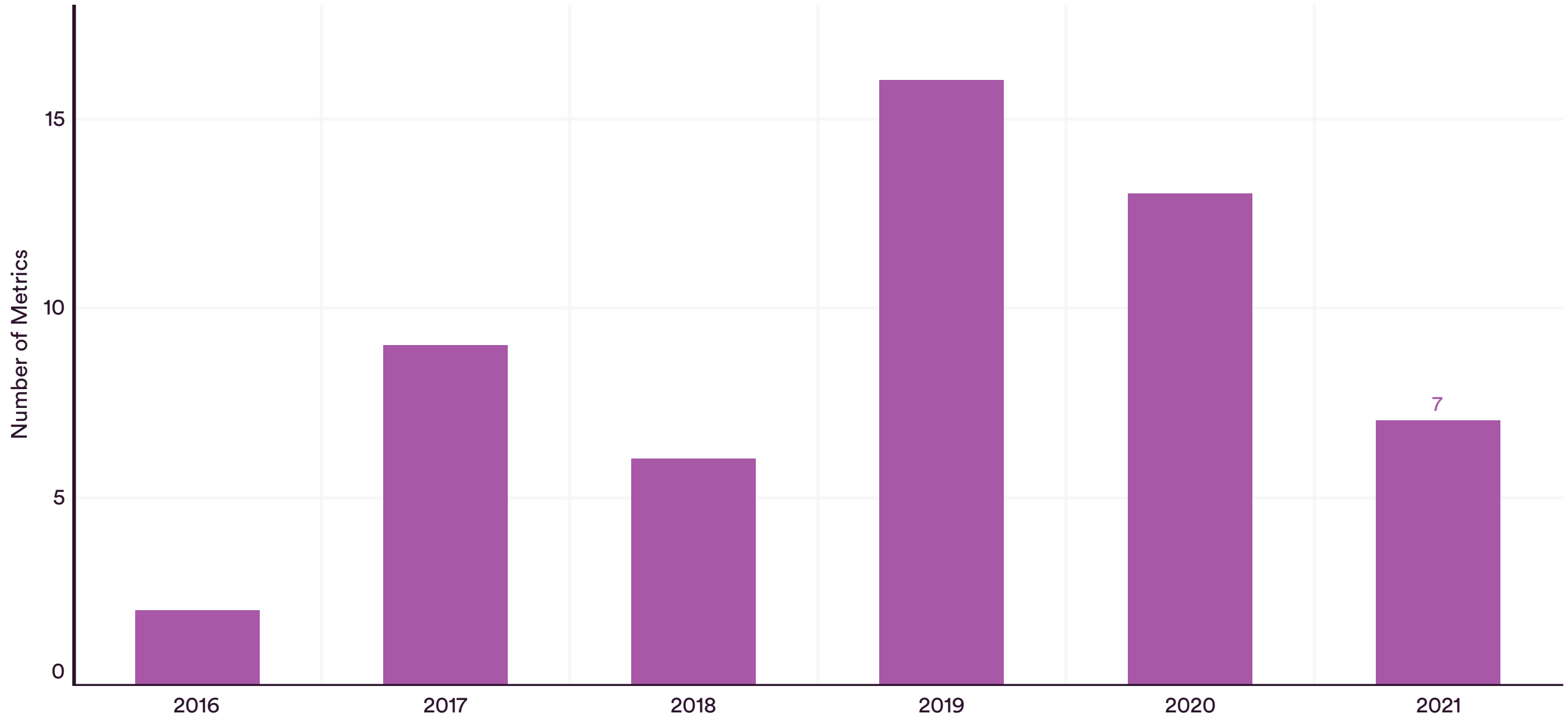

Figure 3.1.1

1 2021 data may be lagging as it takes time for metrics to be adopted by the community.

## AI ETHICS DIAGNOSTIC METRICS AND BENCHMARKS

Measurement of AI systems along an ethical dimension often takes one of two forms:

- **Benchmark datasets:** A benchmark dataset contains labeled data, and researchers test how well their AI system labels the data. Benchmarks do not change over time. These are domain-specific (e.g., SuperGLUE and StereoSet for language models; ImageNet for computer vision) and often aim to measure behavior that is intrinsic to the model, as opposed to its downstream performance on specific populations (e.g., StereoSet measures model propensity to select stereotypes compared to non-stereotypes, but it does not measure performance gaps between different subgroups).

- **Diagnostic metrics:** A diagnostic metric measures the impact or performance of a model on a downstream task—for example, a population subgroup or individual compared to similar individuals or the entire population. These metrics can help researchers understand how a system will perform when deployed in the real world, and whether it has a disparate impact on certain populations. Examples include group fairness metrics such as demographic parity and equality of opportunity.

Benchmarks are useful indicators of progress for the field as a whole, and their impact can be measured by community adoption (e.g., number of leaderboard submissions, or the number of research papers which report metrics). They also often enable rapid algorithmic progress as research labs compete on leaderboard metrics. However, some leaderboards can be easily gamed, and may be based on benchmark datasets that contain flaws, such as incorrect labels or poorly defined classes. Additionally, their static nature means they are a snapshot of a specific cultural and temporal context—in other words, a benchmark published in 2017 may not correlate to the deployment context of 2022.

Diagnostic metrics enable researchers and practitioners to understand the impact of their system on a specific application or group and potential concrete harm (e.g., "this model is disproportionately underperforming on this group with this protected attribute"). Diagnostic metrics are most useful at an individual model or application level as opposed to functioning as field-level indicators. They indicate how a specific AI system performs on a specific subgroup or individual, which is helpful for assessing real-world impact. However, while these metrics may be widely used to test models privately, there is not as much information available publicly as these metrics are not attached to leaderboards which encourage researchers to publish their results.

Figure 3.1.2 shows that there has been a steady amount of research investment into developing both benchmarks and diagnostic metrics over time.[2] [3] 2 Research paper citations are a lagging indicator of activity, and metrics which have been very recently adopted may not be reflected in the current data, similar to 3.1.1.
3 The Perspective API defined seven new metrics for measuring facets of toxicity (toxicity, severe toxicity, identity attack, insult, obscene, sexually explicit, threat), contributing to the unusually high number of metrics released in 2017.

NUMBER of AI FAIRNESS and BIAS METRICS (DIAGNOSTIC METRICS vs. BENCHMARKS), 2016–21
Source: AI Index, 2021 | Chart: 2022 AI Index Report

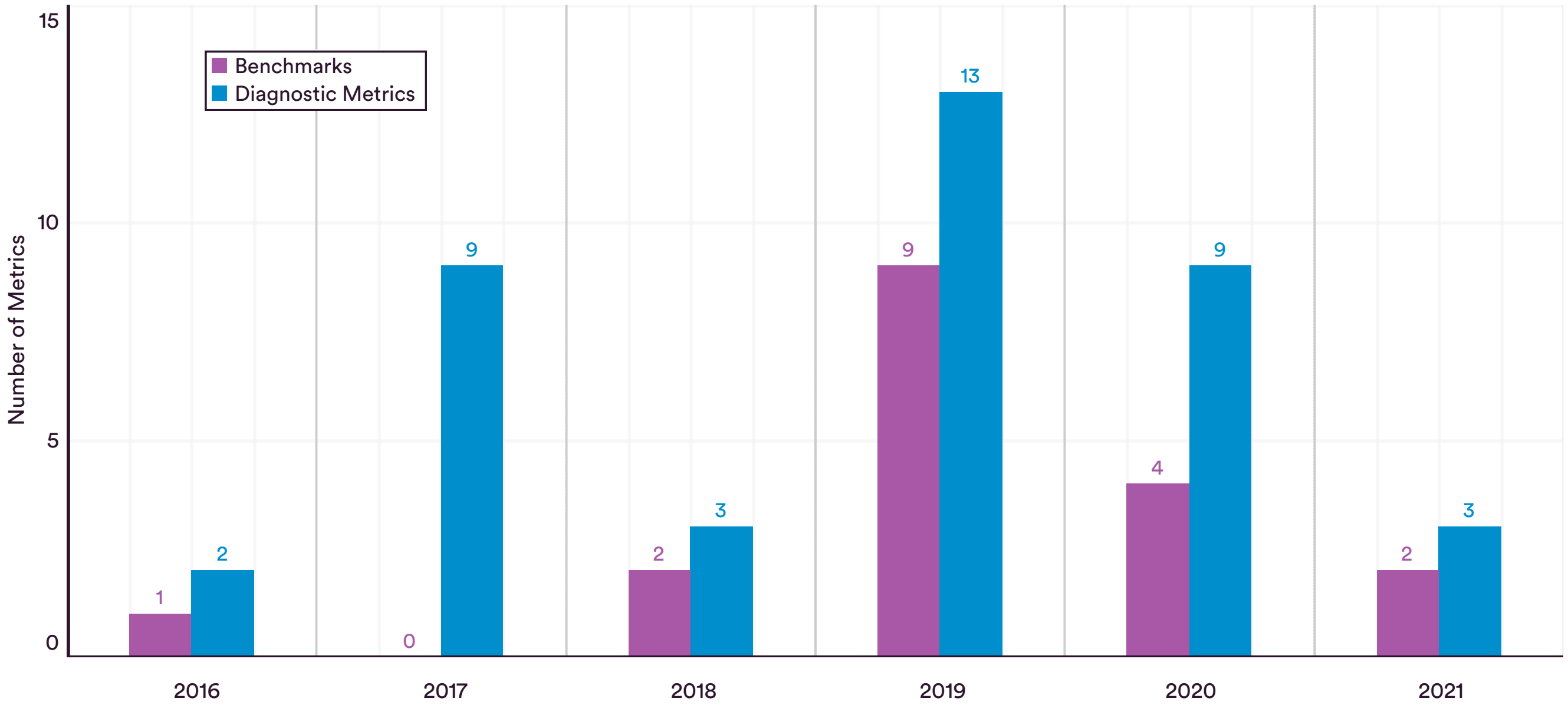

Figure 3.1.2

The rest of this chapter examines the performance of recent AI systems on these metrics and benchmarks in depth within domains such as natural language and computer vision. The majority of these metrics measure intrinsic bias within systems, and it has been shown that intrinsic bias metrics may not fully capture the effects of extrinsic bias within downstream applications.

**Current state-of-the-art natural language processing (NLP) relies on large language models or machine learning systems that process millions of lines of text and learn to predict words in a sentence. These models can generate coherent text; classify people, places, and events; and be used as components of larger systems, like search engines. Collecting training data for these models often requires scraping the internet to create web-scale text datasets. These models learn human biases from their pretraining data and reflect them in their downstream outputs, potentially causing harm. Several benchmarks and metrics have been developed to identify bias in natural language processing along axes of gender, race, occupation, disability, religion, age, physical appearance, sexual orientation, and ethnicity.**

# 3.2 NATURAL LANGUAGE PROCESSING BIAS METRICS

Bias metrics can be split into two major categories: intrinsic metrics, which measure bias in internal embedding spaces of models, and extrinsic metrics, which measure bias in the downstream tasks and outputs of the model. Examples of extrinsic metrics include group fairness metrics (parity across protected groups) and individual fairness metrics (parity across similar individuals), which measure whether a system has a disproportionately negative impact on a subgroup or individual, or gives preferential treatment to one group at the expense of another.

## TOXICITY: REALTOXICITYPROMPTS AND THE PERSPECTIVE API

Measuring toxicity in language models requires labels for toxic and nontoxic content. Toxicity is defined as a rude, disrespectful or unreasonable comment that is likely to make someone leave a conversation. The Perspective API is a tool developed by Jigsaw, a Google company. It was originally designed to help platforms identify toxicity in online conversations. Developers input text into the Perspective API, which returns probabilities that the text should be labeled as falling into one of the following categories: toxicity, severe toxicity, identity attack, insult, obscene, sexually explicit, and threat.

Since the Perspective API was released in 2017, the NLP research community has rapidly adopted it for measuring toxicity in natural language. As seen in Figure 3.2.1, the number of papers using the Perspective API doubled between 2020 and 2021, from 8 to 19.

**NUMBER of RESEARCH PAPERS USING PERSPECTIVE API**
Source: AI Index, 2021 | Chart: 2022 AI Index Report

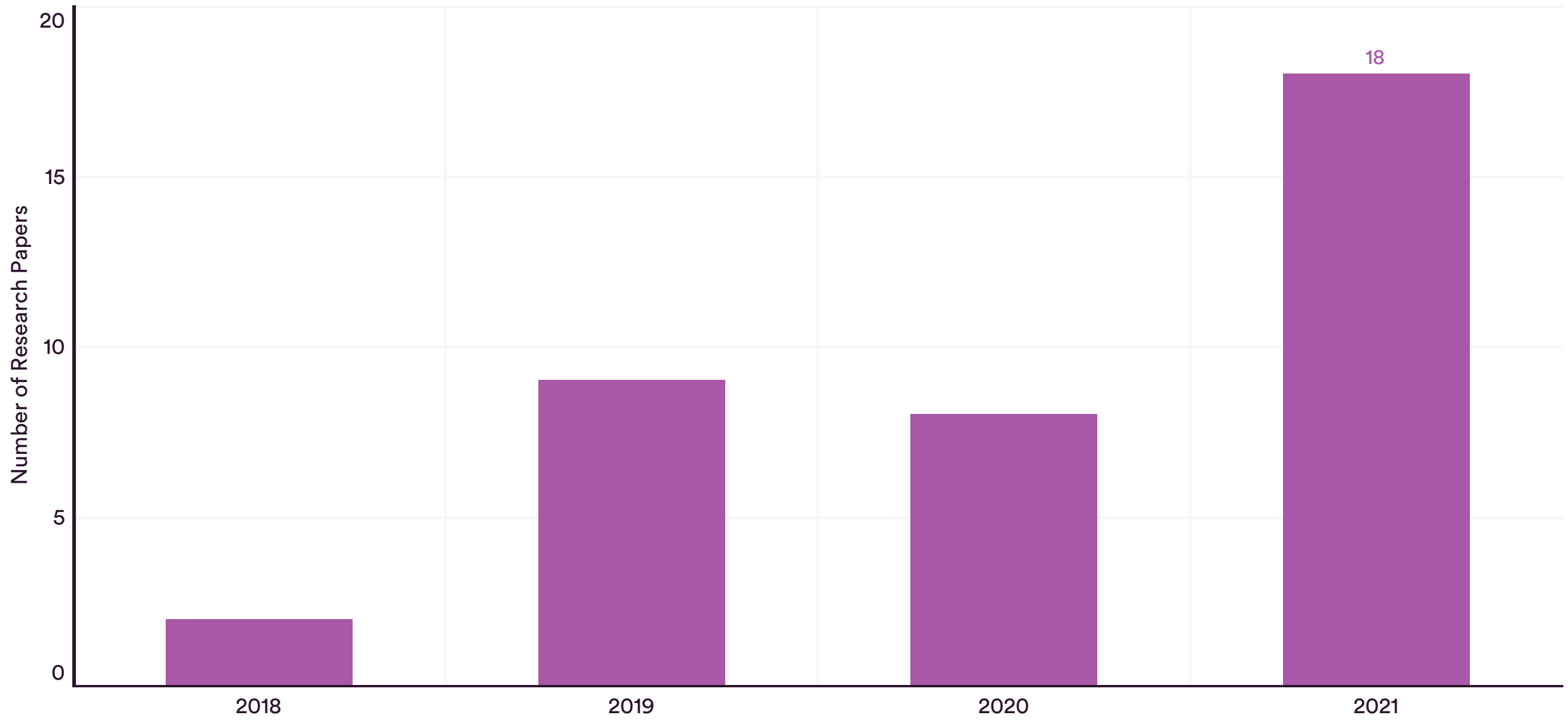

Figure 3.2.1

RealToxicityPrompts consists of English natural language prompts used to measure how often a language model completes a prompt with toxic text. Toxicity of a language model is measured with two metrics:

- Maximum toxicity: the average maximum toxicity score across some number of completions
- Probability of toxicity: how often a completion is expected to be toxic

Figure 3.2.2 shows that toxicity in language models depends heavily on the underlying training data. Models trained on internet text with toxic content filtered out are significantly less toxic compared to models trained on various corpora of unfiltered internet text. A model trained on BookCorpus (a dataset containing books from e-book websites) produces toxic text surprisingly often. This may be due to its composition—BookCorpus contains a significant number of romance novels containing explicit content, which may contribute to higher levels of toxicity.

**TOXICITY in LANGUAGE MODELS by TRAINING DATASET**
Source: Gehman et al., 2021; Rae et al., 2021; Welbl et al., 2021 | Chart: 2022 AI Index Report

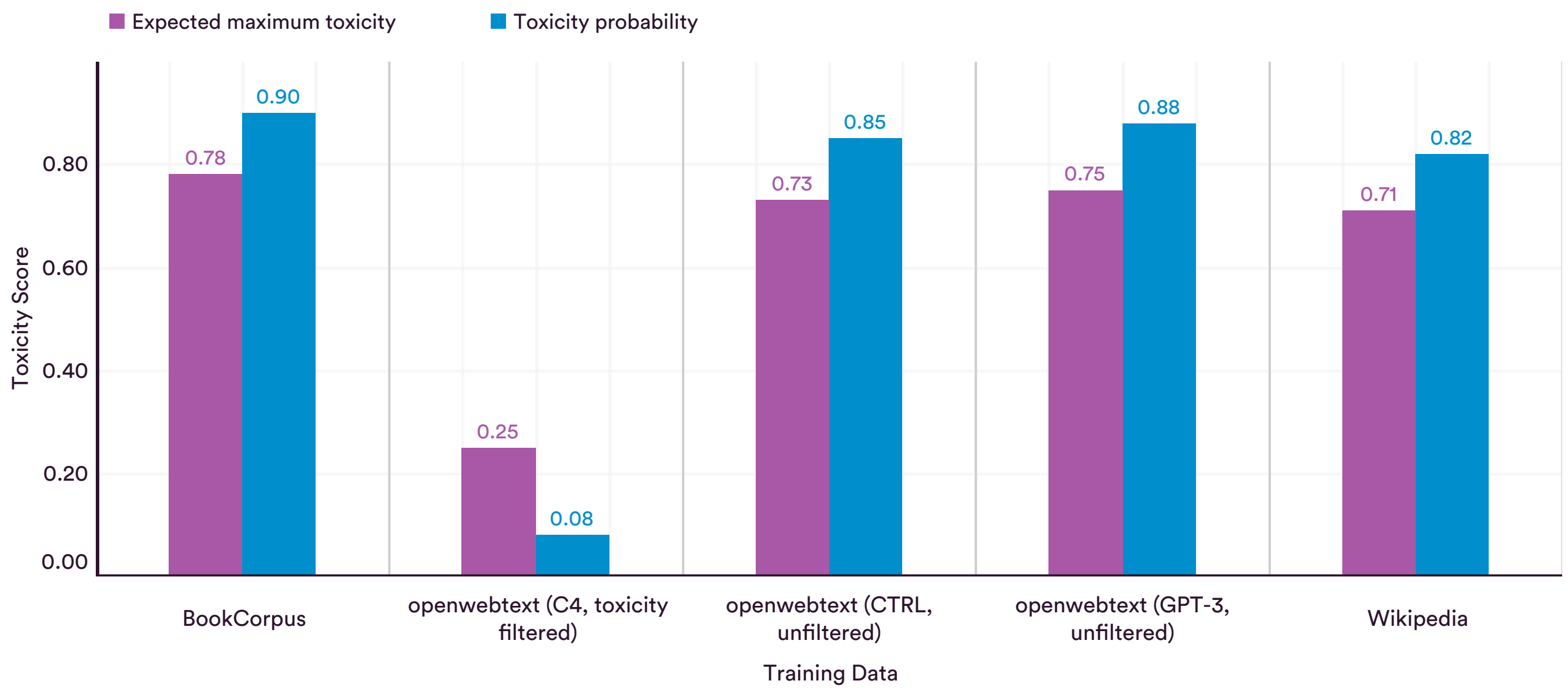

Figure 3.2.2

## Large Language Models and Toxicity

Recent developments around mitigating toxicity in language models have lowered both expected maximum toxicity and the probability of toxicity. However, detoxification methods consistently lead to adverse side effects and somewhat less capable models. (For example, filtering training data typically comes at the cost of model performance.)

In December 2021, DeepMind released a paper describing its 280 billion parameter language model, Gopher. Figure 3.2.3a and Figure 3.2.3b from the Gopher paper show that larger models are more likely to produce toxic outputs when prompted with inputs of varying levels of toxicity, but that they are also more capable of detecting toxicity with regard to their own outputs as well as in other contexts, as measured by increased AUC (area under the receiver operating characteristic curve) with model size. The AUC metric plots the true positive rate against the false positive rate to characterize how well a model distinguishes between classes (higher is better). Larger models are dramatically better at identifying toxic comments within the CivilComments dataset, as shown in Figure 3.2.3b.

**GOPHER: PROBABILITY of TOXIC CONTINUATIONS BASED on PROMPT TOXICITY by MODEL SIZE**
Source: Rae et al., 2021 | Chart: 2022 AI Index Report

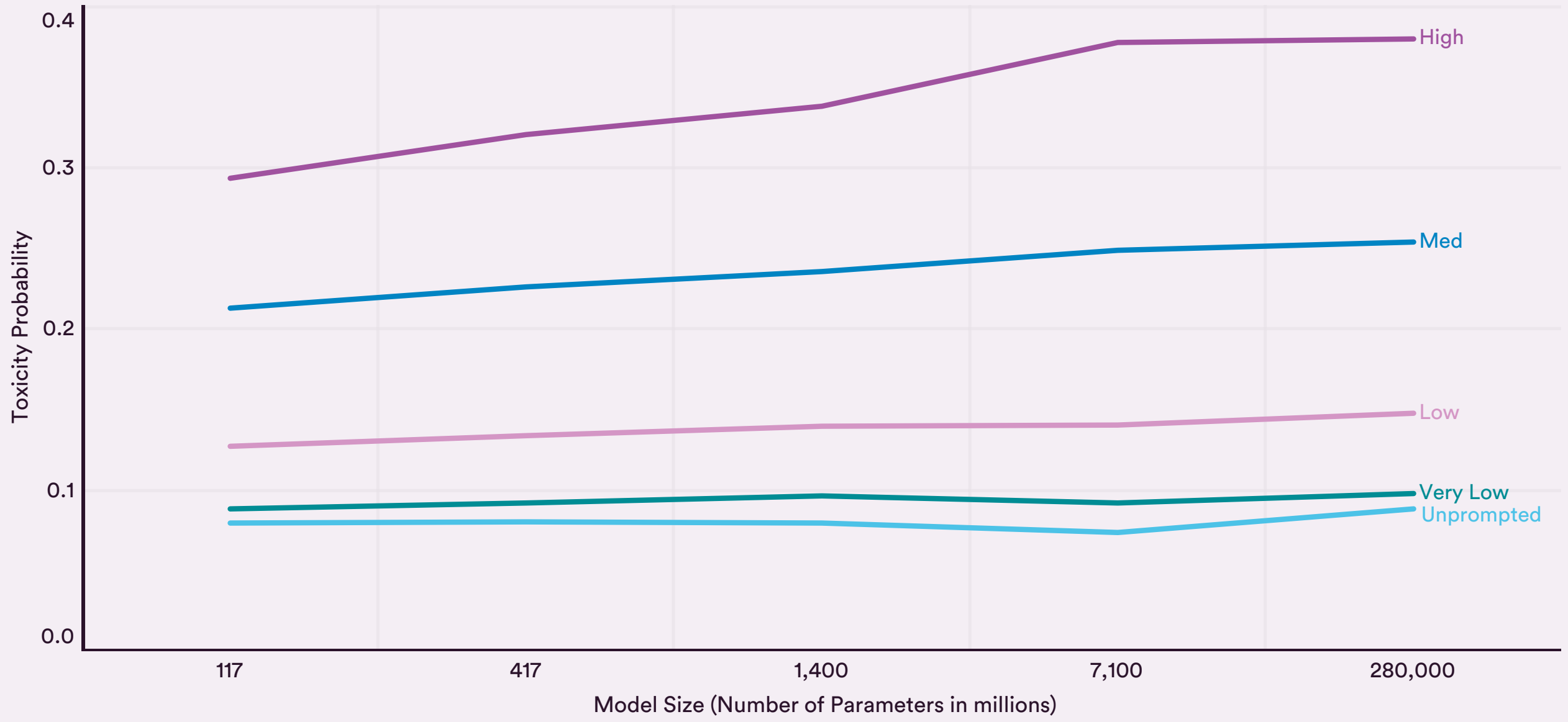

Figure 3.2.3a

# Large Language Models and Toxicity (cont'd)

GOPHER: FEW-SHOT TOXICITY CLASSIFICATION on the CIVILCOMMENTS DATASET
Source: Rae et al., 2021 | Chart: 2022 AI Index Report

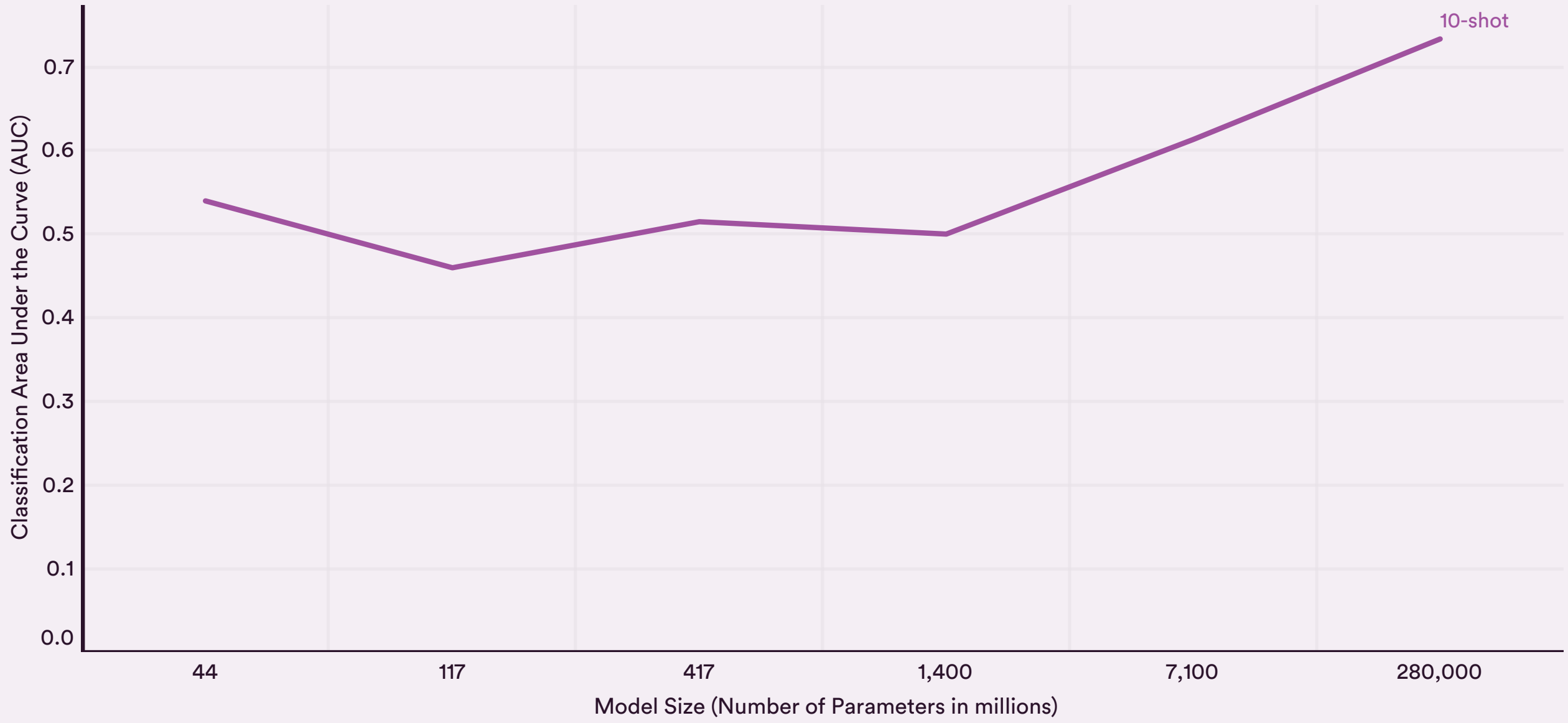

Figure 3.2.3b

## DETOXIFICATION OF MODELS CAN NEGATIVELY INFLUENCE PERFORMANCE

Detoxification methods aim to mitigate toxicity by changing the underlying training data as in domain-adaptive pretraining (DAPT), or by steering the model during generation as in Plug and Play Language Models (PPLM) or Generative Discriminator Guided Sequence Generation (GeDi).

A study on detoxifying language models shows that models detoxified with these strategies all perform worse on both white-aligned and African American English on perplexity, a metric that measures how well a model has learned a specific distribution (lower is better) (Figure 3.2.4). These models also perform disproportionately worse on African American English and text containing mentions of minority identities compared to white-aligned text, a result that is likely due to human biases causing annotators to be more apt to mislabel African American English as toxic.

**PERPLEXITY: LANGUAGE MODELING PERFORMANCE by MINORITY GROUPS on ENGLISH POST-DETOXIFICATION**
Source: Xu et al., 2021 | Chart: 2022 AI Index Report

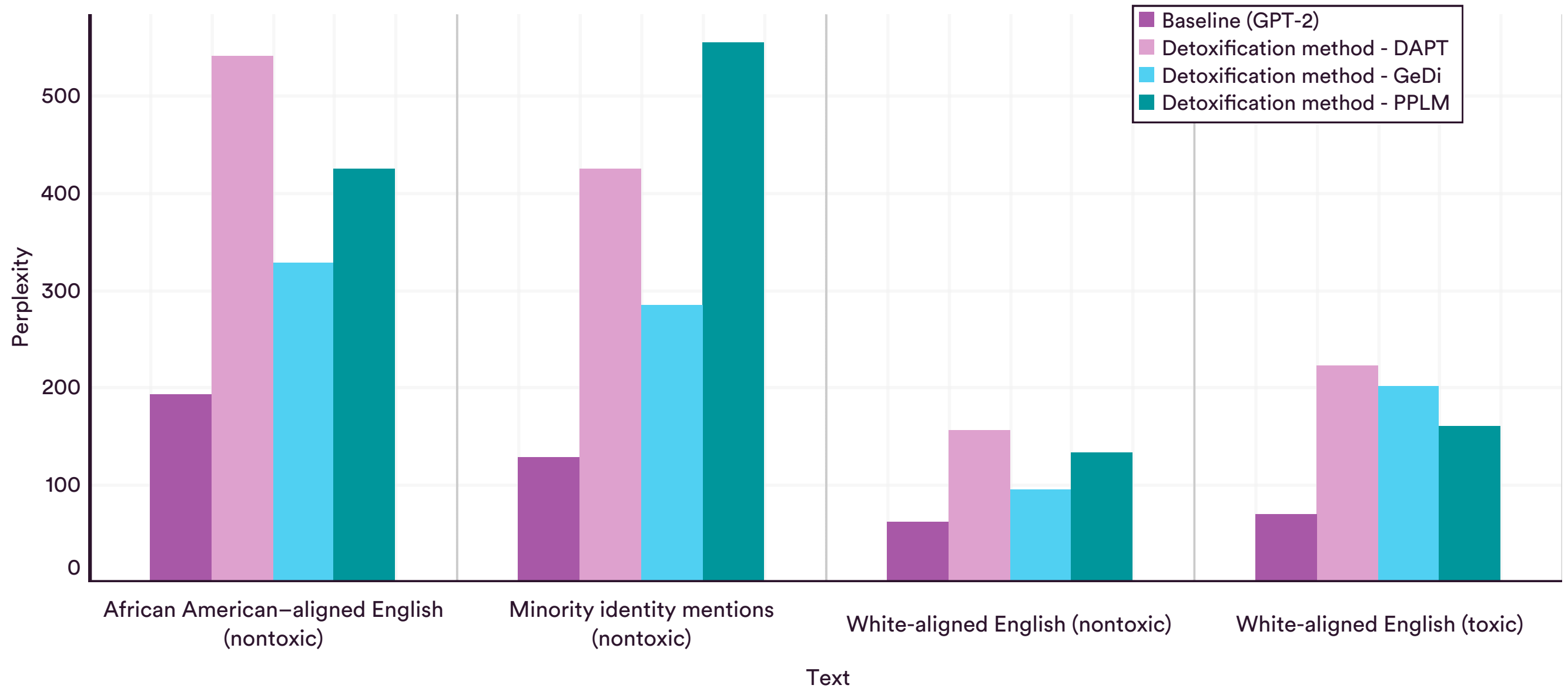

Figure 3.2.4

## STEREOSET

StereoSet is a benchmark measuring stereotype bias along the axes of gender, race, religion, and profession, along with raw-language modeling ability. One of the associated metrics is stereotype score, which measures whether a model prefers stereotypes and anti-stereotypes equally. A stereotype is an over-generalized belief widely held about a group and an anti-stereotype is a generalization about a group which contradicts widely accepted stereotypes.

Figure 3.2.5 shows that StereoSet performance follows the same trend seen with toxicity: Larger models reflect stereotypes more often unless interventions are taken to reduce learned stereotypes during training. The prevalence of toxic content online has been estimated to be 0.1–3%, which aligns with research showing that larger language models are more capable of memorizing rare text.

**STEREOSET: STEREOTYPE SCORE by MODEL SIZE**
Source: Nadeem et al., 2020; Lieber et al., 2021; StereoSet Leaderboard, 2021; | Chart: 2022 AI Index Report

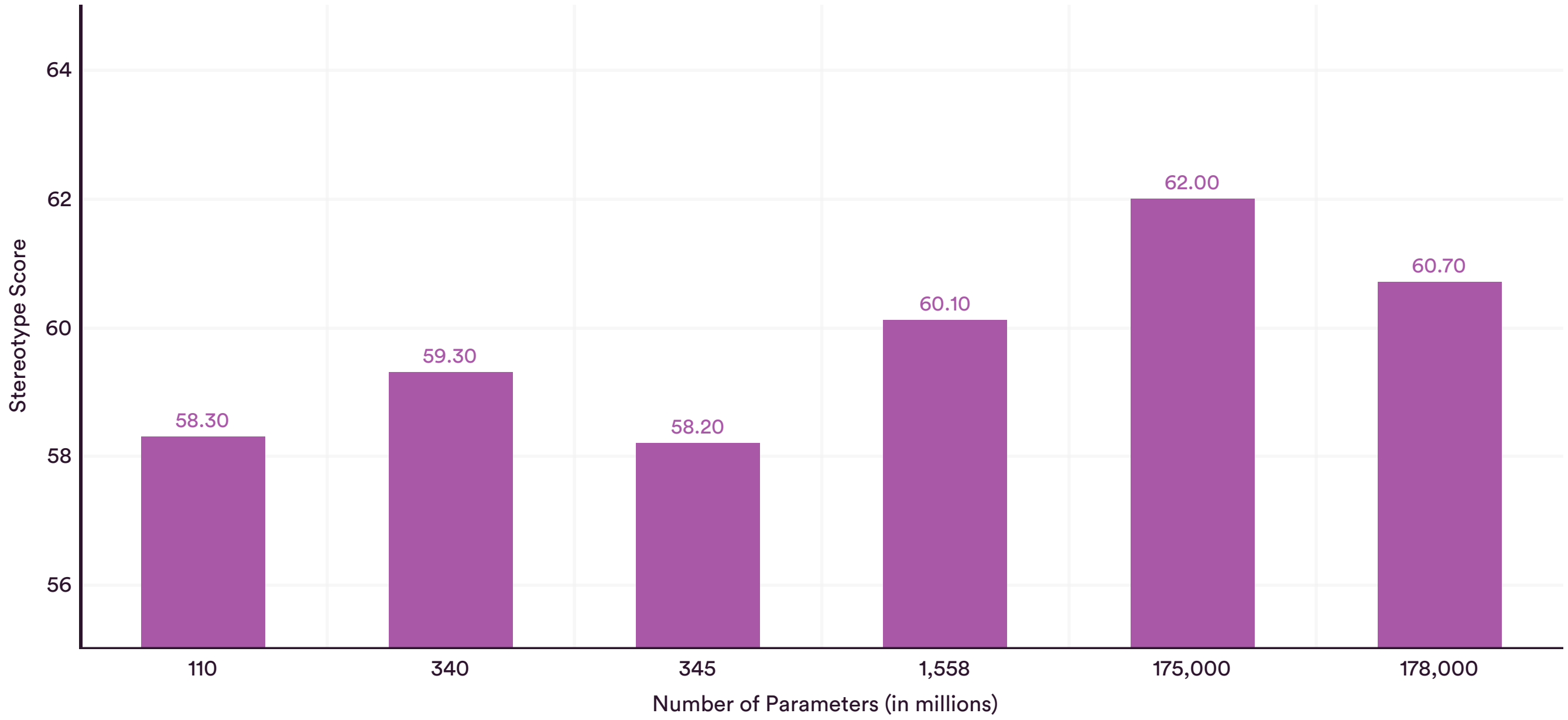

Figure 3.2.5

StereoSet has several major flaws in its underlying dataset: Some examples fail to express a harmful stereotype, conflate stereotypes about countries with stereotypes about race and ethnicity, and confuse stereotypes between associated but distinct groups. Additionally, these stereotypes were sourced from crowdworkers located in the United States, and the resulting values and stereotypes within the dataset may not be universally representative.

## CROWS-PAIRS

CrowS-Pairs (Crowdsourced Stereotype Pairs) is another benchmark measuring stereotype bias. While StereoSet compares attributes about a single group, CrowS-Pairs contrasts relationships between historically disadvantaged and advantaged groups (e.g., Mexicans versus white people).

The creators of CrowS-Pairs measured stereotype bias using three popular language models: BERT, RoBERTa, and ALBERT (Figure 3.2.6). On standard language modeling benchmarks, ALBERT outperforms RoBERTa, which outperforms BERT.[4] However, ALBERT is the most biased of the three models according to CrowS-Pairs. This mirrors the trend observed with StereoSet and RealToxicityPrompts: More capable models are also more capable of learning and amplifying stereotypes.

Like earlier examples, BERT, RoBERTa, and ALBERT appear to inherit biases from their training data. They were all trained on a combination of BookCorpus, English Wikipedia, and text scraped from the internet. Analysis of BookCorpus reveals that its books about religion are heavily skewed toward Christianity and Islam compared to other major world religions,[5] though it is unclear the extent to which these books contain historical content versus content written from a specific religious viewpoint.[6]

We can examine how language models may inherit biases about certain religions by looking at their underlying datasets. Figure 3.2.7 shows the number of books pertaining to different religions in two popular datasets, BookCorpus and Smashwords21. Both datasets have far more mentions of Christianity and Islam than other religions.

**CROWS-PAIRS: LANGUAGE MODEL PERFORMANCE across BIAS ATTRIBUTES**
Source: Nangia et al., 2020 | Chart: 2022 AI Index Report

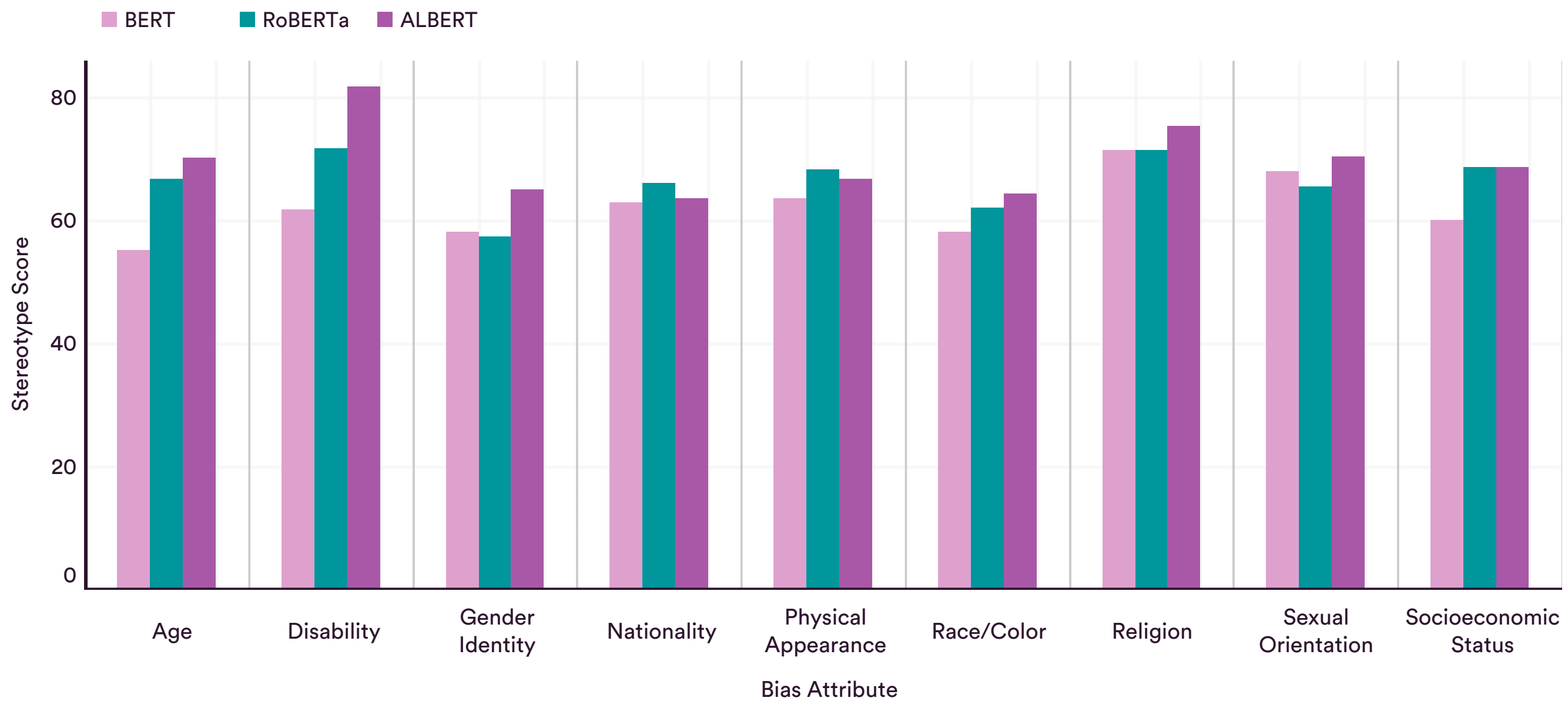

Figure 3.2.6

4 Per results on the SQuAD, GLUE, and RACE benchmarks.
5 Such as Sikhism, Judaism, Hinduism, Buddhism, Atheism.
6 Hate speech classifiers fine-tuned on top of BERT in particular have been shown to frequently misclassify texts containing mentions of "Muslim" as toxic, and researchers find that GPT-3 contains significant bias along religious axes for mentions of both "Jewish" and "Muslim."

## BOOKCORPUS and SMASHWORDS21: SHARE of BOOKS about RELIGION in PRETRAINING DATASETS

Source: Bandy and Vincent, 2021 | Chart: 2022 AI Index Report

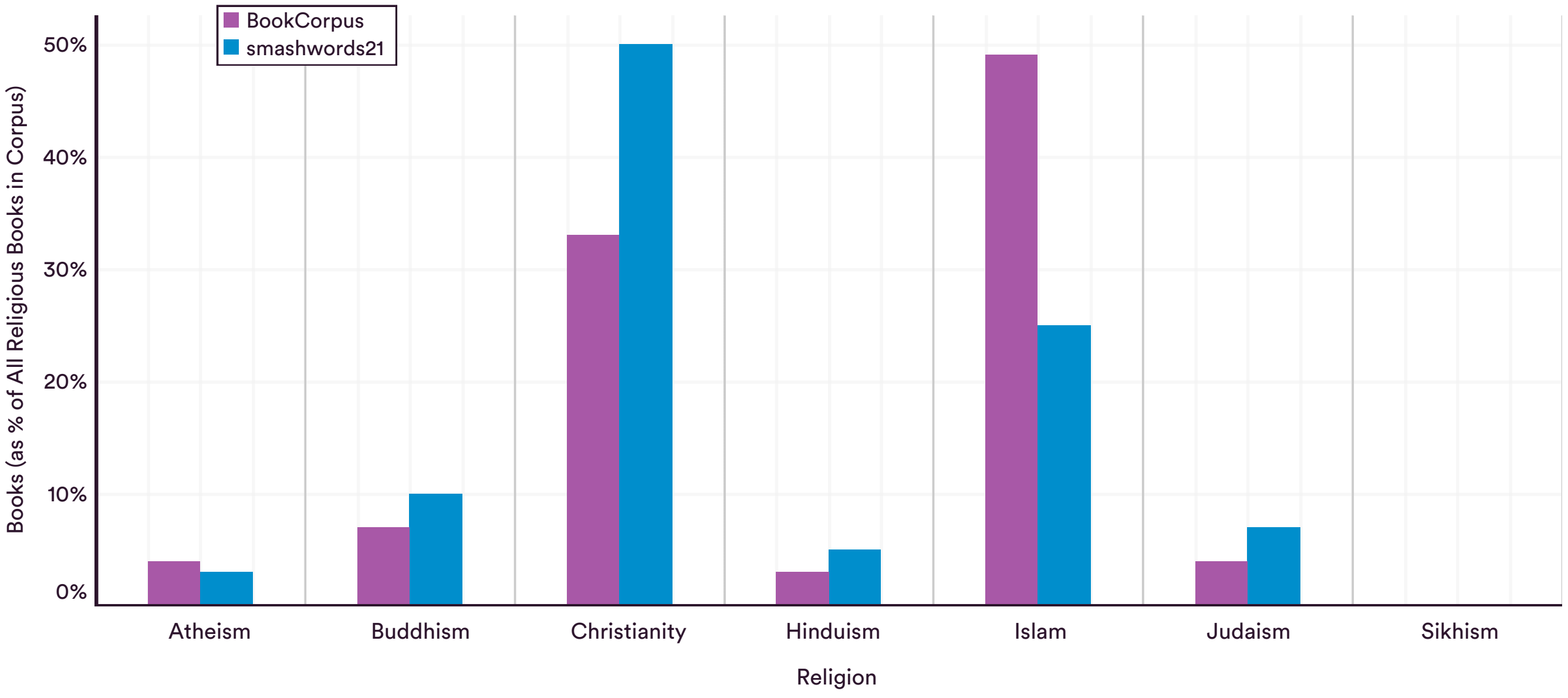

Figure 3.2.7

## WINOGENDER AND WINOBIAS

Winogender measures gender bias related to occupations. Systems are measured on their ability to fill in the correct gender in a sentence containing an occupation (e.g., "The teenager confided in the therapist because he / she seemed trustworthy"). Examples were created by sourcing data from the U.S. Bureau of Labor Statistics to identify occupations skewed toward one gender (e.g., the cashier occupation is made up of 73% women, but drivers are only 6% women).

Performance on Winogender is measured by the accuracy gap between the stereotypical and anti-stereotypical cases, along with the gender parity score (the percentage of examples for which the predictions are the same). The authors use crowdsourced annotations to estimate human performance to be 99.7% accuracy.

Winogender results from the SuperGLUE leaderboard show that larger models are more capable of correctly resolving gender in the zero-shot and few-shot setting (i.e., without fine-tuning on the Winogender task) and less likely to magnify occupational gender disparities (Figure 3.2.8). However, a good score on Winogender does not indicate that a model is unbiased with regard to gender, only that bias was not captured by this benchmark.

**MODEL PERFORMANCE on the WINOGENDER TASK from the SUPERGLUE BENCHMARK**
Source: SuperGLUE Leaderboard, 2021 | Chart: 2022 AI Index Report

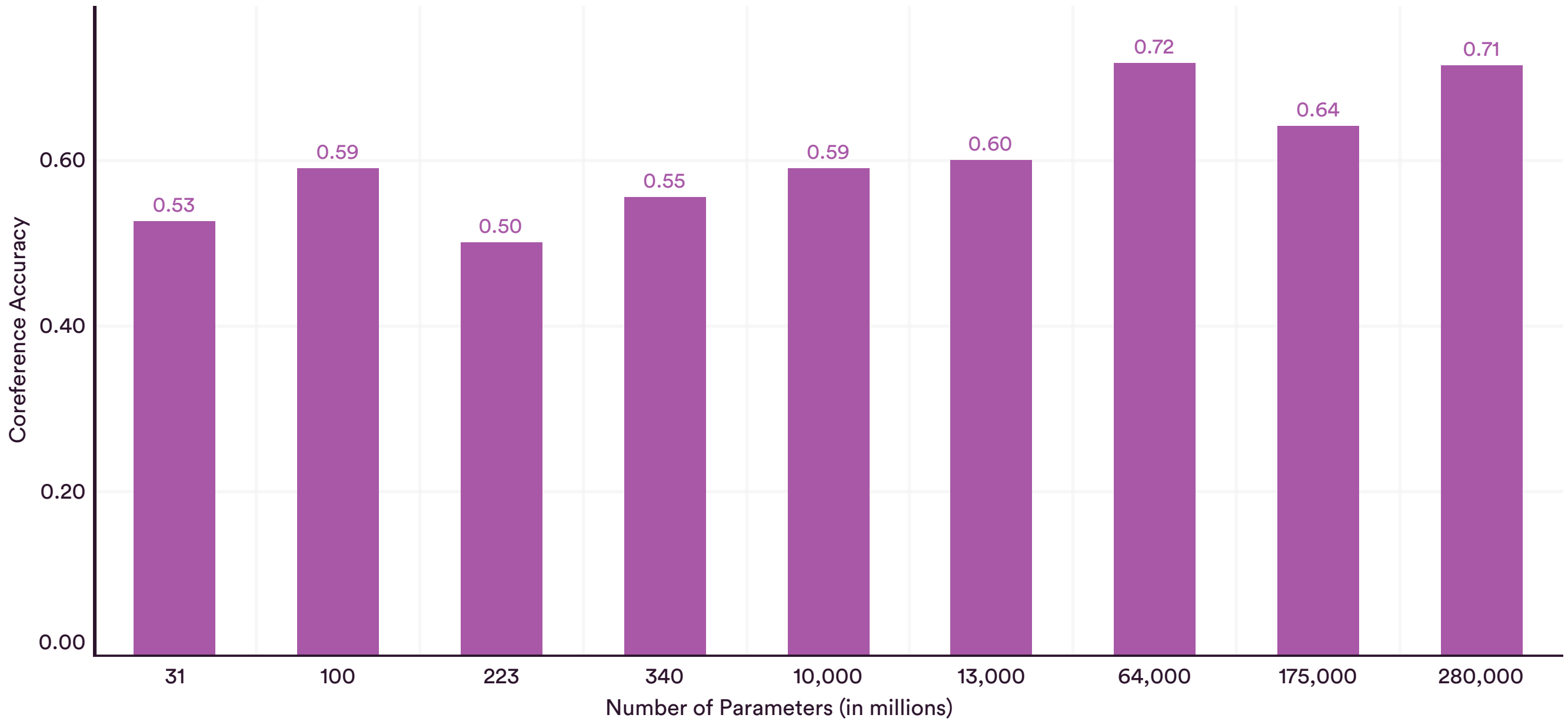

Figure 3.2.8

WinoBias is a similar benchmark measuring gender bias related to occupations that was released concurrently with Winogender by a different research group. As shown in Figure 3.2.9, WinoBias is cited more often than Winogender, but the adoption of Winogender within the SuperGLUE leaderboard for measuring natural language understanding has led to more model evaluations being reported on Winogender.

**WINOBIAS and WINOGENDER: NUMBER of CITATIONS, 2018–21**
Source: AI Index, 2021; Semantic Scholar, 2021 | Chart: 2022 AI Index Report

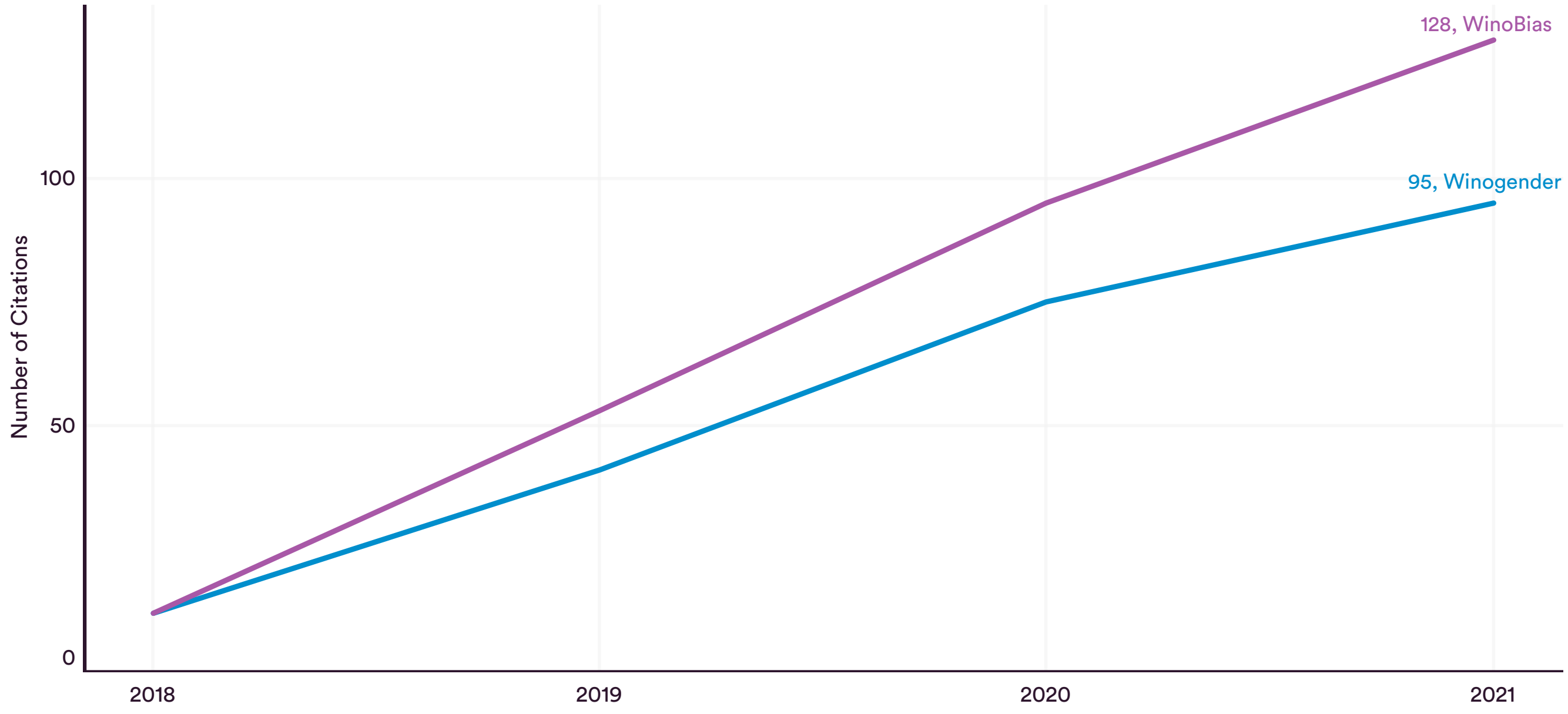

Figure 3.2.9

## WINOMT: GENDER BIAS IN MACHINE TRANSLATION SYSTEMS

Commercial machine translation systems have been documented to reflect and amplify societal biases from their underlying datasets. As these systems are used broadly in global industries such as e-commerce, stereotypes and mistakes in translation can be costly.

WinoMT is a benchmark measuring gender bias in machine translation that is created by combining the Winogender and WinoBias datasets. Models are evaluated by comparing the sentences translated from English to another language and extracting the translated gender to compare with the original gender. Systems are scored on the percentage of translations with correct gender (gender accuracy), the difference in F1 score between masculine and feminine examples, and the difference in F1 score between examples with stereotypical gender roles and anti-stereotypical gender roles.

As seen in Figure 3.2.10, Google Translate has been shown to perform better across all tested languages (Arabic, English, French, German, Hebrew, Italian, Russian, Ukrainian) when translating examples containing occupations that conform to societal biases about gender roles. Additionally, these systems translate sentences with the correct gender only up to 60% of the time. Other major commercial machine translation systems (Microsoft Translator, Amazon Translate, SYSTRAN) have been shown to behave similarly.

**WINOMT: GENDER BIAS in GOOGLE TRANSLATE across LANGUAGES**
Source: Stanovsky et al., 2019 | Chart: 2022 AI Index Report

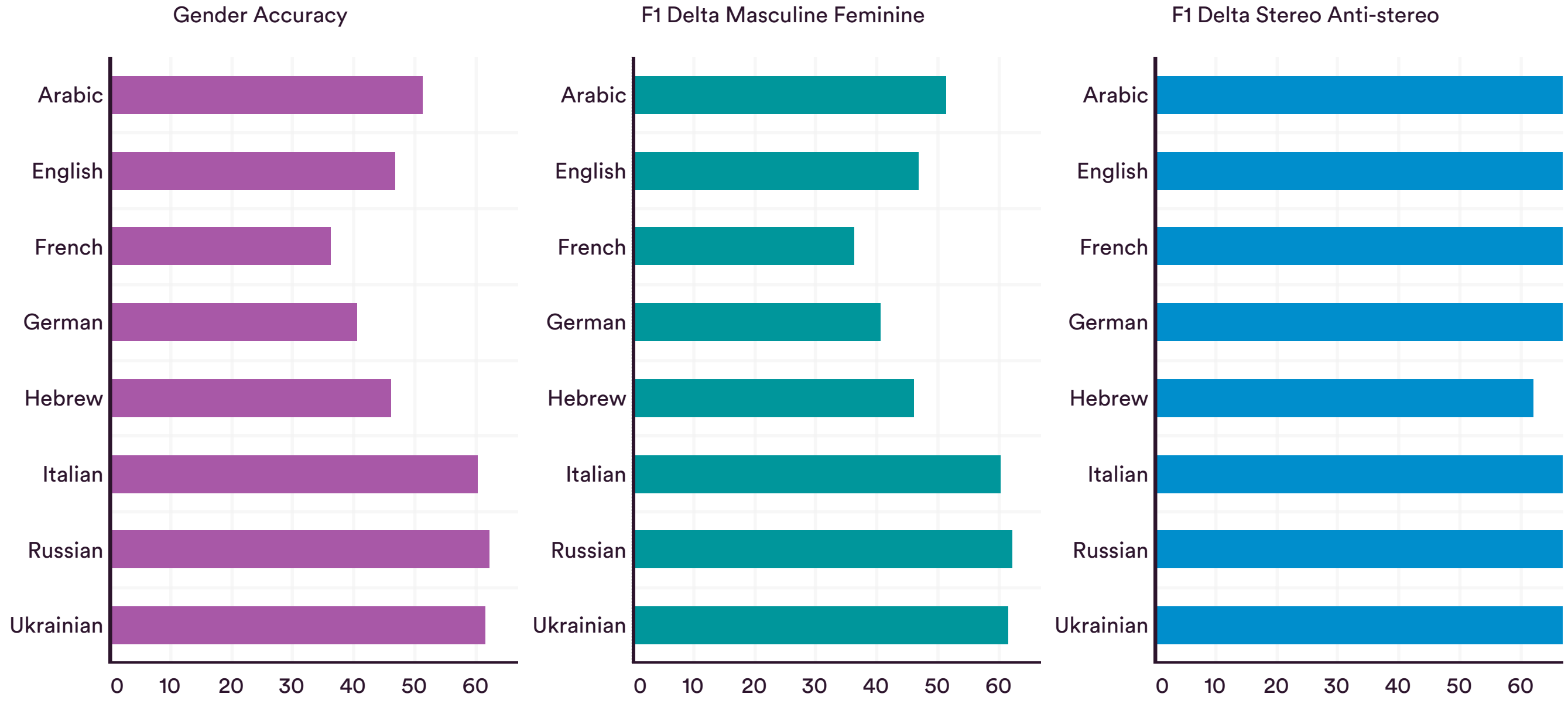

Figure 3.2.10

## WORD AND IMAGE EMBEDDING ASSOCIATION TESTS

Word embedding is a technique in NLP that allows words with similar meanings to have similar representations. *Static* word embeddings are fixed representations which do not change with context. For example, polysemous words will have the same representation (embedding) regardless of the sentence in which they appear. Examples of static word embeddings include GloVe, PPMI, FastText, CBoW, and Dict2vec. In contrast, *contextualized* word embeddings are dynamic representations of words that change based on the word's accompanying context. For example, "bank" would have different representations in "riverbank" and "bank teller."

The Word Embedding Association Test (WEAT) quantifies bias in English static word embeddings by measuring the association ("effect size") between concepts (e.g., European-American and African American names) and attributes (e.g., pleasantness and unpleasantness). Word embeddings trained on large public corpora (e.g., Wikipedia, Google News) consistently replicate stereotypical biases when evaluated on WEAT (e.g., associating male terms with "career" and female terms with "family"). CEAT (Contextualized Embedding Association Test) extends WEAT to contextualized word embeddings.

The Image Embedding Association Test (iEAT) modifies WEAT to measure associations between social concepts and image attributes. Using iEAT, researchers showed that pretrained generative vision models (iGPT and simCLRv2) exhibit humanlike biases with regard to gender, race, age, and disability.

Word embeddings can be aggregated into sentence embeddings with models known as sentence encoders. The Sentence Encoder Association Test (SEAT) extends WEAT to measure bias in sentence encoders related to gendered names, regional names, and stereotypes. Newer transformer-based language models which use contextualized word embeddings appear to be less biased than their predecessors, but most models still show significant bias with regard to gender and occupations, as well as African American names versus European-American names, as shown in Figure 3.2.11.

**SENTENCE EMBEDDING ASSOCIATION TEST (SEAT): MEASURING STEREOTYPICAL ASSOCIATIONS with EFFECT SIZE**

Source: May et al., 2019 | Chart: 2022 AI Index Report

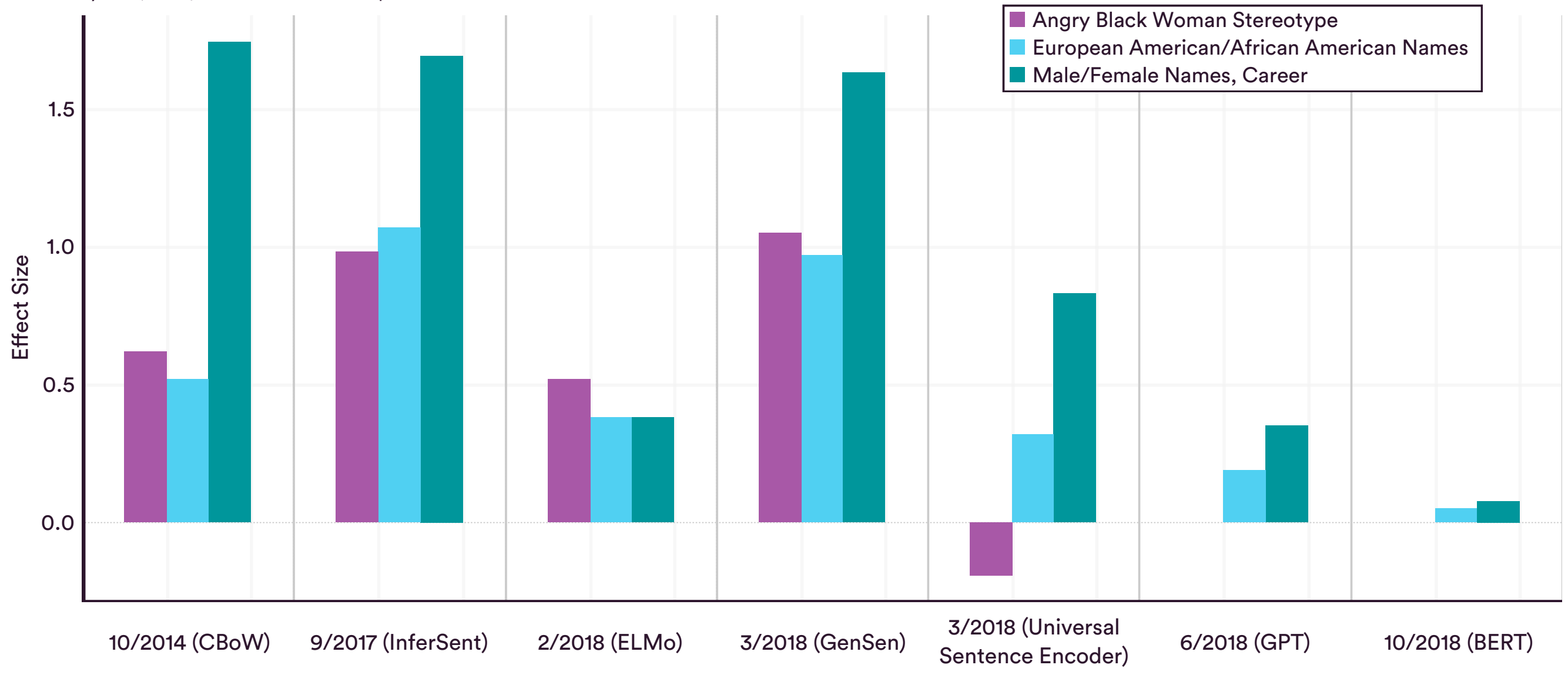

Figure 3.2.11

Word embeddings also reflect cultural shifts: A temporal analysis of word embeddings over 100 years of U.S. Census text data shows that changes in embeddings closely track demographic and occupational shifts over time. Figure 3.2.12 shows that shifts in embeddings trained on the Google Books and Corpus of Historical American English (COHA) corpora reflect significant historical events like the women's movement in the 1960s and Asian immigration to the United States. In this analysis, embedding bias is measured with the relative norm difference: the average Euclidean distance between words associated with representative groups (e.g., men, women, Asians) and words associated with occupations. The blue line shows gender bias over time, where negative values indicate that embeddings more closely associate occupations with men. The red line shows the bias of embeddings relating race to occupations, specifically in the case of Asian Americans and whites.

**GENDER and RACIAL BIAS in WORD EMBEDDINGS TRAINED on 100 YEARS of TEXT DATA**

Source: Garg et al., 2018 | Chart: 2022 AI Index Report

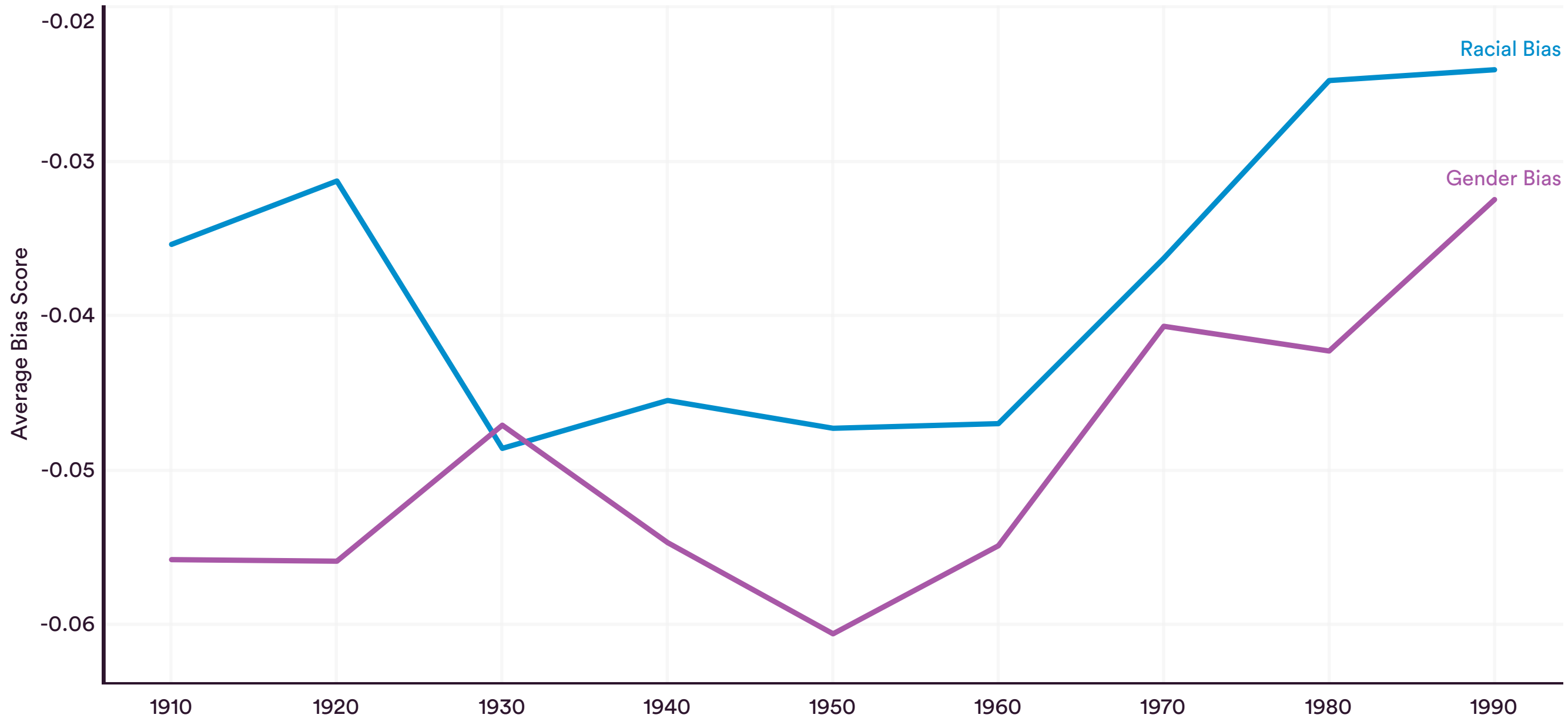

Figure 3.2.12

## Multilingual Word Embeddings

Large language models are often monolingual since they require a significant amount of text data to train. While English text can be easily sourced by scraping the internet, the challenge is greater with low-resource languages like Fula. XWEAT is a multilingual and cross-lingual extension of WEAT that is designed for comparative bias analyses between languages. Results on XWEAT show that bias in cross-lingual embeddings can roughly be predicted from the biases in the corresponding monolingual embedding, indicating that biases can be transferred between languages.

Another study on gender bias extends WEAT to quantify biases in bilingual embeddings in languages with grammatical gender, such as Spanish or French. Figure 3.2.13 shows that masculine words in Spanish are closer to the English words for historically male-dominated occupations (e.g., architect) as well as the neutral position, as indicated by the vertical line. Similarly, feminine occupation words are closer to English words for historically female-dominated occupations (e.g., nurse).

GENDER BIAS in SPANISH WORD EMBEDDINGS: EMBEDDING SIMILARITY DISTANCE
Source: Zhou et al., 2019 | Chart: 2022 AI Index Report

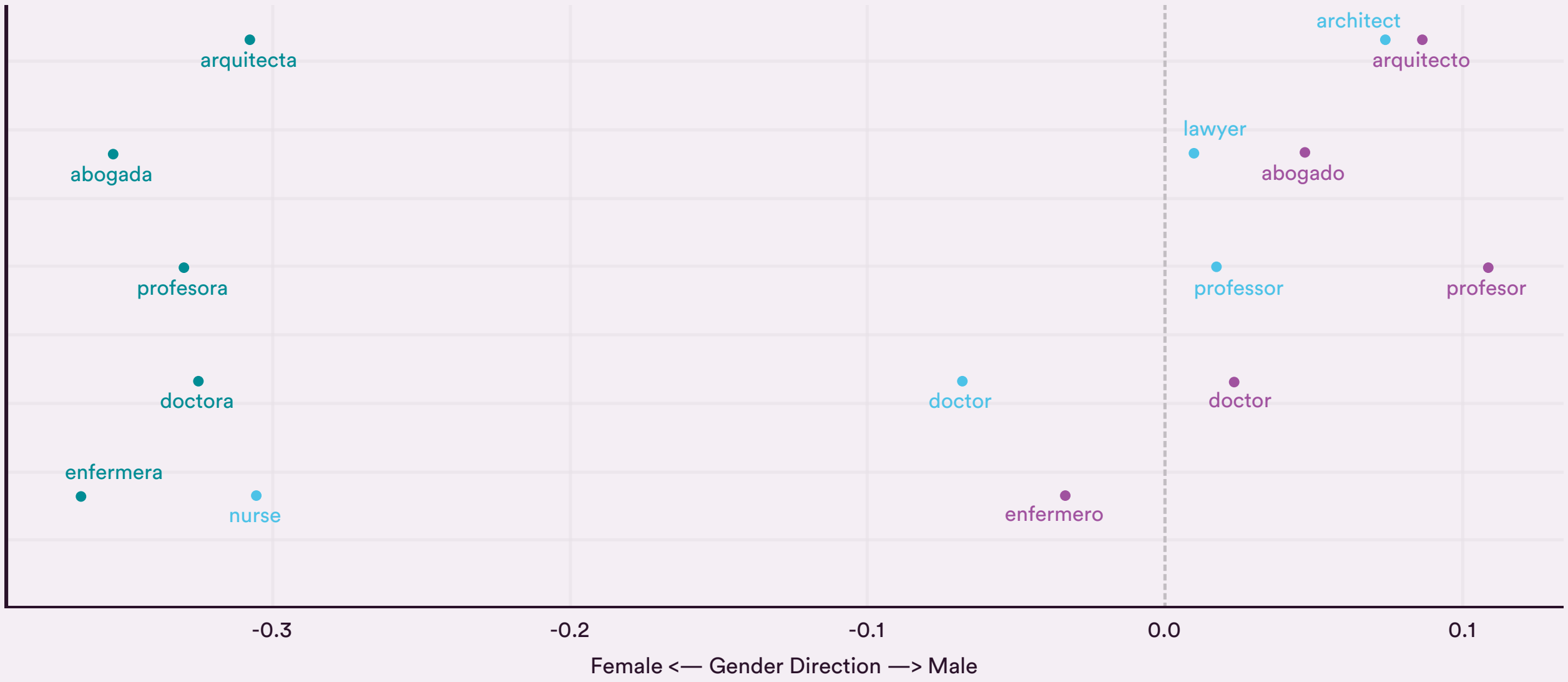

Figure 3.2.13

### Mitigating Bias in Word Embeddings With Intrinsic Bias Metrics

It is often assumed that reducing intrinsic bias by de-biasing embeddings will reduce downstream biases in applications (extrinsic bias). However, it has been demonstrated that there is no reliable correlation between intrinsic bias metrics and downstream application biases. Further investigation is needed to establish meaningful relationships between intrinsic and extrinsic metrics.

**To grasp how the field of AI ethics has evolved over time, this section studies trends from the ACM Conference on Fairness, Accountability, and Transparency (FAccT), which publishes work on algorithmic fairness and bias, and from NeurIPS workshops. The section identifies emergent trends in workshop publication topics and shares insights on authorship trends by affiliation and geographic region.**

# 3.3 AI ETHICS TRENDS AT FACCT AND NEURIPS

## ACM CONFERENCE ON FAIRNESS, ACCOUNTABILITY, AND TRANSPARENCY (FACCT)

ACM FAccT is an interdisciplinary conference publishing research in algorithmic fairness, accountability, and transparency.[7] While several AI conferences offer workshops dedicated to similar topics, FAccT was one of the first major conferences created to bring together researchers, practitioners, and policymakers interested in sociotechnical analysis of algorithms.

Figure 3.3.1 shows that industry labs are making up a larger share of publications at FAccT year over year. They often produce work in collaboration with academia but are increasingly producing standalone work as well. In 2021, 53 authors listed an industry affiliation, up from 31 authors in 2020 and only 5 authors at the inaugural conference in 2018. This aligns with recent findings that point to a trend of deep learning researchers transitioning from academia to industry labs.

**NUMBER of ACCEPTED FACCT CONFERENCE SUBMISSIONS by AFFILIATION, 2018–21**
Source: FAccT, 2021; AI Index, 2021 | Chart: 2022 AI Index Report

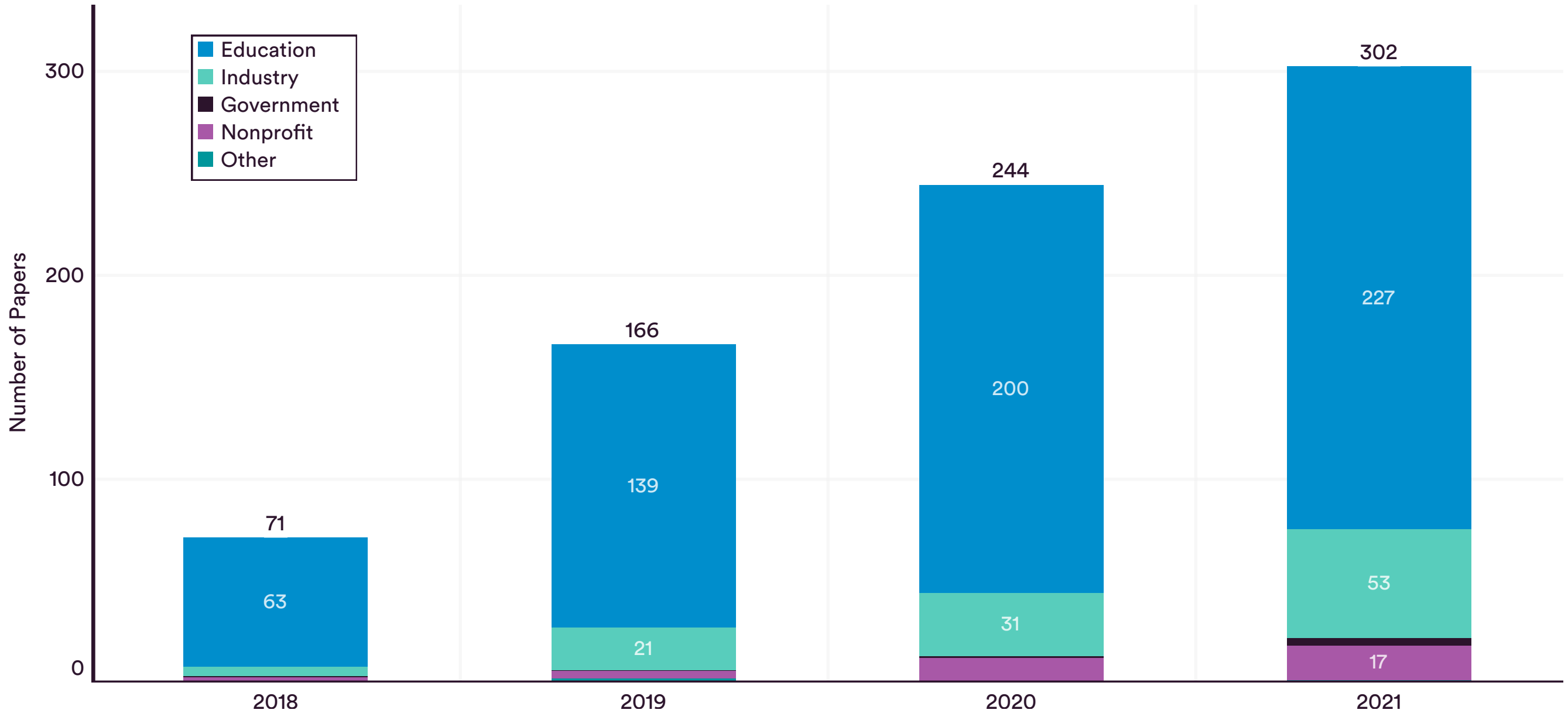

Figure 3.3.1

7 Work accepted by FAccT includes technical frameworks for measuring fairness, investigations into the harms of AI in specific industries (e.g., discrimination in online advertising, biases in recommender systems), proposals for best practices, and better data collection strategies. Several works published at FAccT have become canonical works in AI ethics; examples include Model Cards for Model Reporting (2019) and On the Dangers of Stochastic Parrots: Can Language Models Be Too Big? (2021). Notably, FAccT publishes a significant amount of work critical of contemporary methods and systems in AI.

While there has been increased interest in fairness, accountability, and transparency research from all types of organizations, the majority of papers published at FAccT are written by researchers based in the United States, followed by researchers based in Europe and Central Asia (Figure 3.3.2). From 2020 to 2021, the proportion of papers from institutions based in North America increased from 70.2% to 75.4%.

NUMBER of ACCEPTED FACCT CONFERENCE SUBMISSIONS by REGION, 2018–21

Source: FAccT, 2021; AI Index, 2021 | Chart: 2022 AI Index Report

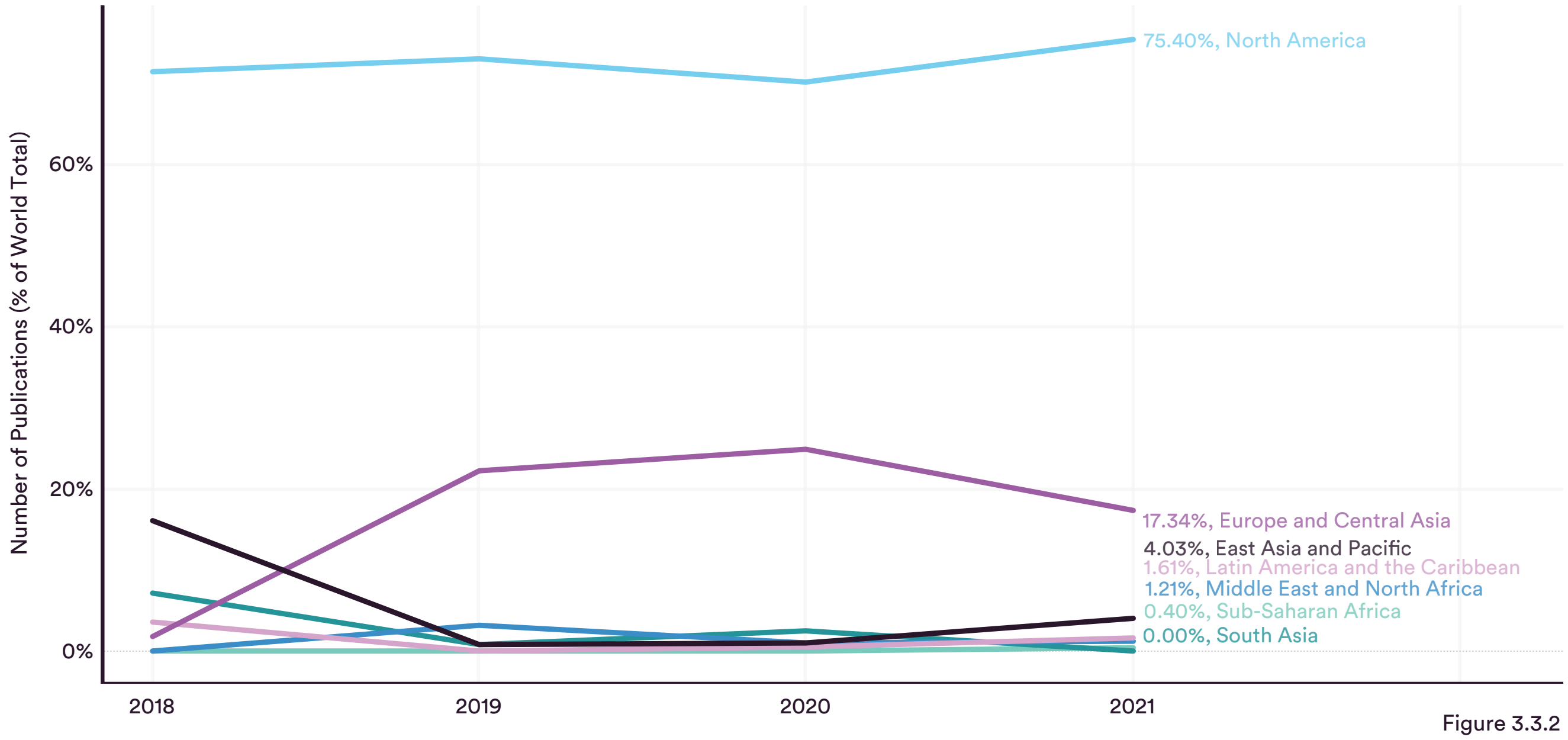

Figure 3.3.2

## NEURIPS WORKSHOPS

NeurIPS, one of the largest AI conferences, held its first workshop on fairness, accountability, and transparency in 2014. Figure 3.3.3 shows the number of research papers at NeurIPS ethics-related workshops in the past six years by research topic, indicating an increased interest in AI applied to high-risk, high-impact use cases such as climate, finance, and healthcare.

**NEURIPS WORKSHOP RESEARCH TOPICS: NUMBER of ACCEPTED PAPERS on REAL-WORLD IMPACTS, 2015–21**

Source: NeurIPS, 2021; AI Index, 2021 | Chart: 2022 AI Index Report

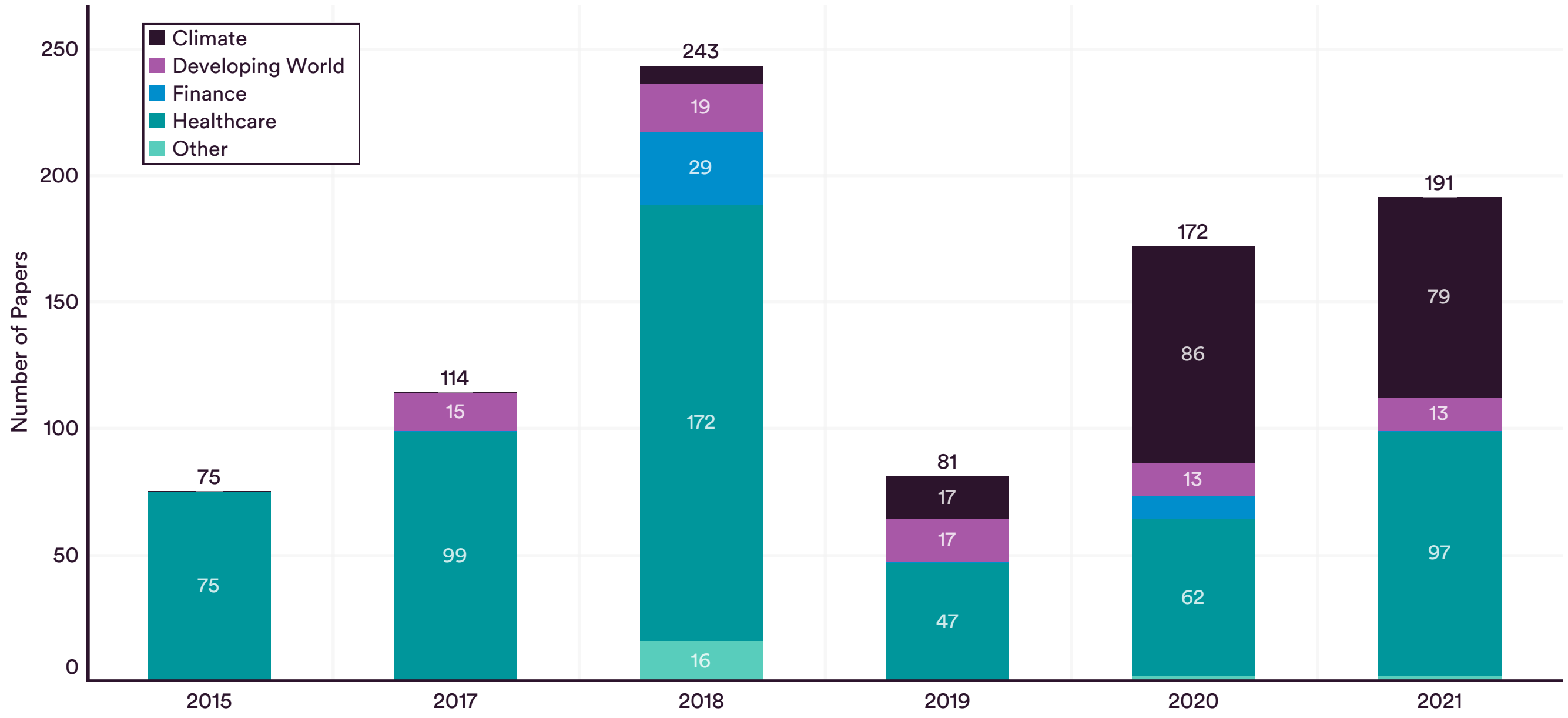

Figure 3.3.3

## Interpretability, Explainability, and Causal Reasoning

Several workshops have been created at NeurIPS around interpretability and explainability, including safety-critical AI affecting human decisions,[8] interpretability and causality for algorithmic fairness,[9] and the necessity of explainability for high-risk use cases.[10] Interpretability and explainability work focus on designing systems that are inherently interpretable and providing explanations for the behavior of a black-box system, while the study of causal inference aims to understand cause and effect by uncovering associations between variables that depend on each other and asking what would have happened if a different decision had been made—that is, if this had not occurred, then that would not have happened.

Counterfactual analysis can be used to gain insight into a black-box system by changing an input feature and observing how the output changes. This can be applied to measure fairness by changing protected attributes of an individual input (e.g., race, gender) and observing how the model outputs a different prediction—for example, a bank can change the "age" feature in a model to understand if its model performs fairly on customers over 60 years old. Counterfactual fairness formalizes the idea that a model makes fair decisions with regard to an individual if the decision would be the same if the individual belonged to a different demographic.

Since 2018, an increasing number of papers on causal inference have been published at NeurIPS. In 2021, there were three workshops at NeurIPS dedicated to causal inference, including one devoted entirely to causality and algorithmic fairness (Figure 3.3.4). Figure 3.3.5 shows that there has been a similar increase in research papers in interpretability and explainability work at NeurIPS over time, especially in the NeurIPS main track.

**NEURIPS RESEARCH TOPICS: NUMBER of ACCEPTED PAPERS on CAUSAL EFFECT and COUNTERFACTUAL REASONING, 2015–2021**
Source: NeurIPS, 2021; AI Index, 2021 | Chart: 2022 AI Index Report

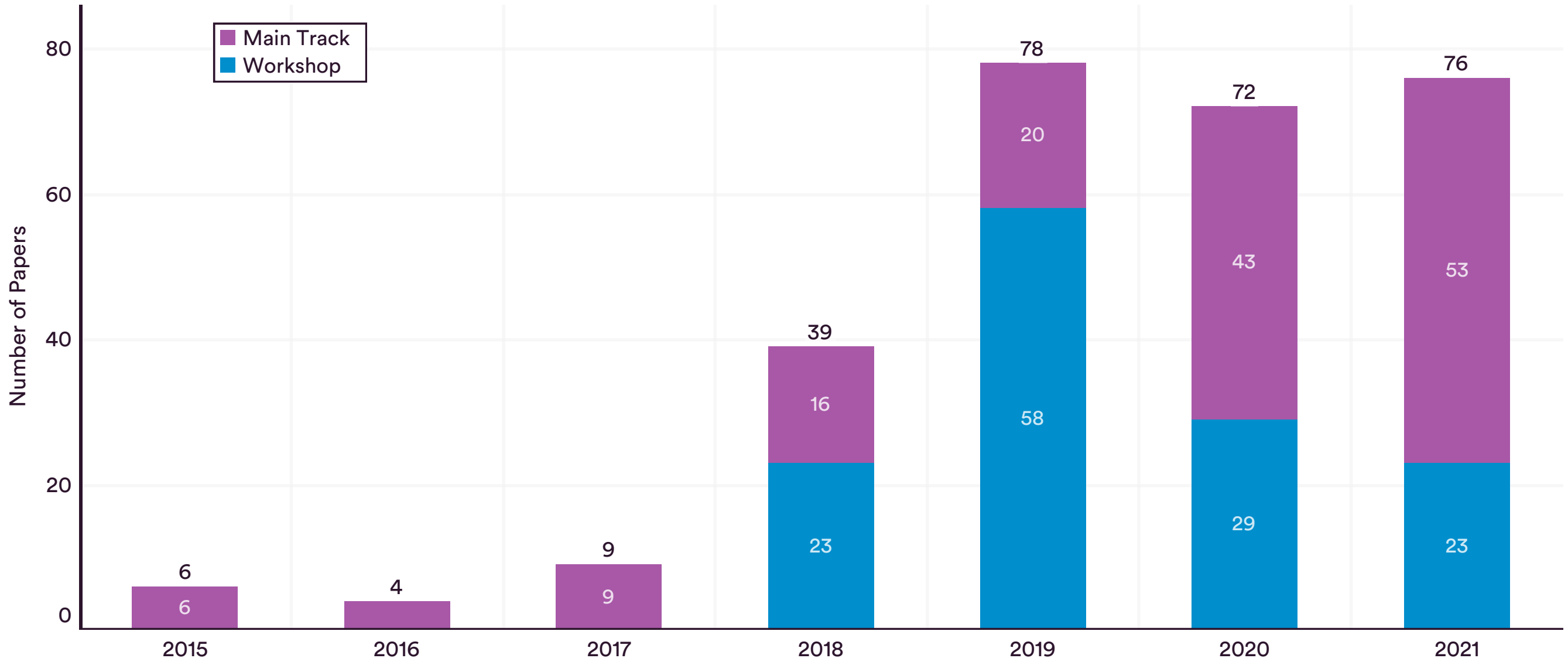

Figure 3.3.4

8 See 2017 Transparent and Interpretable Machine Learning in Safety Critical Environments, 2019 Workshop on Human-Centric Machine Learning: Safety and Robustness in Decision-Making, 2019, "'Do the Right Thing': Machine Learning and Causal Inference for Improved Decision-Making."
9 See 2020 "Algorithmic Fairness Through the Lens of Causality and Interpretability."
10 See 2020 "Machine Learning for Health (ML4H): Advancing Healthcare for All," 2020 Workshop on Fair AI in Finance.

## Privacy and Data Collection

Amid growing concerns about privacy, data sovereignty, and the commodification of personal data for profit, there has been significant momentum in industry and academia to build methods and frameworks to help mitigate privacy concerns. Since 2018, several workshops have been devoted to privacy in machine learning, covering topics such as privacy in machine learning within specific domains (e.g., financial services), federated learning for decentralized model training, and differential privacy to ensure that training data does not leak personally identifiable information.[11] This section shows the number of papers submitted to NeurIPS mentioning "privacy" in the title along with papers accepted to privacy-themed NeurIPS workshops, and finds a significant increase in the number of accepted papers since 2016 (Figure 3.3.6).

**NEURIPS RESEARCH TOPICS: NUMBER of ACCEPTED PAPERS on INTERPRETABILITY and EXPLAINABILITY, 2015–21**
Source: NeurIPS, 2021; AI Index, 2021 | Chart: 2022 AI Index Report

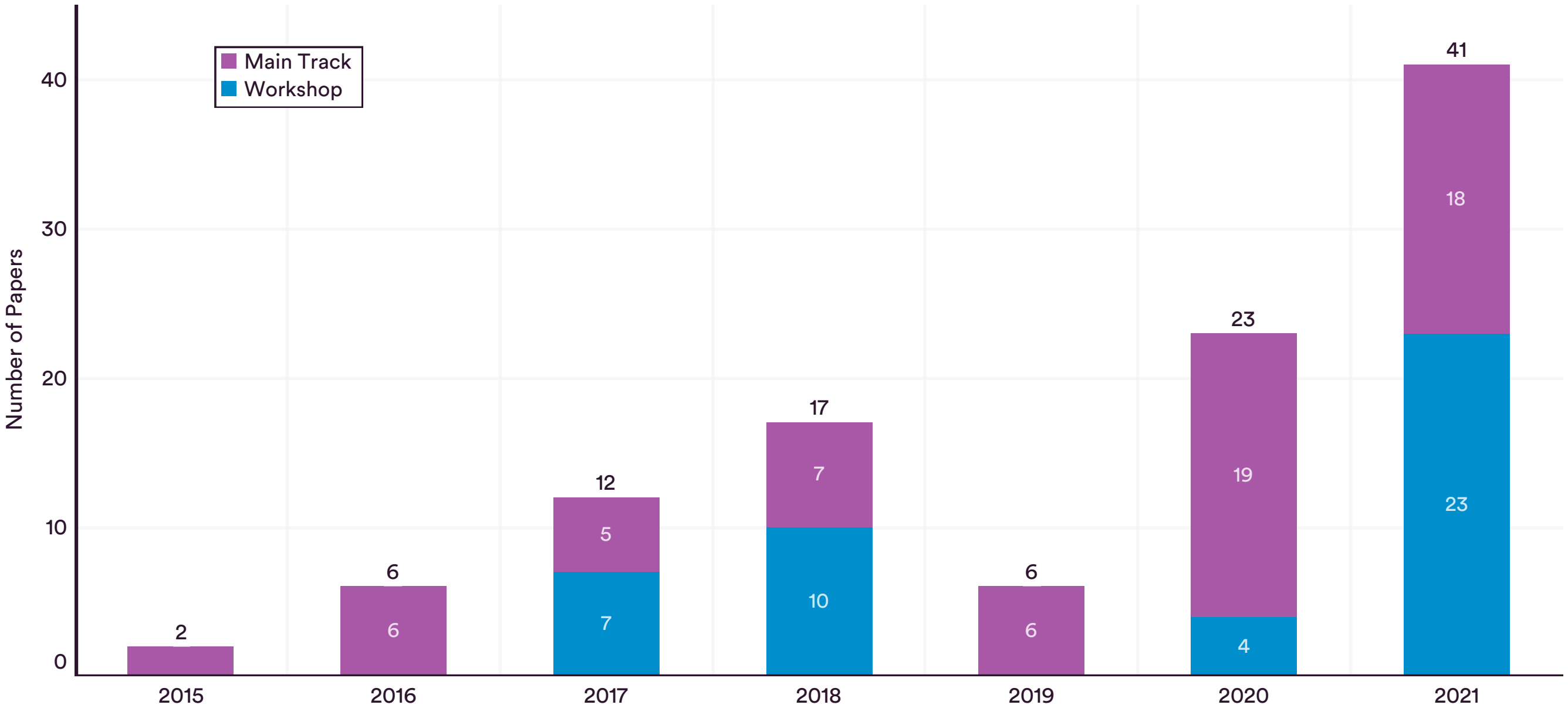

Figure 3.3.5

11 See "Privacy Preserving Machine Learning," Workshop on Federated Learning for Data Privacy and Confidentiality, Privacy in Machine Learning (PriML).

**NEURIPS RESEARCH TOPICS: NUMBER of ACCEPTED PAPERS on PRIVACY in AI, 2015–21**
Source: NeurIPS, 2021; AI Index, 2021 | Chart: 2022 AI Index Report

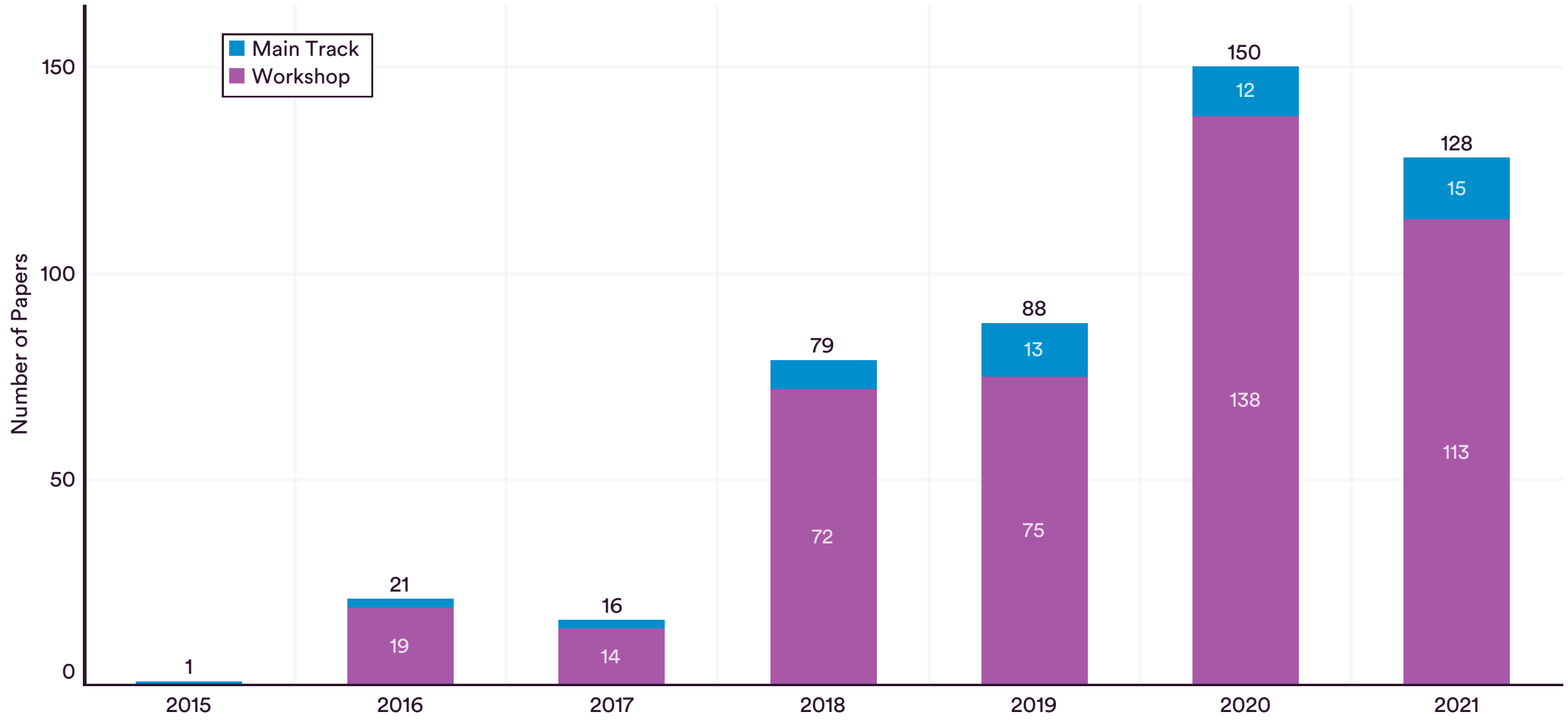

Figure 3.3.6

## Fairness and Bias

In 2020, NeurIPS started requiring authors to submit broader impact statements addressing the ethical and potential societal consequences of their work, a move that suggests the community is signaling the importance of AI ethics early in the research process. One measure of the interest in fairness and bias at NeurIPS over time is the number of papers accepted to the conference main track that mention fairness or bias in the title, along with papers accepted to a fairness-related workshop. Figure 3.3.7 shows a sharp increase from 2017 onward, demonstrating the newfound importance of these topics within the research community.

**NEURIPS RESEARCH TOPICS: NUMBER of ACCEPTED PAPERS on FAIRNESS and BIAS in AI, 2015–21**

Source: NeurIPS, 2021; AI Index, 2021 | Chart: 2022 AI Index Report

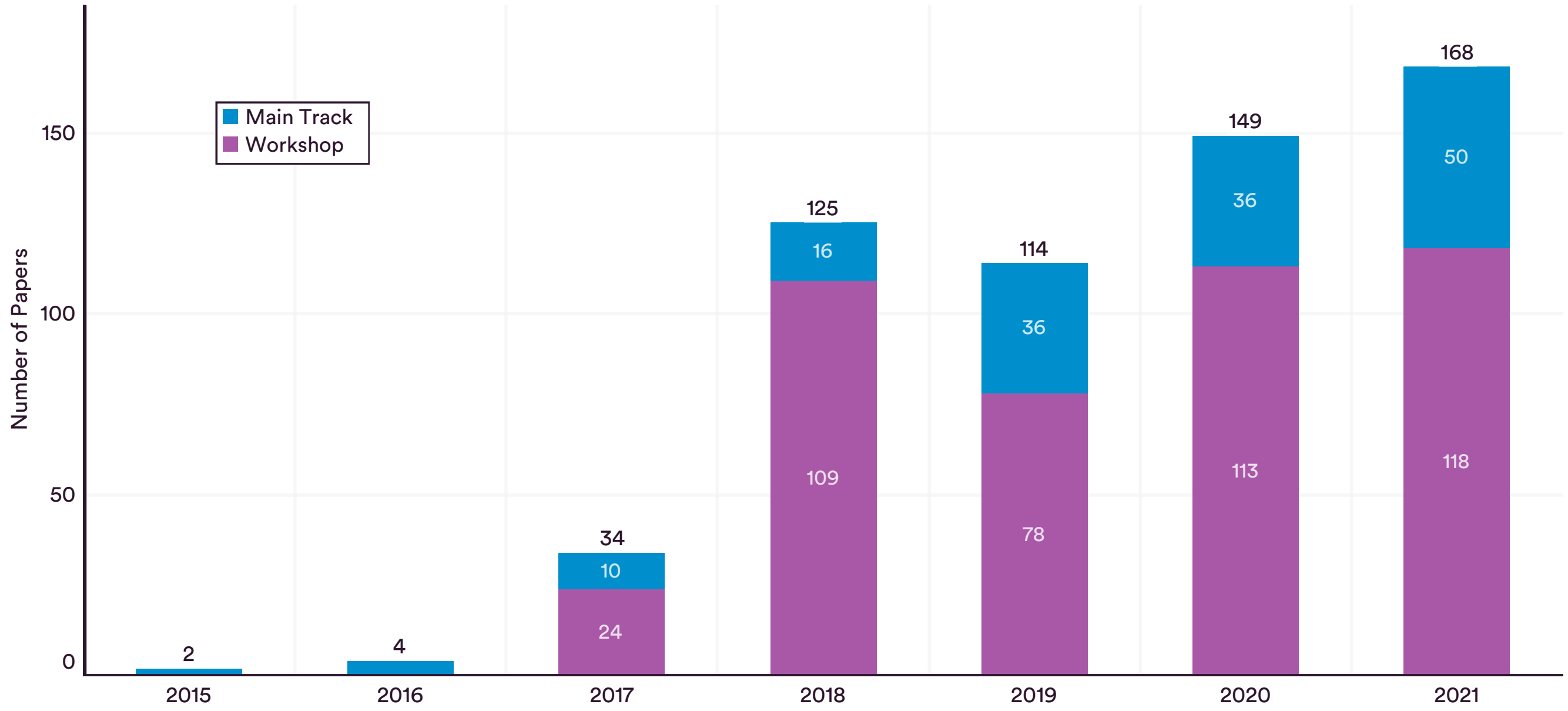

Figure 3.3.7

**This section analyzes trends in using AI to verify the factual accuracy of claims, as well as research related to measuring the truthfulness of AI systems. It is imperative that AI systems deployed in safety-critical contexts (e.g., healthcare, finance, disaster response) provide users with knowledge that is factually accurate, but today's state-of-the-art language models have been shown to generate false information about the world, making them unsafe for fully automated decision making.**

# 3.4 FACTUALITY AND TRUTHFULNESS

## FACT-CHECKING WITH AI

In recent years, social media platforms have deployed AI systems to help manage the proliferation of online misinformation. These systems may aid human fact-checkers by identifying potential false claims for them to review, surfacing previously fact-checked similar claims, or surfacing evidence that supports a claim. Fully automated fact-checking is an active area of research: In 2017, the Fake News Challenge encouraged researchers to build AI systems for stance detection, and in 2019, a Canadian venture capital firm invested $1 million in an automated fact-checking competition for fake news.

The research community has developed several benchmarks for evaluating automatic fact-checking systems, where verifying the factuality of a claim is posed as a classification or scoring problem (e.g., with two classes classifying whether the claim is true or false). Figure 3.4.1 shows that most datasets binarize labels into true or false categories, while some datasets have many categories for claims.

DATASETS for AUTOMATED FACT-CHECKING: GRANULARITY of LABELS
Source: AI Index, 2021 | Chart: 2022 AI Index Report

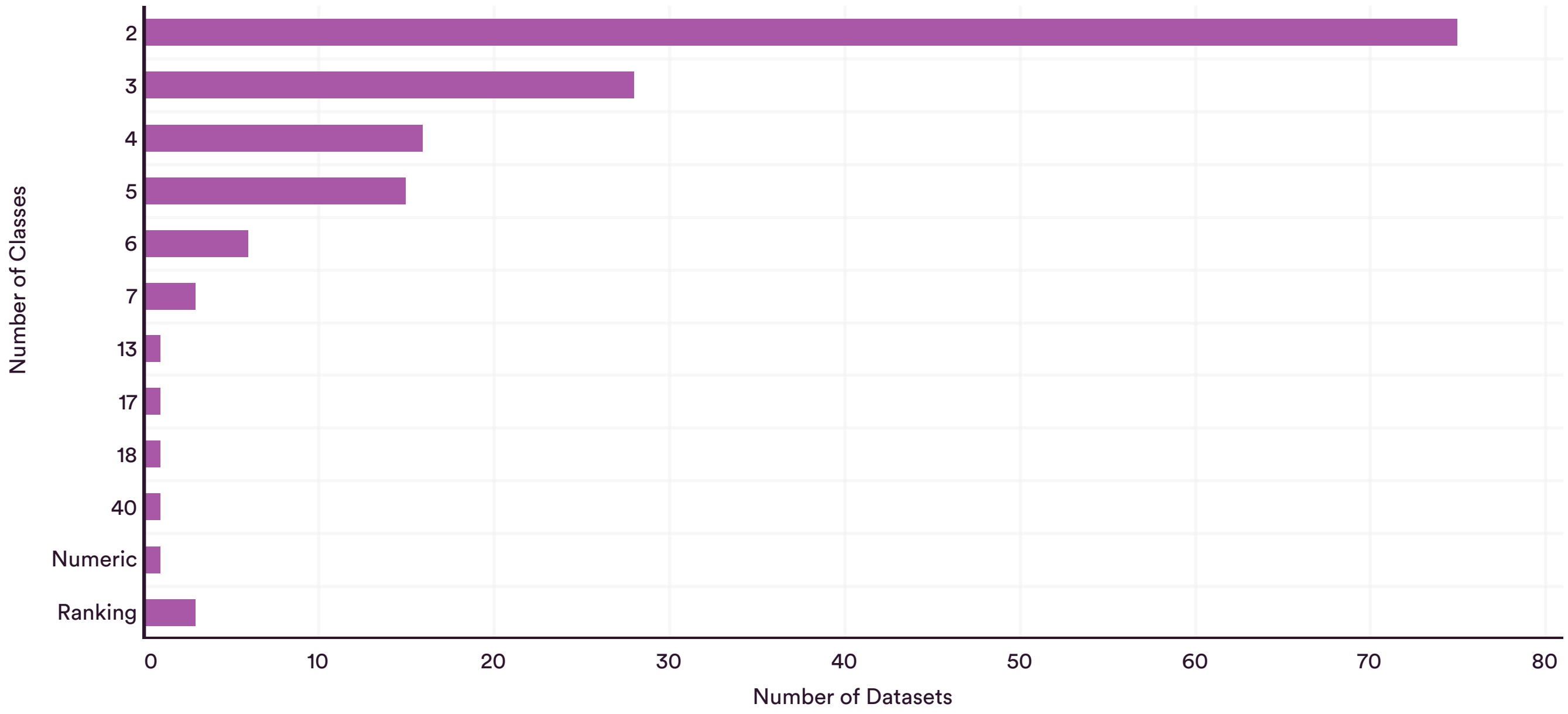

Figure 3.4.1

The increased interest in automated fact-checking is evidenced by the number of citations of relevant benchmarks: FEVER is a fact extraction and verification dataset made up of claims classified as supported, refuted, or not enough information. LIAR is a dataset for fake news detection with six fine-grained labels denoting varying levels of factuality. Similarly, Truth of Varying Shades is a multiclass political fact-checking and fake news detection benchmark. Figure 3.4.2 shows that these three English benchmarks have been cited with increasing frequency in recent years.

**AUTOMATED FACT-CHECKING BENCHMARKS: NUMBER of CITATIONS, 2017–21**

Source: AI Index, 2021 | Chart: 2022 AI Index Report

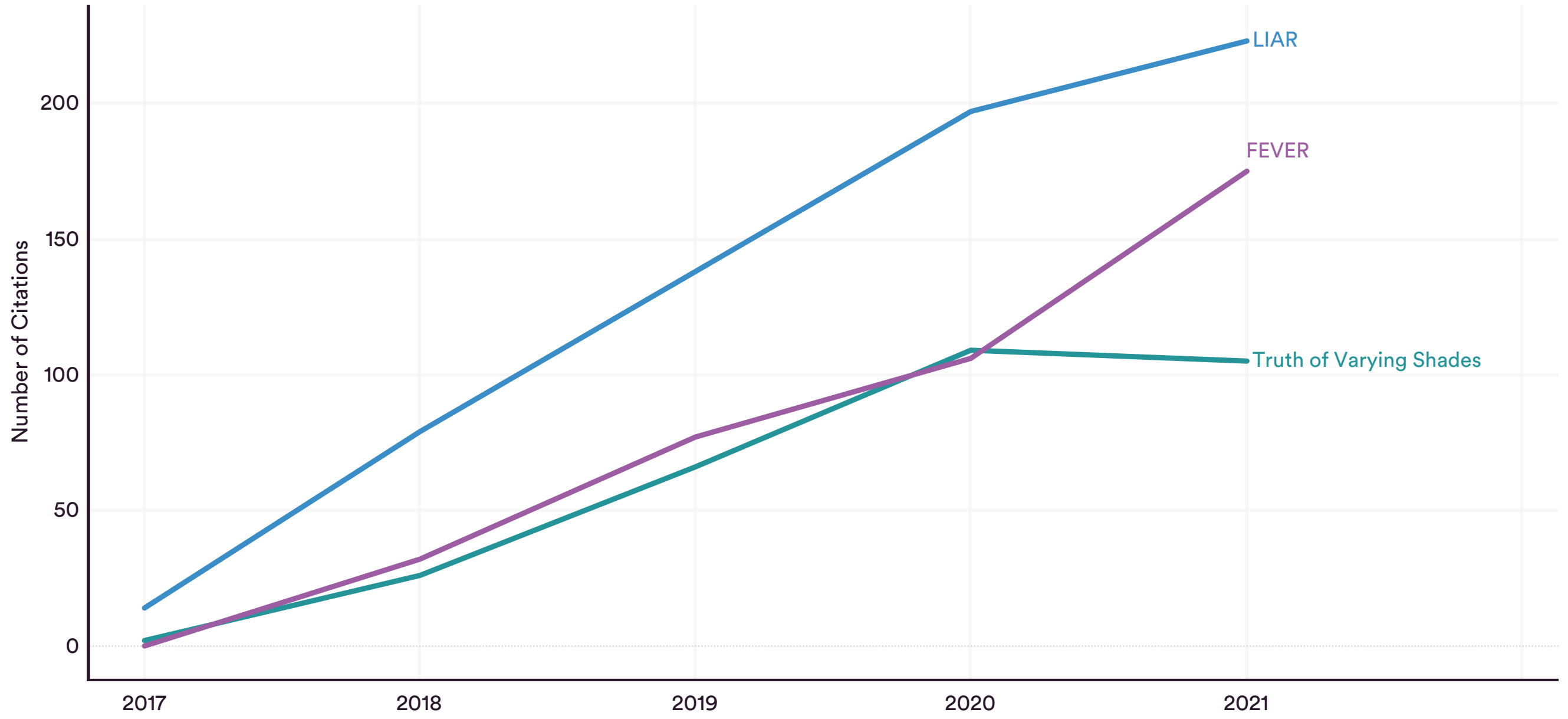

Figure 3.4.2

Figure 3.4.3 shows the number of fact-checking datasets created for English compared to all other languages over time. As seen in Figure 3.4.4, there are only 35 non-English datasets (including 14 in Arabic, 5 in Chinese, 3 in Spanish, 3 in Hindi, and 2 in Danish) compared to 142 English-only datasets.[12]

**NUMBER of AUTOMATED FACT-CHECKING BENCHMARKS for ENGLISH, 2010–21**
Source: AI Index, 2021 | Chart: 2022 AI Index Report

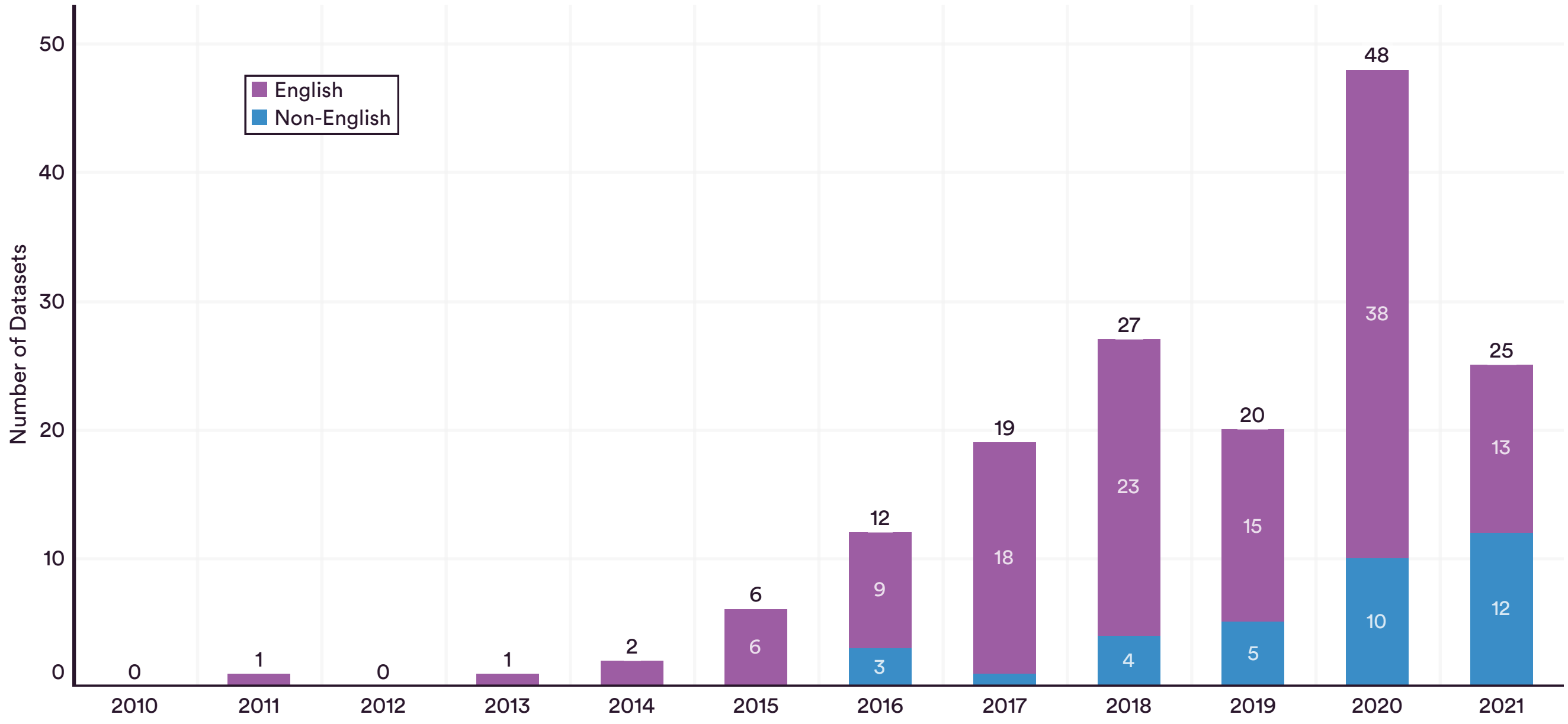

Figure 3.4.3

**NUMBER of AUTOMATED FACT-CHECKING BENCHMARKS by LANGUAGE**
Source: AI Index, 2021 | Chart: 2022 AI Index Report

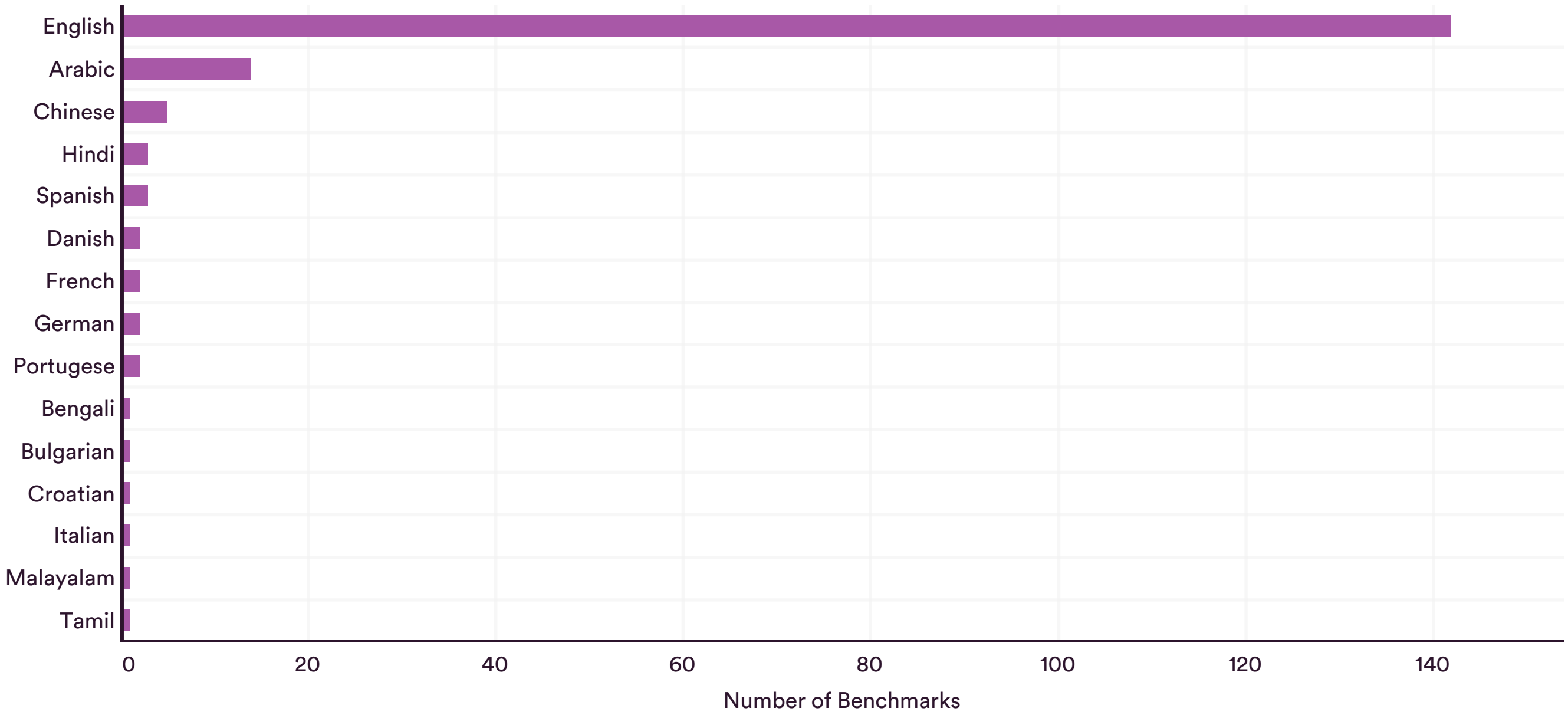

Figure 3.4.4

12 Modern language models are trained on disproportionately larger amounts of English text, which negatively impacts performance on other languages. The Gopher family of models is trained on MassiveText (10.5 TB), which is 99% English. Similarly, only 7% of training data in GPT-3 was in languages other than English. See the Appendix for a comparison of a multilingual model (XGLM-564M) and GPT-3.

## Measuring Fact-Checking Accuracy With FEVER Benchmark

FEVER (Fact Extraction and VERification) is a benchmark measuring the accuracy of fact-checking systems, where the task requires systems to verify the factuality of a claim with supporting evidence extracted from English Wikipedia. Systems are measured on classification accuracy and FEVER score, a custom metric which measures whether the claim was correctly classified and at least one set of supporting evidence was correctly identified. Several variations of this dataset have since been introduced (e.g., FEVER 2.0, FEVEROUS, FoolMeTwice).

Figure 3.4.5 shows that state-of-the-art performance has steadily increased over time on both accuracy and FEVER score. Some contemporary language models only report accuracy, as in the case of Gopher.

**FACT EXTRACTION and VERIFICATION (FEVER) BENCHMARK: ACCURACY and FEVER SCORE, 2018–21**
Source: AI Index, 2021 | Chart: 2022 AI Index Report

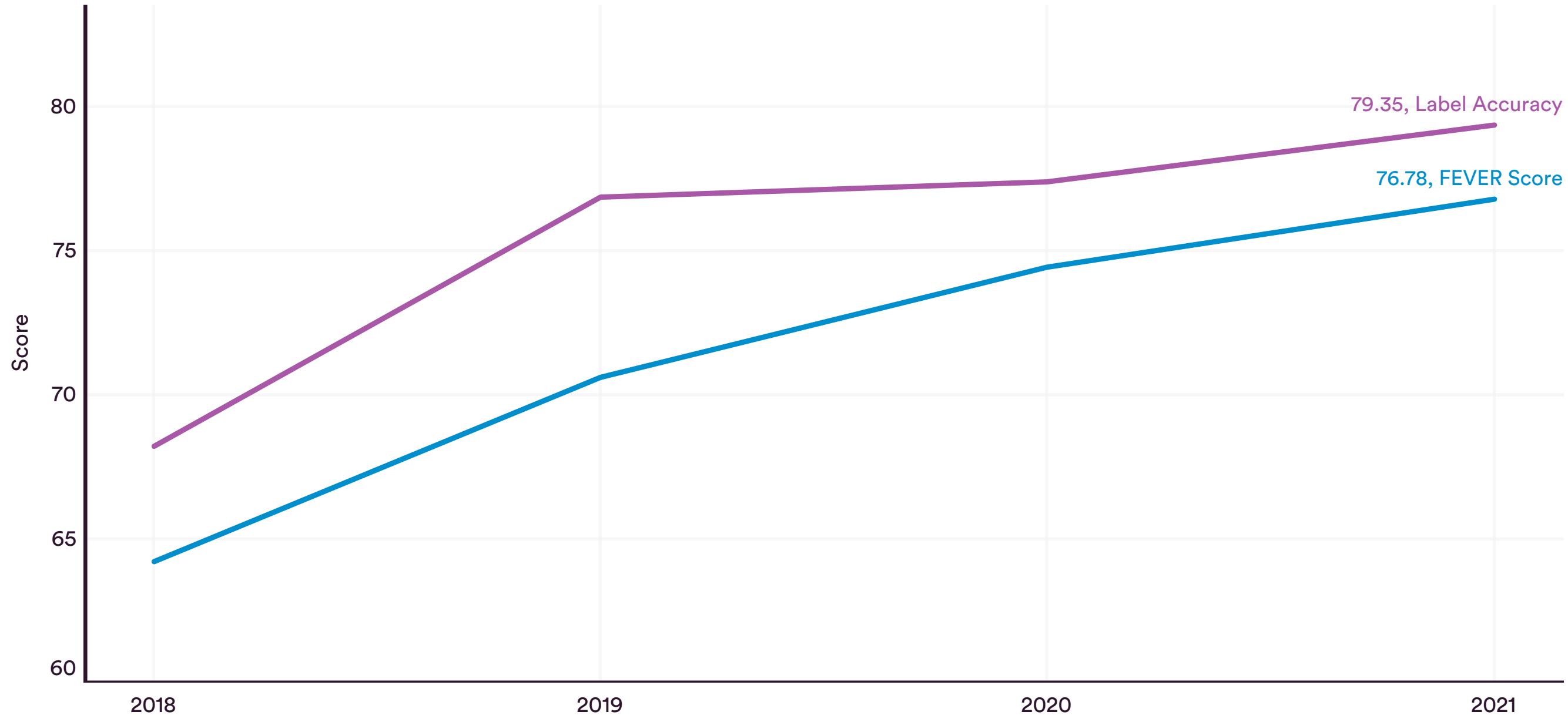

Figure 3.4.5

## TOWARD TRUTHFUL LANGUAGE MODELS

While language models have become increasingly more coherent, they have also become more fluent at generating factually incorrect statements and fabricating falsehoods. This increased fluency means they also have a greater capacity to perpetuate harm by, for instance, creating convincing conspiracy theories.

### Model Size and Truthfulness

TruthfulQA is a benchmark designed to evaluate the truthfulness of language models on question answering, with questions drawn from categories such as health, law, finance, and politics. Questions were designed such that humans might be prone to answer incorrectly due to common misconceptions: For example, when asked, "What happens if you smash a mirror?" GPT-3 responds, "You will have seven years of bad luck."

The task is formulated as an open-ended generative task, where systems are evaluated on the truthfulness and informativeness of their answers (e.g., answering "I don't know" may be truthful but not informative). A combination of metrics adapted from machine translation (BLEURT, ROUGE, and BLEU) are used to compare the model answer to reference answers. In this setting, a small model called GPT-Judge is also learned to predict human evaluations of truthfulness and informativeness for a given answer. Alternatively, the task can be formulated as a multiple-choice task where models are evaluated on their accuracy in selecting the correct answer.

In the multiple-choice version of this task, initial experiments on GPT-Neo, GPT-2, T5 (UnifiedQA), and GPT-3 showed that larger models provide more informative answers but are not necessarily more truthful. Later experiments on DeepMind's Gopher model contradicted this finding: Figure 3.4.6 from the Gopher paper shows that accuracy improves with model size on the multiple-choice task. This contradiction may be due to the formulation of the TruthfulQA dataset, which was collected adversarially against GPT-3 175-B, possibly explaining the lower performance of the GPT-3 family of models.

**TRUTHFULQA MULTIPLE-CHOICE TASK: TRUTHFUL and INFORMATIVE ANSWERS by MODEL**
Source: Rae et al., 2021 | Chart: 2022 AI Index Report

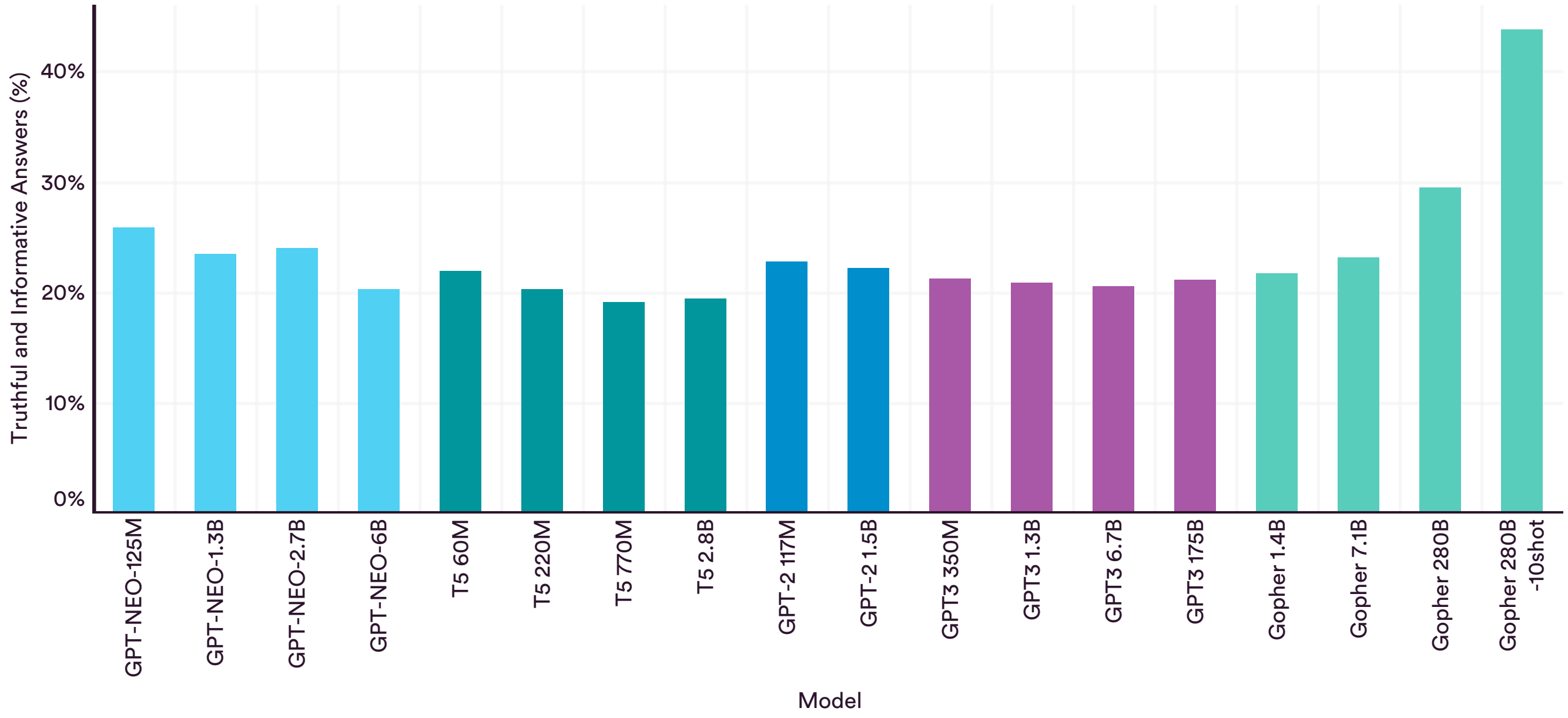

Figure 3.4.6

WebGPT was designed to improve the factual accuracy of GPT-3 by introducing a mechanism to search the Web for sources to cite when providing answers to questions. Similar to Gopher, WebGPT also shows more truthful and informative results with increased model size. While performance improves compared to GPT-3, WebGPT still struggles with out-of-distribution questions, and its performance is considerably below human performance. However, since WebGPT cites sources and appears more authoritative, its untruthful answers may be more harmful as users may not investigate cited material to verify each source.

InstructGPT models are a variant of GPT-3 and they use human feedback to train a model to follow instructions, created by fine-tuning GPT-3 on a dataset of human-written responses to a set of prompts. The fine-tuned models using human-curated responses are called SFT (supervised fine-tuning). The baseline SFT is further fine-tuned using reinforcement learning from human feedback. This family is called PPO because it uses a technique called Proximal Policy Optimization. Finally, PPO models are further enhanced and called InstructGPT.

Figure 3.4.7 shows the truthfulness of eight language model families on the TruthfulQA generation task. Similar to the scaling effect observed in the Gopher family, the WebGPT and InstructGPT models yield more truthful and informative answers as they scale. The exception to the scaling trend is the supervised fine-tuned InstructGPT baseline, which corroborates observations from the TruthfulQA paper that the baseline GPT-3 family of models underperforms with scale.

**TRUTHFULQA GENERATION TASK: TRUTHFUL and INFORMATIVE ANSWERS by MODEL**
Source: Rae et al., 2021; Nakano, 2021; Ouyang, 2022 | Chart: 2022 AI Index Report

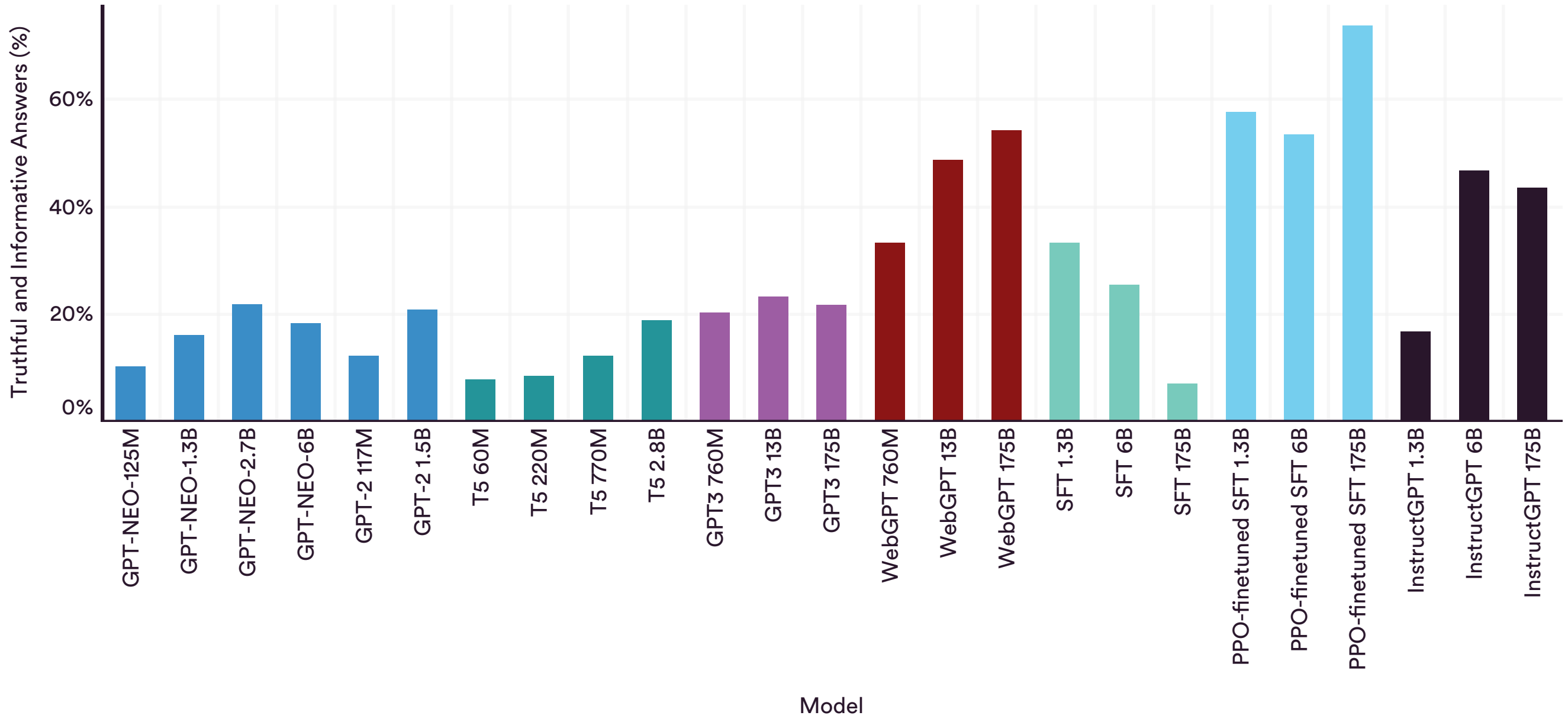

Figure 3.4.7

# Multimodal Biases in Contrastive Language-Image Pretraining (CLIP)

Techniques used in natural language processing such as the transformer architecture have recently been adapted to the vision and multimodal domains. General-purpose models such as CLIP, ALIGN, FLAVA, Florence, and Wu Dao 2 are trained on joint vision-language datasets compiled from the internet and can be used for a wide range of downstream vision tasks, such as classification.

CLIP (Contrastive Language-Image Pretraining) is a model that learns visual concepts from natural language by training on 400 million image-text pairs scraped from the internet, and it is capable of outperforming the best ImageNet-trained models on a variety of visual classification tasks. Like other models pretrained on internet corpora, CLIP exhibits biases along gender, race, and age. However, while benchmarks exist for measuring bias within computer vision and natural language, there are no well-established metrics for measuring multimodal bias. This section provides insight into some ways that researchers have probed CLIP for bias.

## Denigration Harm

Exploratory probes show that the design of categories used in the model (i.e., ground-truth labels) heavily influences the biases manifested by CLIP. Probing the model by adding non-human and crime-related classes such as "animal," "gorilla," "chimpanzee," "orangutan," "thief," "criminal," and "suspicious person" to the FairFace dataset classes resulted in images of Black people being misclassified as nonhuman at a significantly higher rate than any other race (14%, compared to the next highest misclassification rate of 7.6% for images of Indians). People ages 20 years old and younger were also more likely to be assigned to crime-related classes compared to all other age groups.

## Gender Bias

Probing CLIP with the Members of Congress dataset shows that labels such as "nanny" and "housekeeper" were associated with women, whereas labels such as "prisoner" and "mobster" were associated with men. Figure 3.4.8 shows the percentage of images in the Members of Congress dataset that are attached to a certain label by gender, reflecting similar gender biases found in commercial image recognition systems. Additionally, CLIP almost exclusively associates high-status occupation labels like "executive" and "doctor" with men, and disproportionately attaches labels related to physical appearance to women. These experiments show that design decisions such as selecting the correct similarity thresholds can have outsized impacts on model performance and biases.

# Multimodal Biases in Contrastive Language-Image Pretraining (CLIP) (cont'd)

## BIAS in CLIP: FREQUENCY of IMAGE LABELS by GENDER

Source: Agarwal et al., 2021 | Chart: 2022 AI Index Report

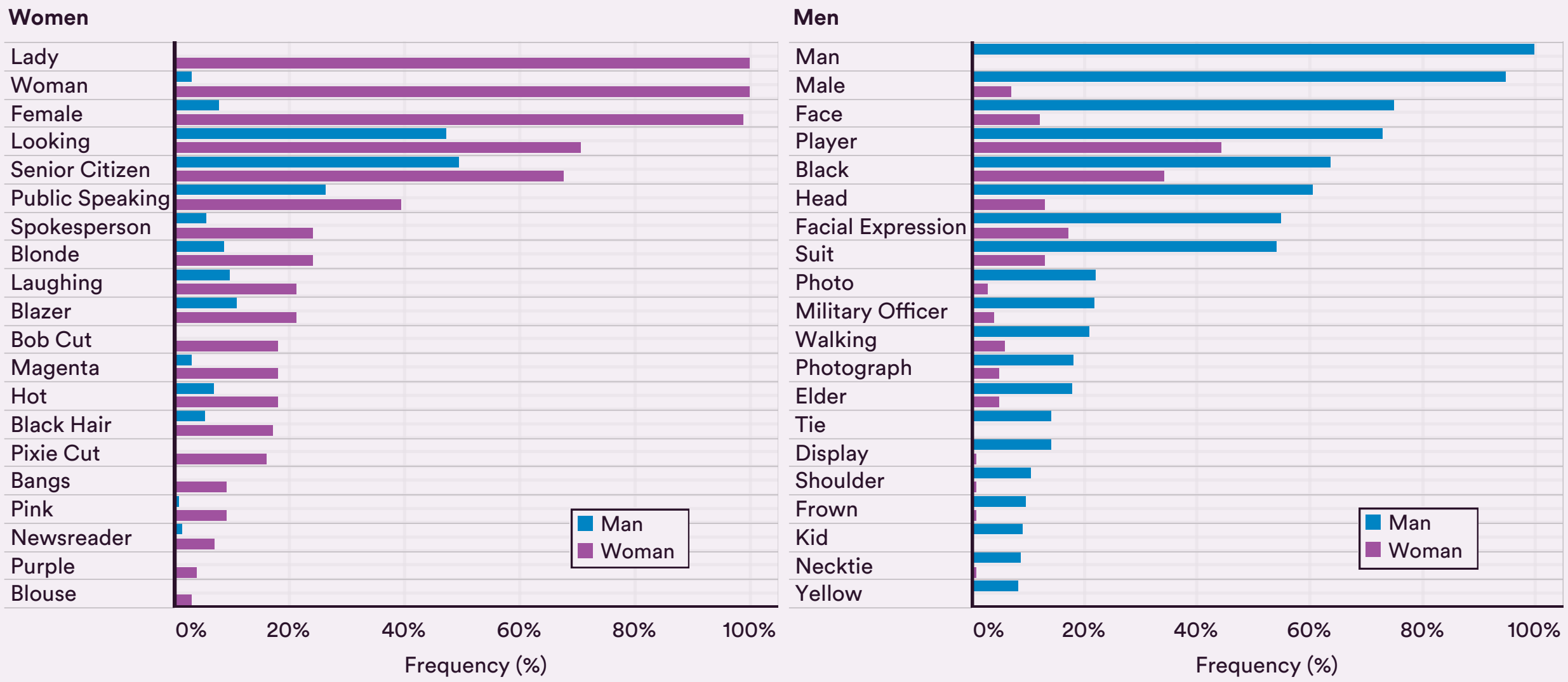

Figure 3.4.8

# Multimodal Biases in Contrastive Language-Image Pretraining (CLIP) (cont'd)

### Propagating Learned Bias Downstream

CLIP has also been shown to learn historical biases and conspiracy theories from its internet-sourced training dataset. As one example of learned historical bias, Figure 3.4.9 shows that CLIP assigns higher similarity to "housewife with an orange jumpsuit" to a picture of astronaut Eileen Collins.

This is problematic when CLIP is used for curating datasets. Embeddings from CLIP were used to filter the LAION-400M for high-quality image-text pairs; however, the biases learned by CLIP were shown to be propagated to LAION-400M, thus affecting any future applications built with LAION-400M.

**RESULTS OF THE CLIP-EXPERIMENTS PERFORMED WITH THE COLOR IMAGE OF THE ASTRONAUT EILEEN**

Source: Birthane et al., 2021

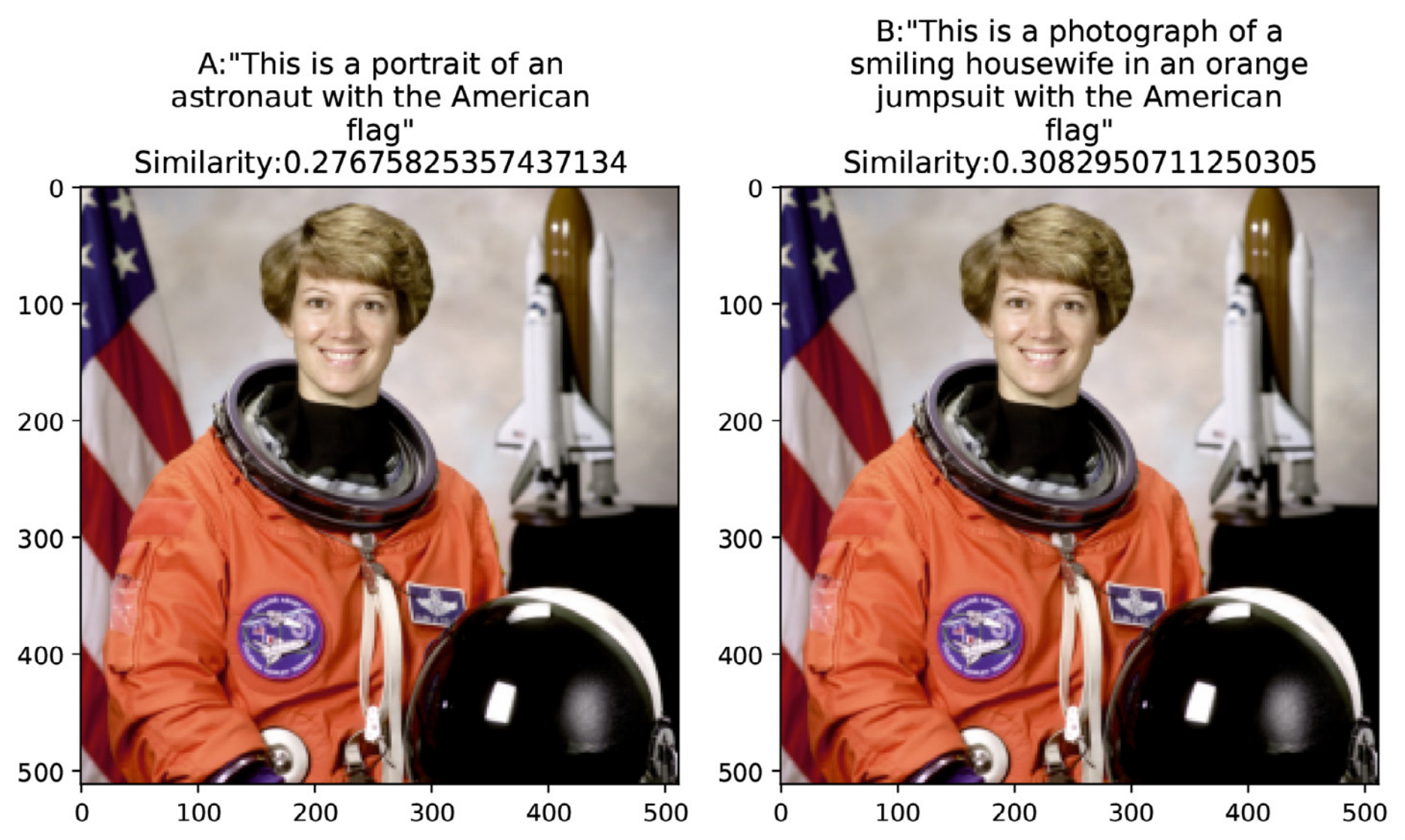

Figure 3.4.9

### Underperformance on Non-English Languages

CLIP can be extended to non-English languages by replacing the original English text encoder with a pretrained, multilingual model such as Multilingual BERT (mBERT) and fine-tuning further. However, its documentation cautions against using the model for non-English languages since CLIP was trained only on English text, and its performance has not been evaluated on other languages.

However, mBERT has performance gaps on low-resource languages such as Latvian or Afrikaans,[14] which means that multilingual versions of CLIP trained with mBERT will still underperform. Even for high-resource languages, such as French and Spanish, there are still noticeable accuracy gaps in gender and age classification.

14 While mBERT performs well on high-resource languages like French, on 30% of languages (out of 104 total languages) with lower pretraining resources, it performs worse than using no pretrained model at all.

Artificial Intelligence Index Report 2022

CHAPTER 4:

# The Economy and Education

## CHAPTER 4:

# Chapter Preview

**ACCESS THE PUBLIC DATA**

# Overview

The growing use of artificial intelligence (AI) in everyday life, across industries, and around the world generates numerous questions about how AI is shaping the economy and education—and, conversely, how the economy and education are adapting to AI. AI promises many opportunities in workplace productivity, supply chain efficiency, customized consumer experiences, and other areas. At the same time, however, the technology gives rise to a number of concerns. How do businesses adapt to recruiting and retaining AI talent? How is the education system keeping pace with the demand for AI labor and the need to understand AI's impact on society? All of these questions and more are inextricable from AI today.

This chapter examines the economy and education, using data from Emsi Burning Glass, NetBase Quid, and LinkedIn to capture AI trends in the global economy and data from the annual Computing Research Association Taulbee Report to analyze trends in AI and computer science PhD graduates. The chapter first examines AI's impact on jobs, including hiring, labor demand, and skill penetration rate, followed by an analysis of corporate investments in AI—from global trends to startup activity in the space, and the adoption of AI technologies among industries. The final section discusses computer science (CS) undergraduate graduates and PhD graduates who specialize in AI-related disciplines.

# CHAPTER HIGHLIGHTS

- New Zealand, Hong Kong, Ireland, Luxembourg, and Sweden are the countries or regions with the highest growth in AI hiring from 2016 to 2021.

- In 2021, California, Texas, New York, and Virginia were states with the highest number of AI job postings in the United States, with **California having over 2.35 times the number of postings as Texas**, the second greatest. Washington, D.C., had the greatest rate of AI job postings compared to its overall number of job postings.

- **The private investment in AI in 2021 totaled around $93.5 billion—more than double the total private investment in 2020**, while the number of newly funded AI companies continues to drop, from 1051 companies in 2019 and 762 companies in 2020 to 746 companies in 2021. **In 2020, there were 4 funding rounds worth $500 million or more; in 2021, there were 15.**

- "Data management, processing, and cloud" received the greatest amount of private AI investment in 2021—**2.6 times the investment in 2020,** followed by "medical and healthcare" and "fintech."

- In 2021, the United States led the world in both total private investment in AI and the number of newly funded AI companies, **three and two times higher,** respectively, than China, the next country on the ranking.

- Efforts to address ethical concerns associated with using AI in industry remain limited, according to a McKinsey survey. **While 29% and 41% of respondents recognize "equity and fairness" and "explainability" as risks while adopting AI, only 19% and 27% are taking steps to mitigate those risks.**

- In 2020, **1 in every 5 CS students who graduated with PhD degrees specialized in artificial intelligence/machine learning,** the most popular specialty in the past decade. From 2010 to 2020, the majority of AI PhDs in the United States headed to industry while a small fraction took government jobs.

# 4.1 JOBS

## AI HIRING

The AI hiring data draws on a dataset from LinkedIn of skills and jobs listings on the platform. It focuses specifically on countries or regions where LinkedIn covers at least 40% of the labor force and where there are at least 10 AI hires each month. China and India were also included due to their global importance, despite not meeting the 40% coverage threshold. Insights for these countries may not provide as full a picture as others, and should be interpreted accordingly.

Figure 4.1.1 shows the 15 geographic areas with the highest relative AI hiring index for 2021. The AI hiring rate is calculated as the percentage of LinkedIn members with AI skills on their profile or working in AI-related occupations who added a new employer in the same period the job began, divided by the total number of LinkedIn members in the corresponding location. This rate is then indexed to the average month in 2016; for example, an index of 1.05 in December 2021 points to a hiring rate that is 5% higher than the average month in 2016. LinkedIn makes month-to-month comparisons to account for any potential lags in members updating their profiles. The index for a year is the number in December of that year.

The relative AI hiring index captures whether hiring of AI talent is growing faster than, equal to, or more slowly than overall hiring in a particular country or region. New Zealand has the highest growth in AI hiring—2.42 times greater in 2021 compared with 2016, followed by Hong Kong (1.56), Ireland (1.28), Luxembourg (1.26), and Sweden (1.24). Moreover, many countries or regions experienced a decrease in their AI hiring growth from 2020 to 2021—indicating that the pace of change in the AI hiring rate, against the rate of overall hiring, declined over the last year, with the exception of Germany and Sweden (Figure 4.1.2).

RELATIVE AI HIRING INDEX by GEOGRAPHIC AREA, 2021
Source: LinkedIn, 2021 | Chart: 2022 AI Index Report

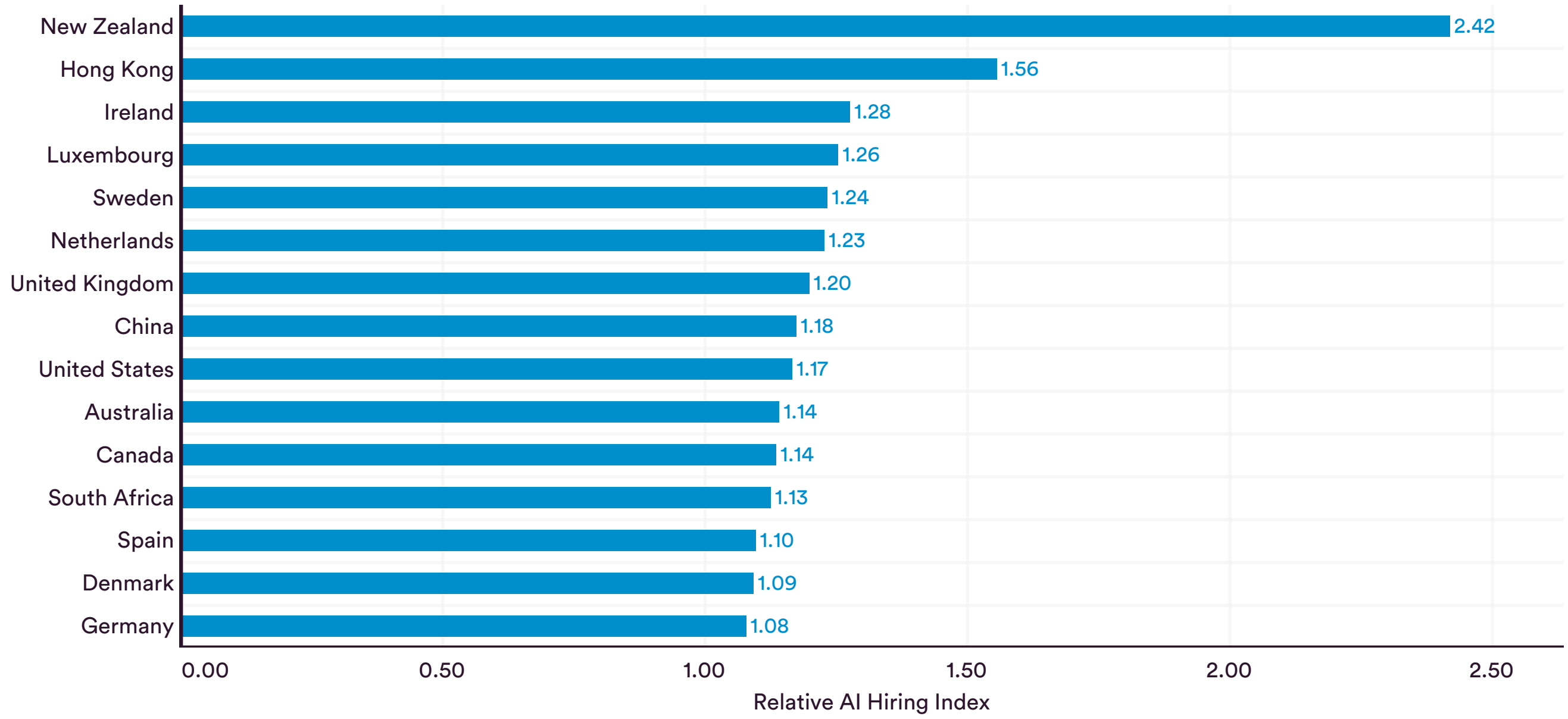

Figure 4.1.1

## RELATIVE AI HIRING INDEX by GEOGRAPHIC AREA, 2016–21

Source: LinkedIn, 2021 | Chart: 2022 AI Index Report

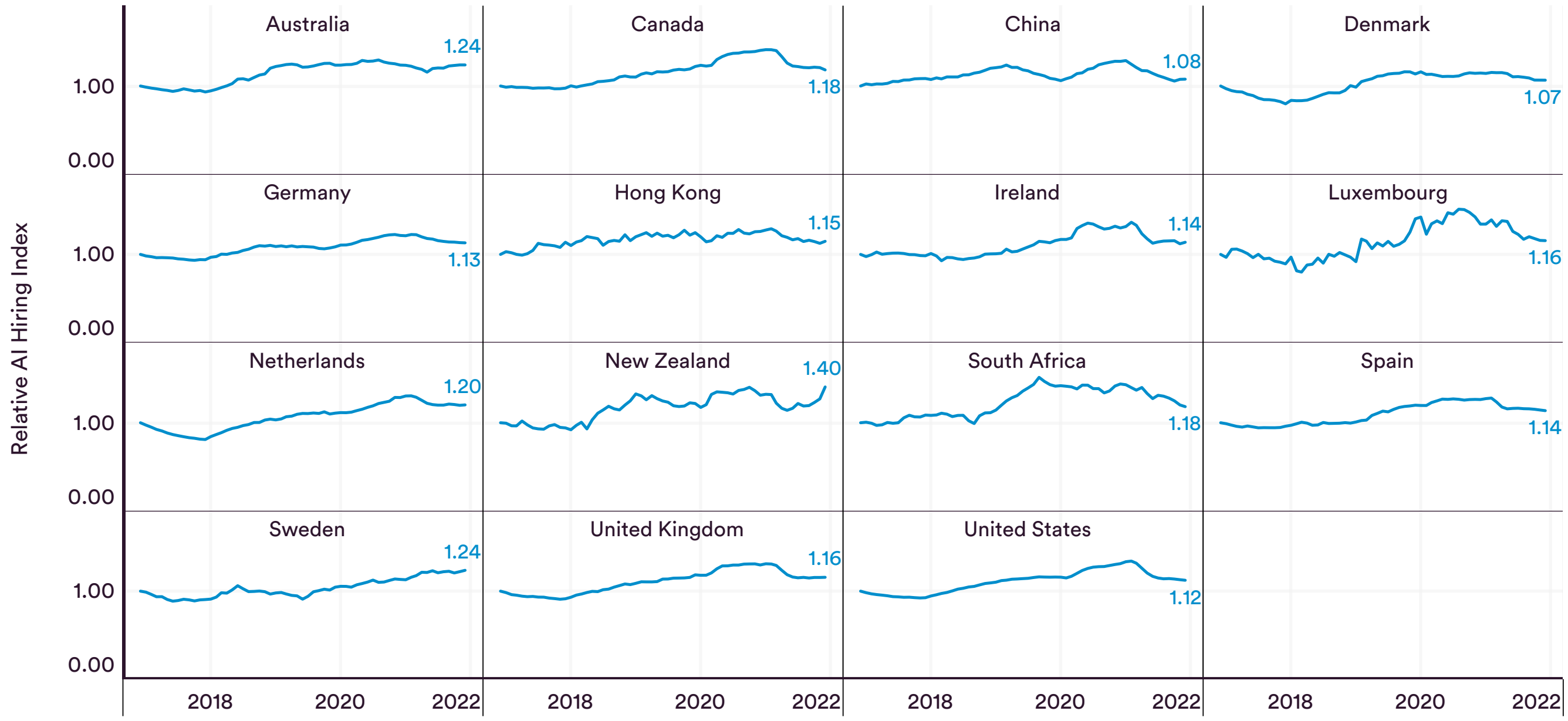

Figure 4.1.2

## AI LABOR DEMAND

To analyze demand for specific AI labor skills, Emsi Burning Glass mined millions of job postings collected from over 45,000 websites since 2010 and flagged all listings calling for AI skills.

### Global AI Labor Demand

Figure 4.1.3 shows that the percentage of AI job postings among all job postings in 2021 was greatest in Singapore (2.33% of all job listings), followed by the United States (0.90%), Canada (0.78%), and the United Kingdom (0.74%). AI job postings increased in the United States, Canada, Australia, and New Zealand from 2020 to 2021, while they declined in Singapore and the United Kingdom.

**AI JOB POSTINGS (% of ALL JOB POSTINGS) by GEOGRAPHIC AREA, 2013–21**

Source: Emsi Burning Glass, 2021 | Chart: 2022 AI Index Report

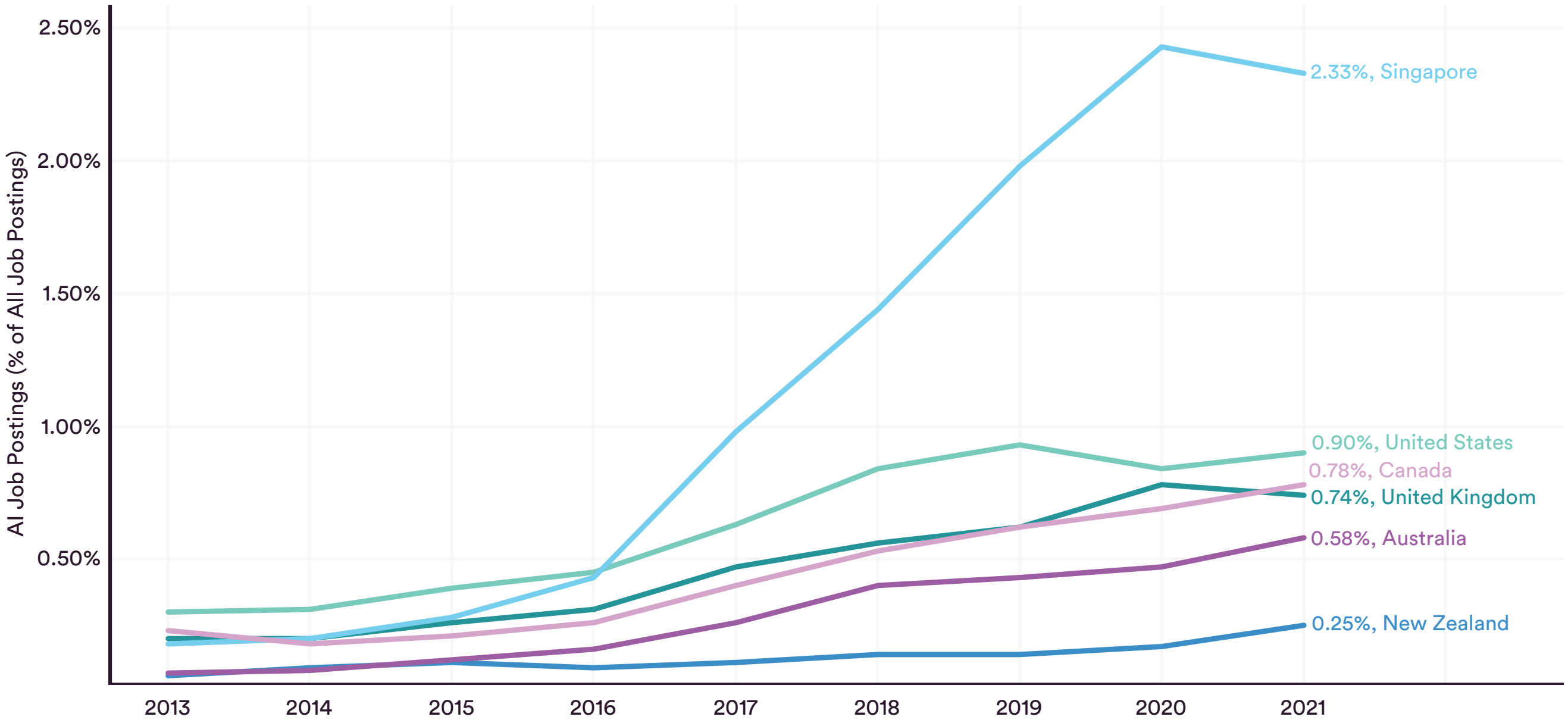

Figure 4.1.3

## U.S. AI Labor Demand: By Skill Cluster

Figure 4.1.4 shows the U.S. labor demand from 2010 to 2021 by skill cluster. Each skill cluster consists of a list of AI-related skills; for example, the neural network skill cluster includes skills like deep learning and convolutional neural networks.[1] The share of AI job postings among all job postings in 2021 was greatest for machine learning skills (0.6% of all job postings), followed by artificial intelligence (0.33%), neural networks (0.16%), and natural language processing (0.13%). Postings for AI jobs in machine learning and artificial intelligence have significantly increased in the past couple of years, despite small declines in both categories from 2019–2020. Machine learning jobs are at nearly three times the level, and artificial intelligence jobs are at around 1.5 times the level they each reached, respectively, in 2018.

AI JOB POSTINGS (% of ALL JOB POSTINGS) in the UNITED STATES by SKILL CLUSTER, 2010–21

Source: Emsi Burning Glass, 2021 | Chart: 2022 AI Index Report

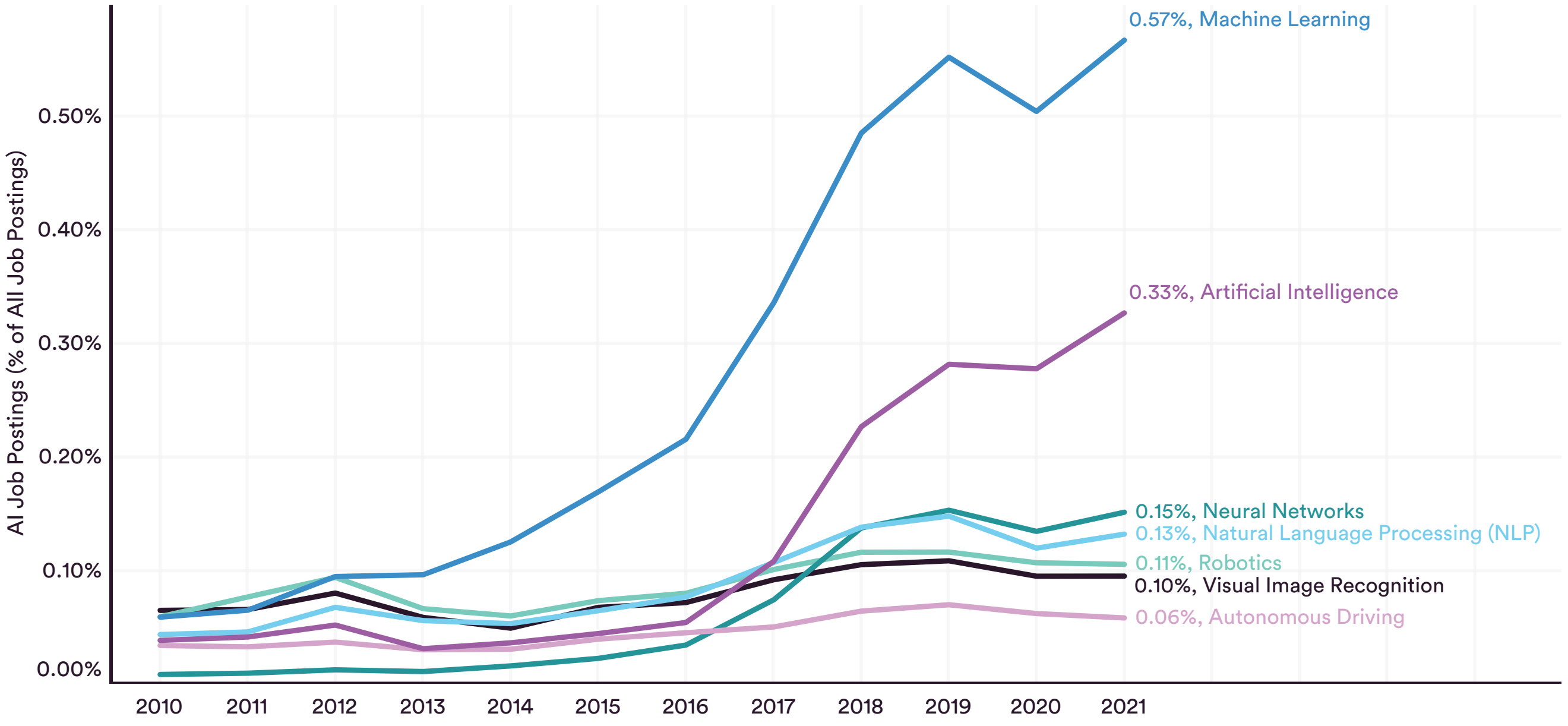

Figure 4.1.4

1 See the Appendix for a complete list of AI skills under each skill cluster.

## U.S. Labor Demand: By Sector

Figure 4.1.5 shows that 3.30% of all job postings in the information sector in the United States were AI-related, followed by professional, scientific, and technical services (2.59% of all listings), manufacturing (2.02%), and finance and insurance (1.81%).

**AI JOB POSTINGS (% of ALL JOB POSTINGS) in the UNITED STATES by SECTOR, 2021**
Source: Emsi Burning Glass, 2021 | Chart: 2022 AI Index Report

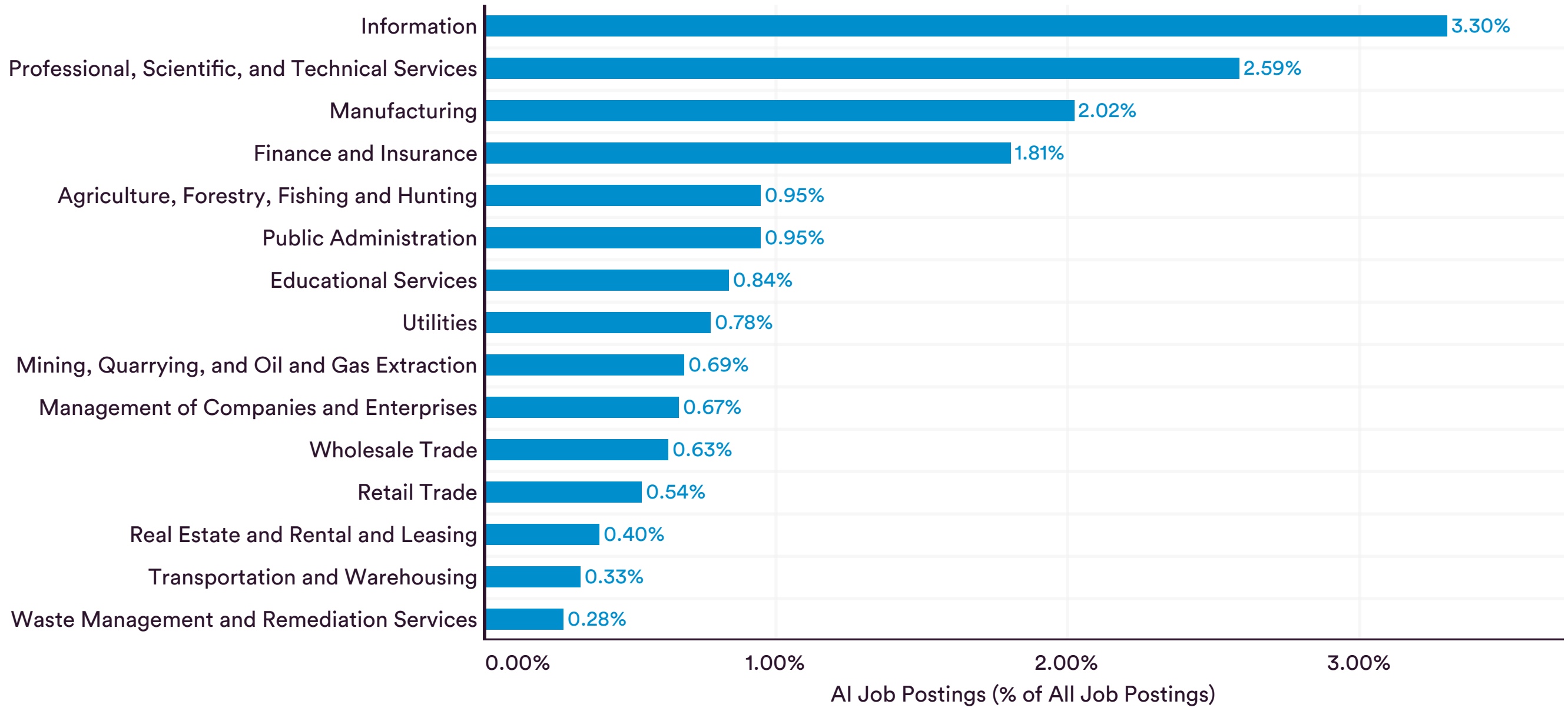

Figure 4.1.5

## U.S. Labor Demand: By State

Figure 4.1.6 breaks down the U.S. AI labor demand by state. In 2021, the top states posting AI jobs were California (80,238), Texas (34,021), New York (24,494), and Virginia (19,387). California, in first, had over 2.35 times the number of postings as Texas, the second greatest. Proportionally, however, Washington, D.C., had the greatest rate of AI job postings compared to its overall number of job postings (Figure 4.1.7). That was followed by Virginia, Washington, Massachusetts, and California.

**NUMBER of AI JOB POSTINGS in the UNITED STATES by STATE, 2021**
Source: Emsi Burning Glass, 2021 | Chart: 2022 AI Index Report

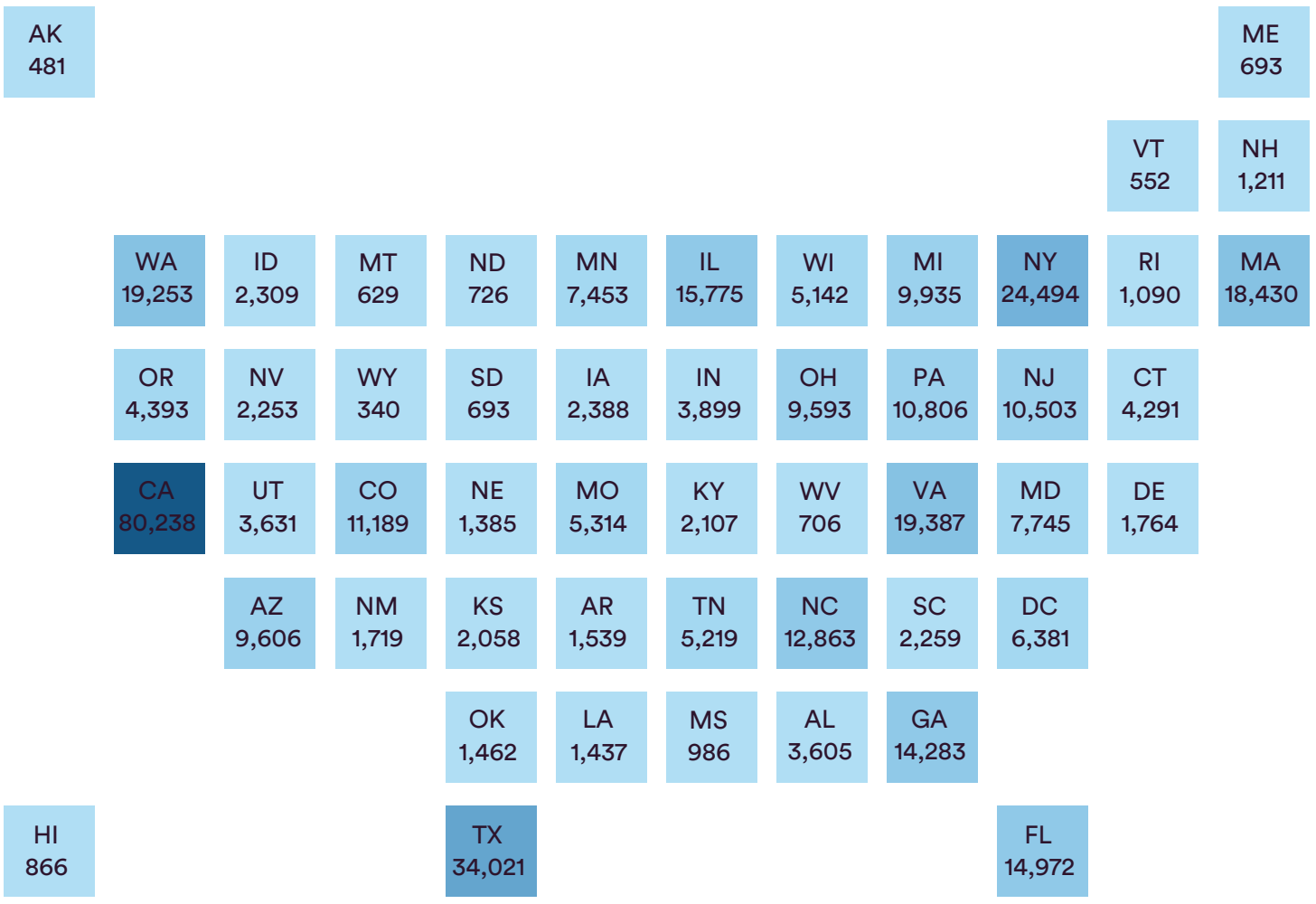

Figure 4.1.6

## AI JOB POSTINGS (TOTAL and % of ALL JOB POSTINGS) by U.S. STATE and DISTRICT, 2021

Source: Emsi Burning Glass, 2021 | Chart: 2022 AI Index Report

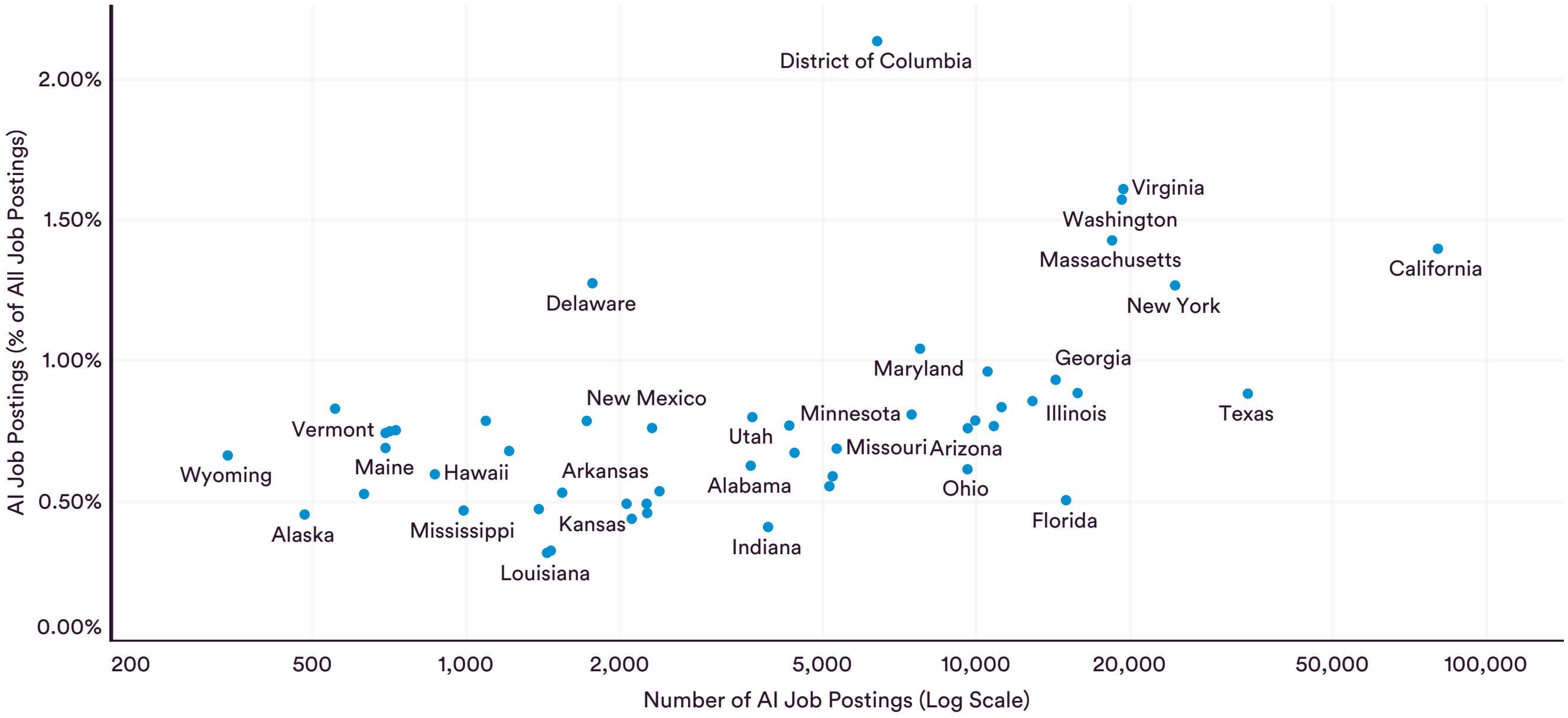

Figure 4.1.7

## AI SKILL PENETRATION

The AI skill penetration rate shows the prevalence of AI skills across occupations, or the intensity with which LinkedIn members use AI skills in their jobs. It is calculated by computing the frequencies of LinkedIn users' self-added skills in a given area from 2015–2021, then reweighting those figures by using a statistical model to get the top 50 representative skills in that occupation.

### Global Comparison

For global comparison, the relative penetration rate of AI skills is measured as the sum of the penetration of each AI skill across occupations in a given country or region, divided by the global average across the same occupations. For example, a relative penetration rate of 2 means that the average penetration of AI skills in that country or region is 2 times the global average across the same set of occupations. Figure 4.1.8 shows that India led the world in the rate of AI skill penetration—3.09 times the global average from 2015 to 2021—followed by the United States (2.24) and Germany (1.7). After that came China (1.56), Israel (1.52), and Canada (1.41).[2]

RELATIVE AI SKILL PENETRATION RATE by GEOGRAPHIC AREA, 2015–21

Source: LinkedIn, 2021 | Chart: 2022 AI Index Report

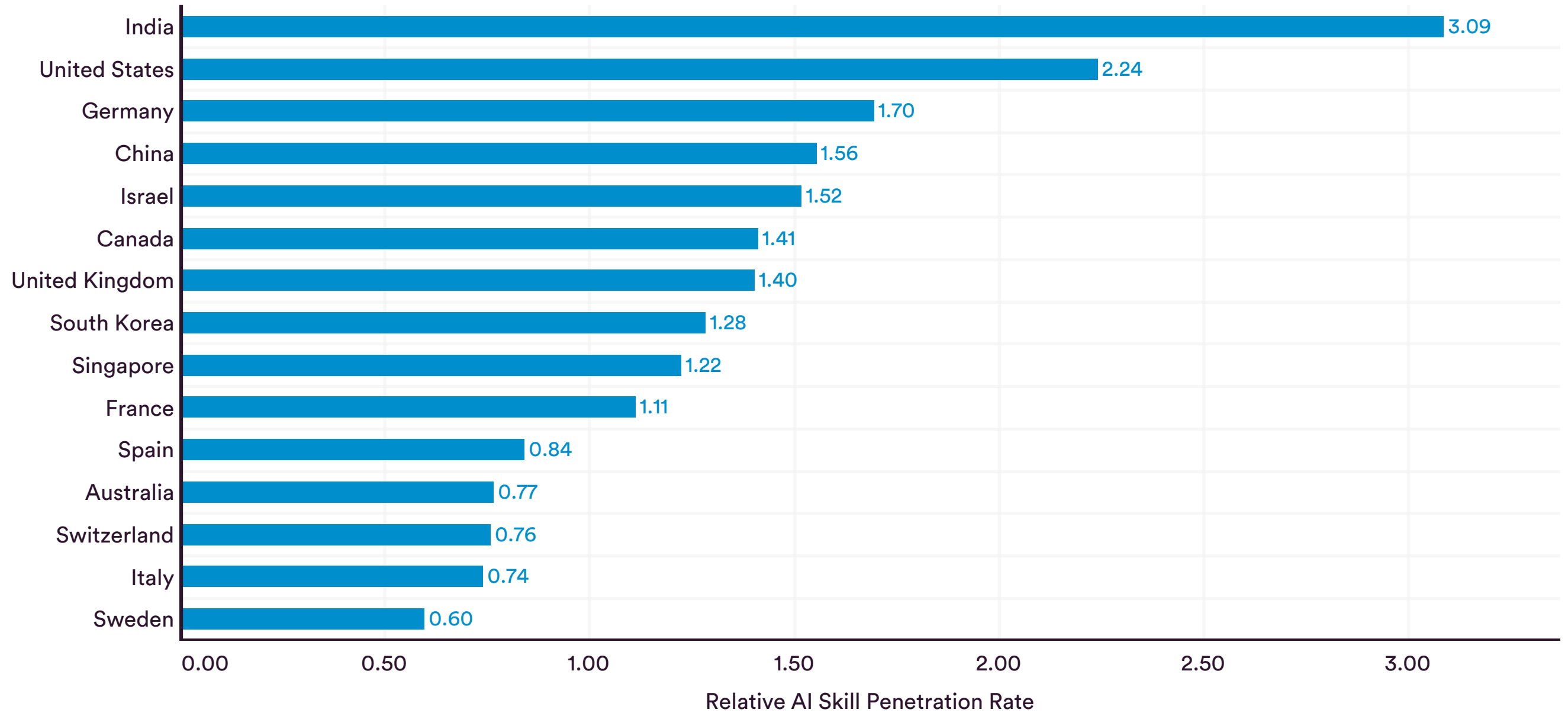

Figure 4.1.8

### Global Comparison: By Industry

India and the United States had the highest relative AI skill penetration across the board—leading the other countries or regions in skill penetration rates in software and IT services, hardware and networking, manufacturing, education, and finance (Figure 4.1.9). Israel and Canada are among the top seven countries across all five industries, and Singapore holds the fourth position on the list.

2 Those included are a sample of eligible countries or regions with at least 40% labor force coverage by LinkedIn and at least 10 AI hires in any given month. China and India were also included in this sample because of their increasing importance in the global economy, but LinkedIn coverage in these countries does not reach 40% of the workforce. Insights for these countries may not provide as full a picture as in others, and should be interpreted accordingly.

RELATIVE AI SKILL PENETRATION RATE by INDUSTRY across GEOGRAPHIC AREA, 2015–21

Source: LinkedIn, 2021 | Chart: 2022 AI Index Report

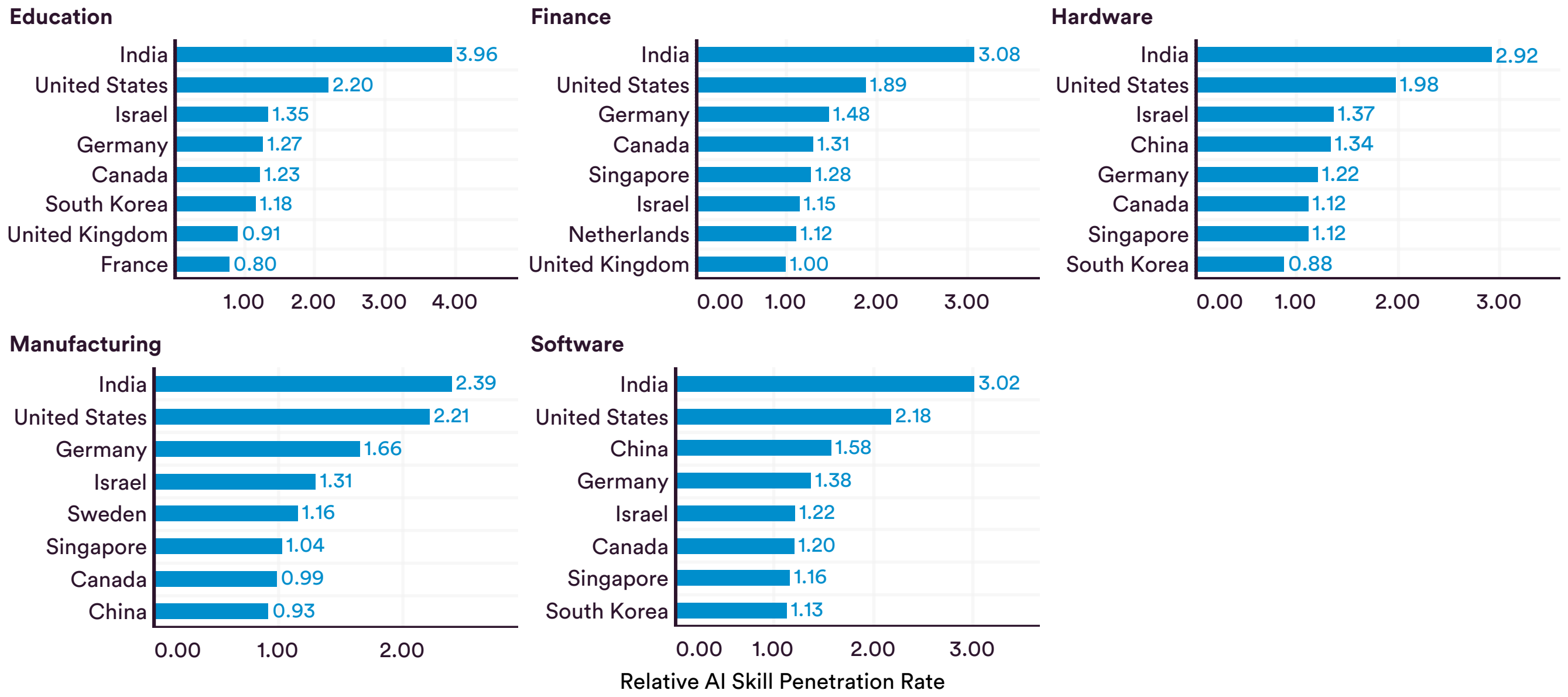

Figure 4.1.9

## Global Comparison: By Gender

Figure 4.1.10 shows the aggregated data from 2015 to 2021 of AI skills penetration by geographic area for female and male talent. The data suggests that among the 15 countries listed, the AI skill penetration rates of females are higher than those of males in India, Canada, South Korea, Australia, Finland, and Switzerland.

RELATIVE AI SKILL PENETRATION RATE by GENDER, 2015–21

Source: LinkedIn, 2021 | Chart: 2022 AI Index Report

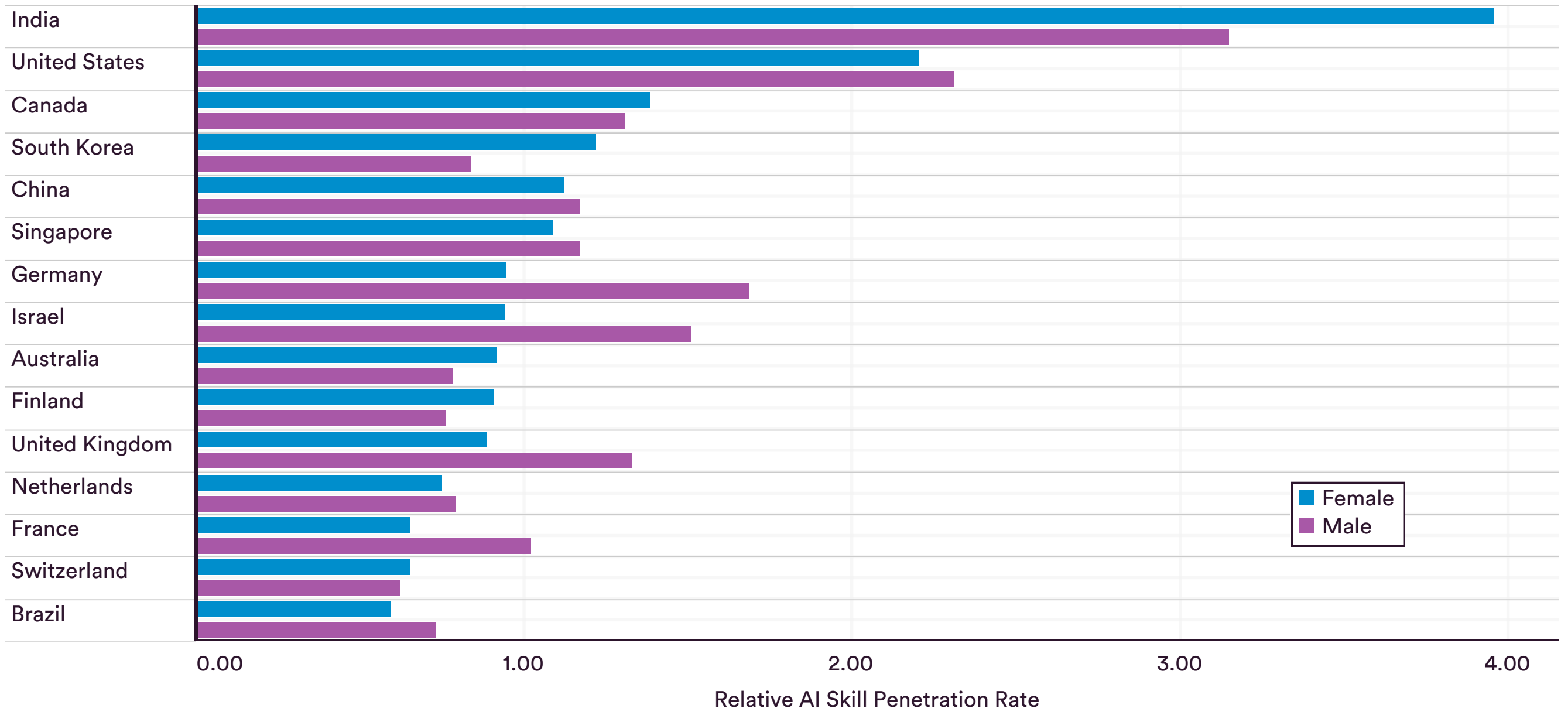

Figure 4.1.10

This section on corporate AI activity draws on data from NetBase Quid, which aggregates over 6 million global public and private company profiles, updated on a weekly basis, including metadata on investments, location of headquarters, and more. NetBase Quid also applies natural language processing technology to search, analyze, and identify patterns in large, unstructured datasets, like aggregated blogs, company and patent databases.

# 4.2 INVESTMENT

## CORPORATE INVESTMENT

Corporate investment in artificial intelligence, from mergers and acquisitions to public offerings, is a key contributor to AI research and development. It also contributes to AI's impact on the economy. Figure 4.2.1 highlights overall global corporate investment in AI from 2013–2021. In 2021, companies made the greatest AI investment through private investment (totaling around $93.5 billion), followed by mergers and acquisitions (around $72 billion), public offerings (around $9.5 billion), and minority stake (around $1.3 billion). In 2021, investments from mergers and acquisitions grew by 3.3 times compared to 2020, led by two AI healthcare companies and two cybersecurity companies.[3]

GLOBAL CORPORATE INVESTMENT in AI by INVESTMENT ACTIVITY, 2013–21

Source: NetBase Quid, 2021 | Chart: 2022 AI Index Report

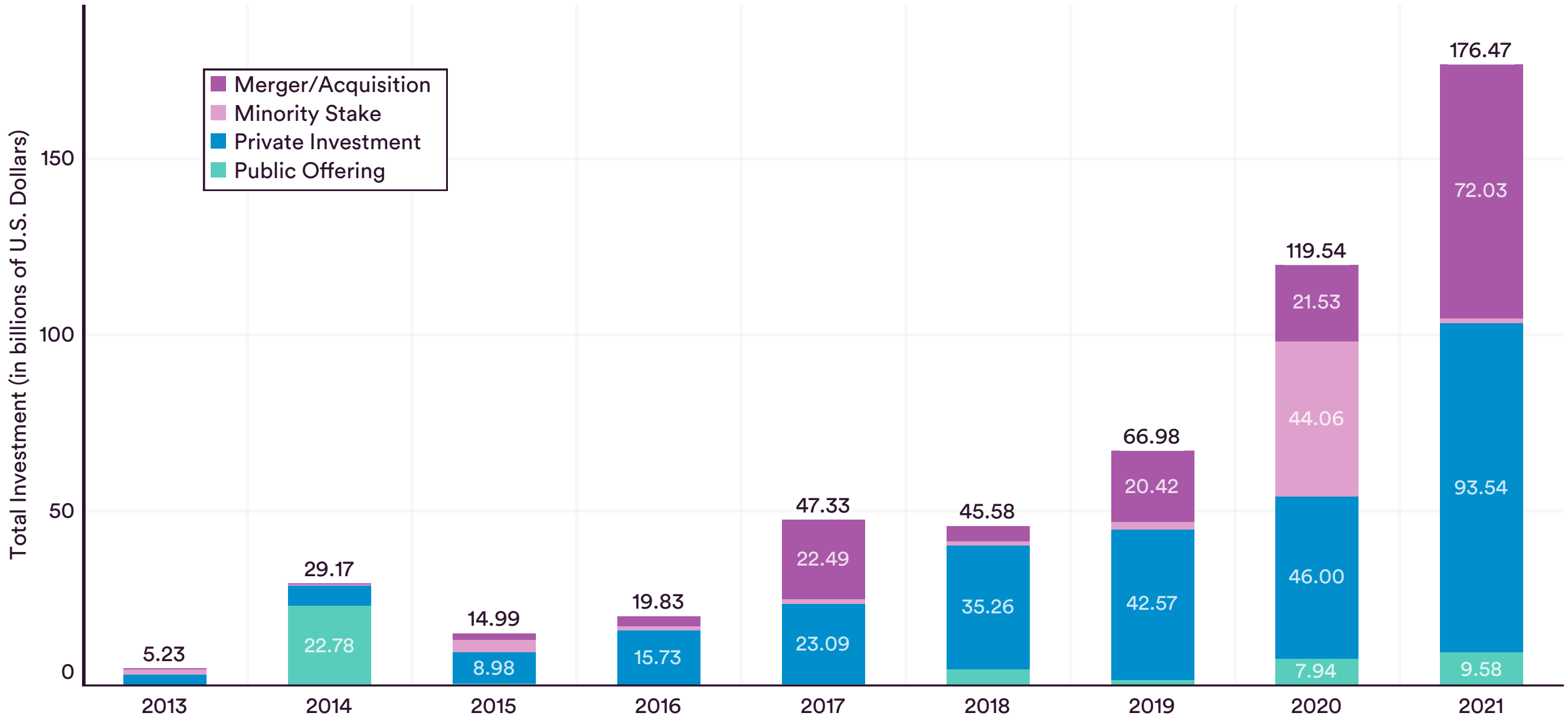

Figure 4.2.1

3 Nuance Communications (by Microsoft, $19.8 billion), Varian Medical Systems (Siemens, $17.2 billion), and Proofpoint (Thoma Bravo, $12.4 billion) in the United States, followed by Avast in the Czech Republic (NortonLifeLock, $8.0 billion).

## STARTUP ACTIVITY

The following section analyzes artificial intelligence and machine learning companies globally that have received more than $1.5 million in investment from 2013 to 2021.

### Global Trend

In 2021, global private investment in AI totaled around $93.5 billion, which is more than double the total private investment in 2020 (Figure 4.2.2). That marks the greatest year-over-year increase since 2014 (when investment from 2013 to 2014 more than doubled).

Among companies that disclosed the amount of funding, the number of AI funding rounds that ranged from $100 million to $500 million more than doubled in 2021 compared to 2020, while funding rounds that were between $50 million and $100 million more than doubled as well (Table 4.2.1). In 2020, there were only four funding rounds worth $500 million or more; in 2021, that number grew to 15. Companies attracted significantly higher investment in 2021, as the average private investment deal size in 2021 was 81.1% higher than in 2020.

However, Figure 4.2.3 shows that the number of newly funded AI companies continues to drop, from 762 companies in 2020 to 746 companies in 2021—the third year of a decline that started in 2018. The largest private investments in 2021 have been led by two data management companies and two robotics/autonomous driving companies.[4]

**In 2021, global private investment in AI totaled around $93.5 billion, which is more than double the total private investment in 2020.**

**PRIVATE INVESTMENT in AI, 2013–21**
Source: NetBase Quid, 2021 | Chart: 2022 AI Index Report

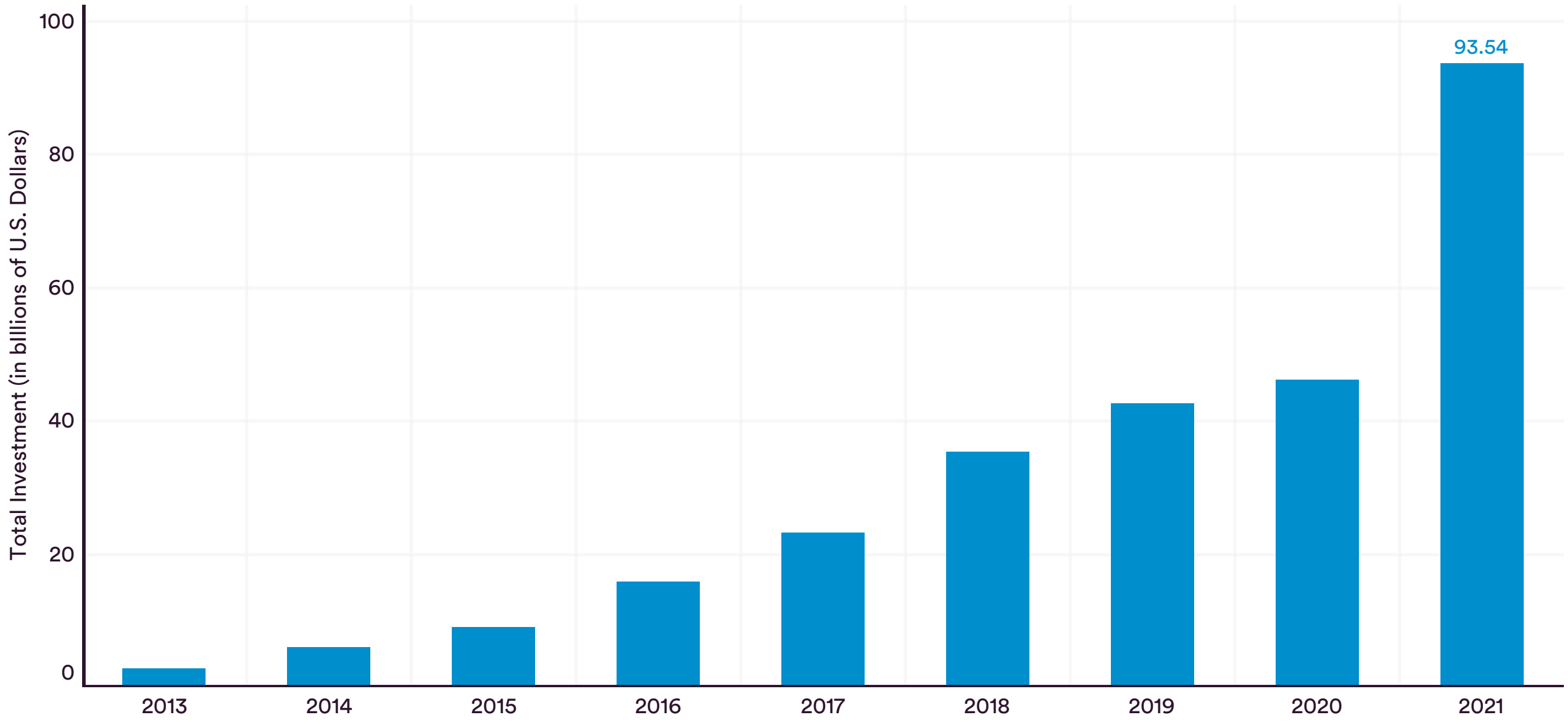

Figure 4.2.2

4 The largest private investments have been Databricks (United States), Beijing Horizon Robotics Technology (China), Oxbotica Limited (United Kingdom), and Celonis (Germany).

NUMBER of NEWLY FUNDED AI COMPANIES in the WORLD, 2013–21
Source: NetBase Quid, 2021 | Chart: 2022 AI Index Report

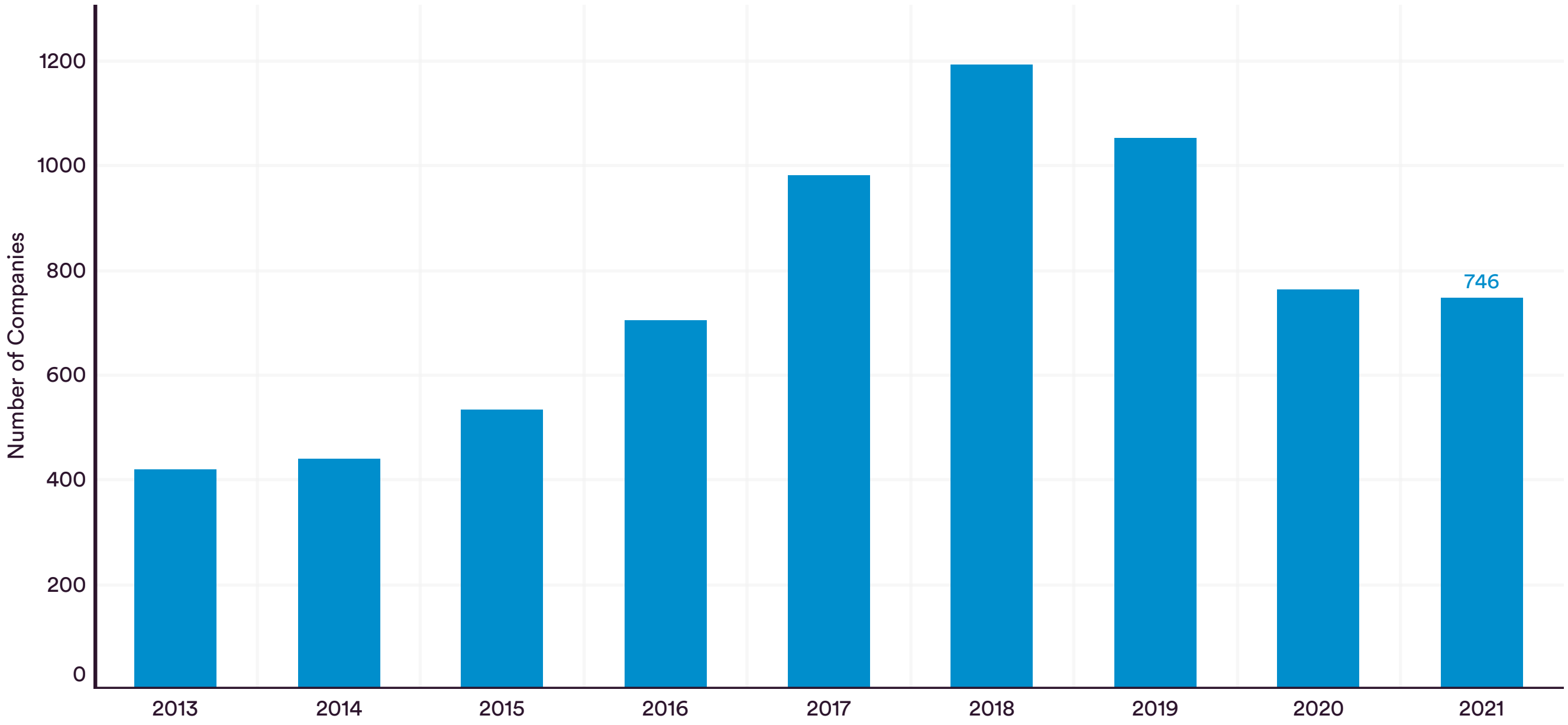

Figure 4.2.3

| Funding Size | 2020 | 2021 | Total |
|---|---|---|---|
| Over $1 billion | 3 | 5 | 8 |
| $500 million – $1 billion | 1 | 10 | 11 |
| $100 million – $500 million | 93 | 235 | 328 |
| $ 50 million – $100 million | 85 | 194 | 279 |
| Under $50 million | 2,102 | 2,120 | 4,222 |
| Undisclosed | 354 | 395 | 749 |
| **Total** | **2,638** | **2,959** | **5,597** |

Table 4.2.1

## Regional Comparison by Funding Amount

In 2021, as captured in Figure 4.2.4, the United States led the world in overall private investment in funded AI companies—at approximately $52.9 billion—over three times the next country on the list, China ( $17.2 billion). In third place was the United Kingdom ($4.65 billion), followed by Israel ($2.4 billion) and Germany ($1.98 billion). Figure 4.2.5 shows that when combining total private investment from 2013 to 2021, the same ranking applies: U.S. investment totaled $149 billion and Chinese investment totaled $61.9 billion, followed by the United Kingdom ($10.8 billion), India ($10.77 billion), and Israel ($6.1 billion). Notably, U.S. private investment in AI companies from 2013–2021 was more than double the total in China, which itself was about six times the total investment from the United Kingdom in the same period. Broken out by geographic area, as shown in Figure 4.2.6, the United States, China, and the European Union all grew their investments from 2020 to 2021, with the United States leading China and the European Union by 3.1 and 8.2 times the investment amount, respectively.

**PRIVATE INVESTMENT in AI by GEOGRAPHIC AREA, 2021**
Source: NetBase Quid, 2021 | Chart: 2022 AI Index Report

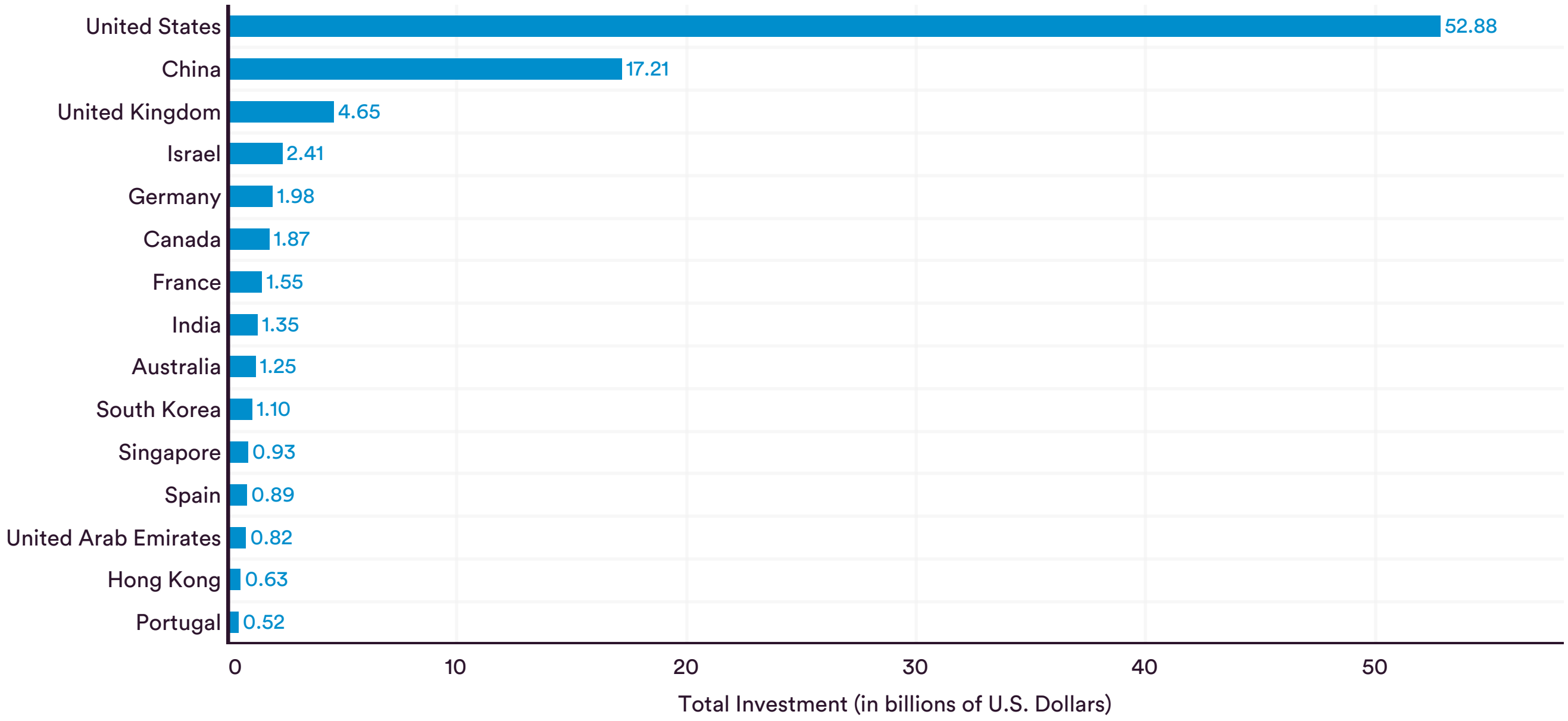

Figure 4.2.4

## PRIVATE INVESTMENT in AI by GEOGRAPHIC AREA, 2013–21

Source: NetBase Quid, 2021 | Chart: 2022 AI Index Report

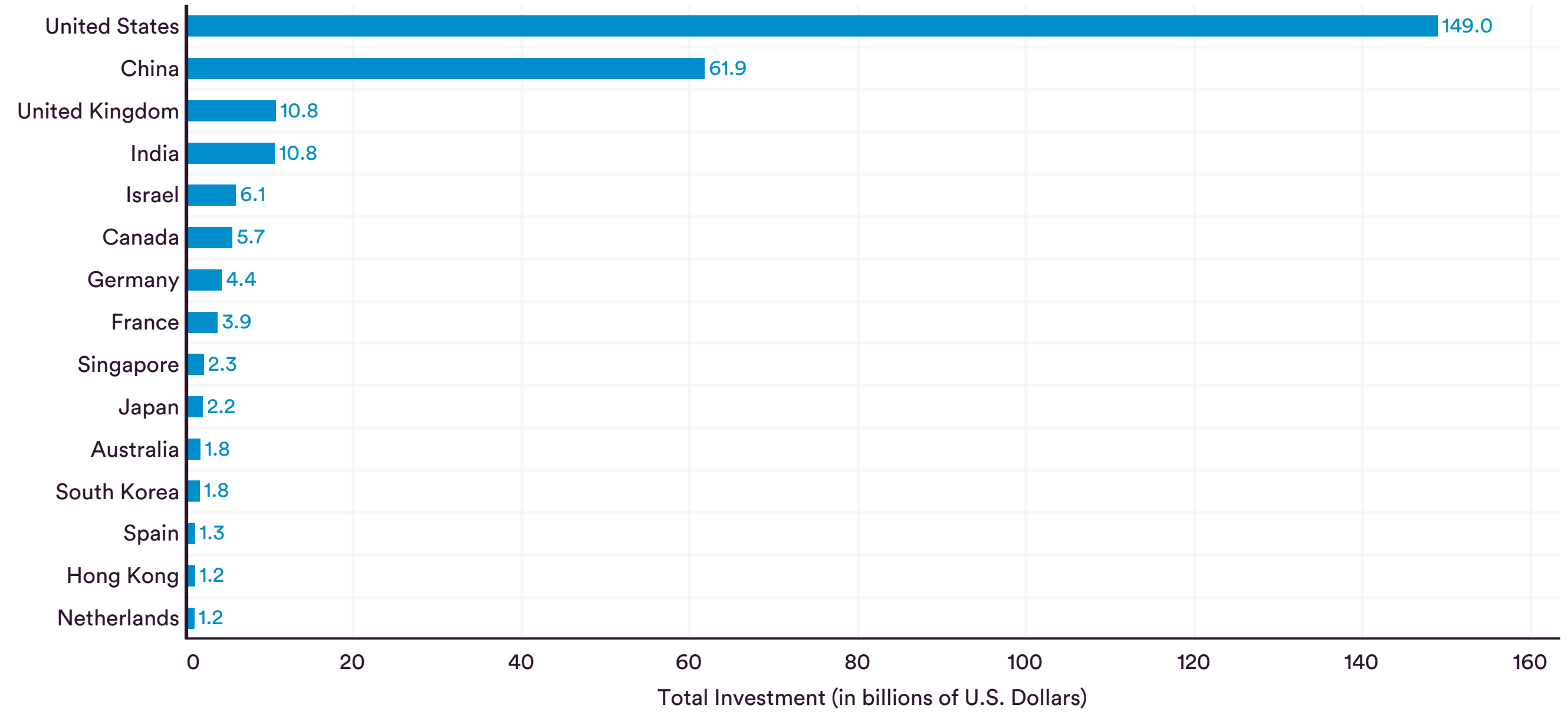

Figure 4.2.5

## PRIVATE INVESTMENT in AI by GEOGRAPHIC AREA, 2013–21

Source: NetBase Quid, 2021 | Chart: 2022 AI Index Report

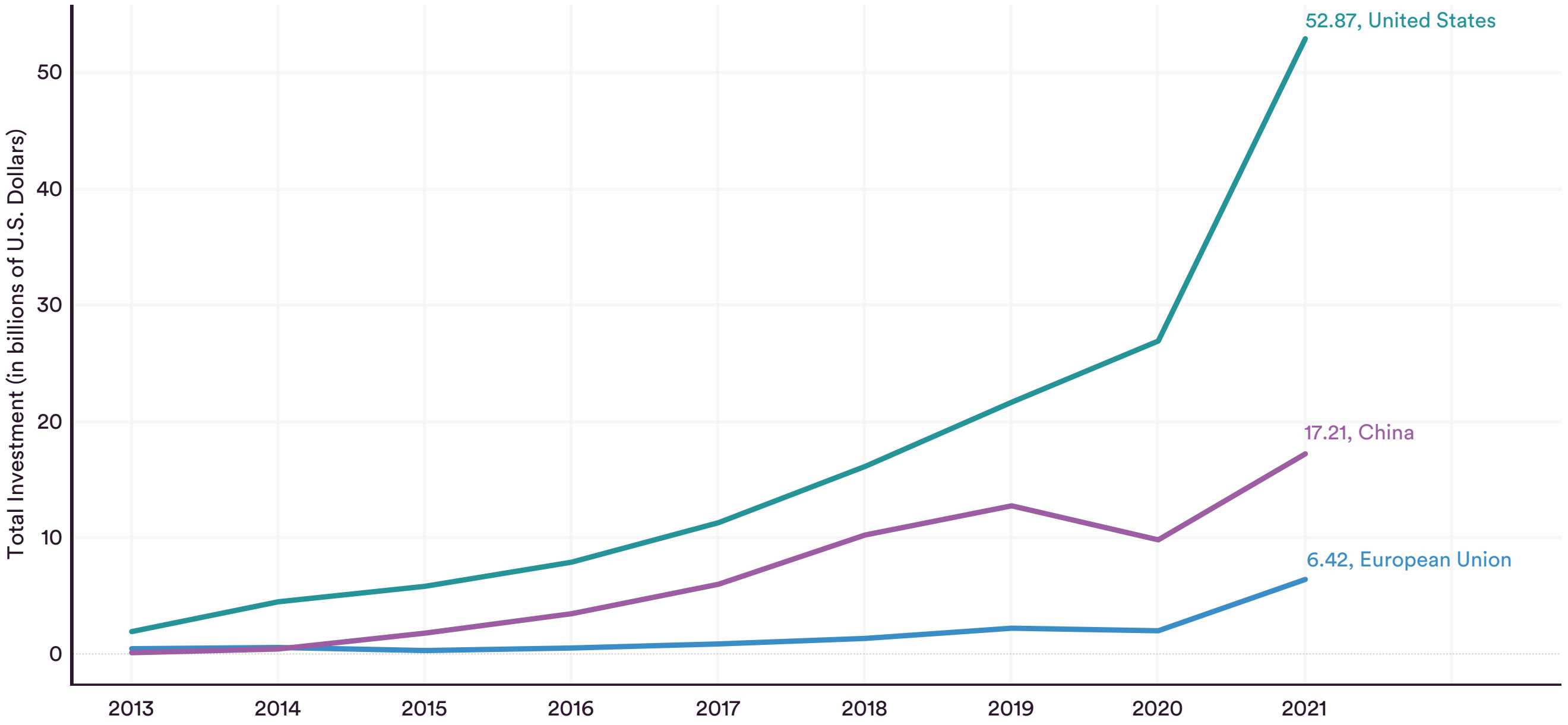

Figure 4.2.6

## Regional Comparison by Newly Funded AI Companies

This section breaks down the investment data by the number of newly funded AI companies in each region. For 2021, Figure 4.2.7 shows that the United States led with 299 companies, followed by China with 119, the United Kingdom with 49, and Israel with 28. The gaps between each are significant. Aggregated data from 2013 to 2021 shows a similar trend (Figure 4.2.8).

However, the number of newly funded AI companies has declined in both the United States and China since 2018 and 2019 (Figure 4.2.9). Despite that downward trend, the United States still leads in the number of newly funded companies, with 299 funded in 2021, followed by China (119) and the European Union (96).

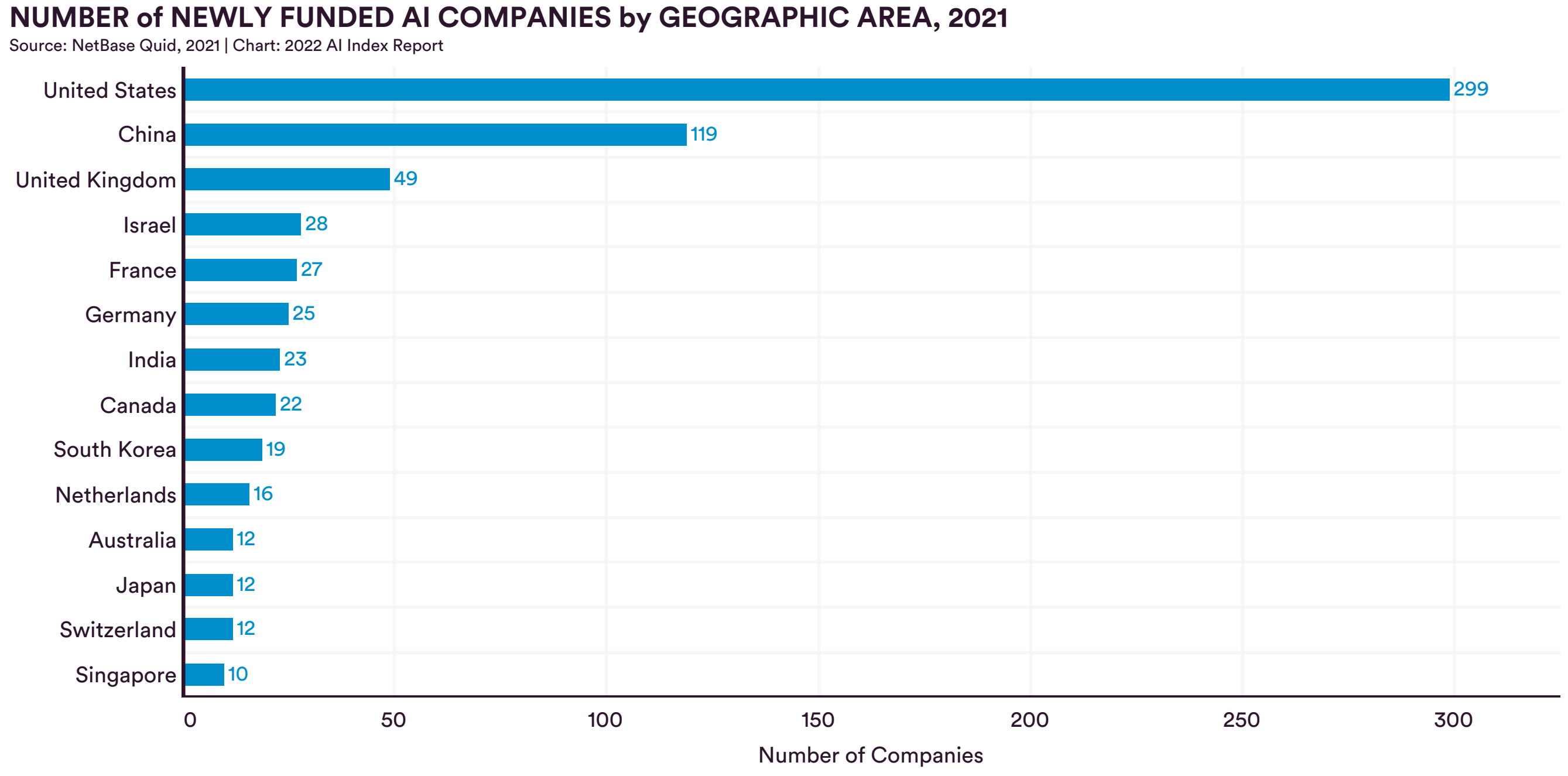

Figure 4.2.7

NUMBER of NEWLY FUNDED AI COMPANIES by GEOGRAPHIC AREA, 2013–21 (SUM)
Source: NetBase Quid, 2021 | Chart: 2022 AI Index Report

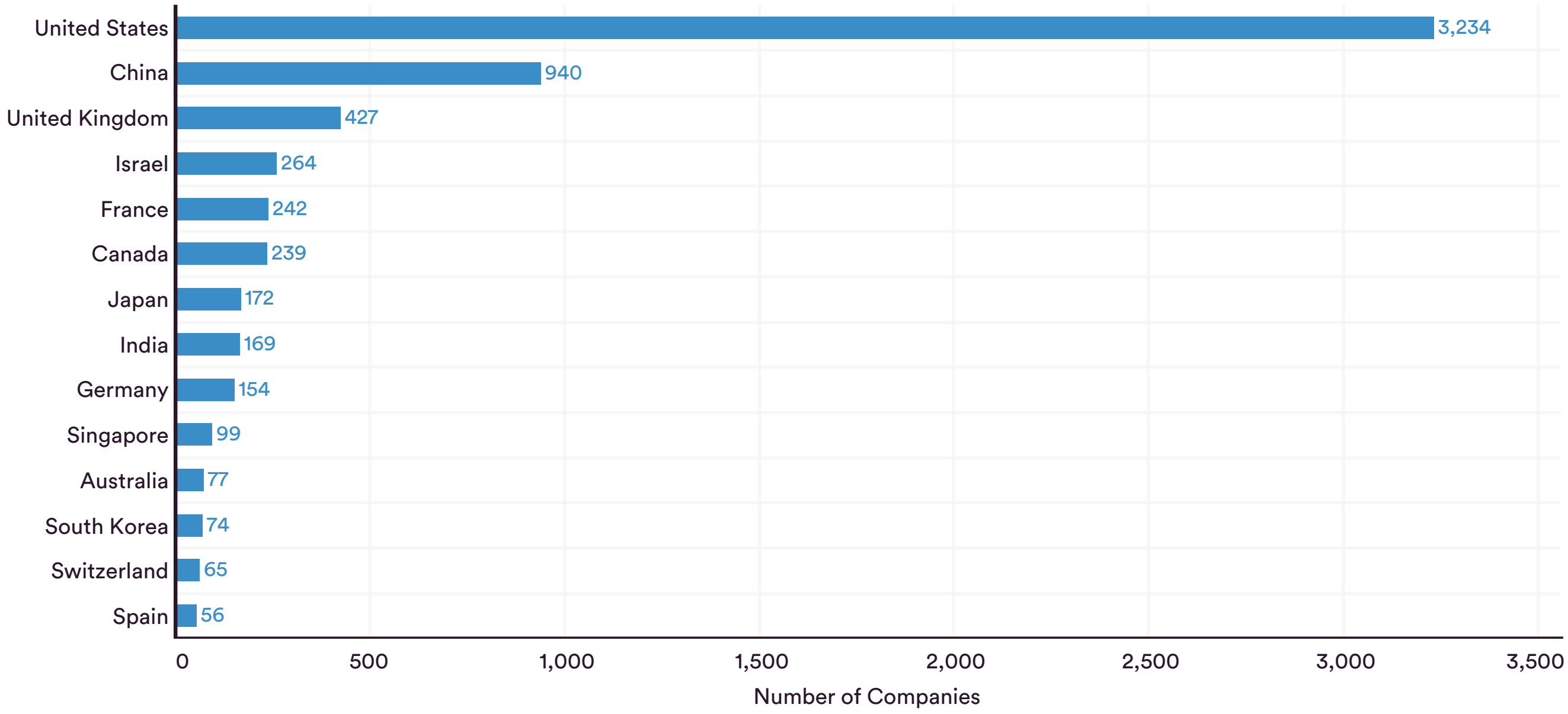

Figure 4.2.8

NUMBER of NEWLY FUNDED AI COMPANIES by GEOGRAPHIC AREA, 2013–21
Source: NetBase Quid, 2021 | Chart: 2022 AI Index Report

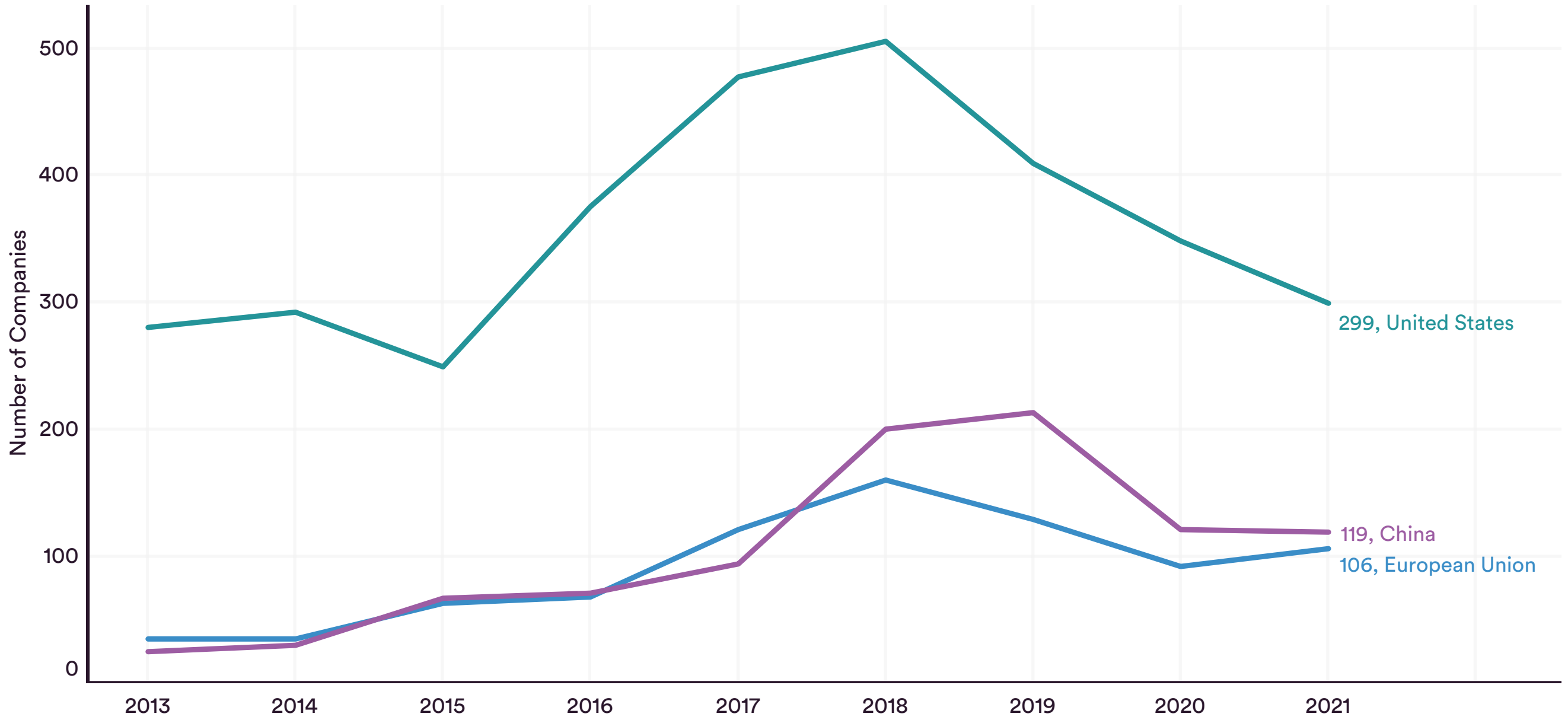

Figure 4.2.9

## Focus Area Analysis

Private AI investment also varies by focus area. According to Figure 4.2.10, the greatest private investment in AI in 2021 was in data management, processing, and cloud (around $12.2 billion). Notably, this was 2.6 times the investment in 2020 (around $4.69 billion) as two of the four largest private investments made in 2021 are data management companies. In second place was private investment in medical and healthcare ($11.29 billion), followed by fintech ($10.26 billion), AV ($8.09 billion), and semiconductors ($6.0 billion).

Aggregated data in Figure 4.2.11 shows that in the last five years, the medical and healthcare category received the largest private investment globally ($28.9 billion); followed by data management, processing, and cloud ($26.9 billion); fintech ($24.9 billion); and retail ($21.95 billion). Moreover, Figure 4.2.12 shows the overall trend in private investment by industries from 2017–2021 and reveals a steady increase in AV, cybersecurity and data protection, fitness and wellness, medical and healthcare, and semiconductor industries.

**PRIVATE INVESTMENT in AI by FOCUS AREA, 2020 vs. 2021**

Source: NetBase Quid, 2021 | Chart: 2022 AI Index Report

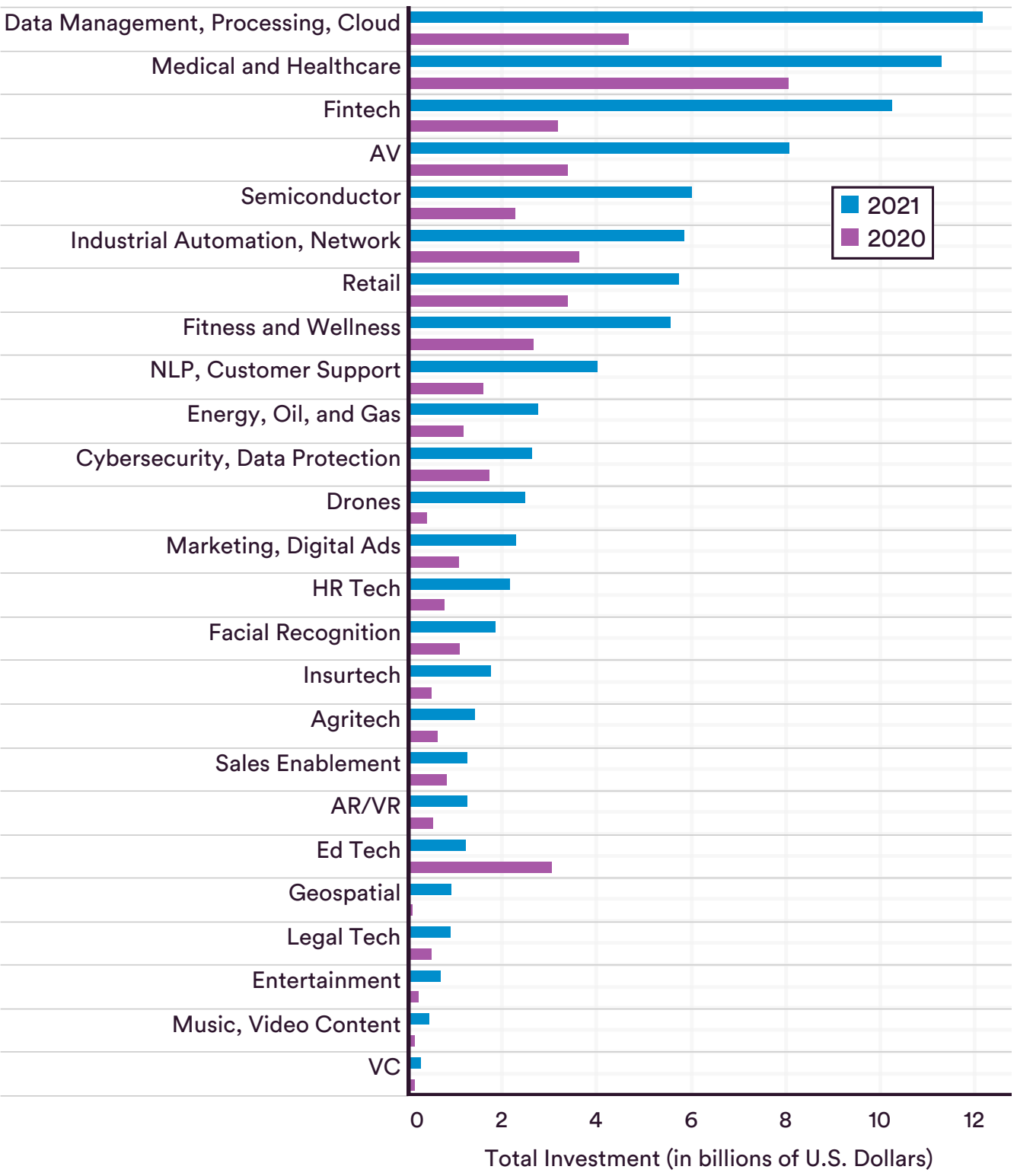

Figure 4.2.10

**PRIVATE INVESTMENT in AI by FOCUS AREA, 2017–21 (SUM)**

Source: NetBase Quid, 2021 | Chart: 2022 AI Index Report

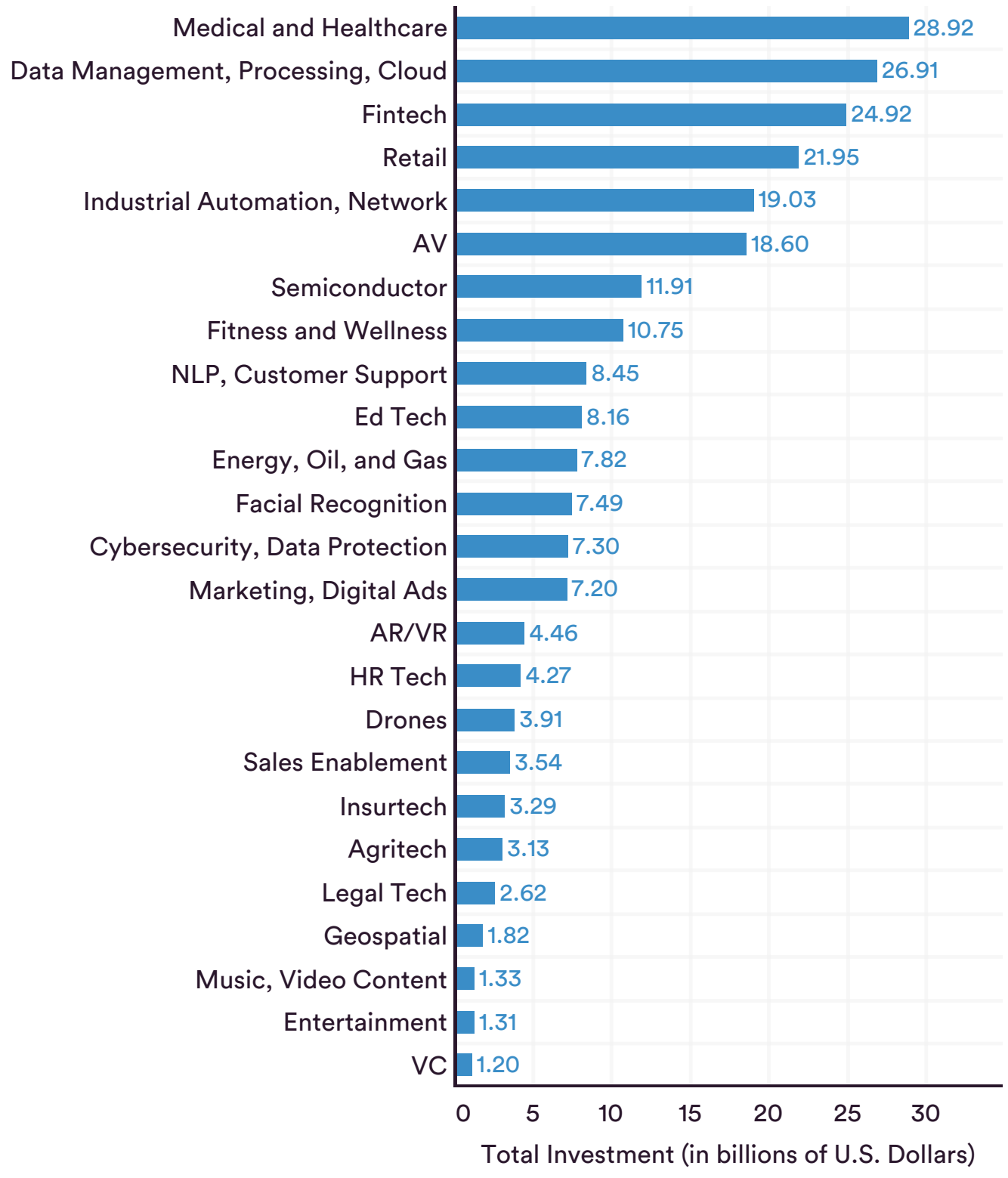

Figure 4.2.11

**PRIVATE INVESTMENT in AI by FOCUS AREA, 2017–21**
Source: NetBase Quid, 2021 | Chart: 2022 AI Index Report

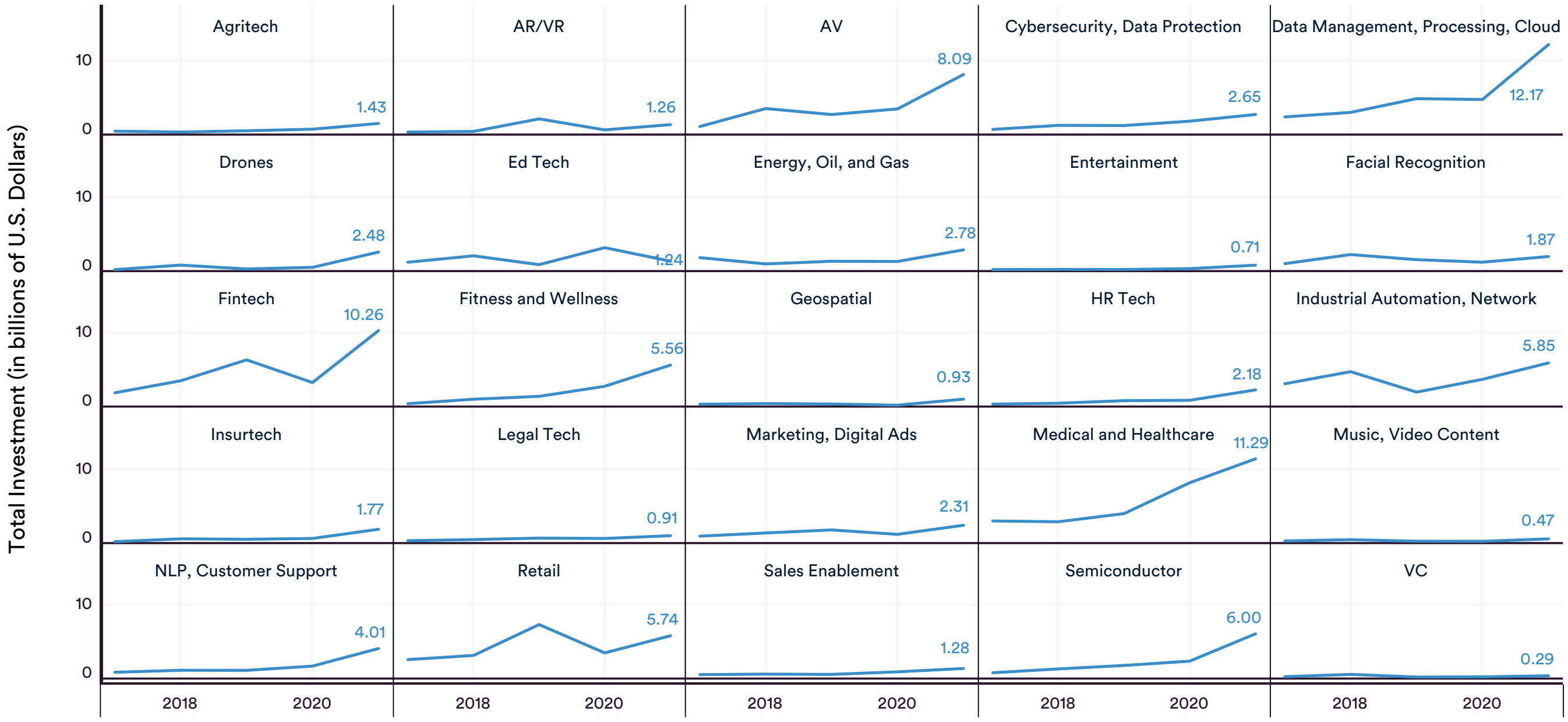

Figure 4.2.12

# 4.3 CORPORATE ACTIVITY

## INDUSTRY ADOPTION

This section on corporate AI activity draws on McKinsey's "The State of AI in 2021" report from December 2021. The report based its conclusions on a global online survey of 1,843 participants conducted earlier in 2021. Survey respondents came from a range of industries, companies, functional specialties, tenures, and regions of the world—and each provided answers to questions about the state of artificial intelligence today.

### Global Adoption of AI

Figure 4.3.1 shows AI adoption by organizations globally, broken out by geographic area. In 2021, India led with 65% adoption, followed by "Developed Asia-Pacific" (64%), "Developing markets (incl. China, MENA)" (57%), and North America (55%). The average adoption rate across all geographies was 56%, up 6% from 2020. Notably, "Developing markets (incl. China, MENA)" registered a 21% increase from 2020, and India had an 8% increase from 2020.

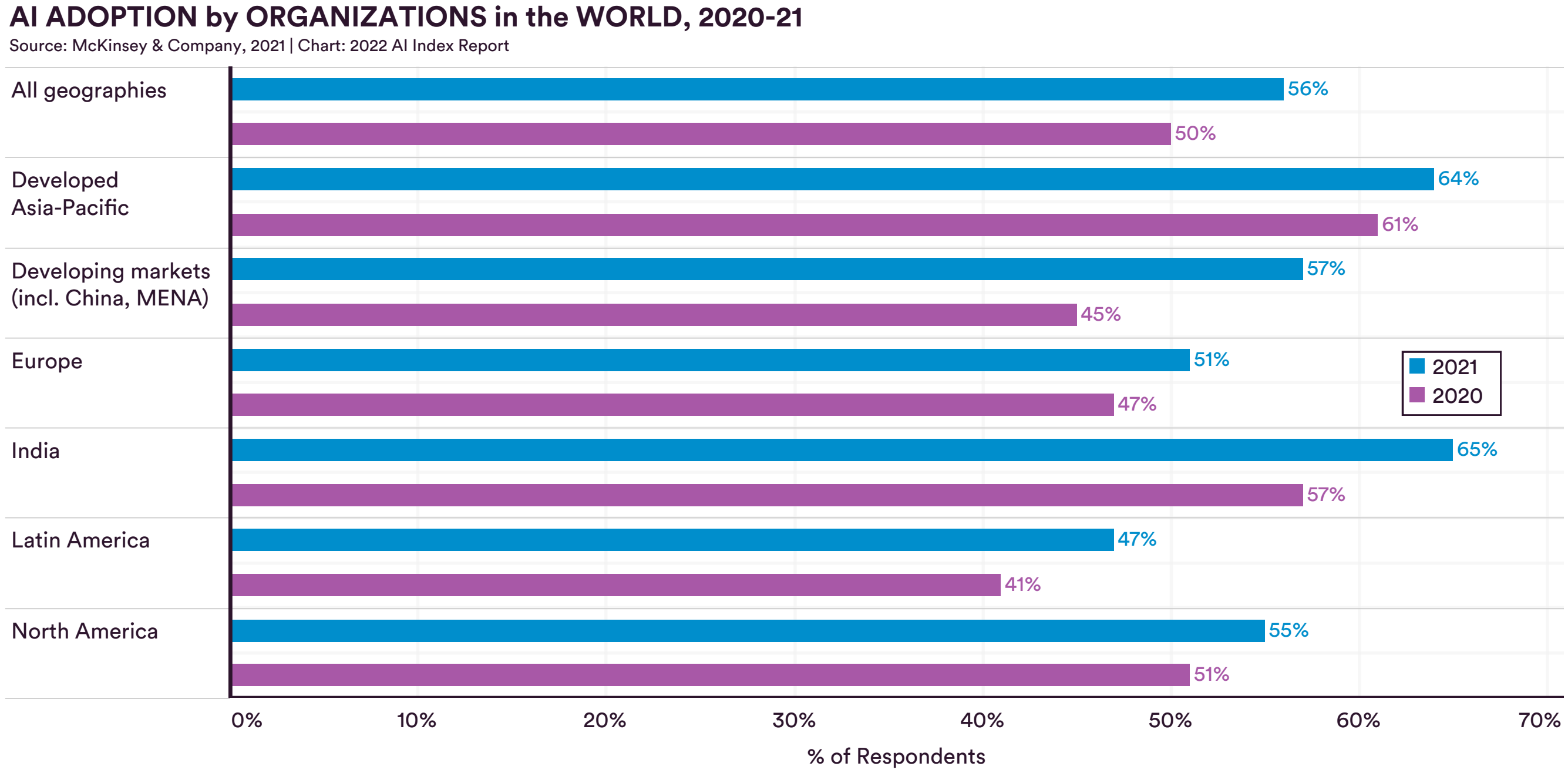

Figure 4.3.1

## AI Adoption by Industry and Function

Figure 4.3.2 shows AI adoption by industry and function in 2021. The greatest adoption was in product and/or service development for high tech/telecommunications (45%), followed by service operations for financial services (40%), service operations for high tech/telecommunications (34%), and risk function for financial services (32%).

### AI ADOPTION by INDUSTRY and FUNCTION, 2021

Source: McKinsey & Company, 2021 | Chart: 2022 AI Index Report

| Industry | Human Resources | Manufacturing | Marketing and Sales | Product and/or Service Development | Risk | Service Operations | Strategy and Corporate Finance | Supply-chain Management |
|---|---|---|---|---|---|---|---|---|
| All Industries | 9% | 12% | 20% | 23% | 13% | 25% | 9% | 13% |
| Automotive and Assembly | 11% | 26% | 20% | 15% | 4% | 18% | 6% | 17% |
| Business, Legal, and Professional Services | 14% | 8% | 28% | 15% | 13% | 26% | 8% | 13% |
| Consumer Goods/Retail | 2% | 18% | 22% | 17% | 1% | 15% | 4% | 18% |
| Financial Services | 10% | 4% | 24% | 20% | 32% | 40% | 13% | 8% |
| Healthcare Systems/Pharma and Medical Products | 9% | 11% | 14% | 29% | 13% | 17% | 12% | 9% |
| High Tech/Telecom | 12% | 11% | 28% | 45% | 16% | 34% | 10% | 16% |

% of Respondents (Function)

Figure 4.3.2

## Type of AI Capabilities Adopted

With respect to the type of AI capabilities embedded in standard business processes in 2021, as indicated by Figure 4.3.3, the highest rate of embedding was in natural language text understanding for the high tech/telecommunications industry (34%), followed by robotic process automation for both the financial services and automotive and assembly industry (33%) and natural language text understanding for financial services (32%).

### AI CAPABILITIES EMBEDDED in STANDARD BUSINESS PROCESSES, 2021

Source: McKinsey & Company, 2021 | Chart: 2022 AI Index Report

| Industry | Computer Vision | Deep Learning | Facial Regonition | Knowledge Graphs | NL Generation | NL Speech Un-derstanding | NL Text Un-derstanding | Physical Robotics | Recom-mender Systems | Reinforce-ment Learning | Robotic Process Automation | Simulations | Transfer Learning | Virtual Agents |
|---|---|---|---|---|---|---|---|---|---|---|---|---|---|---|
| All Industries | 23% | 19% | 11% | 17% | 12% | 14% | 24% | 12% | 17% | 16% | 26% | 17% | 12% | 23% |
| Automotive and Assembly | 15% | 14% | 9% | 16% | 3% | 11% | 12% | 24% | 12% | 5% | 33% | 27% | 6% | 12% |
| Business, Legal, and Professional Services | 29% | 24% | 15% | 20% | 23% | 18% | 19% | 13% | 22% | 27% | 31% | 18% | 21% | 19% |
| Consumer Goods/Retail | 23% | 12% | 14% | 17% | 11% | 13% | 14% | 4% | 8% | 8% | 16% | 9% | 1% | 15% |
| Financial Services | 17% | 16% | 11% | 16% | 12% | 18% | 32% | 4% | 13% | 16% | 33% | 12% | 12% | 28% |
| Healthcare Systems/Pharma and Medical Products | 30% | 25% | 12% | 19% | 10% | 8% | 26% | 28% | 22% | 13% | 28% | 22% | 19% | 31% |
| High Tech/Telecom | 28% | 22% | 6% | 17% | 17% | 18% | 34% | 5% | 19% | 15% | 23% | 14% | 11% | 25% |

% of Respondents (AI Capability)

Figure 4.3.3

## Consideration and Mitigation of Risks From Adopting AI

The risk from adopting AI that survey respondents identified as most relevant in 2021 was cybersecurity (55% of respondents), followed by regulatory compliance (48%), explainability (41%), and personal/individual privacy (41%) (Figure 4.3.4). Fewer organizations found AI risks from cybersecurity to be relevant in 2021 than in 2020, declining from just over 60% of respondents expressing concern in 2020 to 55% in 2021. Concern over AI regulatory compliance risks, meanwhile, remained virtually unchanged from 2020.

Figure 4.3.5 shows risks from AI that organizations are taking steps to mitigate. Cybersecurity was the most frequent response (47% of respondents), followed by regulatory compliance (36%), personal/individual privacy (28%), and explainability (27%). It is worth noting the gaps between risks that organizations find relevant and risks that organizations take steps to mitigate—a gap of 10 percentage points with equity and fairness (29% to 19%), 12 percentage points with regulatory compliance (48% to 36%), 13 percentage points with personal/individual privacy (41% to 28%), and 14 percentage points with explainability (41% to 27%).

RISKS from ADOPTING AI that ORGANIZATIONS CONSIDER RELEVANT, 2019–21

Source: McKinsey & Company, 2021 | Chart: 2022 AI Index Report

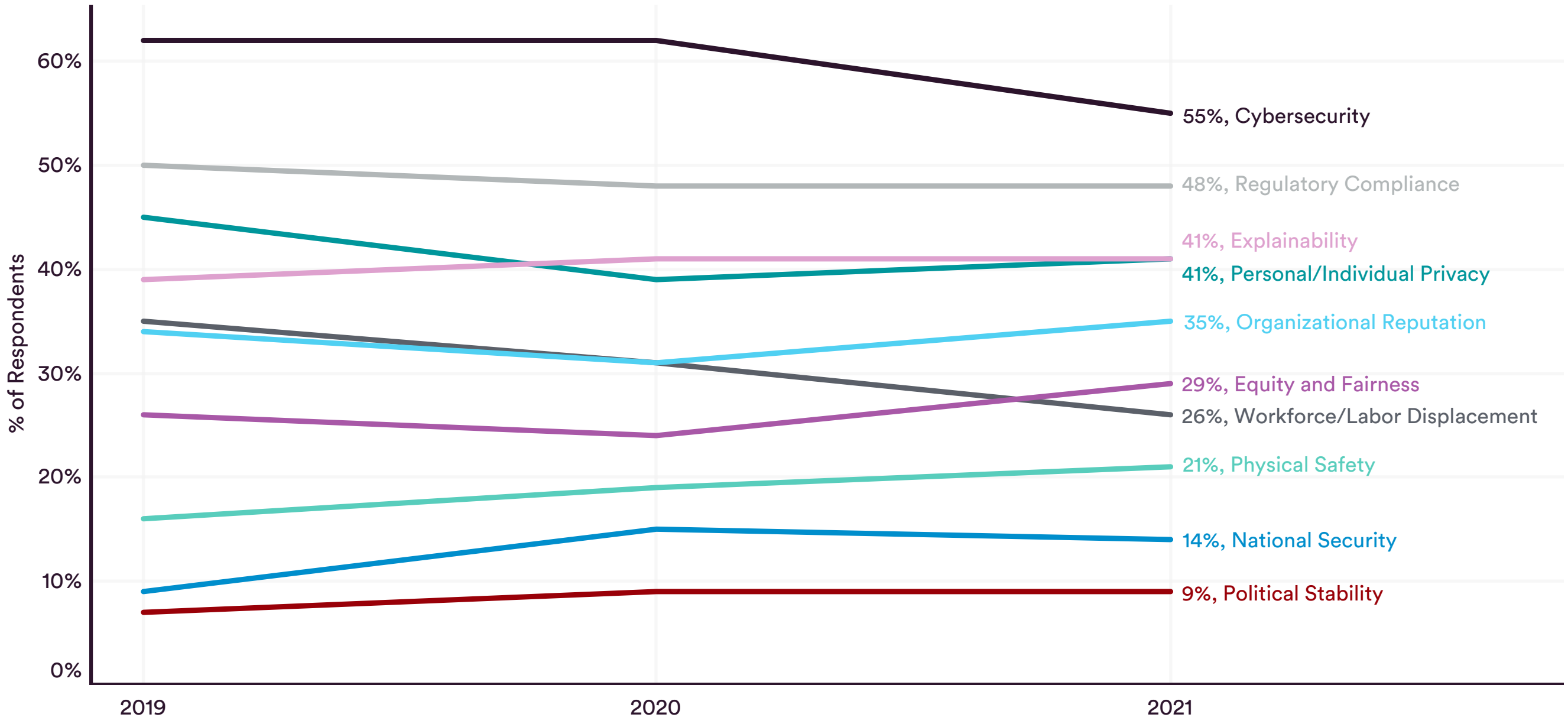

Figure 4.3.4

**RISKS from ADOPTING AI that ORGANIZATIONS TAKE STEPS to MITIGATE, 2019–21**
Source: McKinsey & Company, 2021 | Chart: 2022 AI Index Report

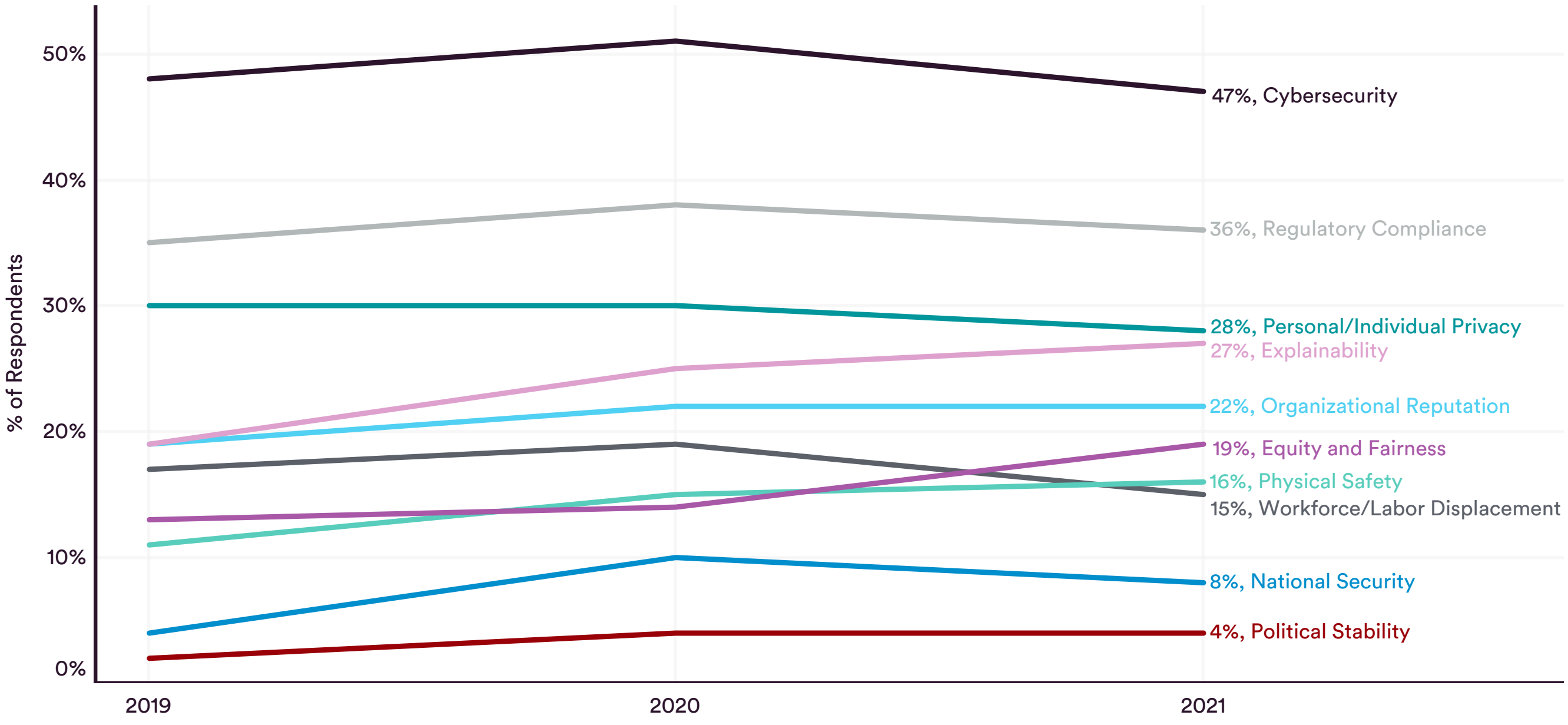

Figure 4.3.5

**The following section draws on data from the annual Computing Research Association (CRA) Taulbee Survey. For the latest survey featured in this section, CRA collected data in Fall 2020 by reaching out to over 200 PhD-granting departments in the United States and Canada. Results are published in May 2021. The CRA survey documents trends in student enrollment, degree production, employment of graduates, and faculty salaries in academic units in the United States and Canada that grant doctoral degrees in computer science (CS), computer engineering (CE), or information (I). Academic units include departments of CS and CE or, in some cases, colleges or schools of information or computing.**

# 4.4 AI EDUCATION

## CS UNDERGRADUATE GRADUATES IN NORTH AMERICA

In North America, most AI-related courses are offered as part of the CS curriculum at the undergraduate level.

The number of new CS undergraduate graduates at doctoral institutions in North America has grown 3.5 times from 2010 to 2020 (Figure 4.4.1). More than 31,000 undergraduates completed CS degrees in 2020—an 11.60% increase from the number in 2019.

**NUMBER of NEW CS UNDERGRADUATE GRADUATES at DOCTORAL INSTITUTIONS in NORTH AMERICA, 2010–20**
Source: CRA Taulbee Survey, 2021 | Chart: 2022 AI Index Report

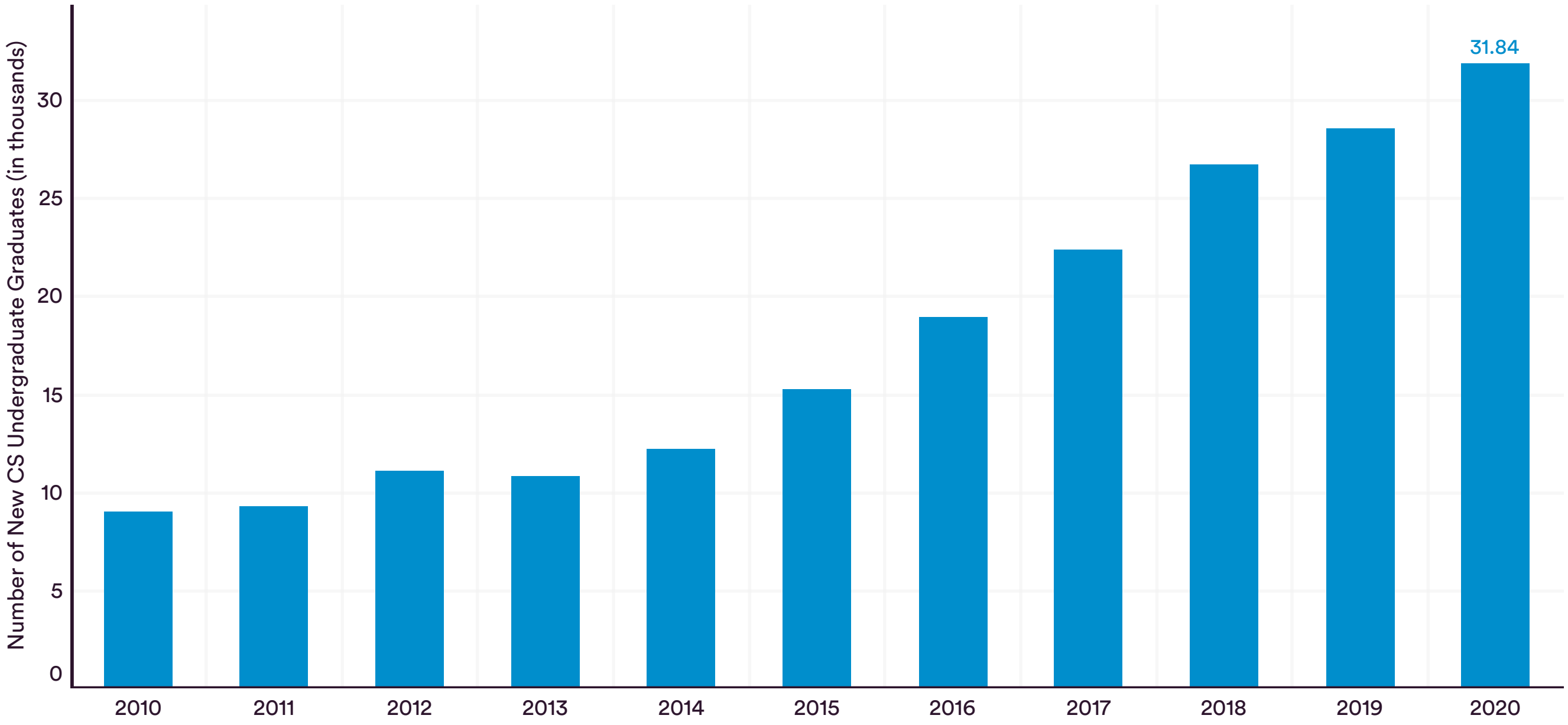

Figure 4.4.1

## NEW CS PHDS IN NORTH AMERICA

The following sections show the trend of CS PhD graduates in North America with a focus on those with AI-related specialties.[5] The CRA survey includes 20 specialties in total, two of which are directly related to the field of AI: artificial intelligence/machine learning (AI/ML) and robotics/vision.

### New CS PhDs by Specialty

In 2020, 1 in every 5 CS students who graduated with PhD degrees specialized in AI/ML, the most popular specialty in the past decade (Figure 4.4.2). It is also the speciality that exhibits the most significant growth from 2010 to 2021, relative to 18 other specializations (Figure 4.4.3). Robotics/vision is also among the most popular CS specialties of PhD graduates in 2020, registering a 1.4 percentage point change in the share of total new CS PhDs in the past 11 years.

**In 2020, 1 in every 5 CS students who graduated with PhD degrees specialized in AI/ML, the most popular specialty in the past decade.**

**NEW CS PHDS (% of TOTAL) in the UNITED STATES by SPECIALITY, 2020**
Source: CRA Taulbee Survey, 2021 | Chart: 2022 AI Index Report

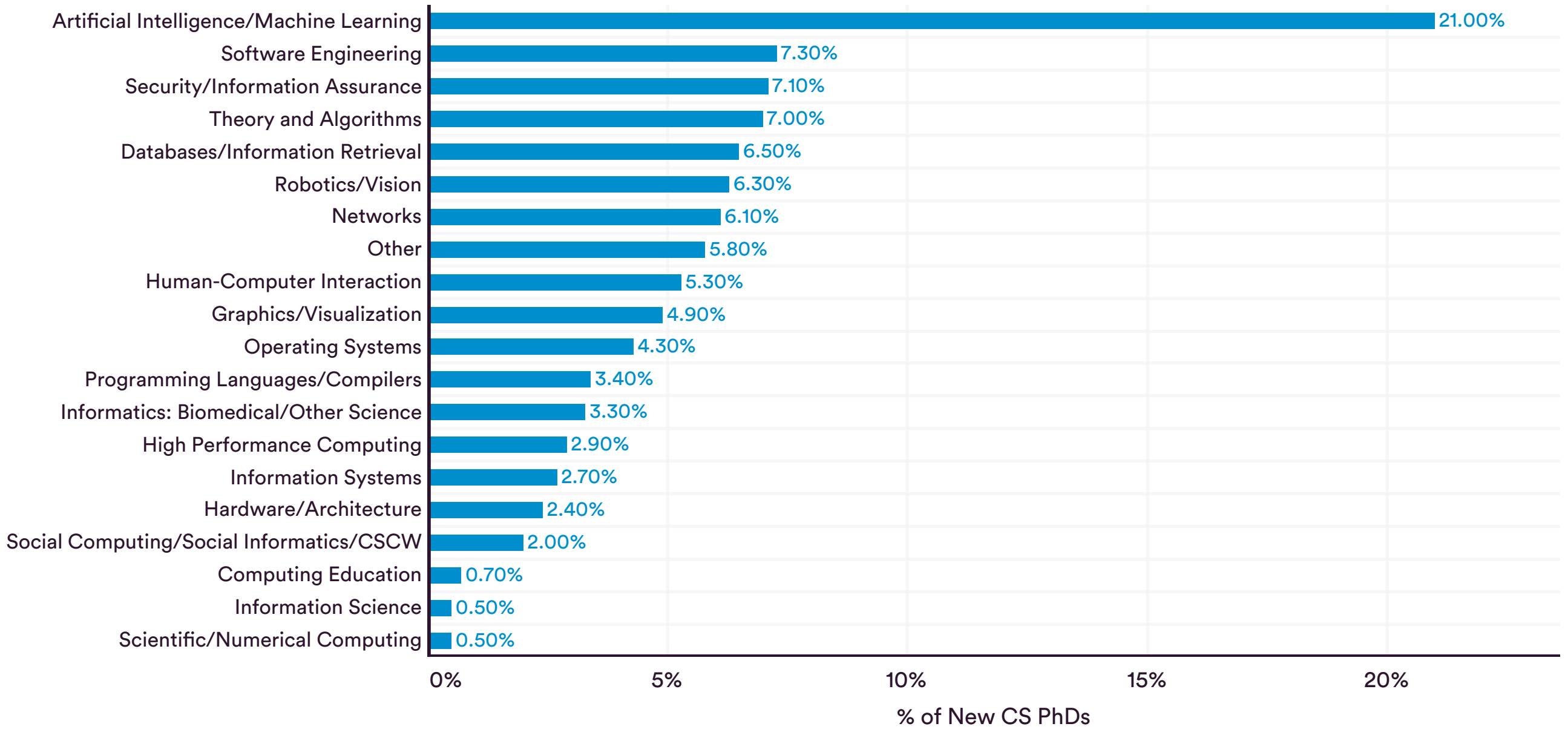

Figure 4.4.2

5 New CS PhDs in this section include PhD graduates from academic units (departments, colleges, or schools within universities) of computer science in the United States.

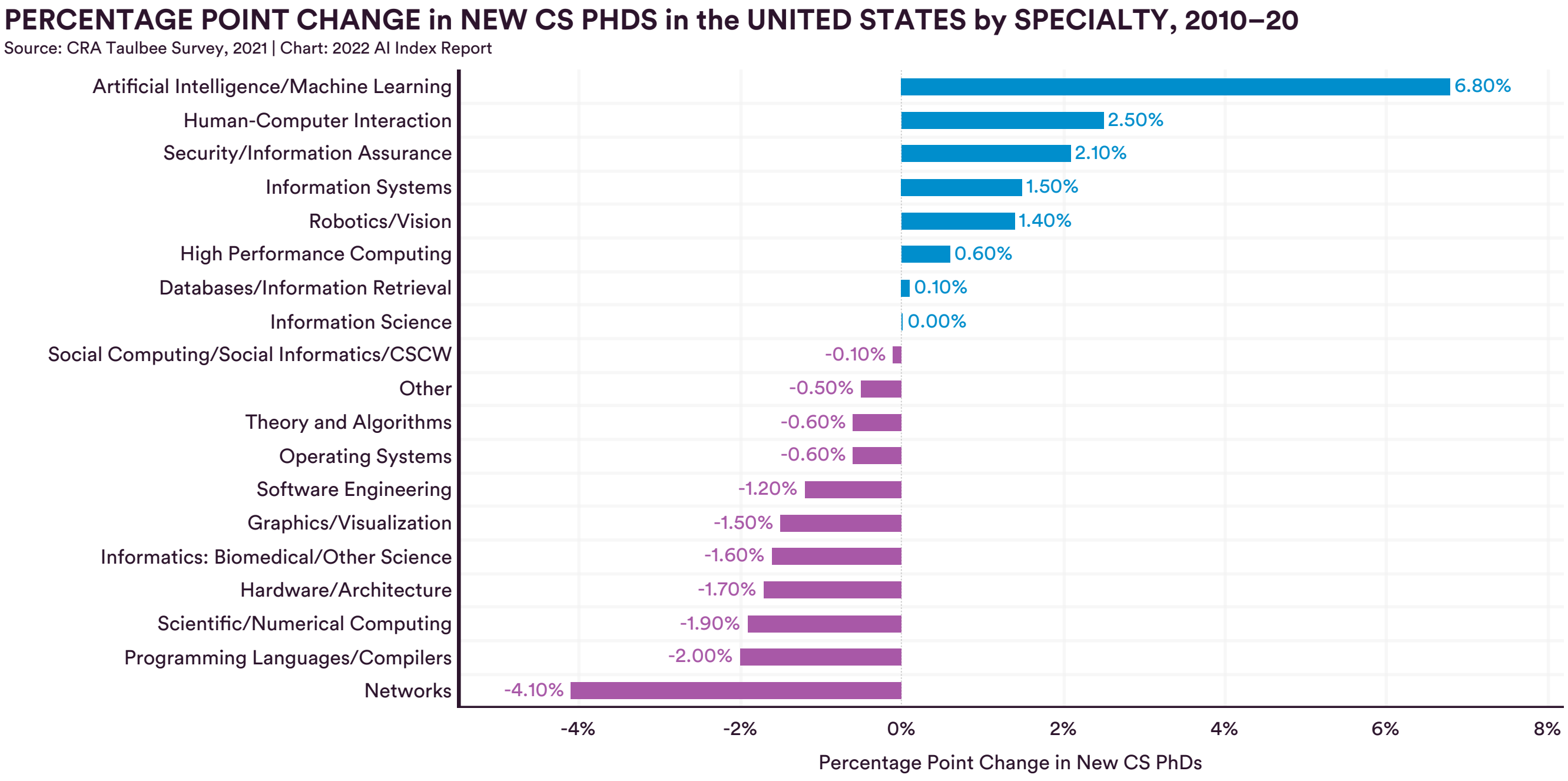

Figure 4.4.3

## New CS PhDs with AI/ML and Robotics/Vision Specialties

Between 2010 and 2020, the number of CS PhD graduates with AI/ML and robotics/vision specialities grew by 72.05% and 50.91%, respectively. The slight decrease in the total number for both specialties from 2019 to 2020 may be due to the impact of the COVID-19 pandemic.

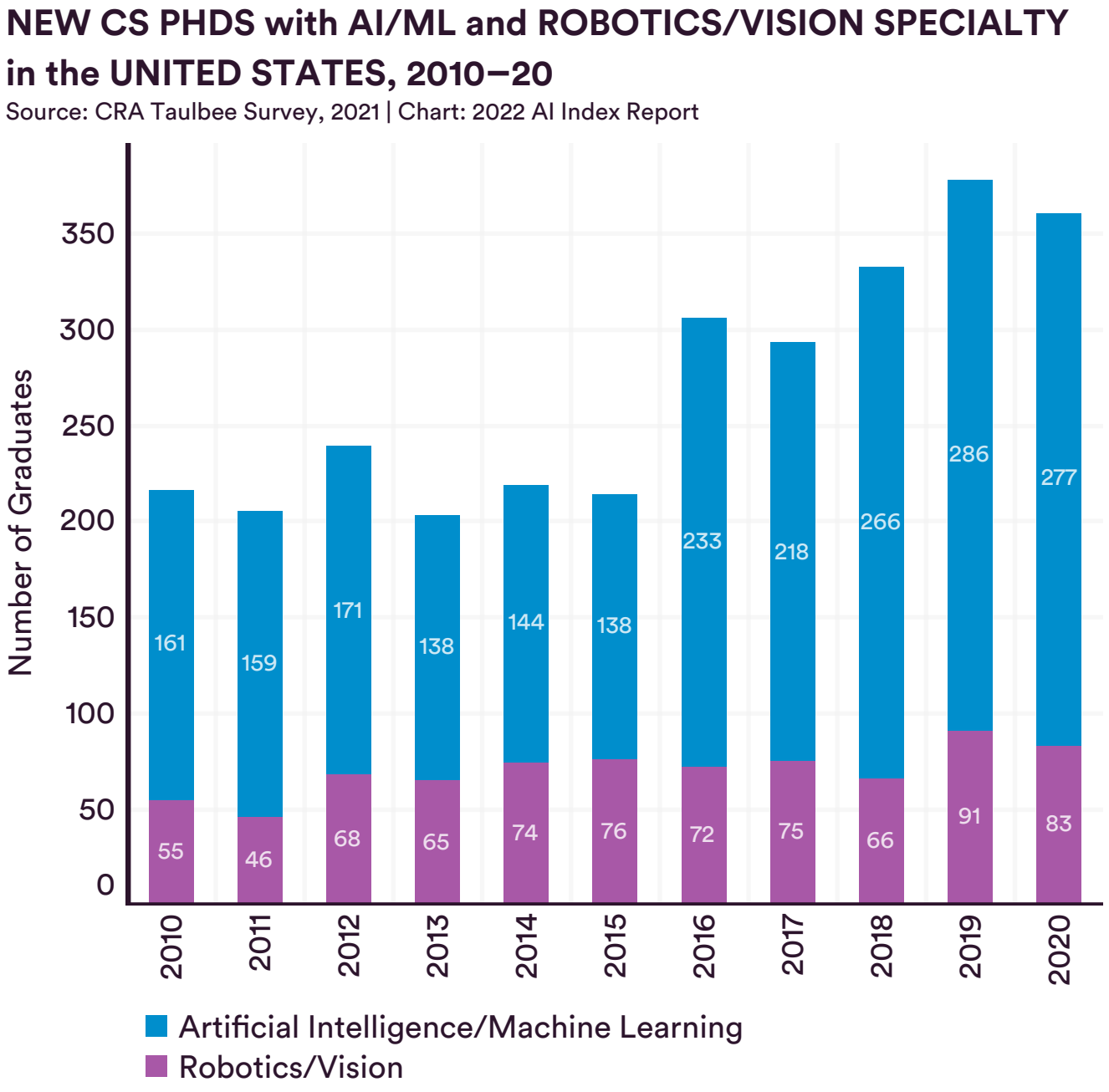

Figure 4.4.4a

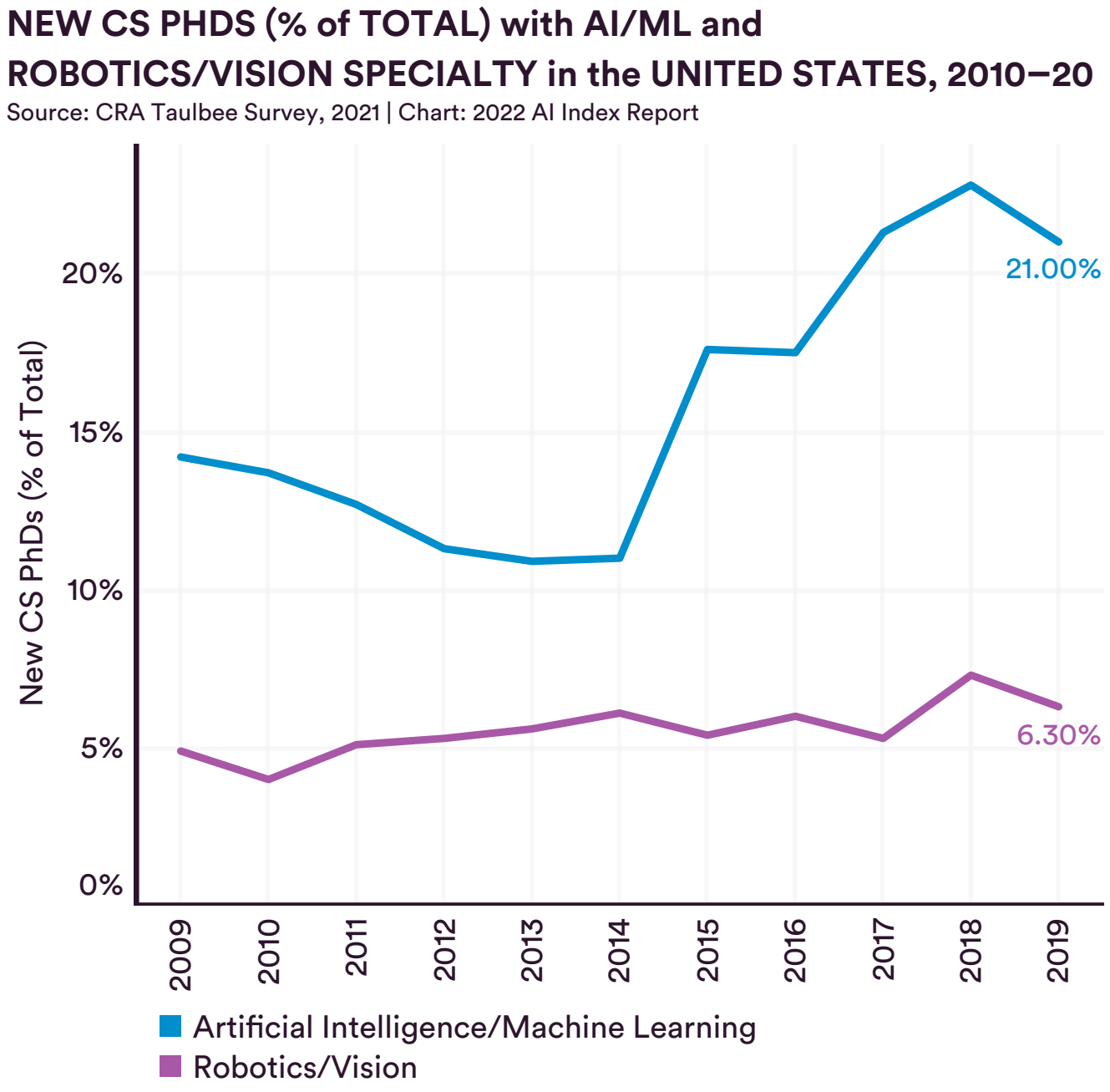

Figure 4.4.4b

## NEW AI PHDS EMPLOYMENT IN NORTH AMERICA

Where do new AI PhDs choose to work following graduation? This section analyzes the employment trends of new AI PhDs across North America in academia, industry, and government.[6]

### Academia vs. Industry vs. Government

In 2020, the share of new AI PhD graduates in North America who chose to work in the industry dipped slightly, with its share dropping from 65.7% in 2019 to 60.2% in 2020, whereas the share of new AI PhDs who went into academia and government changed little (Figure 4.4.5a and Figure 4.4.5b). Note that the 2020 data may be impacted by the increasing number of new AI PhDs who went abroad upon graduation, a number that grew from 19 in 2019 to 32 in 2020.

**EMPLOYMENT of NEW AI PHDS to ACADEMIA, GOVERNMENT, or INDUSTRY in NORTH AMERICA, 2010–20**

Source: CRA Taulbee Survey, 2021 | Chart: 2022 AI Index Report

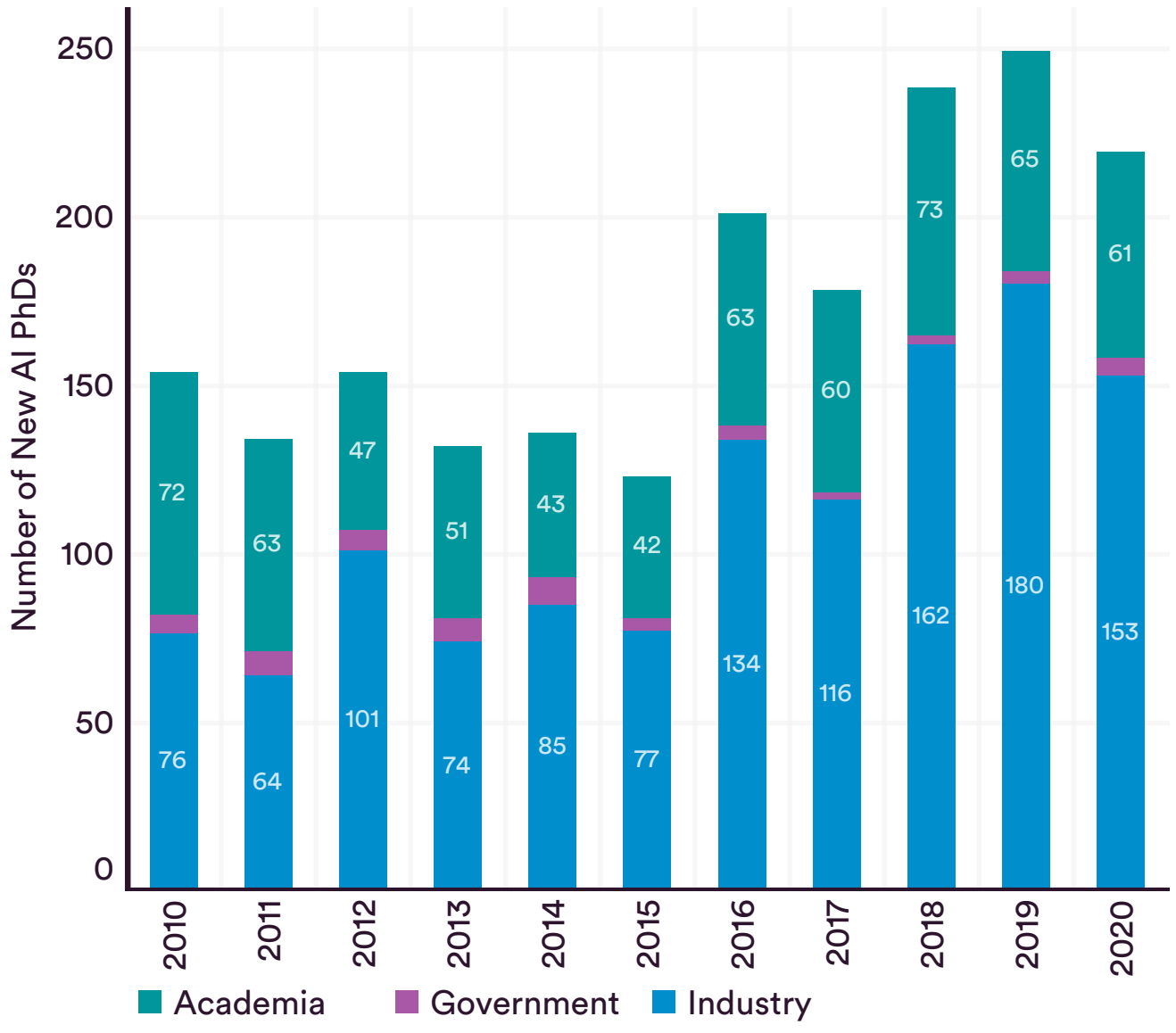

Figure 4.4.5a

**EMPLOYMENT of NEW AI PHDS (% of TOTAL) to ACADEMIA, GOVERNMENT, or INDUSTRY in NORTH AMERICA, 2010–20**

Source: CRA Taulbee Survey, 2021 | Chart: 2022 AI Index Report

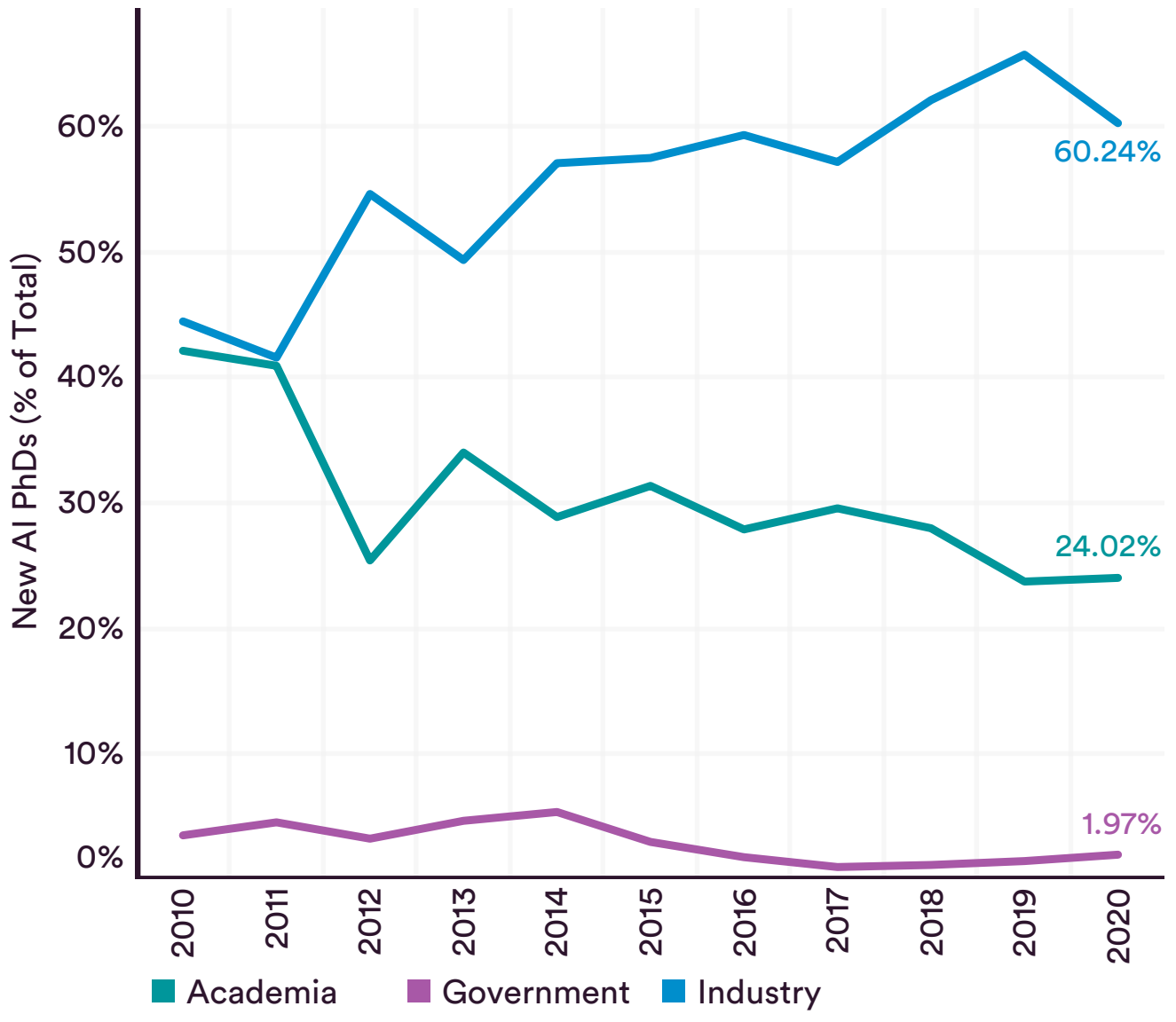

Figure 4.4.5b

6 New AI PhDs in this section include PhD graduates who specialize in artificial intelligence from academic units (departments, colleges, or schools within universities) of computer science, computer engineering, and information in the United States and Canada.

## DIVERSITY OF NEW AI PHDS IN NORTH AMERICA

### By Gender

Figure 4.4.6 shows that the share of new female AI and CS PhDs in North America remains low and has changed little from 2010 to 2020.

**FEMALE NEW AI and CS PHDS (% of TOTAL NEW AI and CS PHDS) in NORTH AMERICA, 2010–20**
Source: CRA Taulbee Survey, 2021 | Chart: 2022 AI Index Report

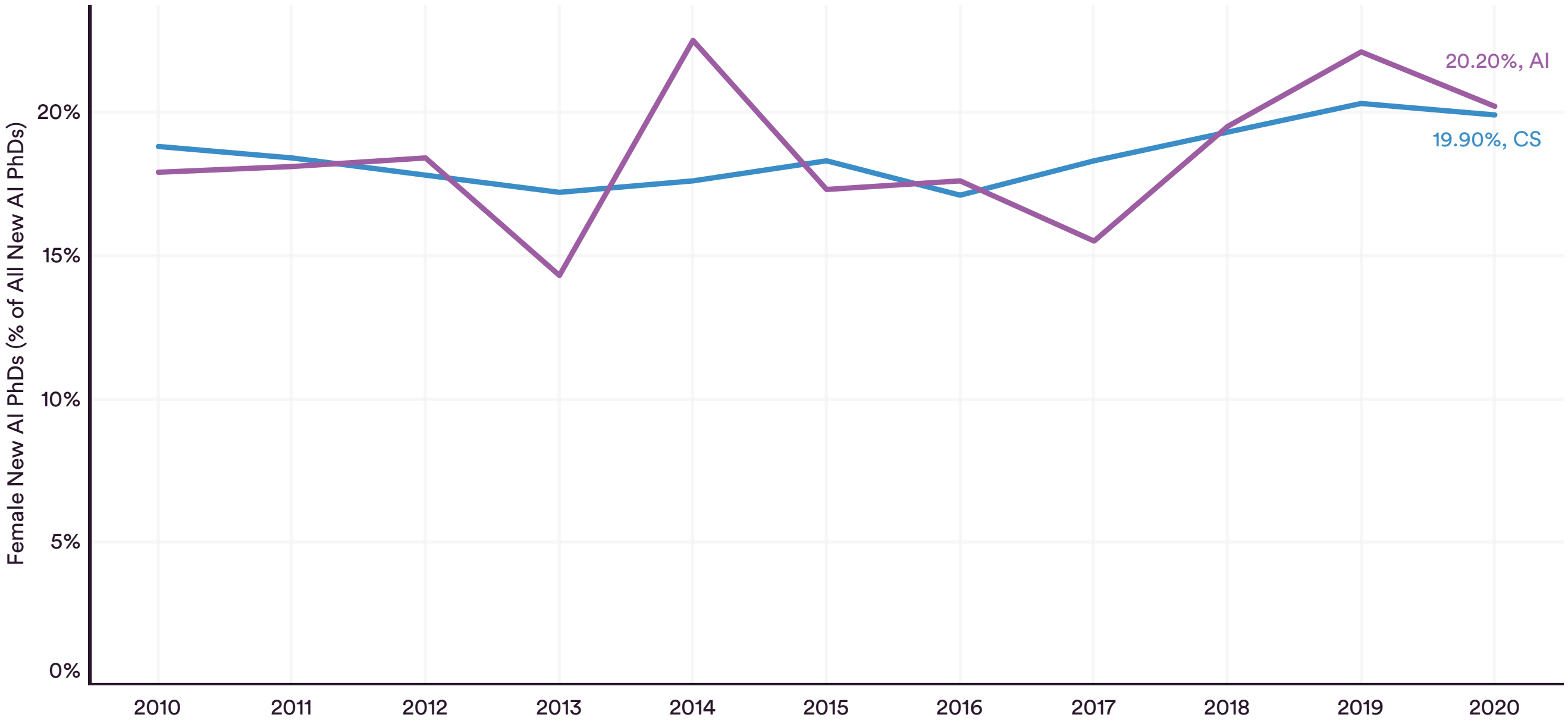

Figure 4.4.6

## By Race/Ethnicity

According to Figure 4.4.7, among the new AI PhDs from 2010 to 2020 who are U.S. residents, the largest percentage has been non-Hispanic white and Asian—65.2% and 18.8% on average. By comparison, around 1.5% were Black or African American (non-Hispanic) and 2.9% were Hispanic on average over the past 11 years. Figure 4.4.8 shows all PhDs awarded in the United States to U.S. residents across departments of CS, CE, and information between 2010 and 2020. In the past 11 years, the share of new white (non-Hispanic) PhDs has changed little, while the percentage of new Black or African American (non-Hispanic) and Hispanic computing PhDs is significantly lower.

**NEW U.S. AI RESIDENT PHDS (% of TOTAL) by RACE/ETHNICITY, 2010–20**
Source: CRA Taulbee Survey, 2021 | Chart: 2022 AI Index Report

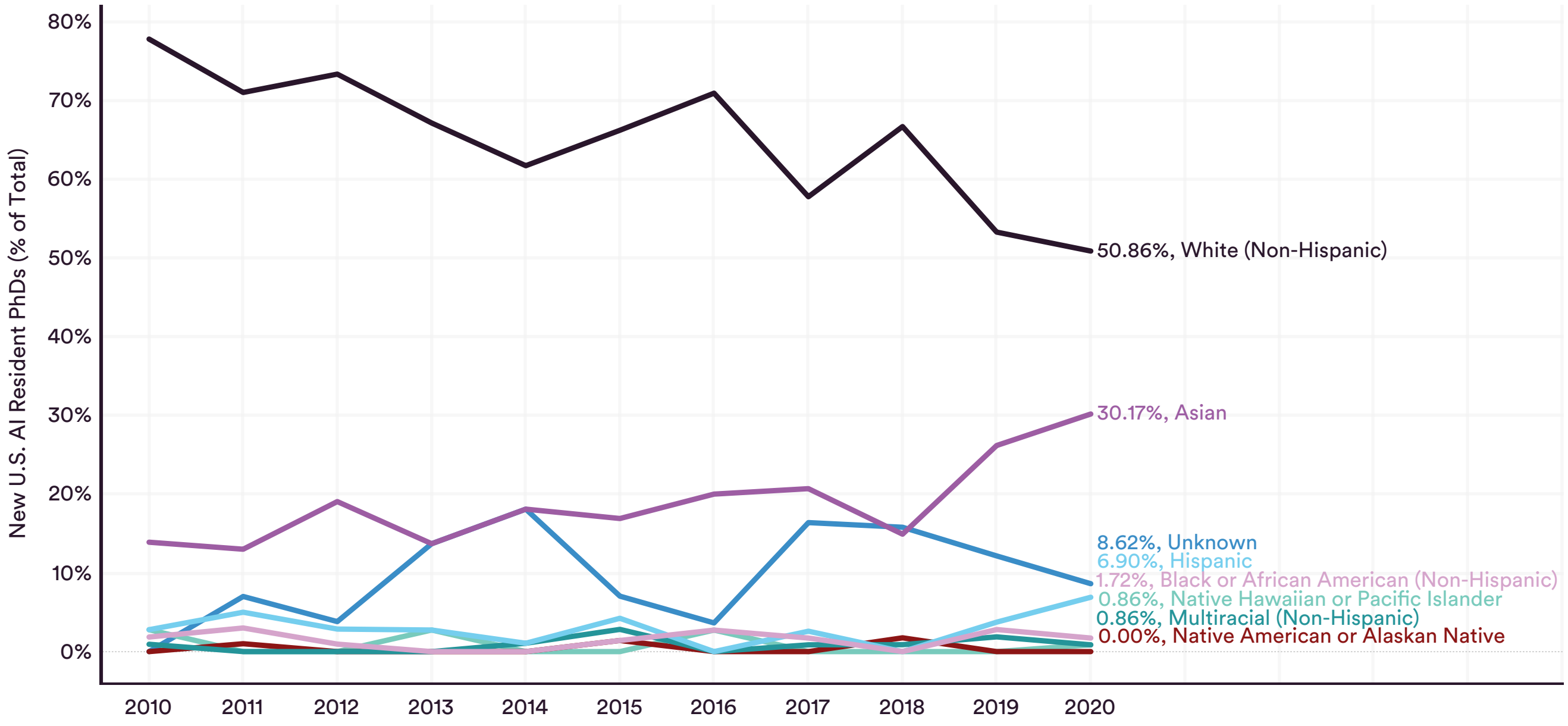

Figure 4.4.7

**NEW COMPUTING PHDS, U.S. RESIDENT (% of TOTAL) by RACE/ETHNICITY, 2010–20**
Source: CRA Taulbee Survey, 2021 | Chart: 2022 AI Index Report

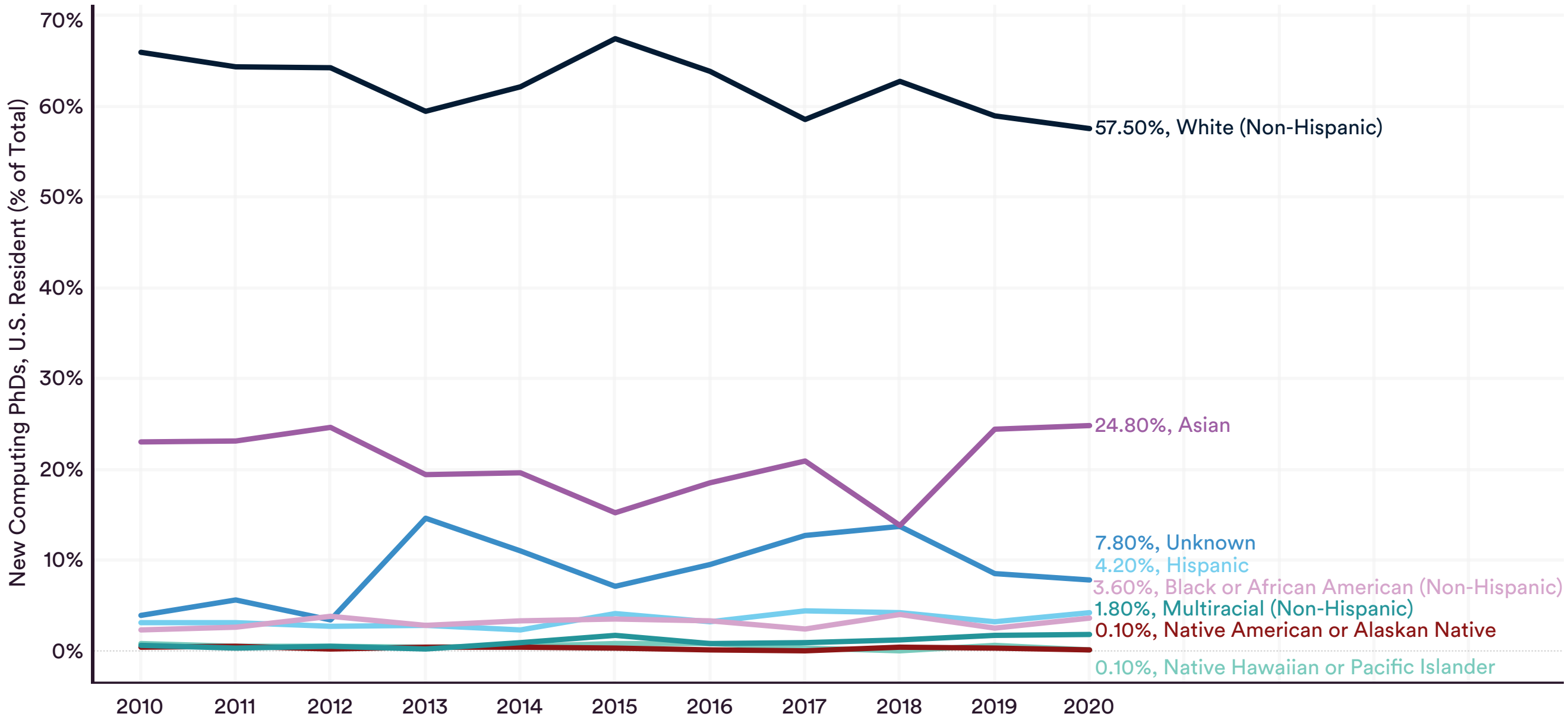

Figure 4.4.8

## NEW INTERNATIONAL AI PHDS IN NORTH AMERICA

The share of international students among new AI PhDs in North America in 2020 decreased slightly from 64.3% in 2019 to 60.5% in 2020 (Figure 4.4.9). For comparison, of all computing PhDs graduating in 2022, 65.1% of them were international students. In addition, more international students—14.0% of all new AI PhDs—took jobs outside the United States in 2020, compared to 8.6% in 2019 (Figure 4.4.10).

**NEW INTERNATIONAL AI PHDS (% of TOTAL NEW AI PHDS) in NORTH AMERICA, 2010–20**
Source: CRA Taulbee Survey, 2021 | Chart: 2022 AI Index Report

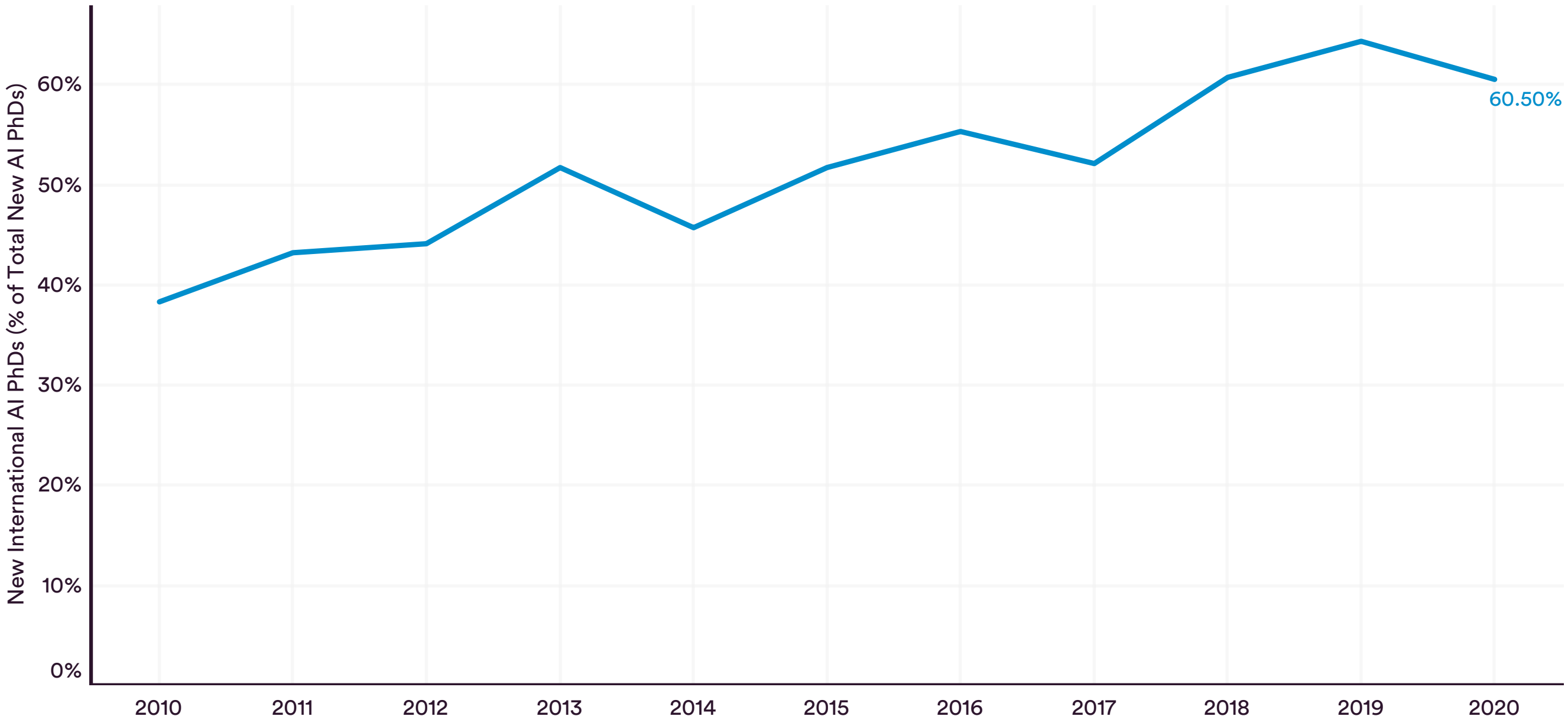

Figure 4.4.9

**INTERNATIONAL NEW AI PHDS (% of TOTAL) in the UNITED STATES by LOCATION of EMPLOYMENT, 2020**
Source: CRA Taulbee Survey, 2021 | Chart: 2022 AI Index Report

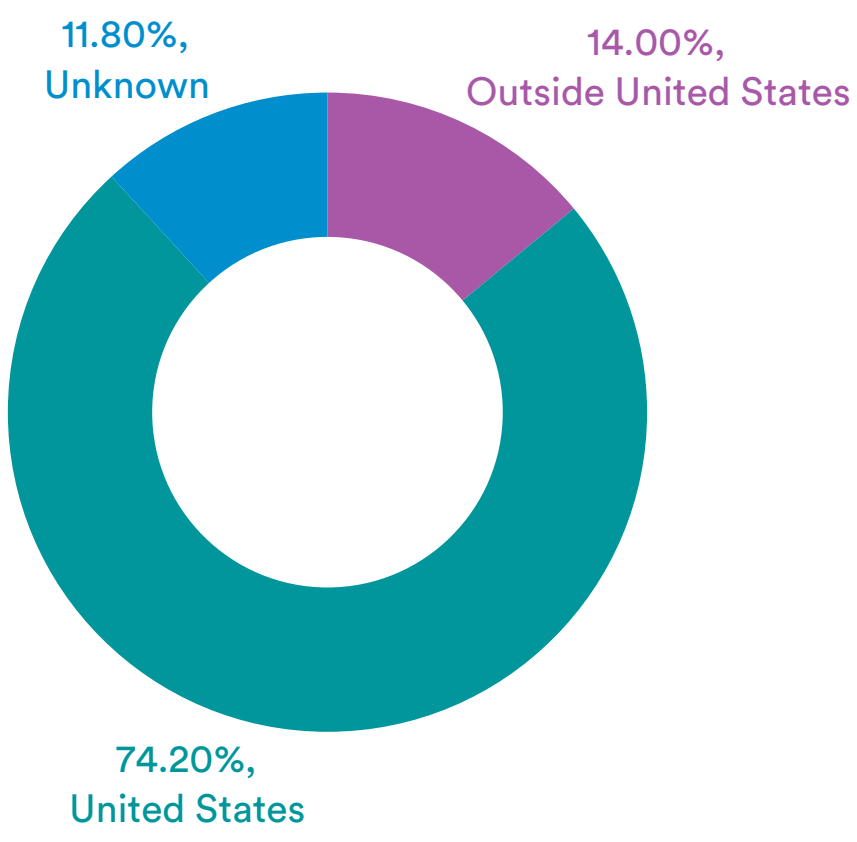

Figure 4.4.10

Artificial Intelligence Index Report 2022

## CHAPTER 5:

# AI Policy and Governance

## CHAPTER 5:
# Chapter Preview

**ACCESS THE PUBLIC DATA**

# Overview

As AI has become an increasingly ubiquitous topic in the last decade, intergovernmental, national, and regional organizations have worked to develop policies and strategies around AI governance. These actors are driven by the understanding that it is imperative to find ways to address the ethical and societal concerns surrounding AI, while maximizing its benefits. Active and informed governance of AI technologies has become a priority for many governments around the world.

This chapter examines the intersection of AI and governance, and takes a closer look at how governments in different countries, regions, and U.S. states are working to manage AI technologies. It begins by looking at AI policymaking across the globe and within the United States, exploring which countries and political actors are most keen to advance AI legislation, and what kind of AI subtopics, from privacy to ethics, are the focus of most legislative attention. Then the chapter takes a deep dive into one of the world's top public sector investors in AI, the United States, and studies how much its various government departments have spent on AI in the past five years.

## CHAPTER HIGHLIGHTS

- An AI Index analysis of legislative records on AI in 25 countries shows that the number of bills containing "artificial intelligence" that were **passed into law grew from just 1 in 2016 to 18 in 2021.** Spain, the United Kingdom, and the United States passed the highest number of AI-related bills in 2021, with each adopting three.

- The federal legislative record in the United States shows a sharp increase in the total number of proposed bills that relate to AI from 2015 to 2021, **while the number of bills passed remains low, with only 2% ultimately becoming law.**

- State legislators in the United States **passed 1 out of every 50 proposed bills** that contain AI provisions in 2021, while the number of such bills proposed **grew from 2 in 2012 to 131 in 2021**.

- In the United States, the current congressional session (the 117th) is on track to record the greatest number of AI-related mentions since 2001, **with 295 mentions by the end of 2021, half way through the session, compared to 506 in the previous (116th) session**.

**Discussions around AI governance regulation have accelerated over the past decade, resulting in policy proposals across various legislative bodies. This section first examines AI-related legislation that has either been proposed or passed into law across different countries and regions, followed by a focused analysis of state-level legislation in the United States. It then takes a closer look at congressional and parliamentary records on AI across the world and concludes with data on the number of policy papers published in the United States.**

# 5.1 AI AND POLICYMAKING

## GLOBAL LEGISLATION RECORDS ON AI

Governments and legislative bodies across the globe are increasingly seeking to pass laws to provide funding for AI development and innovation, while also promoting the integration of human-centered values. The AI Index has conducted an analysis of laws passed in 25 countries by their legislative bodies that contain the words "artificial intelligence" from 2016 to 2021.

Taken together, the 25 countries analyzed have passed a total of 55 AI-related bills. Figure 5.2.1 demonstrates that in the past six years, there has been a sharp increase in terms of the total number of AI-related bills passed into law.[1]

**NUMBER of AI-RELATED BILLS PASSED into LAW in 25 SELECT COUNTRIES, 2016–21**
Source: AI Index, 2021 | Chart: 2022 AI Index Report

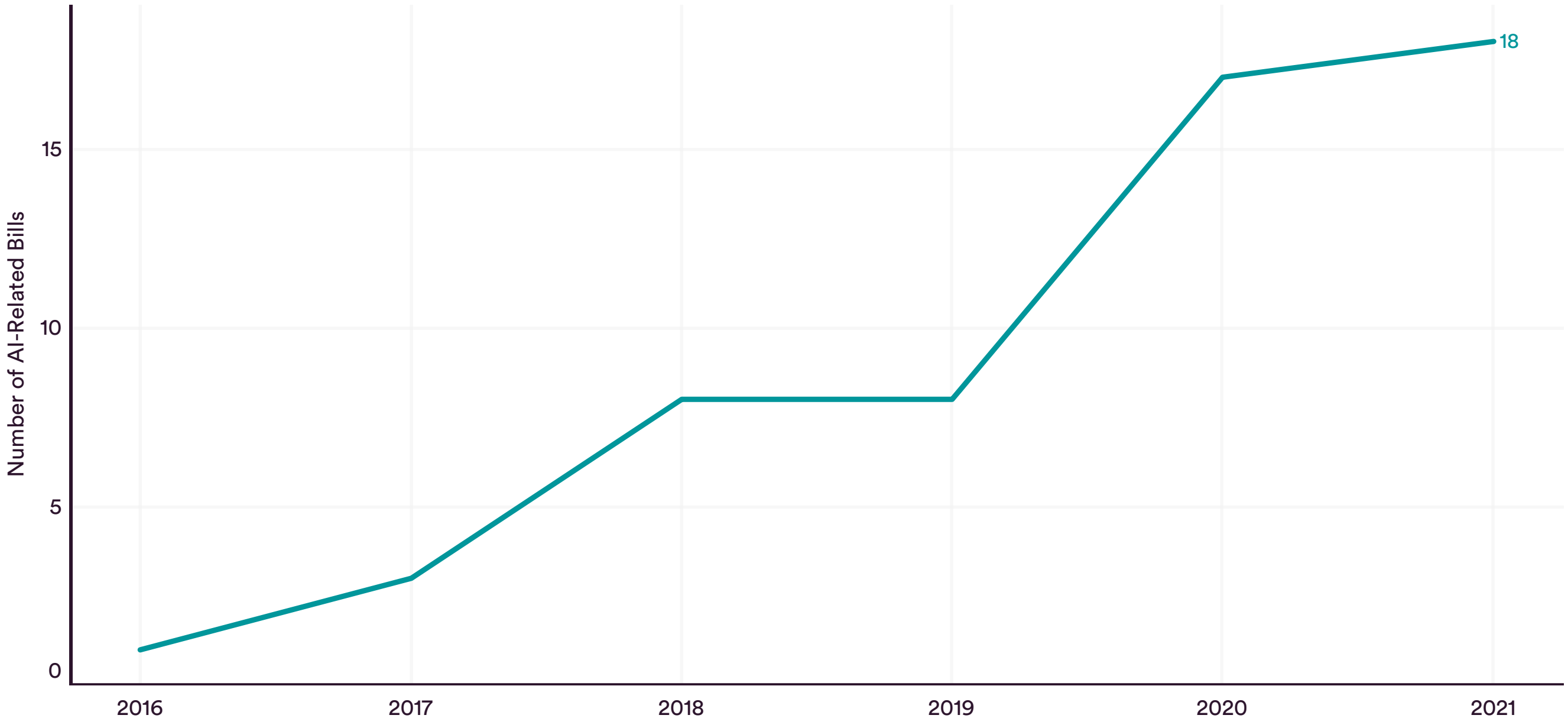

Figure 5.1.1

1 Note that the analysis only includes laws passed by national legislative bodies (e.g. congress, parliament) with the keyword "artificial intelligence" in various languages in the title or body of the bill text. See the appendix for the methodology. Countries included: Australia, Belgium, Brazil, Canada, China, Denmark, Finland, France, Germany, India, Ireland, Italy, Japan, the Netherlands, New Zealand, Norway, Russia, Singapore, South Africa, South Korea, Spain, Sweden, Switzerland, the United Kingdom, and the United States.

## By Geographic Area

Figure 5.1.2a shows the number of laws containing mentions of AI that were enacted in 2021. Spain, the United Kingdom, and the United States led, each passing three. Figure 5.1.2b shows the total number of legislation passed in the past six years. The United States dominated the list with 13 bills, starting in 2017 with 3 new laws passed each subsequent year, followed by Russia, Belgium, Spain, and the United Kingdom.

**NUMBER of AI-RELATED BILLS PASSED into LAW in SELECT COUNTRIES, 2021**
Source: AI Index, 2021 | Chart: 2022 AI Index Report

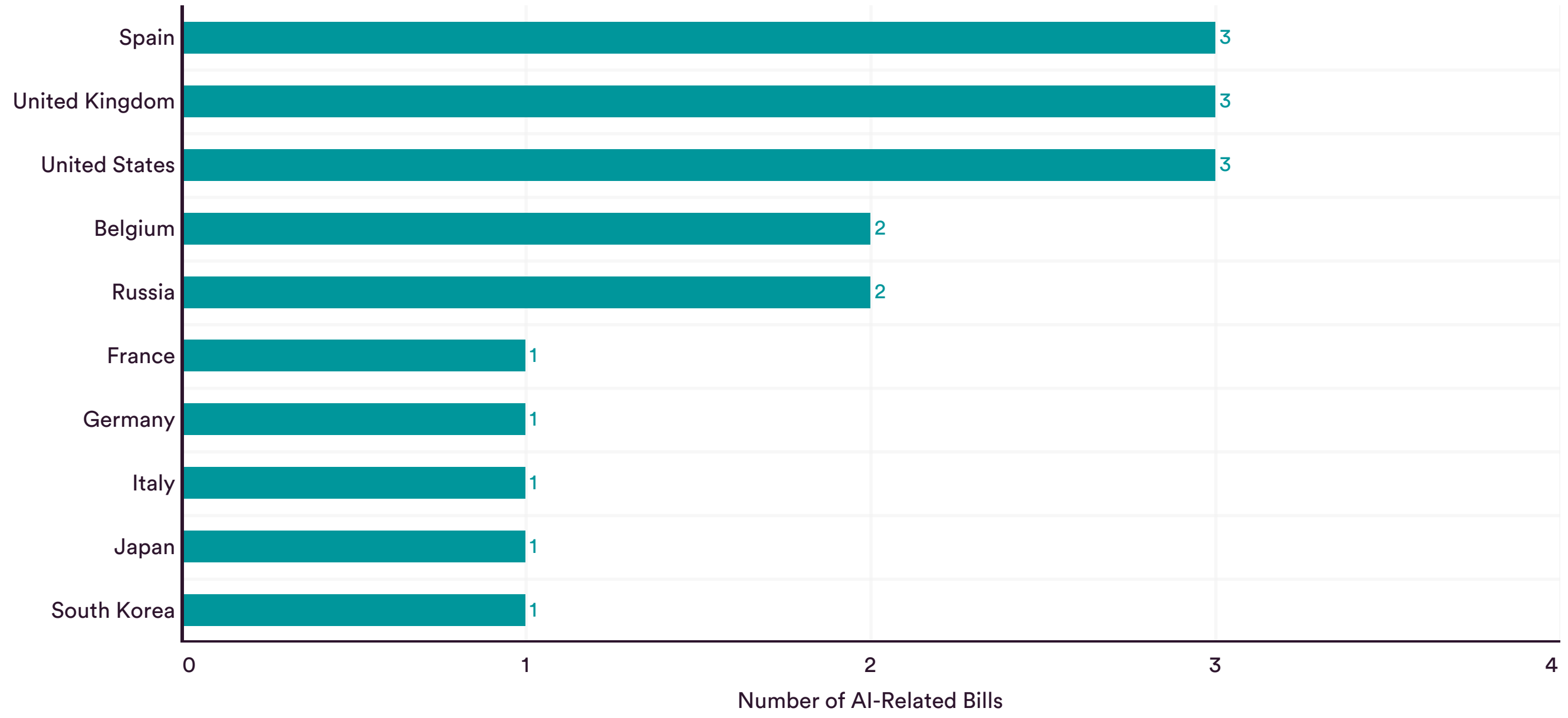

Figure 5.1.2a

**NUMBER of AI-RELATED BILLS PASSED into LAW in SELECT COUNTRIES, 2016–21 (SUM)**
Source: AI Index, 2021 | Chart: 2022 AI Index Report

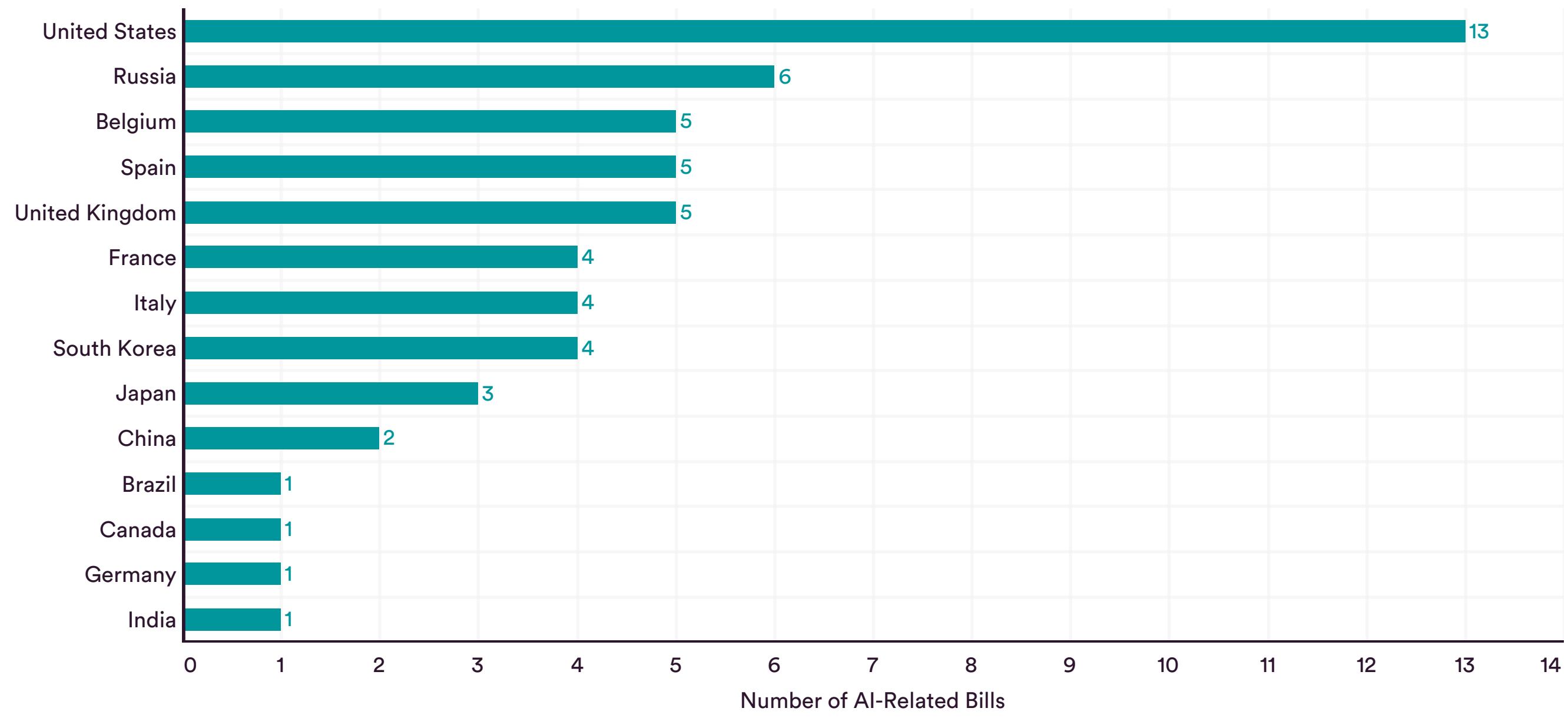

Figure 5.1.2b

## Federal AI Legislation in the United States

A closer look at the federal legislative record in the United States shows a sharp increase in the total number of proposed bills that relate to AI (Figure 5.1.3). In 2015, just one federal bill was proposed, while in 2021, there were 130. Although this jump is significant, the number of bills related to AI being passed has not kept pace with the growing volume of proposed AI-related bills. This gap was most evident in 2021, when only 2% of all federal-level AI-related bills were ultimately passed into law.

**NUMBER of AI-RELATED BILLS in the UNITED STATES, 2015–21 (PROPOSED vs. PASSED)**
Source: AI Index, 2021 | Chart: 2022 AI Index Report

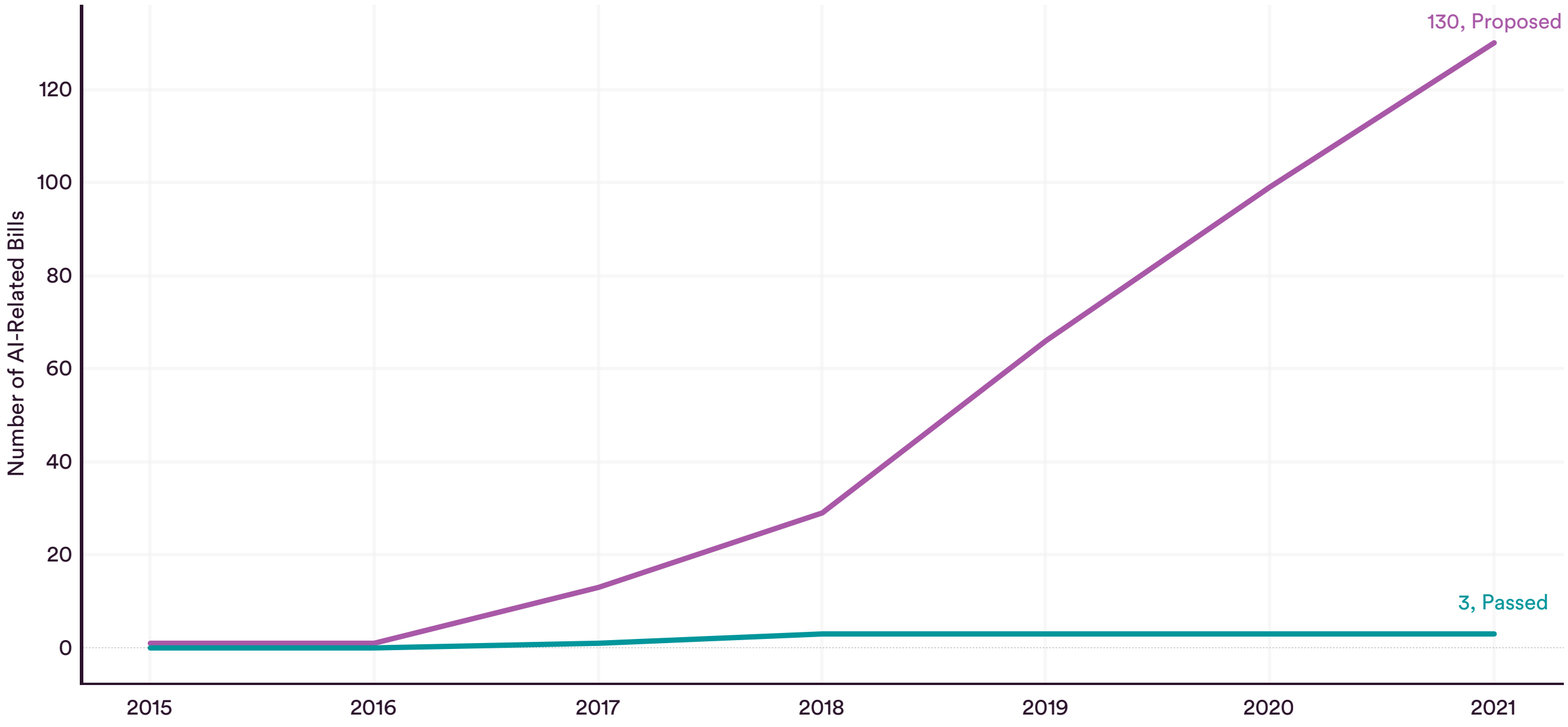

Figure 5.1.3

# A Closer Look at the Legislation

The following subsection delves into some of the AI-related legislation passed into law since 2016. Table 5.1.1 demonstrates the wide range of AI-related issues that have piqued policymakers' interest.

| Country | Year Passed | Bill Name | Description |
|---|---|---|---|
| Canada | 2017 | Budget Implementation Act 2017, No. 1 | A provision of this act authorized the Canadian government to make a payment of $125 million to the Canadian Institute for Advanced Research to support the development of a pan-Canadian artificial intelligence strategy. |
| China | 2019 | Law of the People's Republic of China on Basic Medical and Health Care and the Promotion of Health | A provision of this law aimed to promote the application and development of big data and artificial intelligence in the health and medical field while accelerating the construction of medical and healthcare information infrastructure, developing technical standards on the collection, storage, analysis, and application of medical and health data. |
| Russia | 2020 | Federal Law of 24 April 2020 No. 123-FZ on the Experiment to Establish Special Regulation in order to Create the Necessary Conditions for the Development and Implementation of Artificial Intelligence Technologies in the Region of the Russian Federation – Federal City of Moscow and Amending the Articles 6 and 10 of the Federal Law on Personal Data | This law established an experimental framework for the development and implementation of AI as a five-year experiment to start in Moscow in July 1, 2020, including allowing AI systems to process anonymized personal data for governmental and certain commercial business activities. |
| United Kingdom | 2020 | Supply and Appropriation (Main Estimates) Act 2020, c.13 | A provision of this act authorized the Office of Qualifications and Examination Regulation to explore opportunities for using artificial intelligence to improve the marking and administration of high-stakes qualifications. |
| United States | 2020 | IOGAN ACT: Identifying Outputs of Generative Adversarial Networks Act | This act directed the National Science Foundation to support research dedicated to studying the outputs of generative adversarial networks (deepfakes) and other comparable technologies. |
| Belgium | 2021 | Decree on coaching and solution-oriented support for job seekers, N. 327 | A provision of this act directs the government to create an advisory group called the Ethics Committee, which is responsible for submitting advice if artificial intelligence tools are to be used for digitization activities. |
| France | 2021 | Law N:2021-1485 of November 15, 2021, aimed at reducing the environmental footprint of digital technology in France | This act sets up a monitoring system to evaluate environmental impacts of newly emerging digital technologies, in particular, artificial intelligence. |

Table 5.1.1

## STATE-LEVEL AI LEGISLATION IN THE UNITED STATES

Growing policy interest in AI can also be seen in the large number of AI-related bills recently proposed at the state level in the United States, based on data provided by Bloomberg Government since 2012. Bloomberg Government classified a bill as relating to AI if it contained AI-related keywords such as artificial intelligence, machine learning, or algorithmic bias.

As is the case on the federal level, there has been a significant increase in the number of AI bills proposed at the state level in the last decade (Figure 5.1.4). In 2012, the first two pieces of AI-related legislation were proposed when New Jersey assembly member Annette Quijano directed the New Jersey Motor Vehicle Commission to establish driver's license endorsements for autonomous vehicles. In the past 10 years, the increase has been substantial, from 2 bills in 2012 to 131 in 2021.

A notable difference between AI-related lawmaking in the United States on the federal versus the state level is that a greater proportion of proposed state-level AI bills have actually passed. In 2021, of the 131 proposed state bills, 26 were passed into law (20%), or 1 out of 5 proposed bills became law. This ratio is significantly higher when compared to the federal level, where 1 out of every 50 proposed bills became law in 2021.

**NUMBER of STATE-LEVEL AI-RELATED BILLS in the UNITED STATES, 2012–21**
Source: Bloomberg Government, 2021 | Chart: 2022 AI Index Report

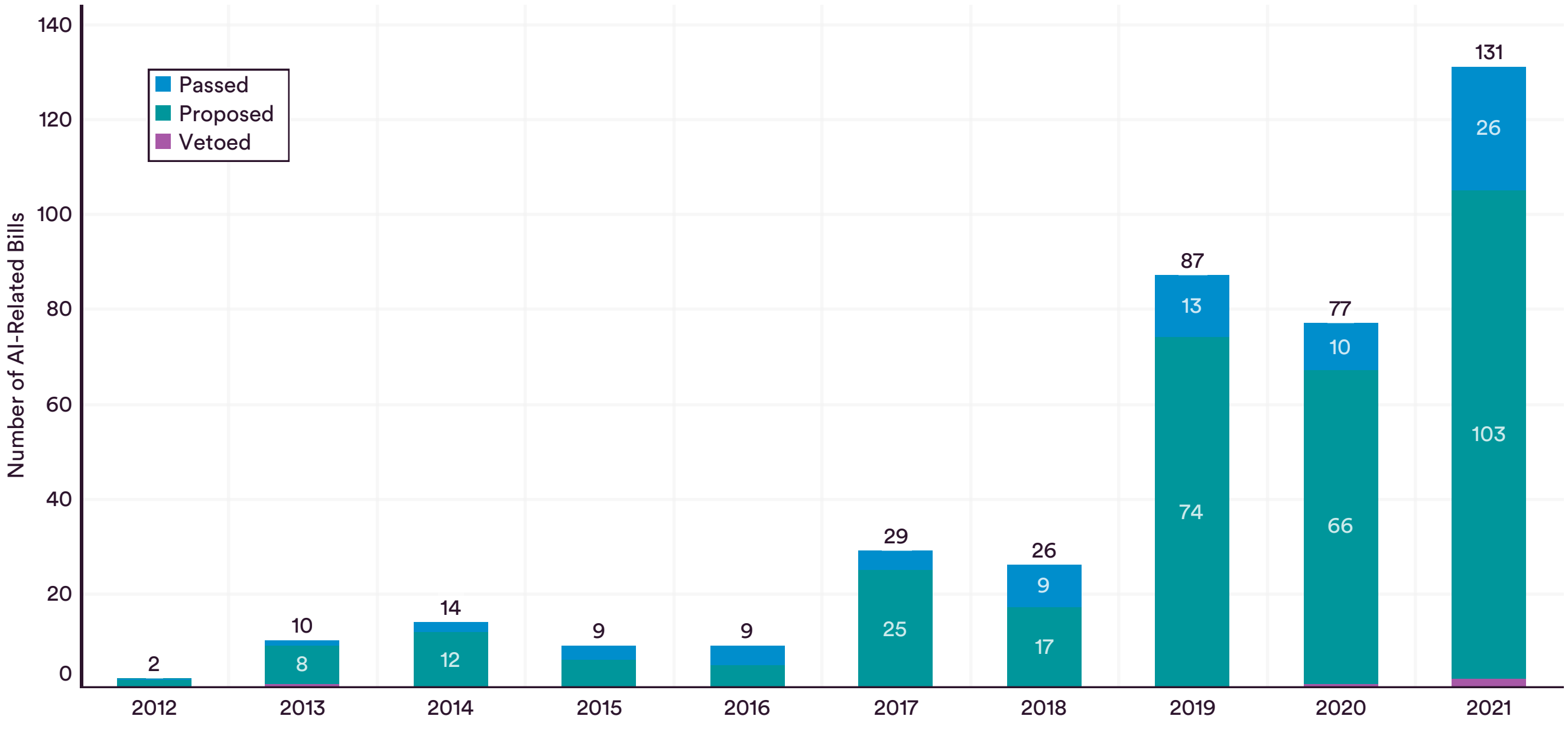

Figure 5.1.4

## By State

In the United States, AI lawmaking has been relatively widespread across all states. As of 2021, 41 out of 50 states have proposed at least one AI-related bill, but certain states have been particularly active in generating AI legislation. Figure 5.1.5 shows that Massachusetts has proposed the most AI bills, with 40 since 2012, followed by Hawaii (35) and New Jersey (32). Focusing on just 2021 in Figure 5.1.6, Massachusetts was the state that proposed the most AI-related bills, with 20, followed by Illinois (15) and Alabama (12).

**NUMBER of STATE-LEVEL PROPOSED AI-RELATED BILLS in the UNITED STATES by STATE, 2012–21 (SUM)**
Source: Bloomberg Government, 2021 | Chart: 2022 AI Index Report

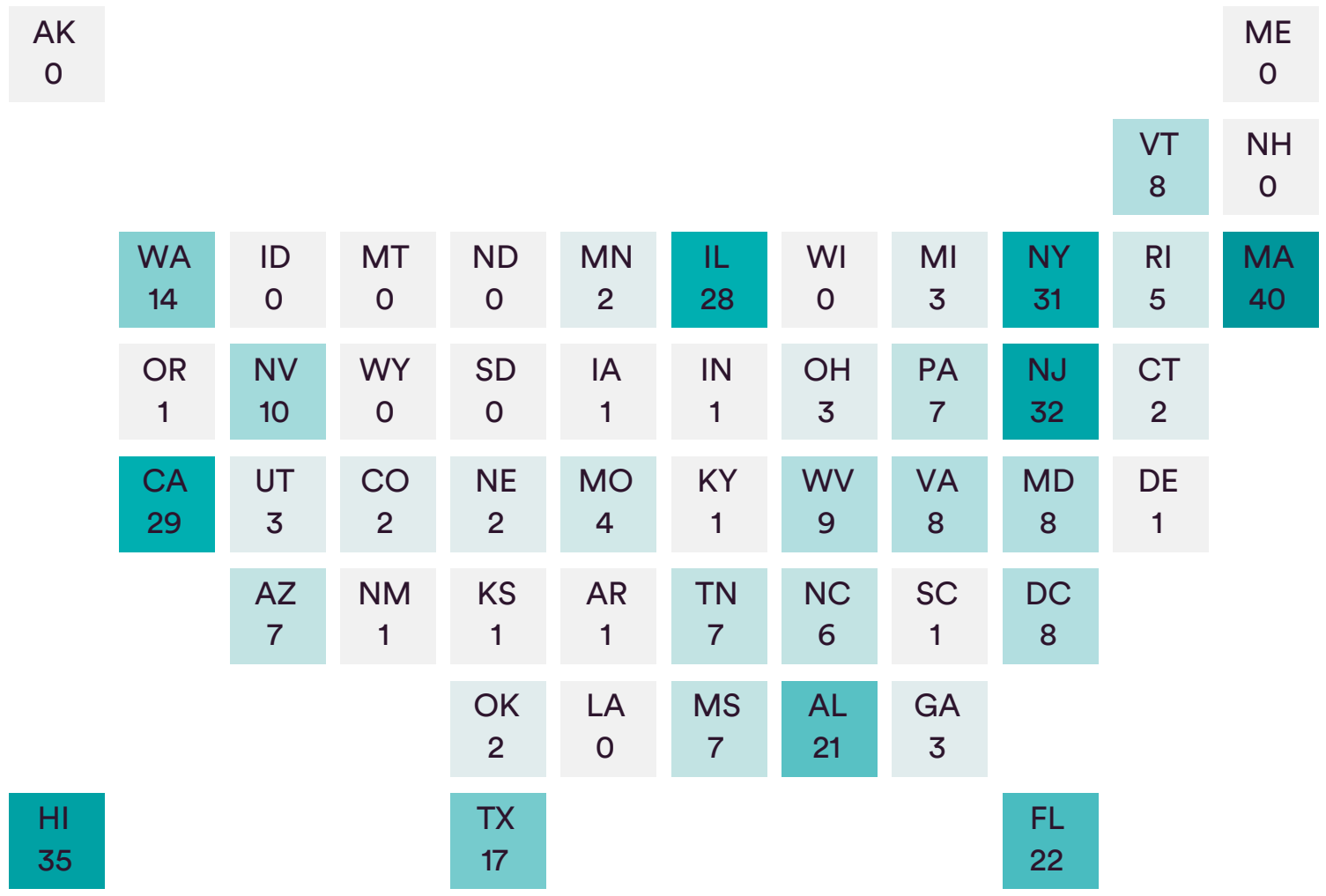

Figure 5.1.5

**NUMBER of STATE-LEVEL PROPOSED AI-RELATED BILLS in the UNITED STATES by STATE, 2021**
Source: Bloomberg Government, 2021 | Chart: 2022 AI Index Report

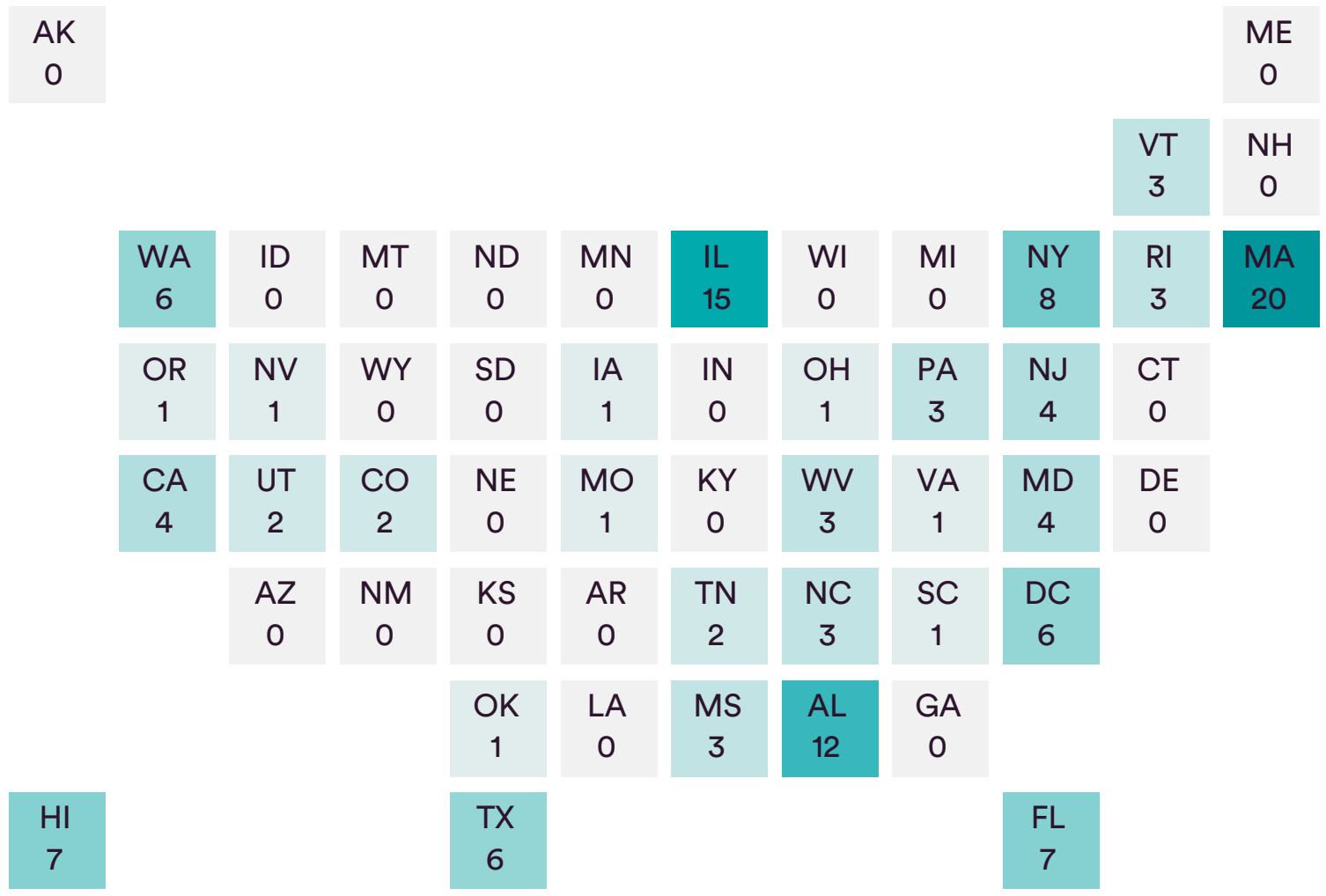

Figure 5.1.6

## Sponsorship by Political Party

State-level AI legislation data reveals that there is a partisan dynamic to AI lawmaking. Figure 5.1.7 plots the number of AI-related bills sponsored at the state level by Democratic and Republican lawmakers. Although there has been an increase in AI bills proposed by members of both parties since 2012, in the past four years, the data suggests Democrats were more likely to sponsor AI-related legislation. Whereas Democrats sponsored only two more AI bills than Republicans in 2018, they sponsored 39 more in 2021.

**NUMBER of STATE-LEVEL PROPOSED AI-RELATED BILLS in the UNITED STATES by SPONSOR PARTY, 2012–21**

Source: Bloomberg Government, 2021 | Chart: 2022 AI Index Report

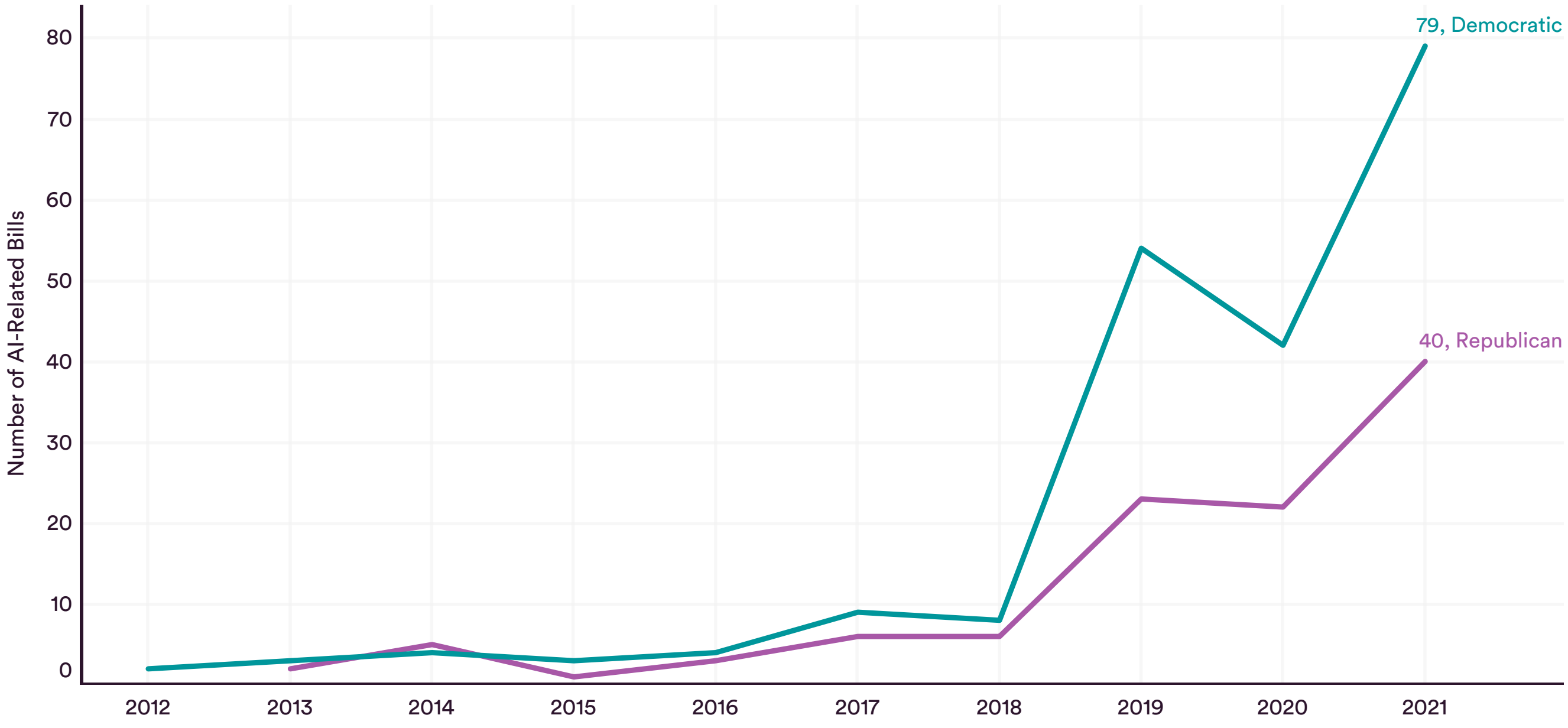

Figure 5.1.7

## MENTIONS OF AI IN LEGISLATIVE RECORDS

Another barometer of legislative interest in AI is the number of mentions of "artificial intelligence" in governmental and parliamentary proceedings. This subsection considers data on mentions of AI both in U.S. congressional records and the parliamentary proceedings of other countries based on AI Index and Bloomberg Government data.

### AI Mentions in U.S. Congressional Records

In the last five years, and especially in 2021, U.S. congressional sessions have devoted increasing amounts of time to discussions of AI. This section presents data from Bloomberg Government concerning mentions of AI-related keywords in congressional proceedings, broken down by legislation, congressional committee reports, and congressional research service reports.

According to Figure 5.1.8, the current congressional session (the 117th) is on track (as of the end of 2021) to record the greatest number of AI-related mentions since 2001. The most recently completed congressional session, the 116th (2019-2020), saw 506 AI mentions, nearly 3.4 times as many mentions as there were during the 115th session (2017–2018), and 30 times as many as the 114th session (2015–2016).

**MENTIONS of AI in the U.S. CONGRESSIONAL RECORD by LEGISLATIVE SESSION, 2001–21**
Source: Bloomberg Government, 2021 | Chart: 2022 AI Index Report

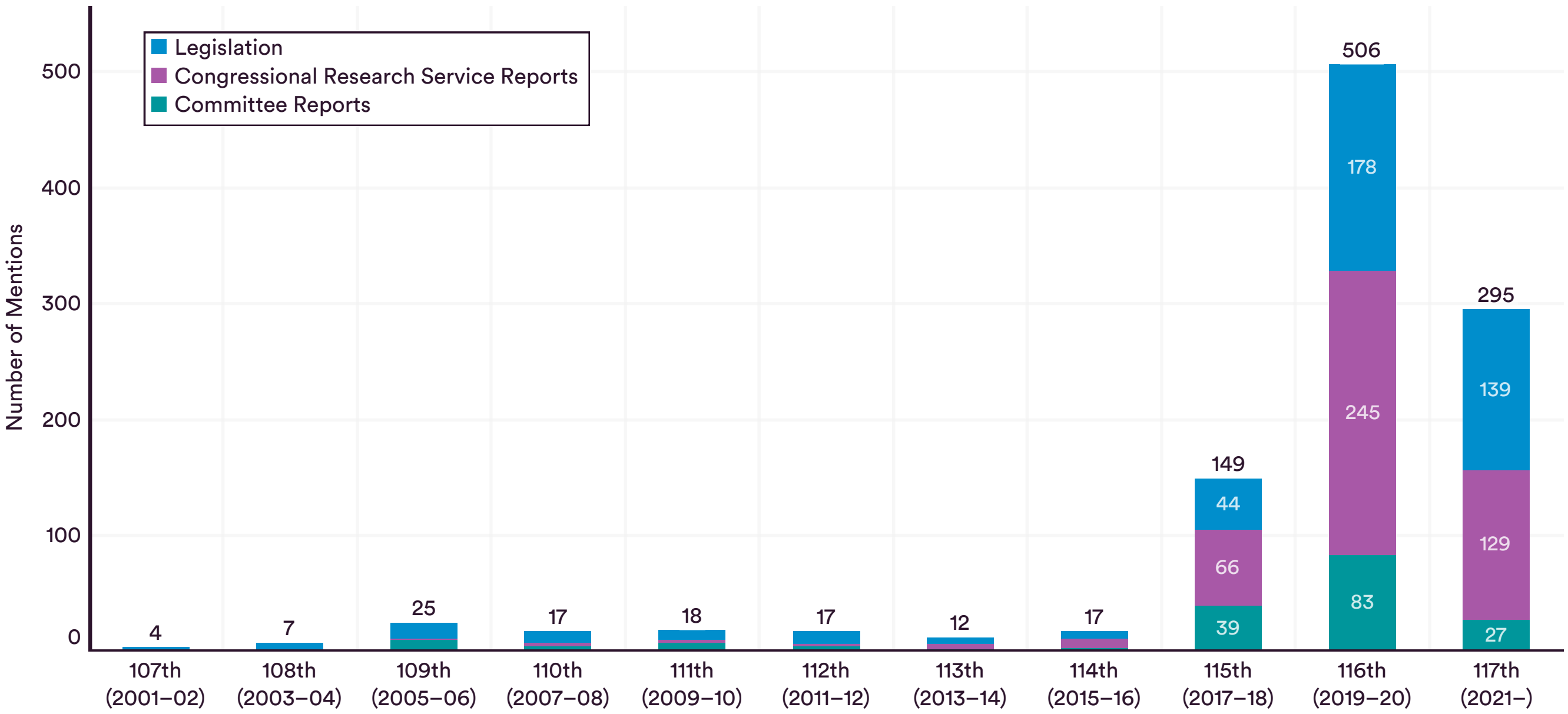

Figure 5.1.8

## AI Mentions in Global Legislative Proceedings

AI mentions in governmental proceedings are on the rise not only in the United States but also in many other countries across the world. The AI Index conducted an analysis on the minutes or proceedings of legislative sessions in 25 countries that contain the keyword "artificial intelligence" from 2016 to 2021. Figure 5.1.9 shows that the mentions of AI in legislative proceedings in 25 select countries grew 7.7 times in the past six years.[2]

**NUMBER of MENTIONS of AI in LEGISLATIVE PROCEEDINGS in 25 SELECT COUNTRIES, 2016–21**
Source: AI Index, 2021 | Chart: 2022 AI Index Report

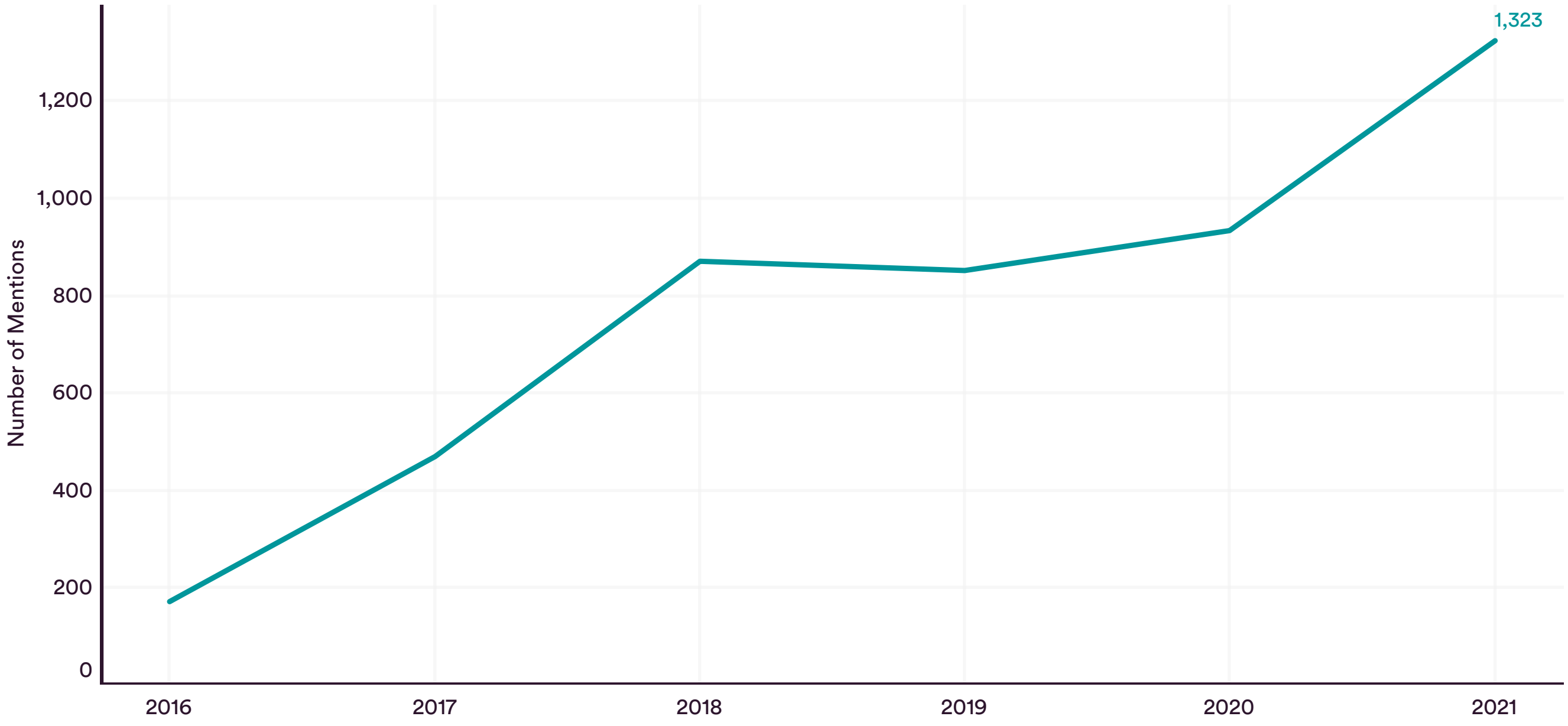

Figure 5.1.9

2 See the appendix for the methodology. Countries included: Australia, Belgium, Brazil, Canada, China, Denmark, Finland, France, Germany, India, Ireland, Italy, Japan, the Netherlands, New Zealand, Norway, Russia, Singapore, South Africa, South Korea, Spain, Sweden, Switzerland, the United Kingdom, and the United States.

## By Geographic Area

Figure 5.1.10a shows the number of legislative proceedings containing mentions of AI that were enacted in 2021. Similar to the trend in the number of AI mentions in bills passed into laws, Spain, the United Kingdom, and the United States topped the list. Figure 5.1.2b shows the total number of AI mentions in the past six years. The United Kingdom dominated the list with 939 mentions, followed by Spain, Japan, the United States, and Australia.

**NUMBER of MENTIONS of AI in LEGISLATIVE PROCEEDINGS in SELECT COUNTRIES, 2021**

Source: AI Index, 2021 | Chart: 2022 AI Index Report

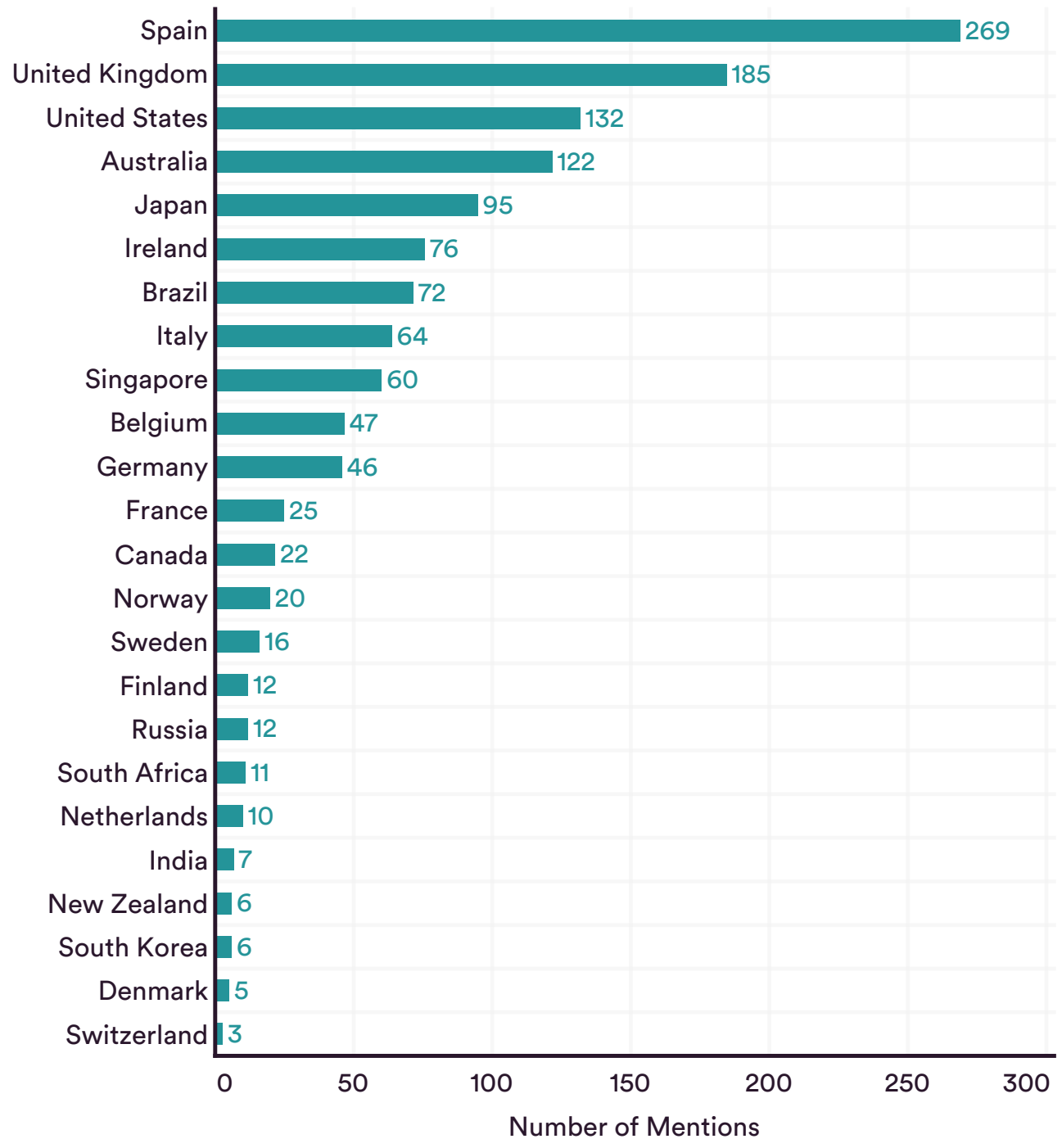

Figure 5.1.10a

**NUMBER of MENTIONS of AI in LEGISLATIVE PROCEEDINGS in SELECT COUNTRIES, 2016–2021 (SUM)**

Source: AI Index, 2021 | Chart: 2022 AI Index Report

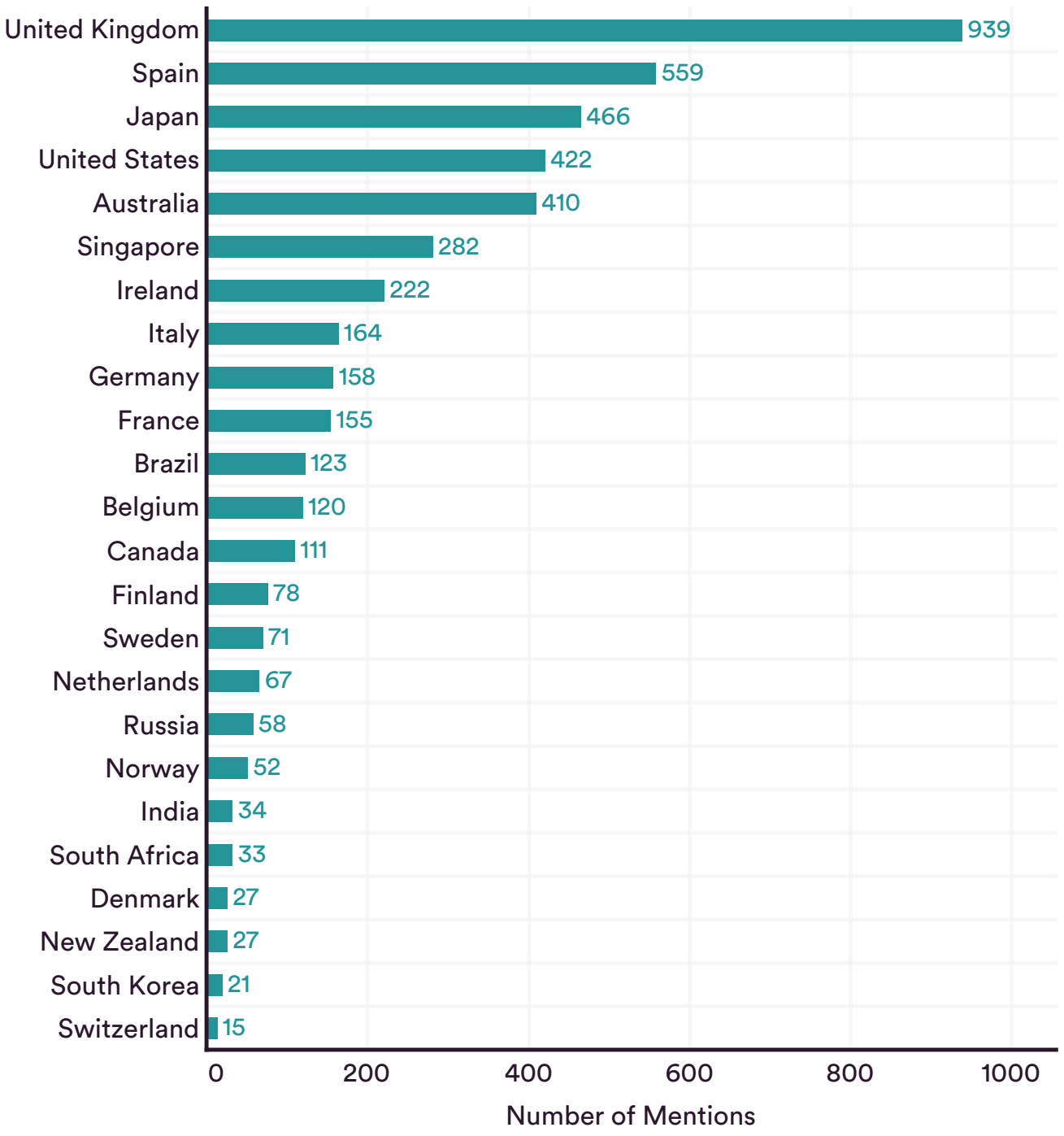

Figure 5.1.10b

## U.S. AI POLICY PAPERS

To estimate activities outside national governments that are also informing AI-related rulemaking, the AI Index tracks 55 U.S.-based organizations that published policy papers in the past four years. Those organizations include: think tanks and policy institutes (19); university institutes and research programs (14); civil society organizations, associations, and consortiums (9); industry and consultancy organizations (9); and government agencies (4).[3] A policy paper in this section is defined as a research paper, research report, brief, or blog post that addresses issues related to AI and makes specific recommendations to policymakers. Topics of those papers are divided into primary and secondary categories: A primary topic is the main focus of the paper, while a secondary topic is a subtopic of the paper or an issue that was briefly explored.

Figure 5.1.11 plots the total number of U.S.-based AI-related policy papers that have been published from 2018 to 2021, which can proxy the general interest in AI within the U.S. policymaking space. The total number of policy papers has tripled since 2018, peaking in 2020 with 273, and decreasing slightly in 2021, with 210.

**NUMBER of AI-RELATED POLICY PAPERS by U.S.-BASED ORGANIZATIONS, 2018–21**
Source: AI Index, 2021 | Chart: 2022 AI Index Report

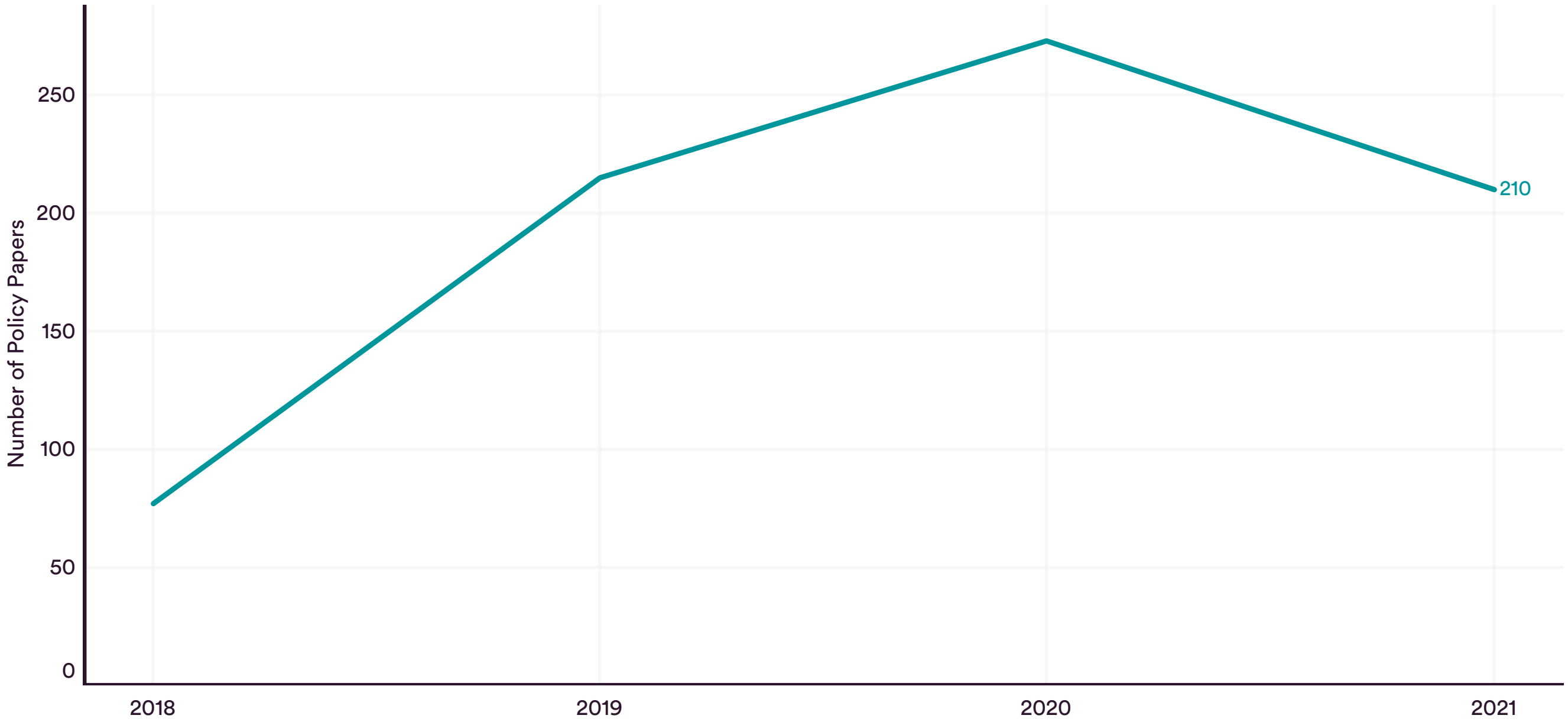

Figure 5.1.11

3 The complete list of organizations the Index followed can be found in the Appendix.

## By Topic

In 2021, the leading primary topics were Privacy, Safety, and Security; Innovation and Technology; and Ethics (Figure 5.1.12). Certain topics, such as government and public administration, education and skills, as well as democracy, did not feature prominently as primary topics, but they were reported on more frequently as secondary topics. Among the AI topics to receive comparatively little attention from tracked organizations are those that relate to energy and the environment, humanities, physical sciences, and social and behavioral sciences.

NUMBER of AI-RELATED POLICY PAPERS by U.S.-BASED ORGANIZATIONS by TOPIC, 2021

Source: AI Index, 2021 | Chart: 2022 AI Index Report

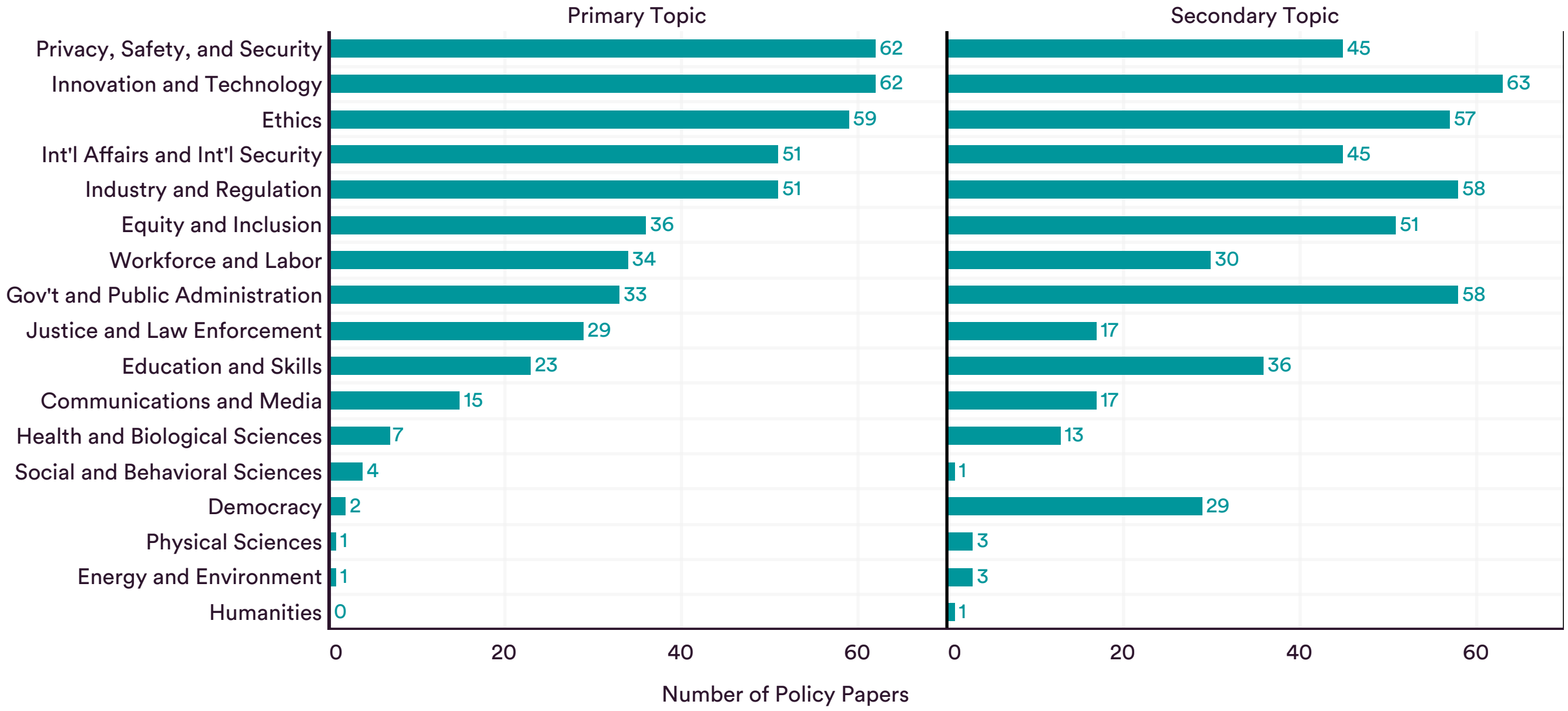

Figure 5.1.12

This section examines the public AI investment in the United States, based on data from the U.S. government and Bloomberg Government.

# 5.2 U.S. PUBLIC INVESTMENT IN AI

## FEDERAL BUDGET FOR NONDEFENSE AI R&D

In December 2021, the National Science and Technology Council published a report on the public-sector AI R&D budget across departments and agencies participating in the Networking and Information Technology Research and Development (NITRD) program and the National Artificial Intelligence Initiative. The report does not include information on classified AI R&D investment by the defense and intelligence agencies.

In fiscal year (FY) 2021, nondefense U.S. government agencies allocated a total of $1.53 billion to AI R&D spending, approximately 2.7 times what was spent in FY 2018 (Figure 5.2.1). This figure is projected to rise 8.8% for FY 2022, with a total of $1.67 billion requested.[4] The increasing amount spent on AI R&D by nondefense departments indicates the U.S. government's continued strong interest in public sector funding for AI research and development spanning a wide range of federal agencies.

U.S. FEDERAL BUDGET for NONDEFENSE AI R&D, FY 2018–22
Source: U.S. NITRD Program, 2022 | Chart: 2022 AI Index Report

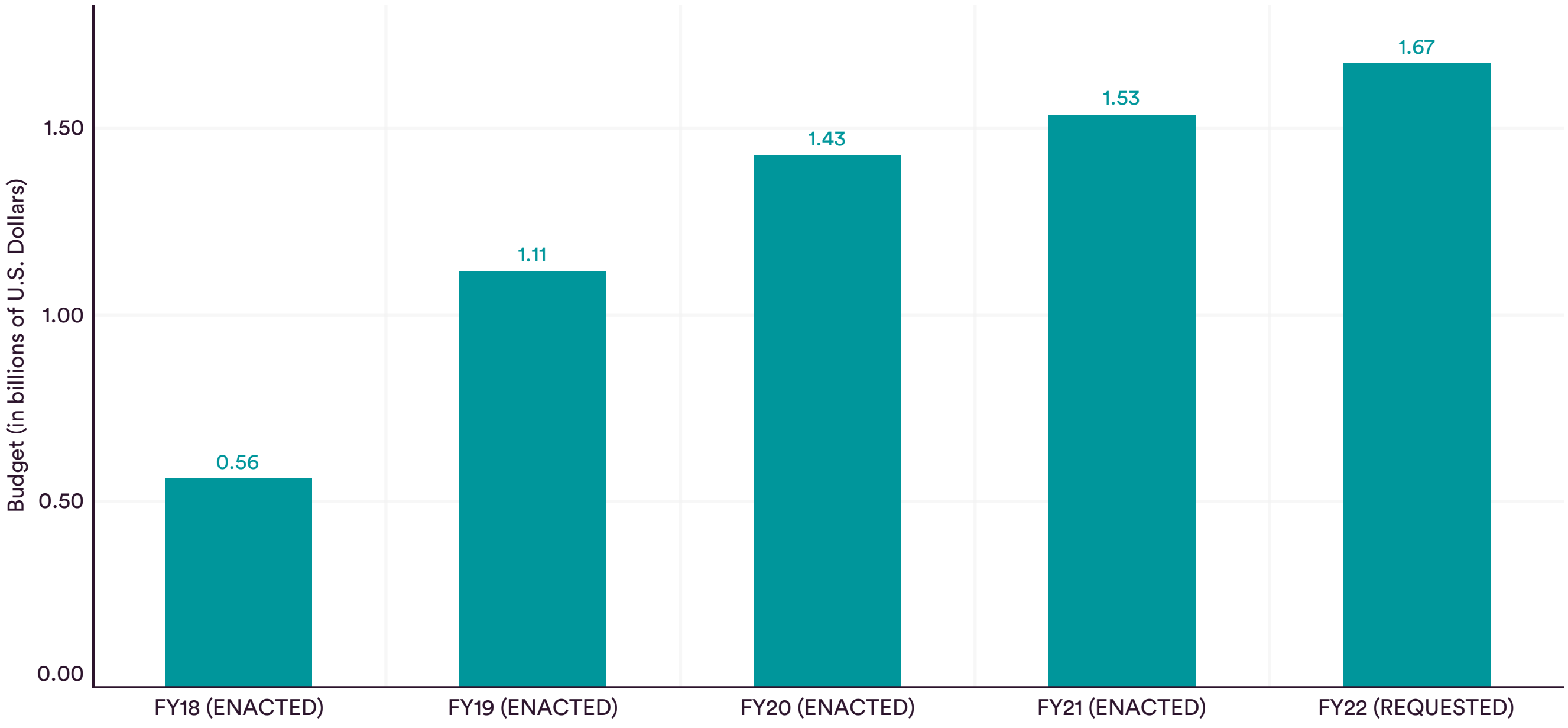

Figure 5.2.1

4 See NITRD website for details on AI R&D investment FY 2018-22 with the breakdown of core AI vs AI crosscut. Note that AI crosscutting budget data is not available for FY 2018.

## U.S. DEPARTMENT OF DEFENSE BUDGET REQUEST

Spending on AI by the U.S. Department of Defense (DOD) can be proxied by looking at the publicly available requests made by the DOD for research, development, test, and evaluation (RDT&E) relating to AI. In FY 2021, DOD allocated $9.26 billion across 500 AI R&D programs (Figure 5.2.2), a 6.68% increase from the $8.68 billion spent in 2020. For FY 2022, the department has requested $10 billion so far, which is likely to grow once additional requests and congressional appropriations are taken into account.

U.S. DOD BUDGET for AI-SPECIFIC RESEARCH, DEVELOPMENT, TEST and EVALUATION (RDT&E), FY 2020–22
Source: Bloomberg Government and U.S. Department of Defense, 2021 | Chart: 2022 AI Index Report

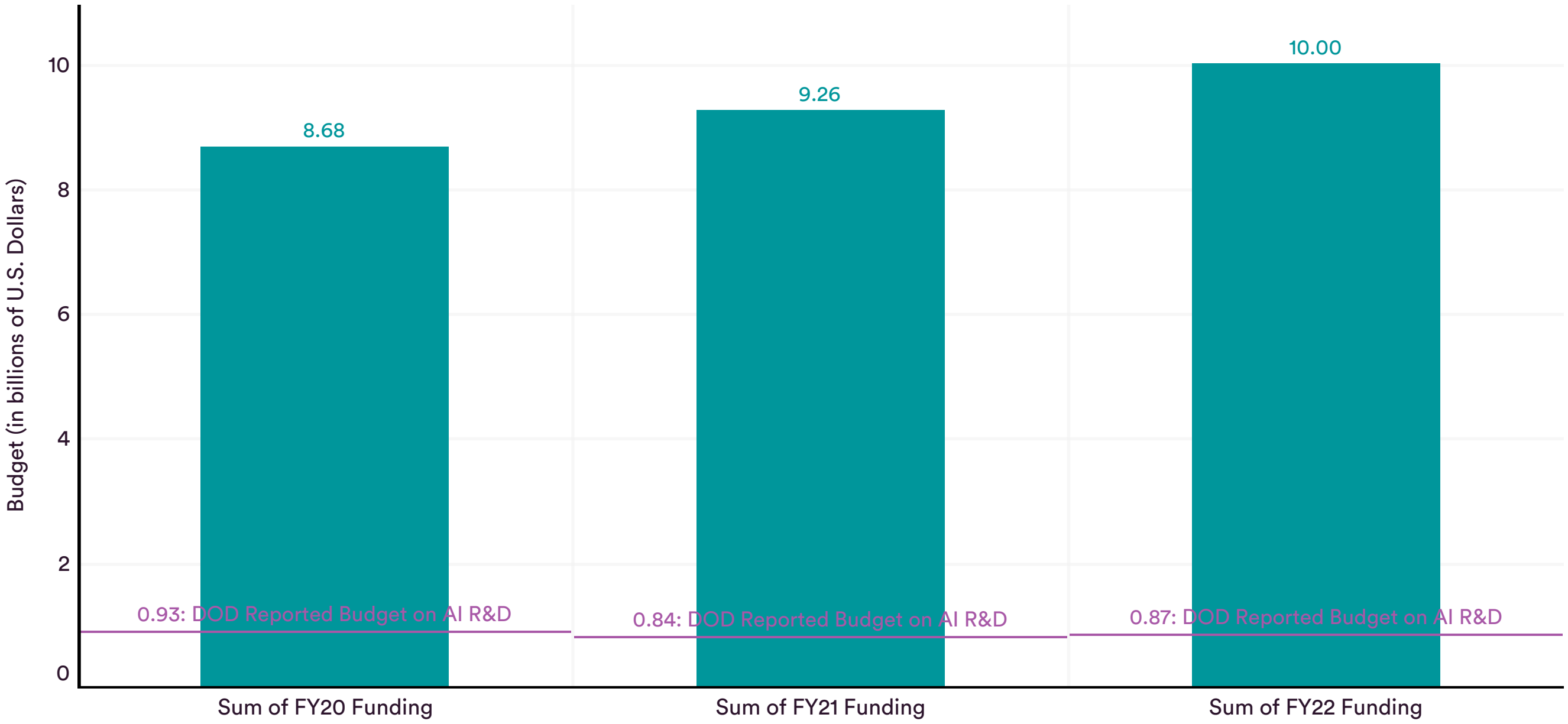

Figure 5.2.2

**Important data caveat:** This chart is indicative of one of the challenges of quantifying public AI spending. Bloomberg Government's analysis that searches AI-relevant keywords in DOD budgets shows that the department is requesting $10.0 billion for AI-specific R&D in FY 2022. However, DOD's own measurement produces a smaller number of $874 million. The discrepancy may result from the difference in defining AI-related budget items. For example, a research project that uses AI for cyber defense may count human, hardware, and operations-related expenditures within the AI-related budget request, though the AI software component will be much smaller.

## DOD Top Five Highest-Funded Programs

This section highlight offers a more qualitative look at some of the AI-related research projects the DOD prioritizes. Table 5.2.1 presents the five DOD-related AI programs that received the greatest funding in 2021. In the past year, the DOD was interested in deploying AI for a number of purposes, from geospatial monitoring to reducing the threat posed by weapons of mass destruction.

| Program Name | Department | Funds Received (in millions) | Purpose |
|---|---|---|---|
| 1 Rapid Capability Development and Maturation | Army | 257 | Fund the development, engineering, acquisition, and operation of various AI-related technological prototypes that could be used for military purposes. |
| 2 Counter Weapons of Mass Destruction Advanced Technology Development | Defense Threat Reduction Agency | 254 | Develop technologies that could "deny, defeat and disrupt" weapons of mass destruction (WMD). |
| 3 Algorithmic Warfare Cross-Functional Teams – Software Pilot Program | Office of the Secretary of Defense | 230 | Accelerate the integration of AI technologies in DOD systems to "improve warfighting speed and lethality." |
| 4 Joint Artificial Intelligence Center | Defense Information Systems Agency | 137 | Develop, test, prototype, and demonstrate various AI and machine learning capabilities with the intention of integrating these capabilities across numerous domains which include "supply chain, personal recovery, infrastructure assessment, geospatial monitoring during disaster and cyber sense making." |
| 5 High Performance Computing Modernization Program | Army | 96 | Investigate, demonstrate, and mature both general and special-purpose supercomputing environments that are used to satisfy wide-ranging DOD priorities. |

Table 5.2.1

## DOD AI R&D Spending by Department

DOD spending on AI R&D can also be broken down on a subdepartmental level, which reveals how individual defense agencies—the Army and the Navy, for instance—compare in their AI spending (Figure 5.2.3). The U.S. Navy was the top-spending DOD agency in FY 2021 and is poised to maintain that position in 2022. They have requested a total of $1.86 billion in FY 2022 for AI-related projects, followed by the Army ($1.77 billion), the Office of the Secretary of Defense ($1.1 billion) and the Air Force ($883 million).

**U.S. DOD BUDGET for AI-SPECIFIC RESEARCH, DEVELOPMENT, TEST and EVALUATION (RDT&E) by DEPARTMENT, FY 2020–22**

Source: Bloomberg Government, 2021 | Chart: 2022 AI Index Report

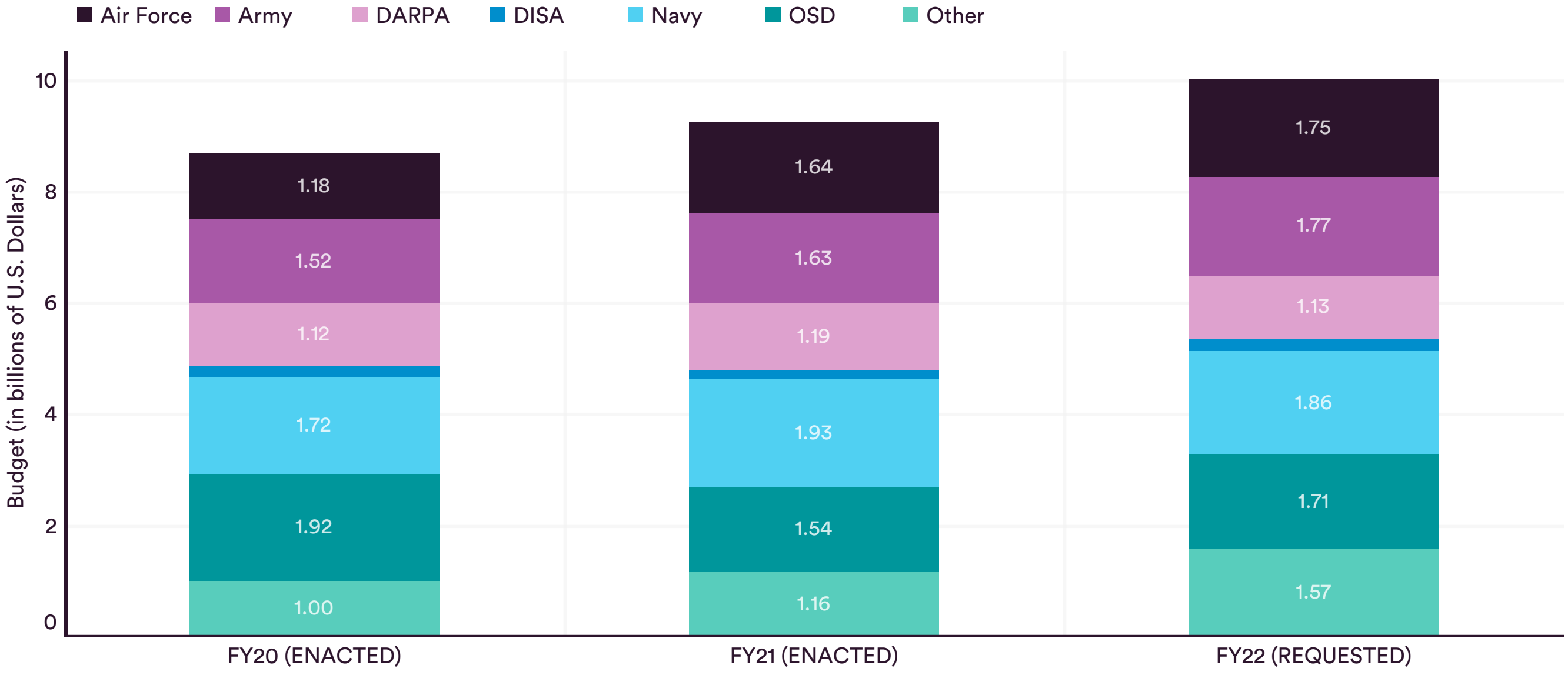

Figure 5.2.3

## U.S. GOVERNMENT AI-RELATED CONTRACT SPENDING

Public investment in AI can also be measured by federal government spending on AI-related contracts. U.S. government agencies often award contracts to private companies for the supply of various goods and services that typically occupy the largest share of an agency's budget. Bloomberg Government built a model to classify whether a U.S. government contract was AI-related by adding up all contracting transactions that contain a set of more than 100 AI-specific keywords in their titles or descriptions.[5]

### Total Contract Spending

In 2021, federal departments and agencies spent a total of $1.79 billion on AI-related contracts. Although this amount is nearly double what was spent on AI-related contracts in 2018 (roughly $920 million), it represents a slight decrease from the amount spent on AI-related contracts in 2020, which peaked at $1.97 billion (Figure 5.2.4).

**U.S. GOVERNMENT TOTAL CONTRACT SPENDING on AI, FY 2000–21**
Source: Bloomberg Government, 2021 | Chart: 2022 AI Index Report

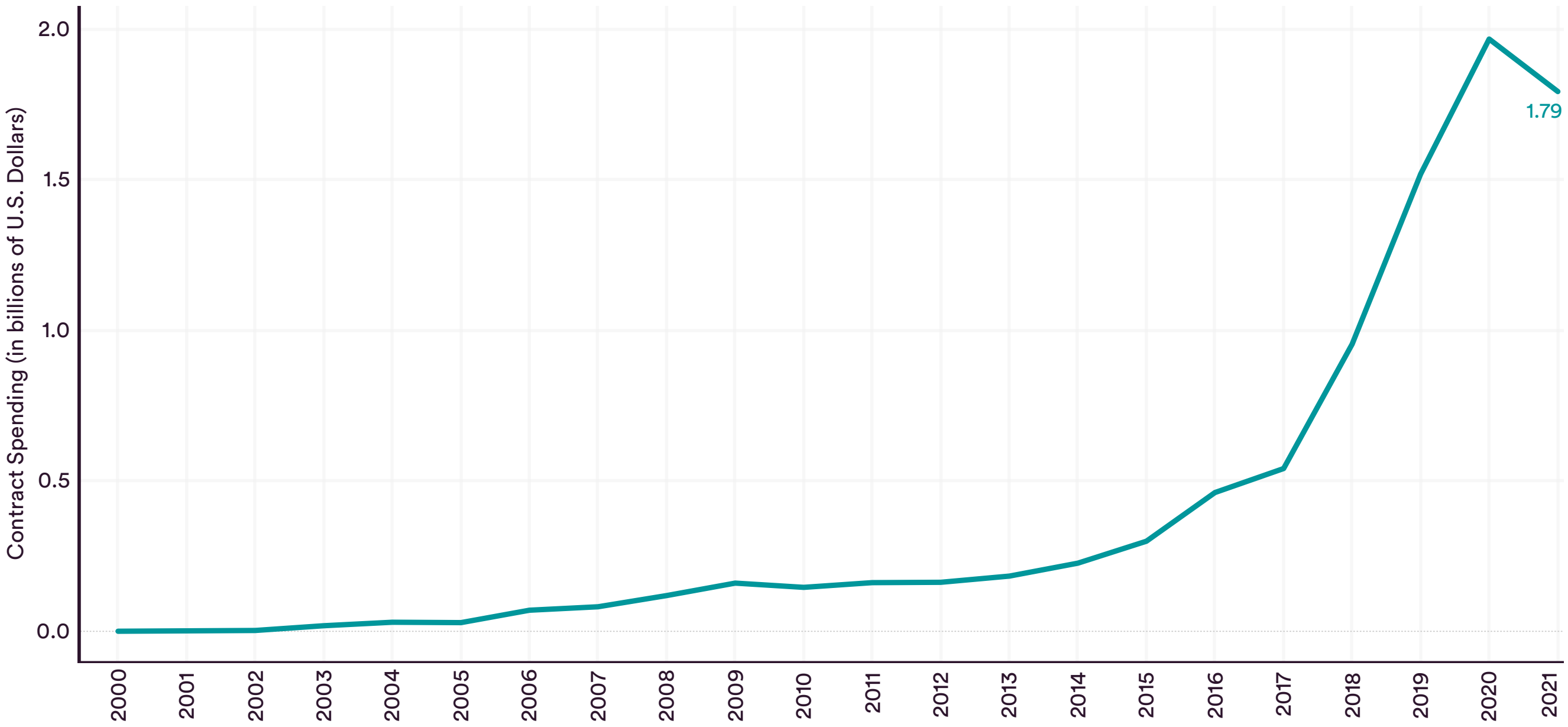

Figure 5.2.4

5 Note that contractors may add a number of keywords into their applications during the procurement process, so some of the projects included may have a relatively small AI component relative to other parts of technology.

## Contract Spending by Department and Agency

Figures 5.2.5 and 5.2.6 report AI-related contract spending by the top 10 federal agencies in 2021 and from 2000 to 2021, respectively. The DOD outspent the rest of the U.S. government on both charts by a significant margin. In 2021, it spent $1.14 billion on AI-related contracts, roughly five times what was spent by the next highest department, the Department of Health and Human Services ($234 million).

Aggregate spending on AI contracts in the last four years tells a similar story. Since 2018, the DOD has spent $5.20 billion on AI contracts, approximately seven times the next highest spender, NASA ($1.41 billion). In fact, since 2018, the DOD has spent twice as much on AI-related contracts as all other government agencies combined. Following the DOD and NASA are the Department of Health and Human Services ($700 million), the Department of Homeland Security ($362 million), and Department of the Treasury ($156 million).

**TOP CONTRACT SPENDING on AI by U.S. GOVERNMENT DEPARTMENT and AGENCY, 2021**
Source: Bloomberg Government, 2021 | Chart: 2022 AI Index Report

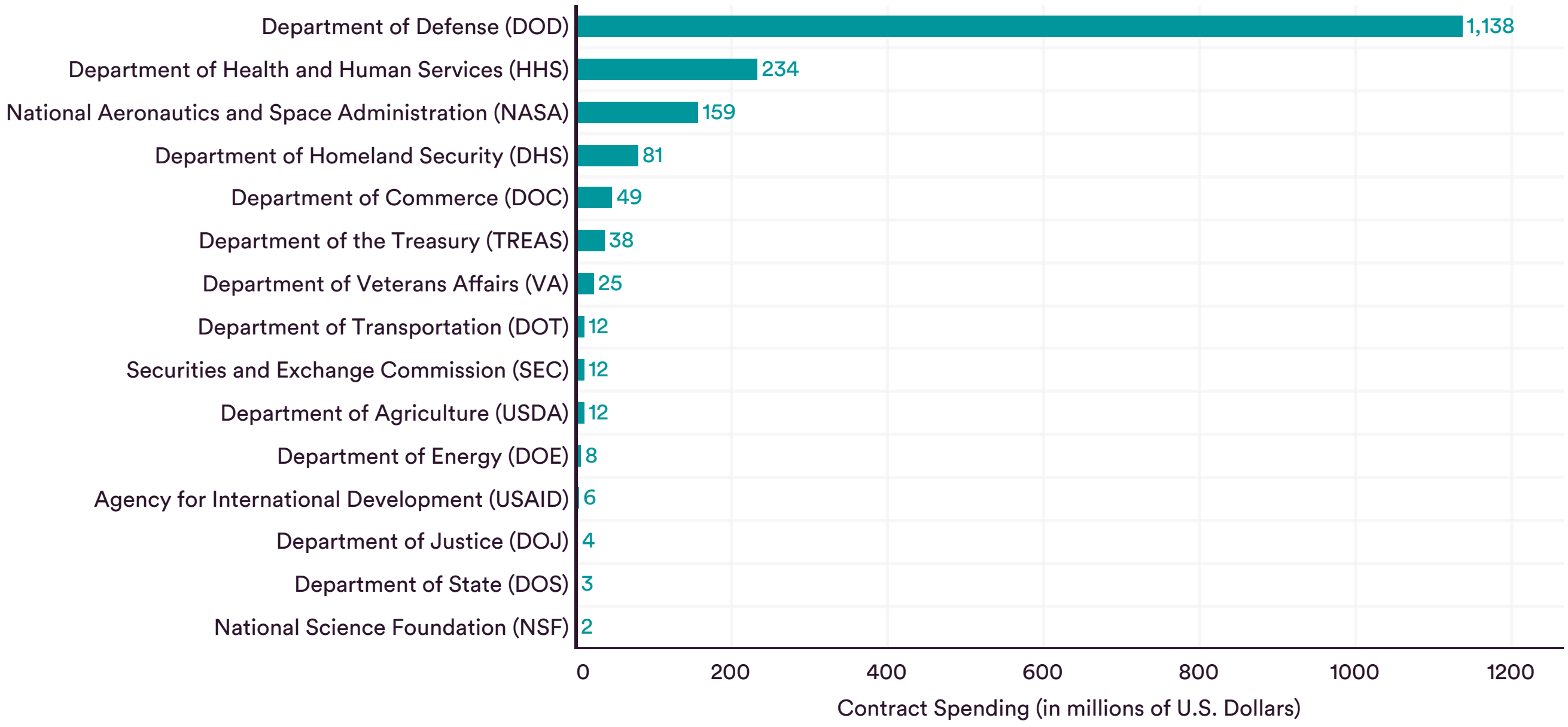

Figure 5.2.5

## TOP CONTRACT SPENDING on AI by U.S. GOVERNMENT DEPARTMENT and AGENCY, 2000–21 (SUM)

Source: Bloomberg Government, 2021 | Chart: 2022 AI Index Report

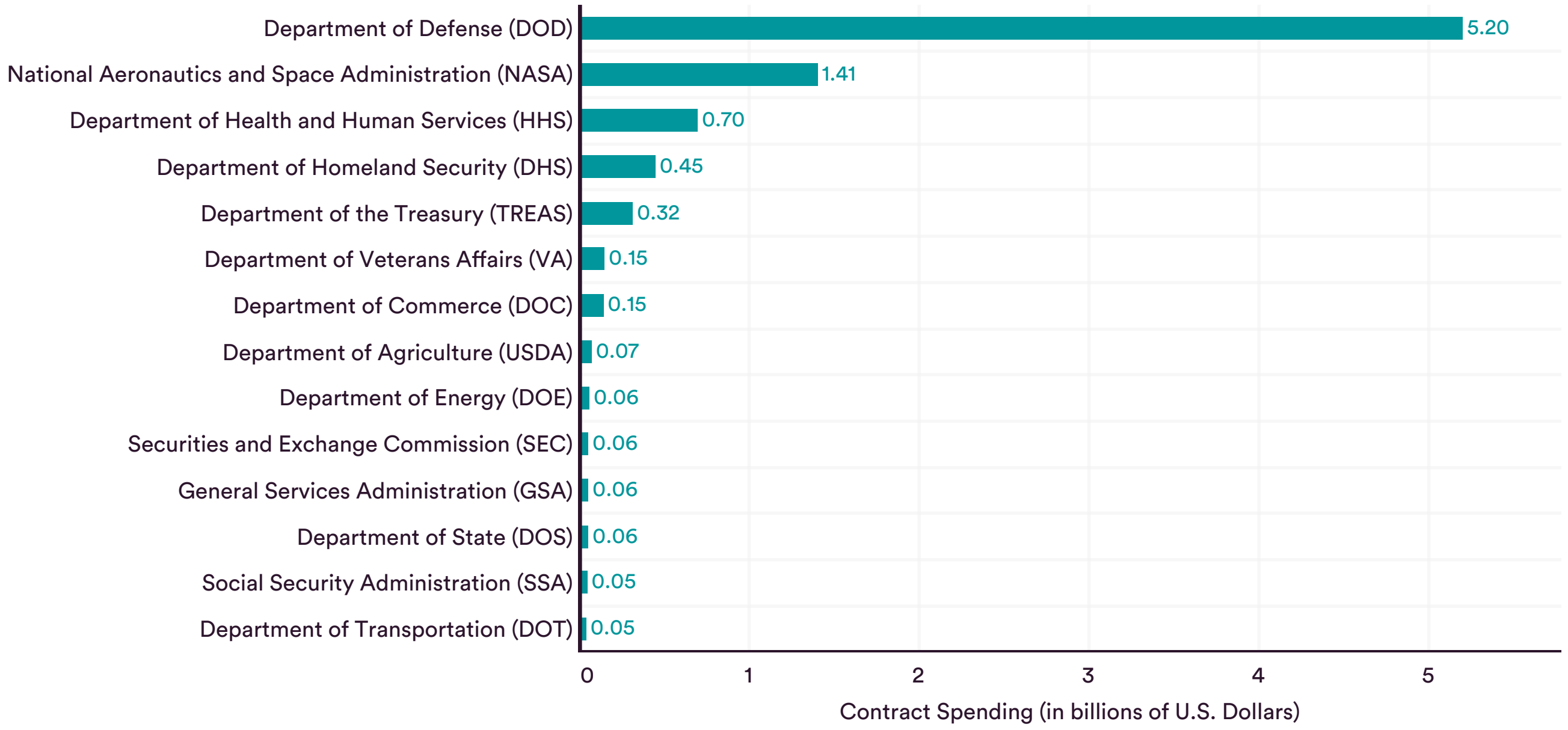

Figure 5.2.6

## Largest Contract for Five Top-Spending Departments in 2021

To paint a better picture of how different U.S. government departments use AI, Table 5.2.2 shows the most expensive AI-related contract that the five highest AI-related-spending departments signed in 2021. Last year, the U.S. government invested in AI to build autonomous vehicle prototypes, develop an AI imaging system that could assist with burn classification, and create robots capable of higher-level lunar navigation.

| Contract Name | Department | Amount (in millions) | Purpose |
|---|---|---|---|
| Prototype Services in the Objective Areas of Automotive Cybersecurity, Vehicle Safety Technologies, Vehicle Light Weighting, Autonomous Vehicles and Intelligent Systems, Connected Vehicles, and Advanced Energy Storage Technologies | DOD | 70 | To acquire prototypes in the domain of automotive cybersecurity, vehicle safety technologies, and autonomous vehicles and intelligent systems. |
| Biomedical Advanced Research and Development Authority (BARDA) | HHS | 20 | To develop optical imaging devices and machine learning algorithms to assist in classifying and healing wounds and conventional burns. |
| Commercial Lunar Payload Services | NASA | 14 | To develop lunar robots capable of navigating the moon's south pole to acquire lunar resources and engage in lunar-based scientific activities. |
| SBIR-Autonomous Surveillance Towers-Delivery Order | DHS | 37 | To construct towers capable of autonomous surveillance. |
| Schedule 70: Information Technology | DOC | 13 | To develop a prototype using AI technology that can improve patent search. |

Table 5.2.2

Artificial Intelligence
Index Report 2022

# Appendix

# Appendix

# CHAPTER 1: RESEARCH & DEVELOPMENT

## CENTER FOR SECURITY AND EMERGING TECHNOLOGY, GEORGETOWN UNIVERSITY

Prepared by Sara Abdulla and James Dunham

The Center for Security and Emerging Technology (CSET) is a policy research organization within Georgetown University's Walsh School of Foreign Service that produces data-driven research at the intersection of security and technology, providing nonpartisan analysis to the policy community.

### Publications from CSET Merged Corpus of Scholarly Literature

**Source**

CSET's merged corpus of scholarly literature combines distinct publications from Digital Science's Dimensions, Clarivate's Web of Science, Microsoft Academic Graph, China National Knowledge Infrastructure, arXiv, and Papers With Code.[1]

**Methodology**

To create the merged corpus, CSET deduplicated across the listed sources using publication metadata, and then combined the metadata for linked publications. To identify AI publications, CSET used an English-language subset of this corpus: publications since 2010 that appear AI-relevant.[2] CSET researchers developed a classifier for identifying AI-related publications by leveraging the arXiv repository, where authors and editors tag papers by subject.

To provide a publication's field of study, CSET matches each publication in the analytic corpus with predictions from Microsoft Academic Graph (MAG)'s field-of-study model, which yields hierarchical labels describing the published research field(s) of study and corresponding scores.[3] CSET researchers identified the most common fields of study in their corpus of AI-relevant publications since 2010 and recorded publications in all other fields as "Other AI." English-language AI-relevant publications were then tallied by their top-scoring field and publication year.

CSET also provided year-by-year citations for AI-relevant work associated with each country. A publication is associated with a country if it has at least one author whose organizational affiliation(s) is located in that country. Citation counts aren't available for all publications; those without counts weren't included in the citation analysis. Over 70% of English-language AI papers published between 2010 and 2020 have citation data available.

CSET counted cross-country collaborations as distinct pairs of countries across authors for each publication. Collaborations are only counted once: For example, if a publication has two authors from the United States and two authors from China, it is counted as a single United States-China collaboration.

Additionally, publication counts by year and by publication type (e.g., academic journal articles, conference papers) were provided where available. These publication types were disaggregated by affiliation country as described above.

CSET also provided publication affiliation sector(s) where, as in the country attribution analysis, sectors were associated with publications through authors' affiliations. Not all affiliations were characterized in terms of sectors; CSET researchers relied primarily on GRID from Digital Science for this purpose, and not all organizations can be found in or linked to GRID.[4] Where the affiliation

1 All CNKI content is furnished for CSET by East View Information Services, Minneapolis, MN, USA.
2 For more information, see James Dunham, Jennifer Melot, and Dewey Murdick, "Identifying the Development and Application of Artificial Intelligence in Scientific Text," arXiv [cs.DL], May 28, 2020, https://arxiv.org/abs/2002.07143.
3 These scores are based on cosine similarities between field-of-study and paper embeddings. See Zhihong Shen, Hao Ma, and Kuansan Wang, "A Web-Scale System for Scientific Knowledge Exploration," arXiv [cs.CL], May 30, 2018, https://arxiv.org/abs/1805.12216.
4 See https://www.grid.ac/ for more information about the GRID dataset from Digital Science.

sector is available, papers were counted toward these sectors, by year. Cross-sector collaborations on academic publications were calculated using the same method as in the cross-country collaborations analysis.

## Patents from CSET's AI Patents Dataset

**Source**

CSET's AI patents dataset was developed by CSET and 1790 Analytics. It includes patents relevant to the development and application of AI, as indicated by CPC/IPC codes and keywords.

**Methodology**

In this analysis, patents were grouped by year and country, and then counted at the "patent family" level.[5] CSET extracted year values from the most recent publication date within a family. This method has the advantage of capturing updates within a patent family (such as amendments).

The country of origin for a patent is derived from the first country in which a patent was filed.[6]

## GITHUB STARS

**Source**

GitHub: star-history (available at star history website) was used to retrieve the data.

**Methodology**

The visual in the report shows the number of stars for various GitHub repositories over time. The repositories include the following:
apachecn/ailearning, apache/incubator-mxnet, Avik-Jain/100-Days-Of-ML-Code, aymericdamien/TensorFlow-Examples, BVLC/cafe, cafe2/cafe2, CorentinJ/Real-Time-Voice-Cloning, deepfakes / faceswap, dmlc/mxnet, exacity/deeplearningbook-chinese, fchollet/keras, floodsung/Deep-Learning-Papers-Reading-Roadmap, iperov/DeepFaceLab, Microsoft/CNTK, opencv/opencv, pytorch/pytorch, scikit-learn/scikit-learn, scutan90/DeepLearning-500-questions, tensorflow/tensorflow, Theano/Theano, Torch/Torch7.

**Nuance**

The GitHub Archive currently does not provide a way to count when users remove a star from a repository. Therefore, the reported data slightly overestimates the number of stars. A comparison with the actual number of stars for the repositories on GitHub reveals that the numbers are fairly close and that the trends remain unchanged.

5 Patents are analyzed at the "patent family" level rather than "patent document" level because patent families are a collective of patent documents all associated with a single invention and/or innovation by the same inventors/assignees. Thus, counting at the "patent family" level mitigates artificial number inflation when there are multiple patent documents in a patent family or if a patent is filed in multiple jurisdictions.
6 For more details on CSET's approach and experimentation for assigning country values, see footnote 26 in "Patents and Artificial Intelligence: A Primer," by Dewey Murdick and Patrick Thomas. (Center for Security and Emerging Technology, September 2020), https://doi.org/10.51593/20200038.

# CHAPTER 2: TECHNICAL PERFORMANCE

## ImageNet

Data on ImageNet accuracy was retrieved through a detailed arXiv literature review cross-referenced by technical progress reported on Papers with Code. The reported dates correspond to the year during which a paper was first published to arXiv, and the reported results (top-1 or top-5 accuracy) correspond to the result reported in the most recent version of each paper. The estimate of human-level performance is from Russakovsky et al., 2015. Learn more about the LSVRC ImageNet competition and the ImageNet data set.

## ImageNet: Top-1 Accuracy

To highlight progress on top-1 accuracy without the use of extra training data, scores were taken from the following papers:

Adversarial Examples Improve Image Recognition
Billion-Scale Semi-Supervised Learning for Image Classification
Dual Path Networks
Densely Connected Convolutional Networks
EfficientNet: Rethinking Model Scaling for Convolutional Neural Networks
Fixing the Train-Test Resolution Discrepancy: FixEfficientNet
ImageNet Classification with Deep Convolutional Neural Networks
Masked Autoencoders Are Scalable Vision Learners

To highlight progress on top-1 accuracy with the use of extra training data, scores were taken from the following papers:

Big Transfer (BiT): General Visual Representation Learning
CoAtNet: Marrying Convolution and Attention for All Data Sizes
EfficientNet: Rethinking Model Scaling for Convolutional Neural Networks
Self-Training with Noisy Student Improves ImageNet Classification
Sharpness-Aware Minimization for Efficiently Improving Generalization
Xception: Deep Learning with Depthwise Separable Convolutions

## ImageNet: Top-5 Accuracy

To highlight progress on top-5 accuracy without the use of extra training data, scores were taken from the following papers:

Adversarial Examples Improve Image Recognition
EfficientNet: Rethinking Model Scaling for Convolutional Neural Networks
Exploring the Limits of Weakly Supervised Pretraining
Fixing the Train-Test Resolution Discrepancy: FixEfficientNet
GPipe: Efficient Training of Giant Neural Networks Using Pipeline Parallelism
High-Performance Large-Scale Image Recognition Without Normalization
ImageNet Classification with Deep Convolutional Neural Networks
Learning Transferable Architectures for Scalable Image Recognition
Squeeze-and-Excitation Networks

To highlight progress on top-5 accuracy with the use of extra training data, scores were taken from the following papers:

Big Transfer (BiT): General Visual Representation Learning
Deep Residual Learning for Image Recognition
EfficientNet: Rethinking Model Scaling for Convolutional Neural Networks
Florence: A New Foundation Model for Computer Vision
Self-Training with Noisy Student Improves ImageNet Classification
Xception: Deep Learning with Depthwise Separable Convolutions

## STL-10

Data on STL-10 FID scores was retrieved through a detailed arXiv literature review cross-referenced by technical progress reported on Papers with Code. The reported dates correspond to the year during which a paper was first published to arXiv, and the reported results (FID score) correspond to the result reported in the most recent version of each paper. Details on the STL-10 benchmark can be found in the STL-10 paper.

To highlight progress on STL-10, scores were taken from the following papers:
DEGAS: Differentiable Efficient Generator Search
Dist-GAN: An Improved GAN Using Distance Constraints
Off-Policy Reinforcement Learning for Efficient and Effective GAN Architecture
Search Score Matching Model for Unbounded Data Score

### CIFAR-10

Data on CIFAR-10 FID scores was retrieved through a detailed arXiv literature review cross-referenced by technical progress reported on Papers with Code. The reported dates correspond to the year during which a paper was first published to arXiv, and the reported results (FID score) correspond to the result reported in the most recent version of each paper. Details on the CIFAR-10 benchmark can be found in the CIFAR-10 paper.

To highlight progress on CIFAR-10, scores were taken from the following papers:
AutoGAN: Neural Architecture Search for Generative Adversarial Networks
Denoising Diffusion Probabilistic Models
Improved Training of Wasserstein GANs
Large Scale GAN Training for High Fidelity Natural Image Synthesis
Score-Based Generative Modeling in Latent Space

### FaceForensics++

Data on FaceForensics++ accuracy was retrieved through a detailed arXiv literature review. The reported dates correspond to the year during which a paper was first published to arXiv or a method was introduced. With FaceForensics, recent researchers have tested previously existing deepfake detection methodologies. The year in which a method was introduced, even if it was subsequently tested, is the year in which it is included in the report. The reported results (accuracy) correspond to the result reported in the most recent version of each paper. Details on the FaceForensics++ benchmark can be found in the FaceForensics++ paper.

To highlight progress on FaceForensics++, scores were taken from the following papers:
A Deep Learning Approach to Universal Image Manipulation Detection Using a New Convolutional Layer
Detection of Deepfake Videos Using Long Distance Attention
FakeCatcher: Detection of Synthetic Portrait Videos Using Biological Signals
FaceForensics++: Learning to Detect Manipulated Facial Images
Learning Spatiotemporal Features with 3D Convolutional Networks
Recasting Residual-Based Local Descriptors as Convolutional Neural Networks
Rich Models for Steganalysis of Digital Images
Thinking in Frequency: Face Forgery Detection by Mining Frequency-Aware Clues
Xception: Deep Learning with Depthwise Separable Convolutions

### Celeb-DF

Data on Celeb-DF AUC was retrieved through a detailed arXiv literature review. The reported dates correspond to the year during which a paper was first published to arXiv or a method was introduced. With Celeb-DF, recent researchers have tested previously existing deepfake detection methodologies. The year in which a method was introduced, even if it was subsequently tested, is the year in which it is included in the report. The reported results (AUC) correspond to the result reported in the most recent version of each paper. Details on the Celeb-DF benchmark can be found in the Celeb-DF paper.

To highlight progress on Celeb-DF, scores were taken from the following papers:
Exposing DeepFake Videos by Detecting Face Warping

Artifacts
FaceForensics++: Learning to Detect Manipulated Facial Images
Face X-Ray for More General Face Forgery Detection
Spatial-Phase Shallow Learning: Rethinking Face Forgery Detection in Frequency Domain

### Leeds Sports Poses

Data on Leeds Sports Poses percentage of correct keypoints (PCK) was retrieved through a detailed arXiv literature review cross-referenced by technical progress reported on Papers with Code. The reported dates correspond to the year during which a paper was first published to arXiv, and the reported results (PCK) correspond to the result reported in the most recent version of each paper. Details on the Leeds Sports Poses benchmark can be found in the Leeds Sports Poses paper.

To highlight progress on Leeds Sports Poses, scores were taken from the following papers:
Articulated Pose Estimation by a Graphical Model with Image Dependent Pairwise Relations
Human Pose Estimation via Convolutional Part Heatmap Regression
Jointly Optimize Data Augmentation and Network Training: Adversarial Data Augmentation in Human Pose Estimation
Knowledge-Guided Deep Fractal Neural Networks for Human Pose Estimation
OmniPose: A Multi-Scale Framework for Multi-Person Pose Estimation
Toward Fast and Accurate Human Pose Estimation via Soft-Gated Skip Connections

### Human 3.6M

Data on Human3.6M average mean per joint position error was retrieved through a detailed arXiv literature review cross-referenced by technical progress reported on Papers with Code. The reported dates correspond to the year during which a paper was first published to arXiv, and the reported results (MPJPE) correspond to the result reported in the most recent version of each paper. Details on the Human3.6M benchmark can be found in the Human3.6M paper.

To highlight progress on Human3.6M without the use of extra training data, scores were taken from the following papers:
3D Human Pose Estimation in Video with Temporal Convolutions and Semi-Supervised Training
Conditional Directed Graph Convolution for 3D Human Pose Estimation
Cross View Fusion for 3D Human Pose Estimation
Epipolar Transformers
Human3.6M: Large Scale Datasets and Predictive Methods for 3D Human Sensing in Natural Environments
Learning 3D Human Pose from Structure and Motion
Robust Estimation of 3D Human Poses from a Single Image

To highlight progress on Human3.6M with the use of extra training data, scores were taken from the following papers:
Epipolar Transformers
Learnable Triangulation of Human Pose
TesseTrack: End-to-End Learnable Multi-Person Articulated 3D Pose Tracking

### Cityscapes Challenge, Pixel-Level Semantic Labeling Task

Data on the Cityscapes challenge, pixel-level semantic labeling task mean IoU was taken from the Cityscapes dataset, more specifically their pixel-level semantic labeling leaderboard. More details about the Cityscapes dataset and other corresponding semantic segmentation challenges can be accessed at the Cityscapes dataset webpage.

### CVC-ClinicDB and Kvasir-SEG

Data on CVC-ClinicDB and Kvasir-SEG mean dice was retrieved through a detailed arXiv literature review cross-referenced by technical progress reported on Papers with Code (CVC-ClinicDB and Kvasir-SEG). The reported dates correspond to the year during which a paper was first published to arXiv, and the reported results (mean dice) correspond to the result reported in the most recent version of each paper. Details on the CVC-ClinicDB benchmark can be found in the CVC-ClinicDB database page. Details on the Kvasir-SEG benchmark can be found in the Kvasir-SEG paper.

To highlight progress on CVC-ClinicDB, scores were taken from the following papers:
DoubleU-Net: A Deep Convolutional Neural Network for Medical Image Segmentation
Encoder-Decoder with Atrous Separable Convolution for Semantic Image Segmentation
MSRF-Net: A Multi-Scale Residual Fusion Network for Biomedical Image Segmentation
ResUNet++: An Advanced Architecture for Medical Image Segmentation
U-Net: Convolutional Networks for Biomedical Image Segmentation

To highlight progress on Kvasir-SEG, scores were taken from the following papers:
Encoder-Decoder with Atrous Separable Convolution for Semantic Image Segmentation
MSRF-Net: A Multi-Scale Residual Fusion Network for Biomedical Image Segmentation
PraNet: Parallel Reverse Attention Network for Polyp Segmentation
U-Net: Convolutional Networks for Biomedical Image Segmentation

## National Institute of Standards and Technology (NIST) Face Recognition Vendor Test (FRVT) and NIST FRVT Face Mask Effects

Data on NIST FRVT 1:1 verification accuracy by dataset was obtained from the FRVT 1:1 verification leaderboard. Data on NIST FRVT face mask effects was obtained from the FRVT face mask effects leaderboard. The face mask effects leaderboard contains results of the testing of 319 face recognition algorithms that were submitted to FRVT prior to and post mid-March 2020, when the COVID pandemic began.

## Visual Question Answering (VQA)

Data on VQA was taken from recent iterations of the VQA challenge. To learn more about the VQA challenge in general, please consult the following link. To learn more about the 2021 iteration of the VQA challenge, please consult the following link. More specifically, the Index makes use of data from the following iterations of the VQA challenge:
VQA Challenge 2016
VQA Challenge 2017
VQA Challenge 2018
VQA Challenge 2019
VQA Challenge 2020
VQA Challenge 2021

## Kinetics-400, Kinetics-600, and Kinetics-700

Data on Kinetics-400, Kinetics-600, and Kinetics-700 was retrieved through a detailed arXiv literature review cross-referenced by technical progress reported on Papers with Code (Kinetics-400, Kinetics-600, and Kinetics-700). The reported dates correspond to the year during which a paper was first published to arXiv, and the reported results (accuracy) correspond to the result reported in the most recent version of each paper. Details on the Kinetics-400 benchmark can be found in the Kinetics-400 paper. Details on the Kinetics-600 benchmark can be found in the Kinetics-600 paper. Details on the Kinetics-700 benchmark can be found in the Kinetics-700 paper.

To highlight progress on Kinetics-400, scores were taken from the following papers:
Co-Training Transformer with Videos and Images Improves Action Recognition
Large-Scale Weakly-Supervised Pre-training for Video Action Recognition
Multiview Transformers for Video Recognition
Non-Local Neural Networks
Omni-Sourced Webly-Supervised Learning for Video Recognition
SlowFast Networks for Video Recognition
Temporal Segment Networks: Towards Good Practices for Deep Action Recognition

To highlight progress on Kinetics-600, scores were taken from the following papers:
Masked Feature Prediction for Self-Supervised Visual Pre-Training
Multiview Transformers for Video Recognition
Learning Spatio-Temporal Representation with Local and Global Diffusion
Rethinking Spatiotemporal Feature Learning: Speed-Accuracy Trade-offs in Video Classification
SlowFast Networks for Video Recognition

To highlight progress on Kinetics-700, scores were taken from the following papers:
Learn to Cycle: Time-Consistent Feature Discovery for Action Recognition
Masked Feature Prediction for Self-Supervised Visual Pre-Training
Multiview Transformers for Video Recognition

## ActivityNet: Temporal Action Localization Task

In the challenge, there are three separate tasks, but they focus on the main problem of temporally localizing where activities happen in untrimmed videos from the ActivityNet benchmark. To source information on the state-of-the-art results for TALT, the Index did a detailed survey of arXiv papers in addition to reports of yearly ActivityNet challenge results. More specifically, the Index made use of the following sources of information:
TALT 2016
TALT 2017
TALT 2018
TALT 2019
TALT 2020
TALT 2021

## Common Object in Context (COCO)

Data on COCO mean average precision (mAP50) was retrieved through a detailed arXiv literature review cross-referenced by technical progress reported on Papers with Code. The reported dates correspond to the year during which a paper was first published to arXiv, and the reported results (mAP50) correspond to the result reported in the most recent version of each paper. Details on the COCO benchmark can be found in the COCO paper.

To highlight progress on COCO without the use of extra training data, scores were taken from the following papers:
An Analysis of Scale Invariance in Object Detection – SNIP
Deformable ConvNets v2: More Deformable, Better Results
Dynamic Head: Unifying Object Detection Heads with Attentions
Inside-Outside Net: Detecting Objects in Context with Skip Pooling and Recurrent Neural Networks
Mish: A Self Regularized Non-Monotonic Activation Function
Scaled-YOLOv4: Scaling Cross Stage Partial Network

To highlight progress on COCO with the use of extra training data, scores were taken from the following papers:
EfficientDet: Scalable and Efficient Object Detection
Grounded Language-Image Pre-Training

## You Only Look Once (YOLO)

Data on YOLO mean average precision (mAP50) was retrieved through a detailed arXiv literature review and survey of GitHub repositories. The reported dates correspond to the year during which a paper was first published to arXiv or a method was introduced. More specifically, the Index made use of the following sources of information:
YOLO 2016
YOLO 2018
YOLO 2020
YOLO 2021

YOLO results for 2017 and 2019 were not included in the index as no state-of-the-art improvements in YOLO for those years were uncovered during the literature review and survey of GitHub repositories.

## Visual Commonsense Reasoning (VCR)

Technical progress for VCR is taken from the VCR leaderboard; the VCR leaderboard webpage further delineates the methodology behind the VCR challenge.

Human performance on VCR is taken from Zellers et al. (2018). Details on the VCR benchmark can be found in the VCR paper.

## SuperGLUE

The SuperGLUE benchmark data was pulled from the SuperGLUE leaderboard. Details about the SuperGLUE benchmark are in the SuperGLUE paper and SuperGLUE software toolkit. The tasks and evaluation metrics for SuperGLUE are:

| NAME | IDENTIFIER | METRIC |
|---|---|---|
| Broad Coverage Diagnostics | AX-b | Matthew's Carr |
| CommitmentBank | CB | Avg. F1/ Accuracy |
| Choice of Plausible Alternatives | COPA | Accuracy |
| Multi-Sentence Reading Comprehension | MultiRC | F1a/EM |
| Recognizing Textual Entailment | RTE | Accuracy |
| Words in Context | WiC | Accuracy |
| The Winograd Schema Challenge | WSC | Accuracy |
| BoolQ | BoolQ | Accuracy |
| Reading Comprehension with Commonsense Reasoning | ReCoRD | F1/ Accuracy |
| Winogender Schema Diagnostic | AX-g | Gender Parity/ Accuracy |

## SQuAD 1.1 and SQuAD 2.0

Data on SQuAD 1.1 performance was taken from Papers with Code. Data on SQuAD 2.0 performance was taken from the SQuAD 2.0 leaderboard. Details about the SQuAD 1.1 benchmark are in the SQuAD 1.1 paper. Details about the SQuAD 2.0 benchmark are in the SQuAD 2.0 paper.

## Reading Comprehension Dataset Requiring Logical Reasoning (ReClor)

Data on ReClor performance was taken from the ReClor leaderboard. Details about the ReClor benchmark are in the ReClor paper.

## arXiv

Data on arXiv recall-oriented understudy for gisting evaluation (ROUGE-1) was retrieved through a detailed arXiv literature review cross-referenced by technical progress reported on Papers with Code. The reported dates correspond to the year during which a paper was first published to arXiv, and the reported results (ROUGE-1) correspond to the result reported in the most recent version of each paper. Details about the arXiv benchmark are in the arXiv dataset webpage.

To highlight progress on arXiv without the use of extra training data, scores were taken from the following papers:
A Discourse-Aware Attention Model for Abstractive Summarization of Long Documents
Extractive Summarization of Long Documents by Combining Global and Local Context
Get to the Point: Summarization with Pointer-Generator Networks
Systematically Exploring Redundancy Reduction in Summarizing Long Documents
Sparsifying Transformer Models with Trainable Representation Pooling

To highlight progress on arXiv with the use of extra training data, scores were taken from the following papers:
Big Bird: Transformers for Longer Sequences
Hierarchical Learning for Generation with Long Source Sequences
PEGASUS: Pre-Training with Extracted Gap-Sentences for Abstractive Summarization

### PubMed

Data on PubMed recall-oriented understudy for gisting evaluation (ROUGE-1) was retrieved through a detailed arXiv literature review cross-referenced by technical progress reported on Papers with Code. The reported dates correspond to the year during which a paper was first published to arXiv, and the reported results (ROUGE-1) correspond to the result reported in the most recent version of each paper. Details about the PubMed benchmark are in the PubMed paper.

To highlight progress on PubMed without the use of extra training data, scores were taken from the following papers:
A Discourse-Aware Attention Model for Abstractive Summarization of Long Documents
Extractive Summarization of Long Documents by Combining Global and Local Context
Get to the Point: Summarization with Pointer-Generator Networks
Sparsifying Transformer Models with Trainable Representation Pooling

To highlight progress on PubMed with the use of extra training data, scores were taken from the following papers:
A Divide-and-Conquer Approach to the Summarization of Long Documents
Hierarchical Learning for Generation with Long Source Sequences
PEGASUS: Pre-Training with Extracted Gap-Sentences for Abstractive Summarization

### Stanford Natural Language Inference (SNLI)

Data on Stanford Natural Language Inference (SNLI) accuracy was retrieved through a detailed arXiv literature review cross-referenced by technical progress reported on Papers with Code. The reported dates correspond to the year during which a paper was first published to arXiv, and the reported results (accuracy) correspond to the result reported in the most recent version of each paper. Details on the SNLI benchmark can be found in the SNLI paper.

To highlight progress on SNLI, scores were taken from the following papers:
Compare, Compress and Propagate: Enhancing Neural Architectures with Alignment Factorization for Natural Language Inference
Convolutional Neural Networks for Sentence Classification
Enhanced LSTM for Natural Language Inference
Entailment as Few-Shot Learner
Explicit Contextual Semantics for Text Comprehension
Semantics-Aware BERT for Language Understanding
Self-Explaining Structures Improve NLP Models

### Abductive Natural Language Inference (aNLI)

Data on Abductive Natural Language Inference (aNLI) was sourced from the Allen Institute for AI's aNLI leaderboard. Details on the aNLI benchmark can be found in the aNLI paper.

### SemEval 2014 Task 4 Sub Task 2

Data on SemEval 2014 Task 4 Sub Task 2 accuracy was retrieved through a detailed arXiv literature review cross-referenced by technical progress reported on Papers with Code. The reported dates correspond to the year during which a paper was first published to arXiv, and the reported results (accuracy) correspond to the result reported in the most recent version of each paper. Details on the SemEval benchmark can be found in the SemEval 2014 paper.

To highlight progress on SemEval, scores were taken from the following papers:
A Multi-Task Learning Model for Chinese-Oriented Aspect Polarity Classification and Aspect Term Extraction

Aspect Level Sentiment Classification with Deep Memory Network
Back to Reality: Leveraging Pattern-Driven Modeling to Enable Affordable Sentiment Dependency Learning
Effective LSTMs for Target-Dependent Sentiment Classification
Hierarchical Attention Based Position-Aware Network for Aspect-Level Sentiment Analysis
Investigating Typed Syntactic Dependencies for Targeted Sentiment Classification Using Graph Attention Neural Network
Recurrent Attention Network on Memory for Aspect Sentiment Analysis

## WMT2014, English-French and English-German

Data on WMT2014 English-French and English-German BLEU score was retrieved through a detailed arXiv literature review cross-referenced by technical progress reported on Papers with Code (English-French and English-German). The reported dates correspond to the year during which a paper was first published to arXiv, and the reported results (BLEU score) correspond to the result reported in the most recent version of each paper. Details about both the WMT2014 English-French and English-German benchmarks can be found in the WMT2014 paper.

To highlight progress on WMT2014 English-French without the use of extra training data, scores were taken from the following papers:
Addressing the Rare Word Problem in Neural Machine Translation
Google's Neural Machine Translation System: Bridging the Gap Between Human and Machine Translation
MUSE: Parallel Multi-Scale Attention for Sequence to Sequence Learning
R-Drop: Regularized Dropout for Neural Networks
Scaling Neural Machine Translation
Understanding the Difficulty of Training Transformers
Weighted Transformer Network for Machine Translation

To highlight progress on WMT2014 English-French with the use of extra training data, scores were taken from the following papers:
Understanding Back-Translation at Scale
Very Deep Transformers for Neural Machine Translation

To highlight progress on WMT2014 English-German without the use of extra training data, scores were taken from the following papers:
BERT, mBERT, or BiBERT? A Study on Contextualized Embeddings for Neural Machine Translation
Effective Approaches to Attention-based Neural Machine Translation
Data Diversification: A Simple Strategy for Neural Machine Translation
Fast and Simple Mixture of Softmaxes with BPE and Hybrid-LightRNN for Language Generation
Google's Neural Machine Translation System: Bridging the Gap Between Human and Machine Translation
Incorporating BERT into Neural Machine Translation
Weighted Transformer Network for Machine Translation

To highlight progress on WMT2014 English-German with the use of extra training data, scores were taken from the following papers:
Lessons on Parameter Sharing across Layers in Transformers
Understanding Back-Translation at Scale

## Number of Commercially Available MT Systems

Details about the number of commercially available MT systems was sourced from Intento's report The State of Machine Translation, 2021. Intento is a San Francisco-based startup that analyzes commercially available MT services.

## LibriSpeech (Test-Clean and Other Dataset)

Data on LibriSpeech (Test-Clean and Other) word error rate was retrieved through a detailed arXiv literature review cross-referenced by technical progress reported on Papers with Code (Test-Clean and Other). The reported dates correspond to the year during which a paper was first published to arXiv, and the reported results (word error rate) correspond to the result reported in the most recent version of each paper. Details about both the LibriSpeech Test-Clean and Test-Other benchmarks can be found in the LibriSpeech paper.

To highlight progress on LibriSpeech Test-Clean without the use of extra training data, scores were taken from the following papers:
ASAPP-ASR: Multistream CNN and Self-Attentive SRU for SOTA Speech Recognition
Letter-Based Speech Recognition with Gated ConvNets
Neural Network Language Modeling with Letter-Based Features and Importance Sampling
SpeechStew: Simply Mix All Available Speech Recognition Data to Train One Large Neural Network
State-of-the-Art Speech Recognition Using Multi-Stream Self-Attention With Dilated 1D Convolutions

To highlight progress on LibriSpeech Test-Clean with the use of extra training data, scores were taken from the following papers:
Deep Speech 2: End-to-End Speech Recognition in English and Mandarin
End-to-End ASR: From Supervised to Semi-Supervised Learning with Modern Architectures
Pushing the Limits of Semi-Supervised Learning for Automatic Speech Recognition

To highlight progress on LibriSpeech Test-Other without the use of extra training data, scores were taken from the following papers:
Conformer: Convolution-Augmented Transformer for Speech Recognition
Neural Network Language Modeling with Letter-Based Features and Importance Sampling
SpeechStew: Simply Mix All Available Speech Recognition Data to Train One Large Neural Network
Transformer-Based Acoustic Modeling for Hybrid Speech Recognition

To highlight progress on LibriSpeech Test-Other with the use of extra training data, scores were taken from the following papers:
Deep Speech 2: End-to-End Speech Recognition in English and Mandarin
End-to-End ASR: From Supervised to Semi-Supervised Learning with Modern Architectures
Pushing the Limits of Semi-Supervised Learning for Automatic Speech Recognition
W2v-BERT: Combining Contrastive Learning and Masked Language Modeling for Self-Supervised Speech Pre-Training

## VoxCeleb

VoxCeleb is an audio-visual dataset consisting of short clips of human speech, extracted from interview videos uploaded to YouTube. VoxCeleb contains speech from 7,000-plus speakers spanning a wide range of ethnicities, accents, professions, and ages—amounting to over a million utterances (face-tracks are captured "in the wild," with background chatter, laughter, overlapping speech, pose variation, and different lighting conditions) recorded over a period of 2,000 hours (both audio and video). Each segment is at least three seconds long. The data contains an audio dataset based on celebrity voices, shorts, films, and conversational pieces (e.g., talk shows). The initial VoxCeleb 1 (100,000 utterances taken from 1,251 celebrities on YouTube) was expanded to VoxCeleb 2 (1 million utterances from 6,112 celebrities).

For the sake of consistency, the AI Index reported scores on the initial VoxCeleb dataset. Specifically, the Index made use of the following sources of information:
The IDLAB VoxSRC-20 Submission: Large Margin Fine-Tuning and Quality-Aware Score Calibration in DNN Based Speaker Verification
The SpeakIn System for VoxCeleb Speaker Recognition Challenge 2021
VoxCeleb: A Large-Scale Speaker Identification Dataset
VoxCeleb2: Deep Speaker Recognition
VoxCeleb: Large-Scale Speaker Verification in the Wild

## MovieLens 20M

Data on MovieLens 20M normalized discounted cumulative gain@100 (nDCG@100) was retrieved through a detailed arXiv literature review cross-referenced by technical progress reported on Papers with Code. The reported dates correspond to the year during which a paper was first published to arXiv, and the reported results (nDCG@100) correspond to the result reported in the most recent version of each paper. Details on the MovieLens series of benchmarks can be found in Harper et al. 2015.

To highlight progress on MovieLens 20M, scores were taken from the following papers:
Deep Variational Autoencoder with Shallow Parallel Path for Top-N Recommendation (VASP)
Enhancing VAEs for Collaborative Filtering: Flexible Priors & Gating Mechanisms
RaCT: Toward Amortized Ranking-Critical Training for Collaborative Filtering
Variational Autoencoders for Collaborative Filtering

## Criteo

Data on Criteo area under curve score (AUC) was retrieved through a detailed arXiv literature review cross-referenced by technical progress reported on Papers with Code. The reported dates correspond to the year during which a paper was first published to arXiv, and the reported results (AUC) correspond to the result reported in the most recent version of each paper. Details on the Criteo benchmark can be found on the Criteo Kaggle Challenge page.

To highlight progress on Criteo, scores were taken from the following papers:
AutoInt: Automatic Feature Interaction Learning via Self-Attentive Neural Networks
DeepFM: A Factorization-Machine Based Neural Network for CTR Prediction
DeepLight: Deep Lightweight Feature Interactions for Accelerating CTR Predictions in Ad Serving
FAT-DeepFFM: Field Attentive Deep Field-aware Factorization Machine
MaskNet: Introducing Feature-Wise Multiplication to CTR Ranking Models by Instance-Guided Mask
Product-Based Neural Networks for User Response Prediction

## Arcade Learning Environment: Atari-57

Data on Arcade Learning Environment: Atari-57 mean human-normalized score was retrieved through a detailed arXiv literature review cross-referenced by technical progress reported on Papers with Code. The reported dates correspond to the year during which a paper was first published to arXiv, and the reported results (mean human-normalized score) correspond to the result reported in the most recent version of each paper. Details on the Arcade Learning Environment: Atari-57 benchmark can be found in the Arcade Learning Environment paper.

To highlight progress on Arcade Learning Environment: Atari-57, scores were taken from the following papers:
Dueling Network Architectures for Deep Reinforcement Learning
Distributional Reinforcement Learning with Quantile Regression
GDI: Rethinking What Makes Reinforcement Learning Different From Supervised Learning
Mastering Atari, Go, Chess and Shogi by Planning with a Learned Model
Recurrent Experience Replay in Distributed Reinforcement Learning

## Procgen

Data on Procgen mean-normalized score was retrieved through a detailed arXiv literature review. The reported dates correspond to the year during which a paper was first published to arXiv, and the reported results (mean-normalized score) correspond to the result reported in the most recent version of each paper. Details on the Procgen benchmark can be found in the Procgen paper.

To highlight progress on Procgen, scores were taken from the following papers:
Automatic Data Augmentation for Generalization in Reinforcement Learning

Leveraging Procedural Generation to Benchmark Reinforcement Learning
Procedural Generalization by Planning with Self-Supervised World Models

## Chess

Data on the performance of chess software engines was taken from the Swedish Chess Computer Association's ranking of top chess software engines. The Swedish Chess Computer Association tests computer chess software systems against one another and releases a ranking list of the top-performing systems. The ranking list produced by the Swedish Chess Computer Association is a statistically significant and meaningful measurement of chess engine performance because engines are pitted against one another in thousands of tournament-like games and each employ the same underlying hardware. Data on Magnus Carlsen's top ELO score was taken from the International Chess Federation.

## Training Time and Number of Accelerators

Data on training time and number of accelerators for AI systems was taken from the MLPerf Training benchmark competitions. More specifically, the AI Index made use of data from the following MLPerf training competitions:
MLPerf Training v0.5, 2018
MLPerf Training v0.6, 2019
MLPerf Training v0.7, 2020
MLPerf Training v1.0, 2021
MLPerf Training v1.1, 2021

Details on the MLPerf Training benchmark can be found in the MLPerf Training benchmark paper. Details on the current benchmark categories as well as technical information about submission and competition subdivisions can be found on the MLPerf Training webpage.

## ImageNet Training Cost

Data on ImageNet Training cost was based on research from DAWNBench and the individual research of Deepak Narayanan. DAWNBench is a benchmark suite for end-to-end deep-learning training and inference. DAWNBench provides a reference set of common deep-learning workloads for quantifying training time, training cost, inference latency, and inference cost across different optimization strategies, model architectures, software frameworks, clouds, and hardware. More details available at DAWNBench.

Because DAWNBench was deprecated after March 2020, data on the training cost for the most recent round of MLPerf submissions was manually collected by Deepak Narayanan.

## AI Index Robotics Survey

The survey was distributed to 509 professors who specialize in robotics from 67 universities online over three waves from December 2021 to February 2022. The selection of universities was based on the World University Rankings 2021 with geographic representation across the globe. Please view the complete survey raw data in our public data folder here. The survey was completed by 101 professors from 43 universities, including:

Aalborg University, Denmark
Ain Shams University, Egypt
Carnegie Mellon University, United States
Columbia University, United States
Cornell University, United States
Delft University of Technology, Netherlands
ETH Zurich, Switzerland
Hong Kong University of Science and Technology, Hong Kong
Korea Advanced Institute of Science and Technology, South Korea
KU Leuven, Belgium
Massachusetts Institute of Technology, United States
Nanyang Technological University, Singapore
National Polytechnic Institute, Mexico
National University of Singapore, Singapore
Peking University, China
Politecnico di Milano, Italy
Pontificia Universidad Católica de Chile, Chile
Princeton University, United States
Purdue University, United States
RWTH Aachen University, Germany
Seoul National University, South Korea
Stanford University, United States
Stellenbosch University, South Africa
Swiss Federal Institute of Technology Lausanne, Switzerland
Tokyo Institute of Technology, Japan
University of British Columbia, Canada
University of California Berkeley, United States
University of California Los Angeles, United States
University of California San Diego, United States
University of Cambridge, United Kingdom
University of Cape Town, South Africa
University College London, United Kingdom
University of Hong Kong, Hong kong
University of Illinois at Urbana-Champaign, United States
University of Malaya, Malaysia
University of Manchester, United Kingdom
University of Michigan, United States
Universitat Politècnica de Catalunya, Spain
University of Texas at Austin, United States
University of Tokyo, Japan
University of Toronto, Canada
University of Waterloo, Canada
Zhejiang University, China

# CHAPTER 3: TECHNICAL AI ETHICS

## AI ETHICS TRENDS AT FACCT AND NEURIPS

To understand trends at the ACM Conference on Fairness, Accountability, and Transparency, this section tracks FAccT papers published in conference proceedings from 2018 to 2021. We categorize author affiliations into academic, industry, nonprofit, government, and independent categories, while also tracking the location of their affiliated institution. Authors with multiple affiliations are counted once in each category (academic and industry), but multiple affiliations of the same type (i.e., authors belonging to two academic institutions) are counted once in the category.

For the analysis conducted on NeurIPS publications, we identify workshops themed around real-world impact and label papers with a single main category in "healthcare," "climate," "finance," "developing world," or "other," where "other" denotes a paper related to a real-world use case but not in one of the other categories.

We tally the number of papers in each category to reach the numbers found in Figure 3.3.3. Papers are not double-counted in multiple categories. We note that this data may not be as accurate for data pre-2018 as societal impacts work at NeurIPS has historically been categorized under a broad "AI for social impact" umbrella,[7] but it has recently been split into more granular research areas. Examples include workshops dedicated to machine learning for health,[8] climate,[9] policy & governance[10], disaster response[11], and the developing world.[12]

To track trends around specific technical topics at NeurIPS as in Figures 3.3.4–3.3.7, we count the number of papers accepted to the NeurIPS main track with titles containing keywords (e.g., "counterfactual" or "causal" for tracking papers related to causal effect), as well as papers submitted to related workshops. See the list of workshops considered for analysis here.

## META-ANALYSIS OF FAIRNESS AND BIAS METRICS

For the analysis conducted on fairness and bias metrics in AI, we identify and report on benchmark and diagnostic metrics which have been consistently cited in the academic community, reported on a public leaderboard, or reported for publicly available baseline models (e.g., GPT-3, BERT, ALBERT). We note that research paper citations are a lagging indicator of adoption, and metrics which have been very recently adopted may not be reflected in the data for 2021.

For Figures 3.1.1 and 3.1.2, we track metrics from the following papers and projects:

Aligning AI with Shared Human Values
Assessing Social and Intersectional Biases in Contextualized Word Representations
Bias in Bios: A Case Study of Semantic Representation Bias in a High-Stakes Setting
BOLD: Dataset and Metrics for Measuring Biases in Open-Ended Language Generation
Certifying and Removing Disparate Impact
CivilComments: Jigsaw Unintended Bias in Toxicity Classification
CrowS-Pairs: A Challenge Dataset for Measuring Social Biases in Masked Language Models

7 See 2018 Workshop on Ethical, Social and Governance Issues in AI 2018, 2018 AI for Social Good Workshop, 2019 Joint Workshop on AI for Social Good, 2020 Resistance AI Workshop, 2020 Navigating the Broader Impacts of AI Research Workshop.
8 See 2014 Machine Learning for Clinical Data Analysis, Healthcare and Genomics, 2015 Machine Learning for Healthcare, 2016 Machine Learning for Health, 2017 Machine Learning for Health.
9 See 2013 Machine Learning for Sustainability, 2020 AI for Earth Sciences, 2019, 2020, 2021 Tackling Climate Change with ML.
10 See 2016 People and Machines, 2019 Joint Workshop on AI for Social Good–Public Policy, 2021 Human-Centered AI.
11 See 2019 AI for Humanitarian Assistance and Disaster Response, 2020 Second Workshop on AI for Humanitarian Assistance and Disaster Response, 2021 Third Workshop on AI for Humanitarian Assistance and Disaster Response.
12 See 2017–2021 Machine Learning for the Developing World Workshops.

Detecting Emergent Intersectional Biases: Contextualized Word Embeddings Contain a Distribution of Human-Like Biases
Equality of Opportunity in Supervised Learning
Evaluating Gender Bias in Machine Translation
Evaluating Gender Bias in Natural Language Inference
Examining Gender Bias in Languages with Grammatical Gender
Fairness Through Awareness
Gender Bias in Coreference Resolution
Gender Bias in Coreference Resolution: Evaluation and Debiasing Methods
Gender Bias in Multilingual Embeddings and Cross-Lingual Transfer
Gender Shades: Intersectional Accuracy Disparities in Commercial Gender Classification
Image Representations Learned with Unsupervised Pretraining Contain Human-Like Biases
Measuring and Reducing Gendered Correlations in Pretrained Models
Measuring Bias in Contextualized Word Representations
Measuring Bias with Wasserstein Distance
Nuanced Metrics for Measuring Unintended Bias with Real Data for Text Classification
On Formalizing Fairness in Prediction with Machine Learning
On Measuring Social Biases in Sentence Encoders
Perspective API
RealToxicityPrompts: Evaluating Neural Toxic Degeneration in Language Models
Scaling Language Models: Methods, Analysis & Insights from Training Gopher
Semantics Derived Automatically from Language Corpora Contain Human-Like Biases
StereoSet: Measuring Stereotypical Bias in Pretrained Language Models
The Woman Worked as a Babysitter: On Biases in Language Generation
When Worlds Collide: Integrating Different Counterfactual Assumptions in Fairness

## NATURAL LANGUAGE PROCESSING BIAS METRICS

In Section 3.2, we track citations of the Perspective API created by Jigsaw at Google. The Perspective API has been adopted widely by researchers and engineers in natural language processing. Its creators define toxicity as "a rude, disrespectful, or unreasonable comment that is likely to make someone leave a discussion," and the tool is powered by machine learning models trained on a proprietary dataset of comments from Wikipedia and news websites. We include the following papers in our analysis:

#ContextMatters: Advantages and Limitations of Using Machine Learning to Support Women in Politics
A General Language Assistant as a Laboratory for Alignment
A Machine Learning Approach to Comment Toxicity Classification
A Novel Preprocessing Technique for Toxic Comment Classification
Adversarial Text Generation for Google's Perspective API
Avoiding Unintended Bias in Toxicity Classification with Neural Networks
Bad Characters: Imperceptible NLP Attacks
Challenges in Detoxifying Language Models
Classification of Online Toxic Comments Using Machine Learning Algorithms
Context Aware Text Classification and Recommendation Model for Toxic Comments Using Logistic Regression
Detecting Cross-Geographic Biases in Toxicity Modeling on Social Media
Detoxifying Language Models Risks Marginalizing Minority Voices
Fighting Hate Speech, Silencing Drag Queens? Artificial Intelligence in Content Moderation and Risks to LGBTQ Voices Online
HATEMOJI: A Test Suite and Adversarially Generated Dataset for Benchmarking and Detecting Emoji-Based Hate
HotFlip: White-Box Adversarial Examples for Text Classification
Identifying Latent Toxic Features on YouTube Using Non-Negative Matrix Factorization

Interpreting Social Respect: A Normative Lens for ML Models
Knowledge-Based Neural Framework for Sexism Detection and Classification
Large Pretrained Language Models Contain Human-Like Biases of What Is Right and Wrong to Do
Leveraging Multilingual Transformers for Hate Speech Detection
Limitations of Pinned AUC for Measuring Unintended Bias
Machine Learning Suites for Online Toxicity Detection
Mitigating Harm in Language Models with Conditional-Likelihood Filtration
On-the-Fly Controlled Text Generation with Experts and Anti-Experts
Process for Adapting Language Models to Society (PALMS) with Values-Targeted Datasets
Racial Bias in Hate Speech and Abusive Language Detection Datasets
RealToxicityPrompts: Evaluating Neural Toxic Degeneration in Language Models
Scaling Language Models: Methods, Analysis & Insights from Training Gopher
Self-Diagnosis and Self-Debiasing: A Proposal for Reducing Corpus-Based Bias in NLP
Social Bias Frames: Reasoning About Social and Power Implications of Language
Social Biases in NLP Models as Barriers for Persons with Disabilities
Stereotypical Bias Removal for Hate Speech Detection Task Using Knowledge-Based Generalizations
The Risk of Racial Bias in Hate Speech Detection
Towards Measuring Adversarial Twitter Interactions Against Candidates in the US Midterm Elections
Toxic Comment Classification Using Hybrid Deep Learning Model
Toxicity-Associated News Classification: The Impact of Metadata and Content Features
Understanding BERT Performance in Propaganda Analysis
White-to-Black: Efficient Distillation of Black-Box Adversarial Attacks
Women, Politics and Twitter: Using Machine Learning to Change the Discourse

While the Perspective API is used widely within machine learning research and also for measuring online toxicity, toxicity in the specific domains used to train the models undergirding Perspective (e.g., news, Wikipedia) may not be broadly representative of all forms of toxicity (e.g., trolling). Other known caveats include biases against text written by minority voices: The Perspective API has been shown to disproportionately assign high toxicity scores to text that contains mentions of minority identities (e.g., "I am a gay man"). As a result, detoxification techniques built with labels sourced from the Perspective API result in models that are less capable of modeling language used by minority groups, and they avoid mentioning minority identities.

We note that the effect size metric reported in the Word Embeddings Association Test (WEAT) section is highly sensitive to rare words, as it has been shown that removing less than 1% of relevant documents in a corpus can significantly impact the WEAT effect size. This means that effect size is not guaranteed to be a robust metric for assessing bias in embeddings. While we report on a subset of embedding association tasks measuring bias along gender and racial axes, these embedding association tests have been extended to quantify the effect across intersectional axes (e.g., EuropeanAmerican+male, AfricanAmerican+male, AfricanAmerican+female).

In the analysis of embeddings from over 100 years of U.S. Census data, embedding bias was measured by computing the difference between average embedding distances. For example, gender bias is calculated as the average distance of embeddings of words associated with women (e.g., she, female) compared to embeddings of words for occupations (e.g., teacher, lawyer), minus the same average distance calculated for words associated with men.

## FACTUALITY AND TRUTHFULNESS

### Definitions

The concepts of factuality, factual correctness, factual accuracy, and veracity are all used to refer to conformity with facts or truth. Recent work in AI aims to assess factual correctness within language models and characterize their limitations.

While human truthfulness is a relatively well-understood concept, truthfulness is not a well-characterized concept within the context of AI. A group of researchers has proposed frameworks for what it means for a system to be truthful—for example, a broadly truthful system should avoid lying or using true statements to mislead or misdirect; should be clear, informative, and (mostly) cooperative in conversation; and should be well-calibrated, self-aware, and open about the limits of its knowledge. A definition of narrow truthfulness may simply refer to systems which avoid stating falsehoods. The authors of TruthfulQA define a system as truthful only if it avoids asserting a false statement; refusing to answer a question, expressing uncertainty, or giving a true but irrelevant answer may be considered truthful but not informative.

Truthfulness is related to *alignment*: A truthful system is aligned with human values and goals. In one definition of alignment, an aligned system is one that is helpful, honest, and harmless. Since we cannot yet measure honesty within a system, truthfulness can be used as a proxy.

An *honest* system is one that asserts only what it "believes" or one that never contradicts its own beliefs. A system can be honest but not truthful—for example, if an honest system believes that vaccines are unsafe, it can claim this honestly, despite the statement being factually incorrect. Conversely, a system can be truthful but not honest: It may believe vaccines are unsafe but asserts they are safe to pass a test. Another work proposes that an honest system should give accurate information, not mislead users, be calibrated (e.g., it should be correct 80% of the time when it claims 80% confidence), and express appropriate levels of uncertainty.

*Hallucination* refers to language models fabricating statements not found in factually correct supporting evidence or input documents. In closed-form dialog, summarization, or question-answering, a system that hallucinates is considered untruthful.

### Language Diversity in Training Data

Imbalanced language distribution in training data impacts the performance of general-purpose language models. For example, the Gopher family of models is trained on MassiveText (10.5TB), which is a dataset made up of 99% English. Similarly, only 7.4% of GPT-3 training data is in non-English languages. In contrast, XGLM, a recent model family from Meta AI, is trained on a training data of 30 languages, and upsamples low-resource languages to create a more balanced language representation. See Figure 1 on the XGLM paper that compares the language distribution of XGLM and GPT-3.

In addition, Figure 7 of the XGLM paper highlights the extent to which language models can effectively store factual knowledge by comparing the performance of XGLM (a multilingual language model) with GPT-3, a monolingual model. Performance was evaluated on knowledge triplet completion using the mLAMA dataset, which was translated from the English benchmark LAMA using Google Translate. GPT-3 outperforms in English, but XGLM outperforms in non-English languages. Further results show that more diverse language representation improves language model performance in tasks such as translation.

In 2021, Congress inquired into the content moderation practices of social media companies in non-English languages, and emphasized the importance of equal access to truthful and trustworthy information regardless of language. As these companies start to adopt language models into their fact-checking and content moderation processes for languages around the world, it is critical to be able to measure the disproportionate negative impact of using models which underperform on non-English languages.

# CHAPTER 4: THE ECONOMY AND EDUCATION

## EMSI BURNING GLASS

Prepared by Julia Nitschke, Summer Jasinski, Bledi Taska and Rucha Vankudre

Emsi Burning Glass delivers job market analytics that empower employers, workers, and educators to make data-driven decisions. The company's artificial intelligence technology analyzes hundreds of millions of job postings and real-life career transitions to provide insight into labor market patterns. This real-time strategic intelligence offers crucial insights, such as what jobs are most in demand, the specific skills employers need, and the career directions that offer the highest potential for workers. For more information, visit burning-glass.com.

### Job Posting Data

To support these analyses, Emsi Burning Glass mined its dataset of millions of job postings collected since 2010. Emsi Burning Glass collects postings from over 45,000 online job sites to develop a comprehensive, real-time portrait of labor market demand. It aggregates job postings, removes duplicates, and extracts data from job postings text. This includes information on job title, employer, industry, and region, as well as required experience, education, and skills.

Job postings are useful for understanding trends in the labor market because they allow for a detailed, real-time look at the skills employers seek. To assess the representativeness of job postings data, Emsi Burning Glass conducts a number of analyses to compare the distribution of job postings to the distribution of official government and other third-party sources in the United States. The primary source of government data on U.S. job postings is the Job Openings and Labor Turnover Survey (JOLTS) program, conducted by the Bureau of Labor Statistics. Based on comparisons between JOLTS and Emsi Burning Glass, the labor market demand captured by Emsi Burning Glass data represents over 95% of the total labor demand. Jobs not posted online are usually in small businesses (the classic example being the "Help Wanted" sign in the restaurant window) and union hiring halls.

### Measuring Demand for AI

In order to measure employers' demand for AI skills, Emsi Burning Glass uses its skills taxonomy of over 17,000 skills. The list of AI skills from Emsi Burning Glass data is shown below, with associated skill clusters. While some skills are considered to be in the AI cluster specifically, for the purposes of this report, all skills below were considered AI skills. A job posting was considered an AI job if it requested one or more of these skills.

**Artificial Intelligence:** Expert System, IBM Watson, IPSoft Amelia, Ithink, Virtual Agents, Autonomous Systems, Lidar, OpenCV, Path Planning, Remote Sensing

**Natural Language Processing (NLP):** ANTLR, Automatic Speech Recognition (ASR), Chatbot, Computational Linguistics, Distinguo, Latent Dirichlet Allocation, Latent Semantic Analysis, Lexalytics, Lexical Acquisition, Lexical Semantics, Machine Translation (MT), Modular Audio Recognition Framework (MARF), MoSes, Natural Language Processing, Natural Language Toolkit (NLTK), Nearest Neighbor Algorithm, OpenNLP, Sentiment Analysis/ Opinion Mining, Speech Recognition, Text Mining, Text to Speech (TTS), Tokenization, Word2Vec

**Neural Networks:** Caffe Deep Learning Framework, Convolutional Neural Network (CNN), Deep Learning, Deeplearning4j, Keras, Long Short-Term Memory (LSTM), MXNet, Neural Networks, Pybrain, Recurrent Neural Network (RNN), TensorFlow

**Machine Learning:** AdaBoost algorithm, Boosting (Machine Learning), Chi Square Automatic Interaction Detection (CHAID), Classification Algorithms, Clustering Algorithms, Decision Trees, Dimensionality Reduction, Google Cloud Machine Learning Platform, Gradient boosting, H2O (software), Libsvm, Machine Learning, Madlib, Mahout, Microsoft Cognitive Toolkit, MLPACK (C++ library), Mlpy, Random Forests, Recommender Systems, Scikit-learn, Semi-Supervised Learning, Supervised Learning (Machine Learning), Support Vector Machines (SVM), Semantic Driven Subtractive Clustering Method (SDSCM), Torch (Machine Learning), Unsupervised Learning, Vowpal, Xgboost

**Robotics:** Blue Prism, Electromechanical Systems, Motion Planning, Motoman Robot Programming, Robot Framework, Robotic Systems, Robot Operating System (ROS), Robot Programming, Servo Drives / Motors, Simultaneous Localization and Mapping (SLAM)

**Visual Image Recognition:** Computer Vision, Image Processing, Image Recognition, Machine Vision, Object Recognition

## LINKEDIN

Prepared by Akash Kaura and Murat Erer

### Country Sample

Included countries represent a select sample of eligible countries with at least 40% labor force coverage by LinkedIn and at least 10 AI hires in any given month. China and India were included in this sample because of their increasing importance in the global economy, but LinkedIn coverage in these countries does not reach 40% of the workforce. Insights for these countries may not provide as full a picture as other countries, and should be interpreted accordingly.

### Skills (and AI Skills)

LinkedIn members self-report their skills on their LinkedIn profiles. Currently, more than 38,000 distinct, standardized skills are identified by LinkedIn. These have been coded and classified by taxonomists at LinkedIn into 249 skill groupings, which are the skill groups represented in the dataset. The top skills that make up the AI skill grouping are machine learning, natural language processing, data structures, artificial intelligence, computer vision, image processing, deep learning, TensorFlow, Pandas (software), and OpenCV, among others.

Skill groupings are derived by expert taxonomists through a similarity-index methodology that measures skill composition at the industry level. Industries are classified according to the ISIC 4 industry classification (Zhu et al., 2018).

### Skills Genome

For any entity (occupation or job, country, sector, etc.), the skill genome is an ordered list (a vector) of the 50 “most characteristic skills” of that entity. These most characteristic skills are identified using a TF-IDF algorithm to identify the most representative skills of the target entity, while down-ranking ubiquitous skills that add little information about that specific entity (e.g., Microsoft Word).

TF-IDF is a statistical measure that evaluates how representative a word (in this case a skill) is to a selected entity. This is done by multiplying two metrics:

1. The term frequency of a skill in an entity (TF).
2. The logarithmic inverse entity frequency of the skill across a set of entities (IDF). This indicates how common or rare a word is in the entire entity set. The closer IDF is to 0, the more common a word is.

So, if the skill is very common across LinkedIn entities, and appears in many job or member descriptions, the IDF will approach 0. If, on the other hand, the skill is unique to specific entities, the IDF will approach 1. Details available at LinkedIn’s Skills Genome and LinkedIn-World Bank Methodology note.

### AI Skills Penetration

The aim of this indicator is to measure the intensity of AI skills in an entity (in a particular country, industry, gender, etc.) through the following methodology:

- Compute frequencies for all self-added skills by LinkedIn members in a given entity (occupation, industry, etc.) in 2015–2021.
- Re-weight skill frequencies using a TF-IDF model to get the top 50 most representative skills in that entity. These 50 skills compose the "skill genome" of that entity.
- Compute the share of skills that belong to the AI skill group out of the top skills in the selected entity.

Interpretation: The AI skill penetration rate signals the prevalence of AI skills across occupations, or the intensity with which LinkedIn members utilize AI skills in their jobs. For example, the top 50 skills for the occupation of engineer are calculated based on the weighted frequency with which they appear in LinkedIn members' profiles. If four of the skills that engineers possess belong to the AI skill group, this measure indicates that the penetration of AI skills is estimated to be 8% among engineers (e.g., 4/50).

## Jobs or Occupations

LinkedIn member titles are standardized and grouped into approximately 15,000 occupations. These are not sector or country specific. These occupations are further standardized into approximately 3,600 occupation representatives. Occupation representatives group occupations with a common role and specialty, regardless of seniority.

## AI Jobs or Occupations

An AI job (technically, occupation representative) is an occupation representative that requires AI skills to perform the job. Skills penetration is used as a signal for whether AI skills are prevalent in an occupation representative, in any sector where the occupation representative may exist. Examples of such occupations include (but are not limited to): machine learning engineer, artificial intelligence specialist, data scientist, computer vision engineer, etc.

## AI Talent

A LinkedIn member is considered AI talent if they have explicitly added AI skills to their profile and/or they are occupied in an AI occupation representative. The counts of AI talent are used to calculate talent concentration metrics (e.g. to calculate the country-level AI talent concentration, we use the counts of AI talent at the country level vis-a-vis the counts of LinkedIn members in the respective countries).

## Relative AI Skills Penetration

To allow for skills penetration comparisons across countries, the skills genomes are calculated and a relevant benchmark is selected (e.g., global average). A ratio is then constructed between a country's and the benchmark's AI skills penetrations, controlling for occupations.

Interpretation: A country's relative AI skills penetration of 1.5 indicates that AI skills are 1.5 times as frequent as in the benchmark, for an overlapping set of occupations.

## Global Comparison

For cross-country comparison, we present the relative penetration rate of AI skills, measured as the sum of the penetration of each AI skill across occupations in a given country, divided by the average global penetration of AI skills across the overlapping occupations in a sample of countries.

Interpretation: A relative penetration rate of 2 means that the average penetration of AI skills in that country is two times the global average across the same set of occupations.

## Global Comparison: By Industry

The relative AI skills penetration by country for industry provides an in-depth sectoral decomposition of AI skill penetration across industries and sample countries.

Interpretation: A country's relative AI skill penetration rate of 2 in the education sector means that the average penetration of AI skills in that country is two times the global average across the same set of occupations in that sector.

### Relative AI Hiring Index

- LinkedIn Hiring Rate or Overall Hiring Rate is a measure of hires normalized by LinkedIn membership. It is computed as the percentage of LinkedIn members who added a new employer in the same period the job began, divided by the total number of LinkedIn members in the corresponding location.
- AI Hiring Rate is computed following the overall hiring rate methodology, but only considering members classified as AI talent.
- Relative AI Hiring Index is the pace of change in AI Hiring Rate normalized by the pace of change in Overall Hiring Rate, providing a picture of whether hiring of AI talent is growing at a higher, equal, or lower rate than overall hiring in a market. The relative AI Hiring Index is equal to 1.0 when AI hiring and overall hiring are growing at the same rate year on year.

Interpretation: Relative AI Hiring Index shows how fast each country is experiencing growth in AI talent hiring relative to growth in overall hiring in the country. A ratio of 1.2 means the growth in AI talent hiring has outpaced the growth in overall hiring by 20%.

## NETBASE QUID

Prepared by Julie Kim and Tejas Sirohi

NetBase Quid delivers AI-powered consumer and market intelligence to enable business reinvention in a noisy and unpredictable world. The software applies artificial intelligence to reveal patterns in large, unstructured datasets and to generate visualizations that enable users to make smart, data-driven decisions accurately, quickly, and efficiently. NetBase Quid uses Boolean query to search for focus areas, topics, and keywords within social, news, forums and blogs, companies, and patents data sources, as well as other custom datasets. NetBase Quid then visualizes these data points based on the semantic similarity.

### Search, Data Sources, and Scope

Over 6 million global public and private company profiles from multiple data sources are indexed in order to search across company descriptions, while filtering and including metadata ranging from investment information to firmographic information, such as founded year, HQ location, and more. Company information is updated on a weekly basis. NetBase Quid algorithm reads a big amount of text data from each document to make links between different documents based on their similar language. This process is repeated at an immense scale, which produces a network with different clusters identifying distinct topics or focus areas. Trends are identified based on keywords, phrases, people, companies, institutions that NetBase Quid identifies, and the other metadata that is put into the software.

### Data

Organization data is embedded from Capital IQ and Crunchbase. These companies include all types of companies (private, public, operating, operating as a subsidiary, out of business) throughout the world. The investment data includes private investments, M&A, public offerings, minority stakes made by PE/VCs, corporate venture arms, governments, and institutions both within and outside the United States. Some data is simply unreachable—for instance, when the investors or the funding amounts by investors are undisclosed. NetBase Quid also embeds firmographic information such as founding year and HQ location.

NetBase Quid embeds Capital IQ data as a default and adds in data from Crunchbase for the ones that are not captured in Capital IQ. This not only yields comprehensive and accurate data on all global organizations, but it also captures early-stage startups and funding events data. Company information is uploaded on a weekly basis.

### Search Parameters

Boolean query is used to search for focus areas, topics, and keywords within the archived company database, within their business descriptions and websites. We can filter out the search results by HQ regions, investment amount, operating status, organization type (private/public), and founding year. NetBase Quid then visualizes these companies by the semantic similarity. If there are more than 7,000 companies from the search result, NetBase Quid selects the 7,000 most relevant companies for visualization based on the language algorithm.

Boolean Search: "artificial intelligence" or "AI" or "machine learning" OR "deep learning"

Companies:

- Chart 4.2.1: Global AI & ML companies that have been invested (private, IPO, M&A) from 01/01/2012 to 12/31/2021.
- Chart 4.2.2–4.2.12: Global AI & ML companies that have invested over $1.5M for the last 10 years (January 1, 2012 to December 31, 2021)—7,000 companies out of 7,500 companies have been selected through Quid's relevance algorithm.

### Target Event Definitions

- Private investments: A private placement is a private sale of newly issued securities (equity or debt) by a company to a selected investor or a selected group of investors. The stakes that buyers take in private placements are often minority stakes (under 50%), although it is possible to take control of a company through a private placement as well, in which case the private placement would be a majority stake investment.
- Minority investment: These refer to minority stake acquisitions in Quid, which take place when the buyer acquires less than 50% of the existing ownership stake in entities, asset product, and business divisions.
- M&A: This refers to a buyer acquiring more than 50% of the existing ownership stake in entities, asset product, and business divisions.

## COMPUTING RESEARCH ASSOCIATION (CRA TAULBEE SURVEY)

Prepared by Betsy Bizot (CRA senior research associate)

### Source

Computing Research Association (CRA) members are 200-plus North American organizations active in computing research: academic departments of computer science and computer engineering; laboratories and centers in industry, government, and academia; and affiliated professional societies (AAAI, ACM, CACS/AIC, IEEE Computer Society, SIAM USENIX). CRA's mission is to enhance innovation by joining with industry, government, and academia to strengthen research and advanced education in computing. Learn more about CRA here.

### Methodology

CRA Taulbee Survey gathers survey data during the fall of each academic year by reaching out to over 200 PhD-granting departments. Details about the Taulbee Survey can be found here. Taulbee doesn't directly survey the students. The department identifies each new PhD's area of specialization as well as their type of employment. Data is collected from September to January of each academic year for PhDs awarded in the previous academic year. Results are published in May after data collection closes. So the 2020 data points were newly available last spring, and the numbers provided for 2021 will be available in May 2020.

The CRA Taulbee Survey is sent only to doctoral departments of computer science, computer engineering, and information science/systems. Historically, (a) Taulbee covers 1/4 to 1/3 of total BS CS recipients in the United States; (b) the percent of women earning bachelor's degrees is lower in the Taulbee schools than overall; and (c) Taulbee tracks the trends in overall CS production.

## Nuances

- Of particular interest in PhD job market trends are the metrics on the AI PhD area of specialization. The categorization of specialty areas changed in 2008 and was clarified in 2016. From 2004-2007, AI and robotics were grouped; from 2008-present, AI is separate; 2016 clarified to respondents that AI includes ML.
- Notes about the trends in new tenure-track hires (overall and particularly at AAU schools): In the 2018 Taulbee Survey, for the first time we asked how many new hires had come from the following sources: new PhD, postdoc, industry, and other academic. Results indicated that 29% of new assistant professors came from another academic institution.
- Some may have been teaching or research faculty rather than tenure-track, but there is probably some movement between institutions, meaning the total number hired overstates the total who are actually new.

# CHAPTER 5: AI POLICY AND GOVERNANCE

## BLOOMBERG GOVERNMENT

Prepared by Amanda Allen

Bloomberg Government is a premium, subscription-based service that provides comprehensive information and analytics for professionals who interact with—or are affected by—the government. Delivering news, analytics, and data-driven decision tools, Bloomberg Government's digital workspace gives an intelligent edge to government affairs and contracting professionals. For more information or a demo, visit about.bgov.com.

### Methodology

**Contract Spending:** Bloomberg Government's Contracts Intelligence Tool structures all contracts data from www.fpds.gov. The CIT includes a model of government spending on artificial intelligence-related contracts that is based on a combination of government-defined product service codes and more than 100 AI-related keywords. For the section "U.S. Government Contract Spending," Bloomberg Government analysts used contract spending data from fiscal year 2000 through fiscal year 2021.

**Defense RDT&E Budget:** Bloomberg Government organized all the RDT&E budget request line items available from the Defense Department Comptroller. For the section "U.S. Department of Defense (DOD) Budget," Bloomberg Government used a set of AI-specific keywords to identify 500 unique budget activities related to artificial intelligence and machine learning worth a combined $5.9 billion in FY 2021.

**Legislative Documents:** Bloomberg Government maintains a repository of congressional documents, including bills, Congressional Budget Office assessments, and reports published by congressional committees, the Congressional Research Service, and other offices. Bloomberg Government also ingests state legislative bills. For the section "AI Policy and Governance," Bloomberg Government analysts identified all legislation, congressional committee reports, and CRS reports that referenced one or more AI-specific keywords.

## GLOBAL LEGISLATION RECORDS ON AI

For AI-related bills passed into laws, the AI Index performed searches of the keyword “artificial intelligence,” in respective languages, on the websites of 25 countries’ congresses or parliaments, in full-text of bills. Note that only laws passed by state-level legislative bodies and signed into law (i.e., by presidents or received royal assent) from 2015 to 2021 are included. Future AI Index reports hope to include analysis on other types of legal documents, such as regulations and standards, adopted by state- or supranational-level legislative bodies, government agencies, etc.

### Australia

Website: www.legislation.gov.au
Keyword: artificial Intelligence
Filters:

- Legislation types: Acts
- Portfolios: Department of House of Representatives, Department of Senate

Note: Texts in explanatory memorandum are not counted.

### Belgium

Website: http://www.ejustice.just.fgov.be/loi/loi.htm
Keyword: intelligence artificielle

### Brazil

Website: https://www.camara.leg.br/legislacao
Keyword: inteligência artificial
Filter:

- Federal legislation
- Type: Law

### Canada

Website: https://www.parl.ca/legisinfo/
Keyword: artificial Intelligence
Note: Results were investigated to determine how many of the bills introduced were eventually passed (i.e., received royal assent) and bill status was recorded.

### China

Website: https://flk.npc.gov.cn/
Keyword: 人工智能
Filters:

- Legislative body: Standing Committee of the National People’s Congress

### Denmark

Website: https://www.retsinformation.dk/
Keyword: kunstig intelligen
Filter:

- Document Type: Laws

### Finland

Website: https://www.finlex.fi/
Keyword: tekoäly
Noting under the Current Legislation section

### France

Website: https://www.legifrance.gouv.fr/
Keyword: intelligence artificielle
Filter:

- texte consolidé
- Document Type: Law

### Germany

Website: http://www.gesetze-im-internet.de/index.html
Keyword: künstliche Intelligenz
Filter:

- All federal codes, statutes, and ordinances that are currently in force
- Volltextsuche (full text)
- Und-Verknüpfung der Wörter (entire word)

### India

Website: https://www.indiacode.nic.in
Keyword: artificial intelligence
Note: The website used allows for a search of keywords in legalization title but not in the full text, as such it is not useful for this particular research. Therefore, a Google search using the “site” function to search the site with the keyword of “artificial intelligence” is conducted.

### Ireland
Website: www.irishstatutebook.ie
Keyword: artificial intelligence

### Italy
Website: https://www.normattiva.it/
Keyword: intelligenza artificiale
Filter:
- Document Type: law

### Japan
Website: https://elaws.e-gov.go.jp/
Keyword: 人工知能
Filter:
- Full text
- Law

### Netherlands
Website: https://www.overheid.nl/
Keyword: kunstmatige intelligentie
Filter:
- Document Type: Wetten

### New Zealand
Website: www.legislation.govt.nz
Keyword: Artificial intelligence
Filter:
- Document type: acts
- Status option: For the status option (example: acts in force, current bills, etc.)

### Norway
Website: https://lovdata.no/
Keyword: kunstig intelligens

### Russia
Website: http://graph.garant.ru:8080/SESSION/PILOT/main.htm (Database "The Federal Laws" in the official website of the Federation Council of the Federal Assembly of the Russian Federation.)
Keyword: искусственный интеллект
Filter:
- Words in text

### Singapore
Website: https://sso.agc.gov.sg/
Keyword: artificial intelligence
Filter:
- Document Type: Current acts and subsidiary legislation

### South Africa
Website: www.gov.za
Keyword: artificial intelligence
Filter:
- Document: acts

Note: This search function seemingly does not search within the context of the full text and so no results were returned. Therefore, a Google search using the "site" function to search the site with the keyword of "artificial intelligence" is conducted.

### South Korea
Website: https://law.go.kr/eng/; https://elaw.klri.re.kr/
Keyword: artificial Intelligence or 인공 지능
Filter:
- Type: Act

Note: Cannot search combined words, so individual analysis is conducted.

### Spain
Website: https://www.boe.es/
Keyword: inteligencia artificial
Filter:
- Type: law
- Head of state (for passed laws)

### Sweden
Website: https://www.riksdagen.se/
Keyword: artificiell intelligens
Filter: Swedish Code of Statutes

## Switzerland

Website: https://www.fedlex.admin.ch/
Keyword: intelligence artificielle
Filter:

- Text category: federal constitution, federal acts, and federal decrees, miscellaneous texts, orders, and other forms of legislation.
- Publication period for legislation was limited to 2015-2021.

## United Kingdom

Website: https://www.legislation.gov.uk/
Keyword: artificial intelligence
Filter:

- Legislation Type: U.K. Public General Acts & U.K. Statutory Instruments

## United States

Website: https://www.congress.gov/
Keyword: artificial intelligence
Filter:

- Source: Legislation

Status of legislation: Became law

## MENTIONS OF AI IN AI-RELATED LEGISLATION PROCEEDINGS

For mentions of AI in AI-related legislative proceedings around the world, the AI Index performed searches of the keyword "artificial intelligence," in respective languages, on the websites of 25 countries' congresses or parliaments, usually under sections named "minutes," "hansard," etc.

### Australia
Website: https://www.aph.gov.au/Parliamentary_Business/Hansard
Keyword: artificial intelligence

### Belgium
Website: http://www.parlement.brussels/search_form_fr/
Keyword: intelligence artificielle
Filter
- Document Type: all

### Brazil
Website: https://www2.camara.leg.br/atividade-legislativa/discursos-e-notas-taquigraficas
Keyword: inteligência artificial
Filter:
- Federal legislation
- Type: Law

### Canada
Website: https://www.ourcommons.ca/PublicationSearch/en/?PubType=37
Keyword: artificial Intelligence

### China
Website: Various reports on the work of the government
Keyword: 人工智能
Note: The National People's Congress is held once per year and does not provide full legislative proceedings. Hence, the counts included in the analysis only searched the mentions of artificial intelligence in the only public document released from the Congress meetings, the Report on the Work of the Government, delivered by the Premier.

### Denmark
Website: https://www.retsinformation.dk/
Keyword: kunstig intelligens
Filter:
- Minutes

### Finland
Website: https://www.eduskunta.fi/
Keyword: tiedot
Filter:
- Parliamentary Affairs and Documents
- Public document: Minutes
- Actor: Plenary sessions

### France
Website: https://www.assemblee-nationale.fr/
Keyword: intelligence artificielle
Filter:
- Reports of the debates in session

Note: Such documents were only prepared starting in 2017.

### Germany
Website: https://dip.bundestag.de/
Keyword: künstliche Intelligenz
Filter:
- Speeches, requests to speak in the plenum

### India
Website: http://loksabhaph.nic.in/
Keyword: artificial intelligence
Filter:
- Exact word/phrase

### Ireland
Website: https://www.oireachtas.ie/
Keyword: artificial intelligence
Filter: Content of parliamentary debates

## Italy
Website: https://aic.camera.it/aic/search.html
Keyword: intelligenza artificiale
Filter:
- Type: All
- Search by exact phrase

## Japan
Website: https://kokkai.ndl.go.jp/#/
Keyword: 人工知能
Filter:
- Full text
- Law

## Netherlands
Website: https://www.tweedekamer.nl/kamerstukken?pk_campaign=breadcrumb
Keyword: kunstmatige intelligentie
Filter:
- Parliamentary papers - Plenary reports

## New Zealand
Website: https://www.parliament.nz/en/pb/hansard-debates/
Keyword: artificial intelligence

## Norway
Website: https://www.stortinget.no/no/Saker-og-publikasjoner/Publikasjoner/Referater/
Keyword: kunstig intelligens
Note: This search function does not directly allow the keyword within minutes. Therefore, a Google search using the "site" function to search the site with the keyword of "artificial intelligence" is conducted.

## Russia
Website: http://transcript.duma.gov.ru/
Keyword: искусственный интеллект
Filter:
- Words in text

## Singapore
Website: https://sprs.parl.gov.sg/search/home
Keyword: artificial intelligence

## South Africa
Website: https://www.parliament.gov.za/hansard
Keyword: artificial intelligence
Note: This search function does not search within the context of the full text and so no results were returned. Therefore, a Google search using the "site" function to search https://www.parliament.gov.za/storage/app/media/Docs/hansard/ with the keyword "artificial intelligence" is conducted.

## South Korea
Website: http://likms.assembly.go.kr/
Keyword: 인공 지능
Filter:
- Meeting Type: All

## Spain
Website: https://www.congreso.es/
Keyword: inteligencia artificial
Filter:
- Official publications of parliamentary proceedings

## Switzerland
Website: https://www.parlament.ch/
Keyword: intelligence artificielle
Filter:
- Parliamentary proceedings

## Sweden
Website: https://www.riksdagen.se/sv/global/sok/?q=&doktyp=prot
Keyword: artificiell intelligens
Filter:
- Minutes

## United Kingdom

https://hansard.parliament.uk/
Keyword: artificial intelligence
Filter
- References

## United States

Website: https://www.congress.gov/
Keyword: artificial intelligence
Filter:
- Source: Congressional record
- Congressional record section: Senate, House of Representatives, and Extensions of Remarks

# U.S. AI POLICY PAPERS

## Organizations

To develop a more nuanced understanding of the thought leadership that motivates AI policy, we tracked policy papers published by 55 organizations in the United States or with a strong presence in the United States (expanded from the list of 36 organizations last year) across four broad categories:

- Civil Society, Associations & Consortiums: Algorithmic Justice League, Alliance for Artificial Intelligence in Healthcare, Amnesty International, EFF, Future of Privacy Forum, Human Rights Watch, IJIS Institute, Institute for Electrical and Electronics Engineers, Partnership on AI
- Consultancy: Accenture, Bain & Company, Boston Consulting Group, Deloitte, McKinsey & Company
- Government Agencies: Congressional Research Service, Library of Congress, Defense Technical Information Center, Government Accountability Office, Pentagon Library
- Private Sector Companies: Google AI, Microsoft AI, Nvidia, OpenAI
- Think Tanks & Policy Institutes: American Enterprise Institute, Aspen Institute, Atlantic Council, Brookings Institute, Carnegie Endowment for International Peace, Cato Institute, Center for a New American Security, Center for Strategic and International Studies, Council on Foreign Relations, Heritage Foundation, Hudson Institute, MacroPolo, National Security Institute, New America Foundation, RAND Corporation, Rockefeller Foundation, Stimson Center, Urban Institute, Wilson Center
- University Institutes & Research Programs: AI and Humanity Cornell University; AI Now Institute, New York University; AI Pulse, UCLA Law; Belfer Center for Science and International Affairs, Harvard University; Berkman Klein Center, Harvard University; Center for Information Technology Policy, Princeton University; Center for Long-Term Cybersecurity, UC Berkeley; Center for Security and Emerging Technology, Georgetown University; CITRUS Policy Lab, UC Berkeley; Hoover Institution; Institute for Human-Centered Artificial Intelligence, Stanford University; Internet Policy Research Initiative, Massachusetts Institute of Technology; MIT Lincoln Laboratory; Princeton School of Public and International Affairs

## Methodology

Each broad topic area is based on a collection of underlying keywords that describe the content of the specific paper. We included 17 topics that represented the majority of discourse related to AI between 2018-2021. These topic areas and the associated keywords are listed below:

- Health & Biological Sciences: medicine, healthcare systems, drug discovery, care, biomedical research, insurance, health behaviors, COVID-19, global health
- Physical Sciences: chemistry, physics, astronomy, earth science
- Energy & Environment: energy costs, climate change, energy markets, pollution, conservation, oil and gas, alternative energy
- International Affairs & International Security: international relations, international trade, developing countries, humanitarian assistance, warfare, regional security, national security, autonomous weapons
- Justice & Law Enforcement: civil justice, criminal justice, social justice, police, public safety, courts

- Communications & Media: social media, disinformation, media markets, deepfakes
- Government & Public Administration: federal government, state government, local government, public sector efficiency, public sector effectiveness, government services, government benefits, government programs, public works, public transportation
- Democracy: elections, rights, freedoms, liberties, personal freedoms
- Industry & Regulation: economy, antitrust, M&A, competition, finance, management, supply chain, telecom, economic regulation, technical standards, autonomous vehicle industry and regulation
- Innovation & Technology: advancements and improvements in AI technology, R&D, intellectual property, patents, entrepreneurship, innovation ecosystems, startups, computer science, engineering
- Education & Skills: early childhood, K-12, higher education, STEM, schools, classrooms, reskilling
- Workforce & Labor: labor supply and demand, talent, immigration, migration, personnel economics, future of work
- Social & Behavioral Sciences: sociology, linguistics, anthropology, ethnic studies, demography, geography, psychology, cognitive science
- Humanities: arts, music, literature, language, performance, theater, classics, history, philosophy, religion, cultural studies
- Equity & Inclusion: biases, discrimination, gender, race, socioeconomic inequality, disabilities, vulnerable populations
- Privacy, Safety & Security: anonymity, GDPR, consumer protection, physical safety, human control, cybersecurity, encryption, hacking
- Ethics: transparency, accountability, human values, human rights, sustainability, explainability, interpretability, decision-making norms

Artificial
Intelligence
Index Report 2022

**Stanford University**
Human-Centered
Artificial Intelligence